\documentclass[sigconf,nonacm,screen]{acmart}

\usepackage{xcolor}

\usepackage{graphicx}
\usepackage{paralist}
\usepackage{amsthm}

\usepackage{pifont}
  
\usepackage{color, colortbl}

\usepackage{placeins}

\newsavebox\CBox

\newcommand{\cbit}{\begin{compactitem}}
	\newcommand{\ceit}{\end{compactitem}}
\newcommand{\cben}{\begin{compactenum}}
	\newcommand{\ceen}{\end{compactenum}}

\definecolor{OliveGreen}{rgb}{0,0.6,0}

\newcommand{\bit}{\begin{itemize}}
	\newcommand{\eit}{\end{itemize}}
\newcommand{\ben}{\begin{enumerate}}
	\newcommand{\een}{\end{enumerate}}
\newcommand{\beq}{\begin{equation}}
\newcommand{\eeq}{\end{equation}}

\newcommand{\Xg}{\mathcal{X}_g}
\newcommand{\Xc}{\mathcal{X}_c}
\newcommand{\Xo}{\mathcal{X}_o}

\newcommand{\dista}{\Delta_a}
\newcommand{\distb}{\Delta_b}
\newcommand{\intra}{c}
\newcommand{\intraa}{d}
\newcommand{\intrab}{D}
\newcommand{\inliers}{\mathcal{I}}
\newcommand{\outliers}{\mathcal{O}}

\newcommand{\hide}[1]{}

\newtheorem{definition}{Definition}

\usepackage{enumitem}
\setlist[enumerate]{leftmargin=10mm, label=\alph*)}
\setlist[itemize]{leftmargin=10mm}

\usepackage{hyperref}

\begin{document}
\hypersetup{
    colorlinks=true,
    urlcolor=blue,
    linkcolor=red,
    citecolor=green
}

\title{Outlier Detection Bias Busted: Understanding Sources of Algorithmic Bias through Data-centric Factors}

\author{Xueying Ding}
\affiliation{%
 \institution{Carnegie Mellon University}
   \city{Pittsburgh}
  \state{PA}
  \country{USA}}
\email{xding2@cs.cmu.edu}

\author{Rui Xi}
\affiliation{%
 \institution{Carnegie Mellon University}
   \city{Pittsburgh}
  \state{PA}
  \country{USA}}
\email{rxi2@andrew.cmu.edu}

\author{Leman Akoglu}
\affiliation{%
 \institution{Carnegie Mellon University}
   \city{Pittsburgh}
  \state{PA}
  \country{USA}}
\email{lakoglu@cs.cmu.edu}

\begin{abstract}
  The astonishing successes of ML  have raised growing concern for the fairness of modern methods when deployed in real world settings.
However, studies on fairness have mostly focused on supervised ML, while unsupervised outlier detection (OD), with numerous applications in finance, security, etc., have attracted little attention. While a few studies proposed fairness-enhanced OD algorithms, they remain agnostic to the underlying driving mechanisms or \textit{sources of unfairness}. Even within the supervised ML literature,there exists debate on whether unfairness stems solely from algorithmic biases (i.e. design choices)  or from the  biases encoded in the data on which they are trained.  

To close this gap, this work aims to shed light on the possible sources of unfairness in OD by auditing detection models
under different data-centric factors.By injecting various known biases into the input data---as pertain to sample size disparity, under-representation, feature measurement noise, and group membership obfuscation---we find that the OD algorithms under the study all exhibit fairness pitfalls, although differing in which types of data bias they are more susceptible to. 
Most notable of our study is to demonstrate that OD algorithm bias is \underline{not} merely a data bias problem. A key realization is that the data properties that emerge from bias injection could as well be organic---as pertain to natural group differences w.r.t. sparsity, base rate, variance, and multi-modality. Either natural or biased, such data properties can 
give rise to unfairness as they interact with
certain algorithmic design choices.

Our work provides a deeper  understanding of the possible sources of OD unfairness, and
serves as a framework for assessing the unfairness of future OD algorithms under specific data-centric factors.
It also paves the way for future work on mitigation strategies by underscoring the susceptibility of various design choices.

\end{abstract}


\maketitle

\section{Introduction}
\label{sec:intro}

With ML claiming its unprecedented place in the society,  
there exists 
growing concern for its responsible use and potential harm to already under-served societal groups that may exacerbate the pre-existing inequities.
While much work has been dedicated to measuring and mitigating unfairness of ML algorithms, they remain mostly agnostic to the underlying sources of unfairness---treating the symptom rather than the cause. 
In fact, 
the literature is quite short on understanding the 
underlying \textit{sources of algorithmic unfairness}, i.e. what drives unfairness to emerge in the first place.

Data is widely acknowledged as the key influencer of the fairness of the algorithms that are trained on it \cite{barocas2016big,mehrabi2021survey}. However, what kind of data-centric factors give rise to certain algorithmic behavior is not well understood.
Moreover, the community has mainly focused on supervised ML while the fairness of unsupervised algorithms such as for outlier detection (OD) has attracted significantly less attention, despite the numerous applications 
and punitive decision-making scenarios that OD algorithms are employed in, such as fraud detection, policing and surveillance, to name a few.
\newcommand{\xmark}{\ding{55}}
\begin{figure*}[!t]
    \includegraphics[width=.9\textwidth]{ 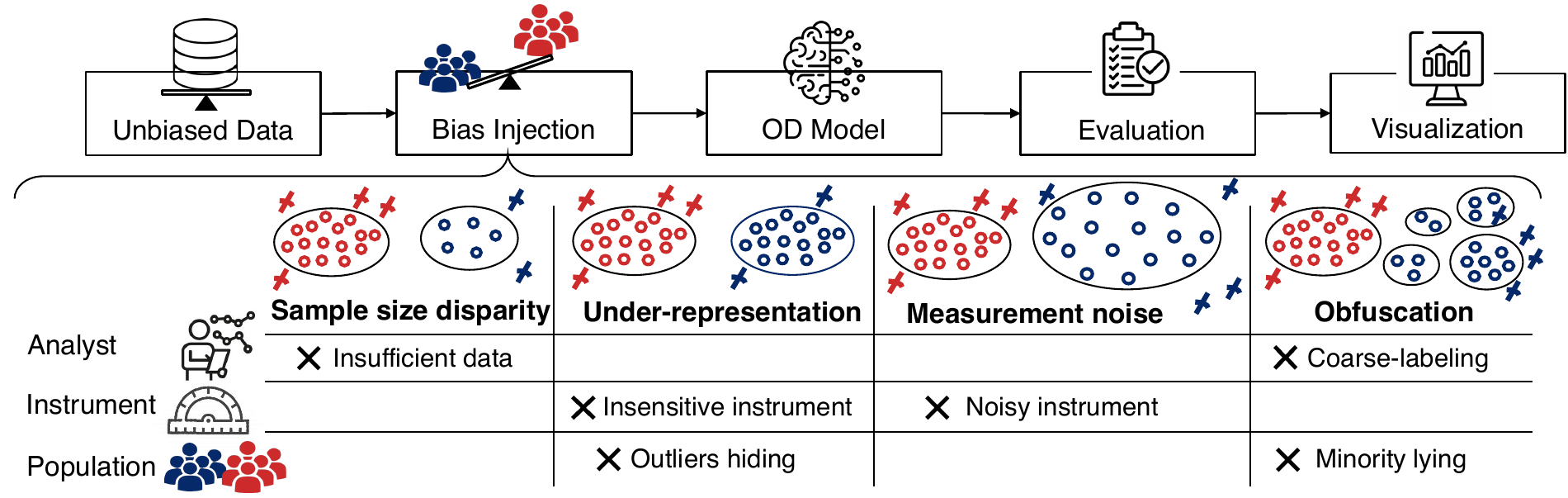}
  \vspace{-0.1in}
  \caption{Overview of our study: Starting with simulated unbiased data containing outliers, we inject known types of bias into the data in a controlled setup. We then evaluate the fairness and performance of outlier detection (OD) models under various measures and report any vulnerabilities.
  We study 
  four different OD models (both shallow\&deep, and both fairness-unaware\&fairness-enhanced), under four different types of data bias with potential implications on OD: size disparity, target under-representation, measurement error, membership obfuscation (\S\ref{ssec:biastypes}). These data biases are graphically illustrated when data points (shown with $\circ$) are composed of two protected groups (in color; red \& blue). Outliers are shown with cross-marks (\xmark).}
    \label{fig:summary}
    \vspace{-0.05in}
\end{figure*}

In this paper, we aim to contribute to a deeper understanding of the possible sources of algorithmic unfairness for unsupervised OD. 
Specifically, we empirically investigate the role of various types of data bias
as potential contributors to OD unfairness.  
Our findings highlight  the shortcomings of several established OD algorithms in the presence of four different carefully injected data biases in a controlled simulation framework. Data biases that we study are namely, group sample size disparity, target under-representation, feature measurement noise, and group membership
obfuscation.  

In addition, we identify certain data-centric properties---in particular, those related to group-wise differences in sparsity, base rate, variance,
and multi-modality---that emerge as a result of each specific data bias, making it easier to understand the interaction between such data properties and the underlying working assumptions of a given OD algorithm.
Perhaps more importantly, we remark that such data properties as pertain to group differences that emerge from data bias could as well be organic, i.e. natural characteristics of the input data. Then our results, when seen through the lens of data-centric factors (and not just from the perspective of data bias), 
provide evidence toward moving beyond the ``algorithmic bias is a data problem'' debate \cite{hooker2021moving}.

Our work is one of the few that presents a 
rigorous empirical framework toward understanding the effect of various data-centric factors on algorithmic bias, and the first one specifically focusing on unsupervised OD algorithms. We summarize our main contributions as follows. Fig. \ref{fig:summary} provides an overview.

\vspace{-0.05in}
\begin{itemize}
\item \textbf{Data bias as possible source of harm (\S\ref{sec:databias}):~} We curate a list of potential data biases in the real world that may lead to OD unfairness; as pertain to  group sample size disparity (inducing minority samples), target under-representation (distorting base rates), feature measurement noise (leading to extreme values), and group
membership obfuscation (giving rise to within-group subpopulations/multi-modality).

\item \sloppy{\textbf{Fairness (stress-)testing popular OD models (\S\ref{sec:prelim}):~} In extensive controlled simulation settings whereby we inject known data biases, we (stress-)test several established OD techniques (both fairness-unaware as well as fairness-enhanced) w.r.t. both fairness and performance metrics.}

\item \textbf{Extensive empirical analysis (\S\ref{sec:empirical}):~} Our analysis shows that all OD methods under study are susceptible to data bias, although their robustness vary notably depending on the type of bias. This suggests that without a clear understanding of the type of biases a dataset may exhibit, it would be a challenge to choose an effective OD model to employ.
Further, not only data bias may be a source of OD unfairness, but in some cases it also severely harms the overall detection performance.

\item \textbf{Theoretical analysis (\S\ref{sec:algobias}):~} We provide detailed analyses of how certain data biases interact with the working mechanism of various OD models, leading them to incur disparate impact on different populations in the data.

\item \textbf{Evidence to move beyond ``(OD) algorithm bias is a data problem'' (\S\ref{sec:algobias}):~} A key understanding derived from our study is that what drives OD algorithm bias is mainly the misalignment between an algorithm's working assumptions/design choices and certain input data characteristics; e.g. whether the detector respects density variability in the feature space or is susceptible to outlier masking (the notion that  outliers get hidden when clustered).
Crucially, while such data-centric factors may arise as a result of data bias, they could also simply be organic. This implies that OD algorithm bias can arise due to reasons beyond data bias.

\end{itemize} 

\textbf{Accessibility and Reproducibility:} 
All code and datasets are open-sourced at \url{https://github.com/xyvivian/ODBias.git}.

\vspace{-0.05in}
\section{Data Bias: Types and Sources}
\label{sec:databias}

\subsection{Preliminaries}   
In general, algorithmic decisions could be punitive (e.g. imprisoning) or assistive (e.g. loan approval).  Our work focuses on the former, considering the scenarios where outlier detection (OD) is often applied on population data to flag risky individuals.

For simplicity, we represent a population as composed of two protected groups, associated with a sensitive attribute $G \in \{a,b\}$. We assume group $b$ is the underprivileged.
We denote by $\mathbf{X} \in \mathbb{R}^d$ and $Y \in \{0,1\}$ the sets of input features and the target, respectively. 
In OD scenarios $Y$ translates to \textit{true risk}, and the task is to effectively flag the individuals with $Y=1$. Let
$S \in \mathbb{R}$ indicate the output outlierness score of an OD algorithm and $O \in \{0,1\}$ is the decided outlier label based on $S$ given a threshold. 

Given the shorthand notation, the \textit{base rate} (also, prevalence) of a group is defined as $br_G = P(Y=1|G)$.
On the other hand, the 
\textit{flag rate} of a group by a detector is given as $fr_G=P(O=1|G)$.

\subsection{Four Types of Data Bias} 
\label{ssec:biastypes}
Based on a survey of the algorithmic fairness literature 
that covers possible sources of algorithmic harm  \cite{suresh2021framework,mehrabi2021survey,wang2020robust,akpinar2022sandbox}, 
we curated a list of data biases most relevant to OD applications in the real world. We provide an overview of four such biases in the context of OD as follows. We also discuss possible sources---the data collector, the measurement instrument or the population---that may drive the presence of each type of bias in the data. Fig. \ref{fig:summary} presents a flowchart of our work, along with a graphical illustration of the data biases discussed in this section.

\vspace{-0.05in}
\subsubsection{\bf Group sample size bias:~}

This type of bias reflects the real-world scenarios with data scarcity for the underprivileged sub-population (group $b$ in our study). In effect, group $b$ with a smaller size constitutes the statistical minority samples in the dataset.
It is easy to see that OD would be susceptible to this bias as its goal is identifying outliers, i.e. statistically rare instances in the data.

The presence of sample size bias can be attributed to the data collector whom might have failed to collect enough samples for some group(s). It may also be driven by the population itself, where some groups are in minority by nature.

\vspace{-0.05in}
\subsubsection{\bf {Target} under-representation bias:~}  This type of bias is similar to the sample size bias, but only impacting the \textit{positively-labeled} (i.e. target) individuals in the underprivileged group $b$. In particular, the rate of positive instances (i.e. the base rate) appear smaller in group $b$ than that in group $a$.  In effect, base rates are purported and the assumption of equal base rates across groups may no longer hold.

This bias could stem in the real world from the measurement instrument doing a poor job in ``sensing'' such individuals, as well as from the fact that  
such individuals may be better at ``hiding'' from being sampled. Alternately, the base rates  may simply be different between the groups by nature.

\vspace{-0.05in}
\subsubsection{\bf Feature measurement %
or response bias:~}
This type of bias is reflective of systematically erroneous or noisy measurements in the real world associated with individuals from the underprivileged group.
Such errors and noise may inflate the variance of certain features, and thereby increasing the propensity of extreme values. Extreme values appearing as outliers, in turn, makes OD susceptible to this type of data bias.

The measurements for some group(s) may also display systematic under- or over-estimations for certain features. Consider for example the number of (re)arrests when used as a feature for risk assessment in various domains, which could be over-represented for African-Americans due to racial disparities in arrests \cite{schleiden2020racial}. Another example is the systematically under-estimated SAT scores for African-American students due to implicit biases in the questions that lead to racial disparities in outcomes \cite{rattani2016sat}.

The source of this bias is often the measurement instrument (such as questions in the survey or test, the camera, the lab test, etc.) which, for various reasons, is not well tuned to measuring underprivileged individuals accurately. In other cases, the source could be the population itself due to self-reporting; for example, when surveys are used to collect the measurements, some group(s) may provide more erroneous or noisy responses.

\vspace{-0.05in}
\subsubsection{\bf Group-membership obfuscation bias:~}

This bias mimics the real world scenarios in which some underprivileged individuals misreport their group membership %
for various reasons including fear of disclosure or discrimination \cite{wang2020robust}. In addition to obfuscating race/gender/etc., we expect them to also alter several of their demographic/proxy features that correlate with the sensitive attribute, to camouflage and better align with this obfuscation. 
For example, individuals who falsify race may also misreport their country of origin, often trying to resemble in some aspects to those individuals in privileged groups. 
While obfuscation may occur in all groups, we assume it to be more prevalent for the underprivileged.

This bias, stemming from the population forging various feature values, induces ``fragments'' in the data, i.e. subgroups within a group. In the presence of obfuscation, an underprivileged group would break into smaller subgroups, resembling other groups in some ways but not others. This can be considered as heterogeneity or multi-modal distribution within the group. Interestingly, such heterogeneity, i.e. a group composed of various different subpopulations, could be an organic property (e.g. in a large country like India, not all Indians are alike). In such a case, it can be seen as a labeling misstep attributed to the data collector to file multiple heterogeneous subgroups under a single sensitive attribute value.

\vspace{-0.05in}
\subsubsection*{\bf Remarks.~} The four data biases we focus on in this study are not necessarily comprehensive. While there may be other OD-relevant data issues that may trigger unfair OD outcomes, there are also other known data biases that are not applicable in OD settings. For example, the problem of ``tainted'' (historical) labels does not apply, since unsupervised OD algorithms do not train on $Y$. Moreover, train/test data distribution mismatch problem is not a concern, as we consider transductive OD wherein outliers are to be detected in a given dataset, with test data being the same as train  data.
While all possible sources of algorithmic harm are not fully understood, it is our plan to expand our current list in the future.

\section{OD under Data Bias: Sandbox Setup}
\label{sec:prelim}

Our goal is to study the effectiveness of OD algorithms under counterfactually injected bias into the simulated unbiased data. Thus, we restrict our study to simulated data. This choice is intentional, because real-world datasets lack an in-depth understanding of whether they exhibit any data bias or what types of bias(es) they may exhibit if any.
In this regard, our work parallels the study by \citet{akpinar2022sandbox} which performed  similar fairness and fidelity analyses on simulated controlled environments for classification.

In this section, we present the details of our (unbiased) data simulation (\S\ref{ssec:simulation}), bias injection steps (\S\ref{ssec:inject}), the OD models under study (\S\ref{ssec:models}), and the evaluation metrics (\S\ref{ssec:metrics}).

\vspace{-0.05in}
\subsection{Data Simulation}
\label{ssec:simulation}

To ensure we start with an unbiased dataset (into which we will inject specific, known biases), we simulate a population with the input features and the target described as follows.

The individuals are represented in a feature space consisting of three types of variables;  $\mathcal{X} = \Xg \cup \Xc \cup \Xo$, where 
\cbit
\item $X_g \in \Xg$ denotes the set of \textit{proxy} variables that correlate with $G$roup membership $G$, but not with the target  $Y$; 
\item $X_c \in \Xc$ depicts the  ``$C$ulprit'' or \textit{incriminating} variables that are relevant (to the OD task) and hence correlate with or are reflective of the target $Y$; and
\item $X_o\in \Xo$ capturing the \textit{non-incriminating} $O$cclusion (or irrelevant) variables that  neither correlate with $G$ nor $Y$. The inclusion of such attributes make the OD task more realistic and non-trivial, as outliers often hide within feature \textit{subspaces}, i.e. they stand out only w.r.t. a few relevant (but not all)  features (in this case $X_c$).
\ceit 

Note that the culprit variables associate with true risk (i.e. the target) by design, while the proxy variables do not. This implies that we assume \textit{equal base rates} across groups in the unbiased dataset. As we will discuss later, this may not always be the case in real world OD settings,  where unequal base rates across groups (e.g. Internet crime propensity by ethnicity)
may exist by nature.

\begin{definition}[Fair Outlier Detection]
   In terms of $X_g, X_c$ and $X_o$, the output of an OD algorithm is considered fair as long as \beq P(O =1| X_c, X_g, X_o) \;=\; P(O =1| X_c) \;, \eeq that is, when the assigned outlier labels are independent of group membership/proxy variables as well as irrelevant features,  given the incriminating variables. 
\end{definition}

To stress-test the fairness of OD algorithms in a controlled setting, we simulate an equal number of ($1000$) samples per group, and also set equal base rates, $br_a=br_b$ at $0.05$ or $0.1$, since outliers are rare.  We use Gaussians to simulate the inliers, with group-wise means respectively at 5 and 20 for $\Xg$, and zero-mean for $\Xc$, both with unit standard deviation. $\Xo$ is drawn from a standard Gaussian uniformly at random.

We create two separate datasets by  injecting  \textbf{clustered} (repetitive or collusive) and \textbf{scattered} outliers, respectively. Clustered outliers have the same distribution for $\Xg$ and $\Xo$ as with the inliers, but in $\Xc$ they are drawn from a Gaussian with a higher mean of 3. 
Scattered outliers are created by randomly sampling a subset of the dimensions in $\Xc$, and inflating the variance (originally, 1) by a factor of $\{3,6,9,12,15\}$ chosen uniformly at random per outlier.

Finally, we set an equal number of dimensions ($5$) for $\Xg$, $\Xc$, and $\Xo$. While it remains a valuable direction to investigate the severity of  unfairness that data bias may drive as a function of the relative number of the proxy and irrelevant variables to the number of incriminating variables, we focus on unfairness for the case when they all have equal proportions in the feature space.

\vspace{-0.05in}
\subsection{Data Bias Injection}
\label{ssec:inject}
The crux of our sandbox is the study of OD  algorithms under different types of known data biases, as discussed in \S\ref{sec:databias}. %
We describe the steps for injecting our unbiased data with each bias type as follows. For brevity, additional details are given in Apdx. \S\ref{asec:exp_details}.

\vspace{0.05in}

\begin{enumerate}[label=\arabic*.]
\item  {{\bf {Group} sample size bias:~}} With probability $\beta_s$, we independently exclude samples from the dataset where $G=b$.
This leads to sample size disparity between groups. 

We inject varying degrees of sample size disparity
using $\beta_s \in \{0.01, 0.05, 0.10, 0.2, 0.4, 0.6, 0.8\}$.

\vspace{0.025in}
\item \textbf{Target under-representation bias:} With probability $\beta_u$, we independently exclude samples where $G=b$ \textit{and} $Y=1$. This is sample size bias impacting only the positively-labeled (i.e., target) individuals in the underprivileged group. We study bias $\beta_u \in \{0.01, 0.05, 0.10, 0.2, 0.4, 0.6, 0.8\}$.

\vspace{0.025in}
\item \textbf{ Feature measurement %
or response bias:~}
To reflect measurement noise, we inject variance-shift; where 
 we  inflate (i.e. multiply) the variance of the distributions of $\Xg$ and $\Xc$ for $G=b$ by a factor of $\beta_v$. To ensure that the supports of the group-wise distributions remain separate after variance inflation, we set the means of $\Xg$ to 5 and 20 for group $a$ and $b$, respectively.
Similarly, to be able to distinguish outlier-vs-inlier distributions upon inflation, we set the means of $\Xc$ to 0 and 10 for inliers and outliers, respectively.
We vary $\beta_v \in \{0, 0.05, 0.1, 0.2, 0.5, 1, 2, 4, 6\}$.

To mimic systematic over-estimation of task-relevant variables, we 
also experiment with mean-shift bias, where the mean of $\Xc$ is shifted additively for $G=b$ by $\beta_m \in \{0,2,4,6,8\}$.

\vspace{0.025in}
\item  \textbf{ Group-membership obfuscation bias:~}
With probability $\beta_g$, we flip/swap the group membership of individuals with $G=b$ to $G=a$. Note that following un-awareness, OD models do not use the group membership indicator/variable $G$ for detection. Thus, for individuals in $G=b$ whose membership has been flipped, we also swap/draw a random subset of their feature values in $\Xg$ from the distributions of $G=a$, which alters the input data. 

This bias mimics the real world scenarios in which some underprivileged individuals misreport their group membership %
for various reasons including fear of disclosure or discrimination \cite{wang2020robust}, in addition to altering several of their demographic/proxy features in $\Xg$ to better align with this obfuscation. While obfuscation may occur in both groups, we 
assume more prevalent obfuscation in the underprivileged group; setting the rate for $G=a$ to be zero for simplicity, and performing our study as $\beta_g$ varies for $G=b$, in $\{0.05, 0.1, 0.15, 0.2, 0.3, 0.4\}$.

\end{enumerate}

\vspace{-0.05in}
\subsection{Outlier Detection Models} %
\label{ssec:models}

In this work, we aim to stress-test the  performance and fairness of OD algorithms in the face of data bias. 
There is a long list of algorithms for OD \cite{aggarwal2016outlier}, including the modern deep neural network  models \cite{pang2021deep}. The vast majority of these do not consider fairness, while only recently a handful of fairness-enhanced OD algorithms are proposed (see \S\ref{sec:related}). 

To be representative of the literature while keeping the computational effort of our study feasible, we consider four OD models comprising both ``shallow''\&deep, and both fairness-unaware\&fairness-
enhanced detectors: Local Outlier Factor (\textbf{LOF}) \cite{breunig2000lof}, Isolation Forest (\textbf{iForest}) \cite{liu2008isolation}, \textbf{DeepAE} \cite{zhou2017anomaly}, and \textbf{FairOD} \cite{shekhar2021fairod}. LOF and iForest are two popular shallow techniques that have been shown to be most effective on benchmark evaluations \cite{emmott2013systematic}. Those two are also \textit{mechanistic}, i.e. directly model/define what an outlier is\footnote{For LOF, a point with larger reachability distance than its neighbors is an outlier. For iForest, an outlier is a point that can be isolated with few randomized axis-splits.}. On the other hand, DeepAE and FairOD are \textit{learning-based} and leverage deep neural networks with end-to-end learnable parameters. Moreover, FairOD is \textit{fairness-enhanced} while others are standard detectors. We present more details as follows. (Further details on LOF and iForest can be found in Apdx. \S\ref{assec:preliminary}, based on which we theoretically analyze these mechanistic OD models 
later in \S\ref{sec:algobias} and Apdx. \S\ref{assec:theory_result}.)

\vspace{-0.025in}
\subsubsection{\bf Local Outlier Factor.~} LOF outlier score is based on the reachability distance of a point to its $k$ nearest neighbors (NN) relative to those distances for its NNs. As its name suggests, LOF evaluates a point w.r.t. the \textit{local density} in the vicinity it resides. 

\vspace{-0.025in}
\subsubsection{\bf Isolation Forest.~}
iForest makes random threshold cuts sequentially on features chosen at random, thus building an ensemble of extremely randomized trees, and considers the  average number of steps required to \textit{isolate} a point from others as its outlier score. 

\vspace{-0.025in}
\subsubsection{\bf DeepAE} Based on a deep autoencoder architecture, DeepAE employs \textit{compression} followed by decompression of the data points, where the reconstruction error is taken as the outlier score. 
The working assumption is that the majority of the data (i.e. inliers) exhibit patterns which can be compressed well while minimizing the total (reconstruction) loss, while outliers that do not obey such patterns receive poor reconstruction.  

\vspace{-0.025in}
\subsubsection{\bf FairOD}
Finally, the {fairness-enhanced} {FairOD} also uses a deep autoencoder as a base model but enhances its loss objective with two additional  terms for fairness regularization; one enforcing  statistical parity and another toward achieving  a heuristic approximation of the equality of opportunity (a.k.a. recall or TPR parity). 

\vspace{-0.025in}
\subsubsection*{\bf Hyperparameter (HP) tuning.~}  
OD algorithms come with HPs; e.g. number of nearest neighbors $k$ for LOF and many others for deep models including architectural (depth and width), regularization (e.g. dropout rate) and optimization (e.g. learning rate) HPs. Critically, OD model performance is quite sensitive to HP choices \cite{ma2023need,ding2022hyperparameter}, which is nontrivial to set without any labeled data.

Notably, we search for the {\em best HP configuration} on each testbed, to prevent the situation where poor HP setting becomes a possible confounding source of unfairness.
In other words, we
carefully tune the HP(s) on the overall input population (without considering group memberships), to be able to attribute observed unfairness %
solely to (injected) data bias, rather than to poor HP tuning.

For standard algorithms LOF, iForest and DeepAE, we set HPs that yield the highest overall performance, and for fairness-aware FairOD we pick the HPs that yield the best performance-fairness trade-off. Details on HP configurations for each model %
is given in Apdx. \S\ref{asec:configs}.
Interestingly, even under optimal HPs, we observe negative implications of data bias not only on fairness but also on detection performance, as we present results shortly in \S\ref{sec:empirical}.

\vspace{-0.05in}
\subsection{Evaluation Metrics}
\label{ssec:metrics}
We evaluate detection performance by AUROC; area under the ROC curve as well as F1; the harmonic mean of Precision and Recall. 
AUROC quantifies the overall ranking, while F1 requires a threshold on the outlier scores. We set a threshold on $S$ to obtain as many flagged outliers with $O=1$ as  number of true outliers with $Y=1$.

We evaluate fairness based on \textit{group-wise ratios} w.r.t. ($i$) positive or flag rates (FR), ($ii$) true positive rates (TPR), ($iii$) false positive rates (FPR) and ($iv$) positive predictive values (PPV, a.k.a. Precision). 

When FR ratio $\frac{fr_a}{fr_b}$ is equal to 1, demographic or statistical parity is satisfied.
One caveat is when group base rates differ, i.e. $br_a\neq br_b$. Then, it is suitable to measure bias amplification \cite{li2022data} as  the ratio between FR ratio and the ground truth base rate ratio. For all other ratios, the ideal/unbiased value is 1.
Unequal rates imply over/under-policing in monitoring for punitive OD settings.

For shallow and deep models, we report results averaged over 10 and 5 independent simulation runs, respectively, as the latter take longer to train and tune (more HPs).

\section{\hspace{-0.1in}OD on Biased Data: Empirical Findings}
\label{sec:empirical}

\definecolor{armygreen}{rgb}{0.0, 0.5, 0.0}

\begin{table}[!t]
\vspace{-0.075in}
	\caption{
 Qualitative summary of results for fairness stress-testing  of various OD algorithms (columns) when different type data bias (rows) is applied on group $b$ with \underline{clustered} outliers. 
 \underline{Shades of colors} \textcolor{red}{red} and \textcolor{blue}{blue} depict the degree of \underline{\textit{unfairness}}, when disproportionately  inflicted on group \textcolor{red}{$a$} or group \textcolor{blue}{$b$}, respectively; blank implying no notable difference. %
 \underline{Arrow counts} reflect relative change in overall detection \underline{\textit{performance}}; no arrows implying no notable change.} 	
			\vspace{-0.1in}
			\begin{tabular}{l|p{1.05cm}|p{1.05cm}|p{1.05cm}|l} 
				\toprule
    Bias type / OD model   &
    	{LOF} & 
     {iForest} & 
     {DeepAE} & 
     FairOD  \\
       \midrule

       Size disparity (Fig. \ref{fig:samplesize_all_cluster})  
       & \cellcolor{red!30}   \centering $\downarrow\downarrow$
       & \cellcolor{blue!80} \centering $\downarrow\downarrow\downarrow$
       &  \cellcolor{blue!50} 
       &  \cellcolor{blue!10} \\ 
       \hline
       
       Under-repres. (\S\ref{assec:urep_cluster})   
       &  \cellcolor{blue!30} %
       &  \cellcolor{blue!20}
       &   \cellcolor{blue!20}
       &   \cellcolor{blue!30} \\ 
       \hline
       
       Meas. noise (\S\ref{assec:meas_cluster})  
       &  
       &  \cellcolor{blue!20}
       &  \cellcolor{blue!5} 
       &  \\ 
       \hline
       Obfuscation (\S\ref{assec:obf_cluster})  
       & \cellcolor{blue!70} \centering $\downarrow\downarrow$    
       & \cellcolor{blue!40} \centering 
       & \cellcolor{blue!60} \centering 
 $\downarrow$  
        & \cellcolor{blue!30} %
        \\ 
         \bottomrule
       \end{tabular}

       \label{tab:summary}
	\vspace{-0.15in}
\end{table}

Starting with the unbiased datasets containing an equal number of samples in each group as well as equal group base rates, we create different biased datasets by varying $\beta_s$, $\beta_u$, $\beta_v$ (or $\beta_m$), and $\beta_g$, depending on the type of bias being injected.
Note that the bias is always injected on group $b$. Each model's hyperparameters are tuned as described in \S\ref{ssec:models} for each resulting dataset separately.

Tables \ref{tab:summary} and \ref{tab:summary_scatter} (best in color)  provide a qualitative summary of our results across OD models and bias types for datasets with \textit{clustered}  and \textit{scattered} outliers, respectively. We discuss several of our notable findings in this section, while the detailed results are given in Apdx. \S\ref{asec:exp_resuls} for brevity.

{Overall, as depicted by the varying shades of blue, we see that the bias-injected group $b$ is impacted disproportionately  across models and bias types (with a couple of exceptions), although the severity of unfairness against $b$ varies. 
Moreover, data bias impacts not only fairness but also detection performance of OD models; in fact, there are cases when performance drops considerably,  as depicted by downward arrows.
}

{It is interesting to note the stark difference between the detectors in regard to susceptibility to different biases. For example, LOF is most susceptible to Obfuscation in the clustered-outliers setting and to Measurement bias for scattered outliers. iForest is susceptible to Sample size disparity, while DeepAE is most sensitive to Obfuscation in both settings. Fairness-enhanced FairOD is no exception;  as it remains comparably brittle under Obfuscation. These suggest the lack of a ``winner'' detector. %
}

\definecolor{armygreen}{rgb}{0.0, 0.5, 0.0}

\begin{table}[!t]
\vspace{-0.075in}
	\caption{
Qualitative summary of results for fairness stress-testing  of various OD algorithms (columns) when different type data bias (rows) is applied on group $b$ with \underline{scattered} outliers. 
 Shades of \textcolor{red}{red} and \textcolor{blue}{blue} depict the degree of \textit{unfairness}, if disproportionately  inflicted on group \textcolor{red}{$a$} or group \textcolor{blue}{$b$}, respectively; blank implying no notable difference. 
 Arrows reflect the relative change in detection \textit{performance}; no arrows implying no notable change.} 	
			\vspace{-0.1in}
			\begin{tabular}{l|p{1.05cm}|p{1.05cm}|p{1.05cm}|l} 
				\toprule
    Bias type / OD model   &
    	{LOF} & 
     {iForest} & 
     {DeepAE} & 
     FairOD  \\
       \midrule

       Size disparity (\S\ref{assec:sizebias_scatter})  
       & \cellcolor{red!30} 
       \centering $\downarrow$
       & \cellcolor{blue!70} \centering $\downarrow\downarrow$
       & \cellcolor{blue!10} 
       & \cellcolor{blue!10}   \\ \hline

       Under-repres. (\S\ref{assec:urep_scatter})   
       &  \cellcolor{blue!20}
       &  \cellcolor{blue!30}
       &  \cellcolor{blue!20}
       &  \cellcolor{blue!30}
       \\ 
       \hline
       Meas. noise (\S\ref{assec:meas_scatter})  
       &  \cellcolor{blue!60} \centering $\downarrow$
       &  \cellcolor{blue!60} \centering $\downarrow\downarrow$
       &  \cellcolor{blue!70} \centering $\downarrow\downarrow$
       &  \cellcolor{blue!40}$\;\;\;\downarrow\downarrow$\\ 
       \hline
       Obfuscation (\S\ref{assec:obf_scatter})  
       & \cellcolor{blue!30} \centering $\downarrow$  
       & \cellcolor{blue!40} \centering $\downarrow$
       & \cellcolor{blue!70} 
       \centering $\downarrow$ 
       & \cellcolor{blue!70}$\;\;\;\;\downarrow$ \\ 
         \bottomrule
       \end{tabular}

       \label{tab:summary_scatter}
	\vspace{-0.2in}
\end{table}

In addition, we find that the models behave quite differently against a certain bias depending on the dataset characteristics. Notable is the Measurement bias, where models are fairly robust when outliers are clustered, which however lead to considerable unfairness as well as performance drop on datasets with scattered outliers.

 We present a detailed analysis of the findings in the following subsections, respectively for each type of data bias.

\vspace{-0.05in}
\subsection{Group sample size bias}
\label{ssec:sizebias}

\begin{figure*}[!ht]
    \centering
    \begin{tabular}{lllll}
     \hspace{-0.1in}{\includegraphics[width=0.19\textwidth]{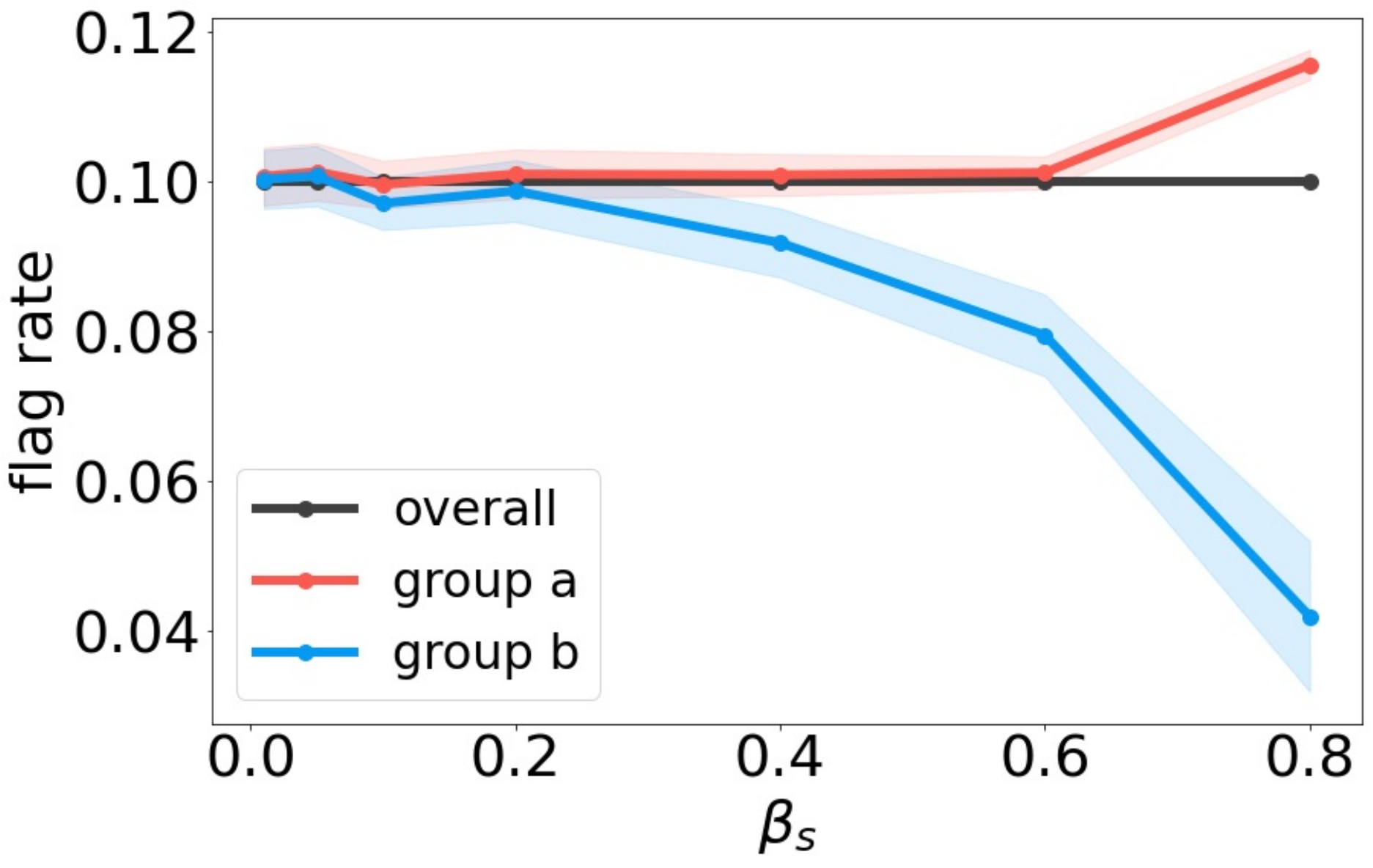}} 
    &
    \hspace{-0.1in}{\includegraphics[width=0.19\textwidth]{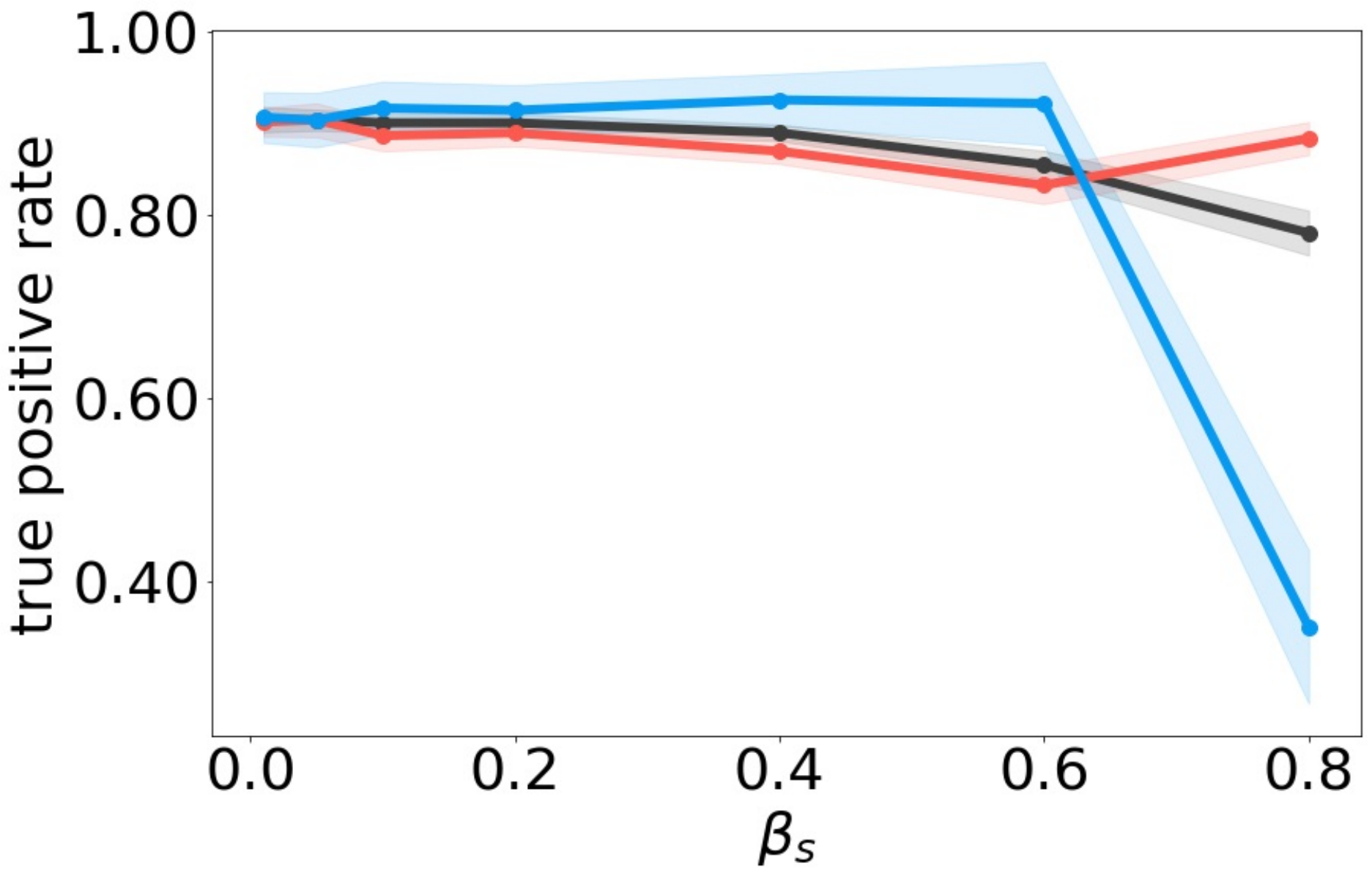}} &
     \hspace{-0.1in}{\includegraphics[width=0.19\textwidth]{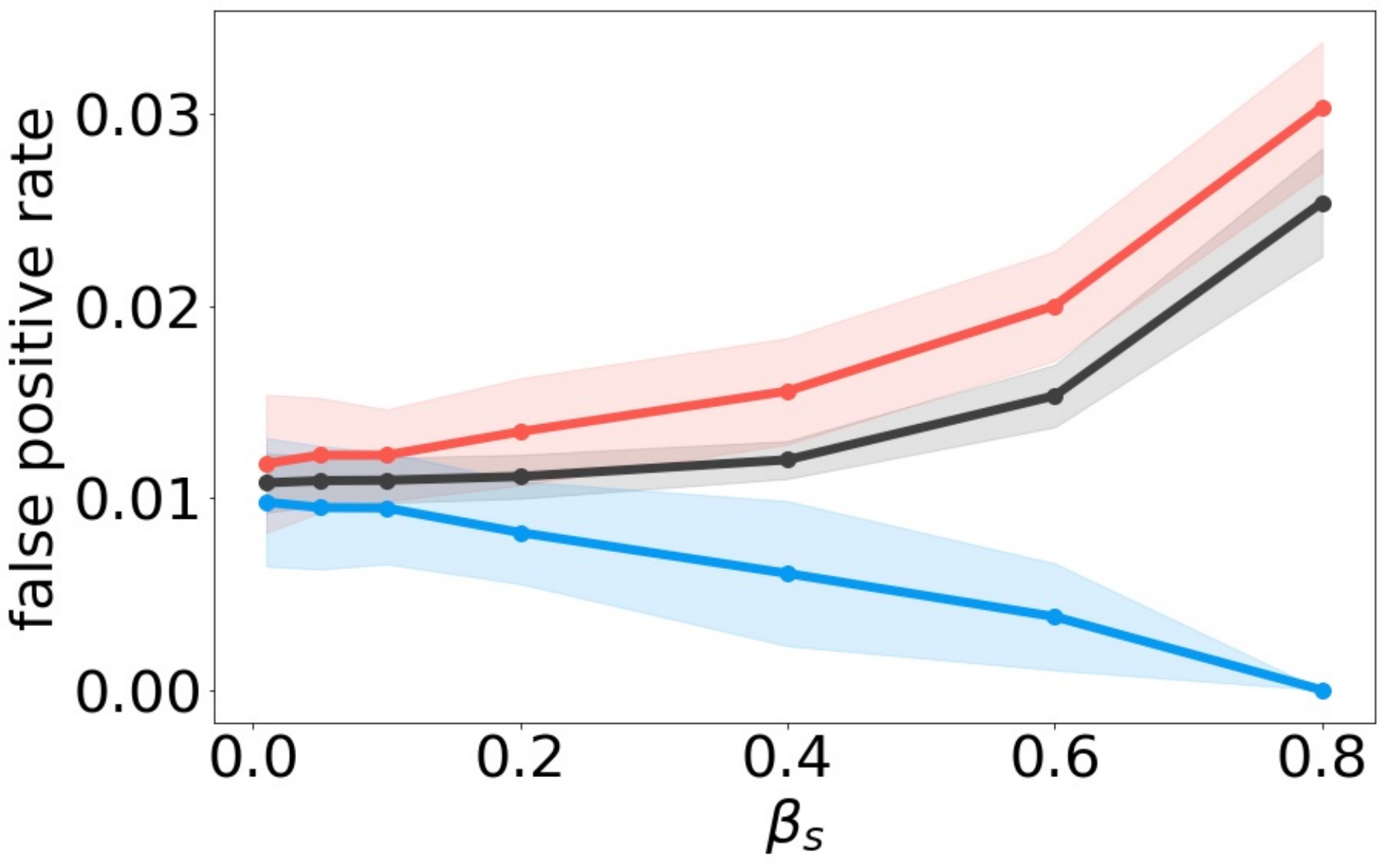}}
    &
   \hspace{-0.1in} {\includegraphics[width=0.19\textwidth]{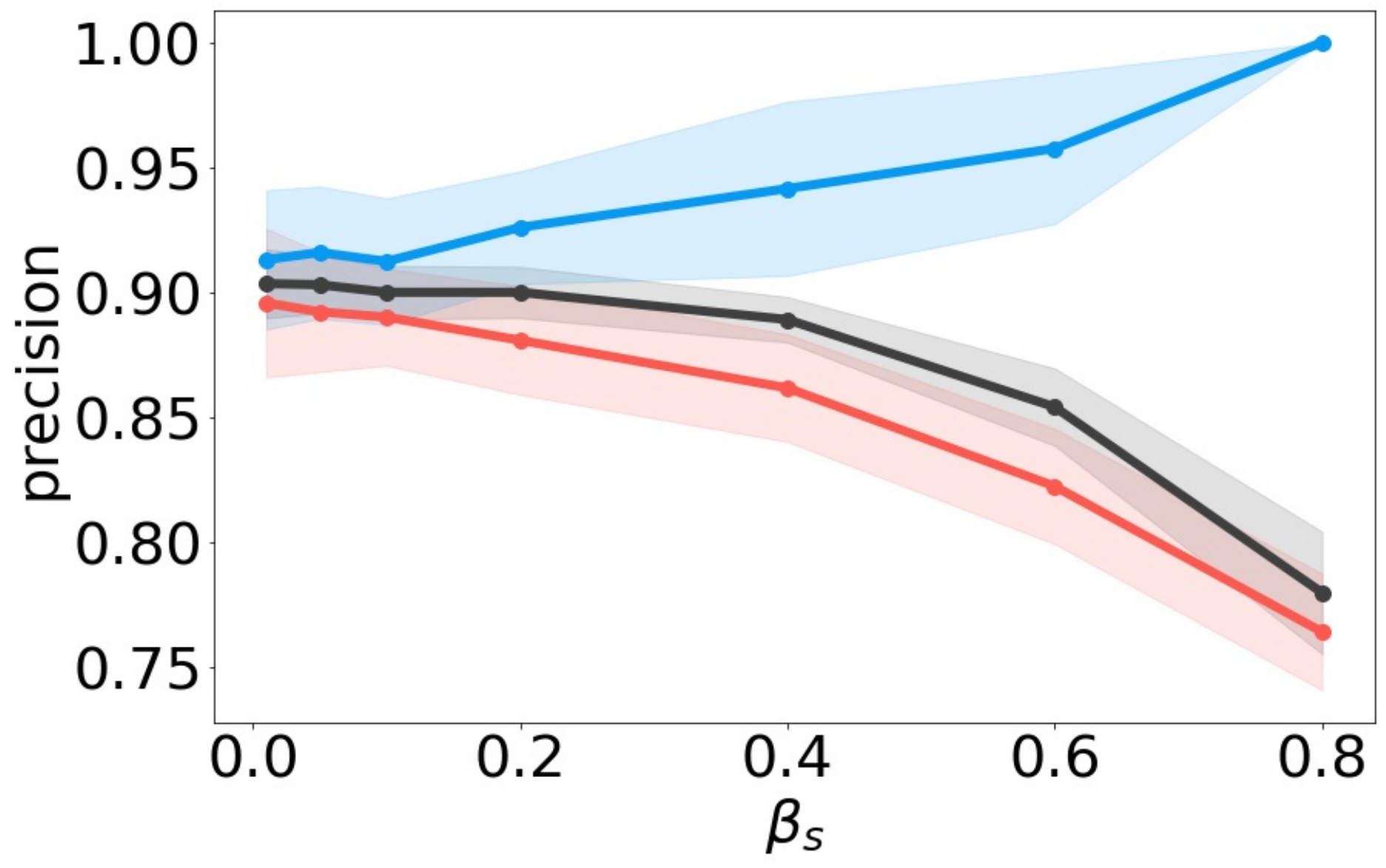}}
   &
    \hspace{-0.1in}{\includegraphics[width=0.19\textwidth]{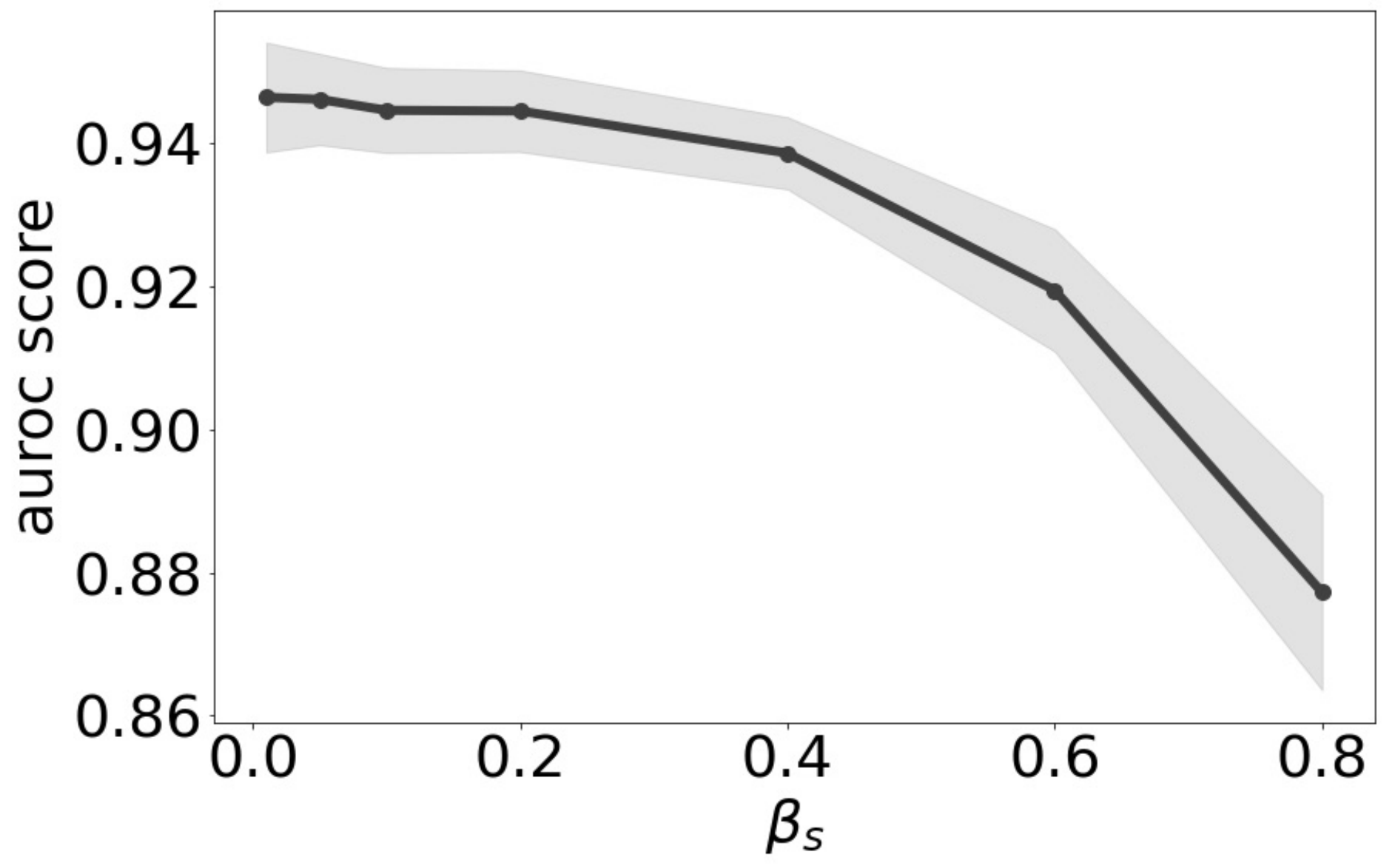}} 
   \\
     \hspace{-0.1in}{\includegraphics[width=0.19\textwidth]{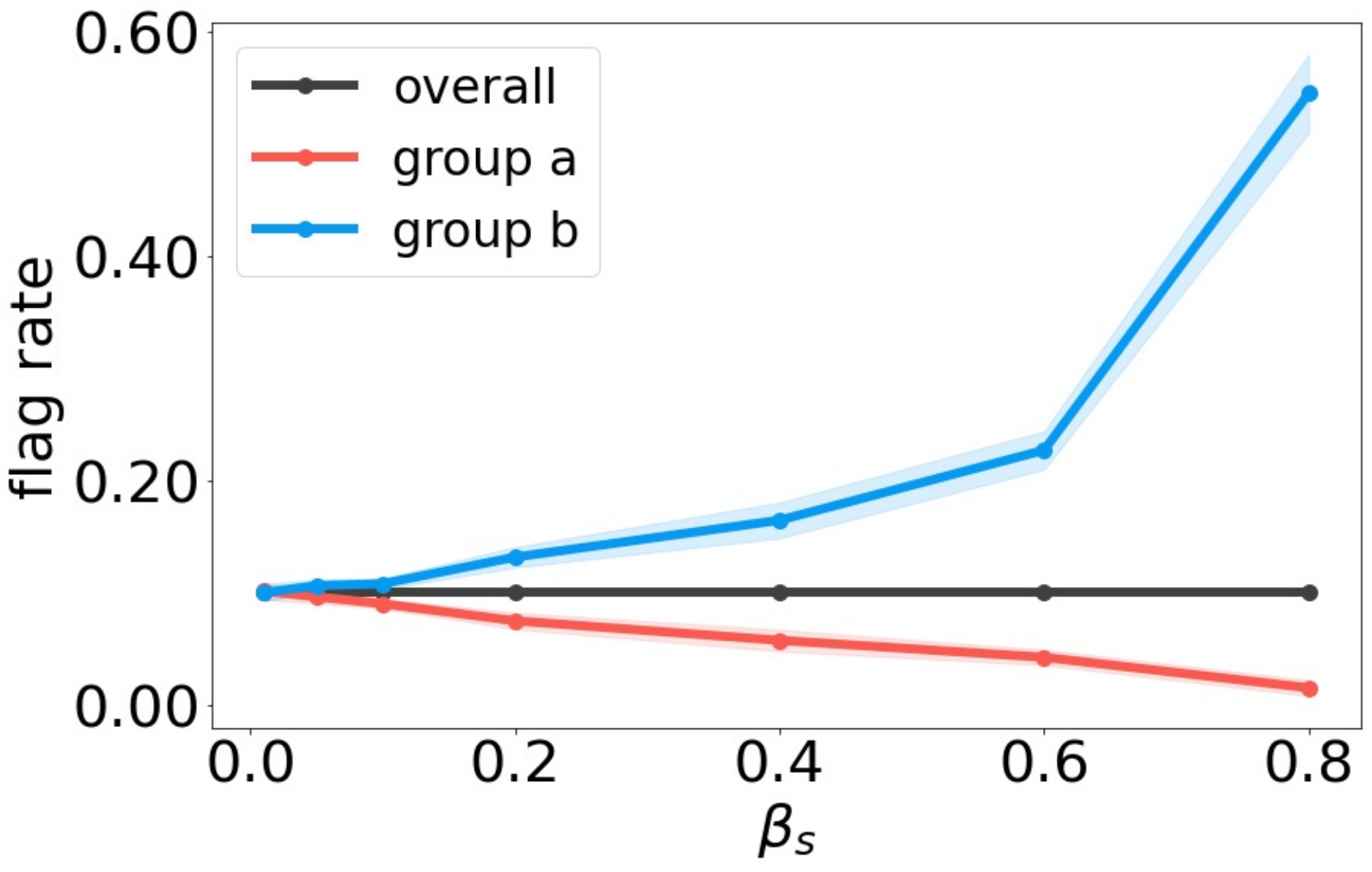}} 
    &
    \hspace{-0.1in}{\includegraphics[width=0.19\textwidth]{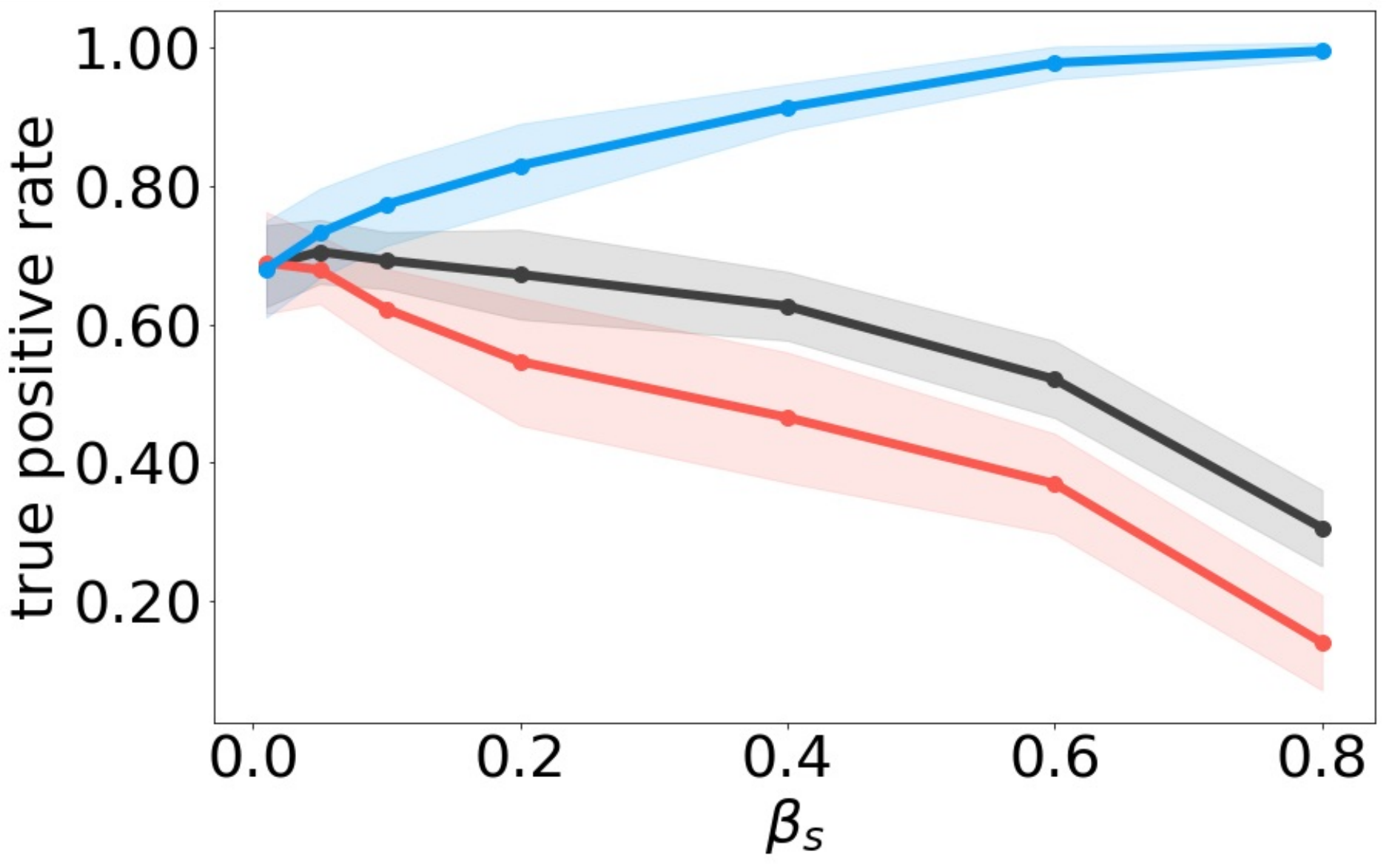}} &
     \hspace{-0.1in}{\includegraphics[width=0.19\textwidth]{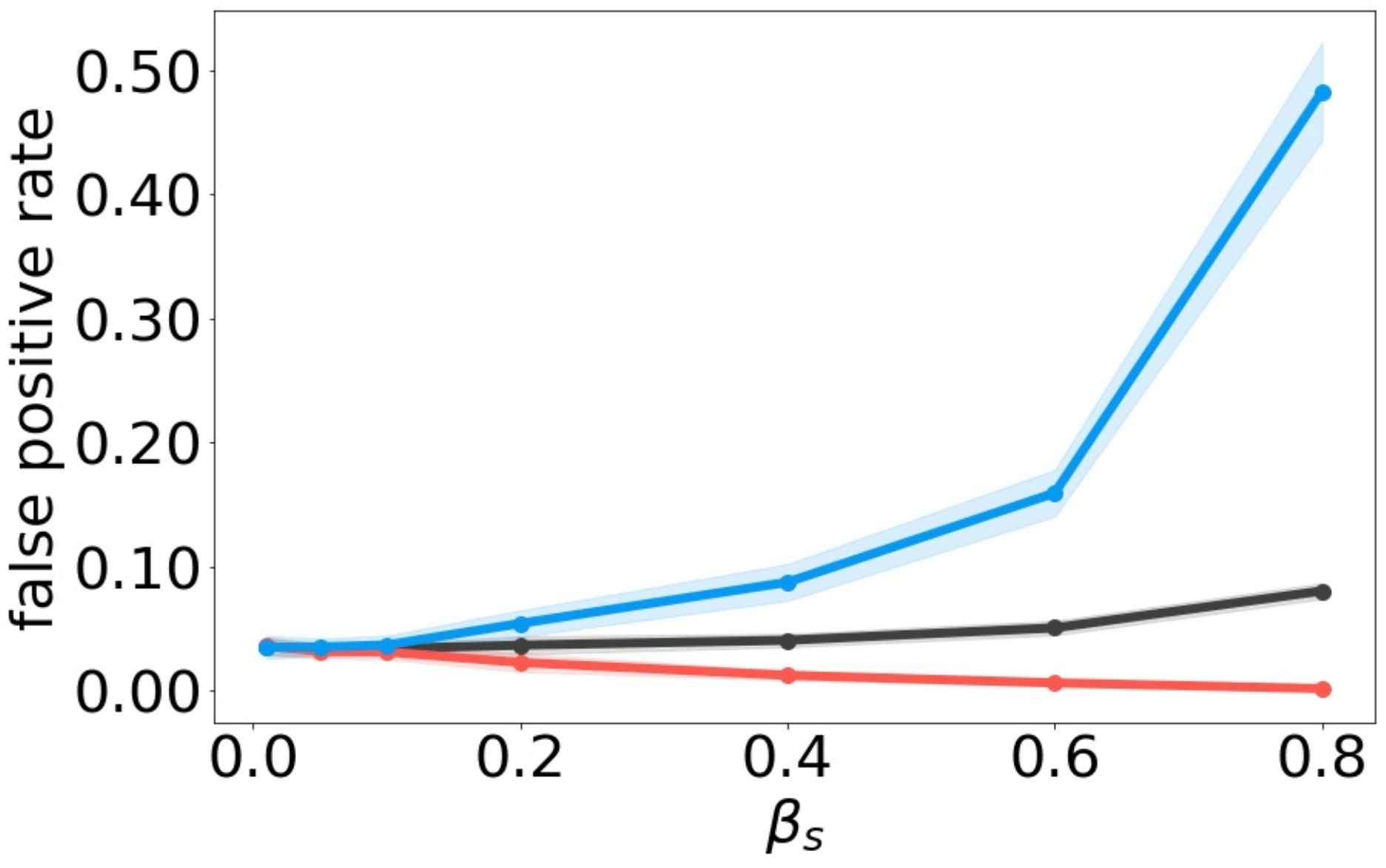}}
    &
   \hspace{-0.1in} {\includegraphics[width=0.19\textwidth]{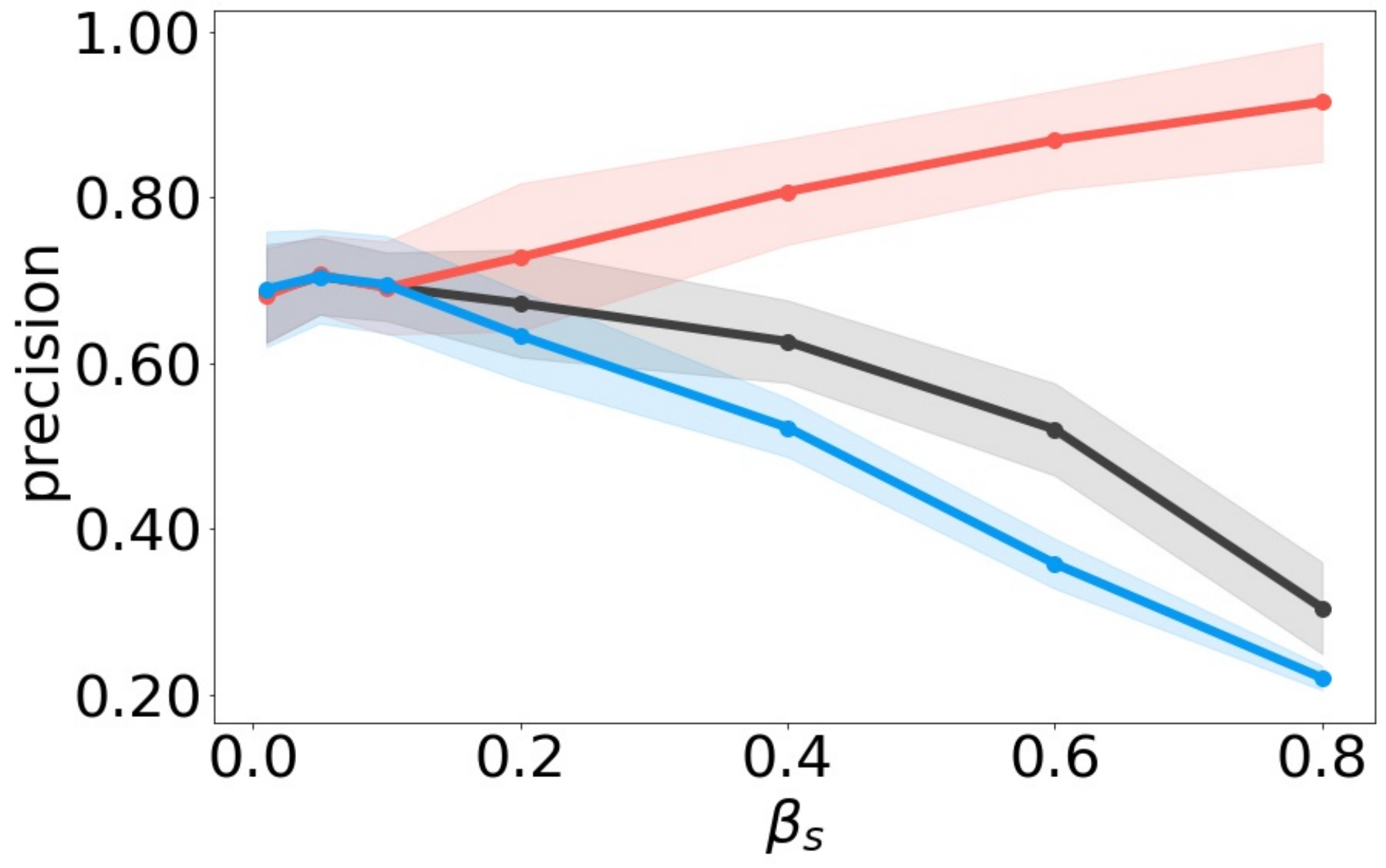}}
   &
    \hspace{-0.1in}{\includegraphics[width=0.19\textwidth]{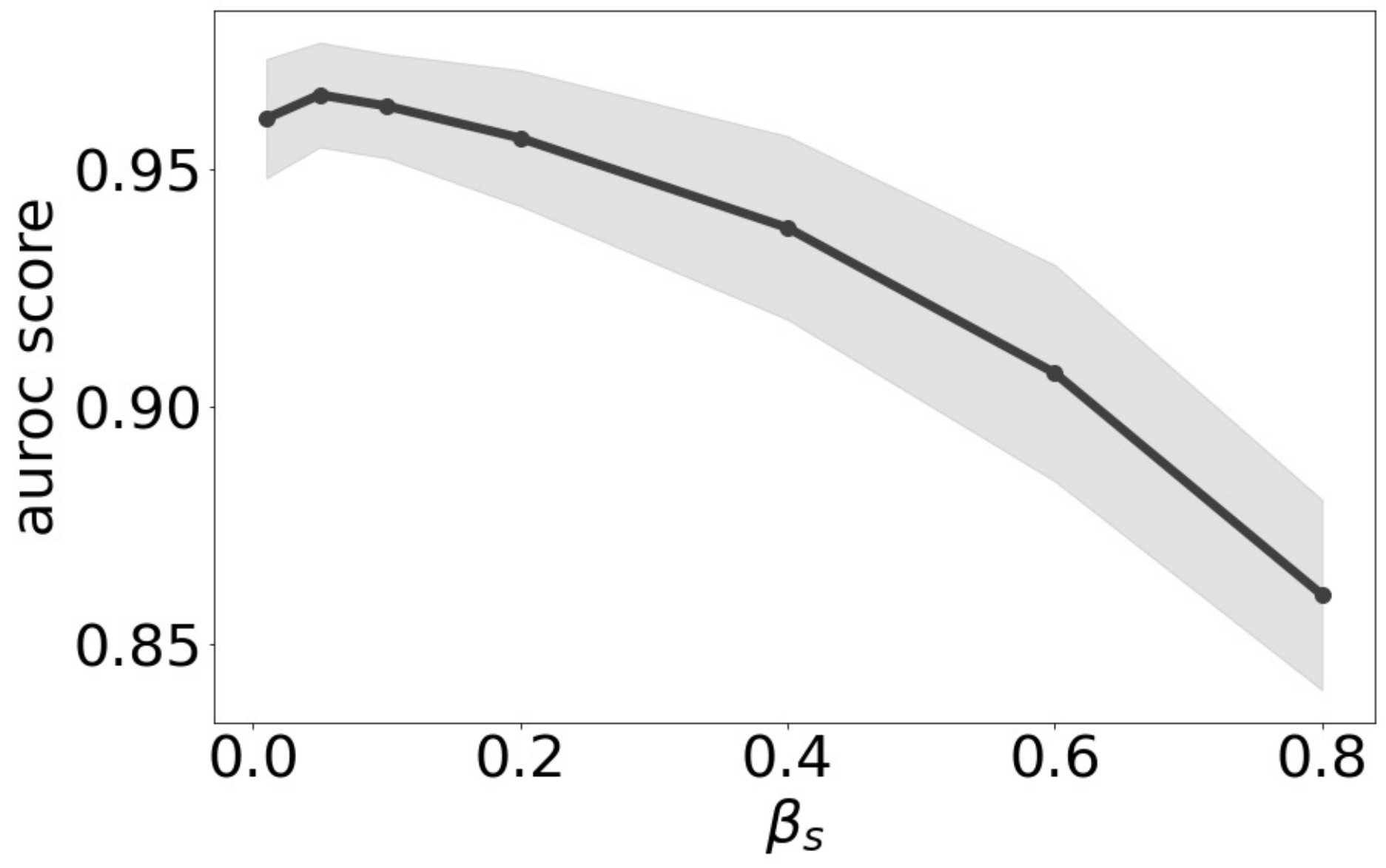}} 
   \\
    \hspace{-0.1in}{\includegraphics[width=0.19\textwidth]{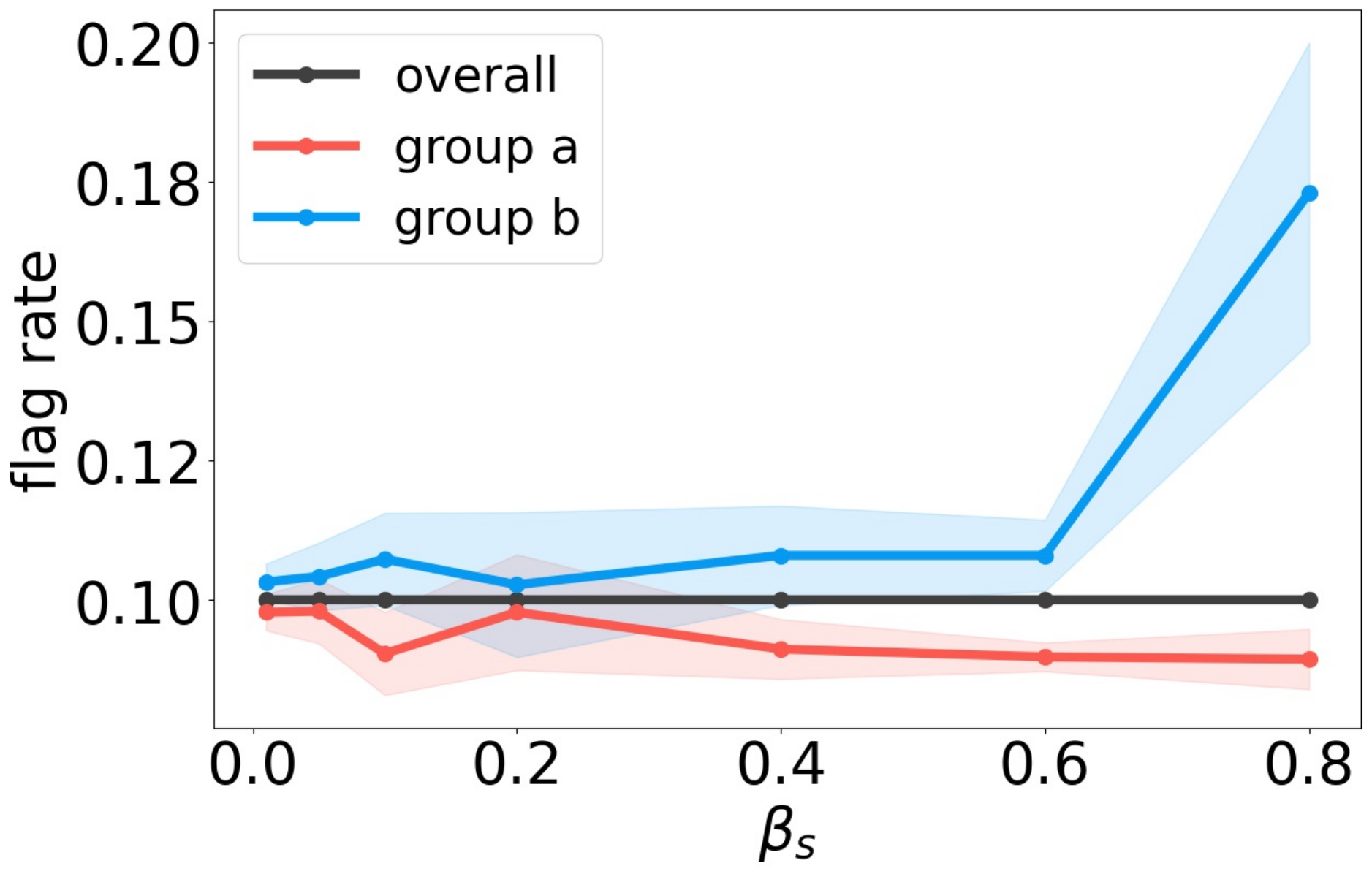}} 
    &
    \hspace{-0.1in}{\includegraphics[width=0.19\textwidth]{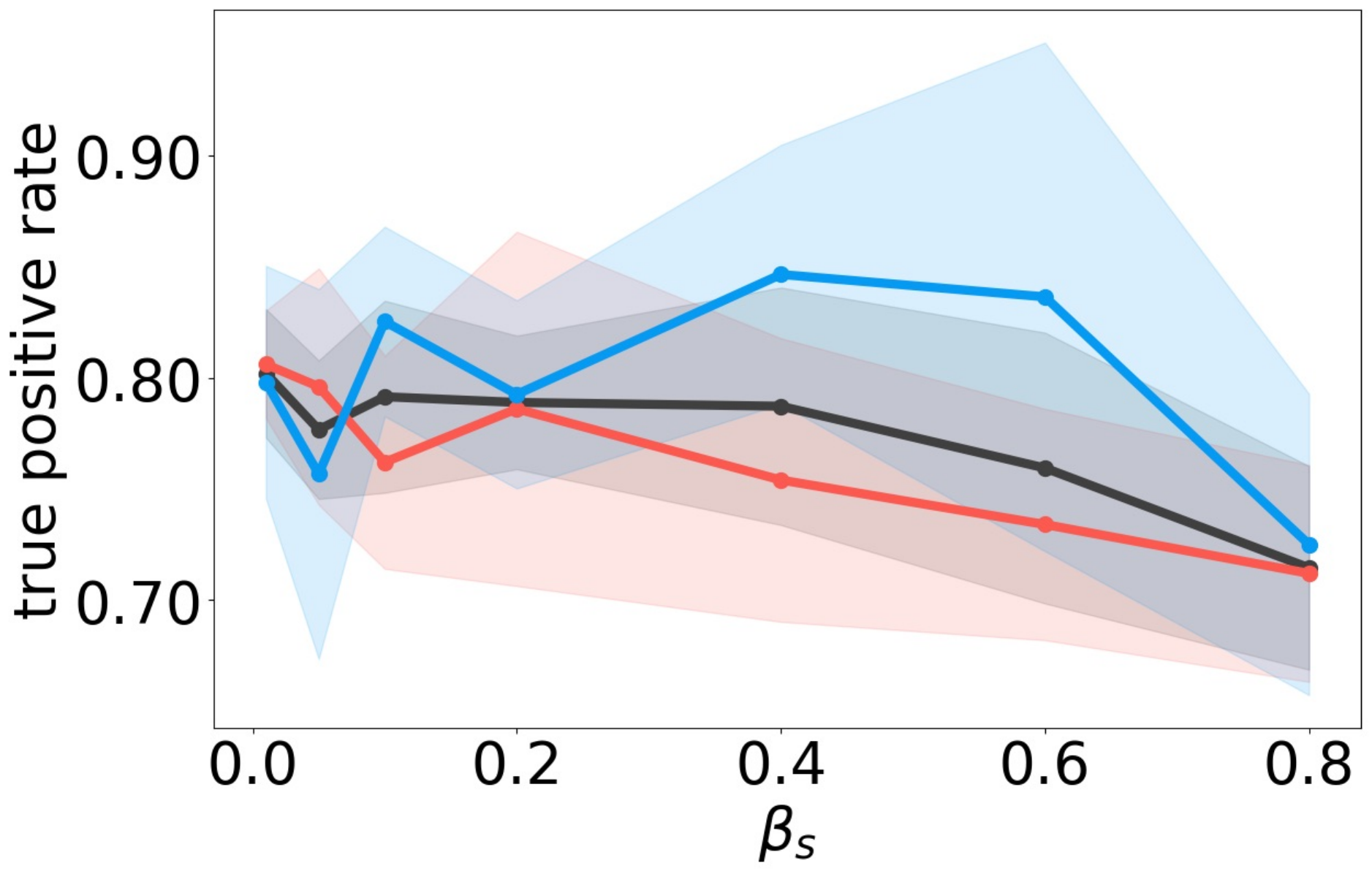}} &
     \hspace{-0.1in}{\includegraphics[width=0.19\textwidth]{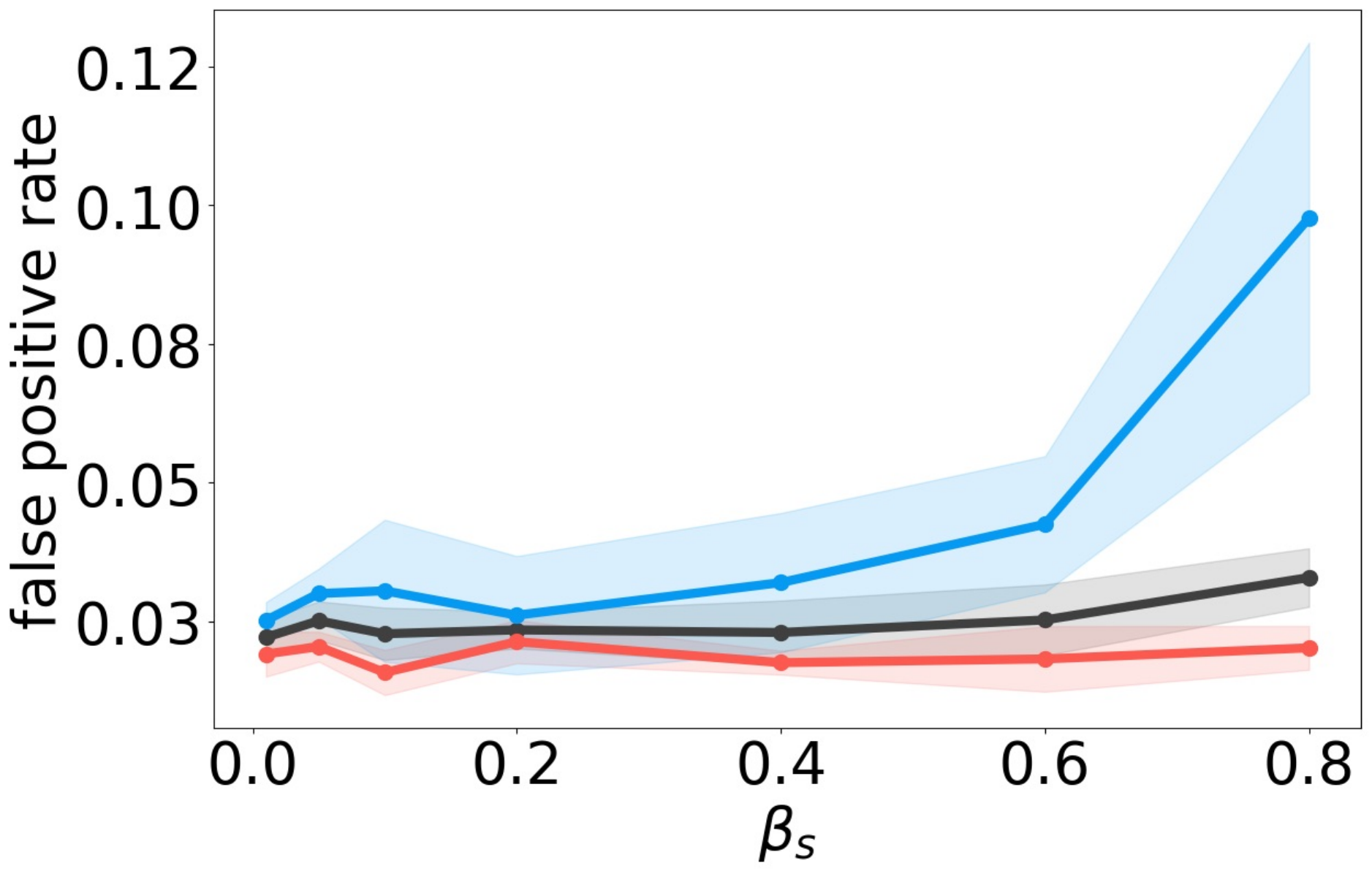}}
    &
   \hspace{-0.1in} {\includegraphics[width=0.19\textwidth]{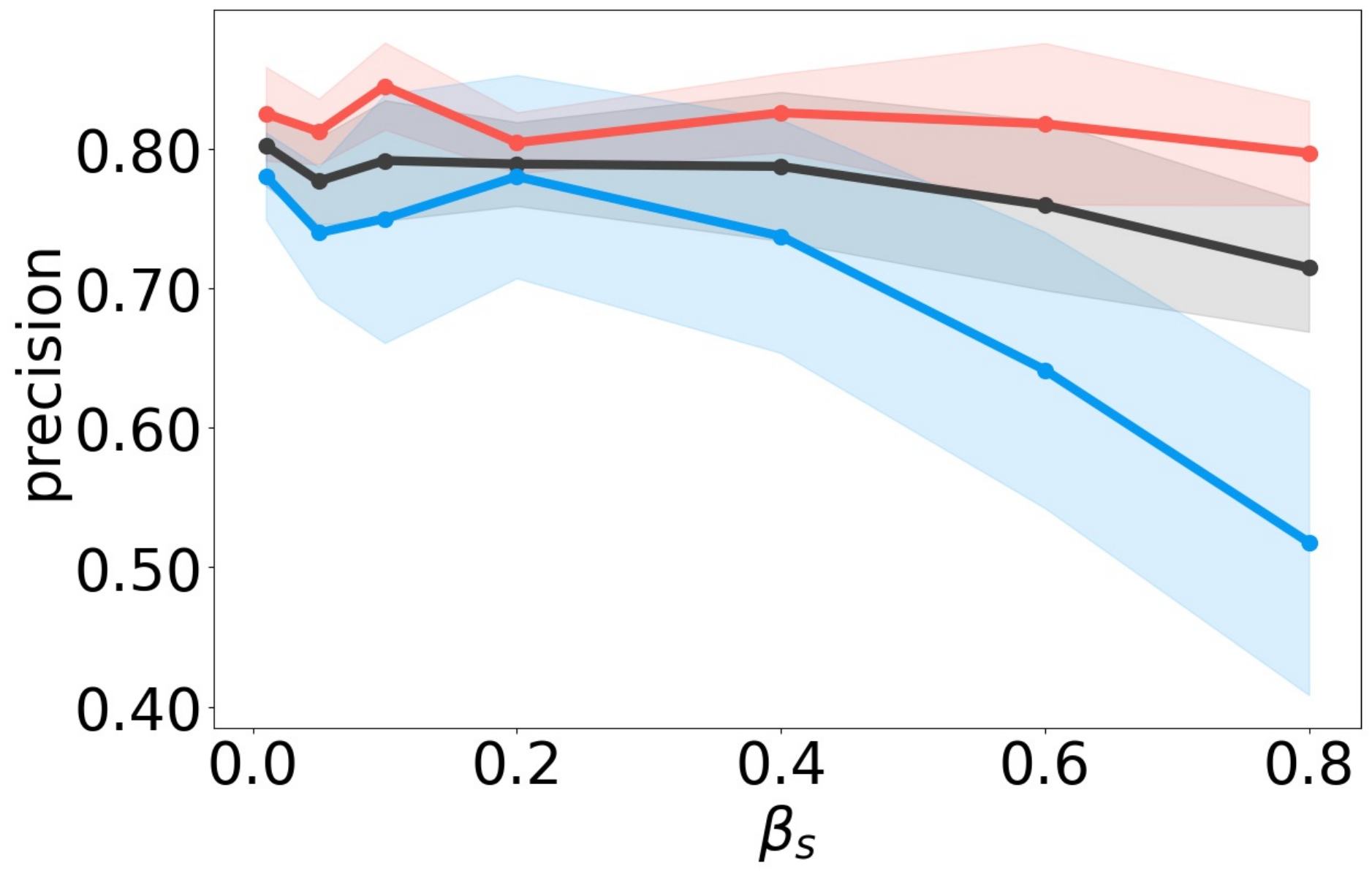}}
   &
     \hspace{-0.1in}{\includegraphics[width=0.19\textwidth]{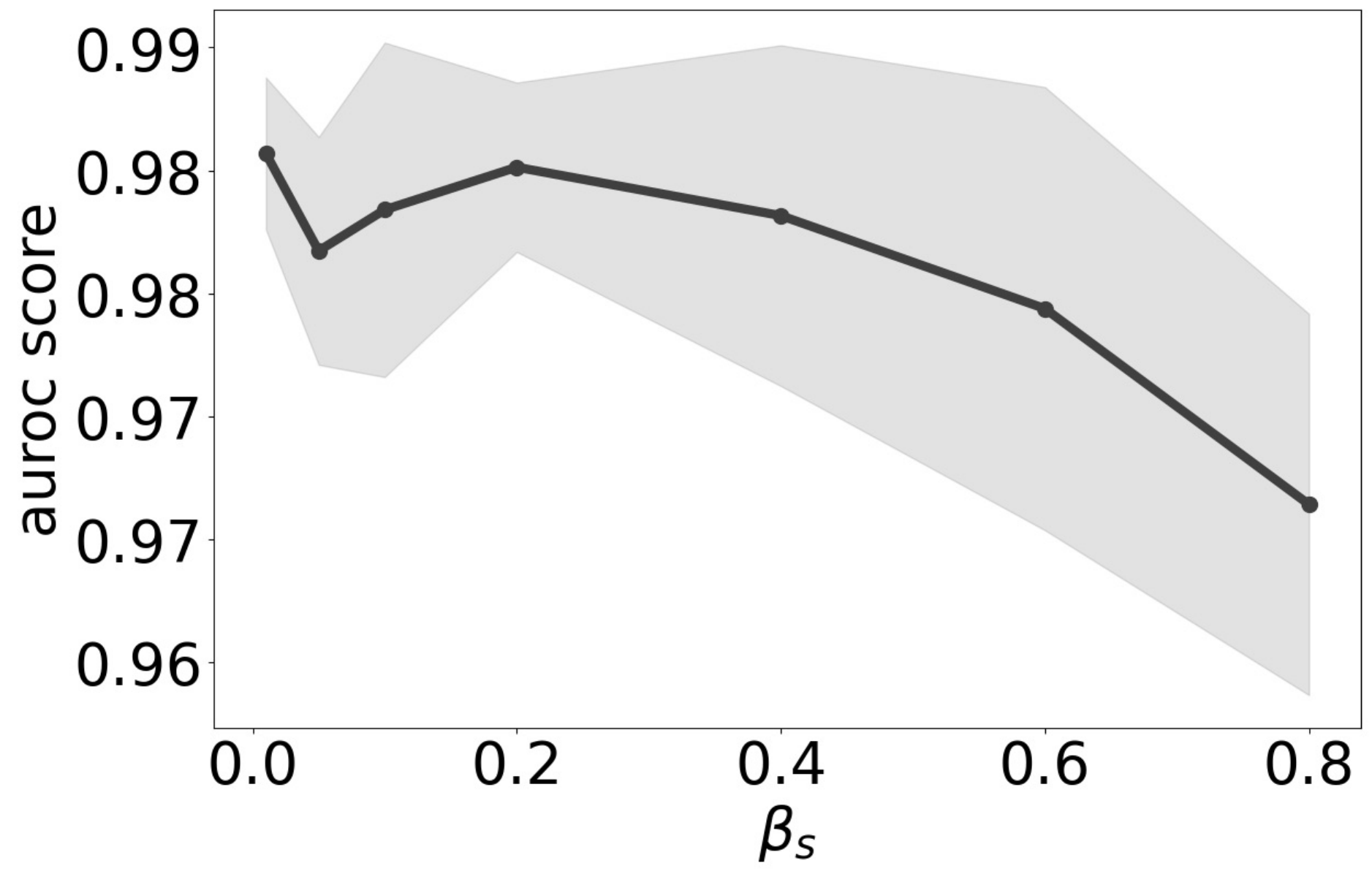}}
   \\
     \hspace{-0.1in}{\includegraphics[width=0.19\textwidth]{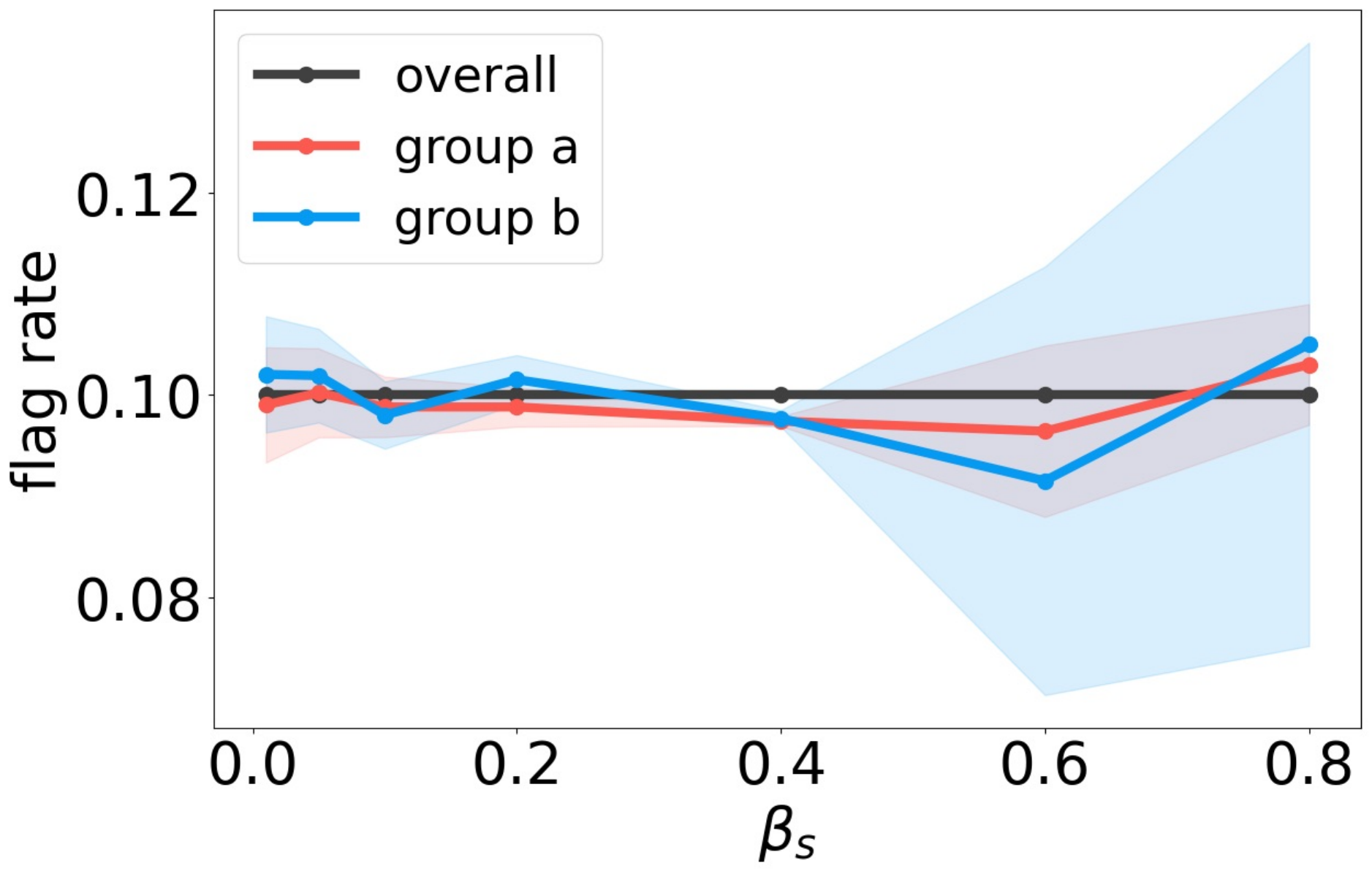}} 
    &
    \hspace{-0.1in}{\includegraphics[width=0.19\textwidth]{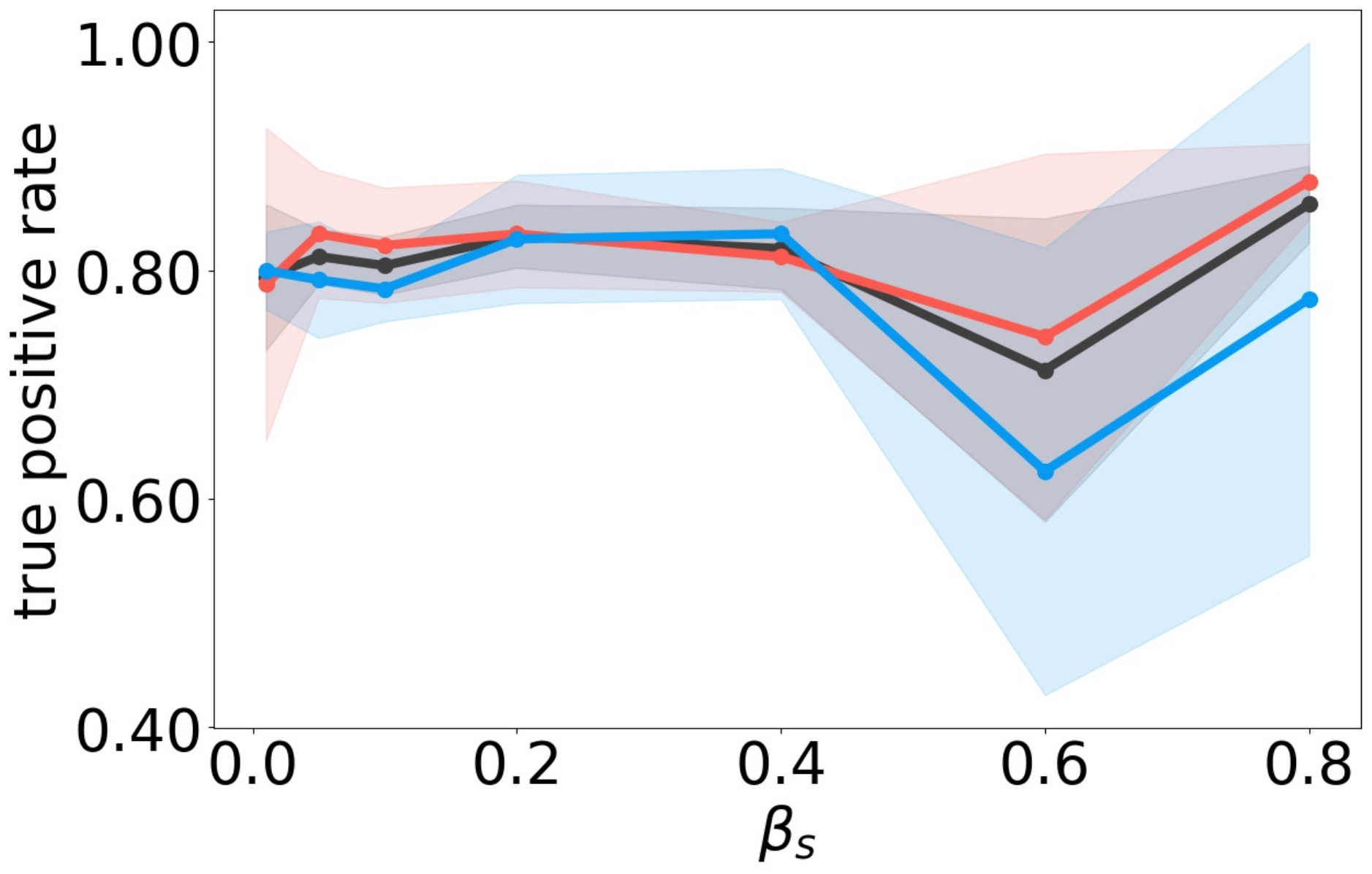}} &
     \hspace{-0.1in}{\includegraphics[width=0.19\textwidth]{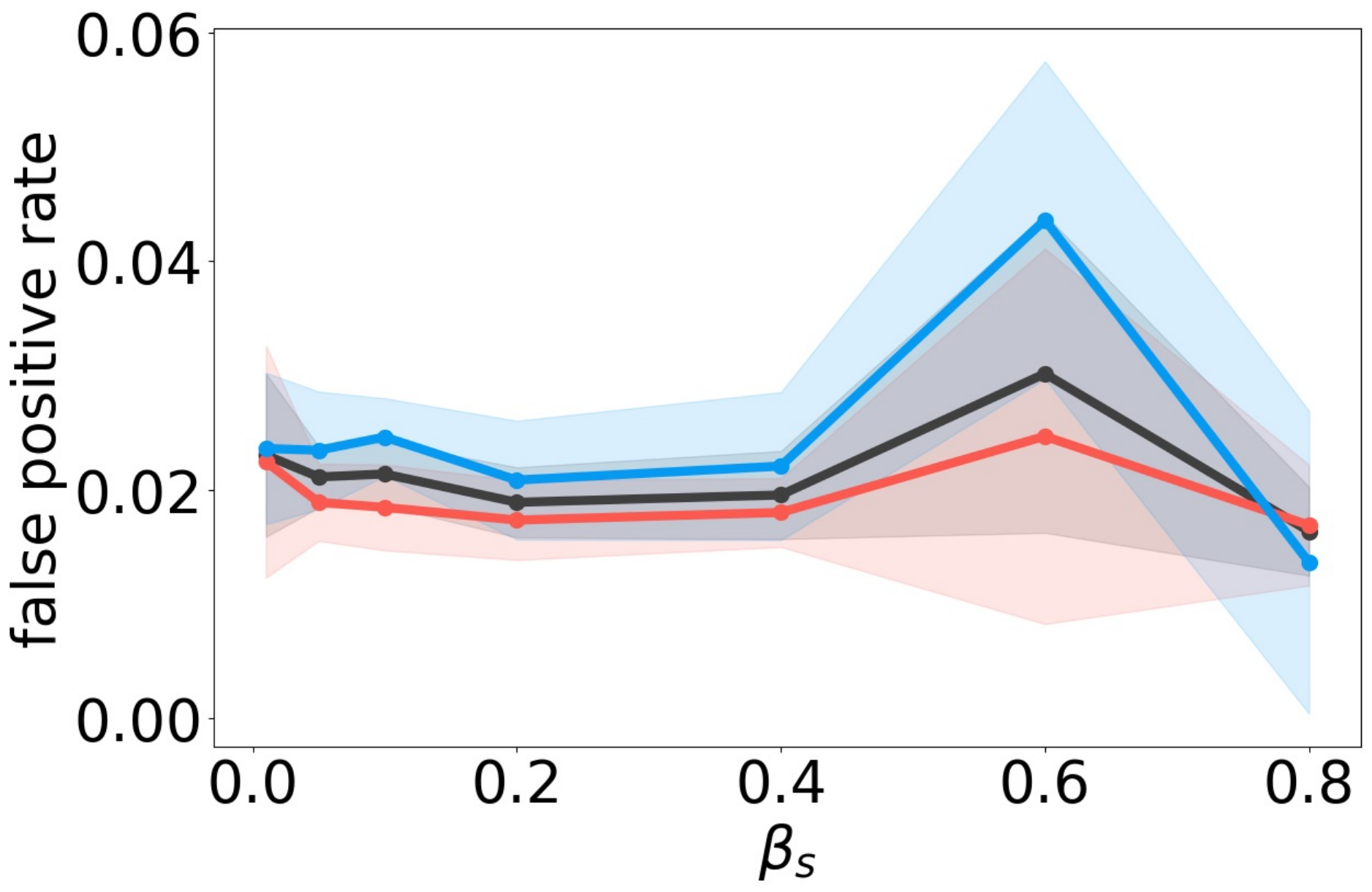}}
    &
   \hspace{-0.1in} {\includegraphics[width=0.19\textwidth]{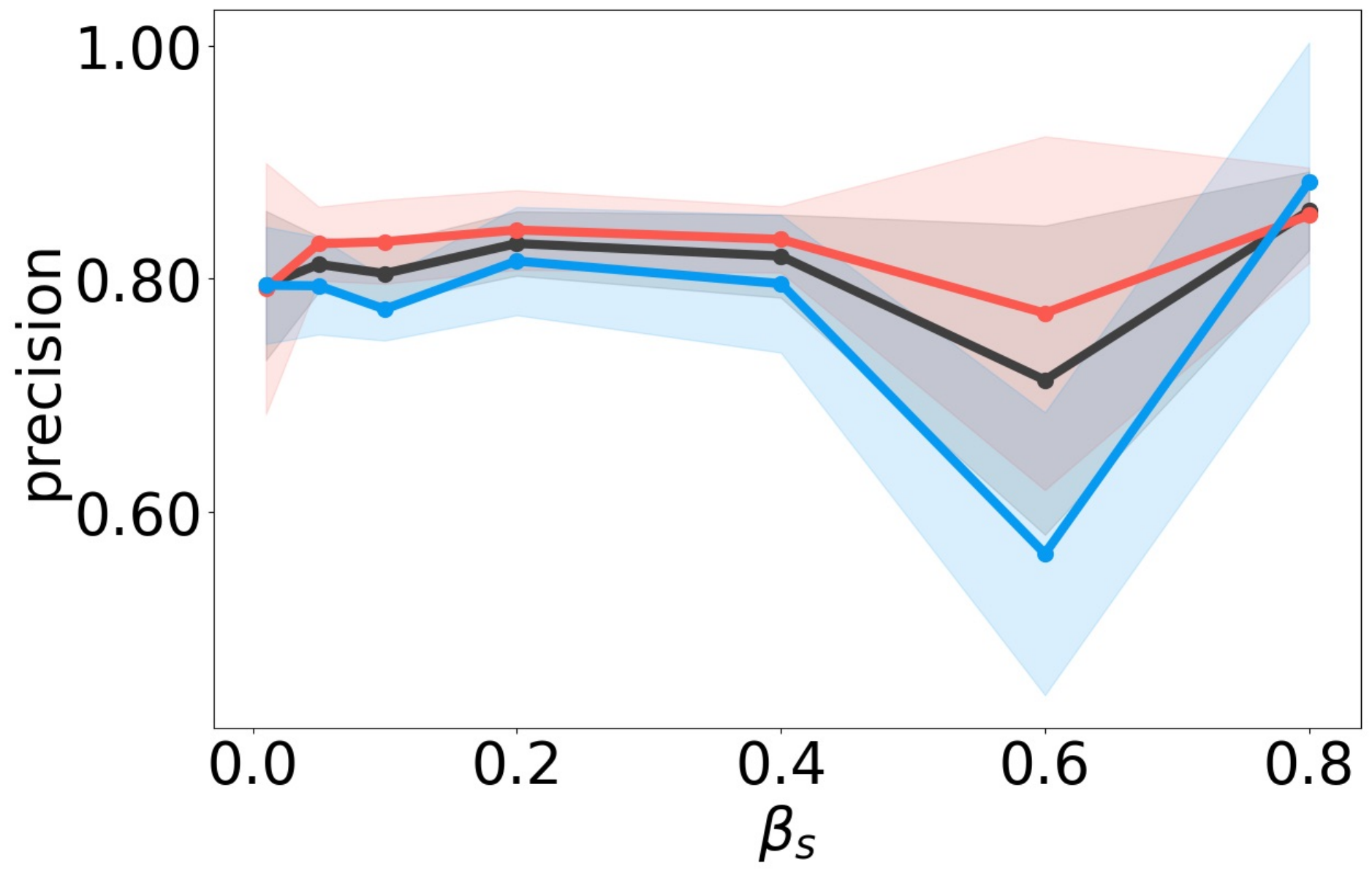}}
    &
   \hspace{-0.1in} {\includegraphics[width=0.19\textwidth]{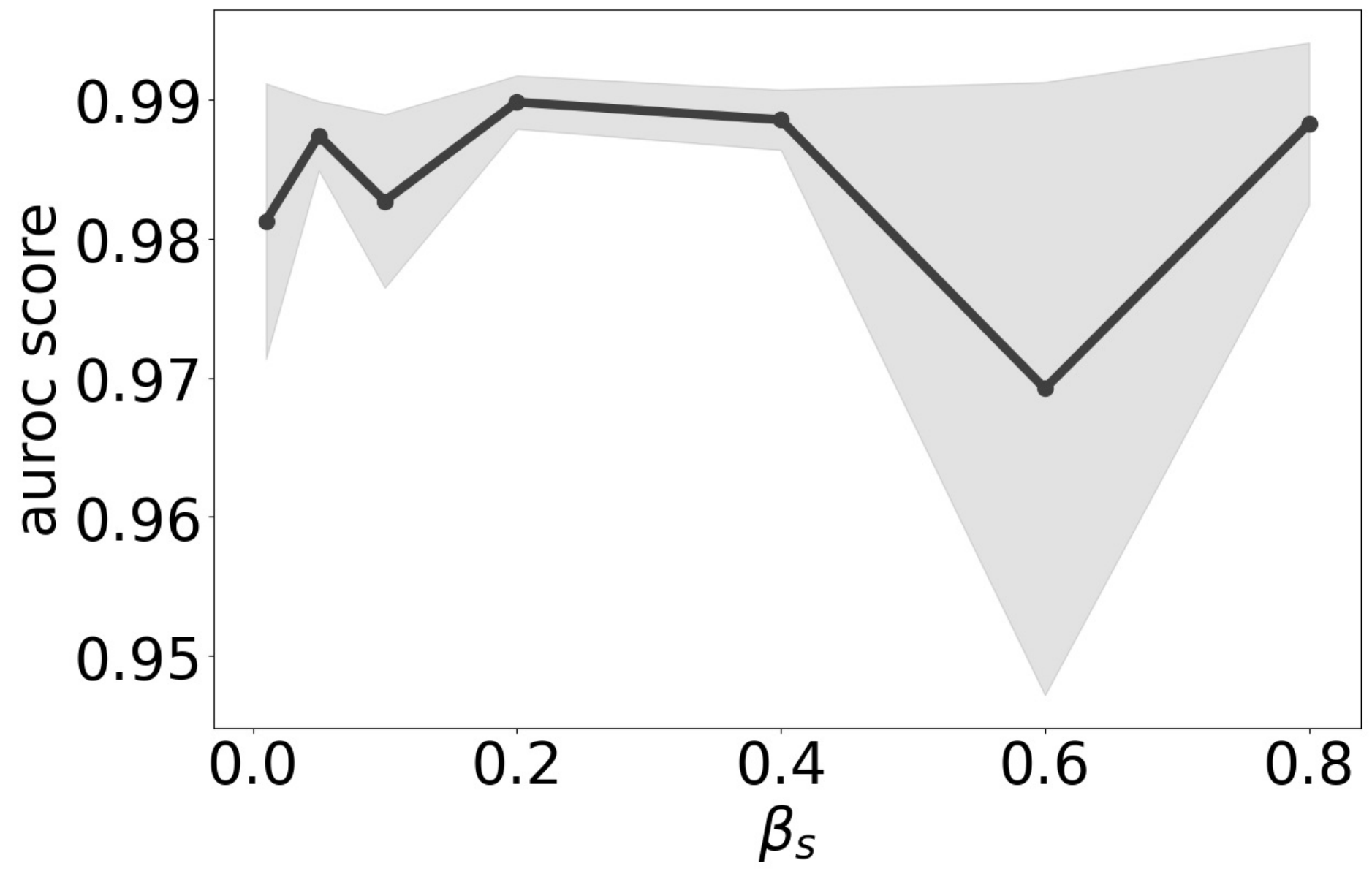}}
    \end{tabular}
    \caption{(best in color) Group-wise fairness metrics and AUROC for (top to bottom) LOF, iForest, DeepAE and FairOD under sample size bias on \textit{clustered} outliers.}
    \label{fig:samplesize_all_cluster}
\end{figure*}

When samples from group $b$ are dropped at random to inject size disparity, the inliers and outliers both sparsify relative to group $a$.
We find that {OD models react to density differences in the feature space  differently}. 
The stark contrast can be explained based on the modeling assumptions of these models, as they interact with density variation (see \S\ref{ssec:models}).

As shown in Fig. \ref{fig:samplesize_all_cluster}, LOF's flag rate for group $b$ \textit{drops}, which in turn disadvantages group $a$ with increased FPR  and decreased Precision (top row). In contrast, iForest behaves  in the \textit{opposite} fashion, disadvantaging group $b$ with significantly higher flag rate, higher FPR and lower Precision (2nd row).
This contrast is due to these models treating density locally or globally.
LOF evaluates outlierness \textit{locally}, by comparing points to their neighbors. When group $b$ inliers sparsify, the clustered $b$-outliers ``hide'' better due to masking \cite{jiang2022d}. Isolation-based iForest, on the other hand, can more quickly isolate now-globally-sparser points in group $b$, overly flagging them, hence maintaining high TPR, but larger FPR and lower Precision.
iForest's brittleness is also evident from its overall performance falling more drastically with increasing bias.

Among the deep models,
DeepAE (3rd row) behaves similar to iForest in terms of group fairness, although through a different mechanism. Since subsampling group $b$ renders them rare \textit{minority} samples, their impact on the total loss of the compression-based DeepAE diminishes. As a result, their poor reconstruction reflects as higher flag rate, larger FPR and lower Precision.  As these group-wise differences are less extreme than for iForest, DeepAE overall detection performance remains relatively stable.
Finally, we find FairOD (bottom row) to be robust against size disparity, where it achieves statistical as well as TPR parity through explicit optimization. The other quantities (FPR and Precision) follow suit, with negligible impact on performance.

These findings on clustered outliers continue to hold on scattered outliers as given in Apdx. \S\ref{assec:sizebias_scatter} Fig. \ref{fig:samplesize_all_scatter}.

\vspace{-0.075in}
\subsection{Target under-representation bias}
\label{ssec:urep}

The algorithm behaviors change considerably when exposed to the under-representation bias, where we  drop only the target (positive) samples from group $b$.
Results are given for all models in Apdx. \S\ref{assec:urep_cluster} Fig. \ref{fig:underrep_all_cluster} and Apdx. \S\ref{assec:urep_scatter} Fig. \ref{fig:underrep_all_scatter} in the clustered and scattered outliers settings, respectively.

Note that dropping outlier samples from $b$ renders group $a$'s observed base rate higher.
With a higher base rate, group $a$'s clustered outliers are slightly masked for LOF, which reduces group $a$'s flag rate and TPR, while increasing those for $b$. FPR increases for both groups, proportionate to the overall increase. The masking effect goes away when outliers are scattered, with no notable TPR and FPR difference between groups. 
However, Precision is specifically lower for group $b$ in both cases, mainly because $b$'s flag rate is larger than its base rate.
iForest, on the other hand, is robust to masking as it can find cuts in the feature space that isolate the outlier clusters at once, with higher TPR but lower Precision for group $b$.

In fact, lower Precision (due to higher flag rate than base rate) for group $b$ is a common trend across all models, including the compression-based DeepAE and FairOD.  
Group $b$ outliers continue to stand out for both models as dropping few outlier samples does not change the bulk of the data and hence the compression quality of group $b$'s inliers.
However, because group $a$'s base rate and hence frequency is higher, its flag rate is relatively lower. This translates to a higher flag rate for $b$ at the cost of lower Precision. 
These results showcase the vulnerability of all OD models in our study in the face of unequal base rates.

We see that the overall detector performances remain quite robust against this type of data bias, in both settings with  clustered and scattered outliers, likely because the overall data distribution is not impacted much, especially compared to the other types of data biases we studied.

\vspace{-0.075in}
\subsection{Feature measurement bias}
\label{ssec:meas}

\subsubsection{\bf Variance shift for measurement noise.~}
In the clustered setting (see Apdx. \S\ref{assec:meas_cluster} Fig. \ref{fig:varianceshift_all_cluster}), 
feature variance loosens the clusters among both inliers and outliers in group $b$. As a result, LOF ranks group $a$ outliers strictly above those from group $b$ as they showcase a starker contrast to their relative inliers, while still retaining the ability to correctly flag almost all true outliers. 
iForest is similarly robust with only slightly higher FPR and slightly lower TPR and Precision for group $b$ than for $a$, while retaining high overall performance.  
Moreover, both DeepAE and FairOD maintain high performance as the smaller outlier clusters do not compress as well as the inliers despite higher variance. As such, deep models are similarly robust with no notable disparity between groups.

\begin{figure*}[!ht]
    \centering
    \begin{tabular}{lllll}
     \hspace{-0.1in}{\includegraphics[width=0.19\textwidth]{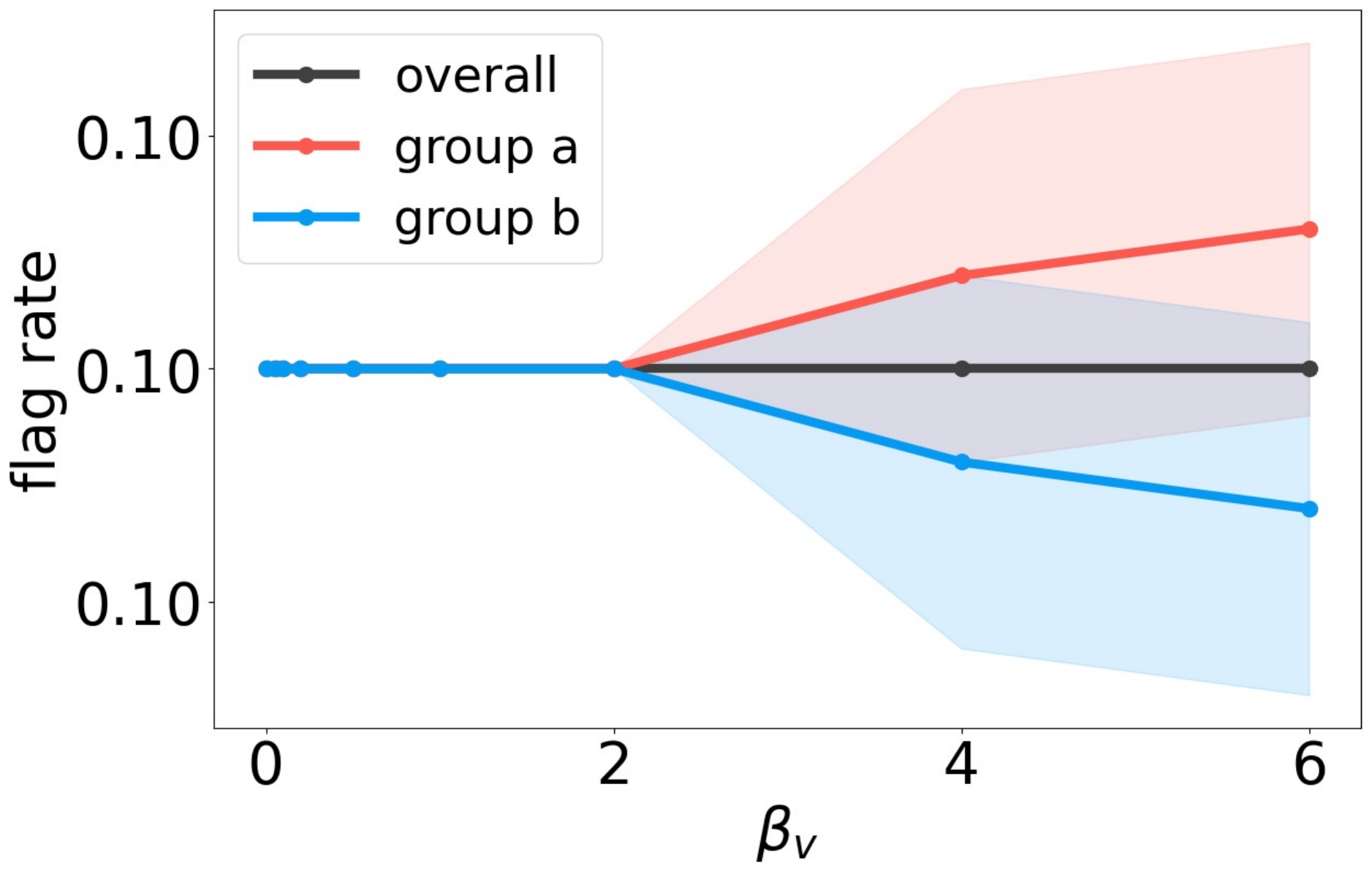}} 
    &
    \hspace{-0.1in}{\includegraphics[width=0.19\textwidth]{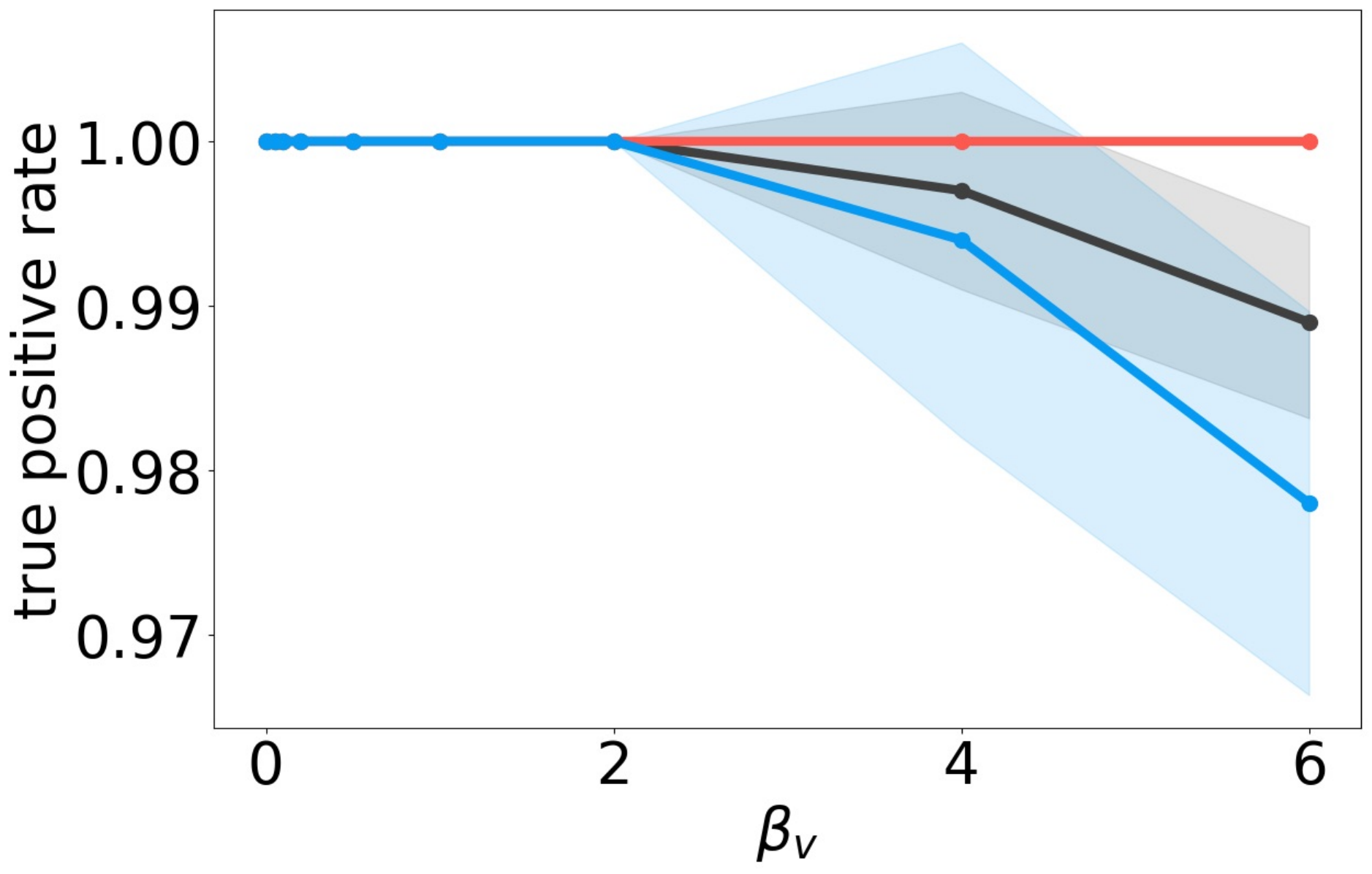}} &
     \hspace{-0.1in}{\includegraphics[width=0.19\textwidth]{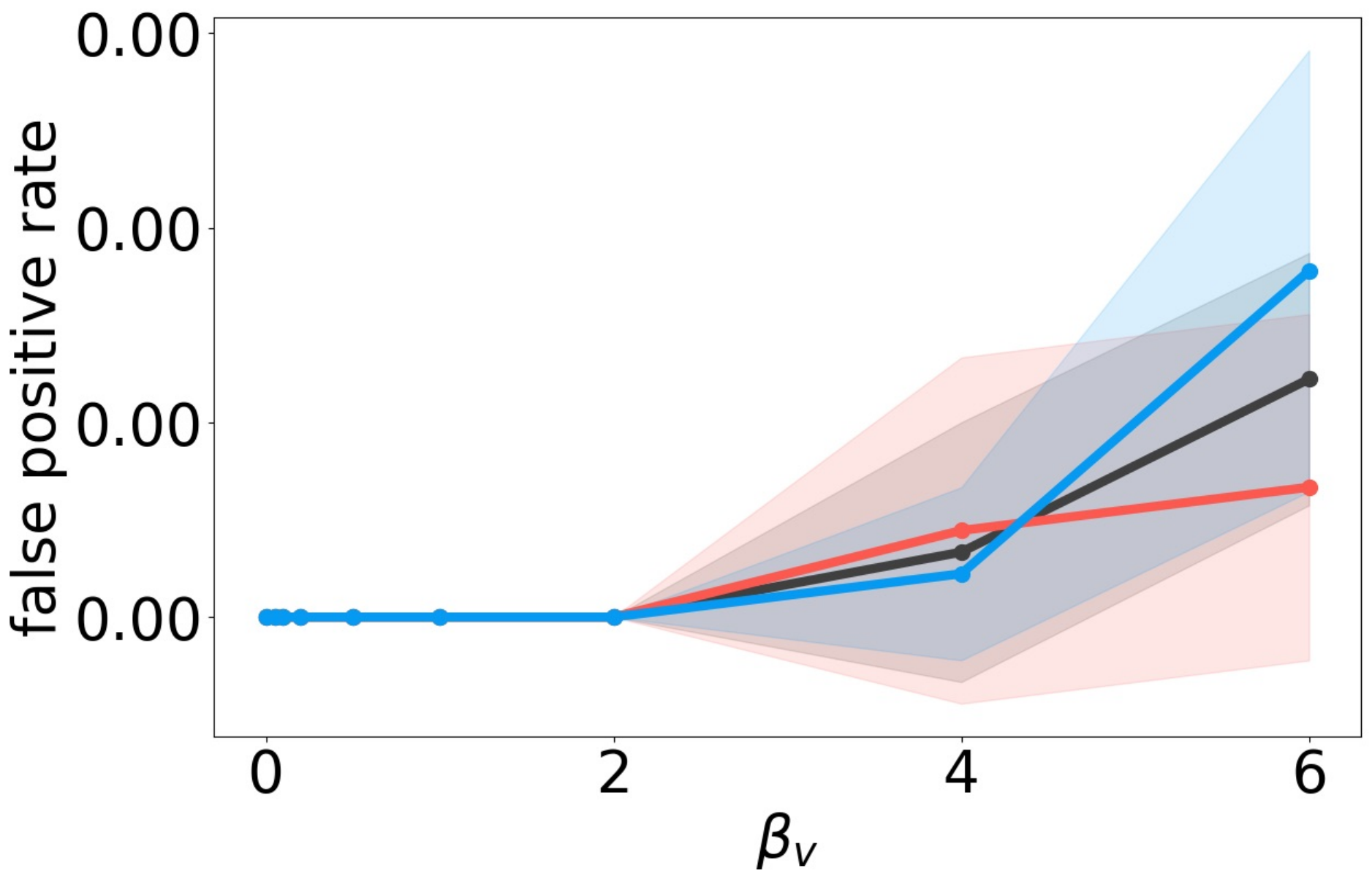}}
    &
   \hspace{-0.1in} {\includegraphics[width=0.19\textwidth]{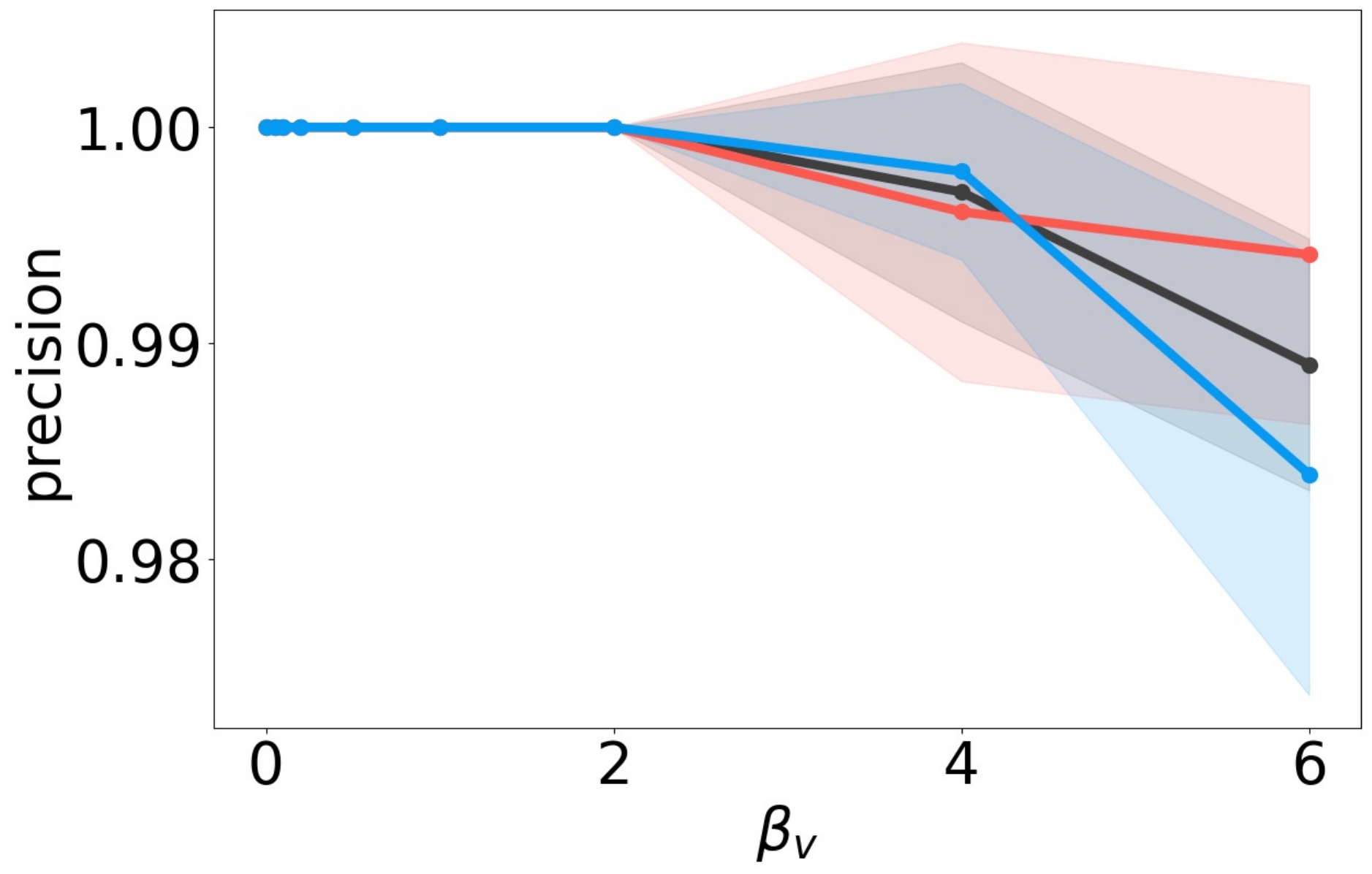}}
   &
     \hspace{-0.1in}{\includegraphics[width=0.19\textwidth]{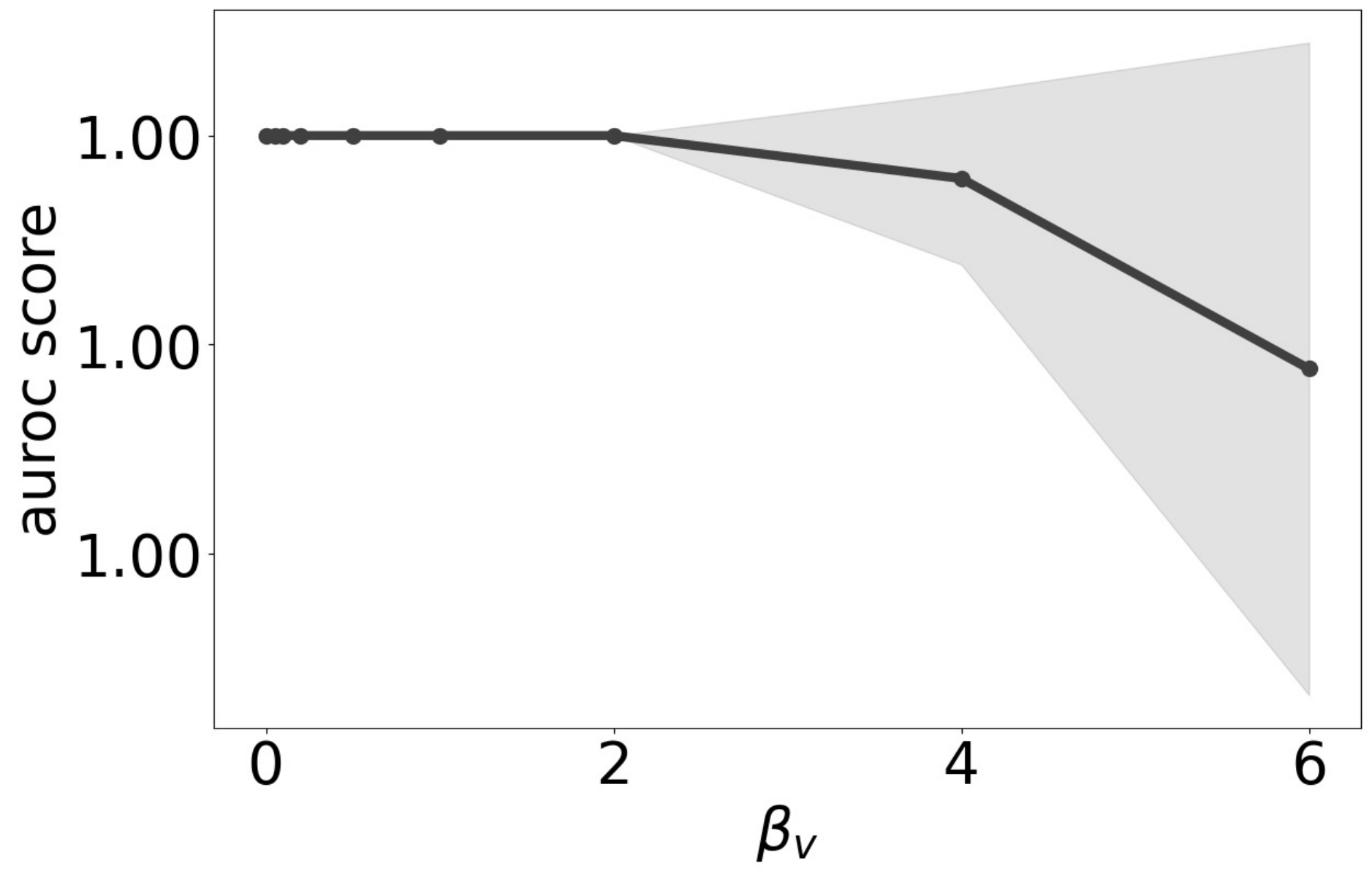}}
   \\
       \hspace{-0.1in}{\includegraphics[width=0.19\textwidth]{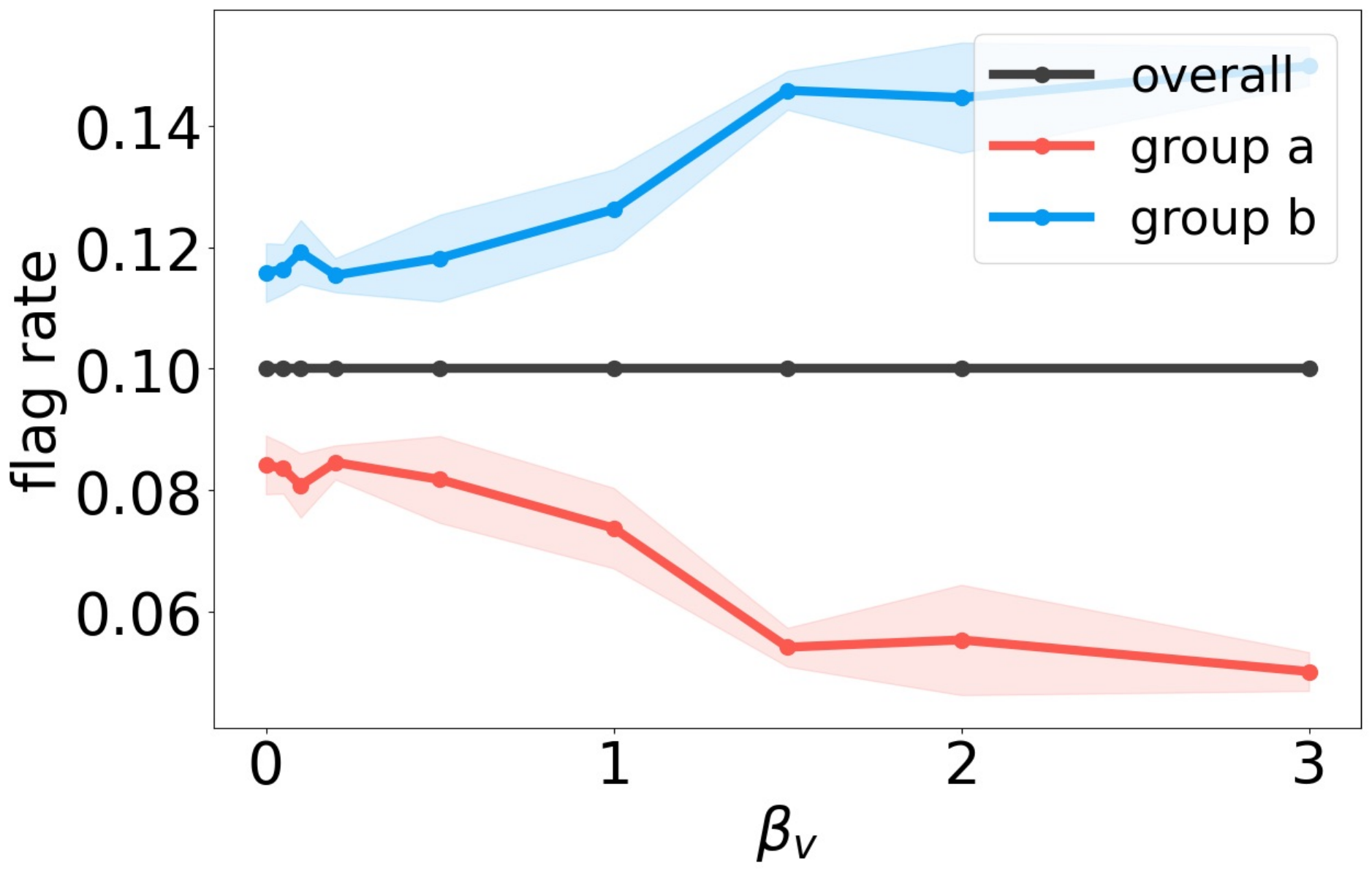}} 
    &
    \hspace{-0.1in}{\includegraphics[width=0.19\textwidth]{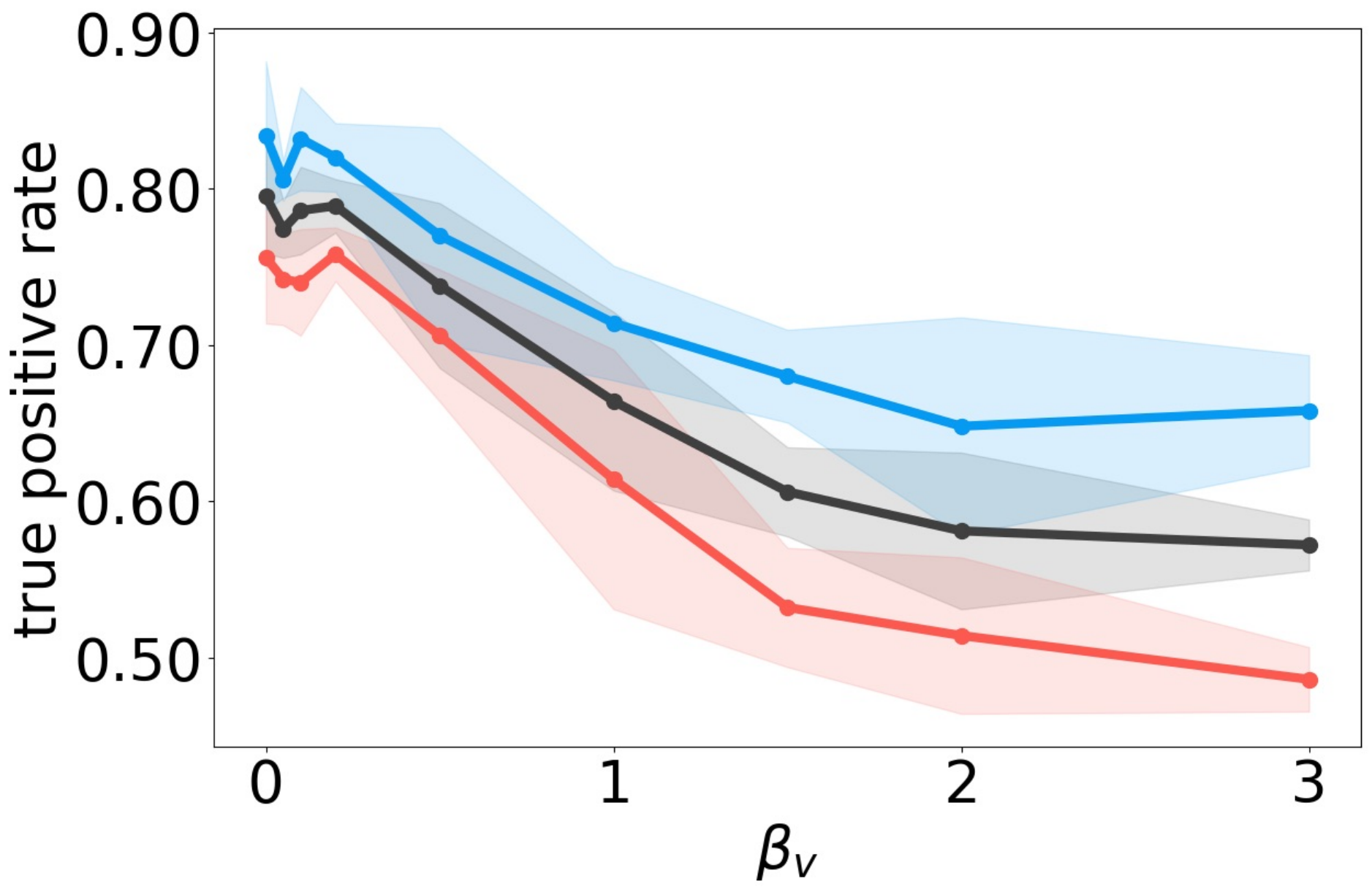}} &
     \hspace{-0.1in}{\includegraphics[width=0.19\textwidth]{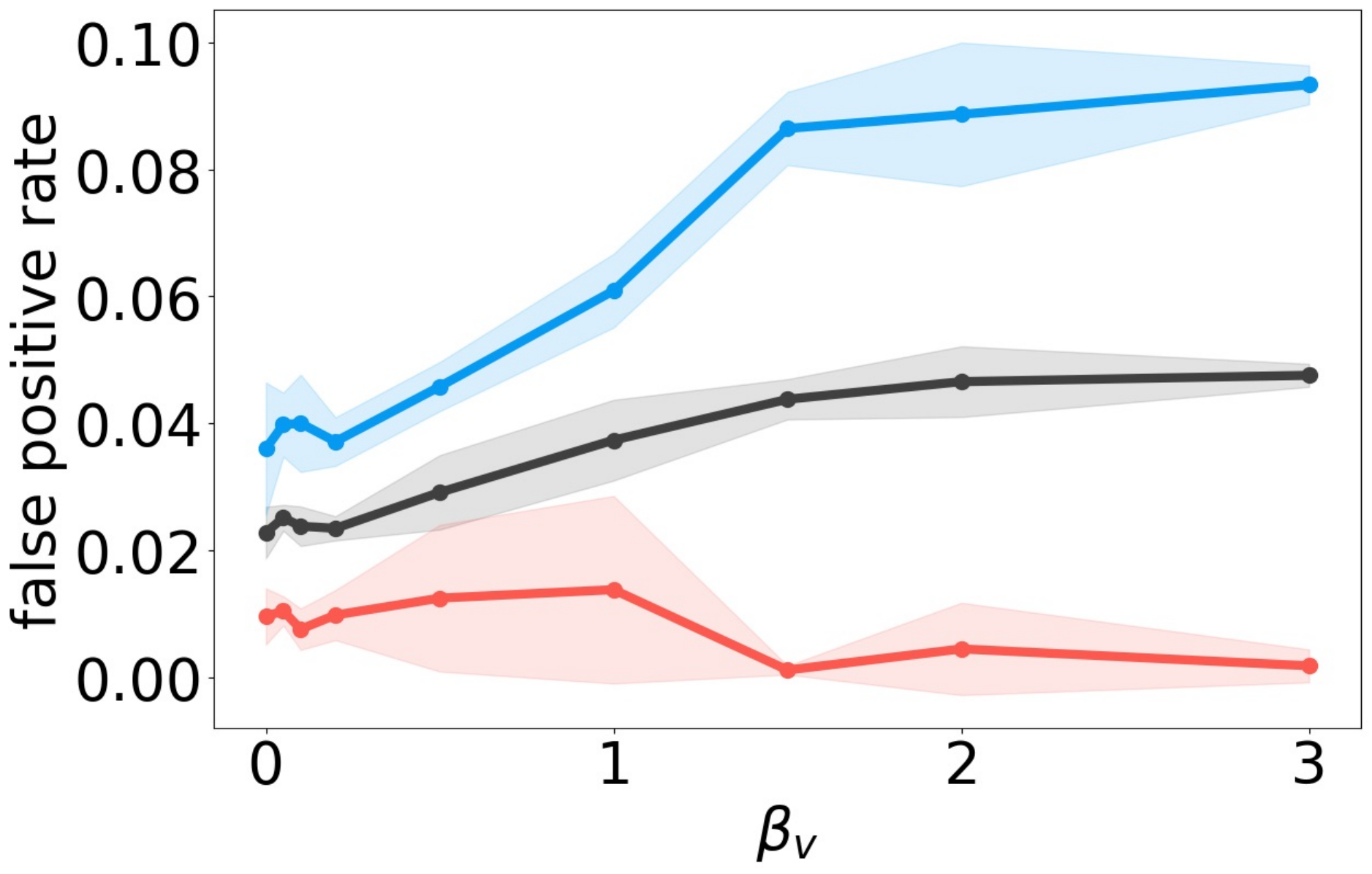}}
    &
   \hspace{-0.1in} {\includegraphics[width=0.19\textwidth]{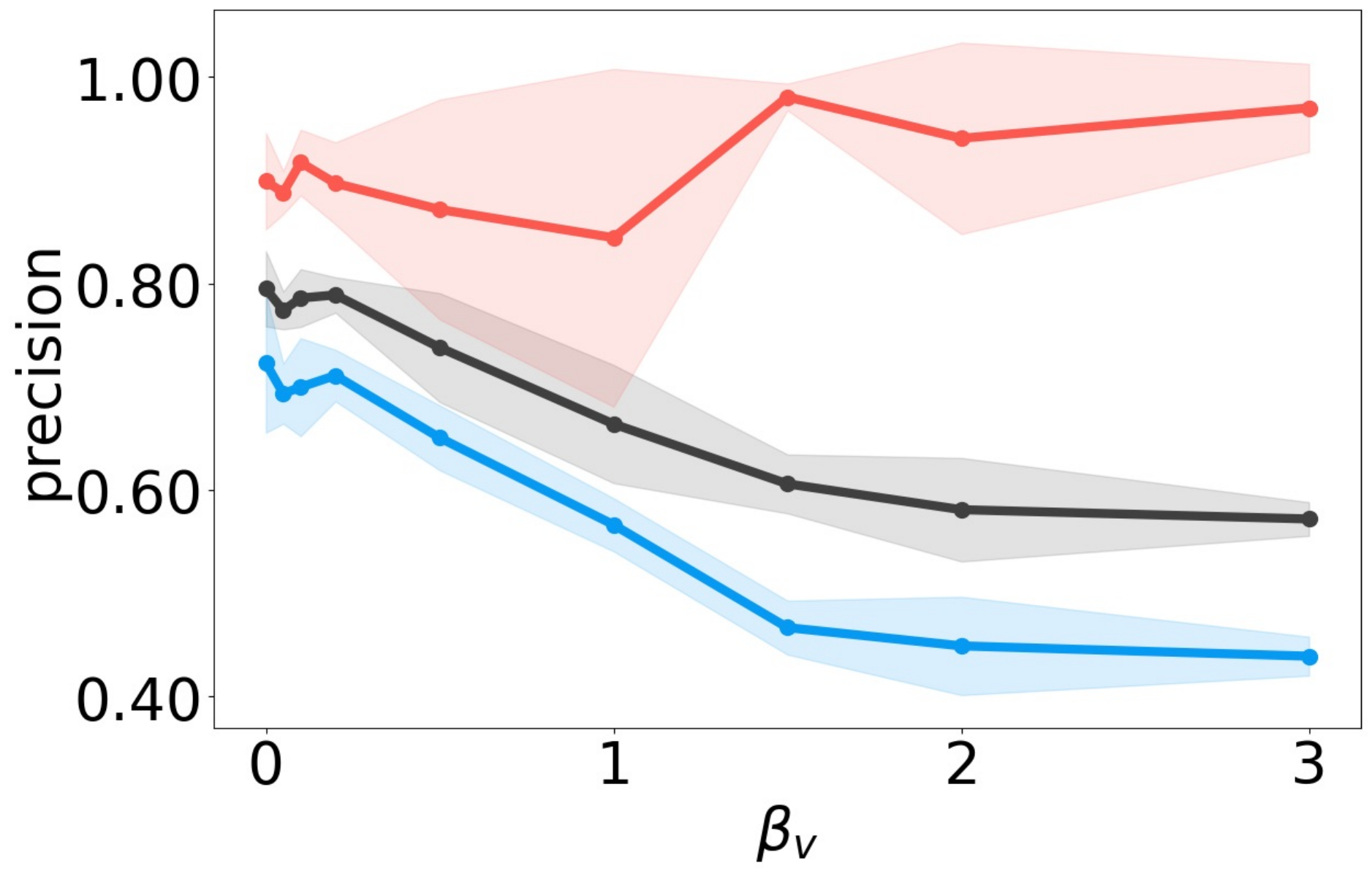}}
   &
     \hspace{-0.1in}{\includegraphics[width=0.19\textwidth]{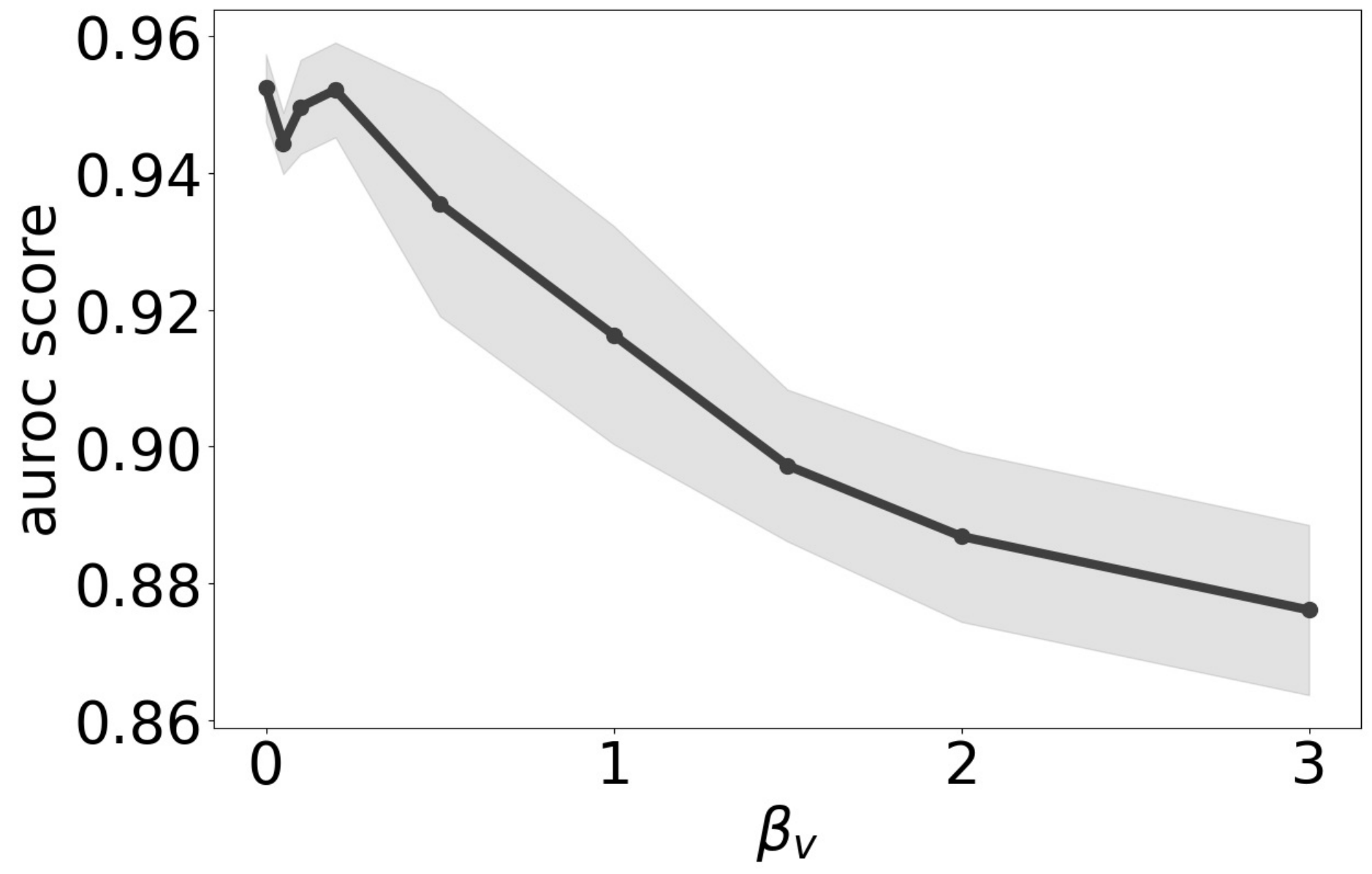}}
    \end{tabular}

    \caption{(best in color) Group-wise fairness metrics and AUROC for DeepAE under feature measurement bias (variance shift) on (top) \textit{clustered} and (bottom) \textit{scattered} outliers. }
    \label{fig:varianceshift_deepae_sidebyside}
\end{figure*}

While we find all detectors to be quite robust against feature variance disparity on clustered outliers, the results differ considerably when outliers are scattered (see Apdx. \S\ref{assec:meas_scatter} Fig. \ref{fig:varianceshift_all_scatter}). 
We show DeepAE as a representative case in 
Fig. \ref{fig:varianceshift_deepae_sidebyside} on clustered versus scattered outliers for comparison.

\begin{figure*}[!ht]
    \centering
    \begin{tabular}{lllll}
    \hspace{-0.1in}{\includegraphics[width=0.188\textwidth]{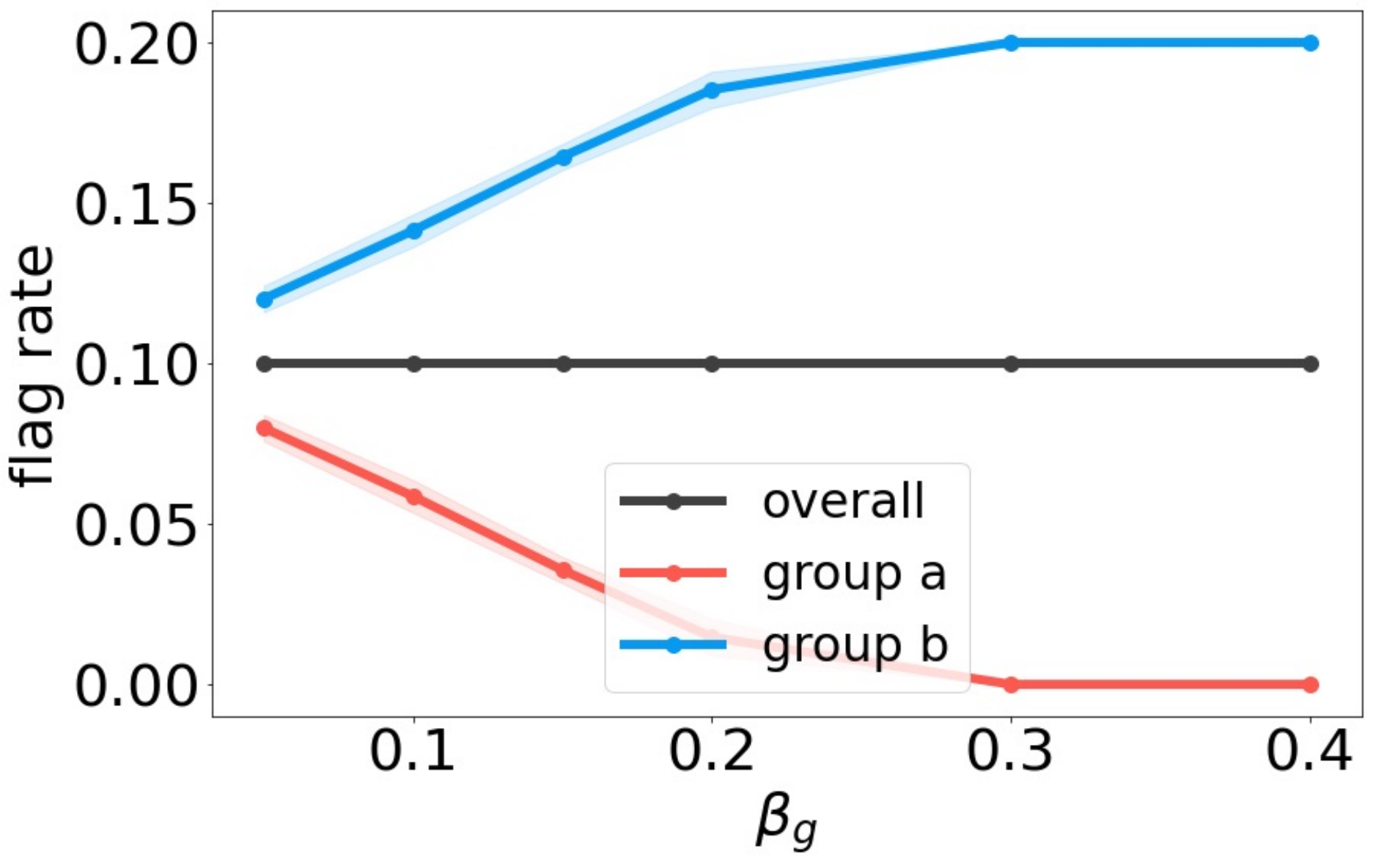}} 
    &
    \hspace{-0.1in}{\includegraphics[width=0.188\textwidth]{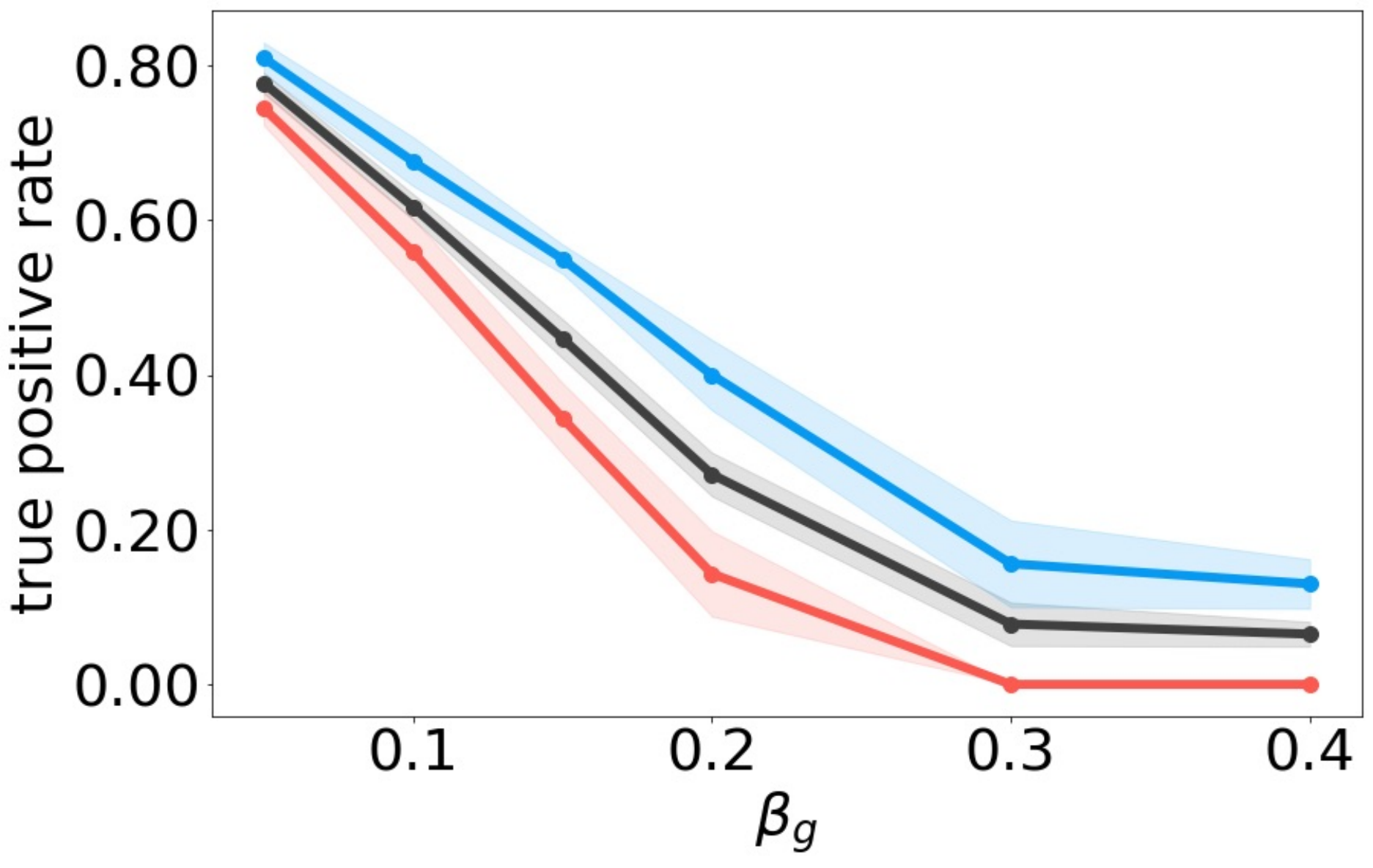}} &
     \hspace{-0.1in}{\includegraphics[width=0.188\textwidth]{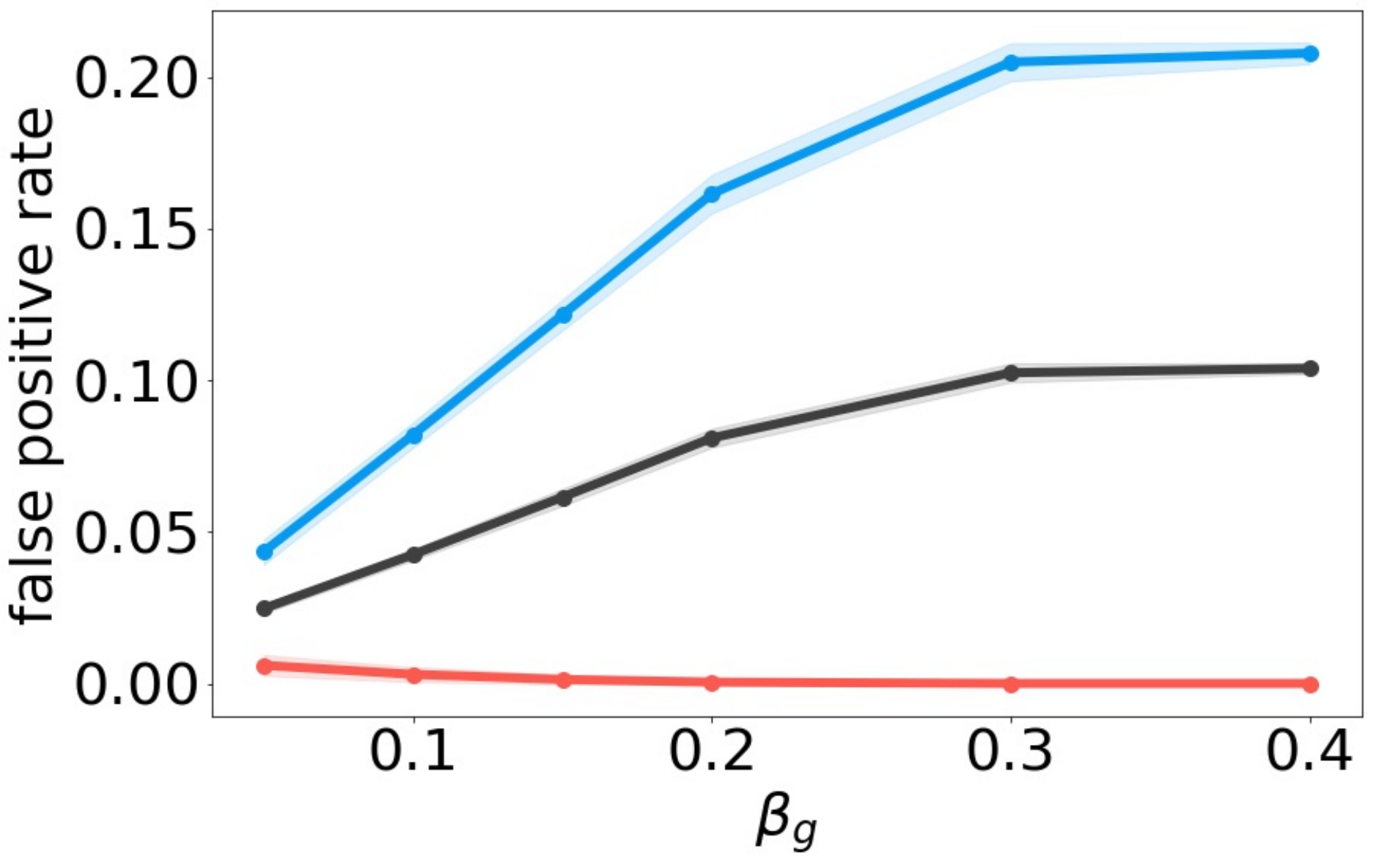}}
    &
   \hspace{-0.1in} {\includegraphics[width=0.188\textwidth]{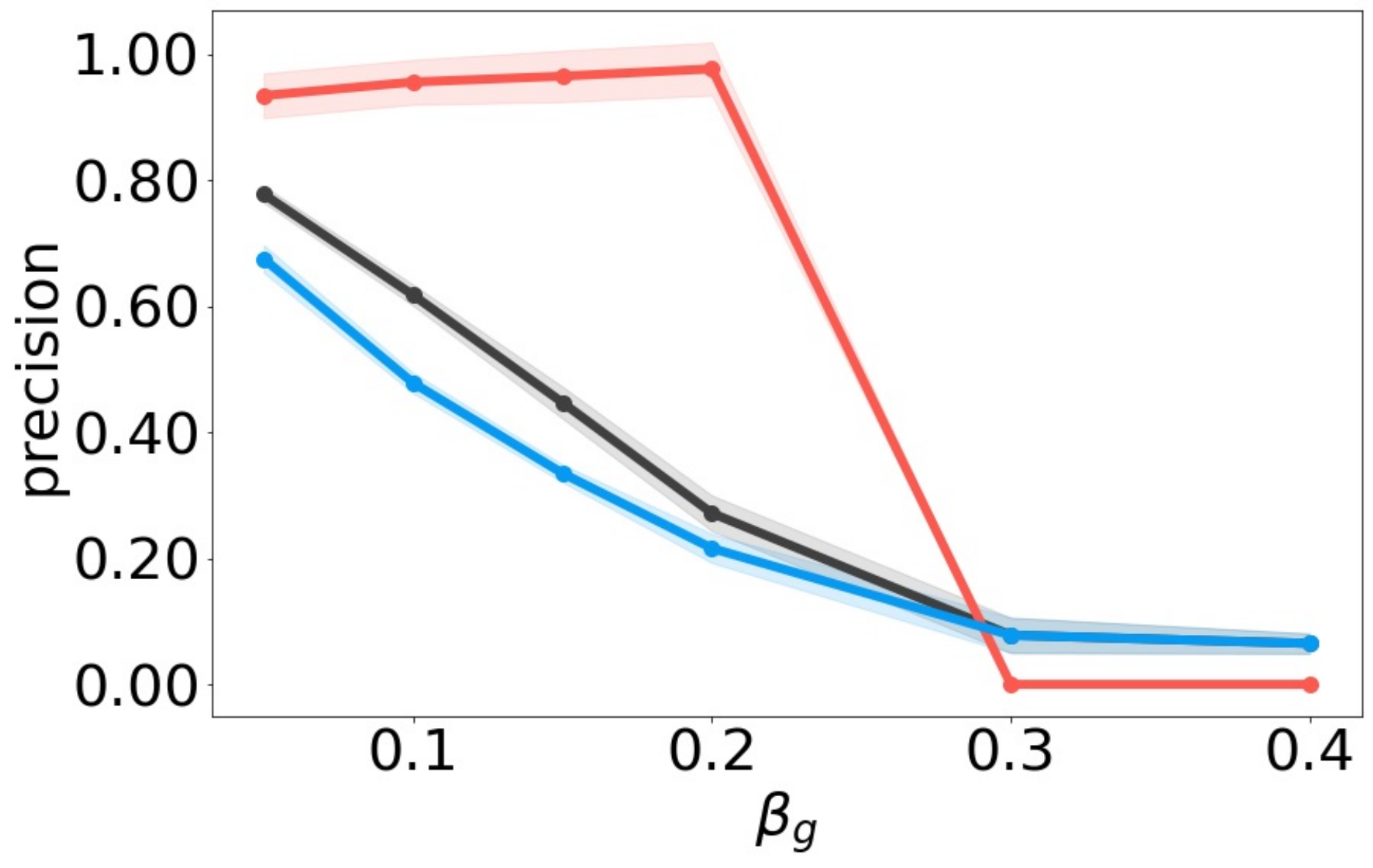}}
   &
   \hspace{-0.1in}
   {\includegraphics[width=0.188\textwidth]{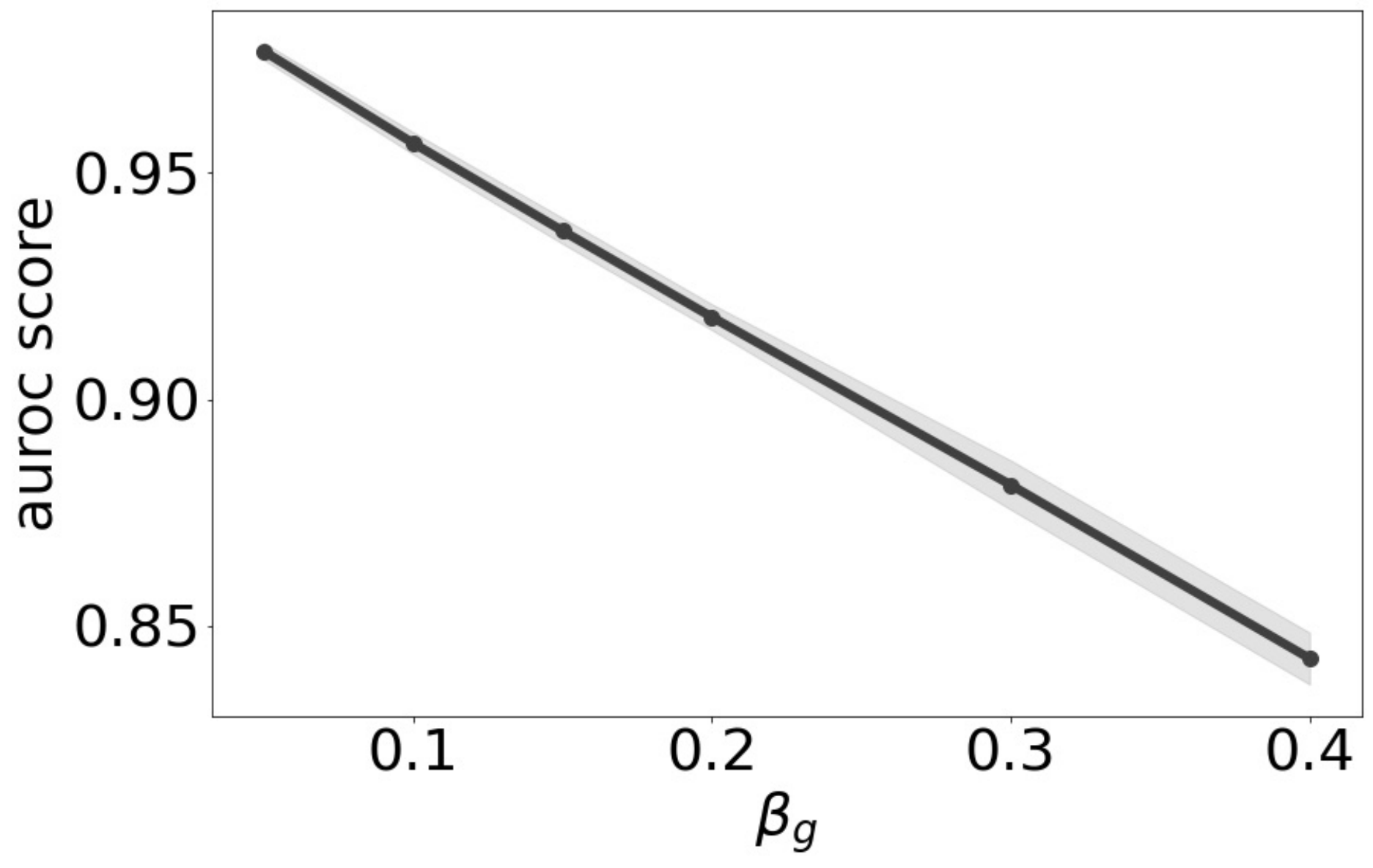}} 
\\
   \hspace{-0.1in}{\includegraphics[width=0.188\textwidth]{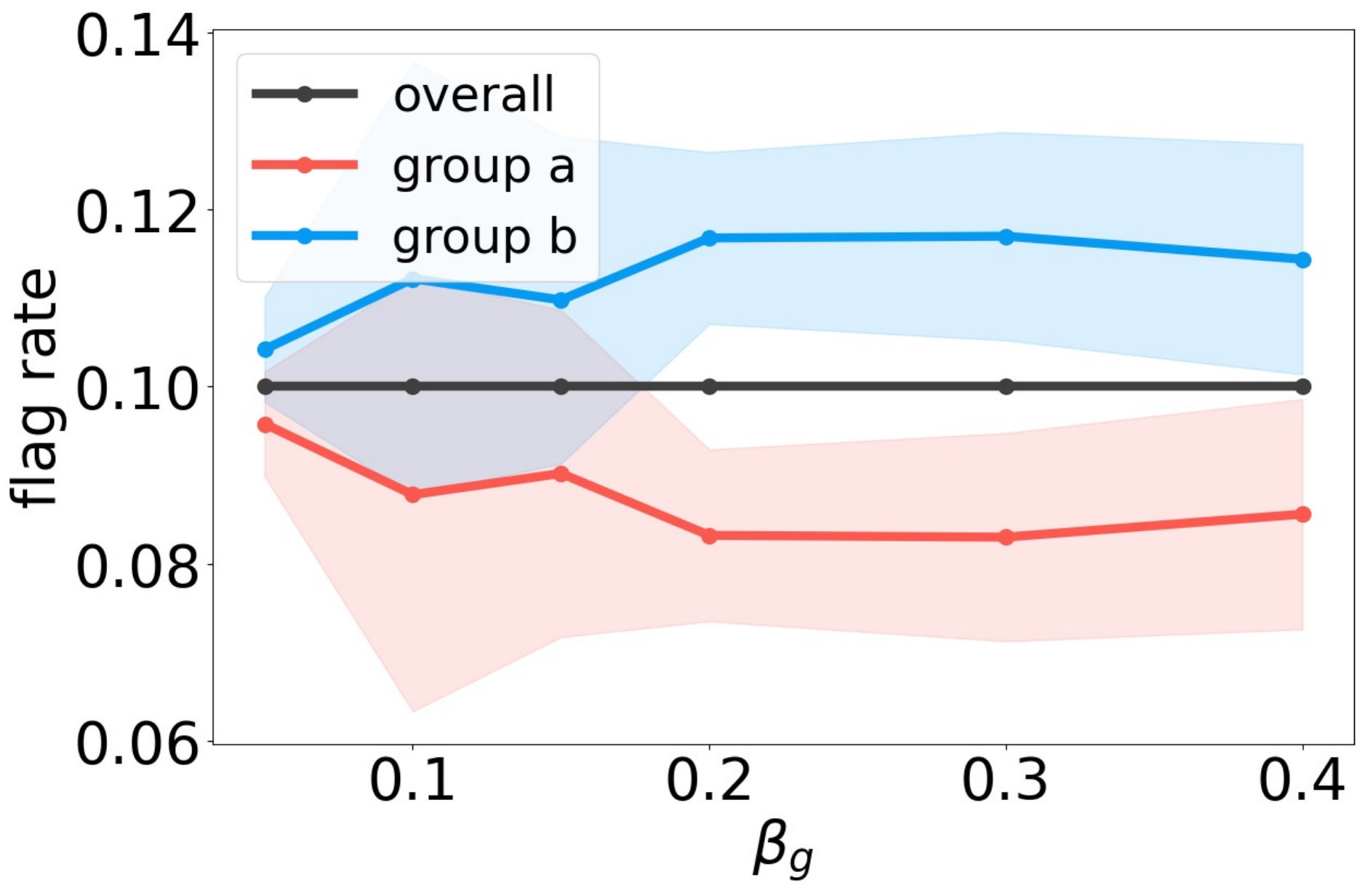} 
   } &
    \hspace{-0.1in}{\includegraphics[width=0.188\textwidth]{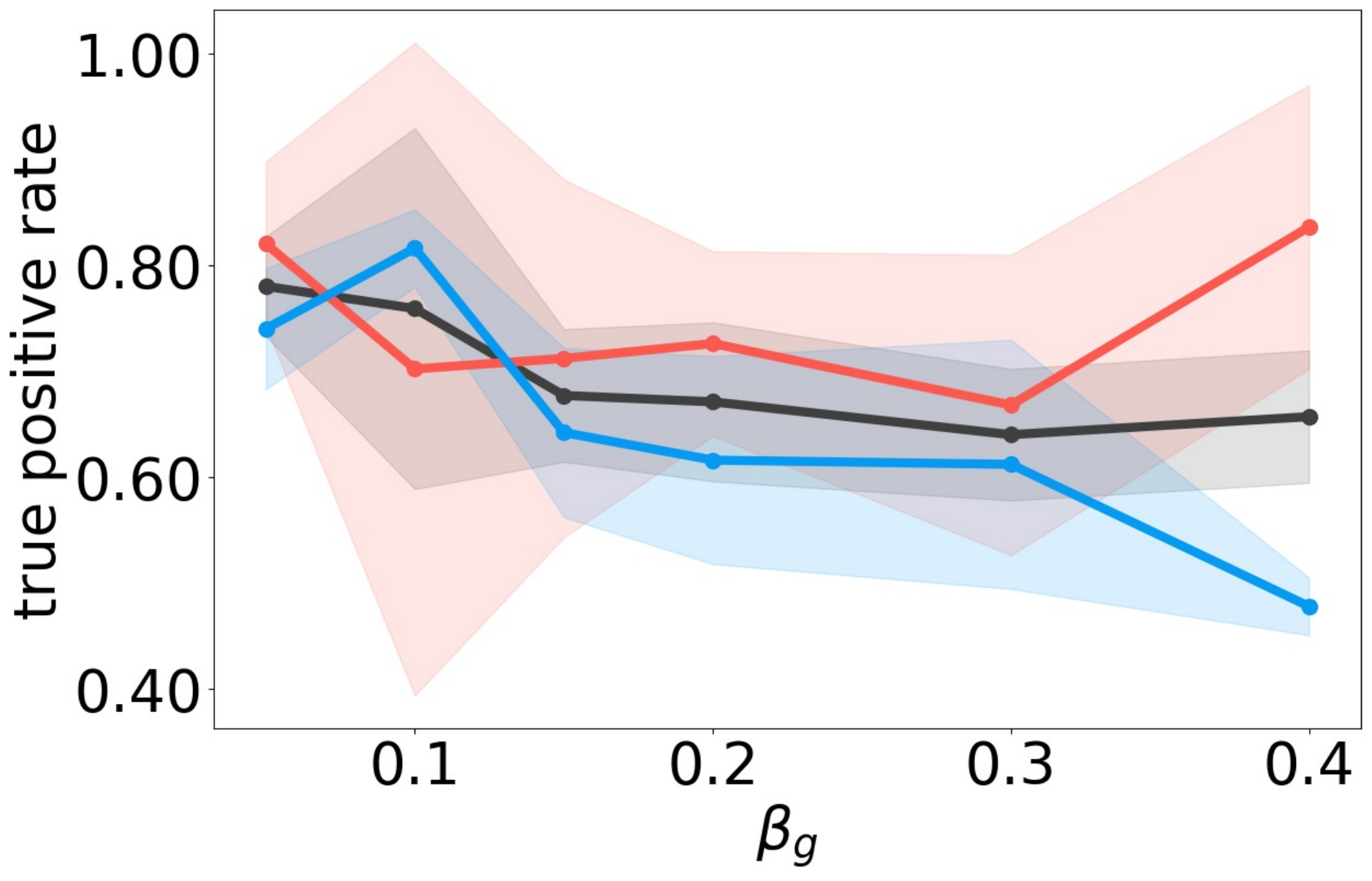}} &
     \hspace{-0.1in}{\includegraphics[width=0.188\textwidth]{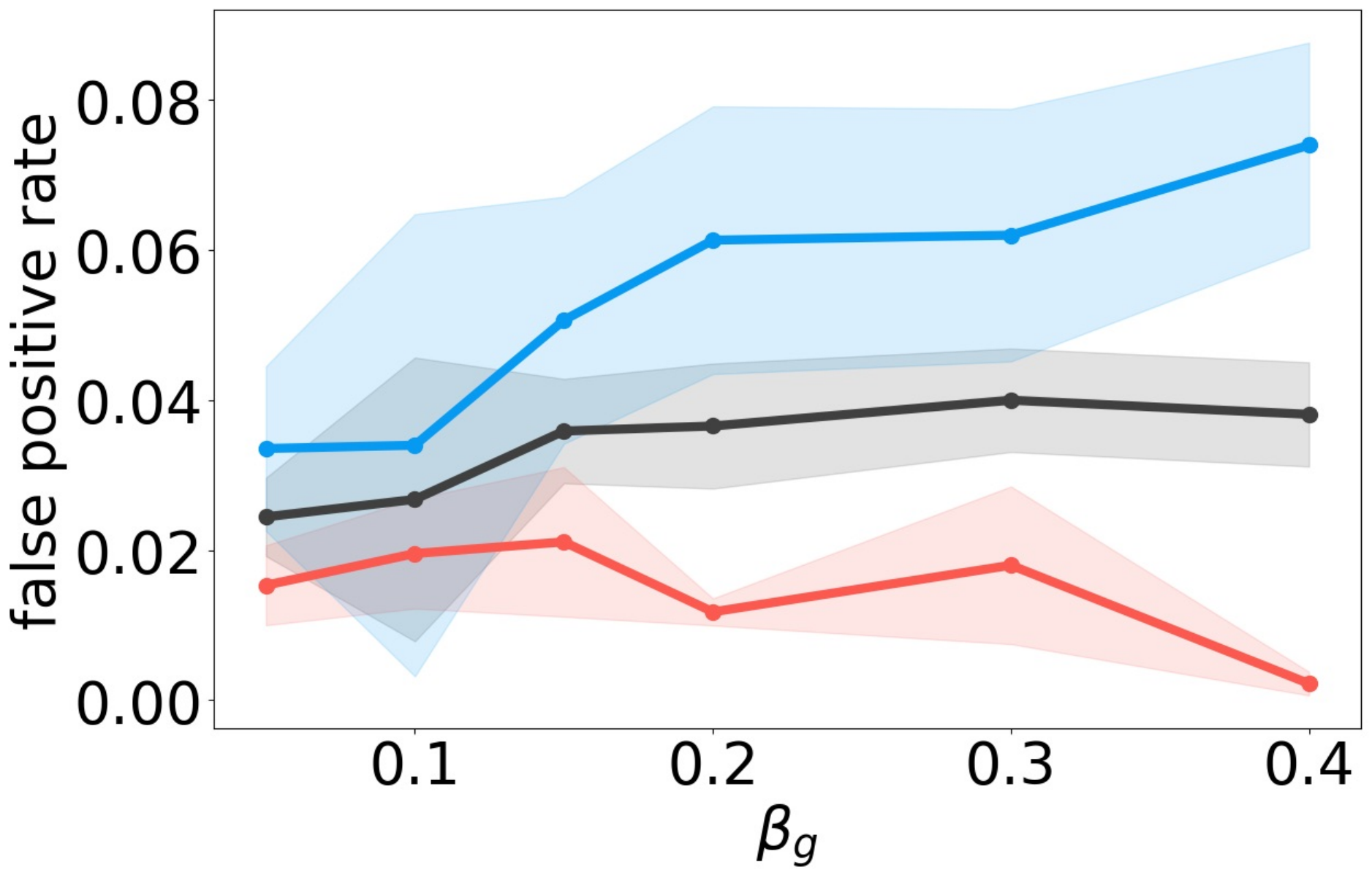}}
    &
   \hspace{-0.1in} {\includegraphics[width=0.188\textwidth]{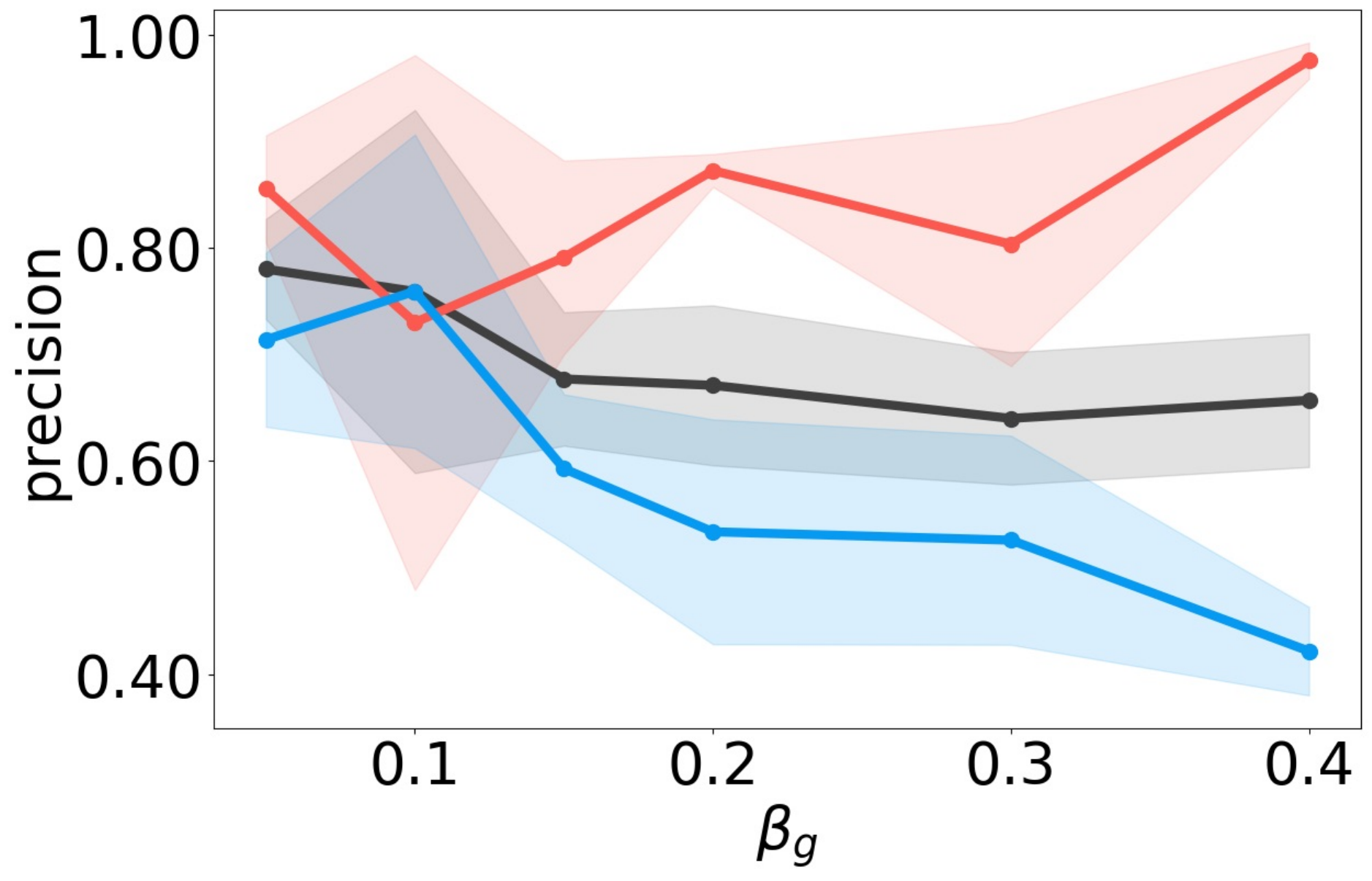}}
   &
   \hspace{-0.1in} {\includegraphics[width=0.187\textwidth]{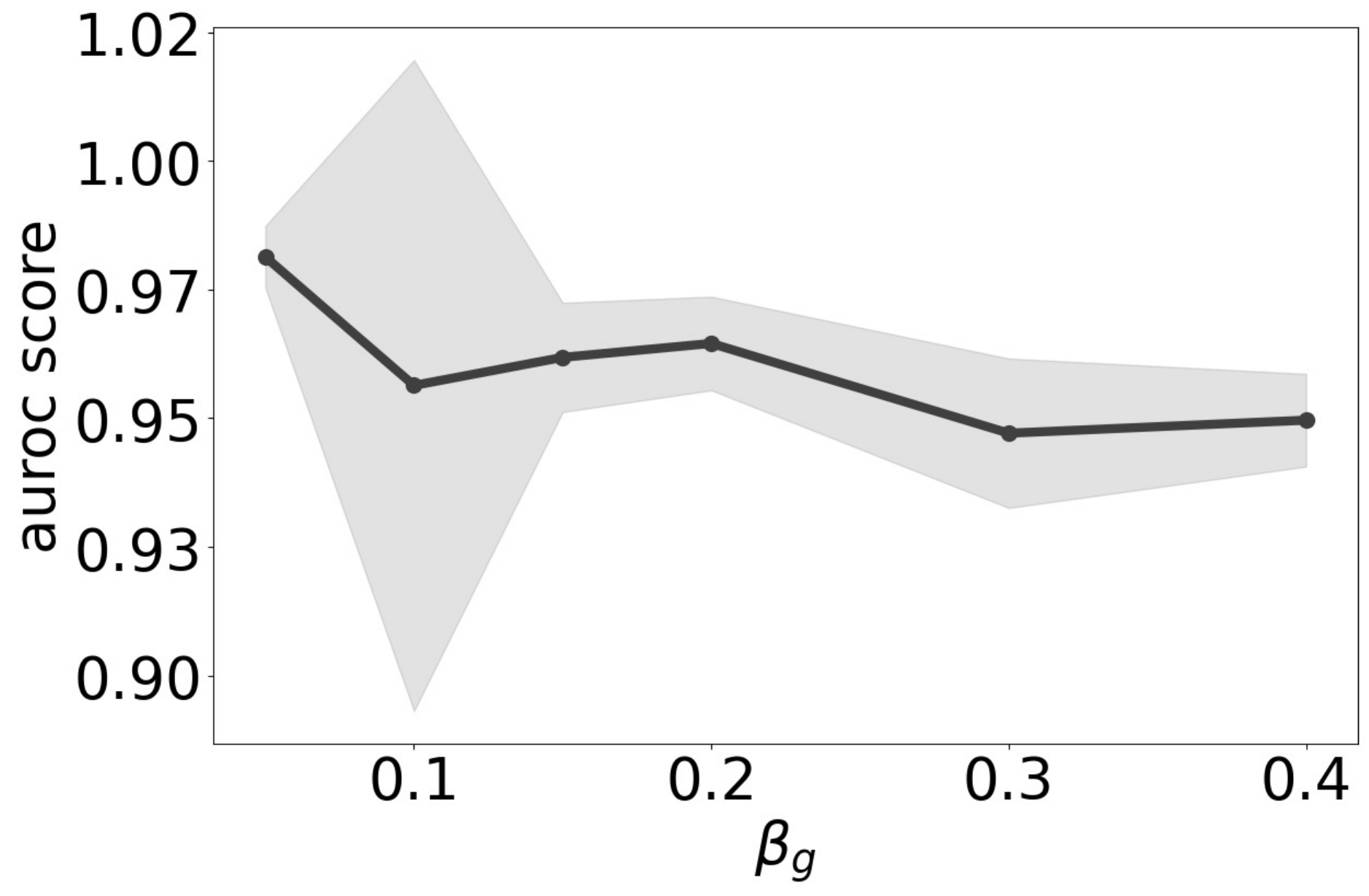}}
    \end{tabular}
    \caption{(best in color) Group-wise fairness metrics and AUROC for (top) LOF and (bottom) FairOD under membership obfuscation bias on \textit{clustered} outliers. }
    \label{fig:obfuscation_loffairod_cluster}
\end{figure*}

In the scattered outliers setting, feature variance loosens the distribution of group $b$ leading to extreme-valued inliers and outliers that are both the at the ``skirts'' of the distribution.
As a result, as group $b$ data sparsifies, 
LOF tends to blur the inliers and outliers at the periphery, with higher FPR, and lower TPR and  Precision for group $b$.
Performance also drops accordingly with increasing bias levels.   
Results are similar for iForest, which overly flags the now-sparser group $b$ samples.
Both deep models are also sensitive in this case, yielding unfair outcomes for group $b$ along with reduced performance. While clustered outliers, despite larger variance, form a separate modality in the data that is hard to compress (hence easier to detect), scattered outliers are harder to detect as they are more similar to the inliers at the outskirts of the inlier distribution. 

The different behavior of models between the two settings demonstrates that unfairness may arise from the interaction between the data distribution, the type of bias, as well as the model assumptions.

\vspace{-0.075in}
\subsubsection{\bf Mean shift for over-estimation.~}
We find that the OD models are \textit{not} impacted by mean-shifting group $b$, provided that the groups are sufficiently apart along the proxy variables. Specifically,  the shift does not impact the locality of points for LOF, the axis-cuts of iForest that isolate the outliers simply shift accordingly once the groups are cut separate along the proxy variable axes, and the compression-based deep models remain intact when one of the modalities shifts in the feature space.
Results are shown for LOF and iForest for brevity   
in Apdx. \S\ref{assec:meas_cluster} Fig. \ref{fig:meanshift_lofif_cluster} and Apdx. \S\ref{assec:meas_scatter} Fig. \ref{fig:meanshift_lofif_scatter} in the clustered and scattered outliers settings, respectively.

\vspace{-0.05in}
\subsection{Membership obfuscation bias}
\label{ssec:obf}

Different from earlier bias scenarios in which the data collector or their instrument may be inflicting the bias, here the individuals in the population themselves bias the data distribution. They do so by inducing mixed-up subpopulations, which resemble in some parts to group $b$ and in other parts disguising as group $a$. As the obfuscation variables are randomized, this leads to \textit{multi-modality} within group $b$, comprising many small subpopulations.
Results can be found
in Apdx. \S\ref{assec:obf_cluster} Fig. \ref{fig:obfuscation_all_cluster} and Apdx. \S\ref{assec:obf_scatter} Fig. \ref{fig:obfuscation_all_scatter} in the clustered and scattered outliers settings, respectively.

As shown in Fig. \ref{fig:obfuscation_loffairod_cluster} (top), LOF is extremely sensitive to this type of bias as it mistakenly flags more and more of those small micro-clusters in group $b$. In fact, it seizes to flag any group $a$ instances beyond a certain bias level as group $b$ breaks down into more subpopulations. This leads to reduced TPR for both groups, and high FPR and low Precision for group $b$. LOF's overall performance also drops with increasing levels of bias. 
iForest, on the other hand, is relatively more robust because it is \textit{subspace}-based; that is, it is able to isolate outliers as long as it makes axis-cuts on the incriminating variables. Put differently, as long as it does not solely pick proxy variables, it does not wrongly flag group $b$ micro-clusters. The probability of this event is low, hence the slight unfairness against group $b$, with high overall performance retained.

Similar to LOF, DeepAE is sensitive to the presence of small micro-clusters as more variability in data patterns imply harder compression. Thus its overall performance drops, with higher flag rate, higher FPR and lower Precision for group $b$.
Higher FPR and lower Precision for group $b$ are common across models including FairOD, as shown in Fig. \ref{fig:obfuscation_loffairod_cluster} (bottom), which better balances flag rate and TPR rates between groups thanks to explicit regularization.

\section{Algorithmic Bias is \underline{not} merely a Data (Bias) problem}
\label{sec:algobias}

When data that is comprised of multiple protected groups is inflicted with data bias on some but not all group(s), it often leads to differences in the observed data distributions between the groups.  
While this can be interpreted as a data bias problem, it is important to note that differences in data distributions between groups may as well be \textit{natural}. 
In fact, assuming the data is not stripped off of all the proxy variables (which is unlikely),  it is perhaps easier to imagine that different groups would follow different data distributions organically, at least with respect to the proxy variables.

In the following subsections, we aim to make connections between specific data properties that emerge as a result of bias and how such properties may also appear in the real world organically. 
These connections support the argument toward moving beyond the ``algorithmic bias is a data (bias) problem'' debate \cite{hooker2021moving}.
That is, algorithmic bias can arise solely from the interaction between  modeling assumptions and  certain properties the input data exhibits, without the data being necessarily inflicted with any bias.

In a nutshell, we argue that group-wise differences in certain data properties (as pertain to sparsity, prevalence or base rate, variance, and multi-modality) --- \textit{either induced by data bias or  exhibited naturally} --- could result in unfair outcomes. Whichever the source may be, it is useful to understand which OD algorithms are more prone to unfairness in the face of such group-wise differences. %

\vspace{-0.05in}
\subsubsection*{\bf Theoretical Analysis.~}  
We  mathematically show the (in)sensitivity of the mechanistic (i.e. non-learning based) models, LOF and iForest, in the presence of sparsity difference and multi-modality in the clustered outliers setting (Propositions 1--4). Analysis can be repeated for scattered outliers, e.g. Proposition 5. 
We refer to Apdx. \ref{asec:theory} for preliminaries, notation, and the detailed proofs.

\vspace{-0.05in}
\subsection{Group Sample Size Bias mimics Sparsity Difference between Groups}
\label{ssec:sizebias2}

In injecting group sample size bias,  
the process of downsampling data from one group creates a population that is sparsely sampled from its underlying manifold. 
This induces density variation in the feature space where one group's distribution is relatively sparser (see Fig. \ref{fig:summary}). As presented in \S\ref{ssec:sizebias}, we observed that this property makes isolation-based iForest quite brittle, as it  tends to more easily (yet mistakenly) isolate the samples in sparser regions. 

However, it may as well be natural that different groups exhibit different  distributions with unequal sparsity. This would be especially realistic for proxy variables. For example hair length, a proxy for gender, could follow  distributions with varying density between groups. A similar argument can be made for income, a potential proxy for race, with different group-wise density distributions.

Note that in group size bias injection, down- or sparsely-sampling group $b$'s distribution, which is the \textit{same} as group $a$'s, induced not only size disparity but also sparsity disparity. Here we argue that OD models could remain prone to producing unfair outcomes when group $a$ and $b$ exhibit \textit{different} density  distributions even in the \textit{absence} of any group size disparity.

\vspace{0.05in}
\textit{\textbf{Proposition 1}: 
In the clustered outliers setting, let groups $a$ and $b$ have equal size $n_a=n_b$ and equal base rate $P(Y=1|a)=P(Y=1|b)$.
Let $\intraa$ and $\intrab$ denote intra-group distance between inlier pairs (same for outlier pairs) for group $a$ and $b$, respectively, where $\intrab>\intraa$ as group $b$ is sparser. 
Also denote by $\dista$ and $\distb$ the average distance between outliers and inliers in each group (See Fig. \ref{fig:prop1}).
Then, assuming LOF hyperparameter $k$ is set s.t. $k > n_a \cdot P(Y=1|a)$ and $\dista = \distb$, LOF tends to assign higher  scores to group $a$-outliers, increasing flag rate $P(O=1|a)$.
Further, when $\Delta_b \approx D$, LOF score of $b$-outliers is $\approx 1$, i.e. close to inlier scores, leading to low TPR for group $b$ due to masking.
}
\vspace{0.05in}

In plain words, in the presence of group-wise sparsity variation, the proposition states that 
 the local reachability distance (i.e. LOF score) of $a$-outliers tends to be larger since their inlier neighbors' reachability distance is smaller when/because group $a$ is denser. Further, as group $b$ continues to sparsify such that the gap between inlier and outlier clusters of group $b$ shrinks, $b$-outliers get harder for LOF to distinguish from inliers, decreasing TPR. 
Proposition 5 in Apdx. \ref{assec:prop5} shows a similar result for LOF for scatter outliers.

\vspace{0.075in}
\textit{\textbf{Proposition 2:} In the clustered outliers setting,
 let $sp$ denote an iTree's split, and $P(x_a < sp < x_b)$ denote the probability for a split to occur between points $x_a$ and $x_b$. Let groups $a$ and $b$ have equal base rate $P(Y=1|a)=P(Y=1|b)$
 and equal average distance between outliers and inliers $\dista = \distb$.
  As group $b$ is sparser, $\intraa< \intrab$ denotes the intra-group distances between inlier and outlier pairs for groups $a$ and $b$, respectively (See Fig. \ref{fig:prop4}). Let $\{\outliers_a$, $\inliers_a\}$, $\{\outliers_b$, $\inliers_b\}$ represent the outliers, inliers sets for groups $a$ and $b$. 
 Then, for $o_b, p_b \in \outliers_b$, and $o_a, p_a \in \outliers_a$, we have $P(p_b < sp < o_b) > P(p_a < sp < o_a)$, indicating an iTree is more likely to split among group-$b$ outliers. The difference between $P(p_b < sp < o_b)$ and $P(p_a < sp < o_a)$ becomes larger when iTree is built in higher dimensions. In addition, for $q_b \in \inliers_b$, $q_a \in \inliers_a$, $P(o_b < sp < q_b) = P(o_a < sp < q_a)$, i.e. an iTree is equally likely to split between the inlier and outlier clusters. 
} 
\vspace{0.025in}

iForest assigns higher outlier scores to points that can be easily isolated with few axis splits.
The proposition shows that an iTree is more likely to split among the sparser $b$-outliers than $a$-outliers, yielding higher flag rate and FPR for group $b$. The discrepancy is more notable in higher dimensional splits. iTree is equally likely to split between the inliers and outliers  for both groups. Due to this special split that isolates the clustered outliers at once, iForest can flag some fraction of $a$-outliers, retaining TPR for group $a$. 

Empirically, variance shift experiments could serve as a similar scenario where samples of group $b$  are more sparsely scattered than those of $a$ due to inflated variance,  which we observed is subject to higher FPR and lower TPR and Precision, consistent across all models (see Apdx. \S\ref{assec:meas_scatter} Fig. \ref{fig:varianceshift_all_scatter}).
This shows one possible scenario, wherein group $b$ organically exhibits greater sparsity, and group-wise sparsity difference leads to disparate OD outcomes.

\begin{figure*}[!ht]
    \centering
    \begin{tabular}{lllll}
        \hspace{-0.1in}{\includegraphics[width=0.19\textwidth]{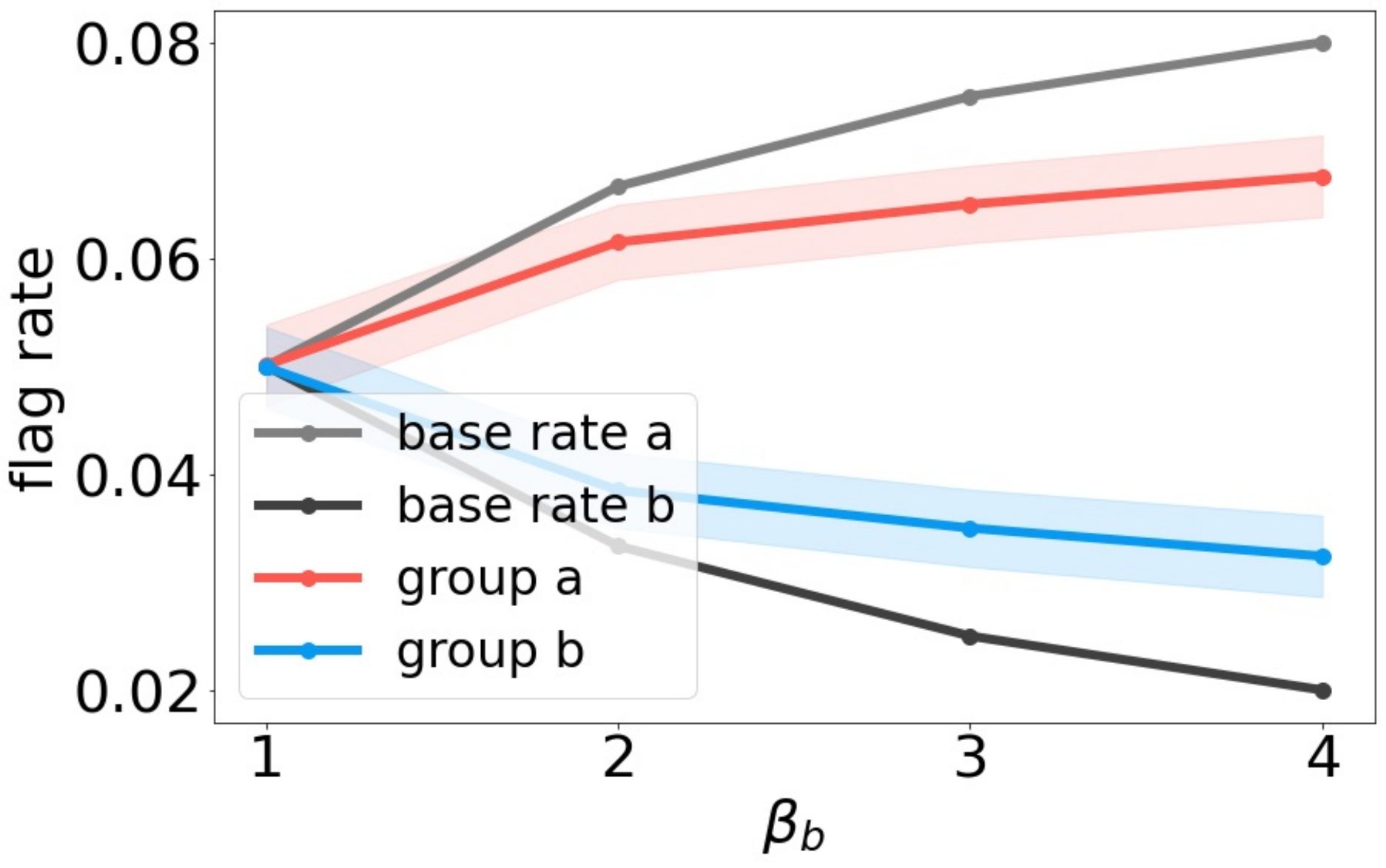}} 
    &
    \hspace{-0.1in}{\includegraphics[width=0.19\textwidth]{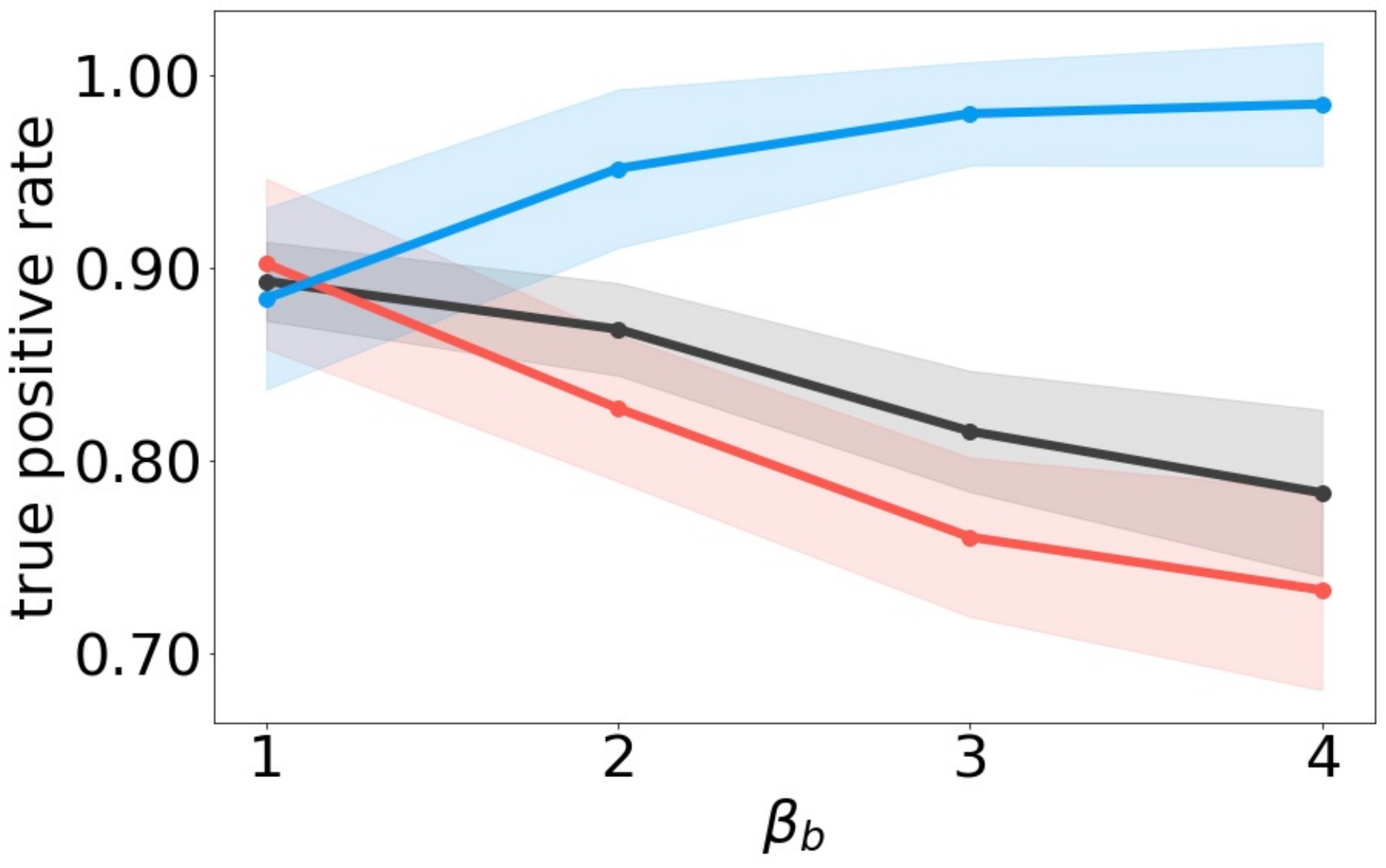}} &
     \hspace{-0.1in}{\includegraphics[width=0.19\textwidth]{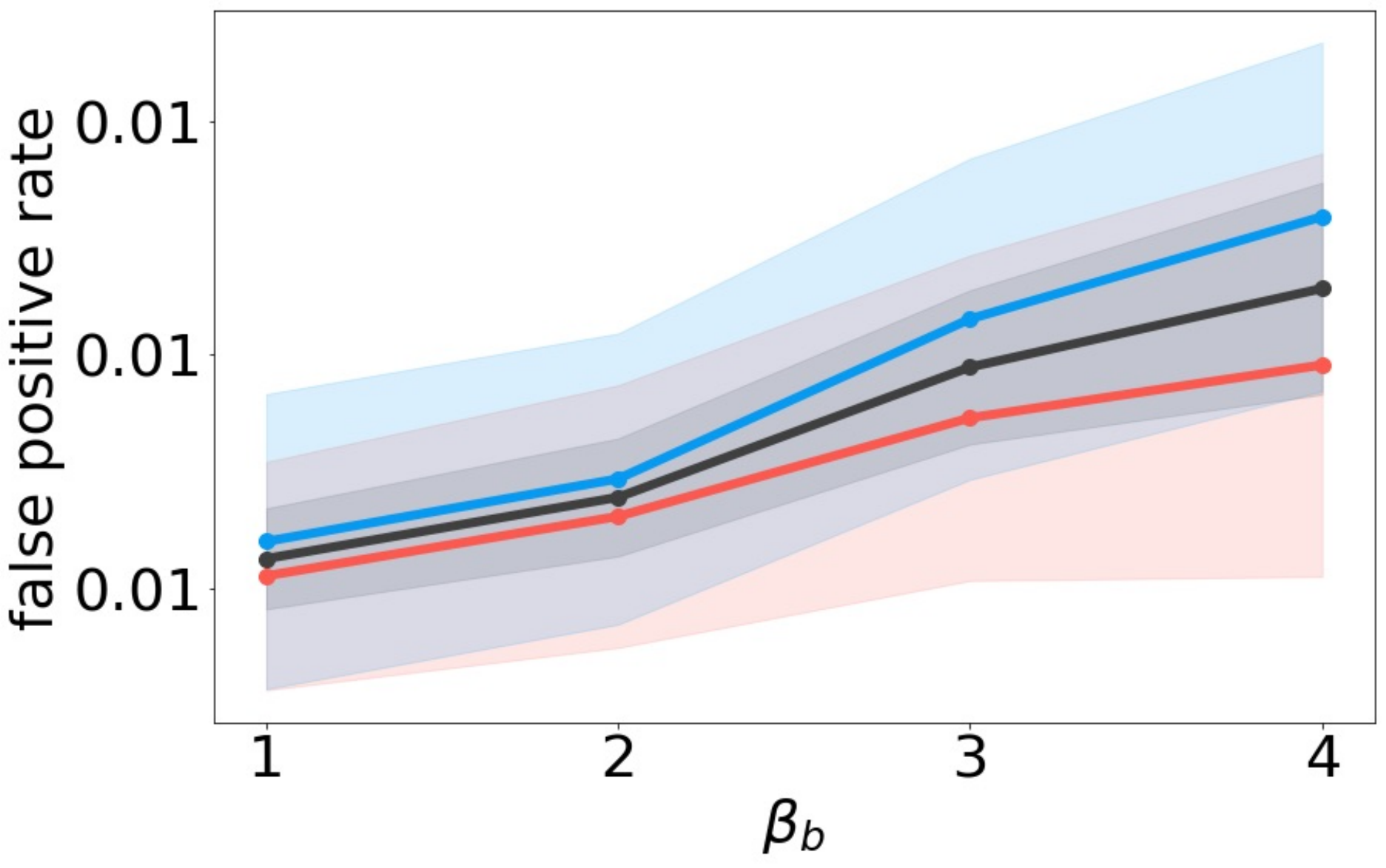}}
    &
   \hspace{-0.1in} {\includegraphics[width=0.19\textwidth]{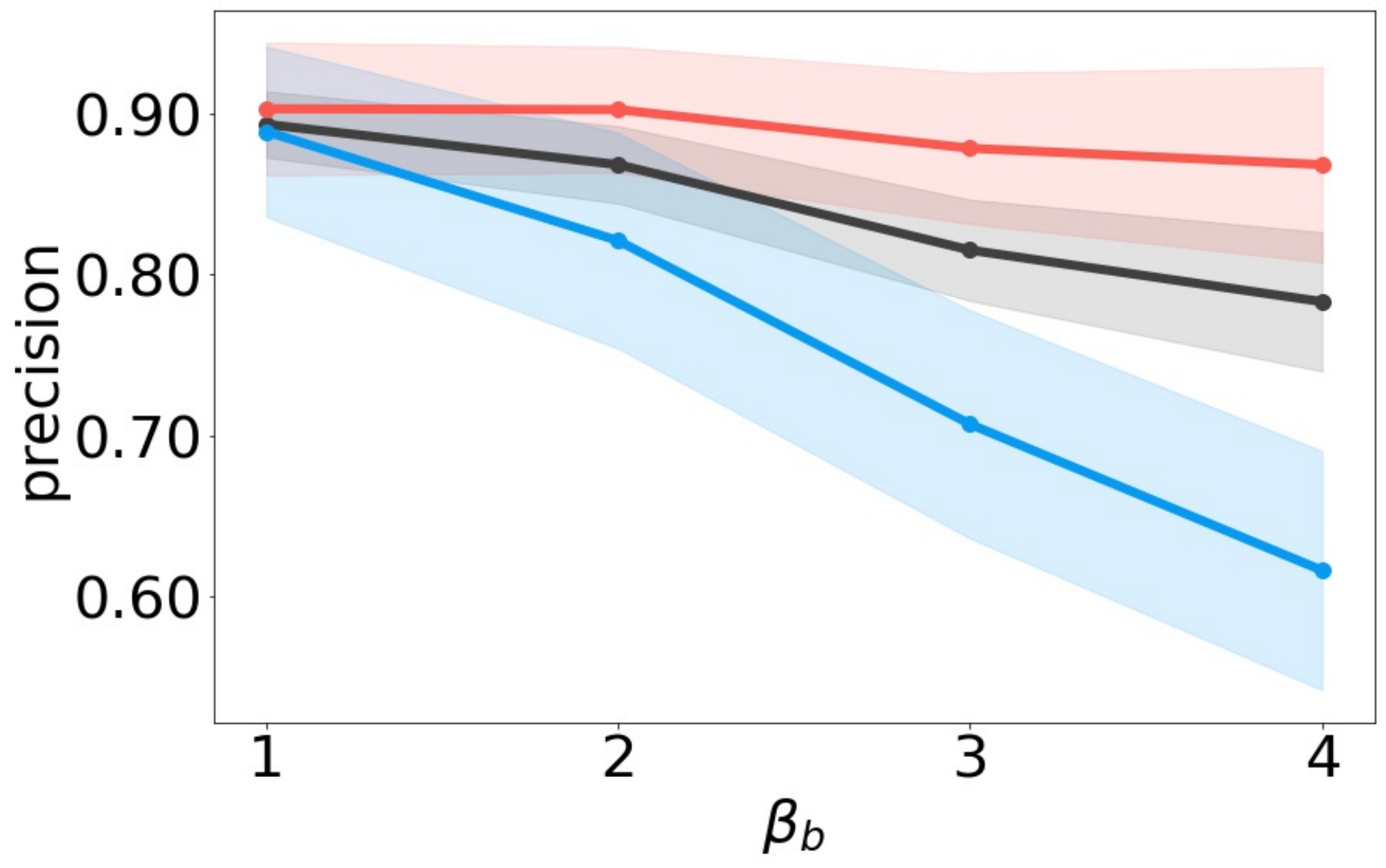}}
   &
   \hspace{-0.1in}
   {\includegraphics[width=0.19\textwidth]{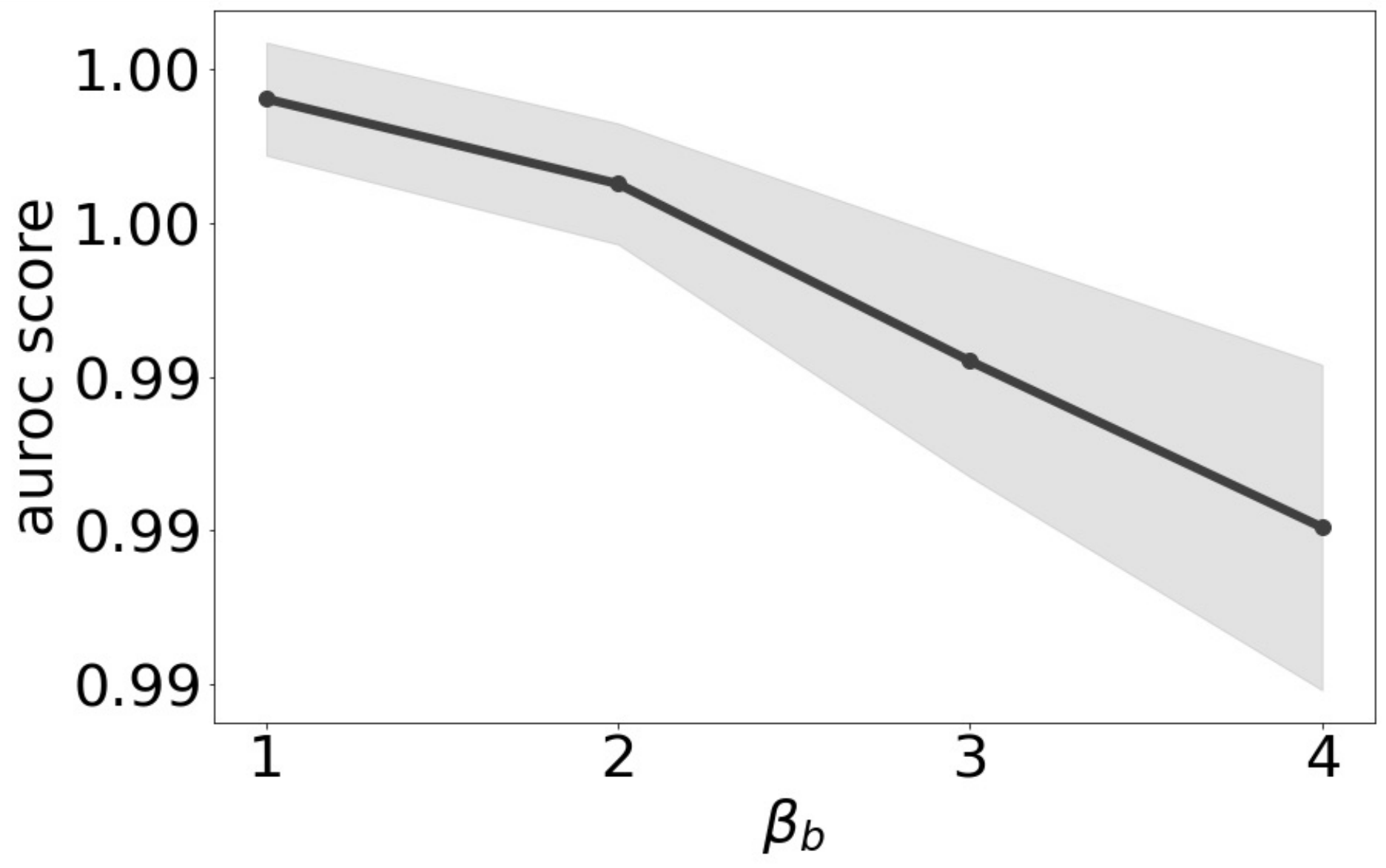}}
   \\
\hspace{-0.1in}{\includegraphics[width=0.19\textwidth]{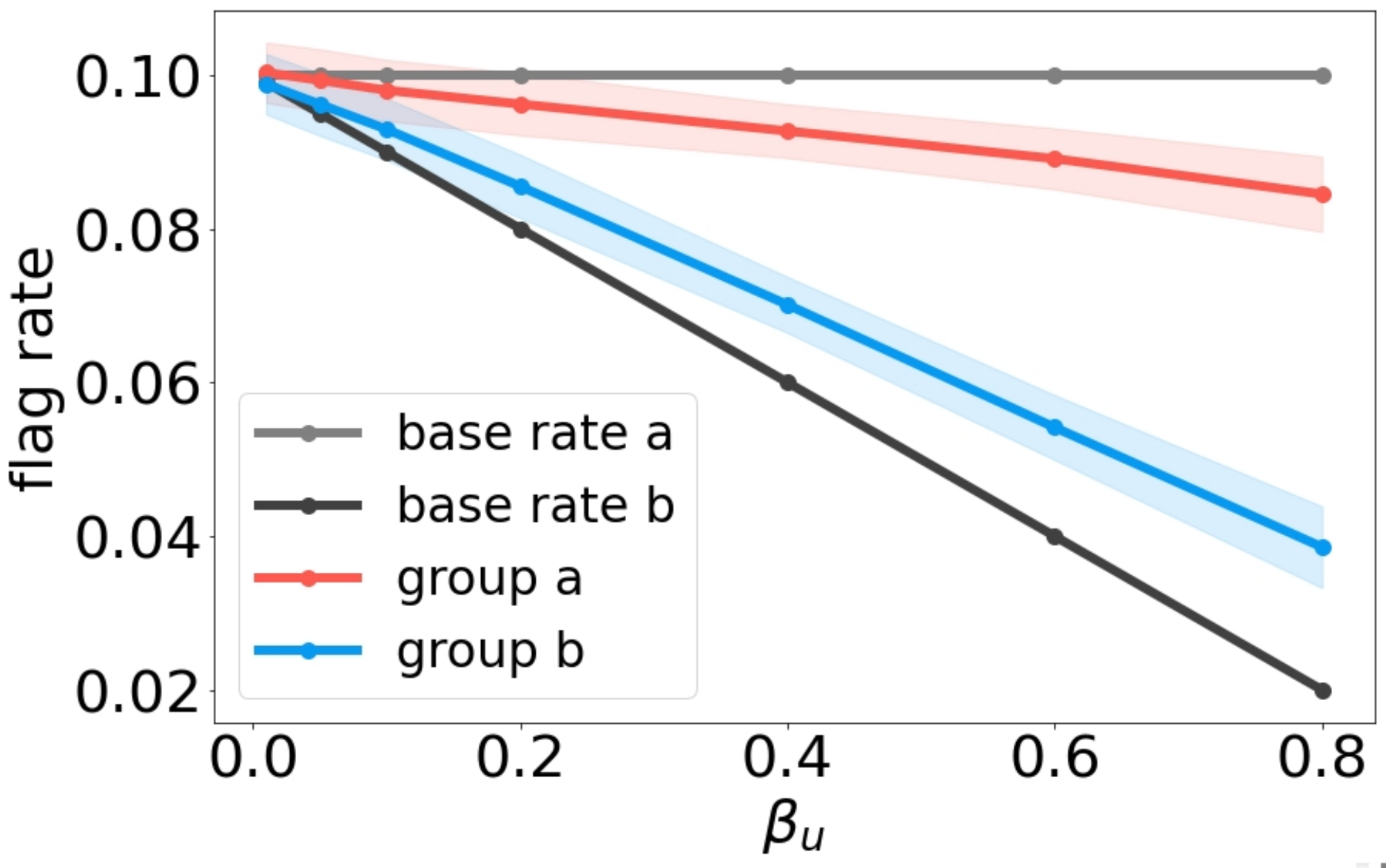}} 
    &
    \hspace{-0.1in}{\includegraphics[width=0.19\textwidth]{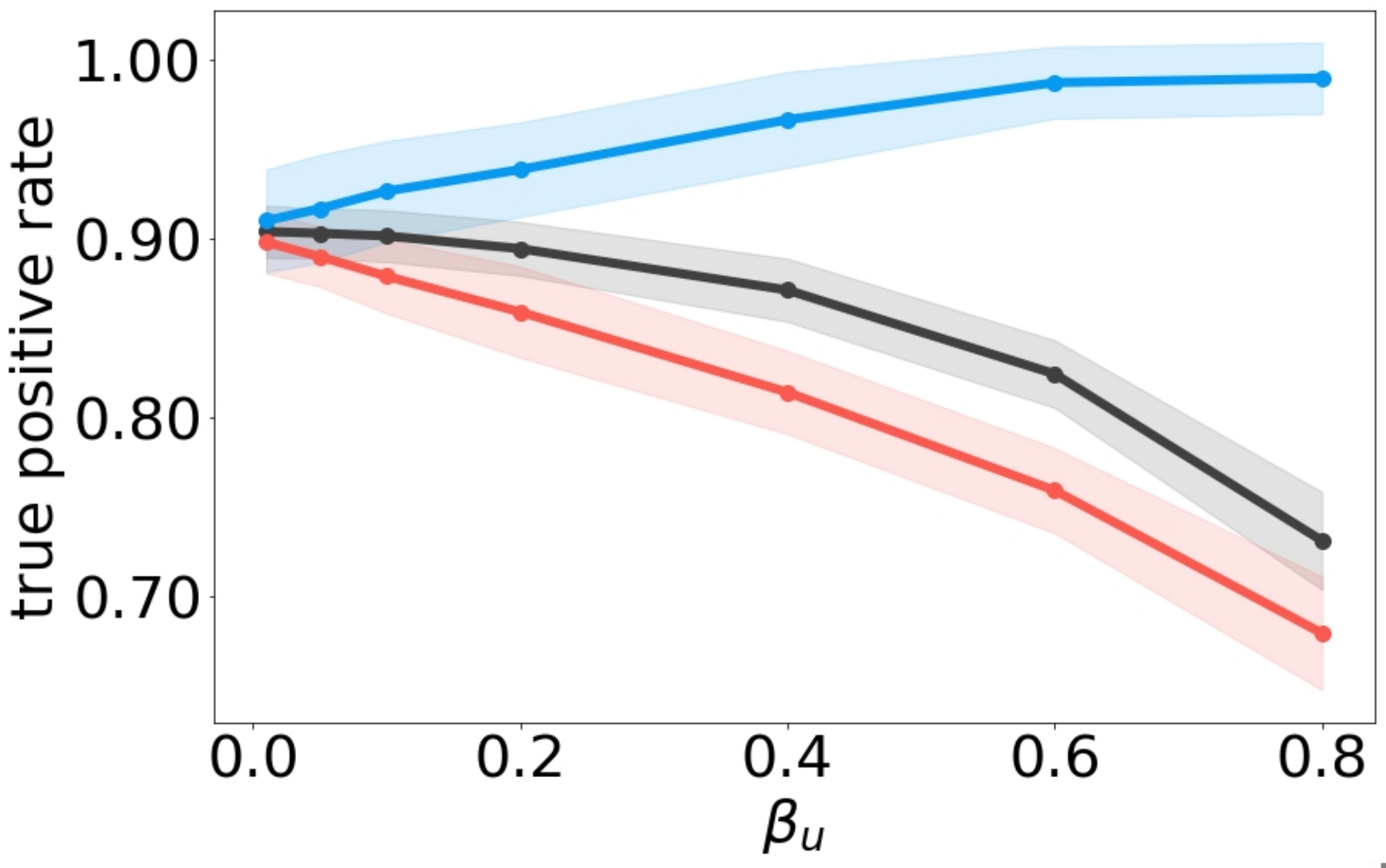}} &
     \hspace{-0.1in}{\includegraphics[width=0.19\textwidth]{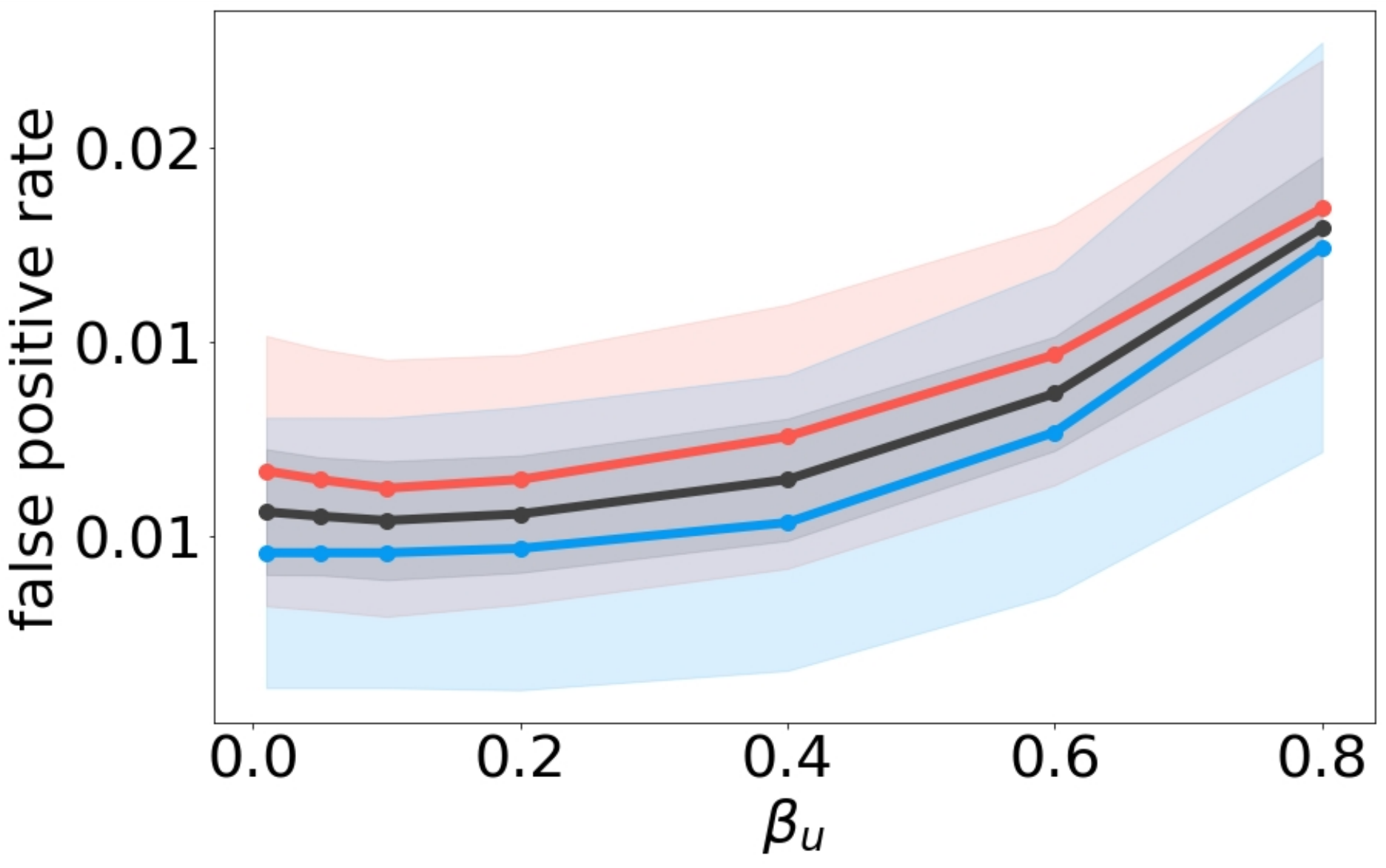}}
    &
   \hspace{-0.1in} {\includegraphics[width=0.19\textwidth]{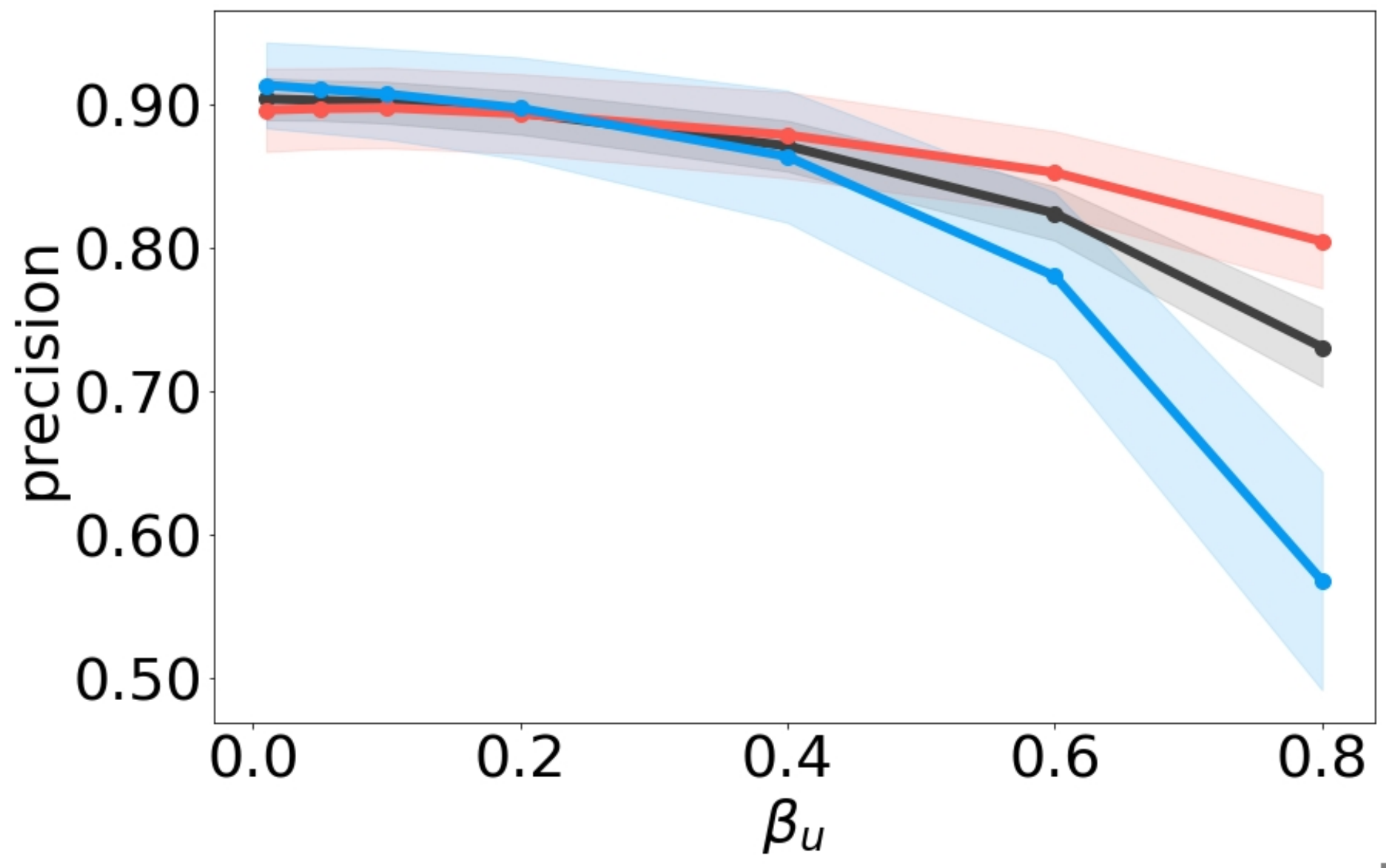}}
   &
   \hspace{-0.1in}
   {\includegraphics[width=0.19\textwidth]{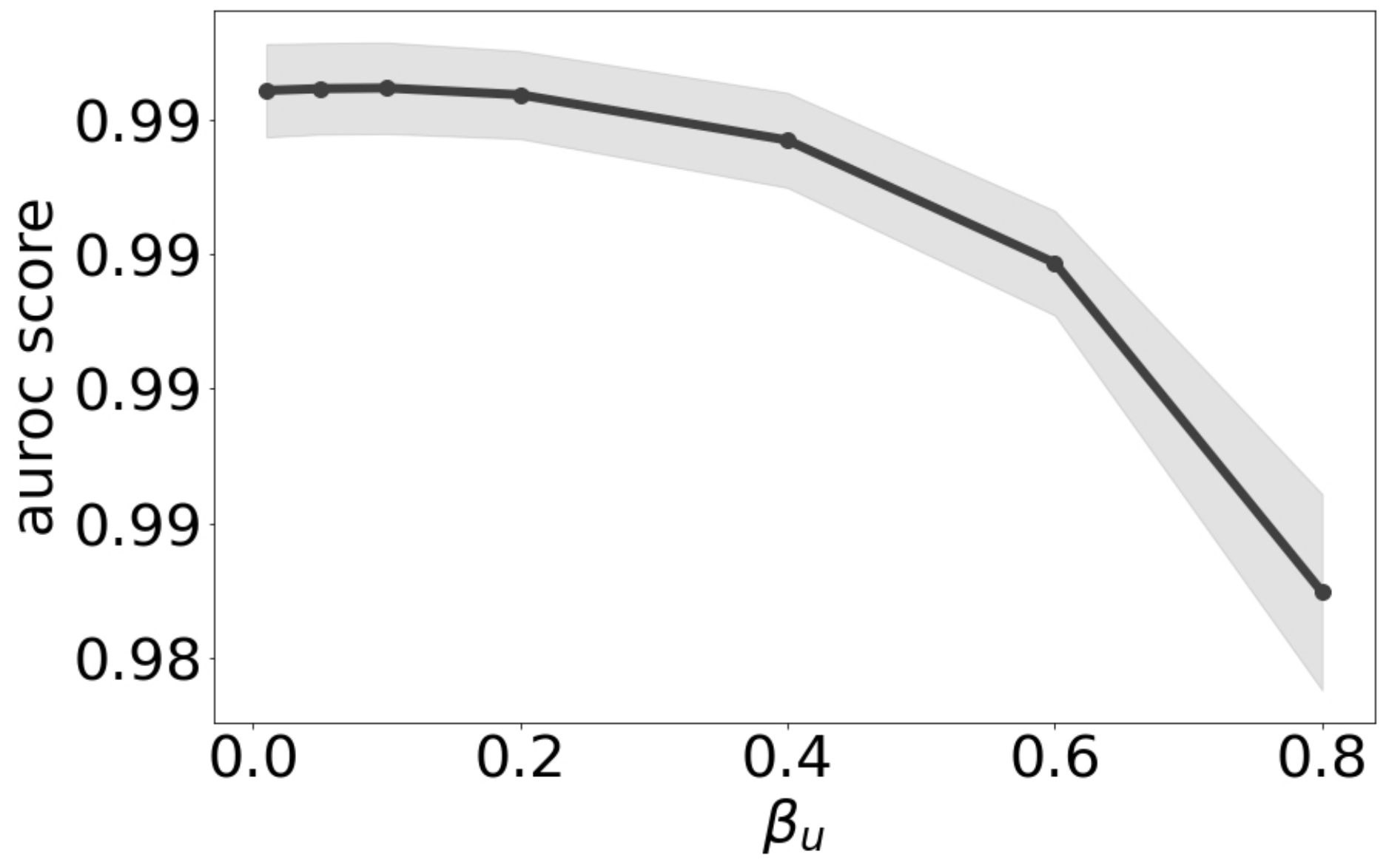}}     \end{tabular}

    \caption{(best in color) Group-wise fairness metrics and AUROC for LOF   under (top) unequal base rates and (bottom) target under-representation on \textit{clustered} outliers. The results are qualitatively similar.}
    \label{fig:baserate_under_lof_cluster}
\end{figure*}

\vspace{-0.1in}
\subsection{Target Under-Representation Bias mimics Base-Rate Difference between Groups}
\label{ssec:urep2}

In injecting target under-representation bias, the process of downsampling group $b$ outliers induces its population to have a lower observed base rate than group $a$. Here we argue that the protected groups may as well exhibit unequal base rates naturally, where 
$P(Y=1|G=a) \neq P(Y=1|G=b)$.

For example, it could be the case that some group(s) are more inclined to criminal activities (such as Internet crime, human trafficking, money laundering, click farming, etc.) than others by nature.
In fact, this is arguably one of the key differences between punitive OD settings and assistive ML settings. While it may be more ethical to provide equal assistance (hiring, loan approval, etc.) by ML algorithms across various protected groups where there is no reason or evidence to suggest one group should necessarily be more deserving than the other,  it could be appropriate to penalize various protected groups at  different rates when there is evidence or reason to believe that the group base rates indeed differ. 

Then, in the presence of (natural) \textit{unequal} base rates,  unfair OD outcomes  observed in \S\ref{ssec:urep} continue to hold. 
To demonstrate this phenomenon, we repeat our experiments by varying the base rates between groups while keeping the total number of outliers (and group-wise inliers) the same.  Results are shown in Fig. \ref{fig:baserate_under_lof_cluster} (top) for LOF for brevity, which remain qualitatively similar to those under target under-representation bias as shown in Fig. \ref{fig:baserate_under_lof_cluster} (bottom). 
Results for all models are in Apdx. \S\ref{assec:ubs_cluster} Fig. \ref{fig:baserate_all_cluster} and Apdx. \S\ref{assec:ubs_scatter} Fig. \ref{fig:baserate_all_scatter} in the clustered and scattered outliers settings, respectively.

\vspace{-0.05in}
\subsection{Feature Measurement Bias mimics  \\ Variance Difference  between Groups}
\label{ssec:meas2}
Variance-shift reflecting measurement noise may change the propensity of extreme values and low-frequency observations.
As studied in \S\ref{ssec:meas}, this may induce unfair OD outcomes for the high-variance group especially when the outliers are scattered. 

In the real world, certain features could also exhibit 
natural variation between groups. 
For example, biomarker features when used for screening may exhibit natural genetic variation by age, gender, or race.
Another, classical example is the 
Nymwars controversy\footnote{\url{https://en.wikipedia.org/wiki/Nymwars}} in 2011, where Google, Facebook, and other tech companies aimed to block users who used uncommon (hence, presumably fake) names. While the policy has been criticized for various reasons, one could see the potential discrimination it could cause given the inherent variability in African first names.

We argue that variance difference between groups mimics the scenario when such difference is induced by additive measurement noise inflicted on some group(s) but not others. This is true for Gaussians; Gaussian noise added to a Gaussian distribution yields another Gaussian with larger variance. In fact, for any distribution from the stable distribution family, a linear combination of two distributions remains in the same distribution, only with different location and scale parameters \cite{Mainardi2007}. This implies that the unfair OD outcomes observed in \S\ref{ssec:meas} would continue to hold when one group exhibits naturally higher-variance  distribution for certain features.

\vspace{-0.15in}
\subsection{Membership Obfuscation Bias mimics  Multi-modality within Groups}
\label{ssec:obf2}

Membership obfuscation -- where members of a group disguising as those from another group -- creates (sub)groups within groups, where each subgroup disguises somewhat differently within the proxy feature space. In effect, obfuscation fragments a population, inducing multi-modality within the group. 

Arguably, members of a group may  exhibit within-group heterogeneity naturally. That is, some group(s) may be composed of many more natural clusters than others. In such cases, it is perhaps a negligence of the analyst to assign them to a single broad sensitive attribute. Consider the Asian or Hispanic ethnicity; those are very coarse classifications of large populations of individuals that in fact break down to many natural, more coherent subpopulations. 

When a protected group comprises multiple smaller subgroups, the risk increases that each may stand out as a minority micro-cluster. As we presented in \S\ref{ssec:obf}, OD models are typically tuned to flagging micro-clusters as (clustered) outliers, leading to many false positives at the expense of low recall for the multi-modal group. 
In the following, we mathematically show that LOF is quite vulnerable in the presence of this data characteristic, supporting the empirical observations in Fig. \ref{fig:obfuscation_loffairod_cluster} (top). (See all proofs in Apdx. \ref{assec:theory_result}).

\vspace{0.05in}
\textit{\textbf{Proposition 3}: In the clustered outliers setting, let groups $a$ and $b$ have equal size $n_a=n_b$ and equal base rate $P(Y=1|a)=P(Y=1|b)$. Assume group $b$-inliers form $f$ smaller populations, $\inliers_b^{i}$, for $i \in [\![1,f]\!]$, and group $b$-outliers form $g$ smaller populations, $\outliers_b^{j}$ for $j \in [\![1,g]\!]$. 
Let $\intraa$ and $\intrab$ denote intra-group distance between inlier pairs (same for outlier pairs) for group $a$ and each sub-population in $b$, respectively, where $\intrab=\intraa$. Denote $\dista$ as the average distance between outliers and inliers in group $a$, and   $\distb$ as the average distance between  each subpopulation in group $b$ for both inliers and outliers (See Fig. \ref{fig:prop3}). Then, assuming LOF hyperparameter $k$ is set s.t. $k > n_a \cdot P(Y=1|a)$, and $\dista = \distb$ while $\dista > \intraa$ and $\distb > \intrab$, LOF tends to assign higher scores to group $b$-outliers, increasing flag rate $P(O=1|b)$.
Further, when $|\inliers_b^j| < |\outliers_b^i| < k$, LOF score of $b$-inliers is larger than of $b$-outliers, leading to both high FPR and low TPR for group $b$.}
\vspace{0.05in}

We also show via Proposition 4 in Appx. \ref{assec:prop4}  that iForest is relatively more robust than LOF against within-group multi-modality since, unlike LOF that relies on nearest neighbor distances in the \textit{full} feature space, it seeks outliers in \textit{sub}spaces. As protected groups form clusters only w.r.t. proxy features, iForest is prone to false positives when axis-splits are limited to proxy features, yet, making axis-splits along the incriminating features allows iForest to isolate true outliers and hence maintain recall.

\section{Related Work}
\label{sec:related}

A vast body of algorithmic fairness literature is on the measures and mitigation of unfairness, with less attention to the driving mechanisms and sources of unfairness. Our work focuses on the latter.  Further, while most work considers supervised ML/classification, we study unfairness in unsupervised ML/outlier detection (OD).

\vspace{0.025in}
\noindent
\textbf{Measuring and mitigating unfairness:}
Various work studied possible definitions and measures of ML fairness \cite{kusner2017counterfactual,verma2018fairness,beutel2019putting,garg2020fairness,tang2023and}, fairness auditing \cite{saleiro2018aequitas,galdon2020auditing,le2023algorithmic}, 
with most emphasis on 
mitigation in ML \cite{mehrabi2021survey,BENCHMARKS2021}, through optimization \cite{xinying2023guide}, regularization \cite{kamishima2011fairness}, constraints \cite{zafar2017fairness}, adversarial learning \cite{delobelle2021ethical}, and representation learning \cite{zemel2013learning}, to name a few.
In contrast, only a handful of work studied 
auditing \cite{davidson2020framework} and   mitigation \cite{abraham2021fairlof,song2021deep,shekhar2021fairod,liu2021fairness,zhang2021towards} for unsupervised OD.

\vspace{0.025in}
\noindent
\textbf{Sources of unfairness:} On developing a deeper understanding of what drives discrimination by ML, various work 
studied the impact of data collection  \cite{chen2018my}, algorithmic factors \cite{blanzeisky2021algorithmic}, as well as model and data interactions \cite{pombal2022understanding}, while others aimed to identify and list possible sources of harm \cite{mehrabi2021survey,suresh2021framework}.  
Our work is inspired by the testbed by Akpinar {\em et al.}
\cite{akpinar2022sandbox} that studied the explicit impact of counterfactually injected biases on \textit{supervised} ML models. In similar vein, our study is the {first} to demonstrate that popular \textit{unsupervised} OD models are susceptible to  certain forms of data bias.

\section{Conclusion and Discussions}
\label{sec:conclusion}

\vspace{0.025in}
\subsubsection*{\bf Summary.~}
We presented a descriptive measurement study that stress-tested the fairness and performance of various OD models when exposed to certain data biases.
Our analyses have been expository, unearthing the pitfalls of various algorithmic design choices as they interact with certain data characteristics---such as group-wise differences in  sparsity, prevalence, variance and multi-modality.
Among the findings, those that stand out are as follows. 

All models (shallow/deep, fairness un/aware) are prone to yield disparate impact that is stacked against the underprivileged group. The impact varies depending on the relation between the type of data bias (and the data characteristics it induces) and the model assumptions. 
In our study, the shallow detectors directly model/define what an outlier is, and yield unfair results when these definitions do not align with the data.
On the other hand, the deep models aim to identify what is normal/compressible and flag outliers indirectly as the samples that do not obey the normal patterns.  
When more than one normal is present and the normal pattern(s) are more dominant/coherent for some protected group(s) than others, these detectors also tend to produce unfair outcomes. 

While our study initially looked at OD unfairness through the lens of data bias, we realized that what is really at play is certain data characteristics that do not ``play well'' with certain algorithmic assumptions and design choices.
Therefore, it is worthy to note that 
while data bias may induce such data characteristics, they could also be natural.
This implies that OD models can naturally fall into the same pitfalls we underscored, even when the analyst, measurement instrument and the population are not at fault of inflicting  bias.

\vspace{-0.025in}
\subsubsection*{\bf Limitations.~}
Our paper presents a controlled measurement study, limited to simulated data. This is motivated by  the fact that it is not known which type(s) of data bias, if any,  real-world datasets exhibit.
 We  studied datasets injected with one (known) type of data bias at a time. 
 Our simulations also included simple data and outlier distributions. 
 While simulations are an excellent way to understand the basics, the design of our simulations may not align too closely with real-world datasets, which limits the external validity of our observations.
 Nevertheless, even in these simple scenarios, we are able to showcase  key shortcomings of various modeling choices that we argued may occur naturally in the real world.

We  did not perform experiments on real-world datasets intentionally. Our aim has been to study which data bias gives rise to unfair outcomes in a controlled setup. Explaining unfairness observed in the wild by pinpointing 
the root-causes would be a reverse-engineering effort, which is not our intended scope. Our work lays out possible sources for investigation, although our bias list is not necessarily comprehensive. Further limitations pertain to our theoretical analyses, as discussed in Apdx. \ref{asec:limitations_additional}.

\vspace{-0.025in}
\subsubsection*{\bf Future directions.~}
Our work %
paves the way to identifying other  types of data biases that may adversely impact fair OD outcomes, and developing
remediation strategies toward mitigating the potential pitfalls we underscored.
We briefly discuss possible directions here, while a deeper investigation is left as future work.

Fairness interventions are often grouped as pre-, post-, and in-processing, which respectively, modify the input data, modify the output scores or decisions, and account for fairness during model training. 
As discussed in \S\ref{sec:algobias}, OD unfairness can stem from algorithmic bias alone when  group-wise differences are natural. Then, pre-processing strategies become voided as it is not clear how to modify organic, unbiased data.
Post-processing could select different thresholds for each group separately \cite{corbett2017algorithmic,menon2018cost}, where the group-specific thresholds could either be selected in a data-driven fashion, or optimized for  demographic parity if it is a desired fairness metric (e.g. assuming equal base rates). Note that metrics based on true labels  cannot be optimized (at least directly) due to lack of ground truth in training unsupervised OD. 
In-processing techniques are also limited to only enforcing demographic parity, which as we showed, remain susceptible to unfairness. A strategy that has not been applied to OD is 
decoupling \cite{dwork2018decoupled,ustun2019fairness}, where a different detector is trained for each group, while optimizing a joint loss. 

We remark that post-processing and decoupling exhibit treatment disparity; they both assume it is ethical and legal to use the sensitive attribute at decision time -- in particular, to select which threshold or detector to employ on a  new sample. 
When there are differences \textit{between} groups, coming to terms with disparate treatment  might be the only way to get around disparate impact, as argued in \cite{lipton2018does}.
These, however, do not address unfairness against multi-modal subpopulations \textit{within} groups, i.e. within-group discrimination. Here, one direction is to explore clustering-based OD. Alternative could be establishing a more nuanced or granular sensitive attribute, assigning a different value for each subpopulation.

\section*{Acknowledgments}{
This work is sponsored by the National Science Foundation IIS-2310481. Any conclusions expressed in this material are those of the authors and do not necessarily reflect the views, expressed or implied, of the funding parties.
}

\bibliographystyle{ACM-Reference-Format}
\bibliography{00ref.bib}

\clearpage

\appendix
\section*{Appendix}

\section{Data Bias Injection Details}
\label{asec:exp_details}

\subsection{Datasets with Clustered Outliers}
\label{asec:cluster_details}

\subsubsection{Group sample size bias}
\label{assec:sizebias}
$\;$\\

Number of samples in group $a$: 1000

Number of samples in group $b$: 1000

$br_a = br_b = 0.1$

$\beta_s \in [0.01, 0.05, 0.1, 0.2, 0.4, 0.6, 0.8]$

Outlier counts in groups ($a$, $b$) respectively for varying $\beta_s$: 

\hspace{0.1in}$[(100, 100), (100, 97), (100, 87), (100, 80), (100, 56), (100, 33), (100, 24)]$

Each algorithm selected hyperparameters that yielded the best \parbox[t]{\linewidth}{
\hspace*{0.1in} performance metrics.}

LOF selected hyperparameter $k$ respectively for varying $\beta_s$: 

\hspace{0.1in}$[240, 237, 227, 220, 196, 173, 154]$

DeepAE selected hyperparameters respectively for varying $\beta_s$:
\vspace{-2mm}
\begin{table}[ht]
\centering
\hspace*{6mm} 
\begin{tabular}{|c|p{7cm}|}  %
\hline
\textbf{$\beta$} & HP \\
\hline
0.01 & \{num\_layer: 4, input\_decay: 2, epochs: 100, lr: 0.001, weight\_decay: 1e-05, dropout: 0, threshold: 1\} \\
0.05 & \{num\_layer: 4, input\_decay: 1.0, epochs: 250, lr: 0.0001, weight\_decay: 1e-05, dropout: 0, threshold: 1\} \\
0.1  & \{num\_layer: 4, input\_decay: 2.5, epochs: 250, lr: 0.001, weight\_decay: 0, dropout: 0, threshold: 1\} \\
0.2 & \{num\_layer: 4, input\_decay: 1.0, epochs: 250, lr: 0.0001, weight\_decay: 1e-05, dropout: 0, threshold: 1\} \\
0.4  & \{num\_layer: 4, input\_decay: 1.0, epochs: 250, lr: 0.0001, weight\_decay: 1e-05, dropout: 0, threshold: 1\} \\
0.6 & \{num\_layer: 2, input\_decay: 2.5, epochs: 100, lr: 0.001, weight\_decay: 0, dropout: 0, threshold: 1\} \\
0.8  & \{num\_layer: 4, input\_decay: 1.0, epochs: 100, lr: 0.001, weight\_decay: 0, dropout: 0.2, threshold: 1\} \\
\hline
\end{tabular}
\end{table}

FairOD selected hyperparameters $\alpha$ and $\gamma$ respectively   for varying $\beta_s$:
\vspace{-2mm}
\begin{table}[ht]
\centering
\hspace*{6mm} 
\begin{tabular}{|c|p{7cm}|}  %
\hline
\textbf{$\beta$} & HP \\
\hline
0.01 & \{num\_layer: 4, input\_decay: 2, epochs: 100, lr: 0.001, weight\_decay: 1e-05, dropout: 0, threshold: 1\, $\alpha$: 0.01, $\beta$: 0.8\} \\
0.05 & \{num\_layer: 4, input\_decay: 1.0, epochs: 250, lr: 0.0001, weight\_decay: 1e-05, dropout: 0, threshold: 1, $\alpha$: 0.05, $\beta$: 0.5\} \\
0.1  & \{num\_layer: 4, input\_decay: 2.5, epochs: 250, lr: 0.001, weight\_decay: 0, dropout: 0, threshold: 1\, $\alpha$: 0.5, $\beta$: 0.8\} \\
0.2 & \{num\_layer: 4, input\_decay: 1.0, epochs: 250, lr: 0.0001, weight\_decay: 1e-05, dropout: 0, threshold: 1, $\alpha$: 0.01, $\beta$: 0.5\} \\
0.4  & \{num\_layer: 4, input\_decay: 1.0, epochs: 250, lr: 0.0001, weight\_decay: 1e-05, dropout: 0, threshold: 1, $\alpha$: 0.01, $\beta$: 0.8\} \\
0.6 & \{num\_layer: 2, input\_decay: 2.5, epochs: 100, lr: 0.001, weight\_decay: 0, dropout: 0, threshold: 1, $\alpha$: 0.05, $\beta$: 0.5\} \\
0.8  & \{num\_layer: 4, input\_decay: 1.0, epochs: 100, lr: 0.001, weight\_decay: 0, dropout: 0.2, threshold: 1, $\alpha$: 0.01, $\beta$: 0.8\} \\
\hline
\end{tabular}
\end{table}

\subsubsection{Target under-representation bias}
\label{assec:urep}
$\;$\\

Number of samples in group $a$: 1000

Number of samples in group $b$: 1000

$br_a = br_b = 0.1$

$\beta_u \in [0.01, 0.05, 0.10, 0.2, 0.4, 0.6, 0.8]$

Outlier counts in groups ($a$, $b$) respectively for varying $\beta_u$: 

\hspace{0.1in}$[(100, 99), (100, 95), (100, 90), (100, 80), (100, 60), (100, 40), (100, 20)]$

Each algorithm selected hyperparameters that yielded the best \parbox[t]{\linewidth}{
\hspace*{0.1in} performance metrics.}

LOF selected hyperparameter $k$ respectively for varying $\beta_u$: 

\hspace{0.1in}$[239, 235, 230, 220, 200, 180, 160]$

DeepAE selected hyperparameters respectively   for varying $\beta_u$:
\vspace{-2mm}
\begin{table}[ht]
\centering
\hspace*{6mm} 
\begin{tabular}{|c|p{7cm}|}  %
\hline
\textbf{$\beta$} & HP \\
\hline
0.01 & \{num\_layer: 4, input\_decay: 2, epochs: 100, lr: 0.001, weight\_decay: 0, dropout: 0, threshold: 1\} \\
0.05 & \{num\_layer: 4, input\_decay: 2, epochs: 100, lr: 0.001, weight\_decay: 0, dropout: 0, threshold: 1\} \\
0.1  & \{num\_layer: 4, input\_decay: 1.0, epochs: 250, lr: 0.0001, weight\_decay: 0, dropout: 0, threshold: 1\} \\
0.2 & \{num\_layer: 4, input\_decay: 2, epochs: 250, lr: 0.001, weight\_decay: 1e-05, dropout: 0, threshold: 1\} \\
0.4  & \{num\_layer: 4, input\_decay: 1.0, epochs: 250, lr: 0.0001, weight\_decay: 1e-05, dropout: 0, threshold: 1\} \\
0.6 & \{num\_layer: 4, input\_decay: 1.0, epochs: 100, lr: 0.001, weight\_decay: 1e-05, dropout: 0.2, threshold: 1\} \\
0.8  & \{num\_layer: 4, input\_decay: 1.0, epochs: 250, lr: 0.0001, weight\_decay: 1e-05, dropout: 0, threshold: 1\} \\
\hline
\end{tabular}
\end{table}

FairOD selected hyperparameters $\alpha$ and $\gamma$ respectively for varying $\beta_u$:
\vspace{-2mm}
\begin{table}[ht]
\centering
\hspace*{6mm} 
\begin{tabular}{|c|p{7cm}|}  %
\hline
\textbf{$\beta$} & HP \\
\hline
0.01 & \{num\_layer: 4, input\_decay: 2, epochs: 100, lr: 0.001, weight\_decay: 1e-05, dropout: 0, threshold: 1\, $\alpha$: 0.01, $\beta$: 0.01\} \\
0.05 & \{num\_layer: 4, input\_decay: 1.0, epochs: 250, lr: 0.0001, weight\_decay: 1e-05, dropout: 0, threshold: 1, $\alpha$: 0.01, $\beta$: 0.2\} \\
0.1  & \{num\_layer: 4, input\_decay: 2.5, epochs: 250, lr: 0.001, weight\_decay: 0, dropout: 0, threshold: 1\, $\alpha$: 0.2, $\beta$: 0.01\} \\
0.2 & \{num\_layer: 4, input\_decay: 1.0, epochs: 250, lr: 0.0001, weight\_decay: 1e-05, dropout: 0, threshold: 1, $\alpha$: 0.8, $\beta$: 0.8\} \\
0.4  & \{num\_layer: 4, input\_decay: 1.0, epochs: 250, lr: 0.0001, weight\_decay: 1e-05, dropout: 0, threshold: 1, $\alpha$: 0.5, $\beta$: 0.01\} \\
0.6 & \{num\_layer: 2, input\_decay: 2.5, epochs: 100, lr: 0.001, weight\_decay: 0, dropout: 0, threshold: 1, $\alpha$: 0.5, $\beta$: 0.01\} \\
0.8  & \{num\_layer: 4, input\_decay: 1.0, epochs: 100, lr: 0.001, weight\_decay: 0, dropout: 0.2, threshold: 1, $\alpha$: 0.05, $\beta$: 0.01\} \\
\hline
\end{tabular}
\end{table}

\newpage

\subsubsection{Feature measurement bias}
\label{assec:meas}
$\;$\\

\subsubsection{Mean-shift}

Number of samples in group $a$: 1000

Number of samples in group $b$: 1000

$br_a = br_b = 0.1$

$\beta_m \in [0, 2, 4, 6, 8]$

Each algorithm selected hyperparameters that yielded the best \parbox[t]{\linewidth}{
\hspace*{0.1in} performance metrics.}

LOF selected hyperparameter $k$ respectively for varying $\beta_m$: 

\hspace{0.1in}$[240, 240, 240, 240, 240]$
\\

\subsubsection{Variance-shift}

Number of samples in group $a$: 1000

Number of samples in group $b$: 1000

$br_a = br_b = 0.1$

$\beta_v \in [0, 0.05, 0.1, 0.2, 0.5, 1, 2, 4, 6]$

Each algorithm selected hyperparameters that yielded the best \parbox[t]{\linewidth}{
\hspace*{0.1in} performance metrics.}

LOF selected hyperparameter $k$ respectively for varying $\beta_v$: 

\hspace{0.1in}$[120, 120, 120, 120, 120, 130, 130, 150, 200]$

DeepAE selected hyperparameters respectively  for varying $\beta_v$:
\vspace{-2mm}
\begin{table}[ht]
\centering
\hspace*{6mm} 
\begin{tabular}{|c|p{7cm}|}  %
\hline
\textbf{$\beta$} & HP \\
\hline
0 & \{num\_layer: 4, input\_decay: 2.5, epochs: 100, lr: 0.001, weight\_decay: 0, dropout: 0, threshold: 1\} \\
0.05 & \{num\_layer: 4, input\_decay: 2, epochs: 100, lr: 0.001, weight\_decay: 0, dropout: 0, threshold: 1\} \\
0.1 & \{num\_layer: 4, input\_decay: 2, epochs: 100, lr: 0.001, weight\_decay: 0, dropout: 0, threshold: 1\} \\
0.2 & \{num\_layer: 2, input\_decay: 1.0, epochs: 250, lr: 0.0001, weight\_decay: 1e-05, dropout: 0, threshold: 1\} \\
0.5 & \{num\_layer: 4, input\_decay: 2, epochs: 100, lr: 0.001, weight\_decay: 0, dropout: 0, threshold: 1\} \\
1 & \{num\_layer: 4, input\_decay: 2, epochs: 100, lr: 0.001, weight\_decay: 0, dropout: 0, threshold: 1\} \\
2 & \{num\_layer: 4, input\_decay: 2, epochs: 100, lr: 0.001, weight\_decay: 0, dropout: 0, threshold: 1\} \\
4 & \{num\_layer: 4, input\_decay: 1.0, epochs: 250, lr: 0.0001, weight\_decay: 1e-05, dropout: 0, threshold: 1\} \\
6 & \{num\_layer: 4, input\_decay: 1.0, epochs: 250, lr: 0.0001, weight\_decay: 1e-05, dropout: 0, threshold: 1\} \\
\hline
\end{tabular}
\end{table}

FairOD selected hyperparameters $\alpha$ and $\gamma$ respectively for varying $\beta_v$:
\vspace{-2mm}
\begin{table}[ht]
\centering
\hspace*{6mm} 
\begin{tabular}{|c|p{7cm}|}  %
\hline
\textbf{$\beta$} & HP \\
\hline
0 & \{num\_layer: 4, input\_decay: 2.5, epochs: 100, lr: 0.001, weight\_decay: 0, dropout: 0, threshold: 1, $\alpha$: 0.01, $\beta$: 0.01\} \\
0.05 & \{num\_layer: 4, input\_decay: 2, epochs: 100, lr: 0.001, weight\_decay: 0, dropout: 0, threshold: 1, $\alpha$: 0.01, $\beta$: 0.01\} \\
0.1 & \{num\_layer: 4, input\_decay: 2, epochs: 100, lr: 0.001, weight\_decay: 0, dropout: 0, threshold: 1, $\alpha$: 0.01, $\beta$: 0.01\} \\
\hline
\end{tabular}
\end{table}

\begin{table}[!ht]
\centering
\hspace*{6mm} 
\begin{tabular}{|c|p{7cm}|}  %
\hline
\textbf{$\beta$} & HP \\
\hline
0.2 & \{num\_layer: 2, input\_decay: 1.0, epochs: 250, lr: 0.0001, weight\_decay: 1e-05, dropout: 0, threshold: 1, $\alpha$: 0.01, $\beta$: 0.01\} \\
0.5 & \{num\_layer: 4, input\_decay: 2, epochs: 100, lr: 0.001, weight\_decay: 0, dropout: 0, threshold: 1, $\alpha$: 0.01, $\beta$: 0.01\} \\
1 & \{num\_layer: 4, input\_decay: 2, epochs: 100, lr: 0.001, weight\_decay: 0, dropout: 0, threshold: 1, $\alpha$: 0.01, $\beta$: 0.01\} \\
2 & \{num\_layer: 4, input\_decay: 2, epochs: 100, lr: 0.001, weight\_decay: 0, dropout: 0, threshold: 1, $\alpha$: 0.01, $\beta$: 0.01\} \\
4 & \{num\_layer: 4, input\_decay: 1.0, epochs: 250, lr: 0.0001, weight\_decay: 1e-05, dropout: 0, threshold: 1, $\alpha$: 0.05, $\beta$: 0.8\} \\
6 & \{num\_layer: 4, input\_decay: 1.0, epochs: 250, lr: 0.0001, weight\_decay: 1e-05, dropout: 0, threshold: 1, $\alpha$: 0.8, $\beta$: 0.2\} \\
\hline
\end{tabular}
\end{table}

\subsubsection{Membership obfuscation bias}
\label{assec:obf}
$\;$\\

Number of samples in group $a$: 1000

Number of samples in group $b$: 1000

$br_a = br_b = 0.1$

$\beta_o \in [0.05, 0.1, 0.15, 0.2, 0.3, 0.4]$

Each algorithm selected hyperparameters that yielded the best \parbox[t]{\linewidth}{
\hspace*{0.1in} performance metrics.}

LOF selected hyperparameter $k$ respectively for varying $\beta_o$: 

\hspace{0.1in}$[240, 240, 240, 240, 240, 240]$

DeepAE selected hyperparameters respectively  for varying $\beta_o$:
\vspace{-2mm}
\begin{table}[ht]
\centering
\hspace*{6mm} 
\begin{tabular}{|c|p{7cm}|}  %
\hline
\textbf{$\beta$} & HP \\
\hline
0 & \{num\_layer: 4, input\_decay: 2, epochs: 100, lr: 0.001, weight\_decay: 1e-05, dropout: 0, threshold: 1\} \\
0.05 & \{num\_layer: 4, input\_decay: 2.5, epochs: 250, lr: 0.001, weight\_decay: 0, dropout: 0, threshold: 1\} \\
0.1 & \{num\_layer: 4, input\_decay: 1.5, epochs: 250, lr: 0.0001, weight\_decay: 1e-05, dropout: 0, threshold: 1\} \\
0.2 & \{num\_layer: 4, input\_decay: 1.5, epochs: 250, lr: 0.0001, weight\_decay: 0, dropout: 0, threshold: 1\} \\
0.3 & \{num\_layer: 2, input\_decay: 2, epochs: 250, lr: 0.0001, weight\_decay: 1e-05, dropout: 0, threshold: 1\} \\
0.4 & \{num\_layer: 4, input\_decay: 1.0, epochs: 250, lr: 0.0001, weight\_decay: 1e-05, dropout: 0, threshold: 1\} \\
\hline
\end{tabular}
\label{tab:optimal_hyperparameters}
\end{table}

FairOD selected hyperparameters $\alpha$ and $\gamma$ respectively for varying $\beta_o$:
\vspace{-2mm}
\begin{table}[ht]
\centering
\hspace*{6mm} 
\begin{tabular}{|c|p{7cm}|}  %
\hline
\textbf{$\beta$} & HP \\
\hline
0 & \{num\_layer: 4, input\_decay: 2, epochs: 100, lr: 0.001, weight\_decay: 1e-05, dropout: 0, threshold: 1, $\alpha$: 0.05, $\beta$: 0.01\} \\
0.05 & \{num\_layer: 4, input\_decay: 2.5, epochs: 250, lr: 0.001, weight\_decay: 0, dropout: 0, threshold: 1, $\alpha$: 0.05, $\beta$: 0.01\} \\
\hline
\end{tabular}
\end{table}

\begin{table}[!ht]
\centering
\hspace*{6mm} 
\begin{tabular}{|c|p{7cm}|}  %
\hline
\textbf{$\beta$} & HP \\
\hline
0.1 & \{num\_layer: 4, input\_decay: 1.5, epochs: 250, lr: 0.0001, weight\_decay: 1e-05, dropout: 0, threshold: 1, $\alpha$: 0.2, $\beta$: 0.01\} \\
0.2 & \{num\_layer: 4, input\_decay: 1.5, epochs: 250, lr: 0.0001, weight\_decay: 0, dropout: 0, threshold: 1, $\alpha$: 0.05, $\beta$: 0.01\} \\
0.3 & \{num\_layer: 2, input\_decay: 2, epochs: 250, lr: 0.0001, weight\_decay: 1e-05, dropout: 0, threshold: 1, $\alpha$: 0.2, $\beta$: 0.01\} \\
0.4 & \{num\_layer: 4, input\_decay: 1.0, epochs: 250, lr: 0.0001, weight\_decay: 1e-05, dropout: 0, threshold: 1, $\alpha$: 0.05, $\beta$: 0.2\} \\
\hline
\end{tabular}
\label{tab:optimal_hyperparameters}
\end{table}

\subsection{Datasets with Scattered Outliers}
\label{asec:scatter_details}

\subsubsection{Group sample size bias}        
\label{assec:sizebias}      

$\;$\\

Number of samples in group $a$: 1000

Number of samples in group $b$: 1000

$br_a = br_b = 0.1$

$\beta_s \in [0.01, 0.05, 0.1, 0.2, 0.4, 0.6, 0.8]$

Outlier counts in groups ($a$, $b$) respectively for varying $\beta_s$: 

\hspace{0.1in}$[(100, 100), (100, 97), (100, 87), (100, 80), (100, 56), (100, 33), (100, 24)]$

Each algorithm selected hyperparameters that yielded the best \parbox[t]{\linewidth}{
\hspace*{0.1in} performance metrics.}

LOF selected hyperparameter $k$ respectively for varying $\beta_s$: 

\hspace{0.1in}$[240, 237, 227, 220, 196, 173, 154]$. 

DeepAE selected hyperparameters respectively  for varying $\beta_s$:
\vspace{-2.6mm}
\begin{table}[ht]
\centering
\hspace*{6mm} 
\begin{tabular}{|c|p{7cm}|}  %
\hline
\textbf{$\beta$} & HP \\
\hline
0.01 & \{num\_layer: 2, input\_decay: 1.5, epochs: 250, lr: 0.001, weight\_decay: 1e-05, dropout: 0, threshold: 1\} \\
0.05 & \{num\_layer: 2, input\_decay: 1.0, epochs: 250, lr: 0.001, weight\_decay: 0, dropout: 0, threshold: 1\} \\
0.1  & \{num\_layer: 2, input\_decay: 1.5, epochs: 250, lr: 0.001, weight\_decay: 1e-05, dropout: 0, threshold: 1\} \\
0.2  & \{num\_layer: 2, input\_decay: 1.0, epochs: 250, lr: 0.001, weight\_decay: 1e-05, dropout: 0, threshold: 1\} \\
0.4  & \{num\_layer: 2, input\_decay: 1.0, epochs: 250, lr: 0.001, weight\_decay: 0, dropout: 0, threshold: 1\} \\
0.6  & \{num\_layer: 2, input\_decay: 1.0, epochs: 250, lr: 0.001, weight\_decay: 1e-05, dropout: 0, threshold: 1\} \\
0.8  & \{num\_layer: 2, input\_decay: 1.0, epochs: 250, lr: 0.001, weight\_decay: 1e-05, dropout: 0, threshold: 1\} \\
\hline
\end{tabular}
\end{table}

FairOD selected hyperparameters $\alpha$ and $\gamma$ respectively for varying $\beta_s$:
\vspace{-2mm}
\begin{table}[ht]
\centering
\hspace*{6mm} 
\begin{tabular}{|c|p{7cm}|}  %
\hline
\textbf{$\beta$} & HP \\
\hline
0.01 & \{num\_layer: 2, input\_decay: 1.5, epochs: 250, lr: 0.001, weight\_decay: 1e-05, dropout: 0, threshold: 1, $\alpha$: 0.5, $\beta$: 0.01\} \\
0.05 & \{num\_layer: 2, input\_decay: 1.0, epochs: 250, lr: 0.001, weight\_decay: 0, dropout: 0, threshold: 1, $\alpha$: 0.01, $\beta$: 0.01\} \\
0.1  & \{num\_layer: 2, input\_decay: 1.5, epochs: 250, lr: 0.001, weight\_decay: 1e-05, dropout: 0, threshold: 1, $\alpha$: 0.01, $\beta$: 0.01\} \\
\hline
\end{tabular}
\end{table}

\begin{table}[!ht]
\centering
\hspace*{6mm} 
\begin{tabular}{|c|p{7cm}|}  %
\hline
\textbf{$\beta$} & HP \\
\hline
0.2  & \{num\_layer: 2, input\_decay: 1.0, epochs: 250, lr: 0.001, weight\_decay: 1e-05, dropout: 0, threshold: 1, $\alpha$: 0.01, $\beta$: 0.01\} \\
0.4  & \{num\_layer: 2, input\_decay: 1.0, epochs: 250, lr: 0.001, weight\_decay: 0, dropout: 0, threshold: 1, $\alpha$: 0.2, $\beta$: 0.01\} \\
0.6  & \{num\_layer: 2, input\_decay: 1.0, epochs: 250, lr: 0.001, weight\_decay: 1e-05, dropout: 0, threshold: 1, $\alpha$: 0.2, $\beta$: 0.01\} \\
0.8  & \{num\_layer: 2, input\_decay: 1.0, epochs: 250, lr: 0.001, weight\_decay: 1e-05, dropout: 0, threshold: 1, $\alpha$: 0.05, $\beta$: 0.01\} \\
\hline
\end{tabular}
\end{table}

\subsubsection{Target under-representation bias}
\label{assec:urep}
$\;$\\

Number of samples in group $a$: 1000

Number of samples in group $b$: 1000

$br_a = br_b = 0.1$

$\beta_u \in [0.01, 0.05, 0.10, 0.2, 0.4, 0.6, 0.8]$

Outlier counts in groups ($a$, $b$) respectively for varying $\beta_u$: 

\hspace{0.1in}$[(100, 99), (100, 95), (100, 90), (100, 80), (100, 60), (100, 40), (100, 20)]$

Each algorithm selected hyperparameters that yielded the best \parbox[t]{\linewidth}{
\hspace*{0.1in} performance metrics.}

LOF selected hyperparameter $k$ respectively for varying $\beta_u$: 

\hspace{0.1in}$[149, 145, 140, 130, 110, 90, 70]$

DeepAE selected hyperparameters respectively  for varying $\beta_u$:
\vspace{-2mm}
\begin{table}[ht]
\centering
\hspace*{6mm} 
\begin{tabular}{|c|p{7cm}|}  %
\hline
\textbf{$\beta$} & HP \\
\hline
0.01 & \{num\_layer: 2, input\_decay: 1.0, epochs: 250, lr: 0.001, weight\_decay: 1e-05, dropout: 0, threshold: 1\} \\
0.05 & \{num\_layer: 2, input\_decay: 1.0, epochs: 250, lr: 0.001, weight\_decay: 0, dropout: 0, threshold: 1\} \\
0.1  & \{num\_layer: 2, input\_decay: 1.5, epochs: 250, lr: 0.001, weight\_decay: 1e-05, dropout: 0, threshold: 1\} \\
0.2  & \{num\_layer: 2, input\_decay: 1.0, epochs: 250, lr: 0.001, weight\_decay: 1e-05, dropout: 0, threshold: 1\} \\
0.4  & \{num\_layer: 2, input\_decay: 1.0, epochs: 250, lr: 0.001, weight\_decay: 0, dropout: 0, threshold: 1\} \\
0.6  & \{num\_layer: 2, input\_decay: 1.5, epochs: 250, lr: 0.001, weight\_decay: 1e-05, dropout: 0, threshold: 1\} \\
0.8  & \{num\_layer: 2, input\_decay: 1.5, epochs: 250, lr: 0.001, weight\_decay: 1e-05, dropout: 0, threshold: 1\} \\
\hline
\end{tabular}
\end{table}

FairOD selected hyperparameters $\alpha$ and $\gamma$ respectively for varying $\beta_u$:
\vspace{-2mm}
\begin{table}[ht]
\centering
\hspace*{6mm} 
\begin{tabular}{|c|p{7cm}|}  %
\hline
\textbf{$\beta$} & HP \\
\hline
0.01 & \{num\_layer: 2, input\_decay: 1.0, epochs: 250, lr: 0.001, weight\_decay: 1e-05, dropout: 0, threshold: 1, $\alpha$: 0.5, $\beta$: 0.01\} \\
0.05 & \{num\_layer: 2, input\_decay: 1.0, epochs: 250, lr: 0.001, weight\_decay: 0, dropout: 0, threshold: 1, $\alpha$: 0.05, $\beta$: 0.01\} \\
0.1  & \{num\_layer: 2, input\_decay: 1.5, epochs: 250, lr: 0.001, weight\_decay: 1e-05, dropout: 0, threshold: 1, $\alpha$: 0.01, $\beta$: 0.5\} \\
\hline
\end{tabular}
\end{table}

\begin{table}[!ht]
\centering
\hspace*{6mm} 
\begin{tabular}{|c|p{7cm}|}  %
\hline
\textbf{$\beta$} & HP \\
\hline
0.2  & \{num\_layer: 2, input\_decay: 1.0, epochs: 250, lr: 0.001, weight\_decay: 1e-05, dropout: 0, threshold: 1, $\alpha$: 0.05, $\beta$: 0.01\} \\
0.4  & \{num\_layer: 2, input\_decay: 1.0, epochs: 250, lr: 0.001, weight\_decay: 0, dropout: 0, threshold: 1, $\alpha$: 0.5, $\beta$: 0.5\} \\
0.6  & \{num\_layer: 2, input\_decay: 1.5, epochs: 250, lr: 0.001, weight\_decay: 1e-05, dropout: 0, threshold: 1, $\alpha$: 0.01, $\beta$: 0.01\} \\
0.8  & \{num\_layer: 2, input\_decay: 1.5, epochs: 250, lr: 0.001, weight\_decay: 1e-05, dropout: 0, threshold: 1\, $\alpha$: 0.8, $\beta$: 0.8\} \\
\hline
\end{tabular}
\end{table}

\subsubsection{Feature measurement bias}
\label{assec:meas}
$\;$\\

\subsubsection{Mean-shift}

Number of samples in group $a$: 1000

Number of samples in group $b$: 1000

$br_a = br_b = 0.1$

$\beta_m \in [0, 2, 4, 6, 8]$

Each algorithm selected hyperparameters that yielded the best \parbox[t]{\linewidth}{
\hspace*{0.1in} performance metrics.}

LOF selected hyperparameter $k$ respectively for varying $\beta_m$: 

\hspace{0.1in}$[20, 10, 40, 30, 10]$
\\

\subsubsection{Variance-shift}

Number of samples in group $a$: 1000

Number of samples in group $b$: 1000

$br_a = br_b = 0.1$

$\beta_v \in  [0, 0.05, 0.1, 0.2, 0.5, 1, 1.5, 2, 3]
$

Each algorithm selected hyperparameters that yielded the best \parbox[t]{\linewidth}{
\hspace*{0.1in} performance metrics.}

LOF selected hyperparameter $k$ respectively for varying $\beta_v$: 

\hspace{0.1in}$[10, 10, 30, 30, 70, 30, 90, 70, 140]$

DeepAE selected hyperparameters respectively  for varying $\beta_v$:
\vspace{-2mm}
\begin{table}[ht]
\centering
\hspace*{6mm} 
\begin{tabular}{|c|p{7cm}|}  %
\hline
\textbf{$\beta$} & HP \\
\hline
0 & \{num\_layer: 2, input\_decay: 1.0, epochs: 250, lr: 0.001, weight\_decay: 0, dropout: 0, threshold: 1\} \\
0.05 & \{num\_layer: 2, input\_decay: 1.0, epochs: 250, lr: 0.001, weight\_decay: 0, dropout: 0, threshold: 1\} \\
0.1  & \{num\_layer: 2, input\_decay: 1.5, epochs: 250, lr: 0.001, weight\_decay: 1e-05, dropout: 0, threshold: 1\} \\
0.2  & \{num\_layer: 2, input\_decay: 1.5, epochs: 250, lr: 0.001, weight\_decay: 0, dropout: 0, threshold: 1\} \\
0.5  & \{num\_layer: 2, input\_decay: 1.5, epochs: 250, lr: 0.001, weight\_decay: 1e-05, dropout: 0, threshold: 1\} \\
1  & \{num\_layer: 2, input\_decay: 1.5, epochs: 250, lr: 0.001, weight\_decay: 0, dropout: 0, threshold: 1\} \\
1.5  & \{num\_layer: 2, input\_decay: 1.0, epochs: 250, lr: 0.001, weight\_decay: 0, dropout: 0, threshold: 1\} \\
2  & \{num\_layer: 2, input\_decay: 1.5, epochs: 250, lr: 0.001, weight\_decay: 1e-05, dropout: 0, threshold: 1\} \\
3  & \{num\_layer: 2, input\_decay: 1.5, epochs: 250, lr: 0.001, weight\_decay: 0, dropout: 0, threshold: 1\} \\
\hline
\end{tabular}
\end{table}

FairOD selected hyperparameters $\alpha$ and $\gamma$ respectively for varying $\beta_v$:
\vspace{-2mm}
\begin{table}[ht]
\centering
\hspace*{6mm} 
\begin{tabular}{|c|p{7cm}|}  %
\hline
\textbf{$\beta$} & HP \\
\hline
0 & \{num\_layer: 2, input\_decay: 1.0, epochs: 250, lr: 0.001, weight\_decay: 0, dropout: 0, threshold: 1, $\alpha$: 0.8, $\beta$: 0.01\} \\
0.05 & \{num\_layer: 2, input\_decay: 1.0, epochs: 250, lr: 0.001, weight\_decay: 0, dropout: 0, threshold: 1, $\alpha$: 0.01, $\beta$: 0.01\} \\
0.1  & \{num\_layer: 2, input\_decay: 1.5, epochs: 250, lr: 0.001, weight\_decay: 1e-05, dropout: 0, threshold: 1, $\alpha$: 0.01, $\beta$: 0.01\} \\
0.2  & \{num\_layer: 2, input\_decay: 1.5, epochs: 250, lr: 0.001, weight\_decay: 0, dropout: 0, threshold: 1, $\alpha$: 0.01, $\beta$: 0.5\} \\
0.5  & \{num\_layer: 2, input\_decay: 1.5, epochs: 250, lr: 0.001, weight\_decay: 1e-05, dropout: 0, threshold: 1, $\alpha$: 0.05, $\beta$: 0.01\} \\
1  & \{num\_layer: 2, input\_decay: 1.5, epochs: 250, lr: 0.001, weight\_decay: 0, dropout: 0, threshold: 1, $\alpha$: 0.01, $\beta$: 0.01\} \\
1.5  & \{num\_layer: 2, input\_decay: 1.0, epochs: 250, lr: 0.001, weight\_decay: 0, dropout: 0, threshold: 1, $\alpha$: 0.5, $\beta$: 0.01\} \\
2  & \{num\_layer: 2, input\_decay: 1.5, epochs: 250, lr: 0.001, weight\_decay: 1e-05, dropout: 0, threshold: 1, $\alpha$: 0.01, $\beta$: 0.01\} \\
3  & \{num\_layer: 2, input\_decay: 1.5, epochs: 250, lr: 0.001, weight\_decay: 0, dropout: 0, threshold: 1, $\alpha$: 0.01, $\beta$: 0.01\} \\
\hline
\end{tabular}
\end{table}

\subsubsection{Membership obfuscation bias}
\label{assec:obf}
$\;$\\

Number of samples in group $a$: 1000

Number of samples in group $b$: 1000

$br_a = br_b = 0.1$

$\beta_o \in [0.05, 0.1, 0.15, 0.2, 0.3, 0.4]$

Each algorithm selected hyperparameters that yielded the best \parbox[t]{\linewidth}{
\hspace*{0.1in} performance metrics.}

LOF selected hyperparameter $k$ respectively for varying $\beta_o$: 

\hspace{0.1in}$[10, 20, 10, 10, 20, 10]$

DeepAE selected hyperparameters respectively  for varying $\beta_o$:
\vspace{-2mm}
\begin{table}[ht]
\centering
\hspace*{6mm} 
\begin{tabular}{|c|p{7cm}|}  %
\hline
\textbf{$\beta$} & HP \\
\hline
0.05 & \{num\_layer: 2, input\_decay: 1.5, epochs: 250, lr: 0.001, weight\_decay: 1e-05, dropout: 0, threshold: 1\} \\
0.1 & \{num\_layer: 2, input\_decay: 1.5, epochs: 250, lr: 0.001, weight\_decay: 1e-05, dropout: 0, threshold: 1\} \\
0.15 & \{num\_layer: 2, input\_decay: 1.0, epochs: 250, lr: 0.001, weight\_decay: 1e-05, dropout: 0, threshold: 1\} \\
0.2 & \{num\_layer: 2, input\_decay: 1.0, epochs: 250, lr: 0.001, weight\_decay: 1e-05, dropout: 0, threshold: 1\} \\
0.3 & \{num\_layer: 2, input\_decay: 1.0, epochs: 250, lr: 0.001, weight\_decay: 1e-05, dropout: 0, threshold: 1\} \\
0.4 & \{num\_layer: 2, input\_decay: 1.0, epochs: 250, lr: 0.001, weight\_decay: 1e-05, dropout: 0, threshold: 1\} \\
\hline
\end{tabular}
\end{table}

\newpage
FairOD selected hyperparameters $\alpha$ and $\gamma$ respectively for varying $\beta_o$:
\vspace{-2mm}
\begin{table}[ht]
\centering
\hspace*{6mm} 
\begin{tabular}{|c|p{7cm}|}  %
\hline
\textbf{$\beta$} & HP \\
\hline
0.05 & \{num\_layer: 2, input\_decay: 1.5, epochs: 250, lr: 0.001, weight\_decay: 1e-05, dropout: 0, threshold: 1, $\alpha$: 0.01, $\beta$: 0.01\} \\
0.1 & \{num\_layer: 2, input\_decay: 1.5, epochs: 250, lr: 0.001, weight\_decay: 1e-05, dropout: 0, threshold: 1, $\alpha$: 0.01, $\beta$: 0.01\} \\
0.15 & \{num\_layer: 2, input\_decay: 1.0, epochs: 250, lr: 0.001, weight\_decay: 1e-05, dropout: 0, threshold: 1, $\alpha$: 0.5, $\beta$: 0.01\} \\
0.2 & \{num\_layer: 2, input\_decay: 1.0, epochs: 250, lr: 0.001, weight\_decay: 1e-05, dropout: 0, threshold: 1, $\alpha$: 0.8, $\beta$: 0.01\} \\
0.3 & \{num\_layer: 2, input\_decay: 1.0, epochs: 250, lr: 0.001, weight\_decay: 1e-05, dropout: 0, threshold: 1, $\alpha$: 0.8, $\beta$: 0.01\} \\
0.4 & \{num\_layer: 2, input\_decay: 1.0, epochs: 250, lr: 0.001, weight\_decay: 1e-05, dropout: 0, threshold: 1, $\alpha$: 0.8, $\beta$: 0.01\} \\
\hline
\end{tabular}
\end{table}
$\;$\\

\section{Hyperparameter Configurations}
\label{asec:configs}

\vspace{0.1in}
\textbf{LOF:~} We tune the number of nearest neighbors, $k$, over the candidate set $k\in \{10, 20, ..., 250\}$ and pick the best w.r.t. overall AUROC detection performance. 

\vspace{0.1in}
\noindent
\textbf{Isolation Forest:~} As an ensemble model of extremely randomized trees, iForest has two hyperparameters (HPs); number of trees and number of samples to construct each tree. As with most ensemble models, it has been found to be robust under a wide range of choices for these HPs. Therefore, we use the default values of 100 trees and 256 samples per tree. 

\vspace{0.1in}
\noindent
\textbf{DeepAE:~} We tune the 6 hyperparameters, picking the best among 128 candidate configurations as shown in Table \ref{tab:ae_hps} w.r.t. overall AUROC detection performance.
\begin{table}[h]
\centering
\caption{Hyperparameter search space for DeepAE}
\vspace{-0.1in}
\begin{tabular}{lll}
\toprule
\textbf{Name} & \textbf{Count} & \textbf{Values}\\
\midrule
num\_layers    & 2 & {[}2, 4{]}             \\
weight\_decay & 2 & {[}0, 1e-5{]}          \\
learning\_rate            & 2 & {[}1e-3, 1e-4{]}       \\
num\_epochs        & 2 & {[}100, 250{]}         \\
input\_decay  & 4 & {[}1.0, 1.5, 2, 2.5{]} \\
droupout\_rate     & 2 & {[}0, 0.2{]}    \\
\bottomrule
\end{tabular}
\label{tab:ae_hps}
\end{table}

\vspace{0.1in}
\noindent
\textbf{FairOD:~} Utilizing a deep autoencoder as its base model, FairOD additionally optimizes a fairness-enhanced loss; specifically with two additional terms besides the reconstruction term. Each additional term has a scalar hyperparameter associated with it, $\alpha$ and $\gamma$ respectively, to trade-off performance/fit vs. two fairness notions (statistical parity and a heuristic approximation of equality of opportunity). 
On any given dataset, we use the same best DeepAE configuration for FairOD and only tune these two trade-off HPs in the loss. Specifically, we tune over $\alpha \in \{ 0.01, 0.05, 0.2, 0.5, 0.8\}$ and $\gamma \in \{ 0.01, 0.2, 0.5, 0.8\}$, and pick the configuration that yields the best (AUROC, statistical parity, equality of opportunity) outcome/triplet, with shortest distance to the ideal values of (1, 1, 1) for these three quantities.

\section{Additional Empirical Results}
\label{asec:exp_resuls}

\subsection{Datasets with Clustered Outliers}
\label{asec:cluster_results}

\subsubsection{Group sample size bias}
\label{assec:sizebias_cluster}

The results are presented in the main text in Fig. \ref{fig:samplesize_all_cluster}.

\subsubsection{Target under-representation bias}
\label{assec:urep_cluster}

\begin{figure*}[!ht]
    \centering
    \begin{tabular}{lllll}
    \hspace{-0.1in}{\includegraphics[width=0.195\textwidth]{FIG/underrep/lof/cluster/fr_legend.pdf}} 
    &
    \hspace{-0.1in}{\includegraphics[width=0.195\textwidth]{FIG/underrep/lof/cluster/tpr.pdf}} &
     \hspace{-0.1in}{\includegraphics[width=0.195\textwidth]{FIG/underrep/lof/cluster/fpr.pdf}}
    &
   \hspace{-0.1in} {\includegraphics[width=0.195\textwidth]{FIG/underrep/lof/cluster/prec.pdf}}
   &
   \hspace{-0.1in}
   {\includegraphics[width=0.195\textwidth]{FIG/underrep/lof/cluster/auroc.pdf}} 
\\
\hspace{-0.1in}{\includegraphics[width=0.195\textwidth]{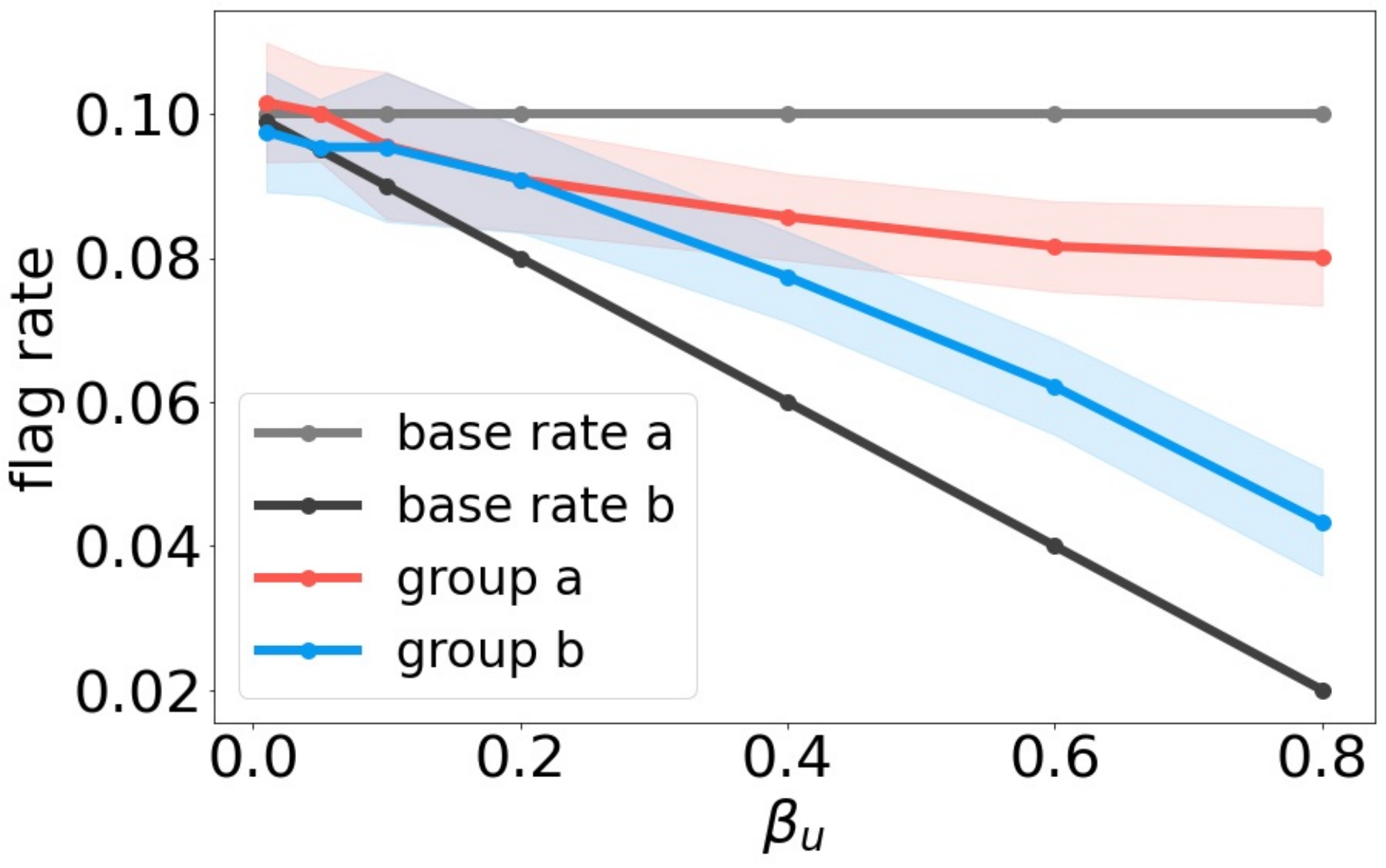}} 
    &
    \hspace{-0.1in}{\includegraphics[width=0.195\textwidth]{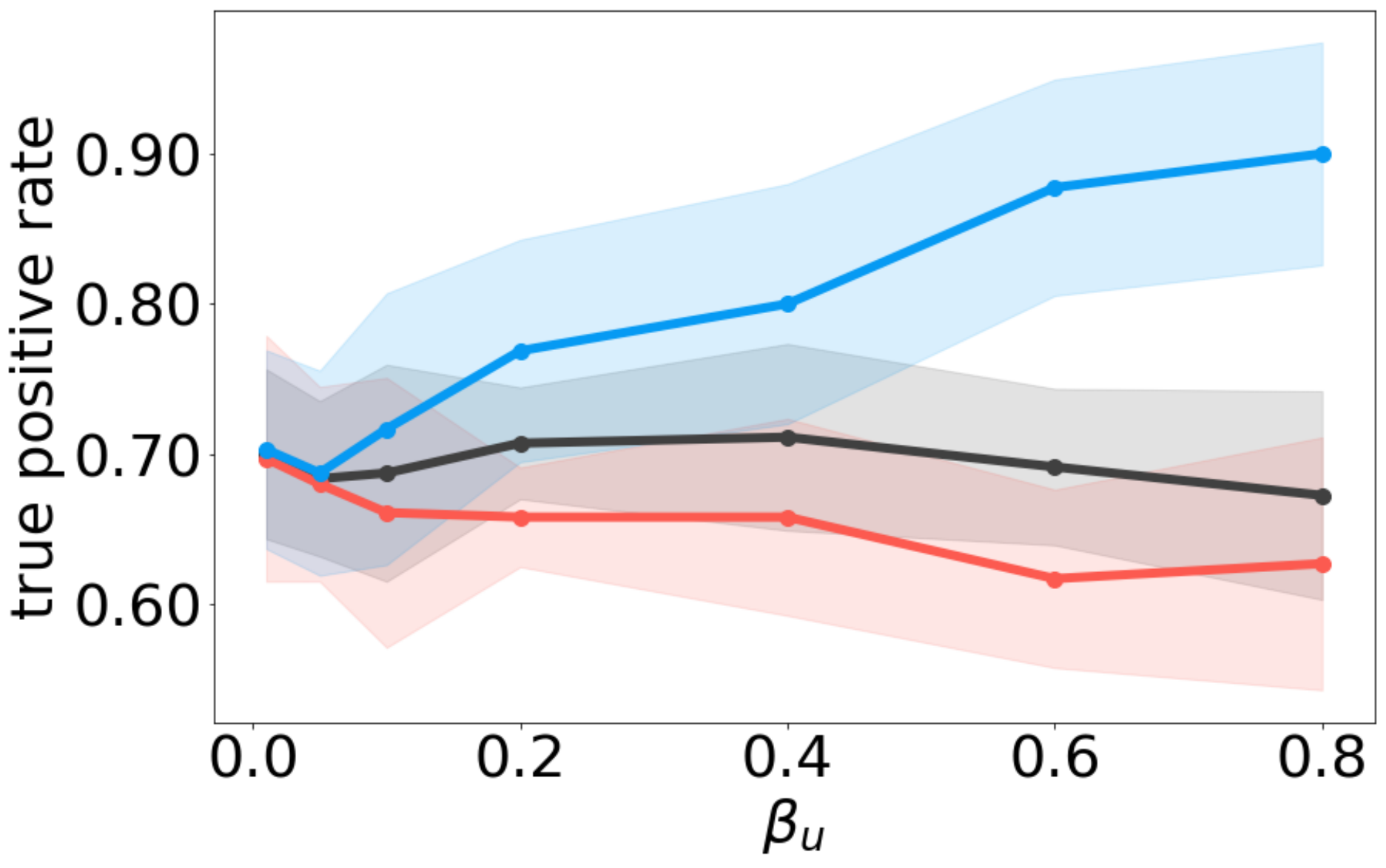}} &
     \hspace{-0.1in}{\includegraphics[width=0.195\textwidth]{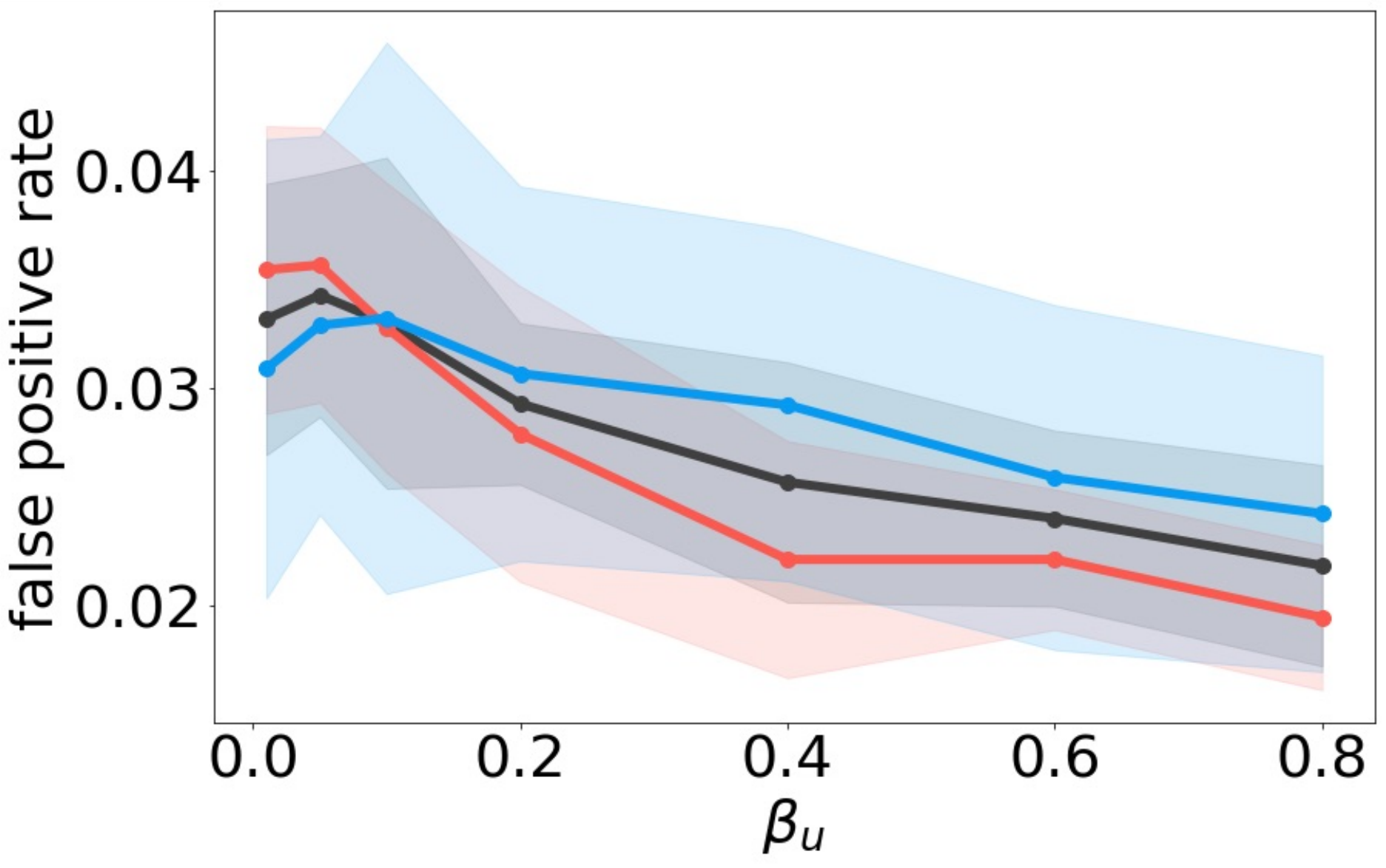}}
    &
   \hspace{-0.1in} {\includegraphics[width=0.195\textwidth]{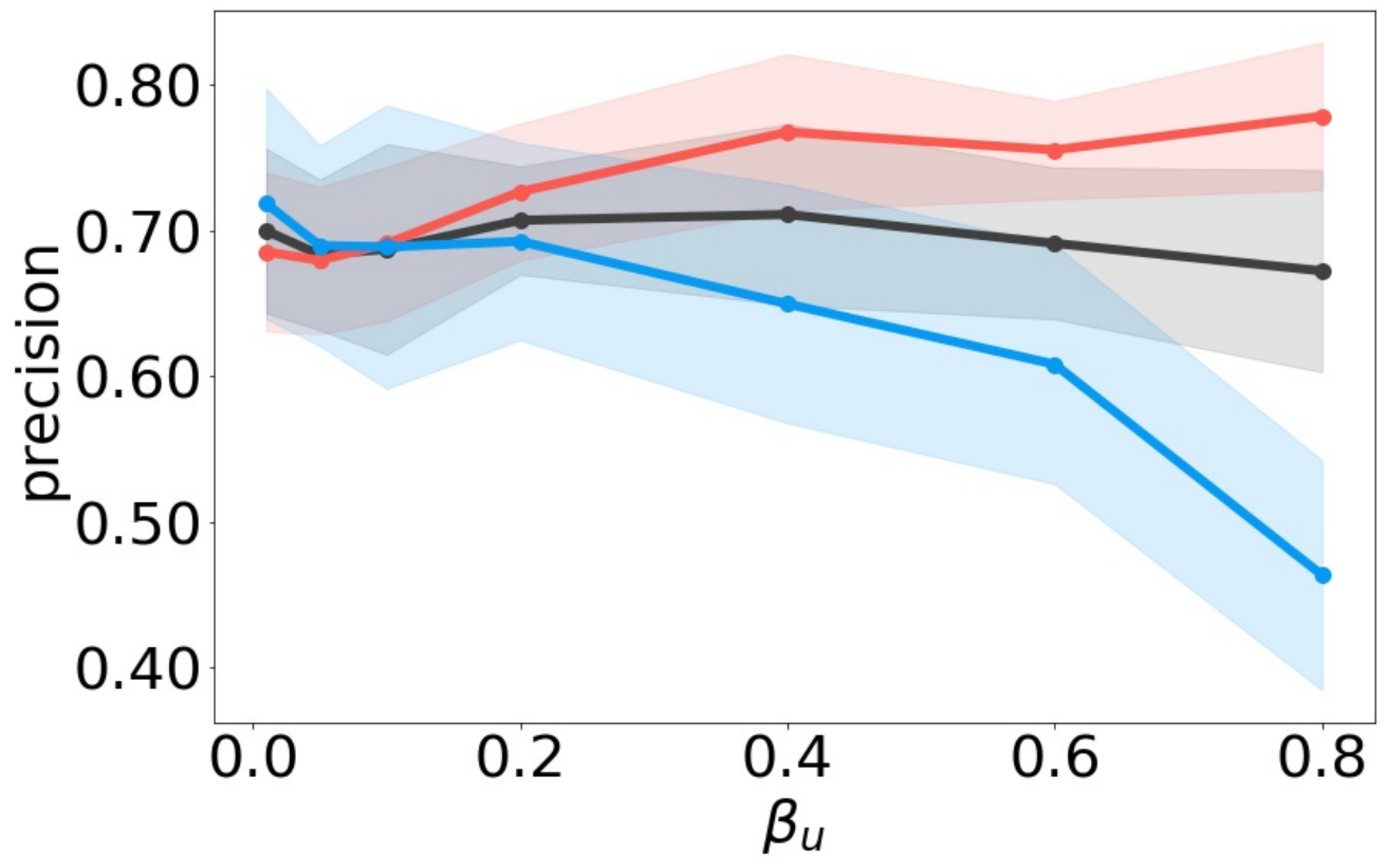}}
   &
    \hspace{-0.1in}{\includegraphics[width=0.195\textwidth]{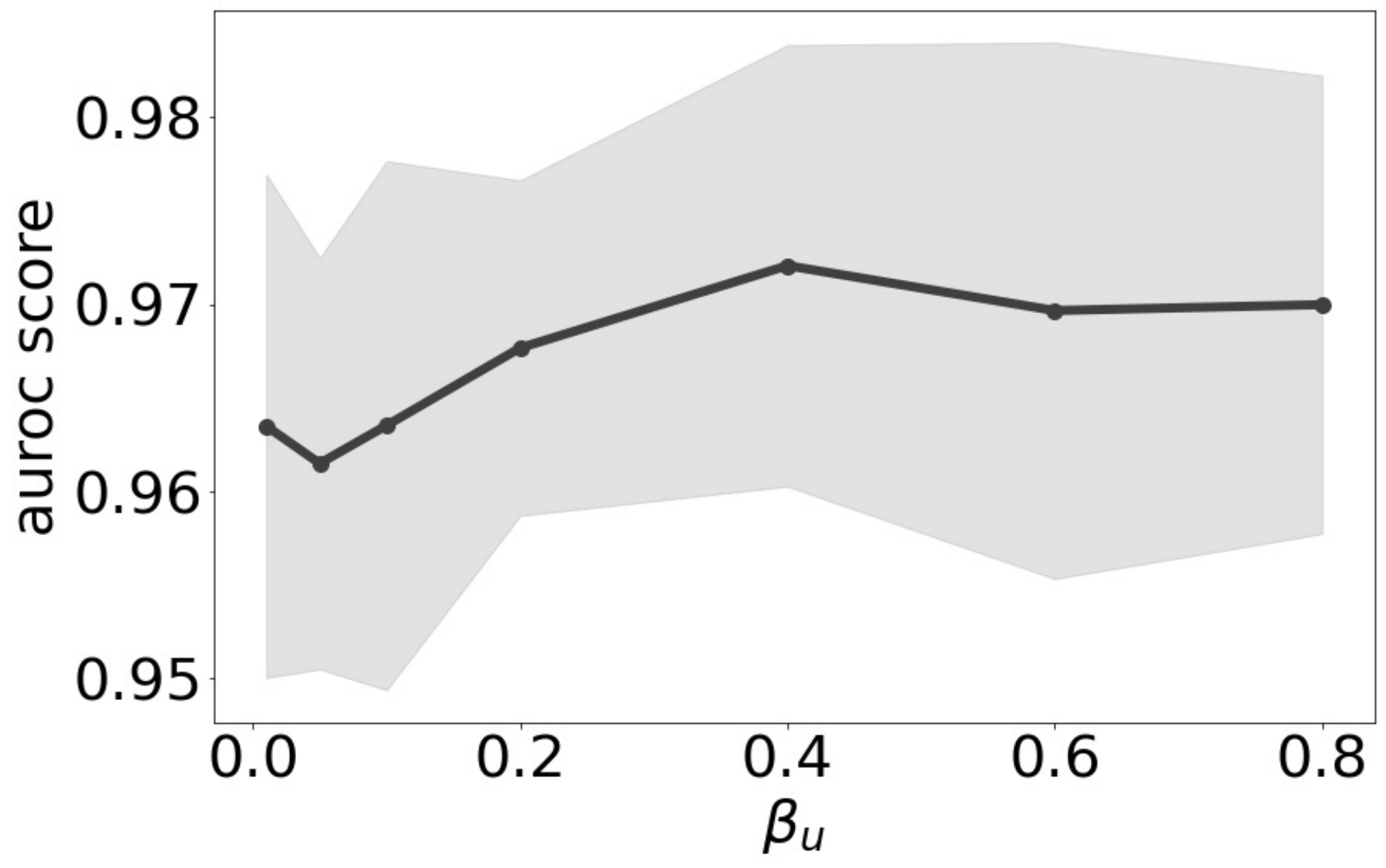}}
   \\
     \hspace{-0.1in}{\includegraphics[width=0.195\textwidth]{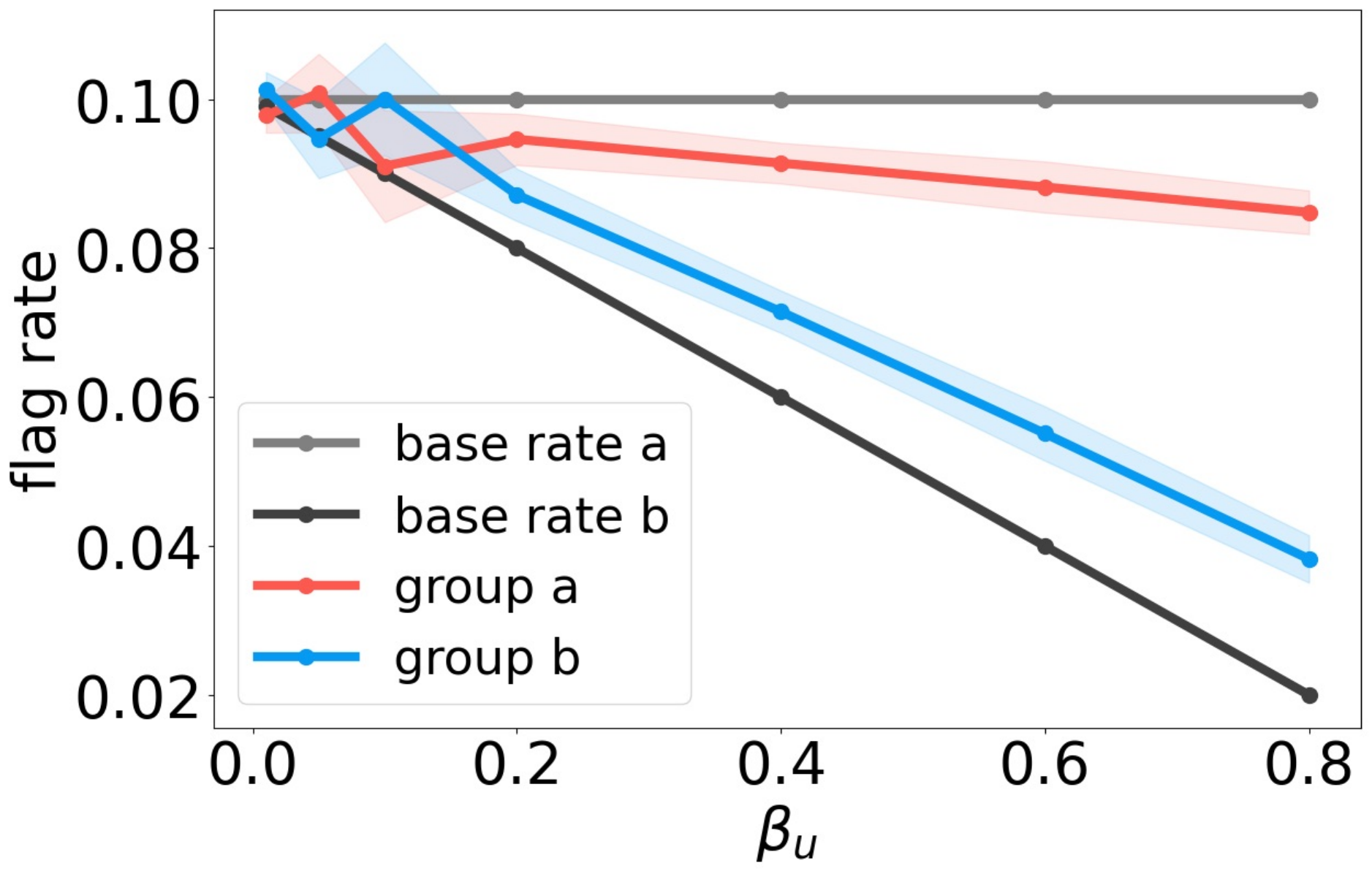}} 
    &
    \hspace{-0.1in}{\includegraphics[width=0.195\textwidth]{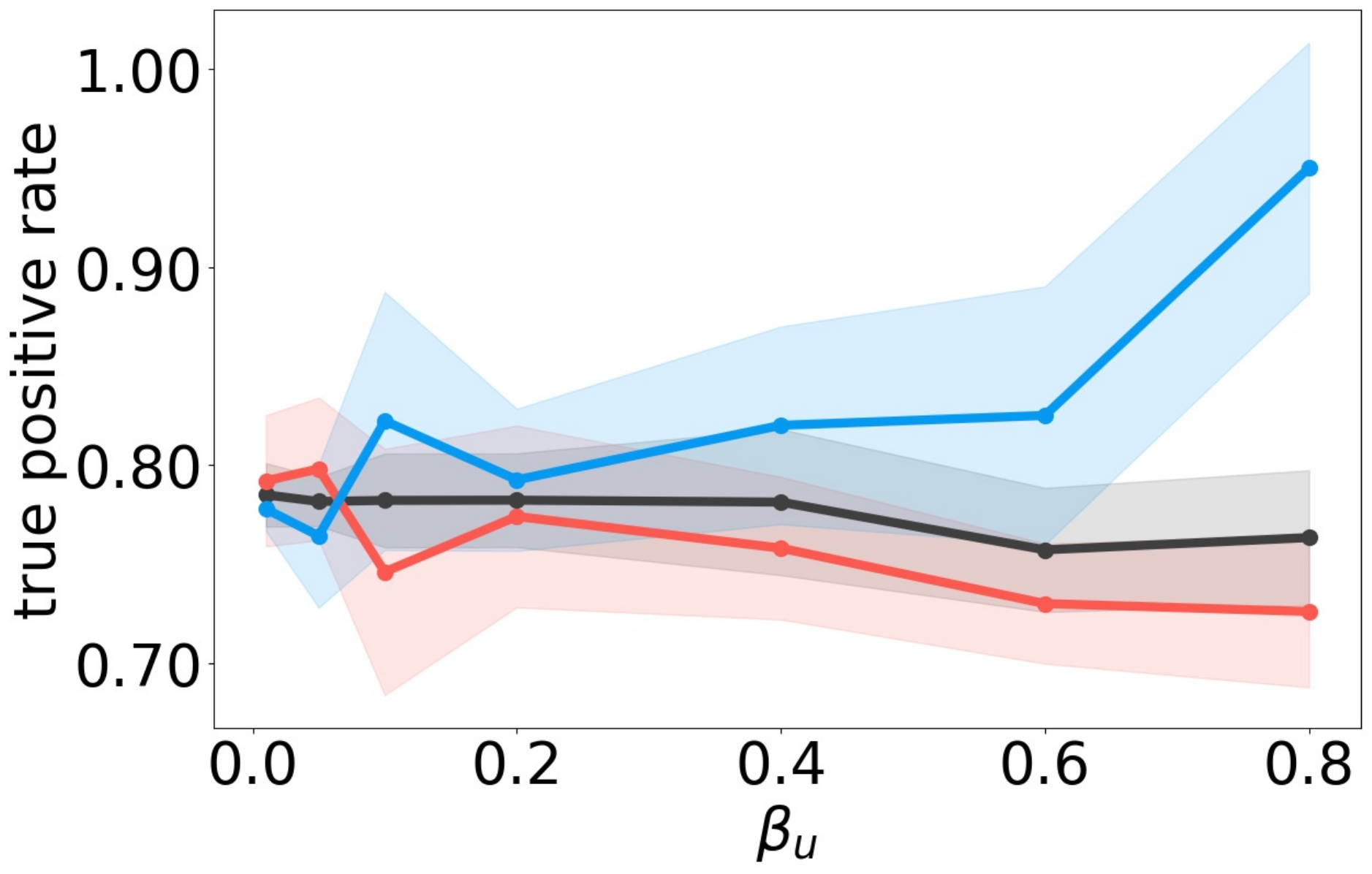}} &
     \hspace{-0.1in}{\includegraphics[width=0.195\textwidth]{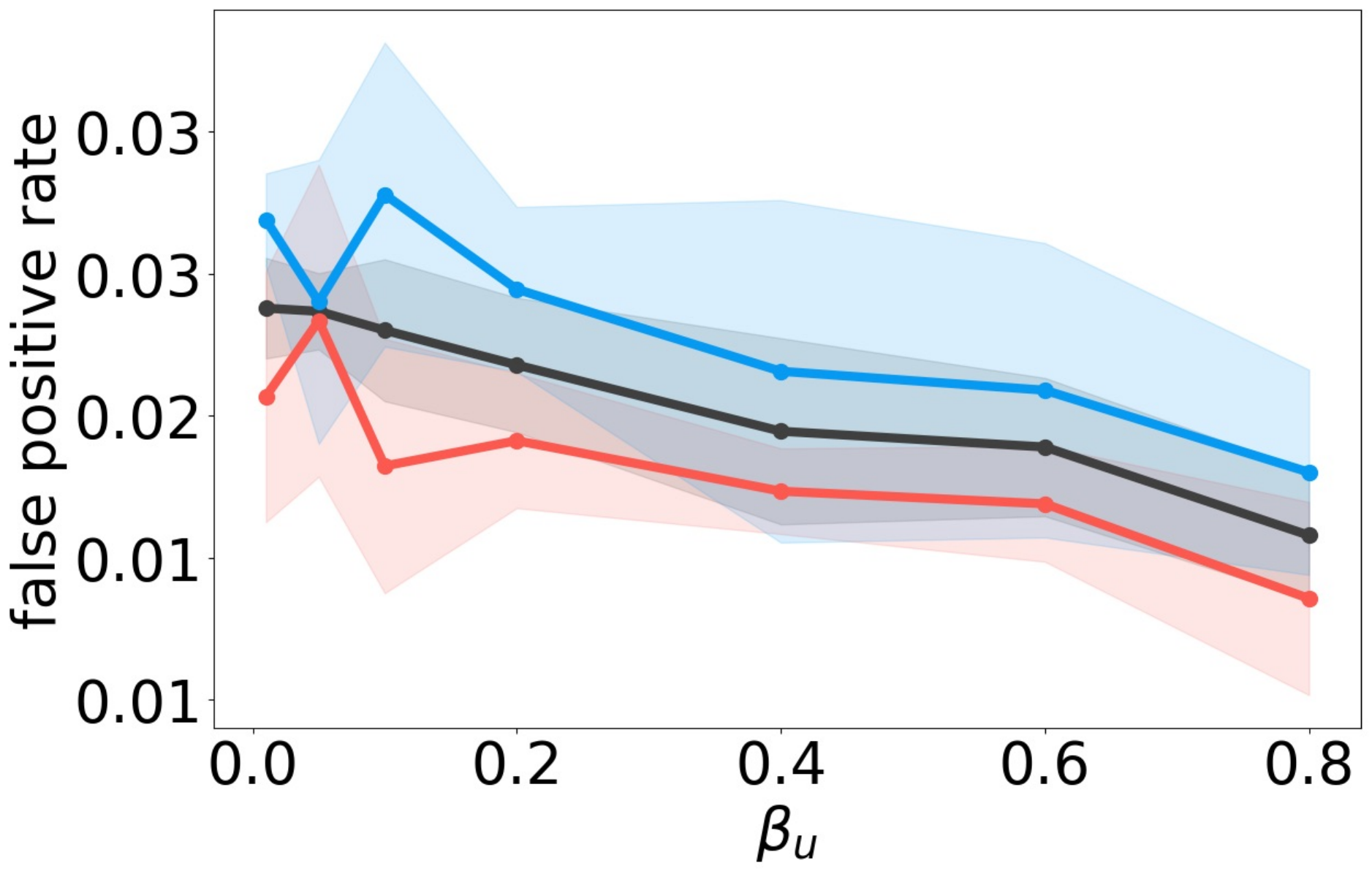}}
    &
   \hspace{-0.1in} {\includegraphics[width=0.195\textwidth]{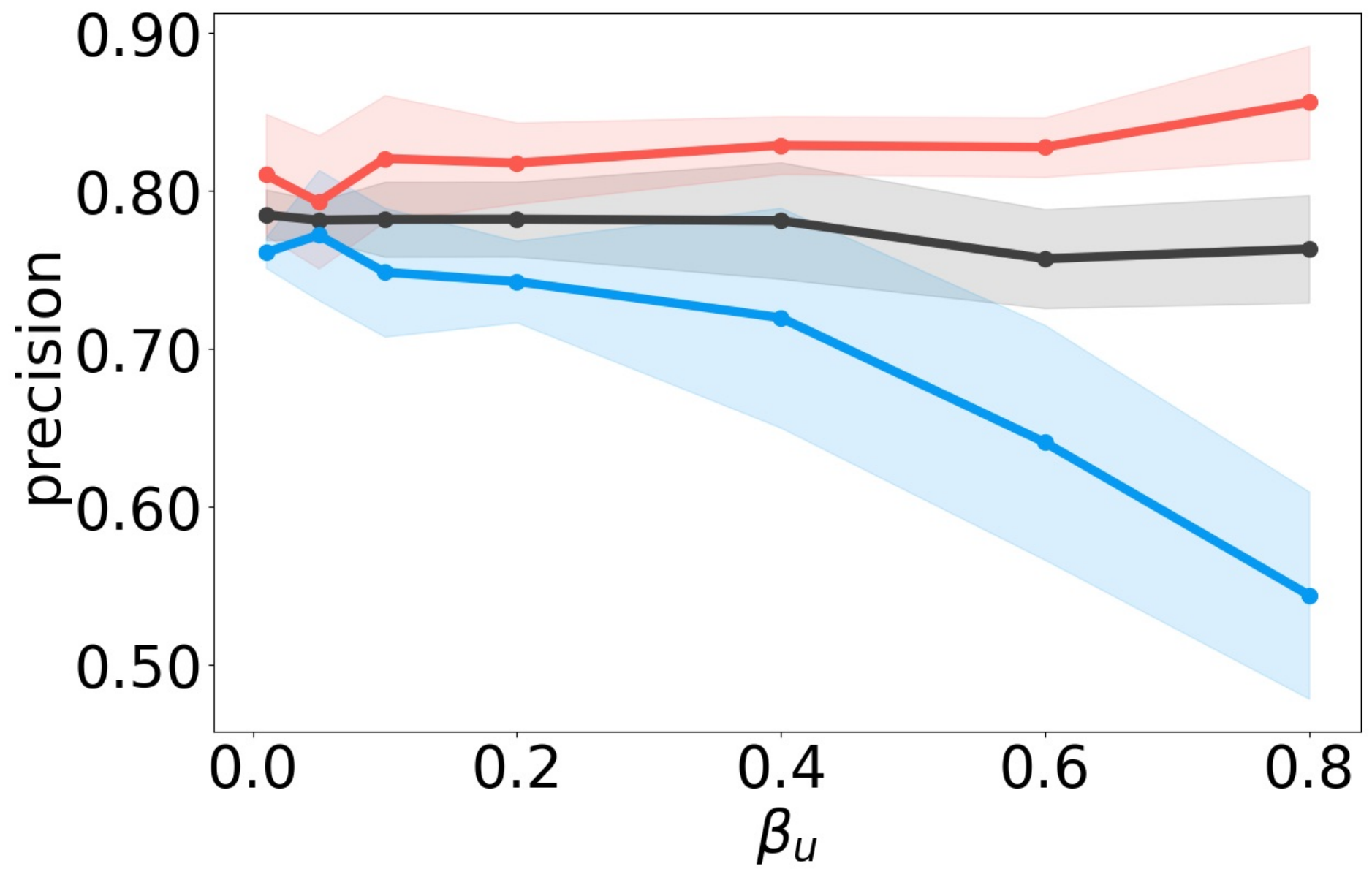}}
   &
     \hspace{-0.1in}{\includegraphics[width=0.195\textwidth]{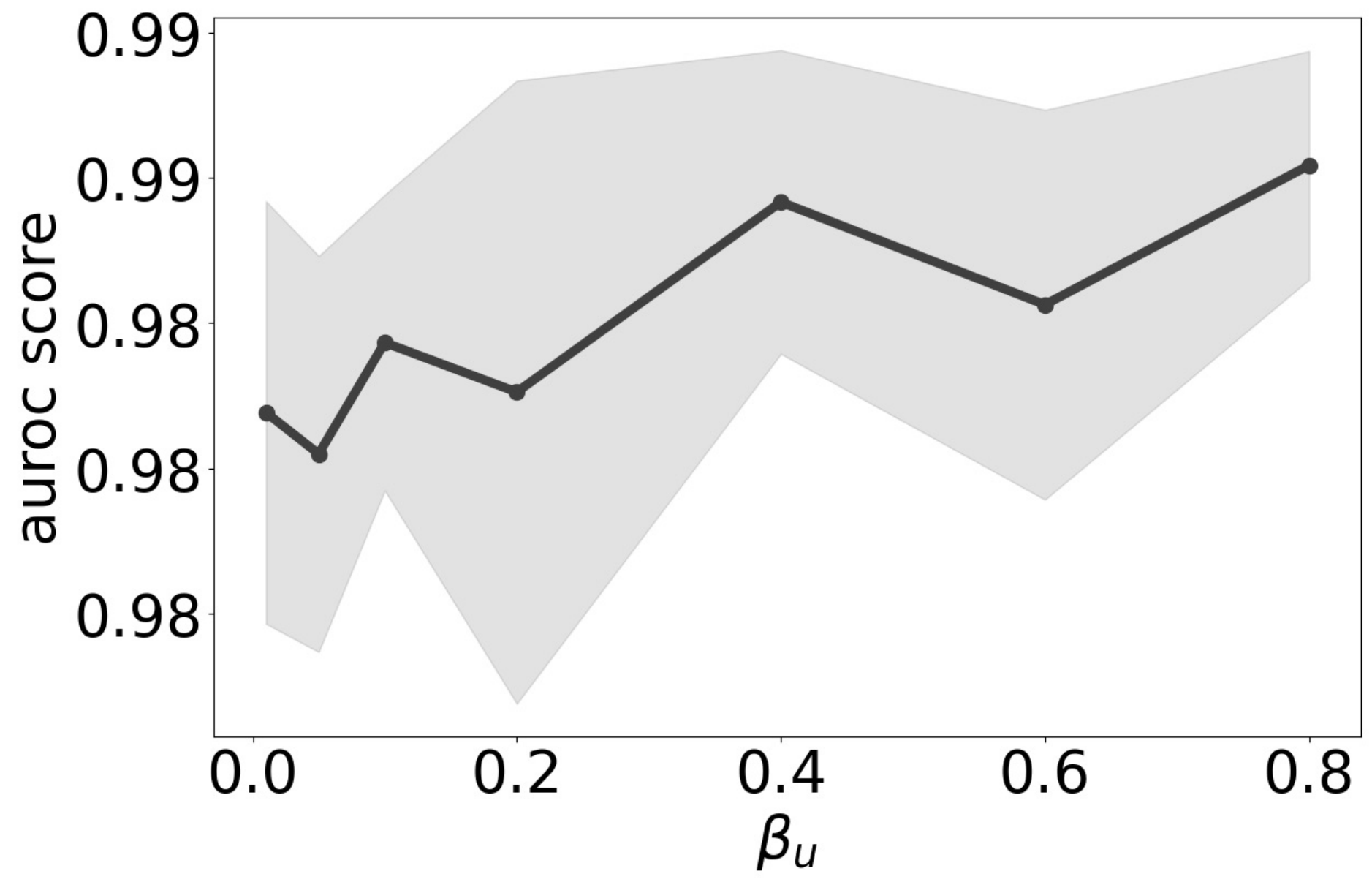}}
   \\
   \hspace{-0.1in}{\includegraphics[width=0.195\textwidth]{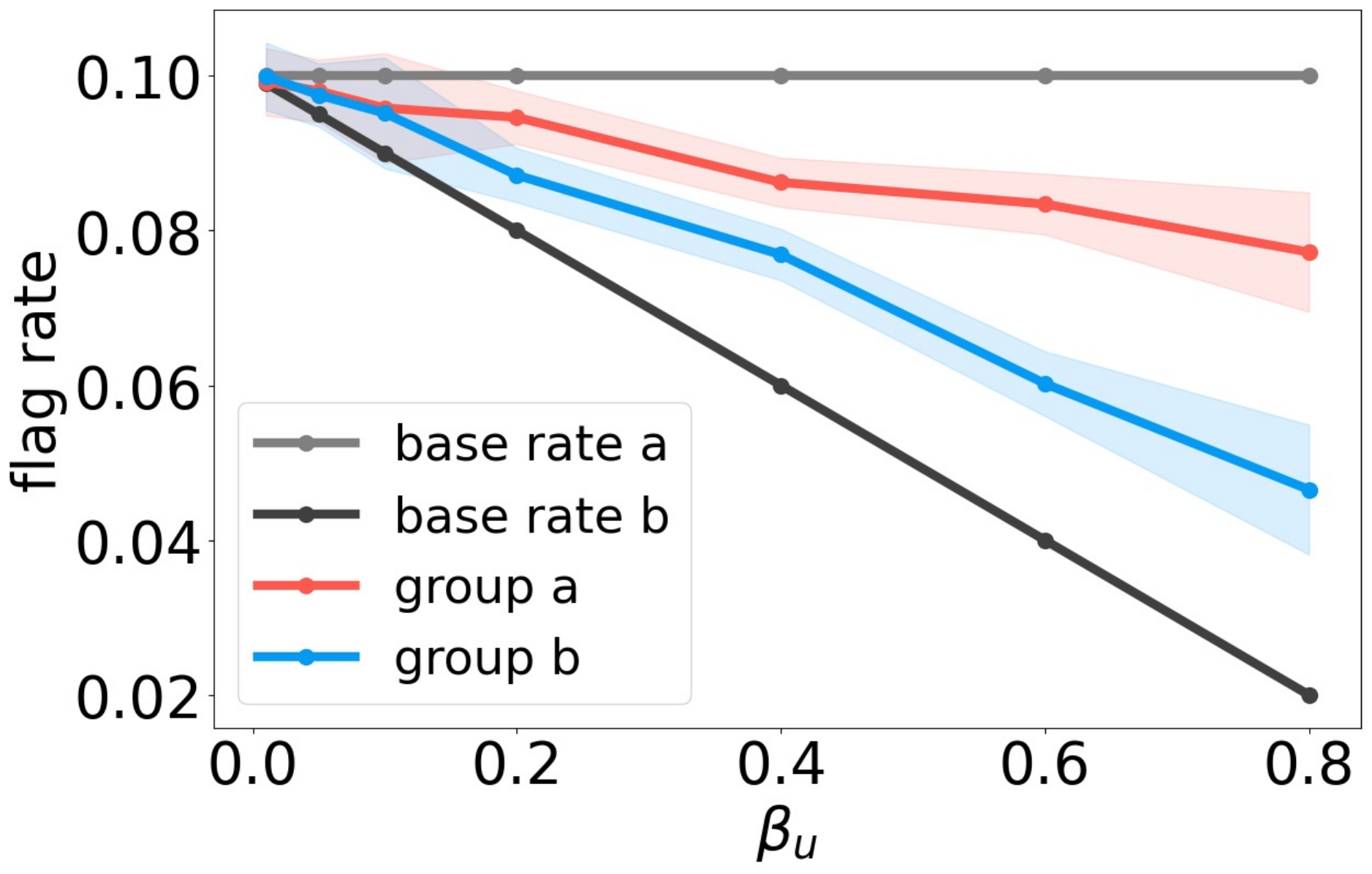} 
   } &
    \hspace{-0.1in}{\includegraphics[width=0.195\textwidth]{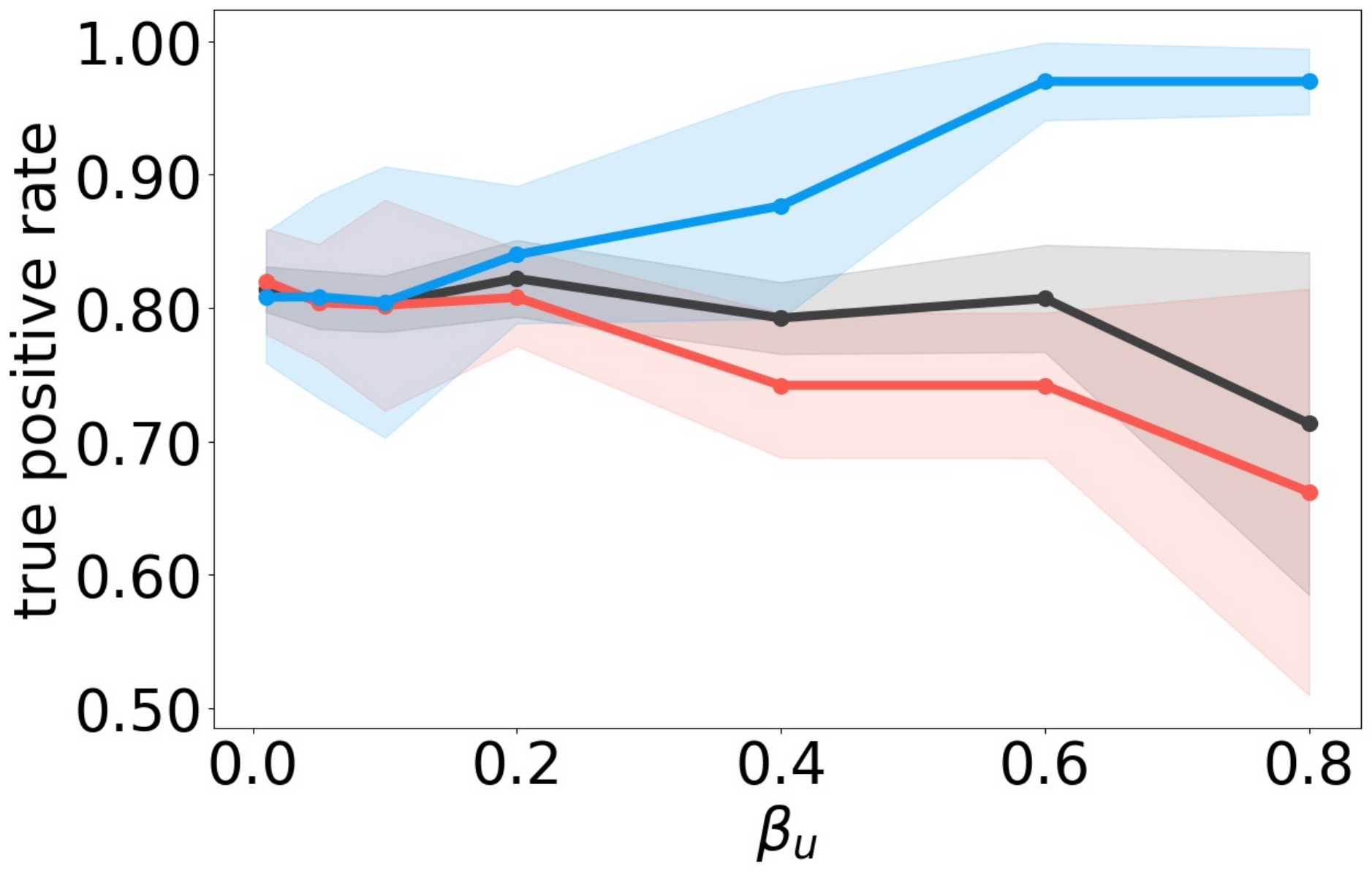}} &
     \hspace{-0.1in}{\includegraphics[width=0.195\textwidth]{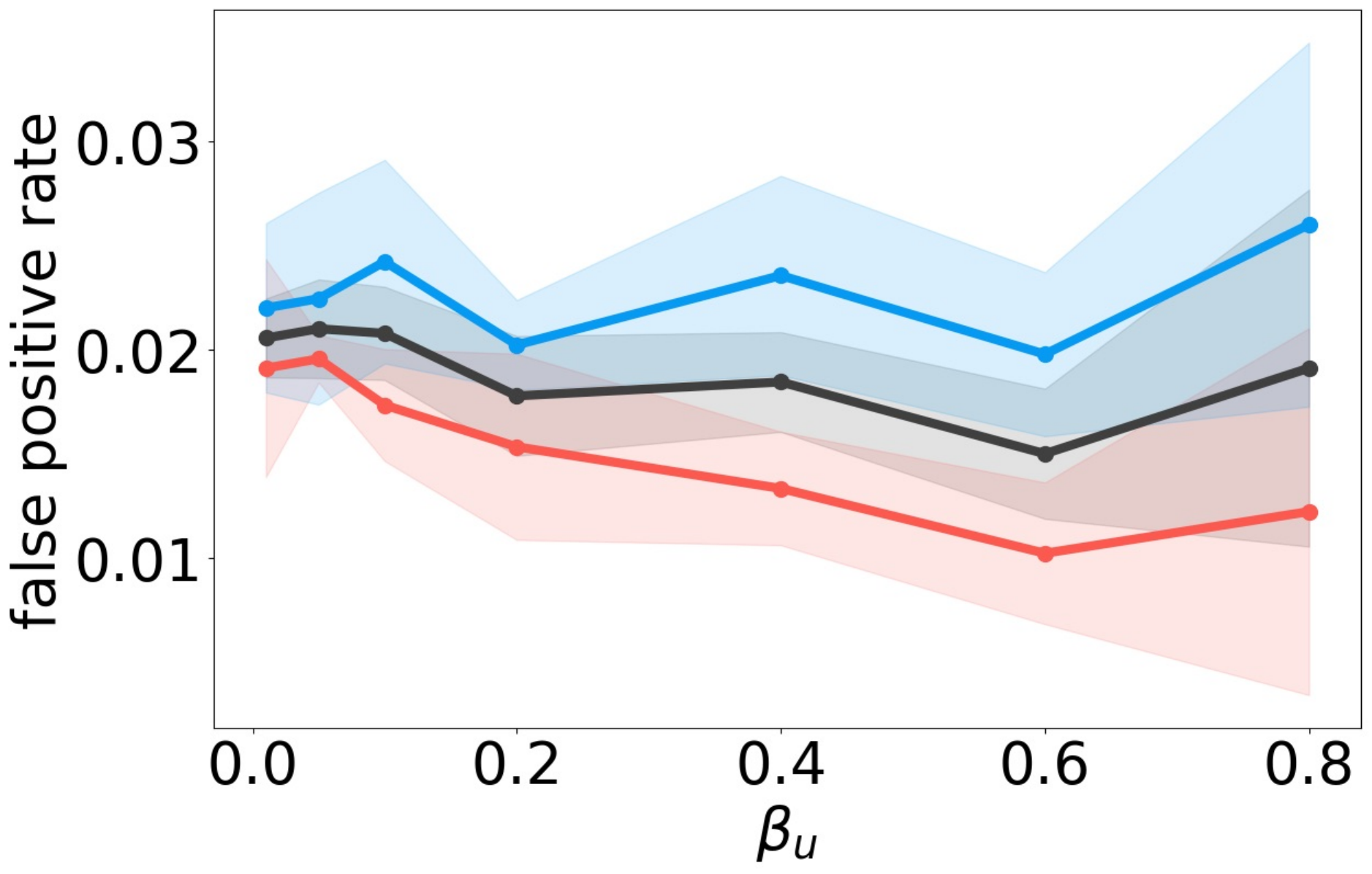}}
    &
   \hspace{-0.1in} {\includegraphics[width=0.195\textwidth]{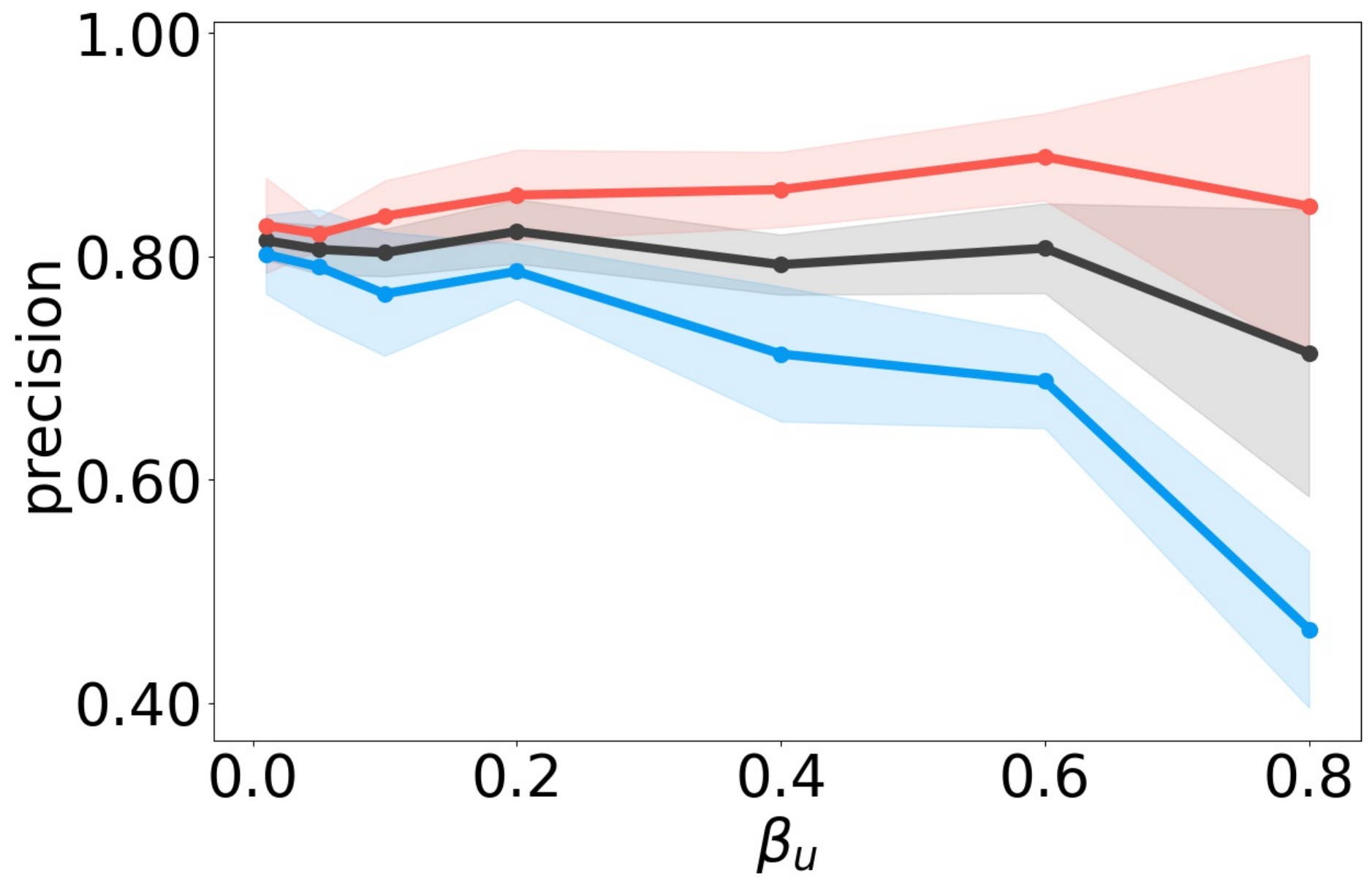}}
   &
   \hspace{-0.1in} {\includegraphics[width=0.195\textwidth]{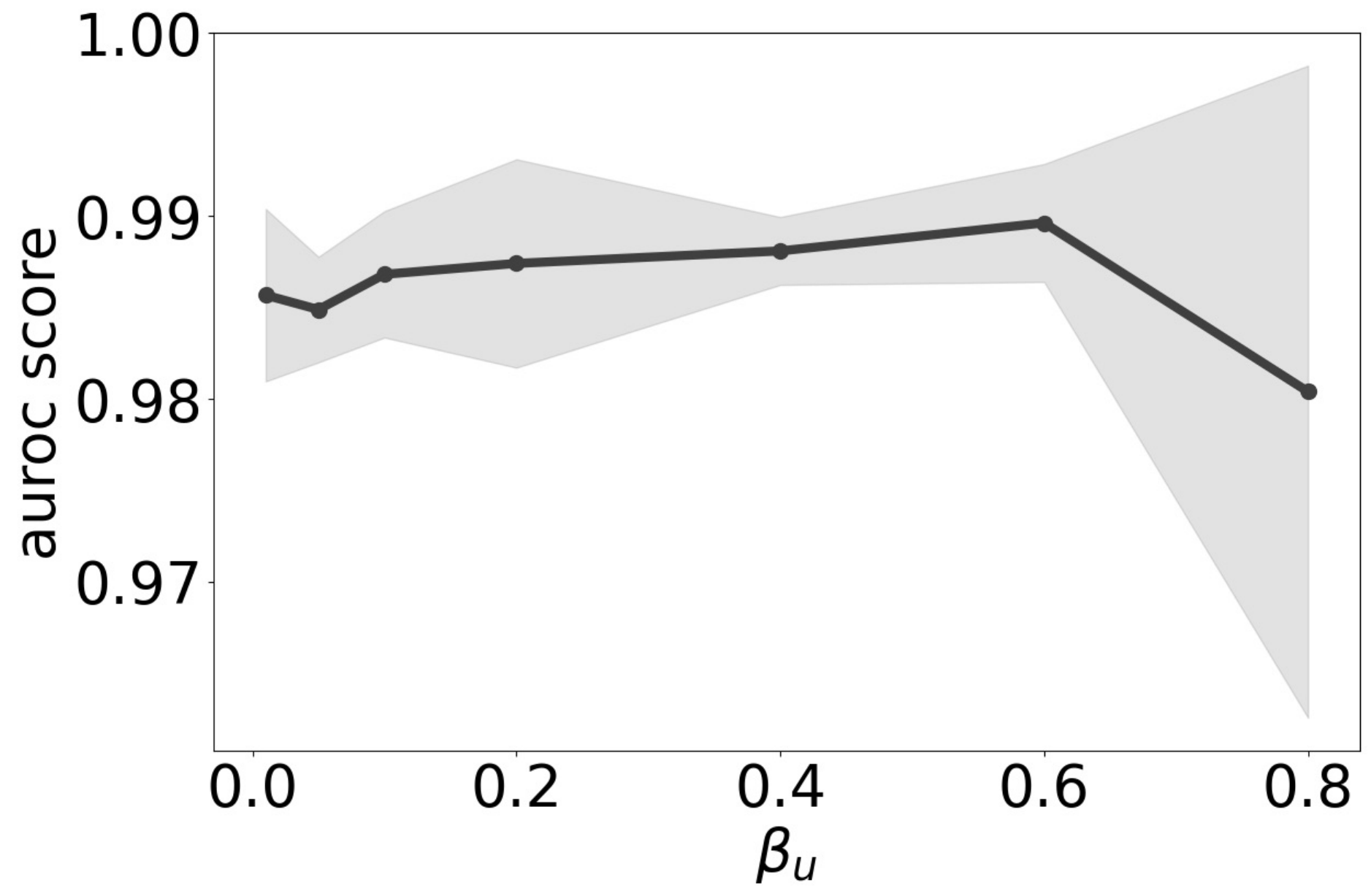}}
    \end{tabular}
    \caption{(best in color) Group-wise fairness metrics and AUROC for (top to bottom) LOF, iForest, DeepAE and FairOD under \underline{target under-representation} on \textit{clustered} outliers. }
    \label{fig:underrep_all_cluster}
\end{figure*}

See Fig \ref{fig:underrep_all_cluster}.

\subsubsection{Feature measurement bias}
\label{assec:meas_cluster}

\begin{figure*}[!ht]
    \centering
    \begin{tabular}{lllll}
    \hspace{-0.1in}{\includegraphics[width=0.195\textwidth]{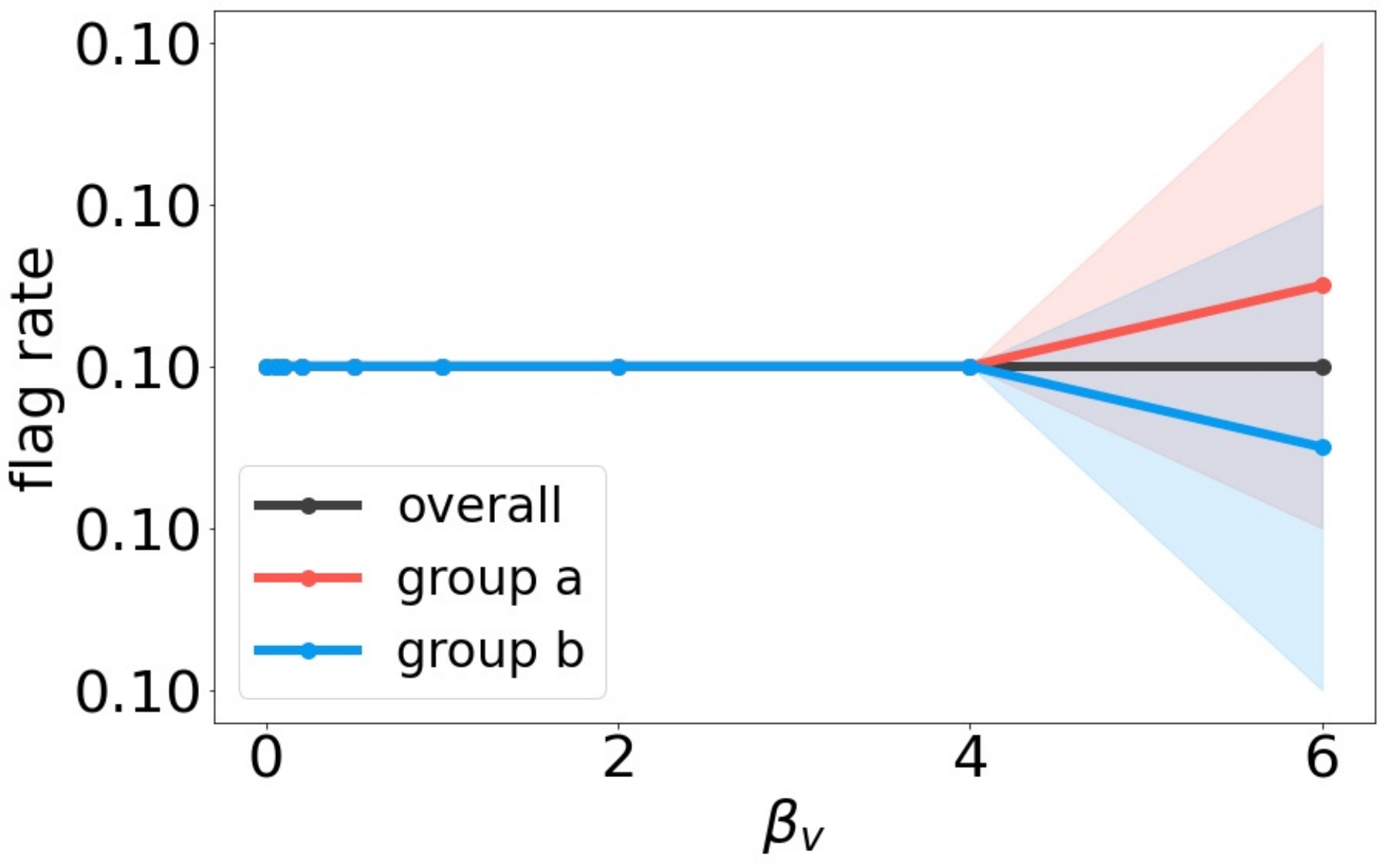}} 
    &
    \hspace{-0.1in}{\includegraphics[width=0.195\textwidth]{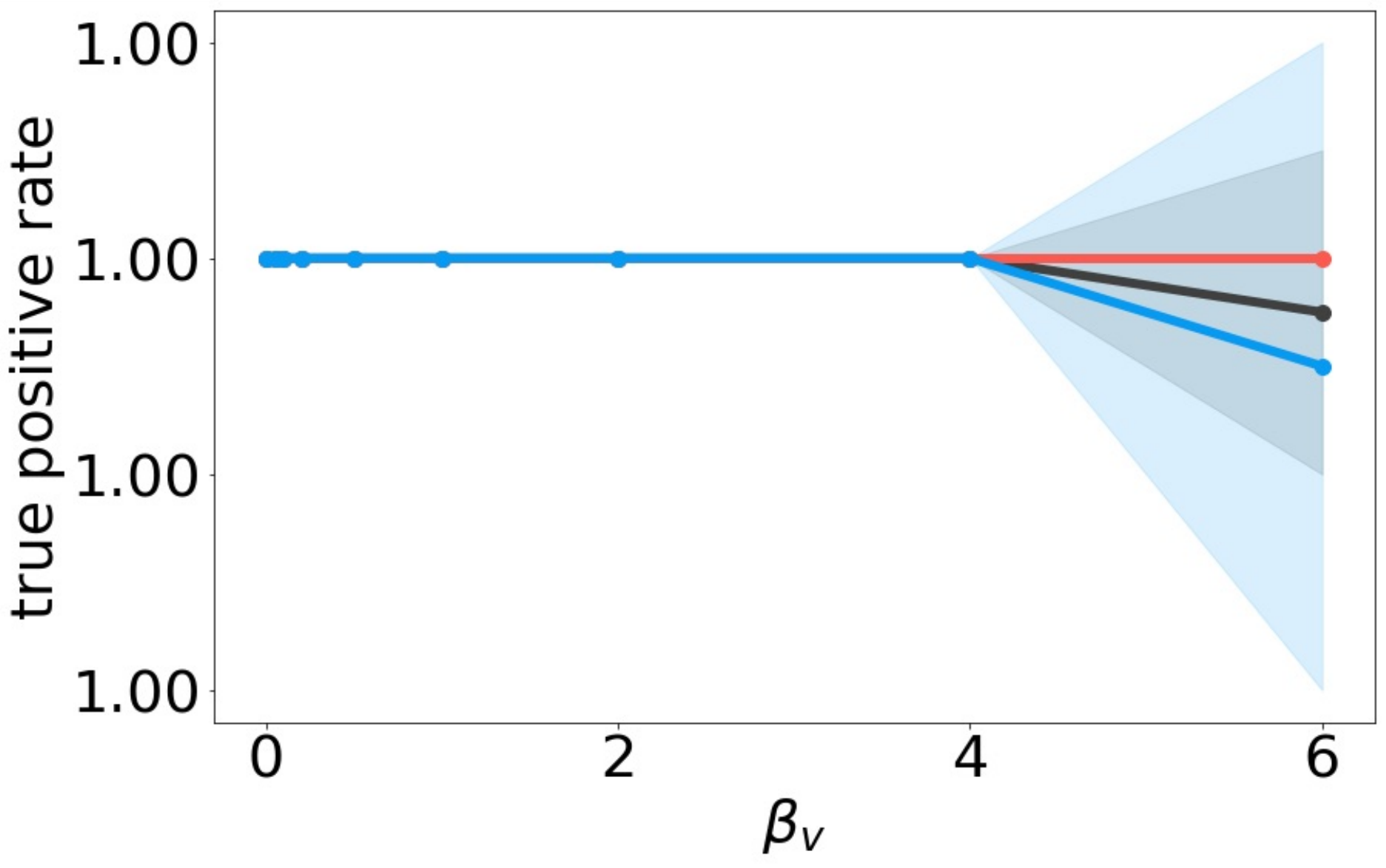}} &
     \hspace{-0.1in}{\includegraphics[width=0.195\textwidth]{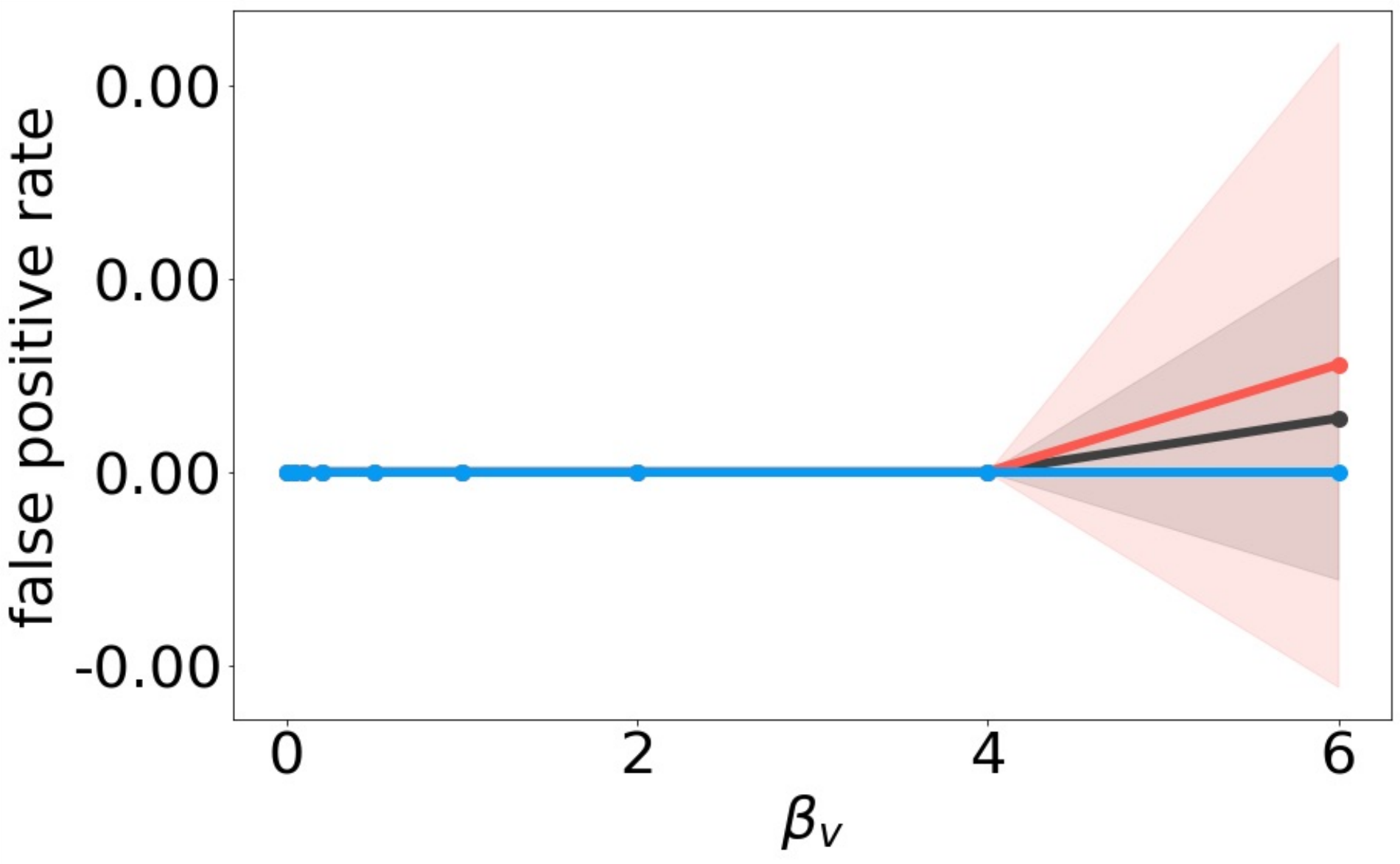}}
    &
   \hspace{-0.1in} {\includegraphics[width=0.195\textwidth]{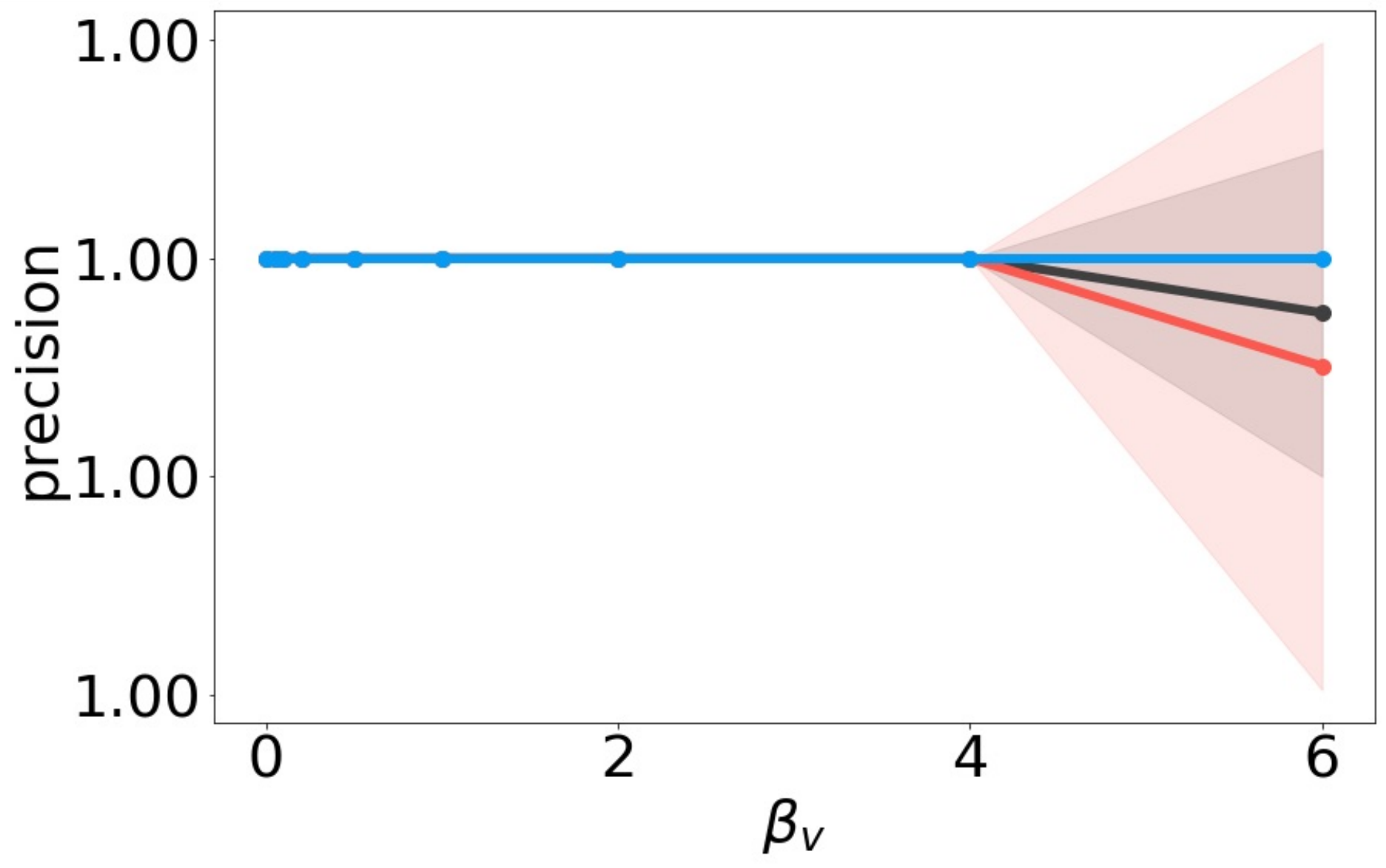}}
   &
   \hspace{-0.1in}
   {\includegraphics[width=0.195\textwidth]{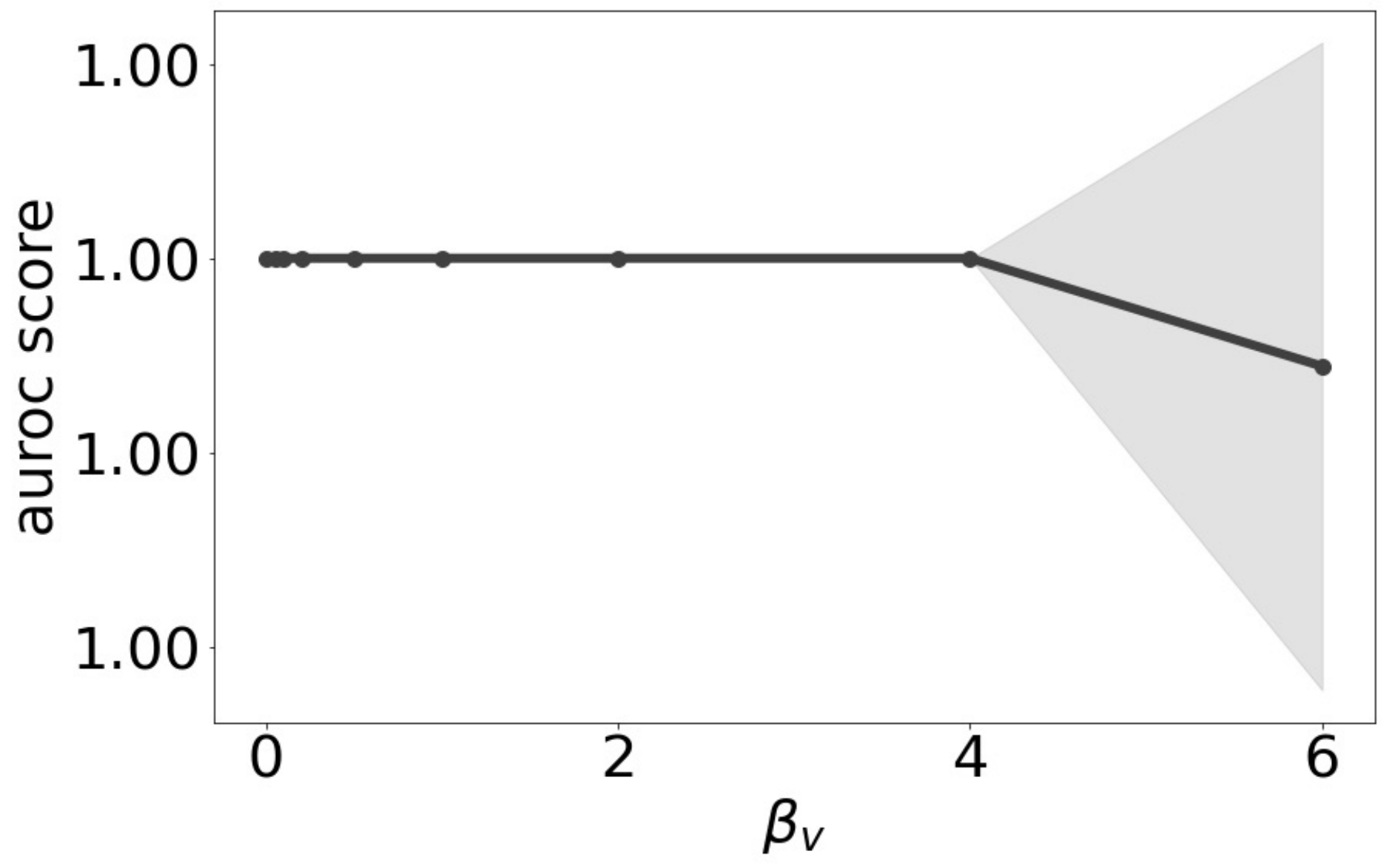}} 
\\
\hspace{-0.1in}{\includegraphics[width=0.195\textwidth]{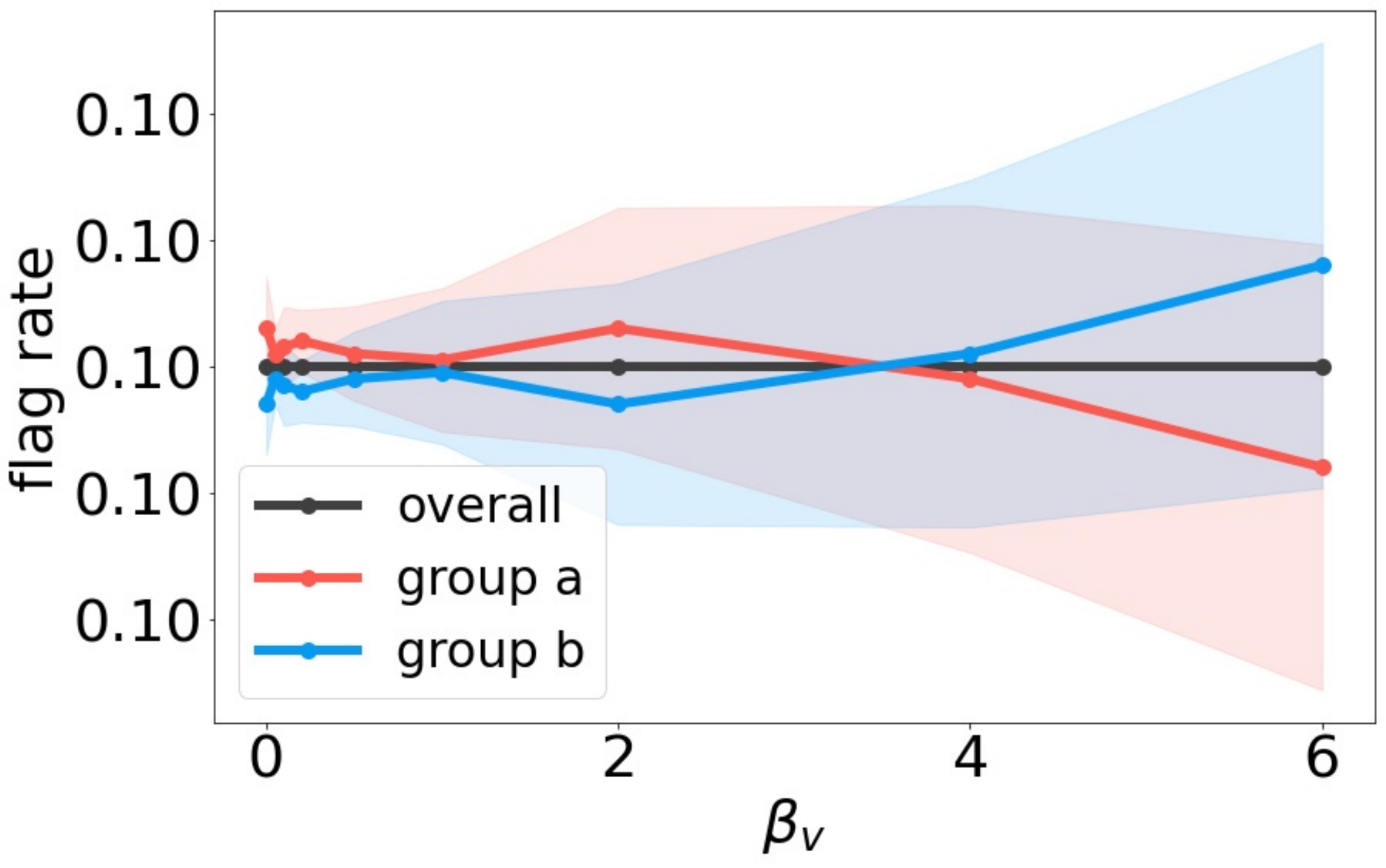}} 
    &
    \hspace{-0.1in}{\includegraphics[width=0.195\textwidth]{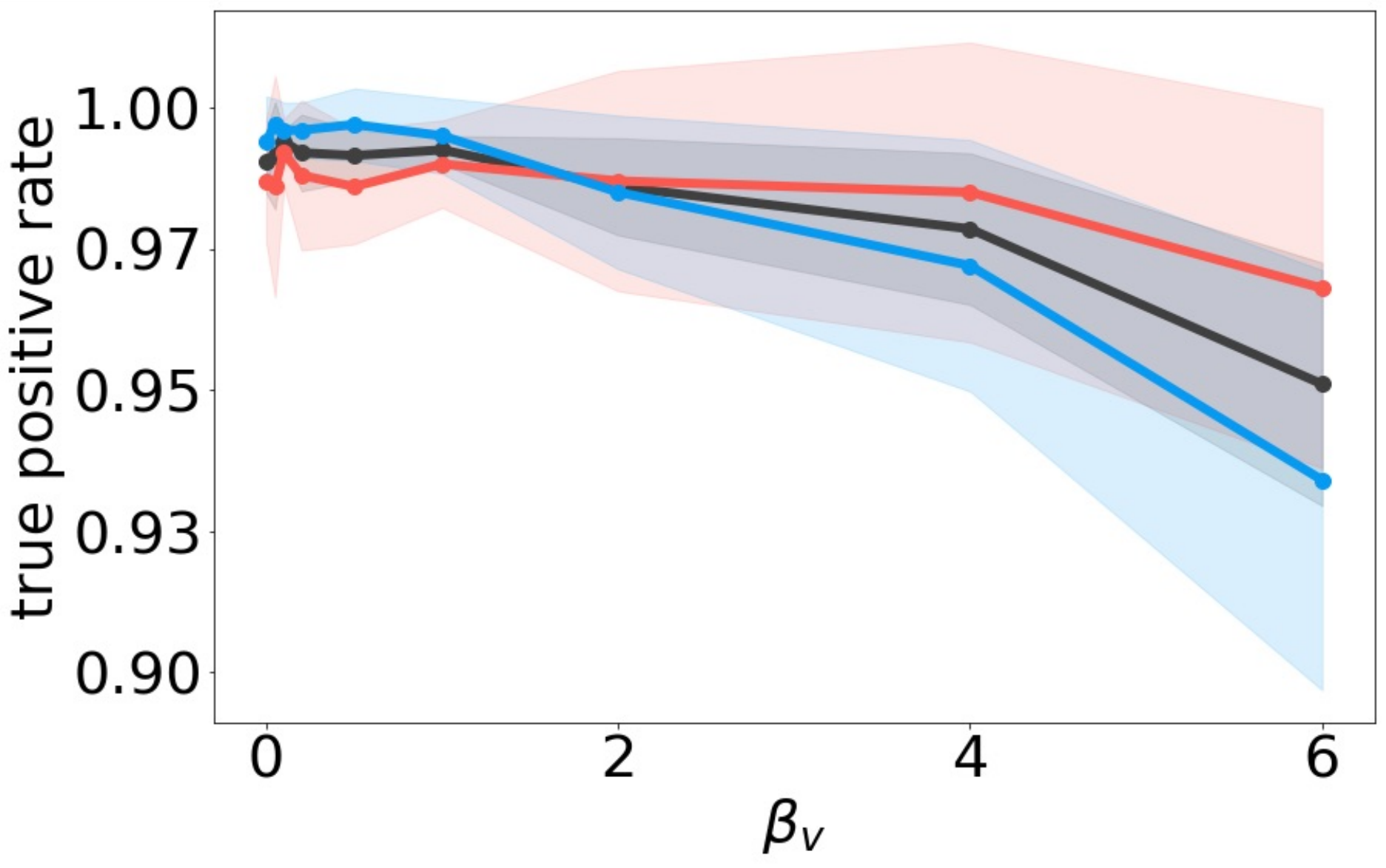}} &
     \hspace{-0.1in}{\includegraphics[width=0.195\textwidth]{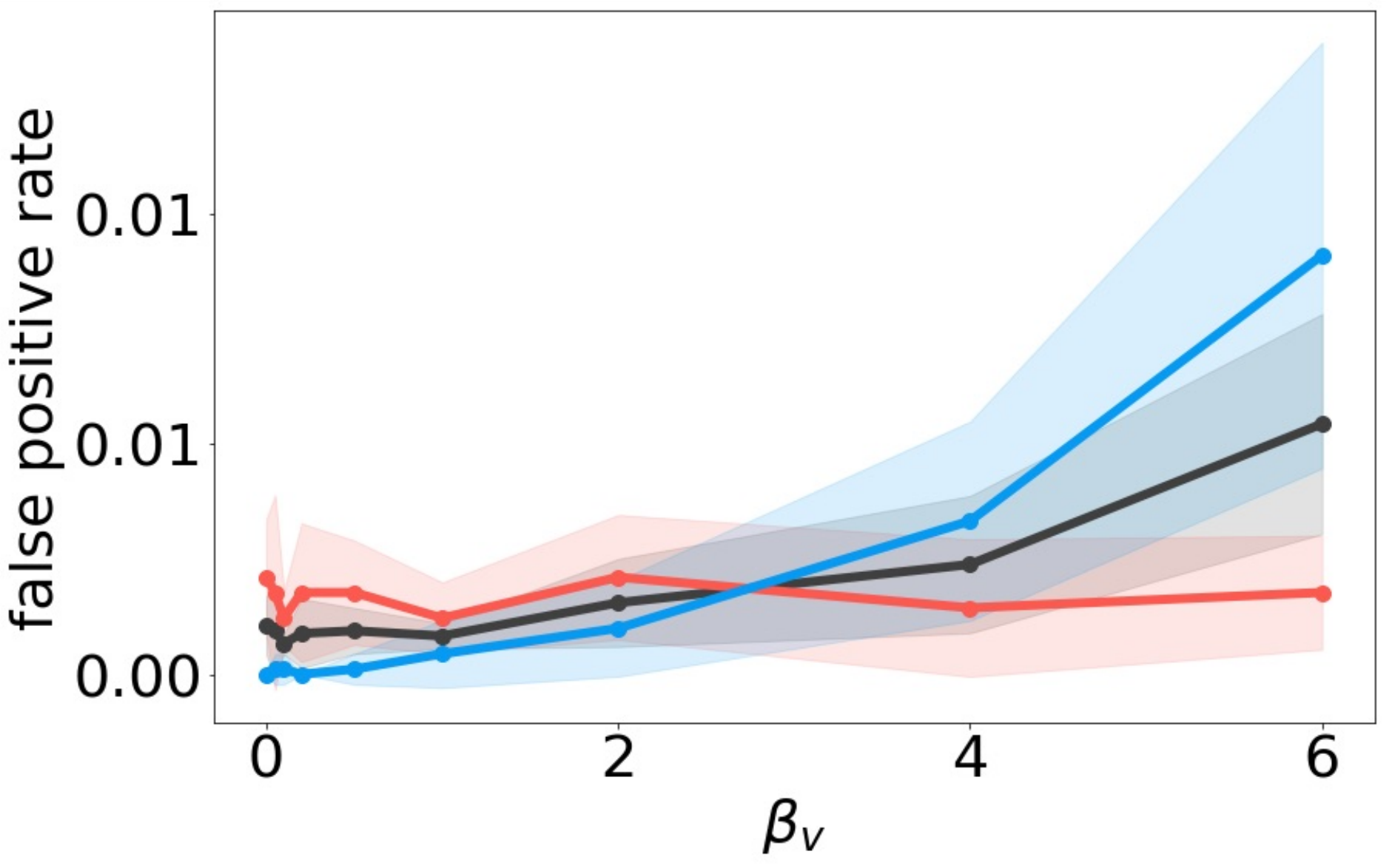}}
    &
   \hspace{-0.1in} {\includegraphics[width=0.195\textwidth]{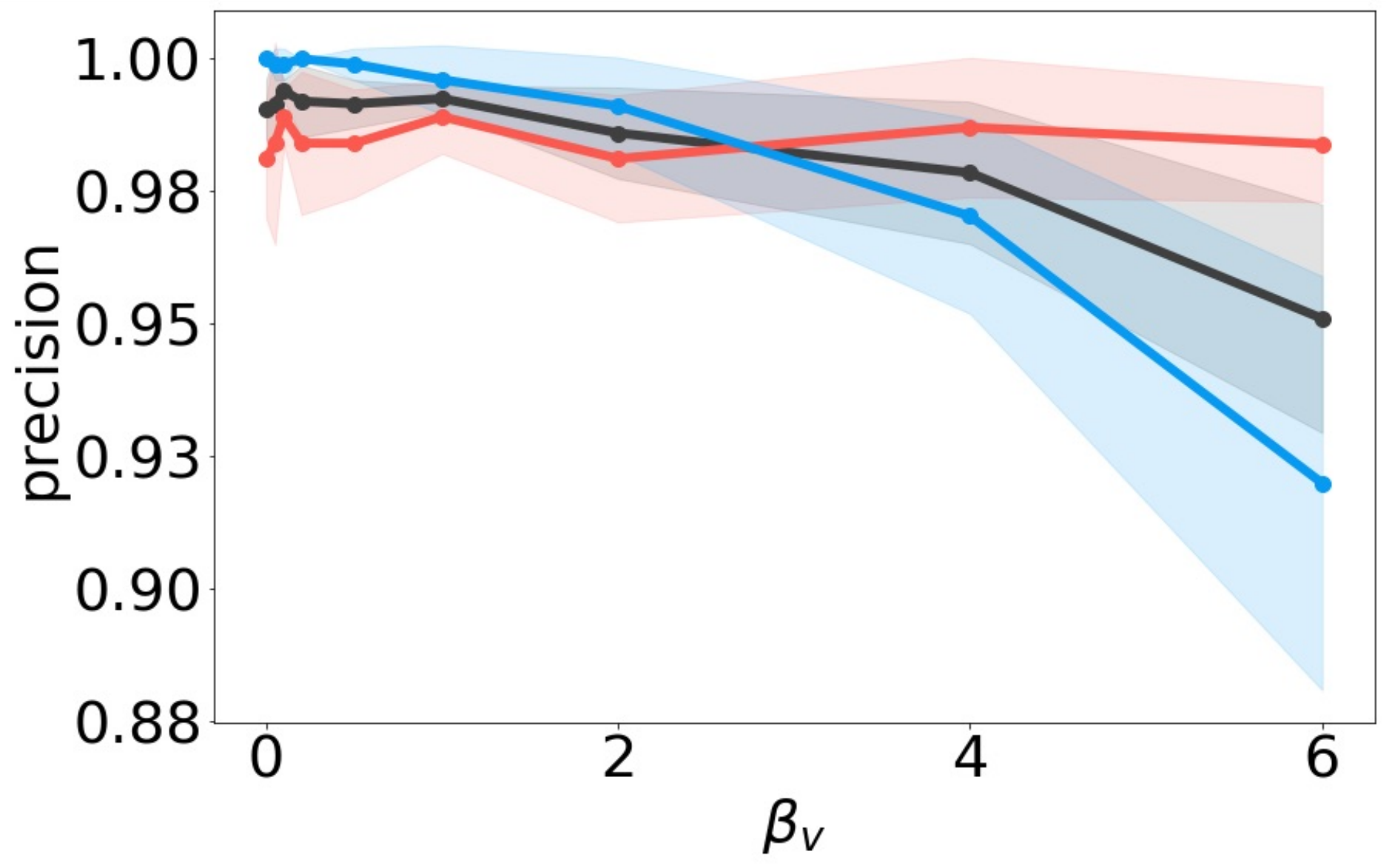}}
   &
    \hspace{-0.1in}{\includegraphics[width=0.195\textwidth]{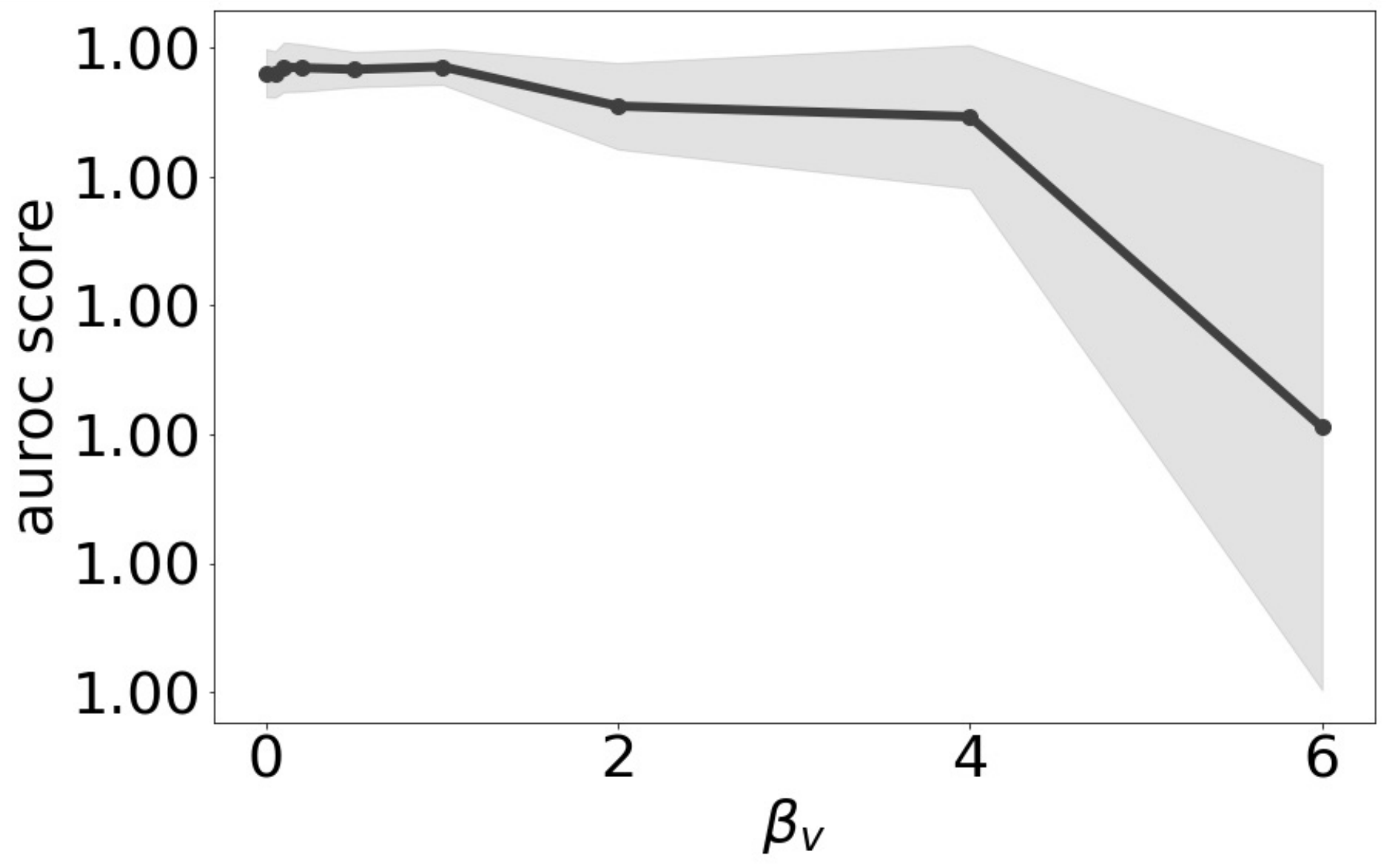}}
   \\
     \hspace{-0.1in}{\includegraphics[width=0.195\textwidth]{FIG/varianceshift/deepae/cluster/fr_legend.pdf}} 
    &
    \hspace{-0.1in}{\includegraphics[width=0.195\textwidth]{FIG/varianceshift/deepae/cluster/tpr.pdf}} &
     \hspace{-0.1in}{\includegraphics[width=0.195\textwidth]{FIG/varianceshift/deepae/cluster/fpr.pdf}}
    &
   \hspace{-0.1in} {\includegraphics[width=0.195\textwidth]{FIG/varianceshift/deepae/cluster/prec.pdf}}
   &
     \hspace{-0.1in}{\includegraphics[width=0.195\textwidth]{FIG/varianceshift/deepae/cluster/auroc.pdf}}
   \\
   \hspace{-0.1in}{\includegraphics[width=0.195\textwidth]{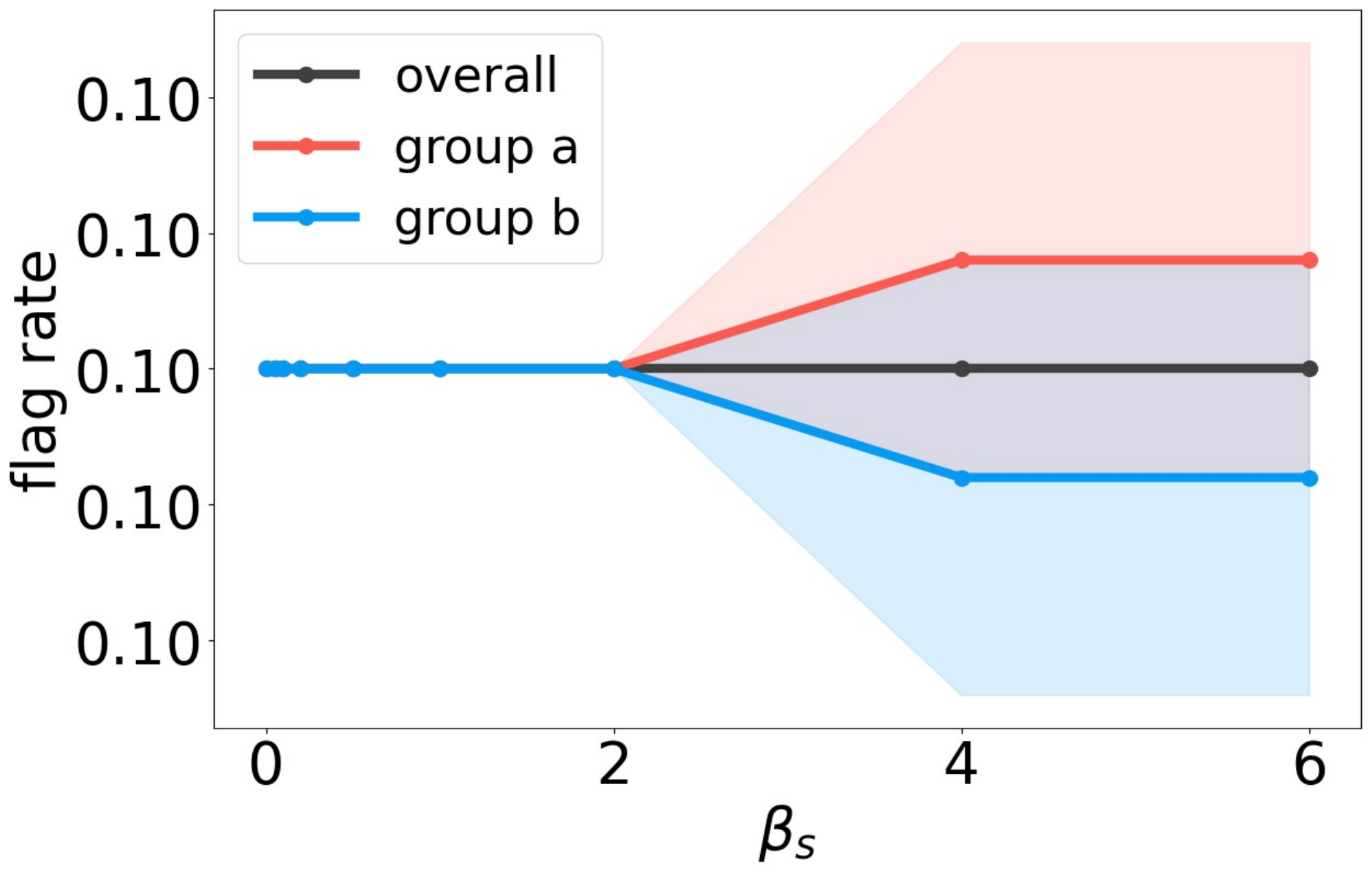} }
    &
    \hspace{-0.1in}{\includegraphics[width=0.195\textwidth]{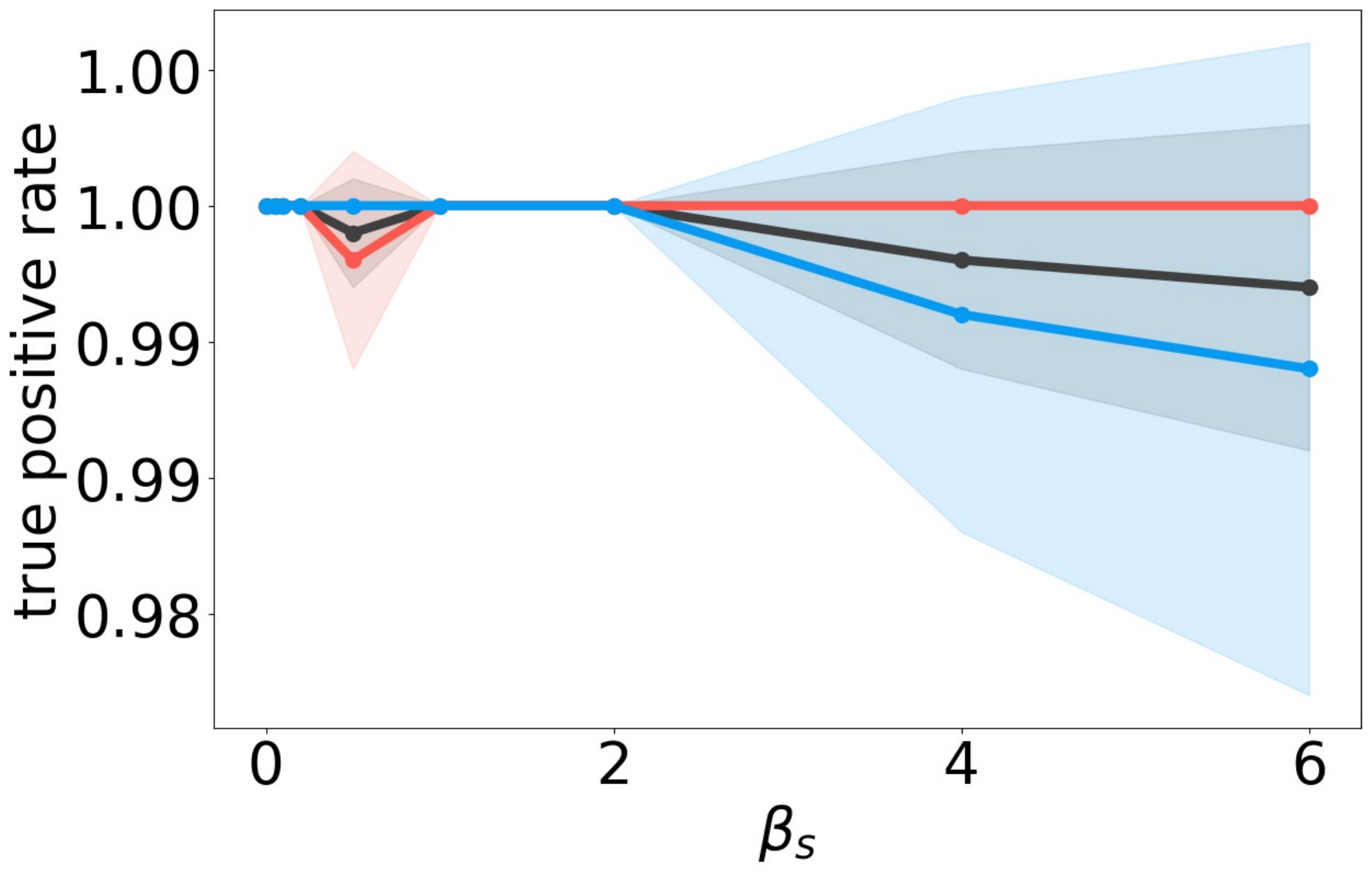}} &
     \hspace{-0.1in}{\includegraphics[width=0.195\textwidth]{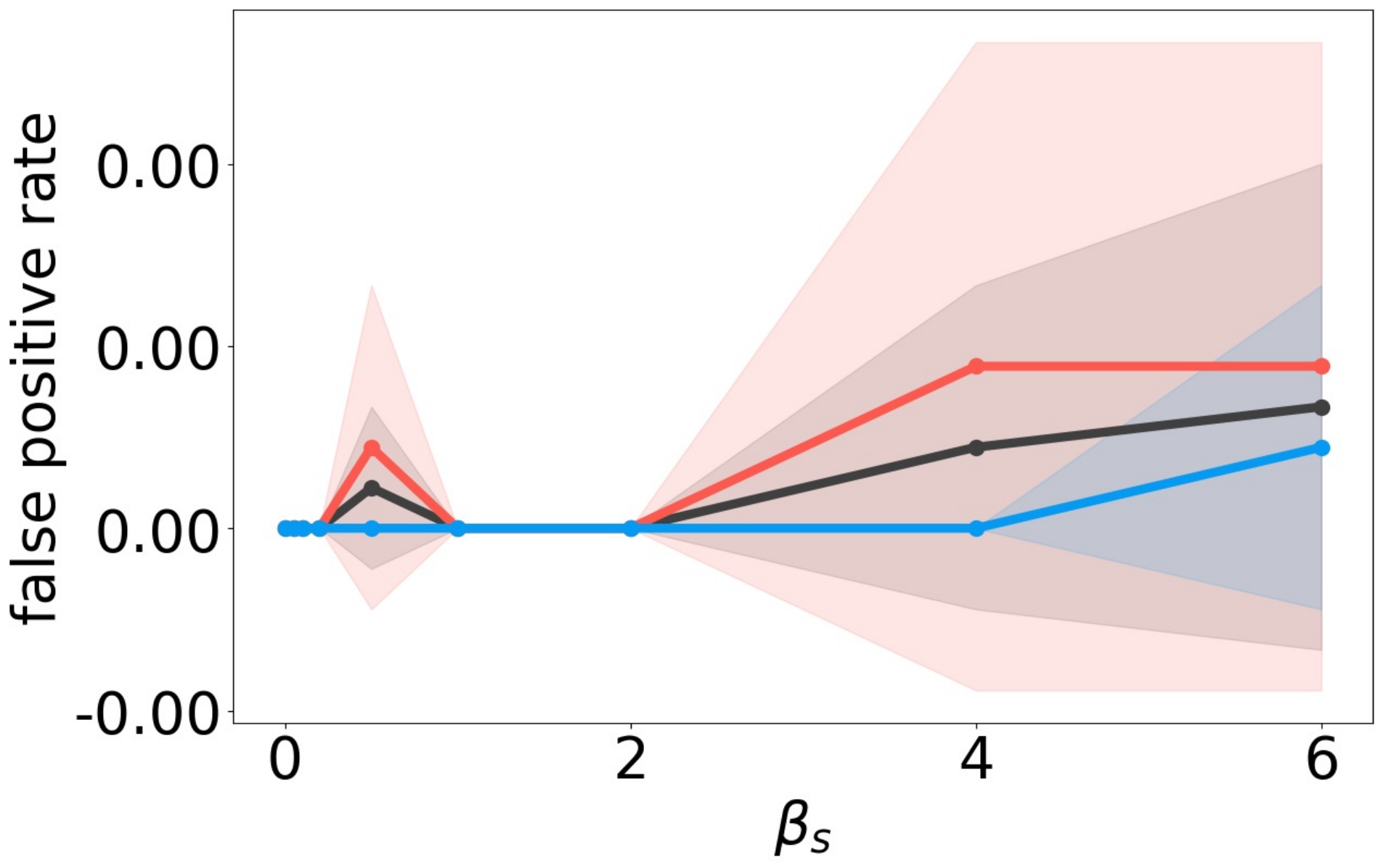}}
    &
   \hspace{-0.1in} {\includegraphics[width=0.195\textwidth]{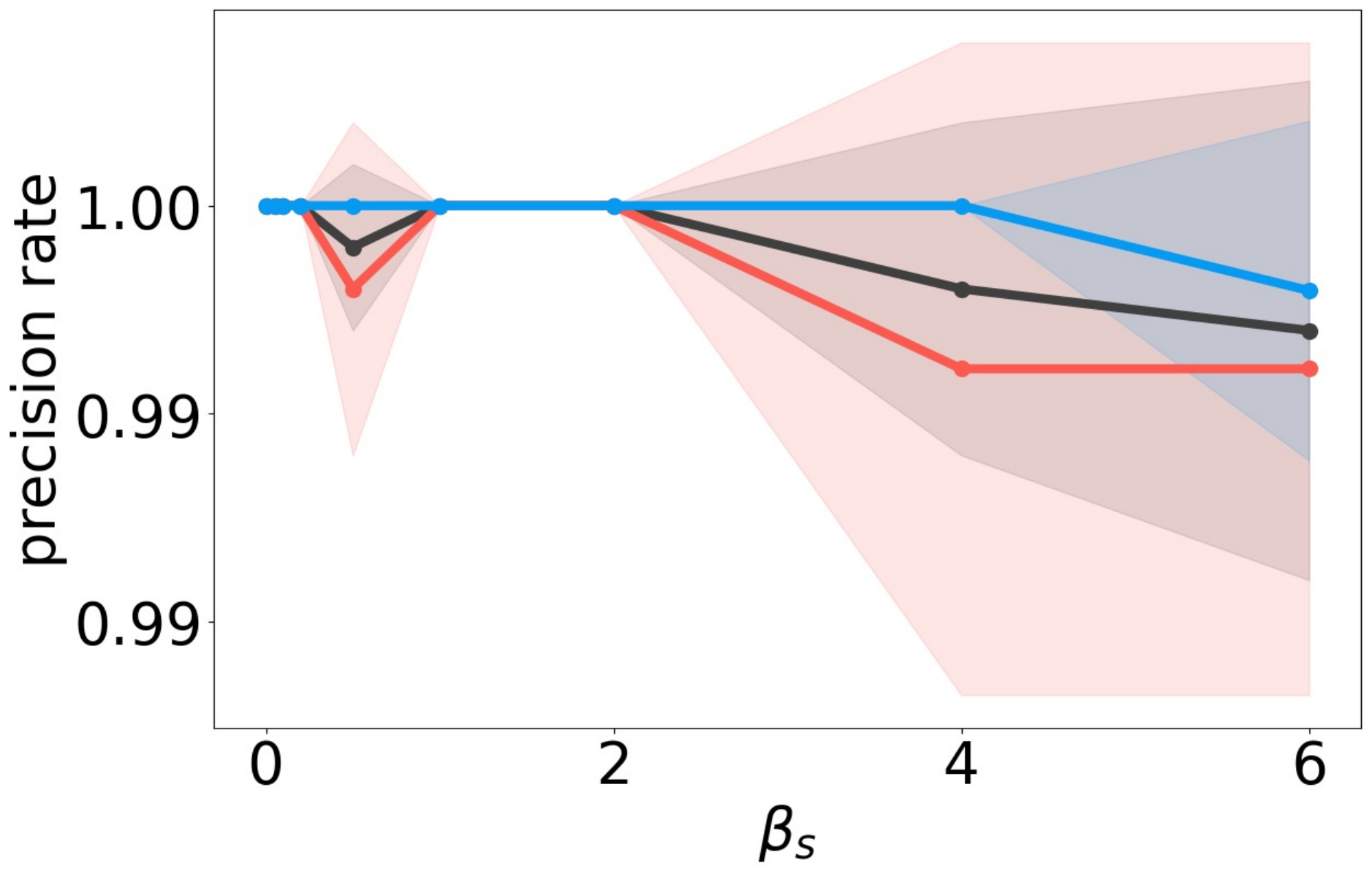}}
   &
   \hspace{-0.1in} {\includegraphics[width=0.195\textwidth]{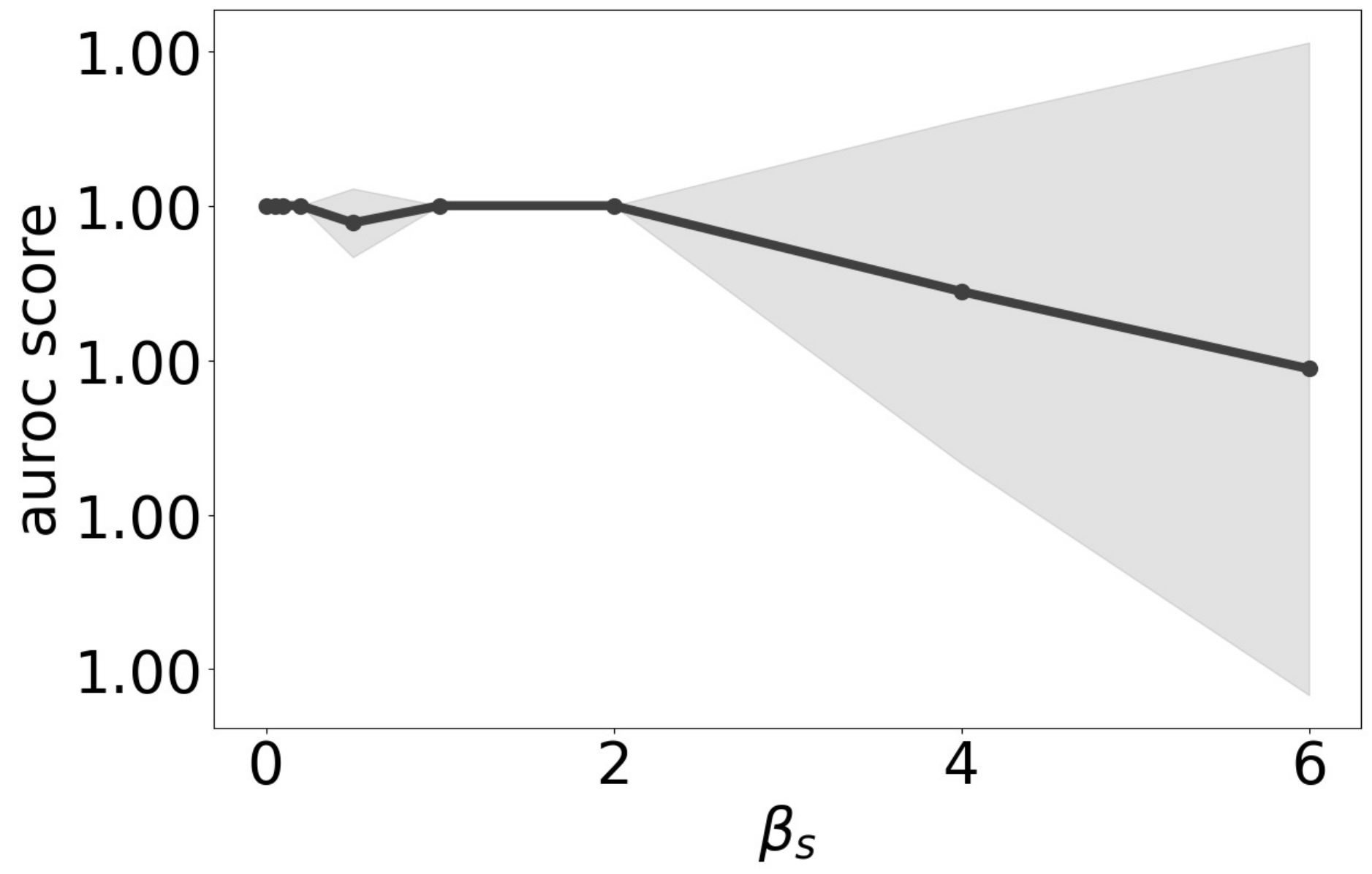}}
    \end{tabular}

    \caption{(best in color) Group-wise fairness metrics and AUROC for (top to bottom) LOF, iForest, DeepAE and FairOD under \underline{feature measurement bias} (variance shift) on \textit{clustered} outliers. }
    \label{fig:varianceshift_all_cluster}
\end{figure*}

\begin{figure*}[!ht]
    \centering
    \begin{tabular}{lllll}
    \hspace{-0.1in}{\includegraphics[width=0.195\textwidth]{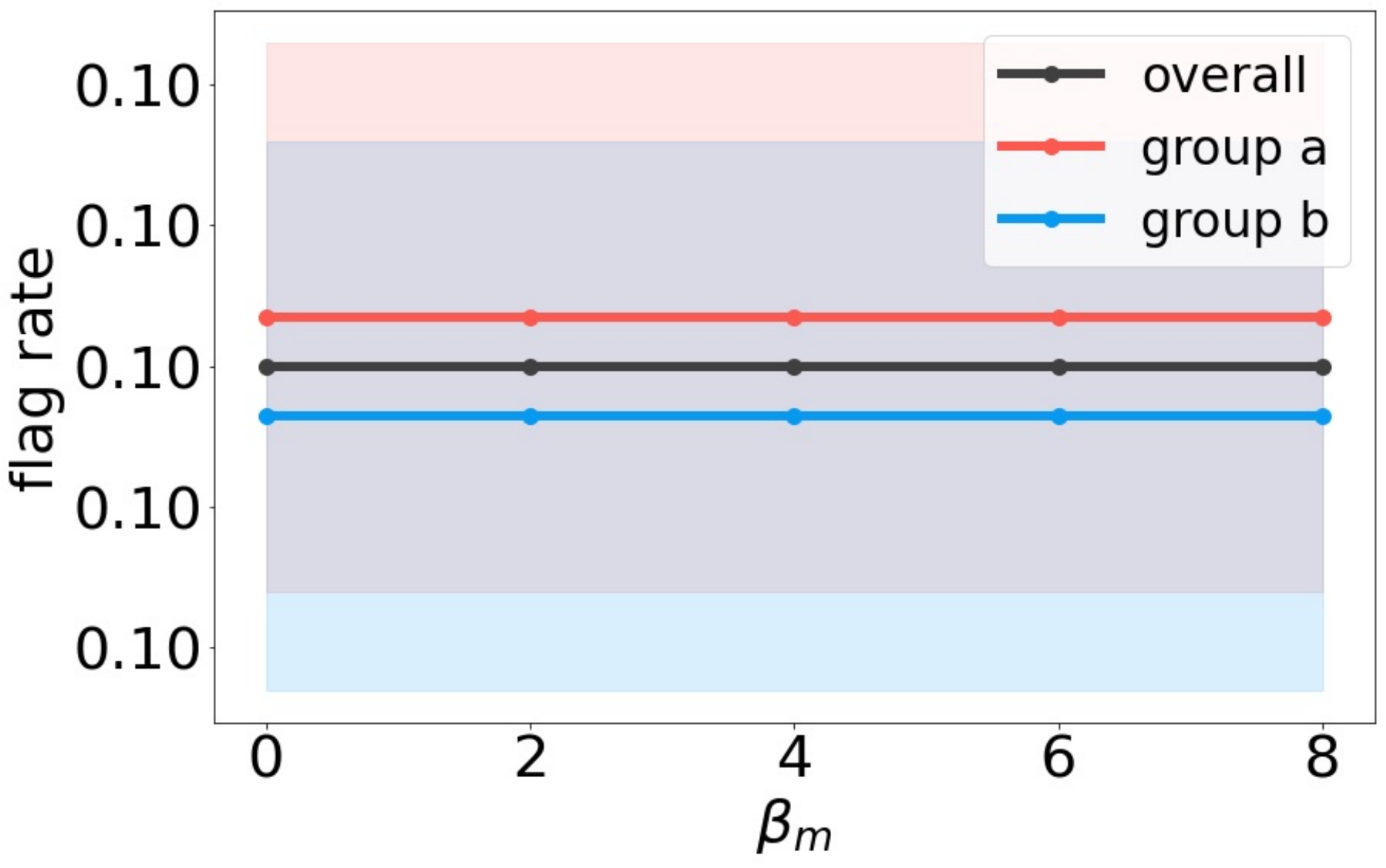}} 
    &
    \hspace{-0.1in}{\includegraphics[width=0.195\textwidth]{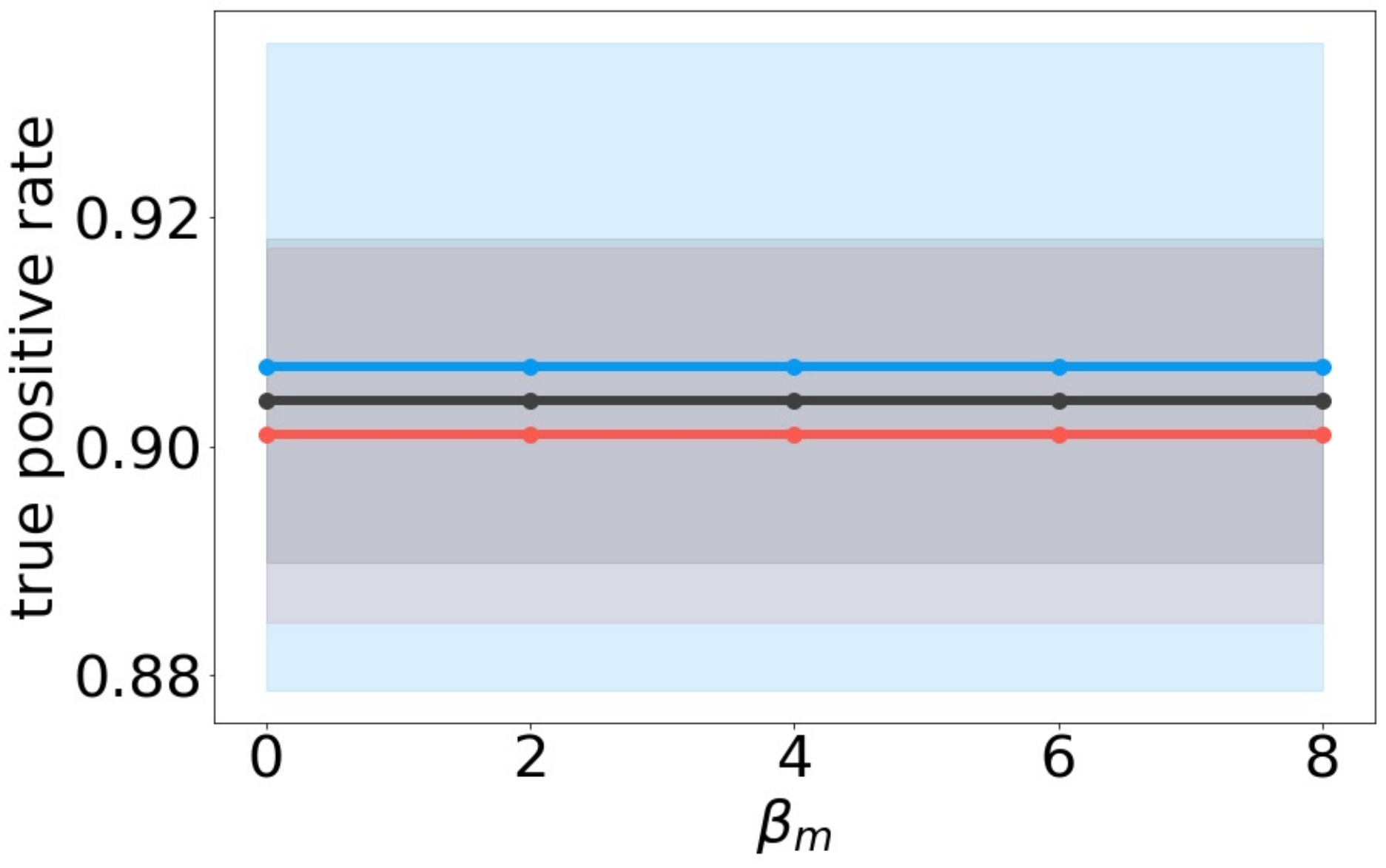}} &
     \hspace{-0.1in}{\includegraphics[width=0.195\textwidth]{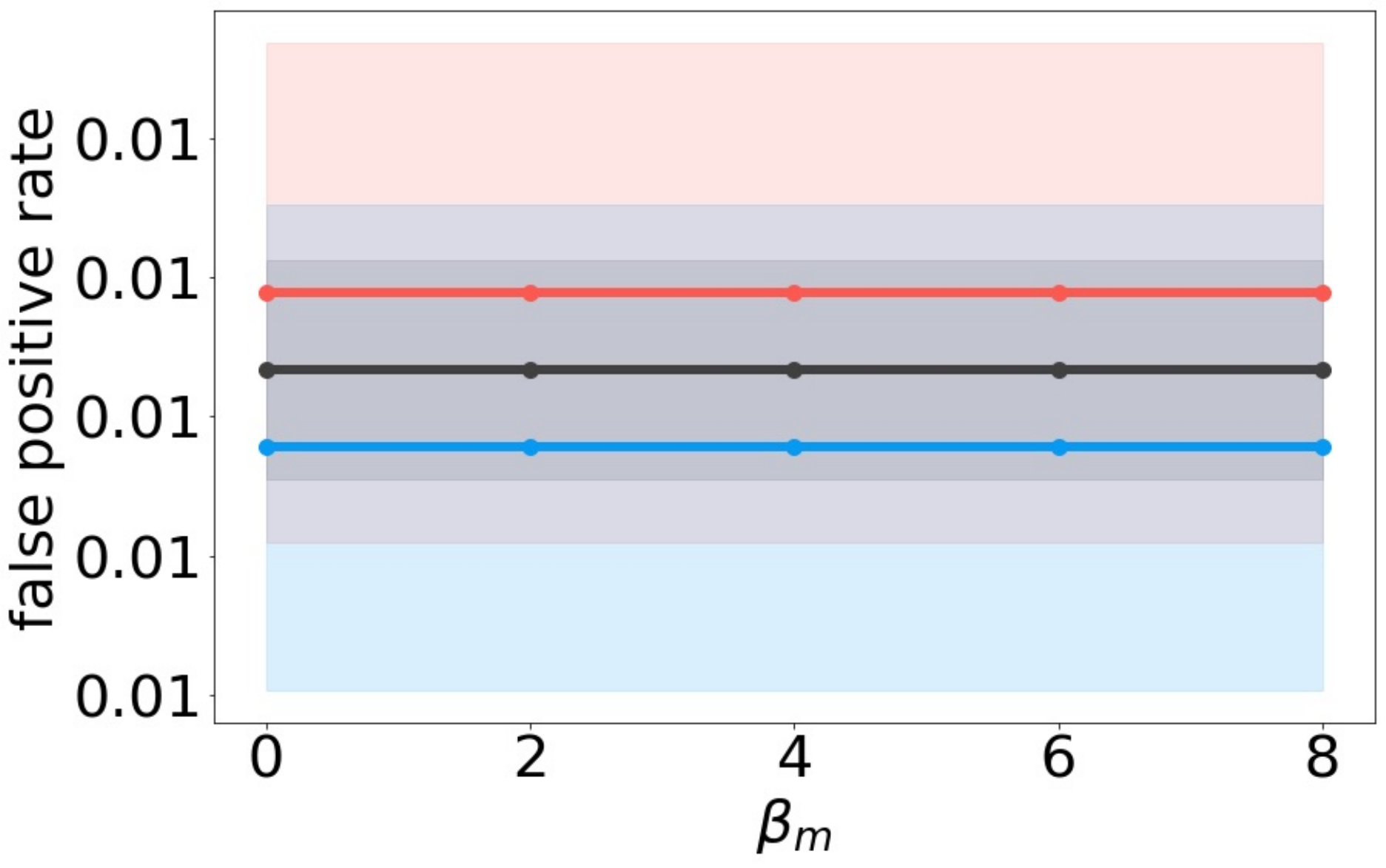}}
    &
   \hspace{-0.1in} {\includegraphics[width=0.195\textwidth]{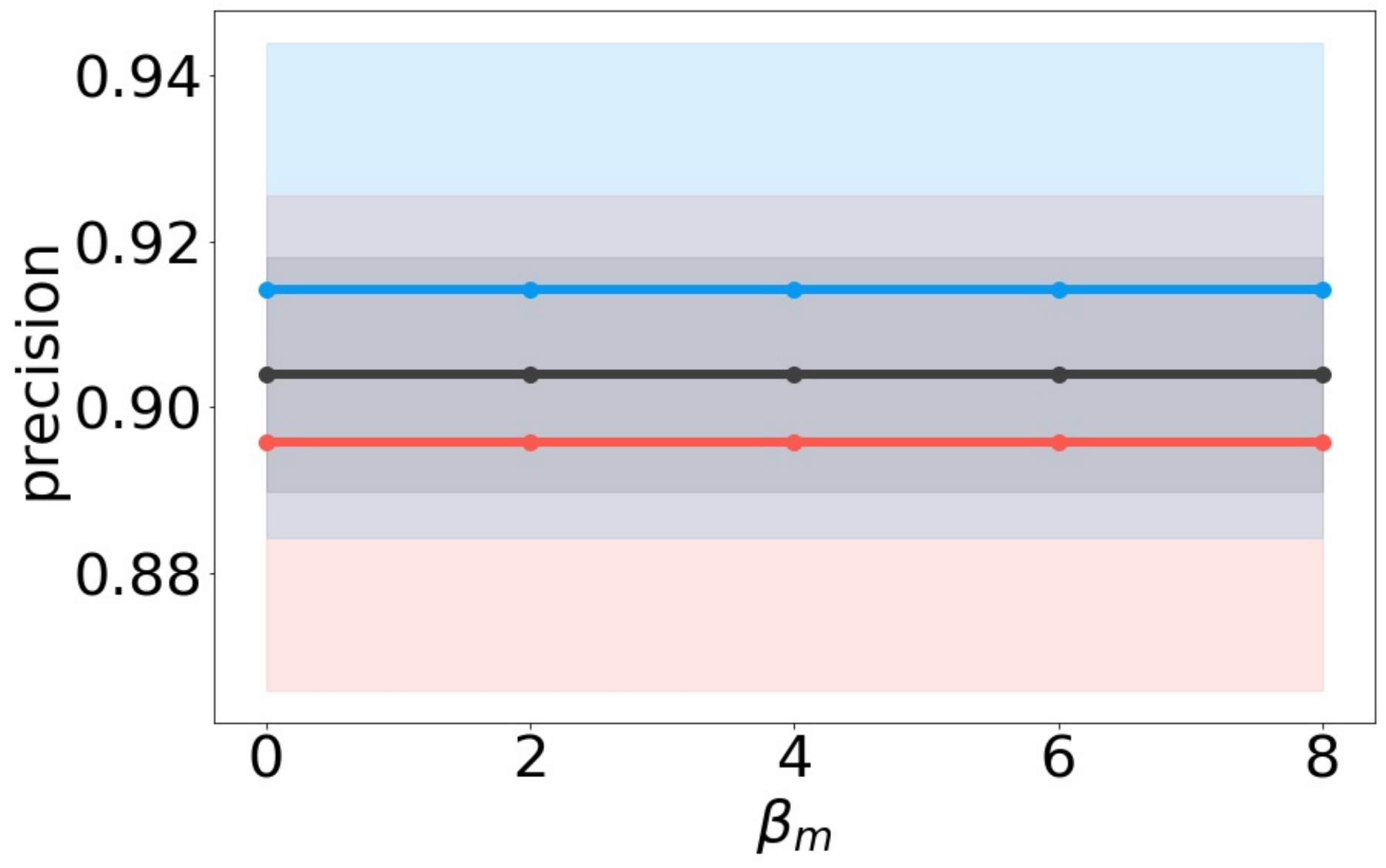}}
   &
   \hspace{-0.1in}
   {\includegraphics[width=0.195\textwidth]{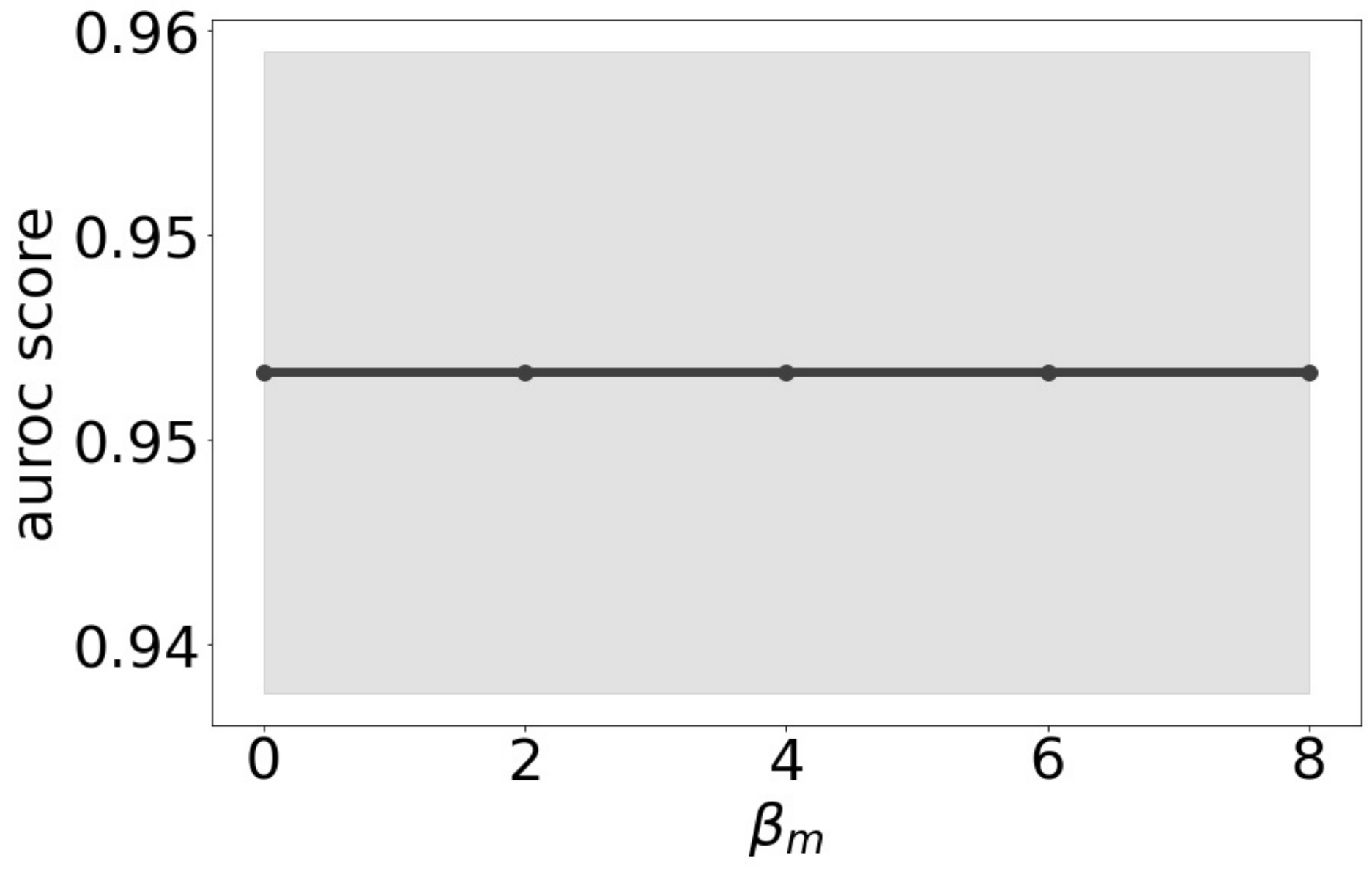}} 
\\
\hspace{-0.1in}{\includegraphics[width=0.195\textwidth]{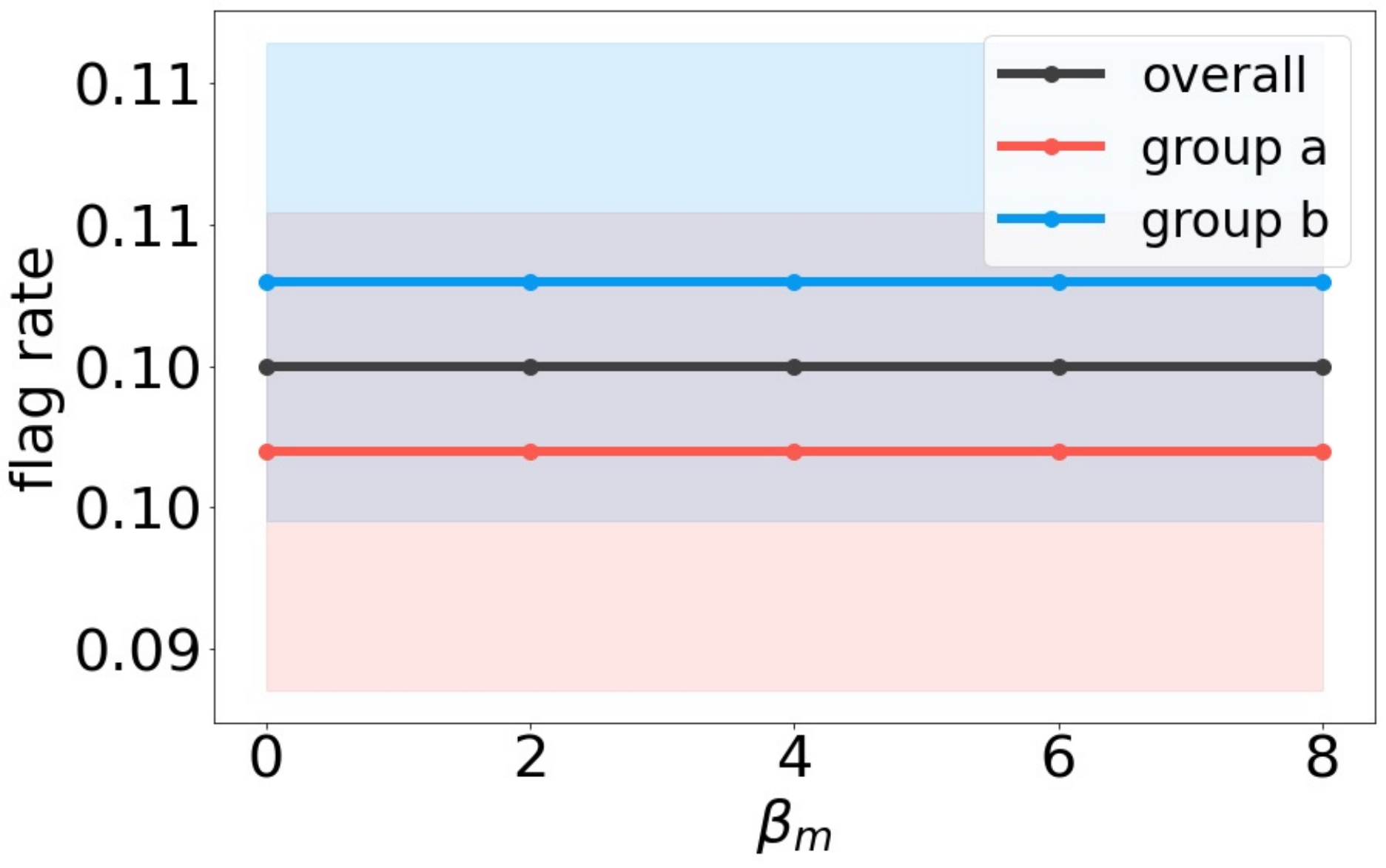}} 
    &
    \hspace{-0.1in}{\includegraphics[width=0.195\textwidth]{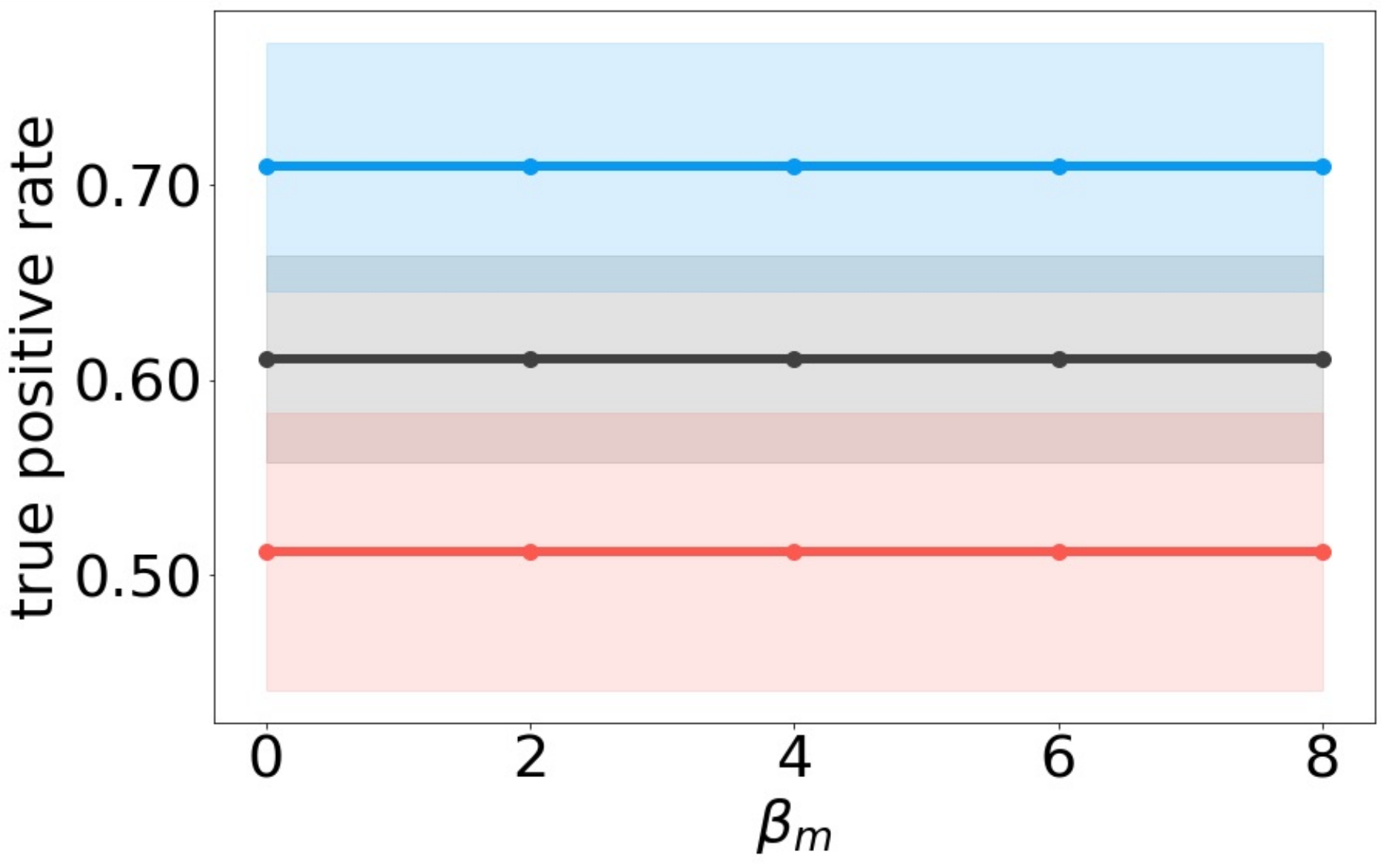}} &
     \hspace{-0.1in}{\includegraphics[width=0.195\textwidth]{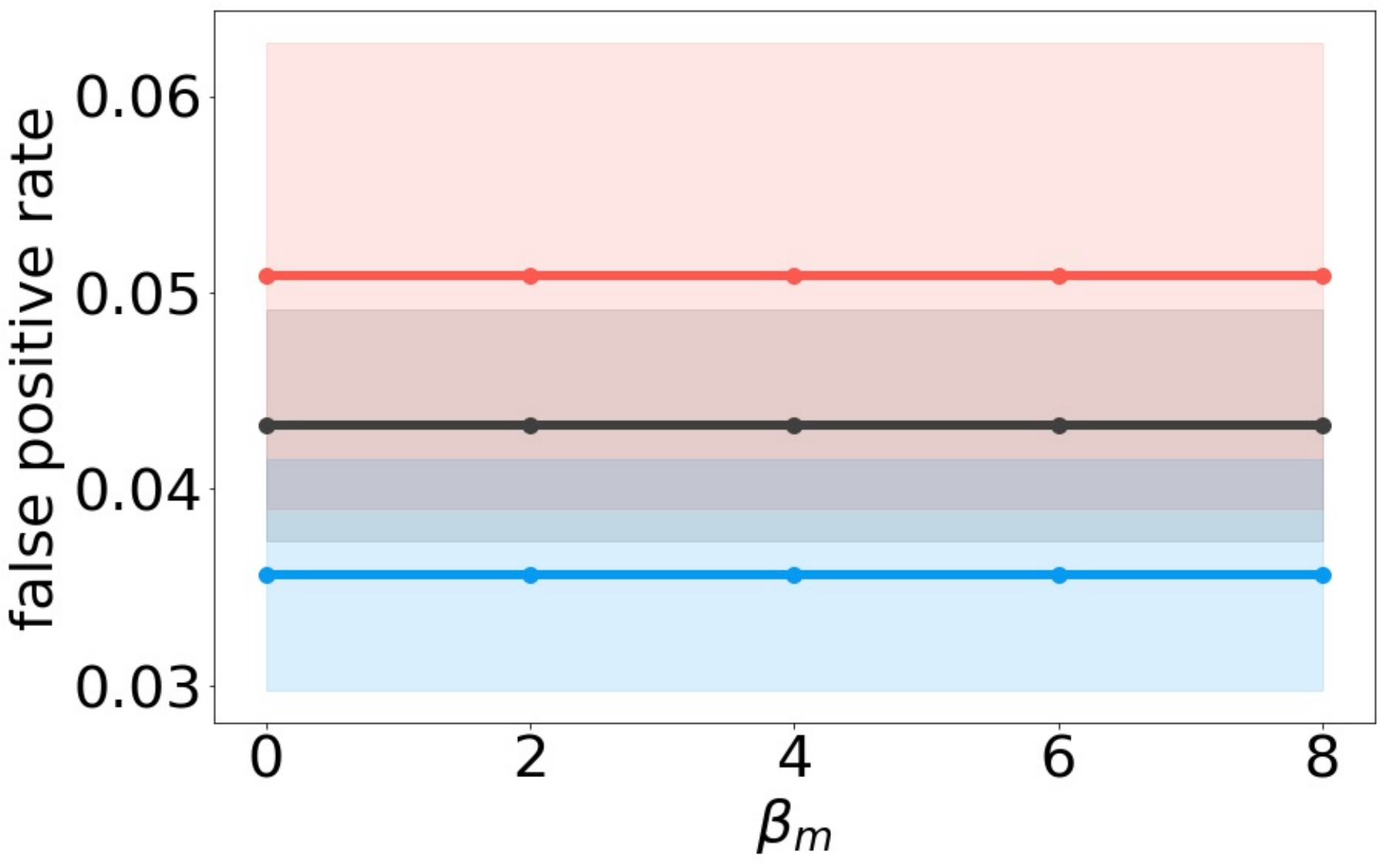}}
    &
   \hspace{-0.1in} {\includegraphics[width=0.195\textwidth]{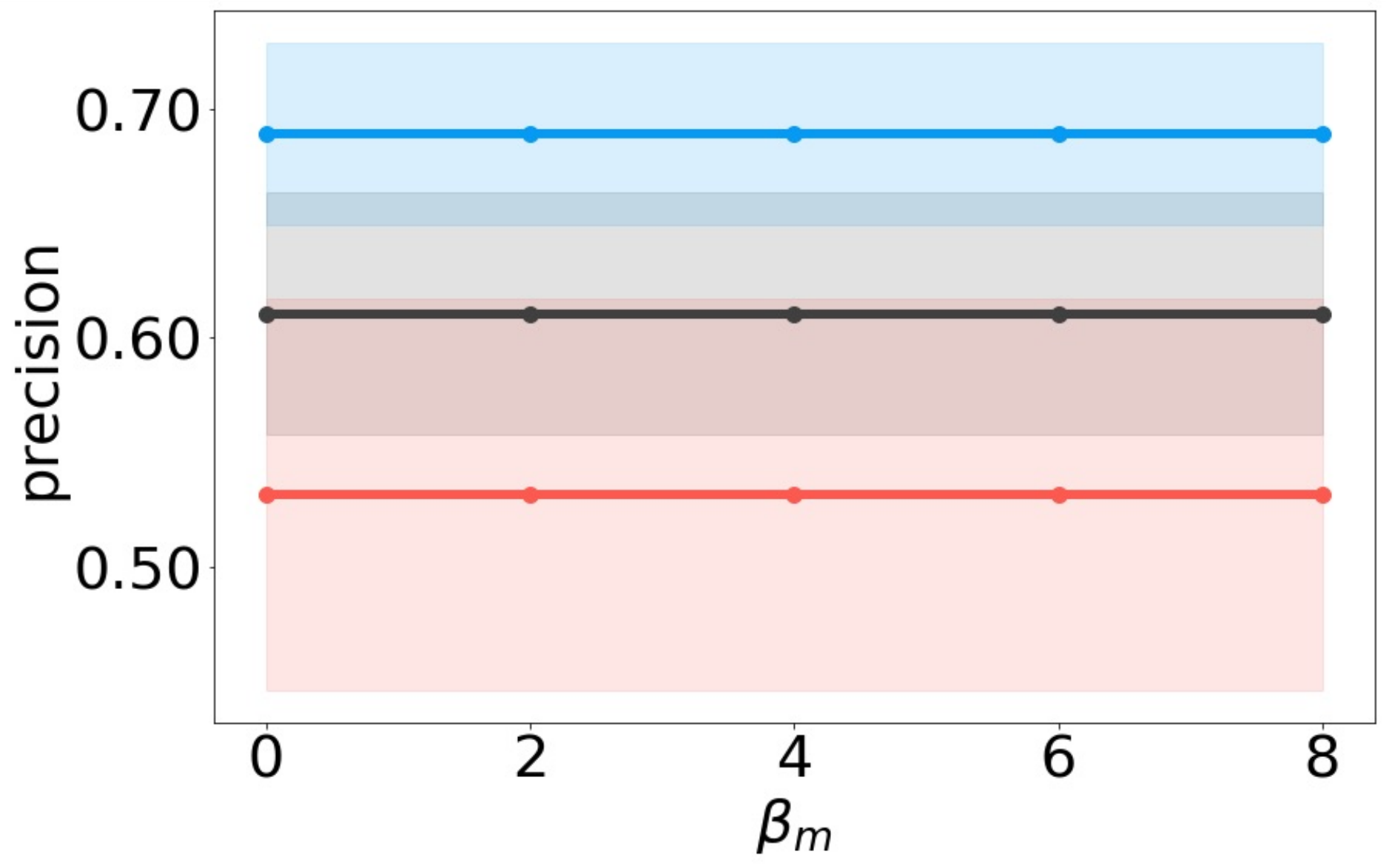}}
   &
    \hspace{-0.1in}{\includegraphics[width=0.195\textwidth]{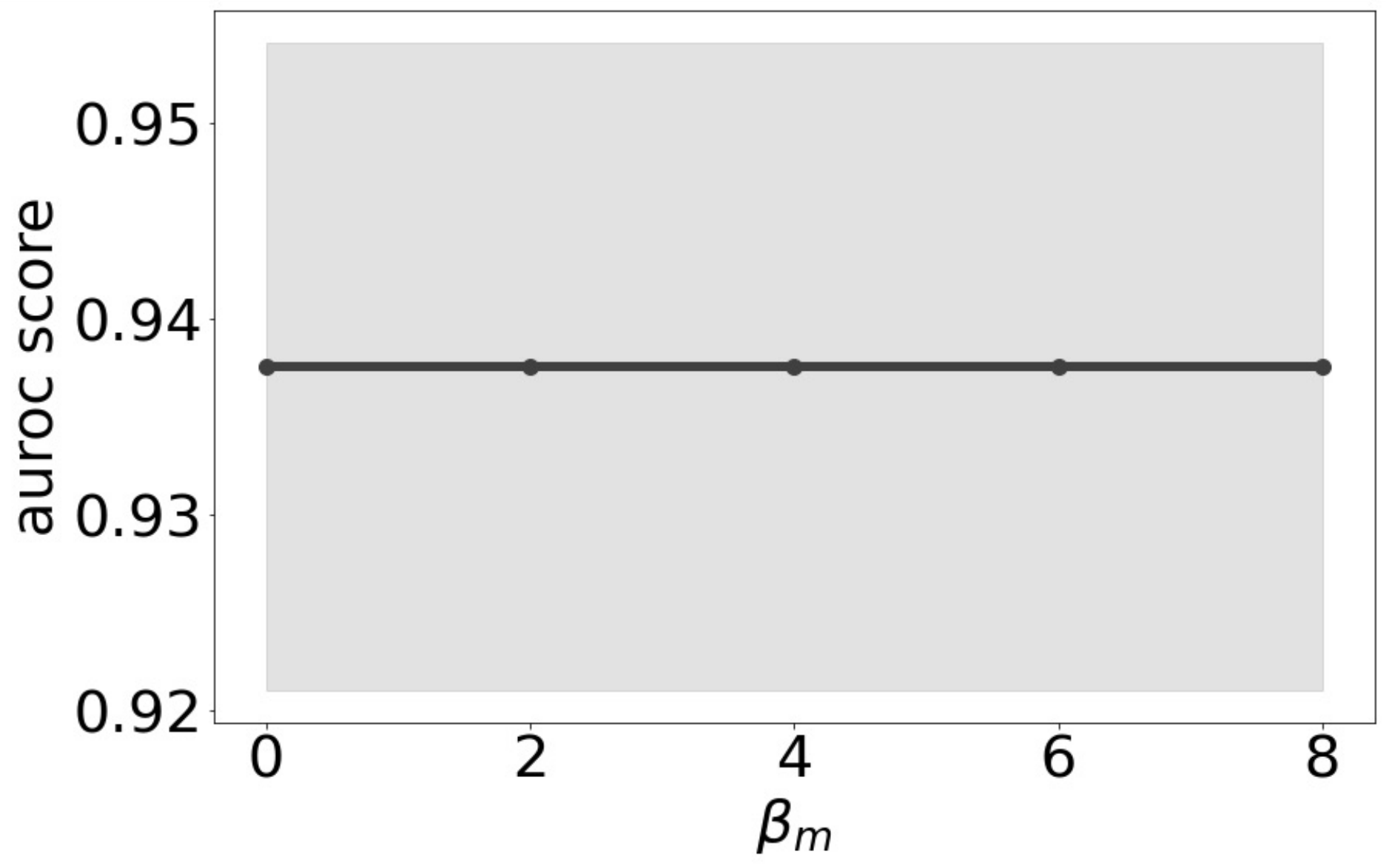}}
    \end{tabular}
    \caption{(best in color) Group-wise fairness metrics and AUROC for (top) LOF and (bottom) iForest under \underline{feature measurement bias} (mean shift) on \textit{clustered} outliers. }
    \label{fig:meanshift_lofif_cluster}
\end{figure*}

See Fig. \ref{fig:varianceshift_all_cluster} for variance shift and Fig. 
\ref{fig:meanshift_lofif_cluster} for mean shift results (LOF and iForest only for brevity).

\subsubsection{Membership obfuscation bias}
\label{assec:obf_cluster}

\begin{figure*}[!ht]
    \centering
    \begin{tabular}{lllll}
    \hspace{-0.1in}{\includegraphics[width=0.195\textwidth]{FIG/obfuscation/lof/cluster/fr_legend.pdf}} 
    &
    \hspace{-0.1in}{\includegraphics[width=0.195\textwidth]{FIG/obfuscation/lof/cluster/tpr.pdf}} &
     \hspace{-0.1in}{\includegraphics[width=0.195\textwidth]{FIG/obfuscation/lof/cluster/fpr.pdf}}
    &
   \hspace{-0.1in} {\includegraphics[width=0.195\textwidth]{FIG/obfuscation/lof/cluster/prec.pdf}}
   &
   \hspace{-0.1in}
   {\includegraphics[width=0.195\textwidth]{FIG/obfuscation/lof/cluster/auroc.pdf}} 
\\
\hspace{-0.1in}{\includegraphics[width=0.195\textwidth]{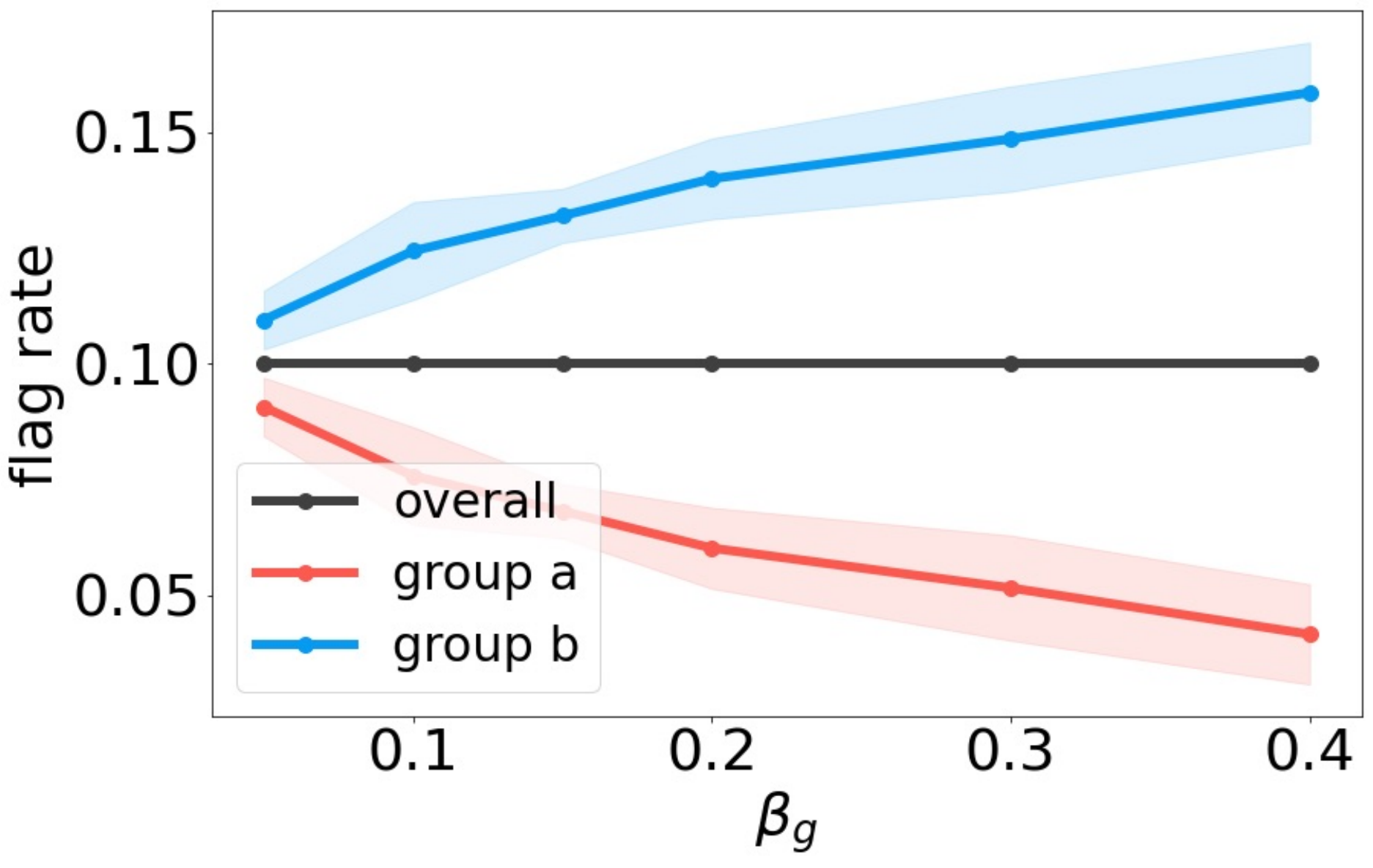}} 
    &
    \hspace{-0.1in}{\includegraphics[width=0.195\textwidth]{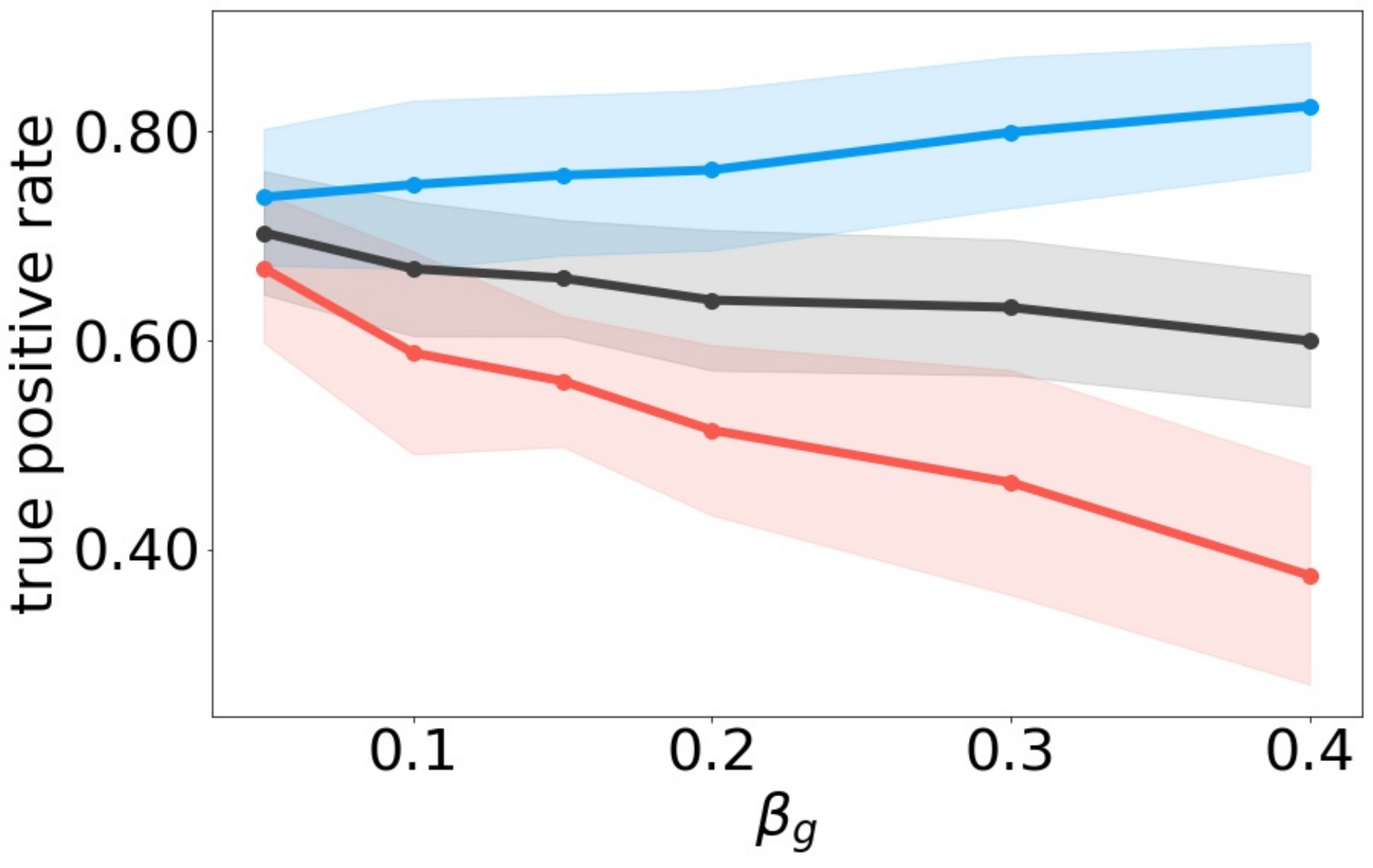}} &
     \hspace{-0.1in}{\includegraphics[width=0.195\textwidth]{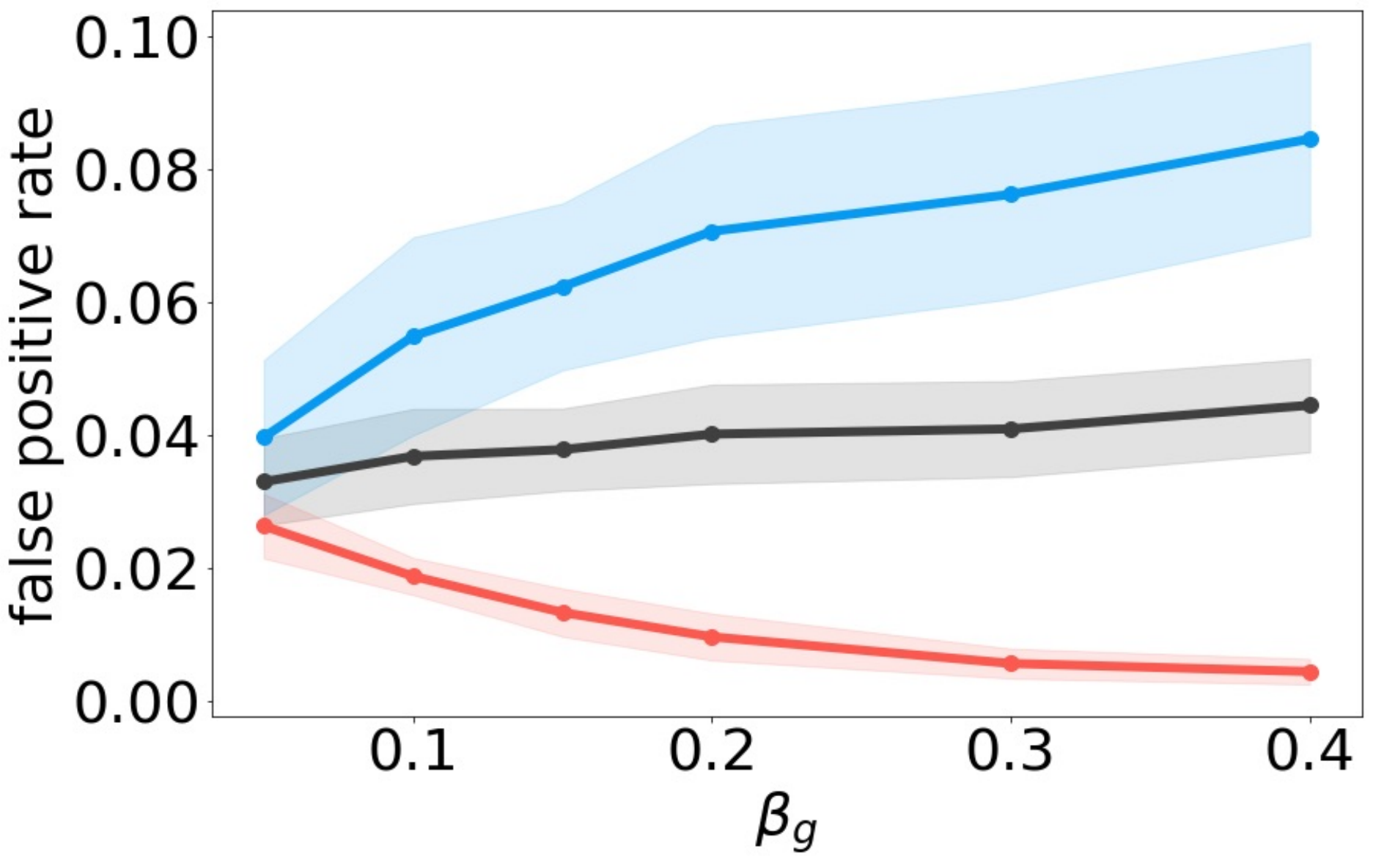}}
    &
   \hspace{-0.1in} {\includegraphics[width=0.195\textwidth]{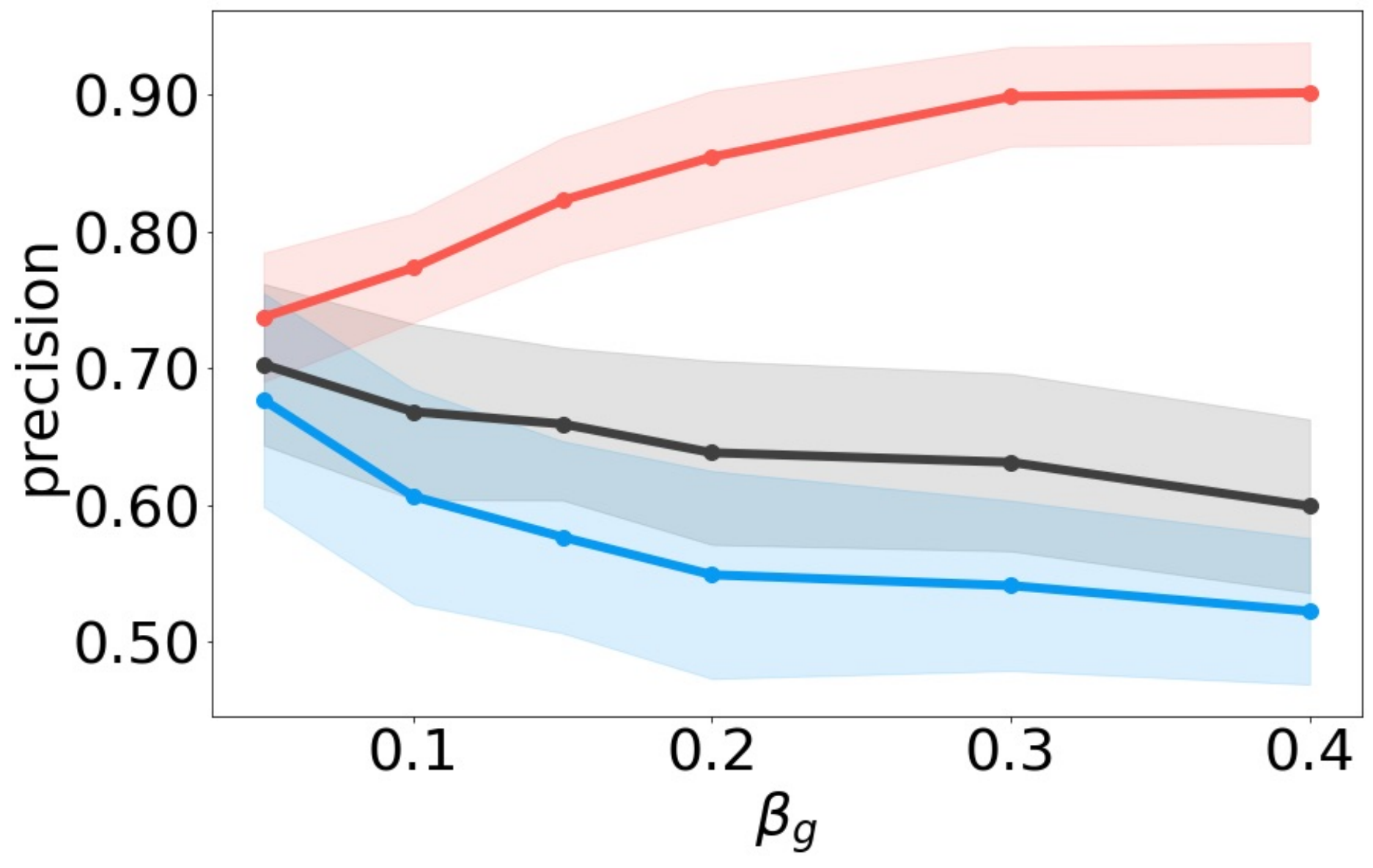}}
   &
    \hspace{-0.1in}{\includegraphics[width=0.195\textwidth]{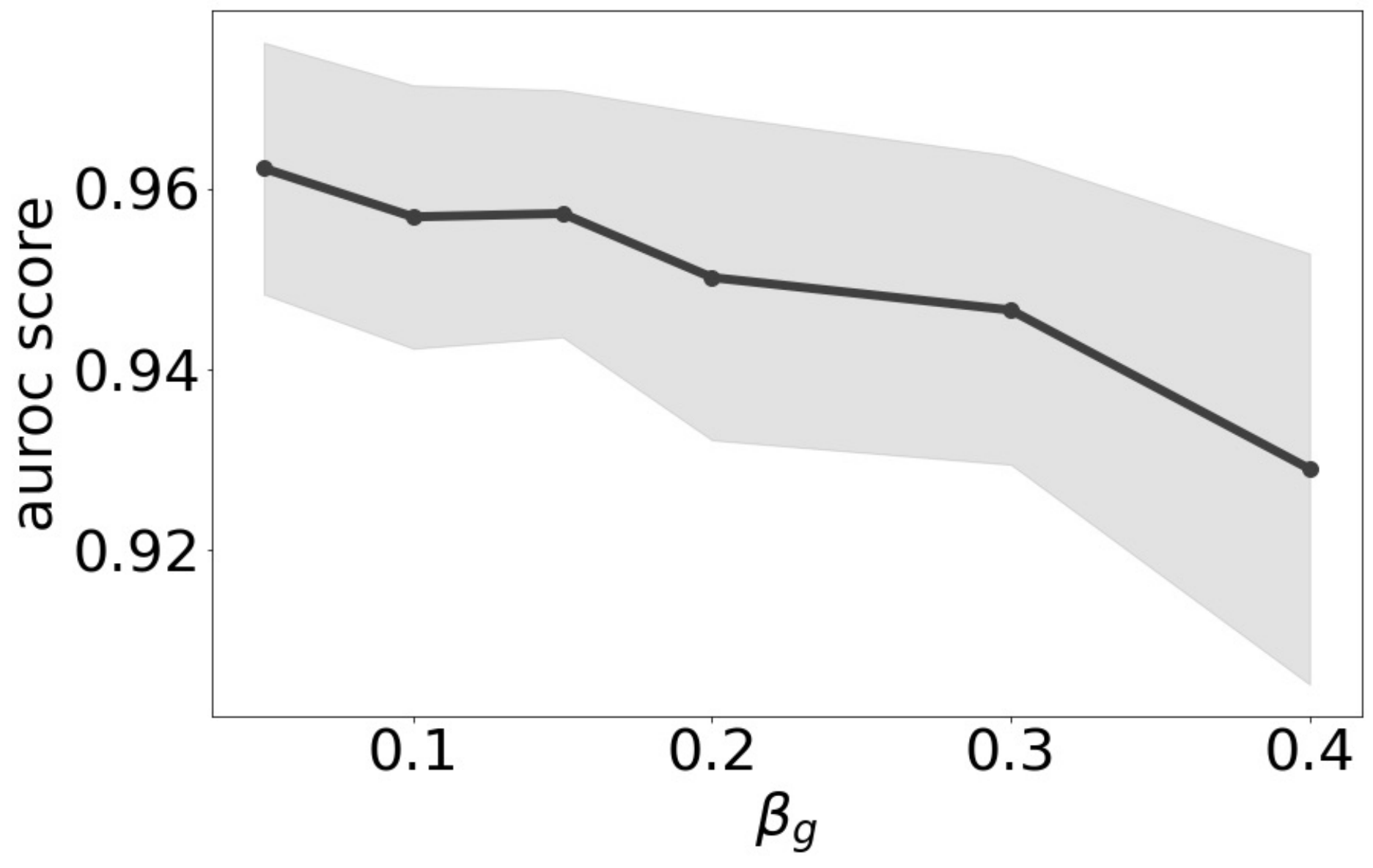}}
   \\
     \hspace{-0.1in}{\includegraphics[width=0.195\textwidth]{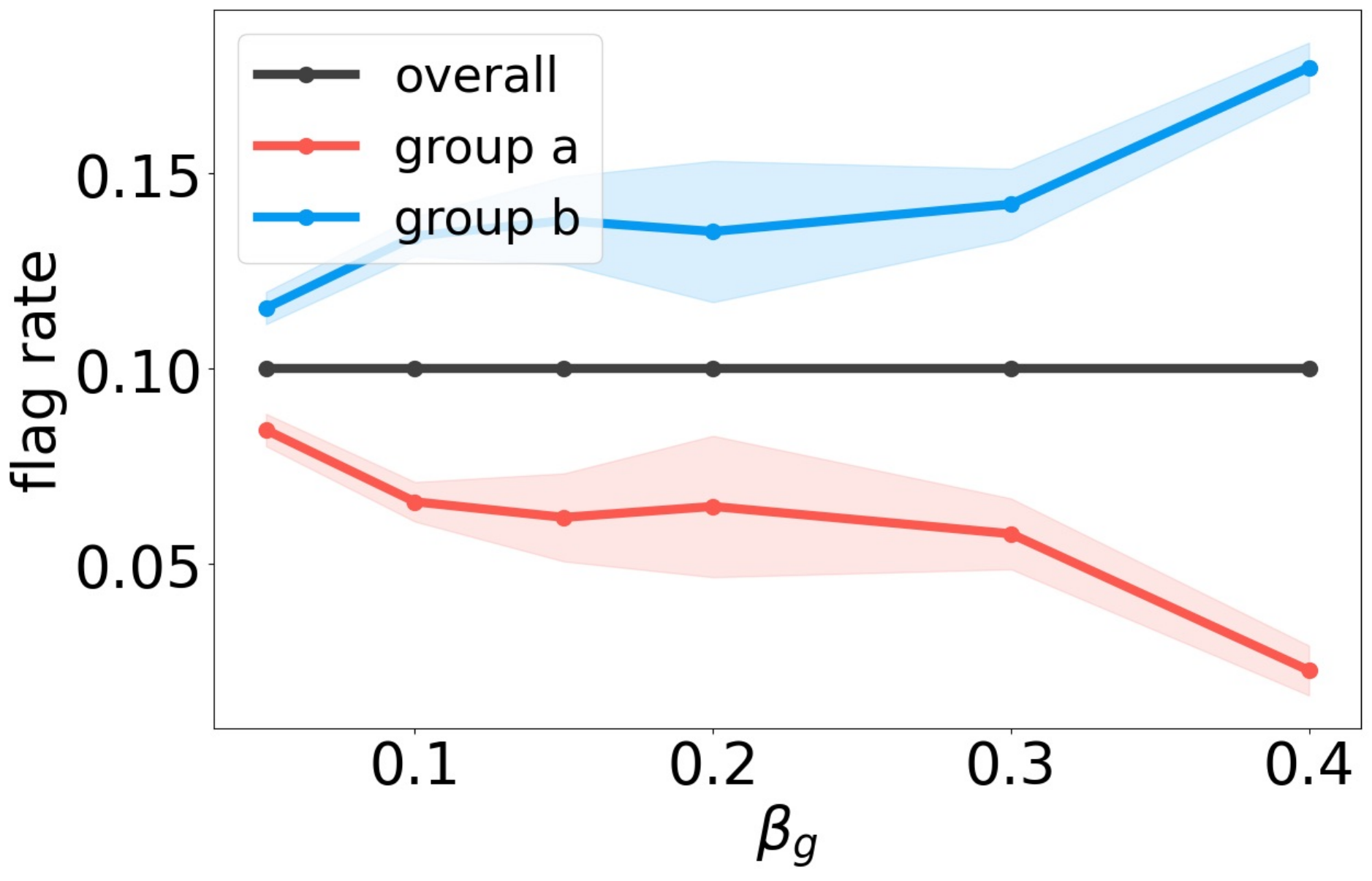}} 
    &
    \hspace{-0.1in}{\includegraphics[width=0.195\textwidth]{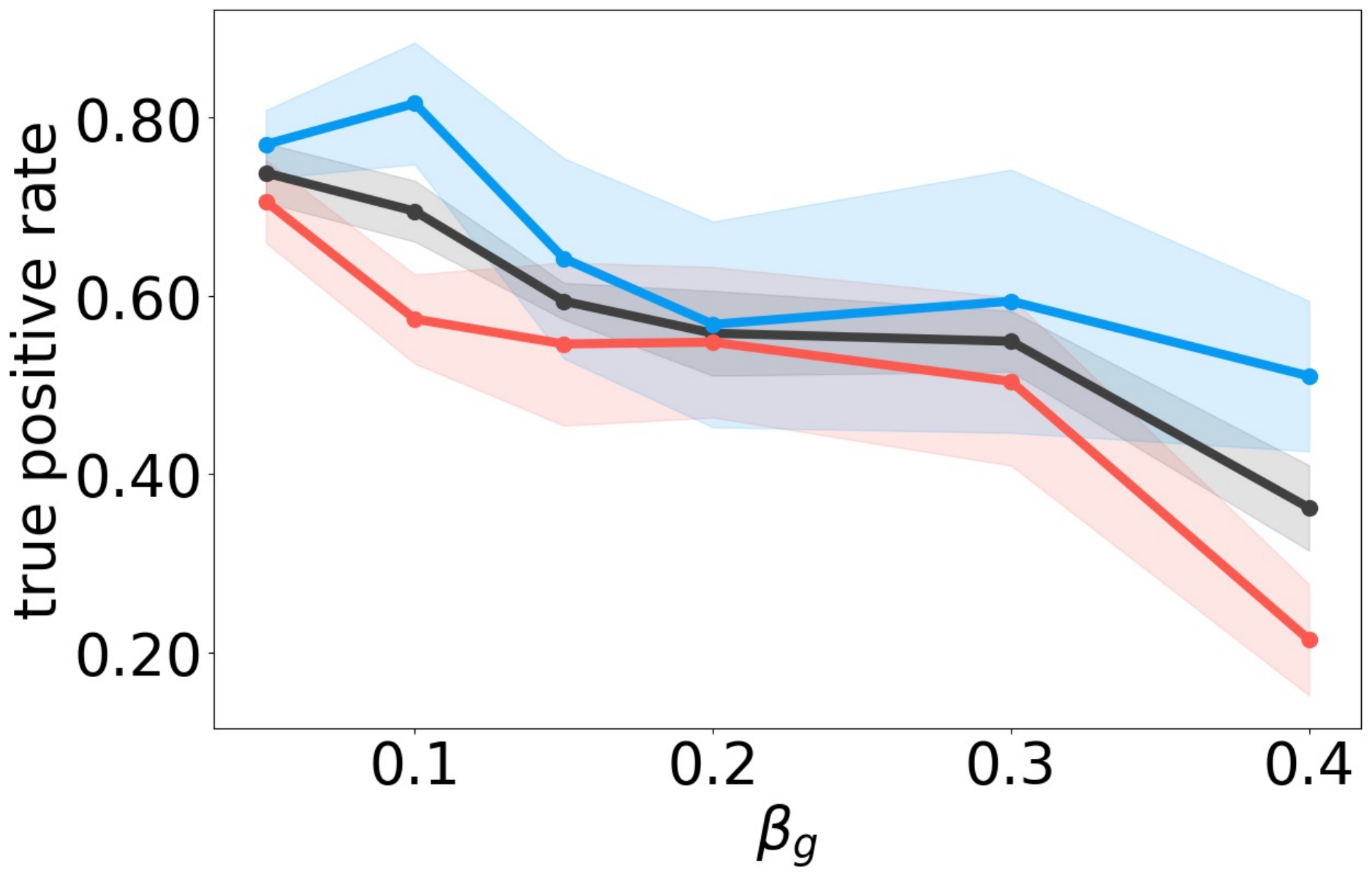}} &
     \hspace{-0.1in}{\includegraphics[width=0.195\textwidth]{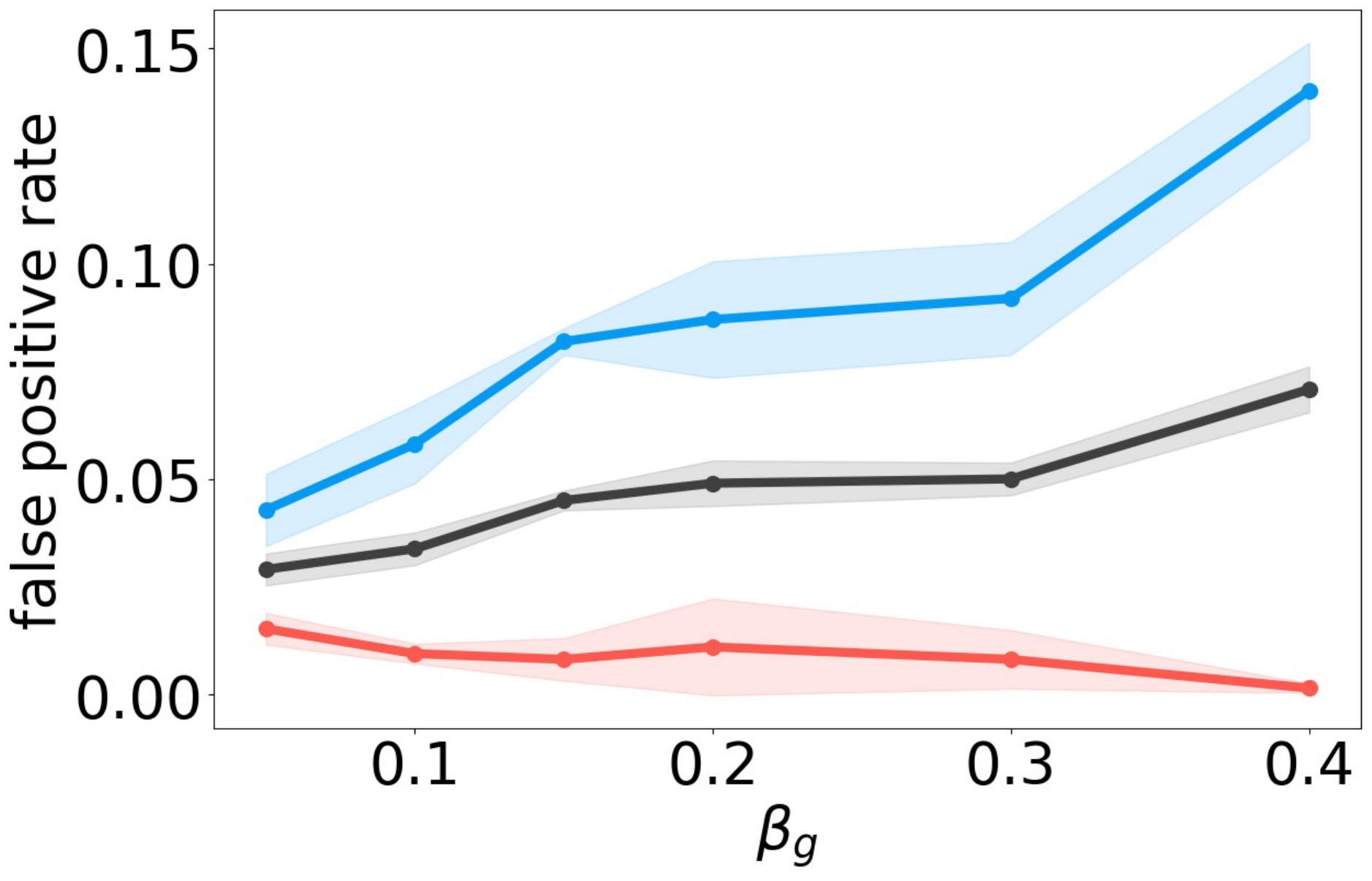}}
    &
   \hspace{-0.1in} {\includegraphics[width=0.195\textwidth]{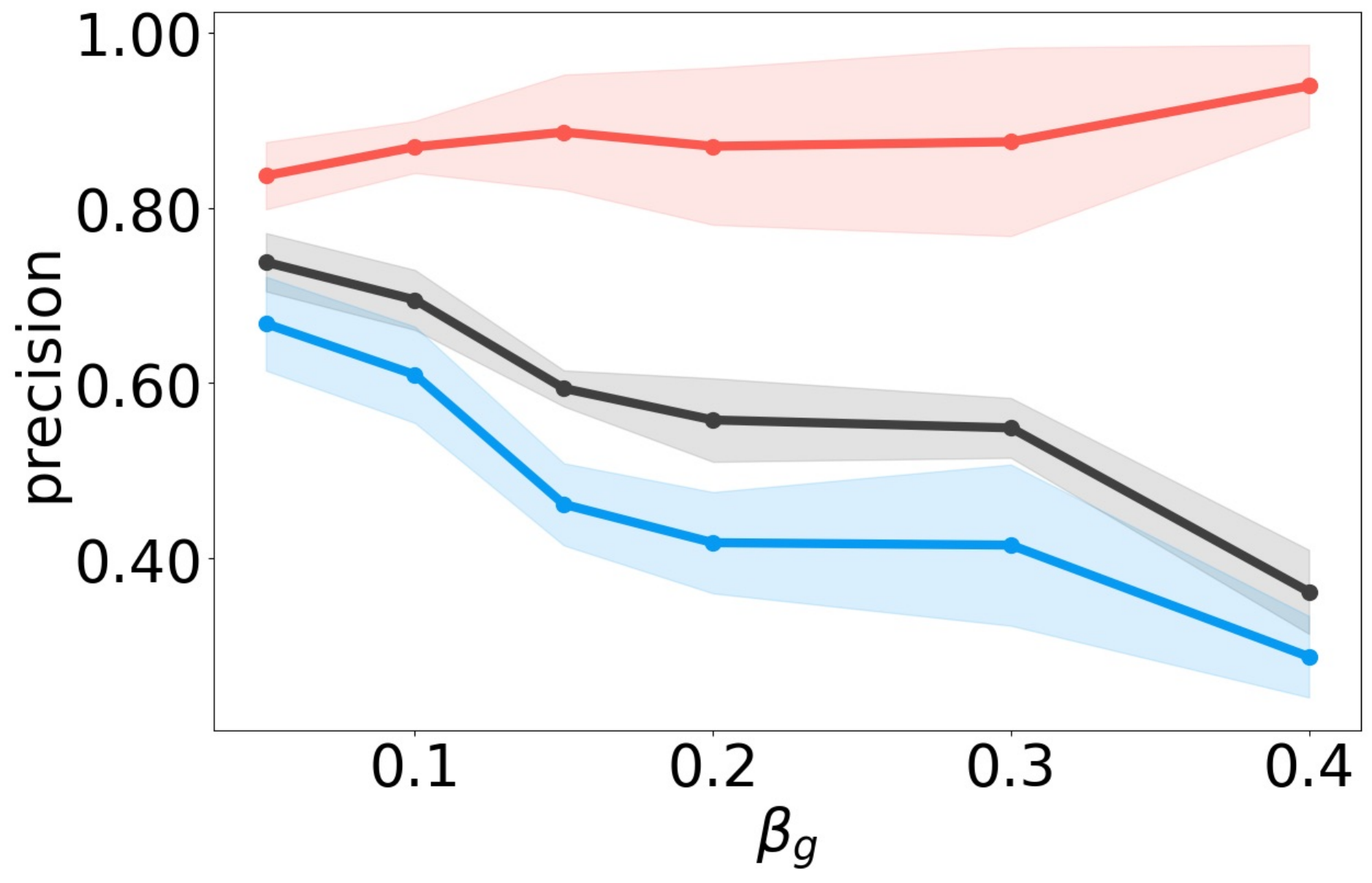}}
   &
     \hspace{-0.1in}{\includegraphics[width=0.195\textwidth]{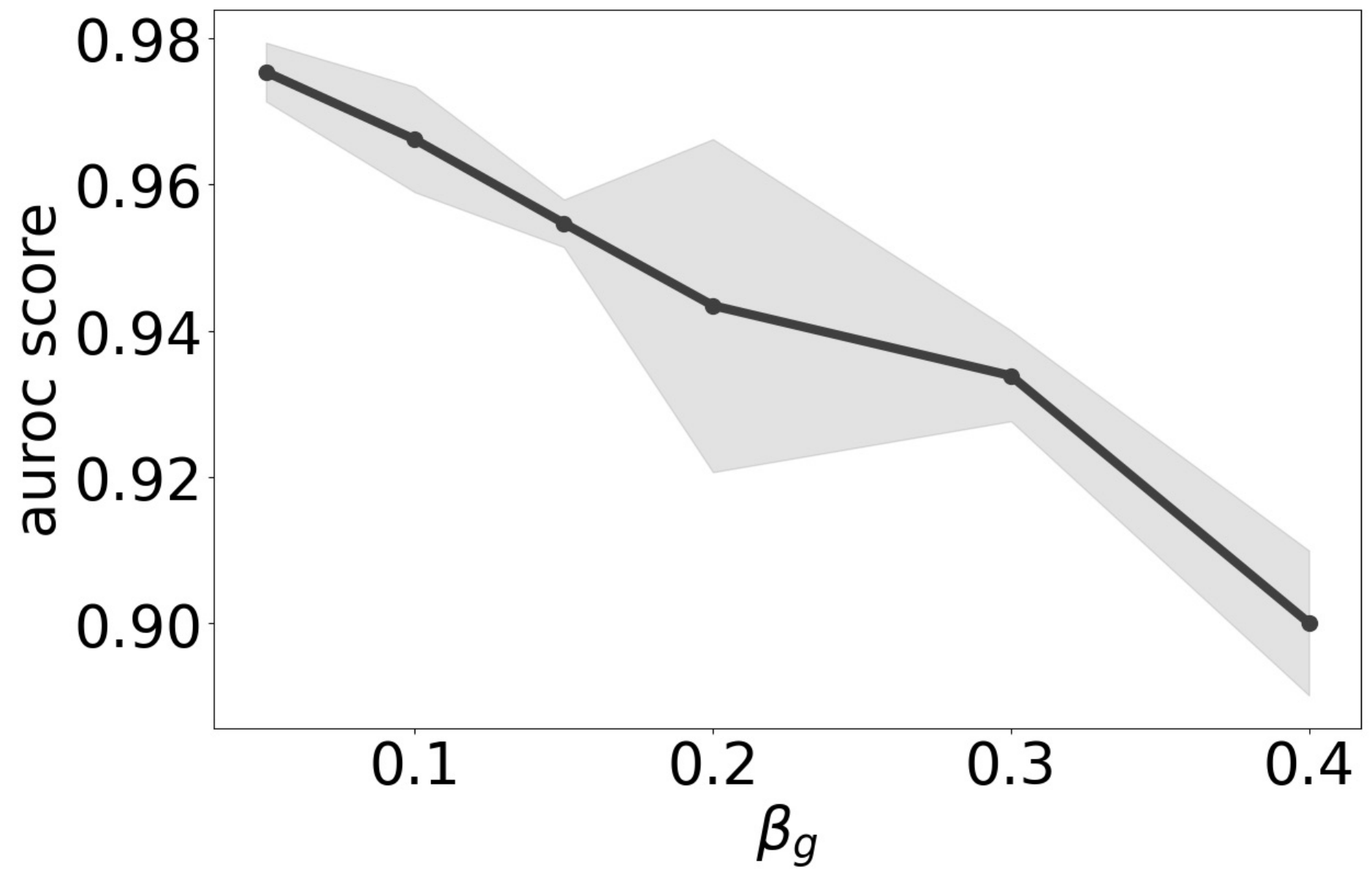}}
   \\
   \hspace{-0.1in}{\includegraphics[width=0.195\textwidth]{FIG/obfuscation/fairod/cluster/fr_legend.pdf} 
}    &
    \hspace{-0.1in}{\includegraphics[width=0.195\textwidth]{FIG/obfuscation/fairod/cluster/tpr.pdf}} &
     \hspace{-0.1in}{\includegraphics[width=0.195\textwidth]{FIG/obfuscation/fairod/cluster/fpr.pdf}}
    &
   \hspace{-0.1in} {\includegraphics[width=0.195\textwidth]{FIG/obfuscation/fairod/cluster/prec.pdf}}
   &
   \hspace{-0.1in} {\includegraphics[width=0.195\textwidth]{FIG/obfuscation/fairod/cluster/auroc.pdf}}
    \end{tabular}

    \caption{(best in color) Group-wise fairness metrics and AUROC for (top to bottom) LOF, iForest, DeepAE and FairOD under \underline{membership obfuscation bias} on \textit{clustered} outliers. }
    \label{fig:obfuscation_all_cluster}
\end{figure*}

See Fig. \ref{fig:obfuscation_all_cluster}.

\subsubsection{Unequal base rates}
\label{assec:ubs_cluster}

\begin{figure*}[!ht]
    \centering
    \begin{tabular}{lllll}
    \hspace{-0.1in}{\includegraphics[width=0.195\textwidth]{FIG/baserate/lof/cluster/fr_legend.pdf}} 
    &
    \hspace{-0.1in}{\includegraphics[width=0.195\textwidth]{FIG/baserate/lof/cluster/tpr.pdf}} &
     \hspace{-0.1in}{\includegraphics[width=0.195\textwidth]{FIG/baserate/lof/cluster/fpr.pdf}}
    &
   \hspace{-0.1in} {\includegraphics[width=0.195\textwidth]{FIG/baserate/lof/cluster/prec.pdf}}
   &
   \hspace{-0.1in}
   {\includegraphics[width=0.195\textwidth]{FIG/baserate/lof/cluster/auroc.pdf}} 
\\
\hspace{-0.1in}{\includegraphics[width=0.195\textwidth]{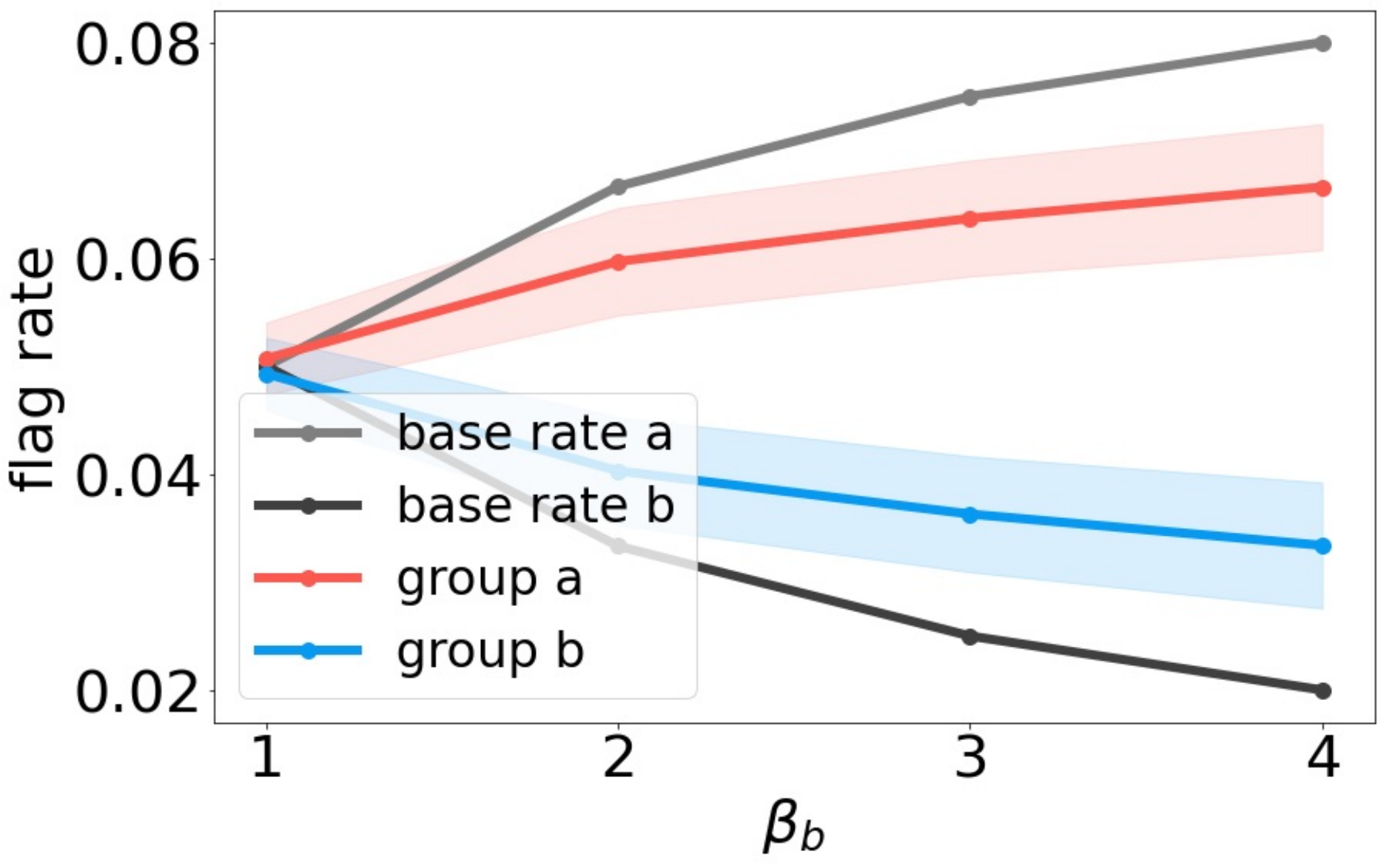}} 
    &
    \hspace{-0.1in}{\includegraphics[width=0.195\textwidth]{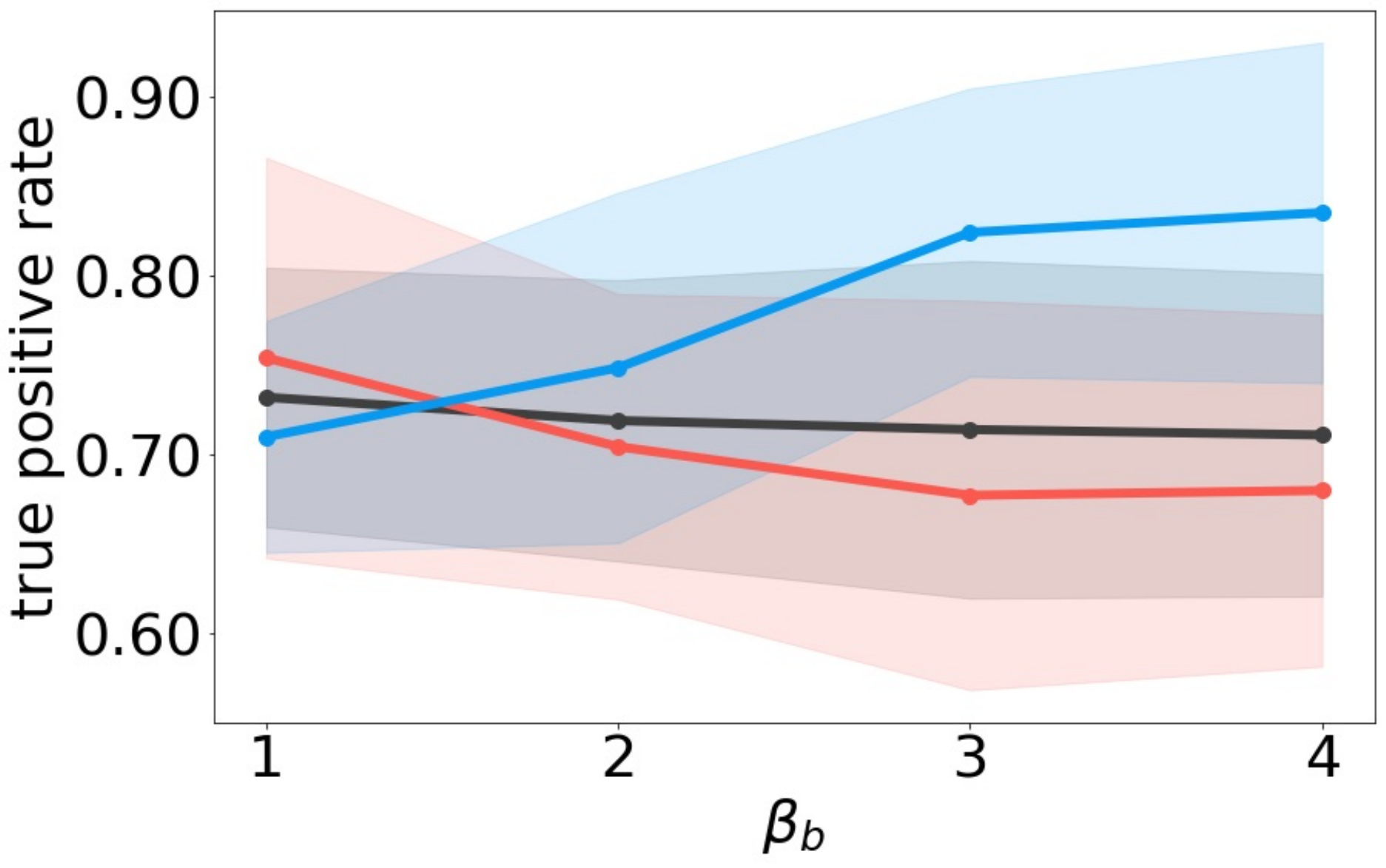}} &
     \hspace{-0.1in}{\includegraphics[width=0.195\textwidth]{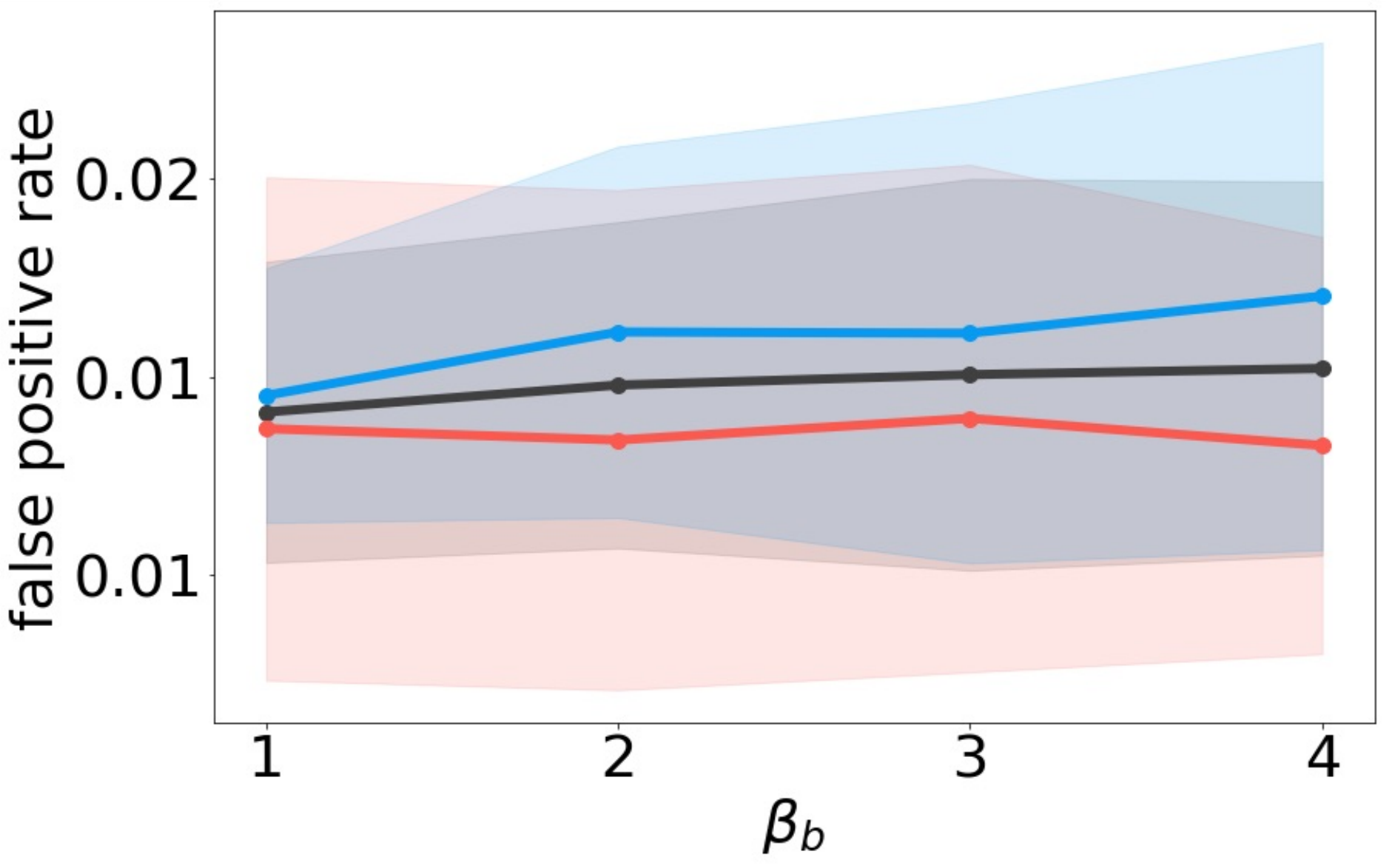}}
    &
   \hspace{-0.1in} {\includegraphics[width=0.195\textwidth]{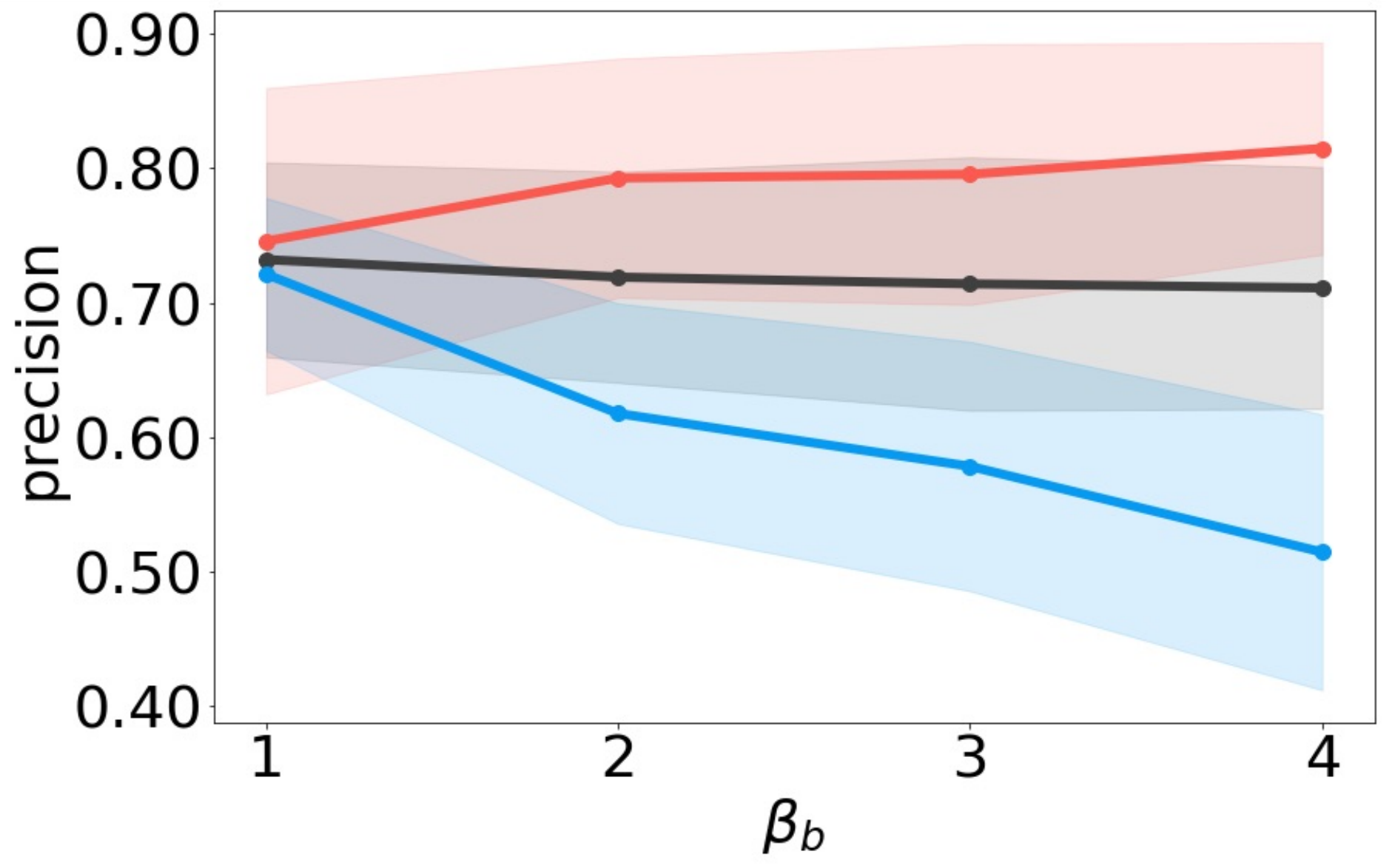}}
   &
    \hspace{-0.1in}{\includegraphics[width=0.195\textwidth]{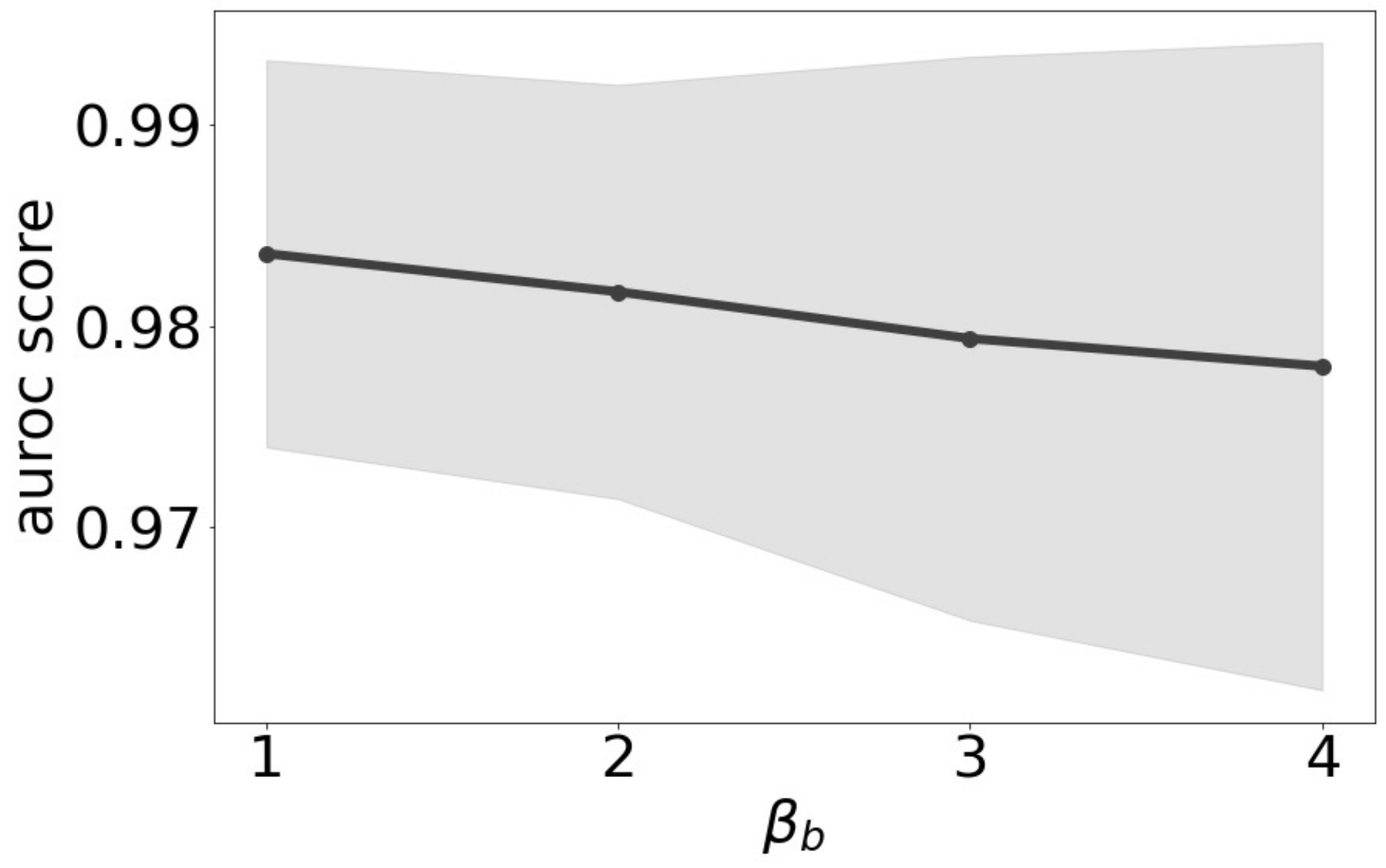}}
   \\
     \hspace{-0.1in}{\includegraphics[width=0.195\textwidth]{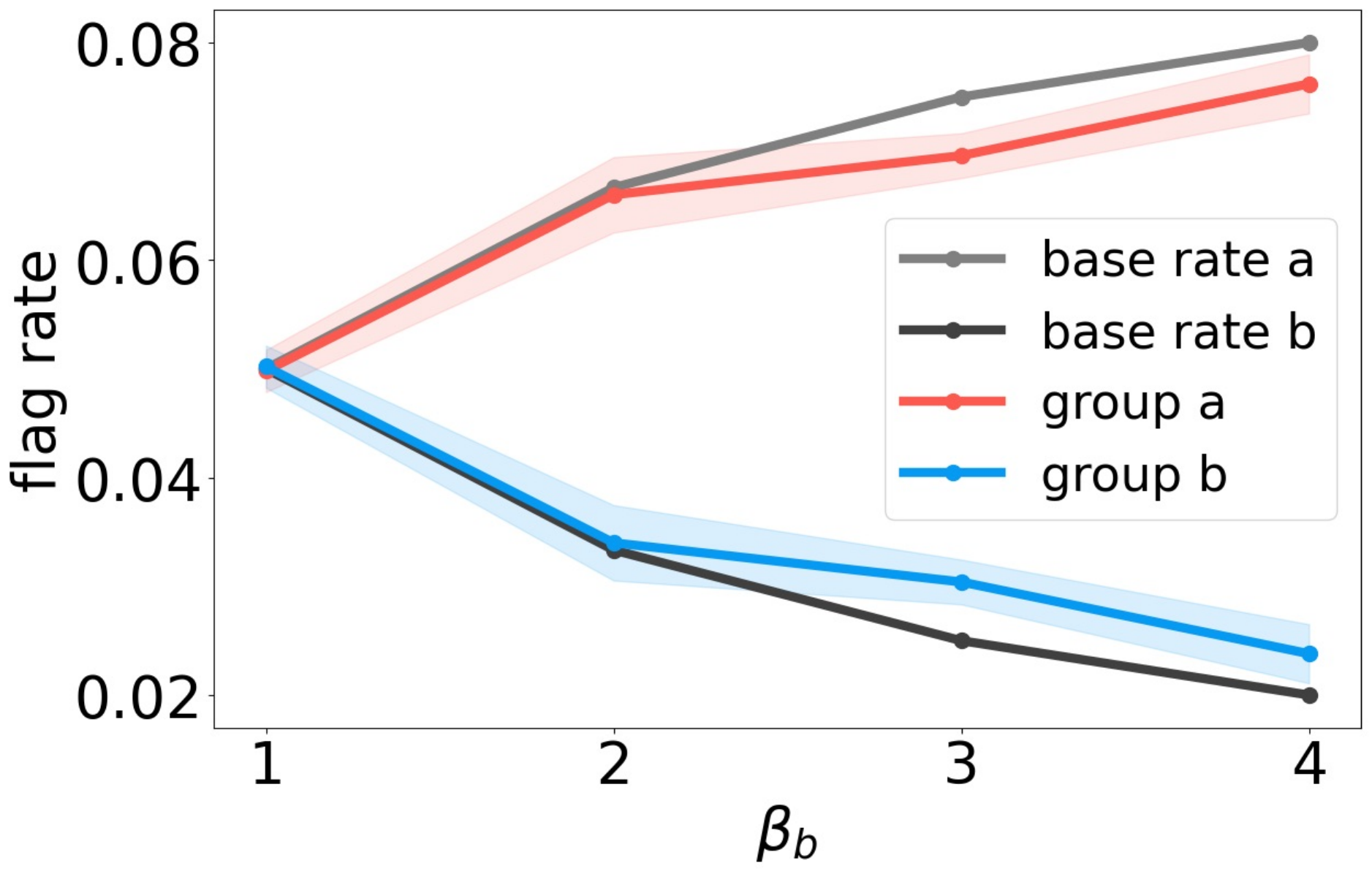}} 
    &
    \hspace{-0.1in}{\includegraphics[width=0.195\textwidth]{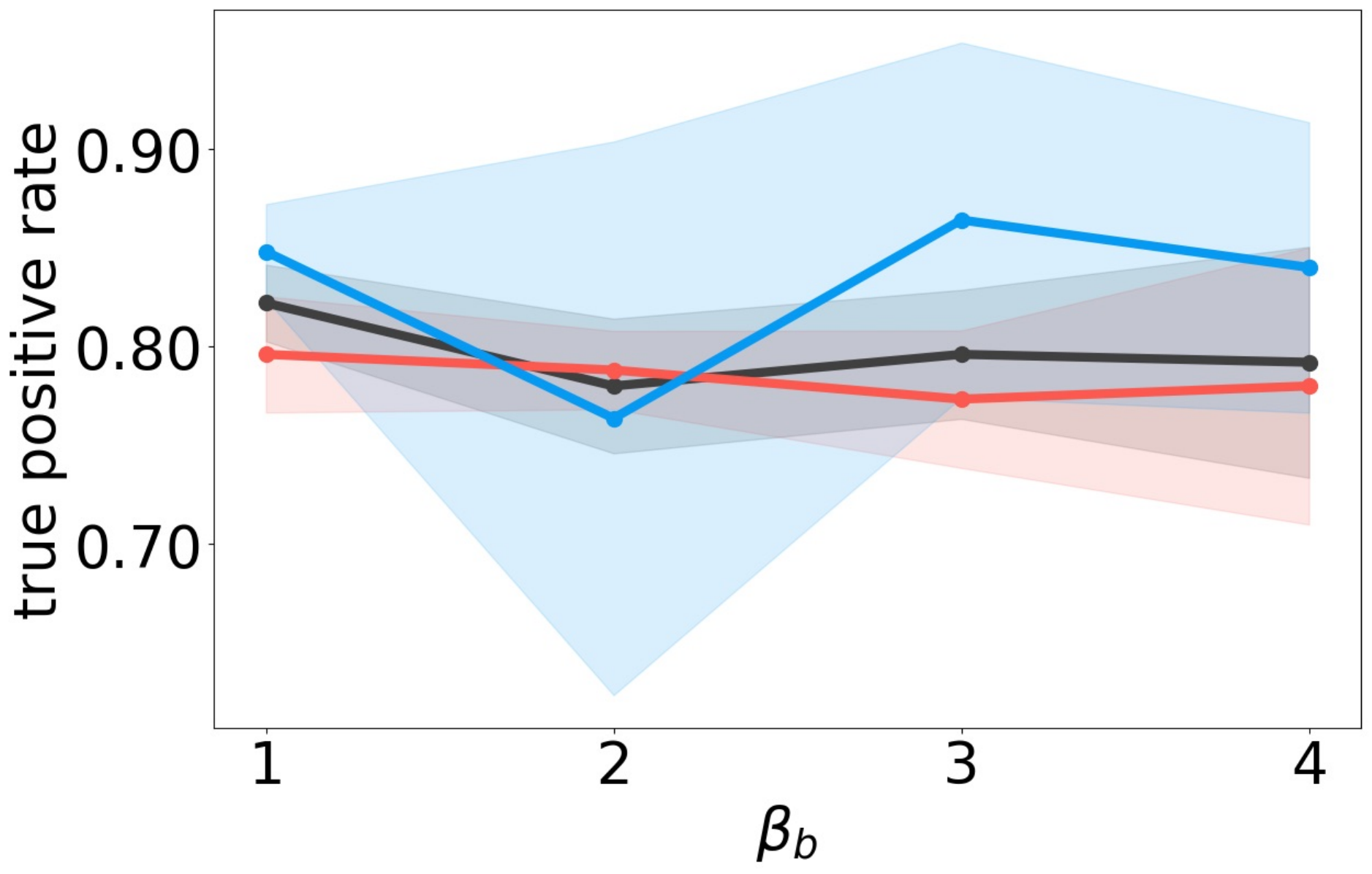}} &
     \hspace{-0.1in}{\includegraphics[width=0.195\textwidth]{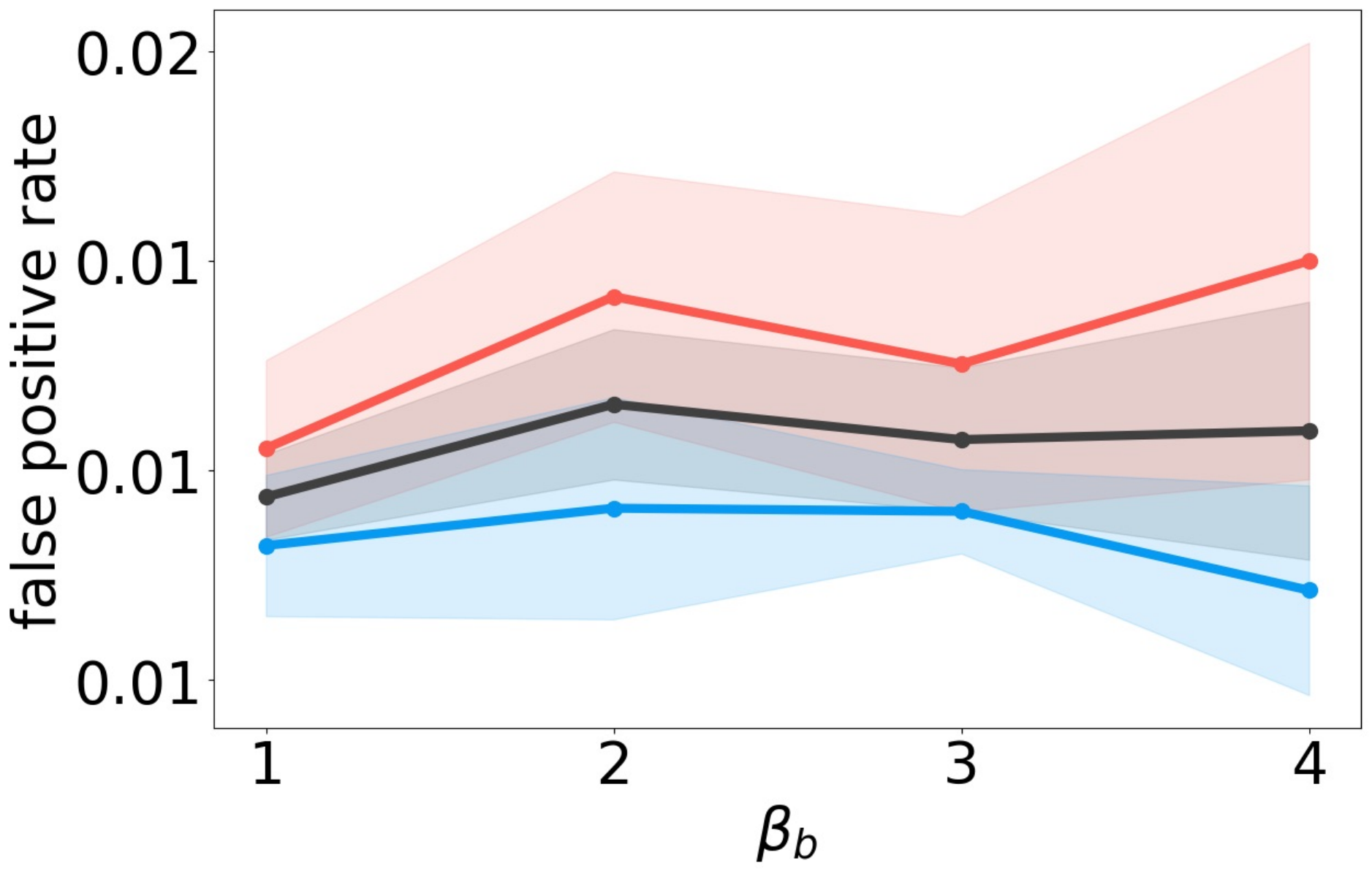}}
    &
   \hspace{-0.1in} {\includegraphics[width=0.195\textwidth]{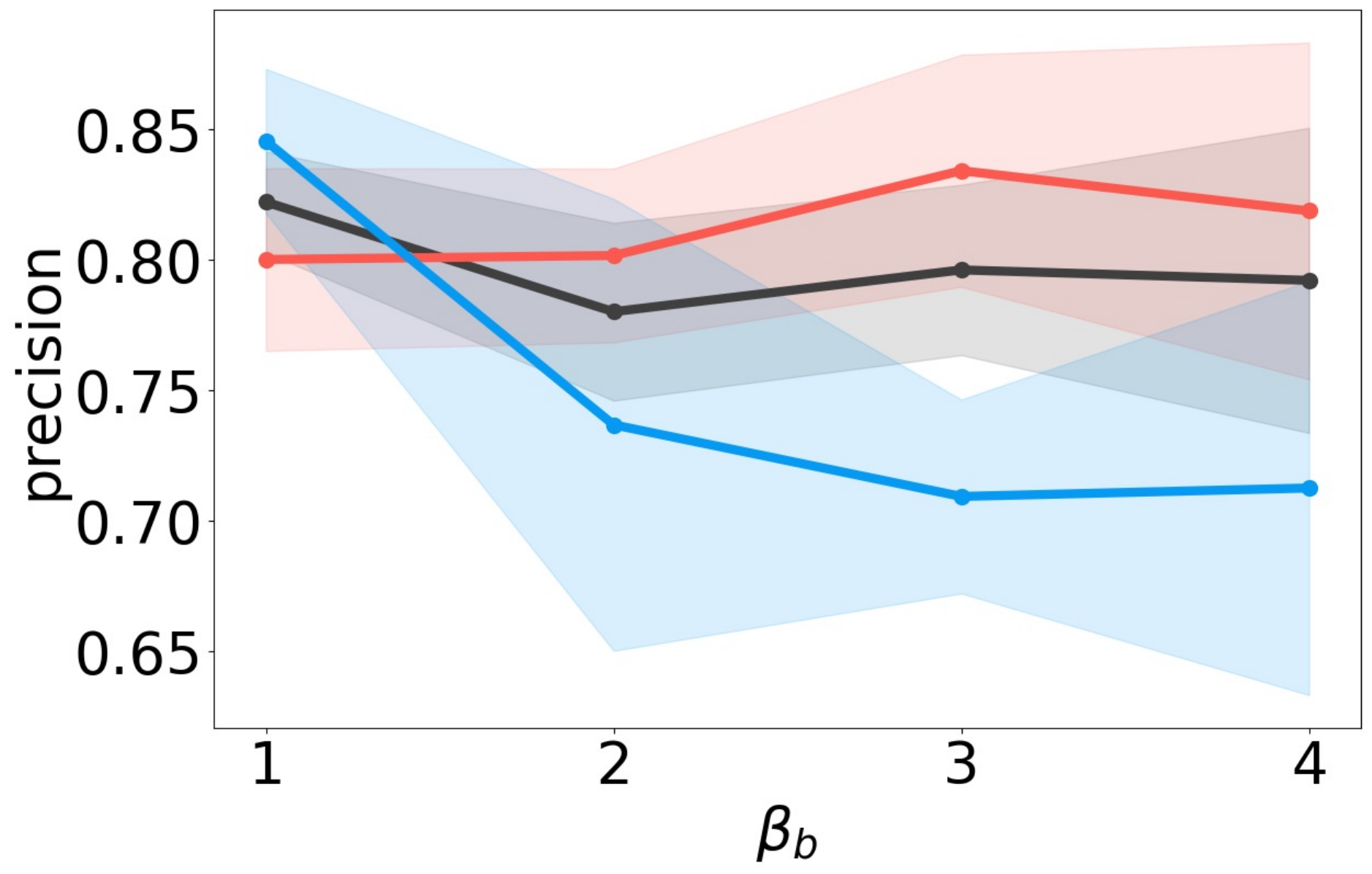}}
   &
     \hspace{-0.1in}{\includegraphics[width=0.195\textwidth]{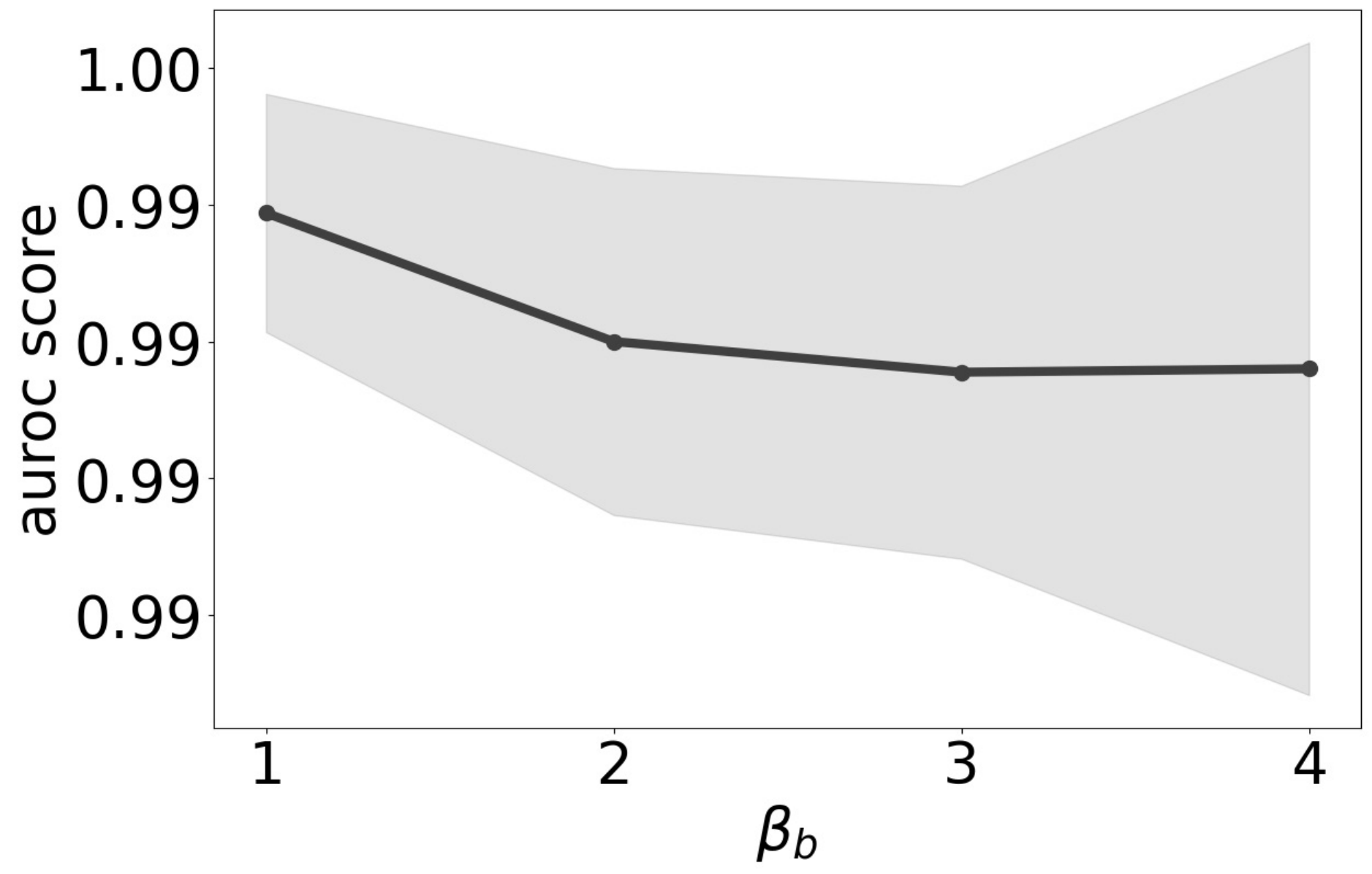}}
   \\
   \hspace{-0.1in}{\includegraphics[width=0.195\textwidth]{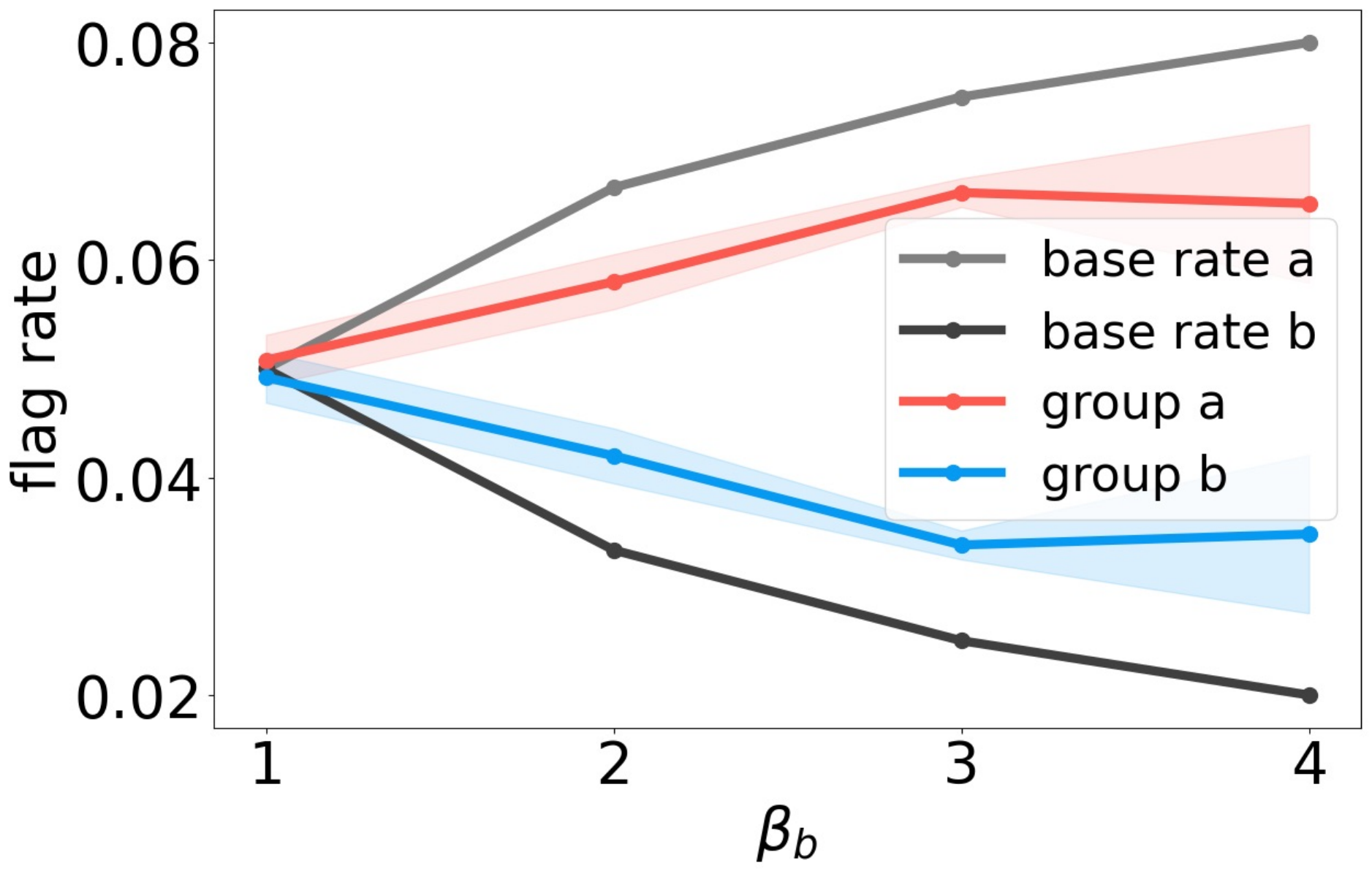} }
    &
    \hspace{-0.1in}{\includegraphics[width=0.195\textwidth]{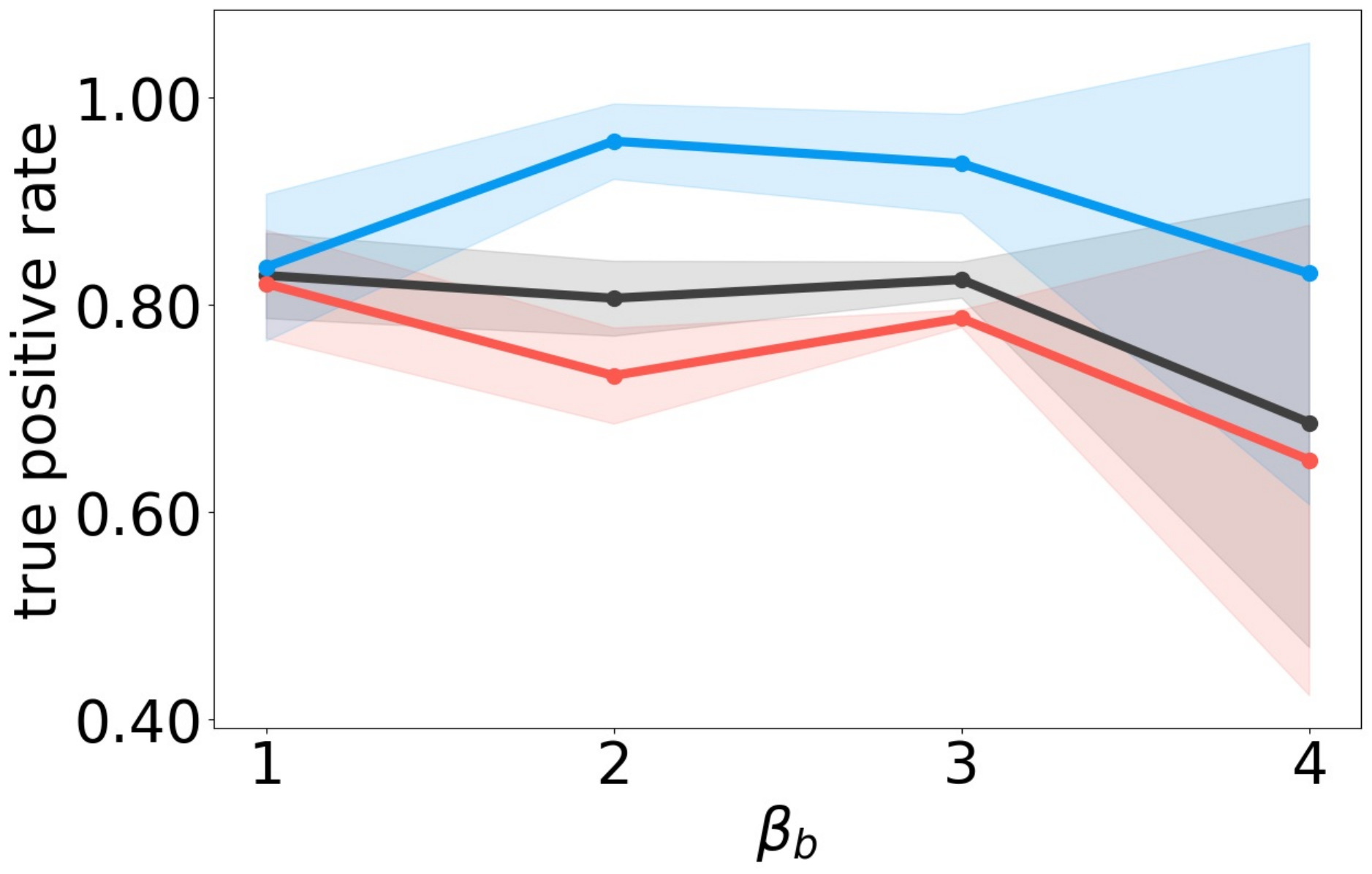}} &
     \hspace{-0.1in}{\includegraphics[width=0.195\textwidth]{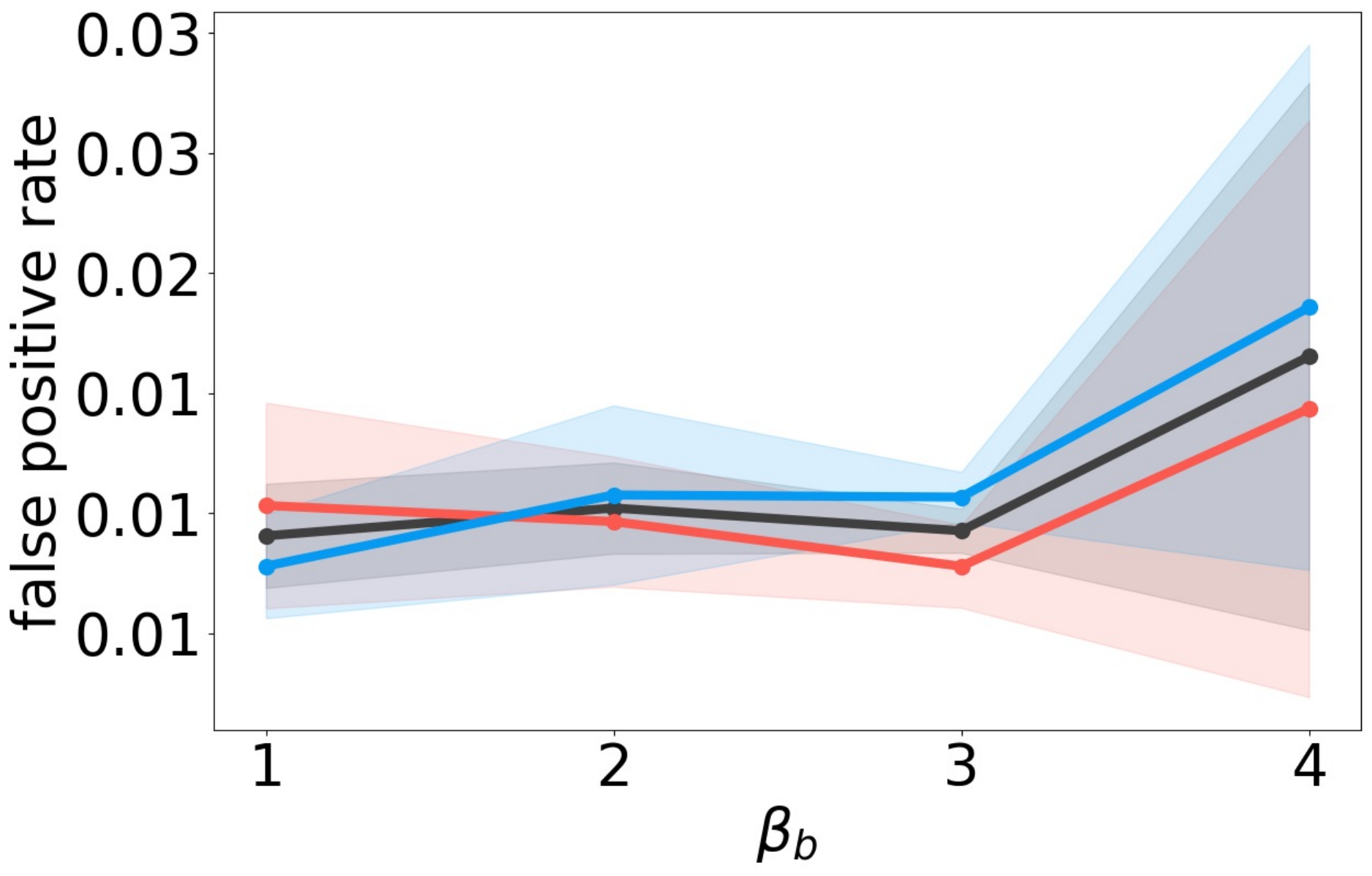}}
    &
   \hspace{-0.1in} {\includegraphics[width=0.195\textwidth]{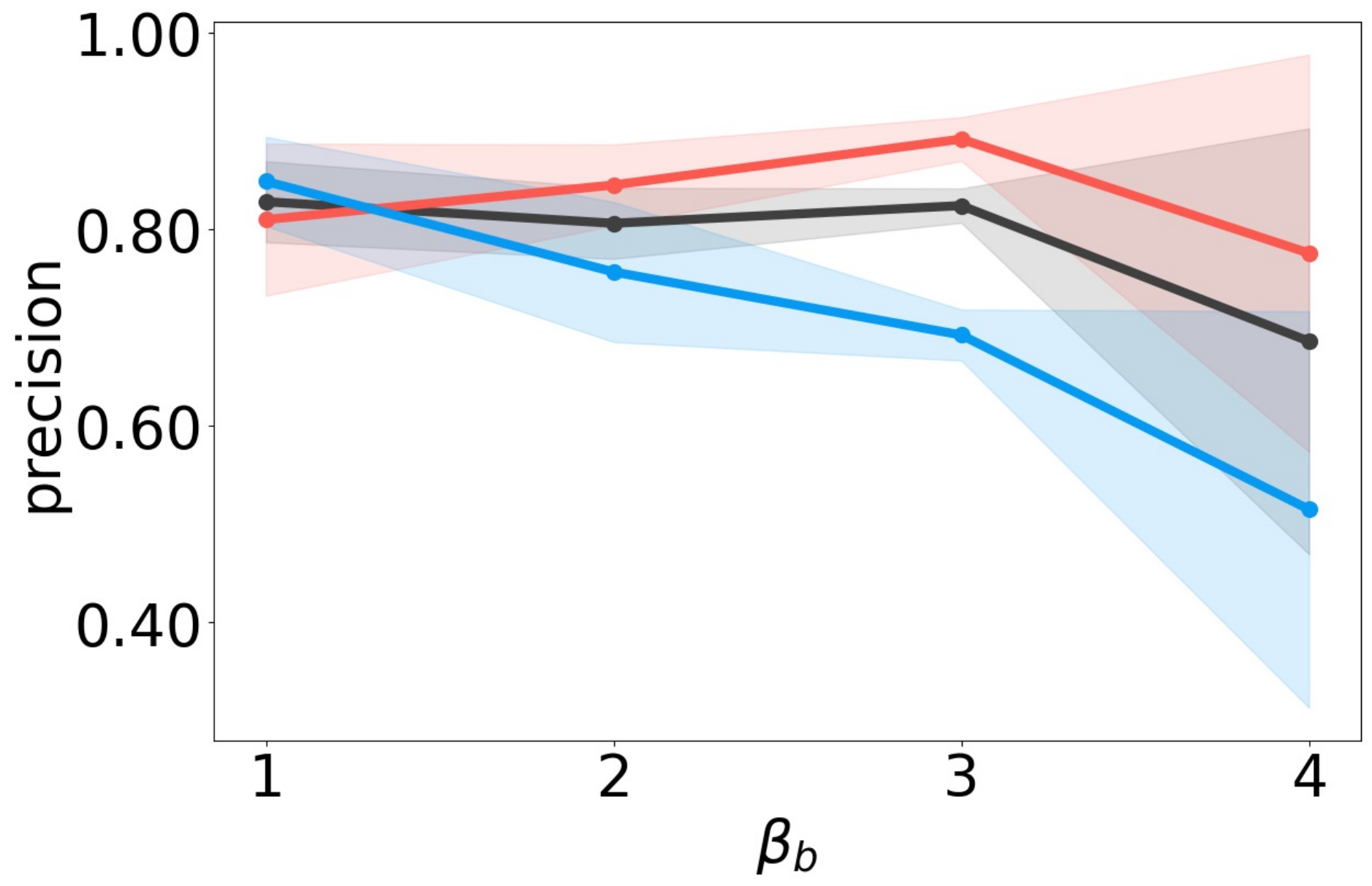}}
   &
   \hspace{-0.1in} {\includegraphics[width=0.195\textwidth]{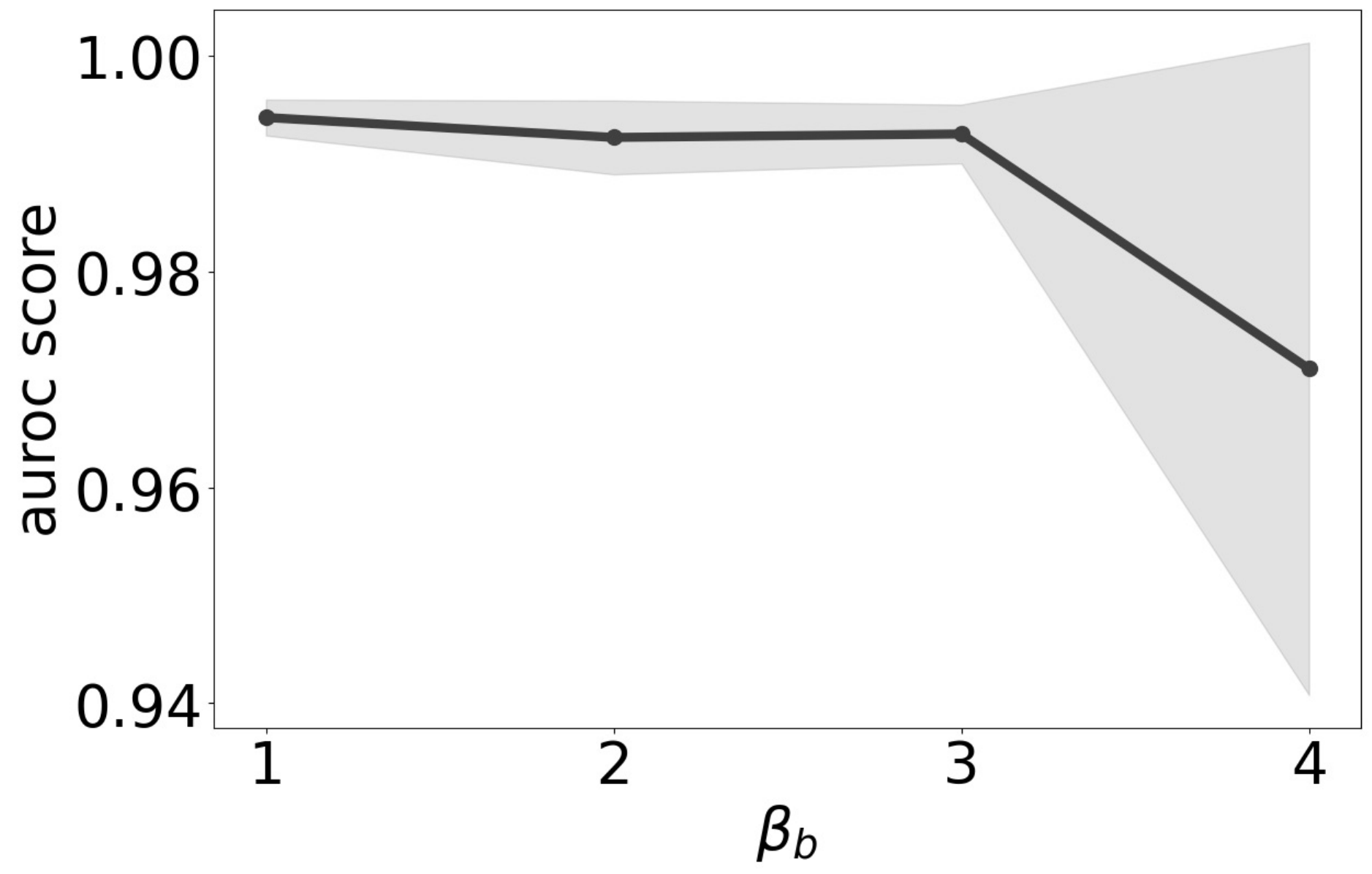}}
    \end{tabular}
    \caption{(best in color) Group-wise fairness metrics and AUROC for (top to bottom) LOF, iForest, DeepAE and FairOD under \underline{unequal base rates} on \textit{clustered} outliers. }
    \label{fig:baserate_all_cluster}
\end{figure*}

See Fig. \ref{fig:baserate_all_cluster}.

\subsection{Datasets with Scattered Outliers}
\label{asec:scatter_results}

\subsubsection{Group sample size bias}
\label{assec:sizebias_scatter}

\begin{figure*}[!ht]
    \centering
    \begin{tabular}{lllll}
     \hspace{-0.1in}{\includegraphics[width=0.195\textwidth]{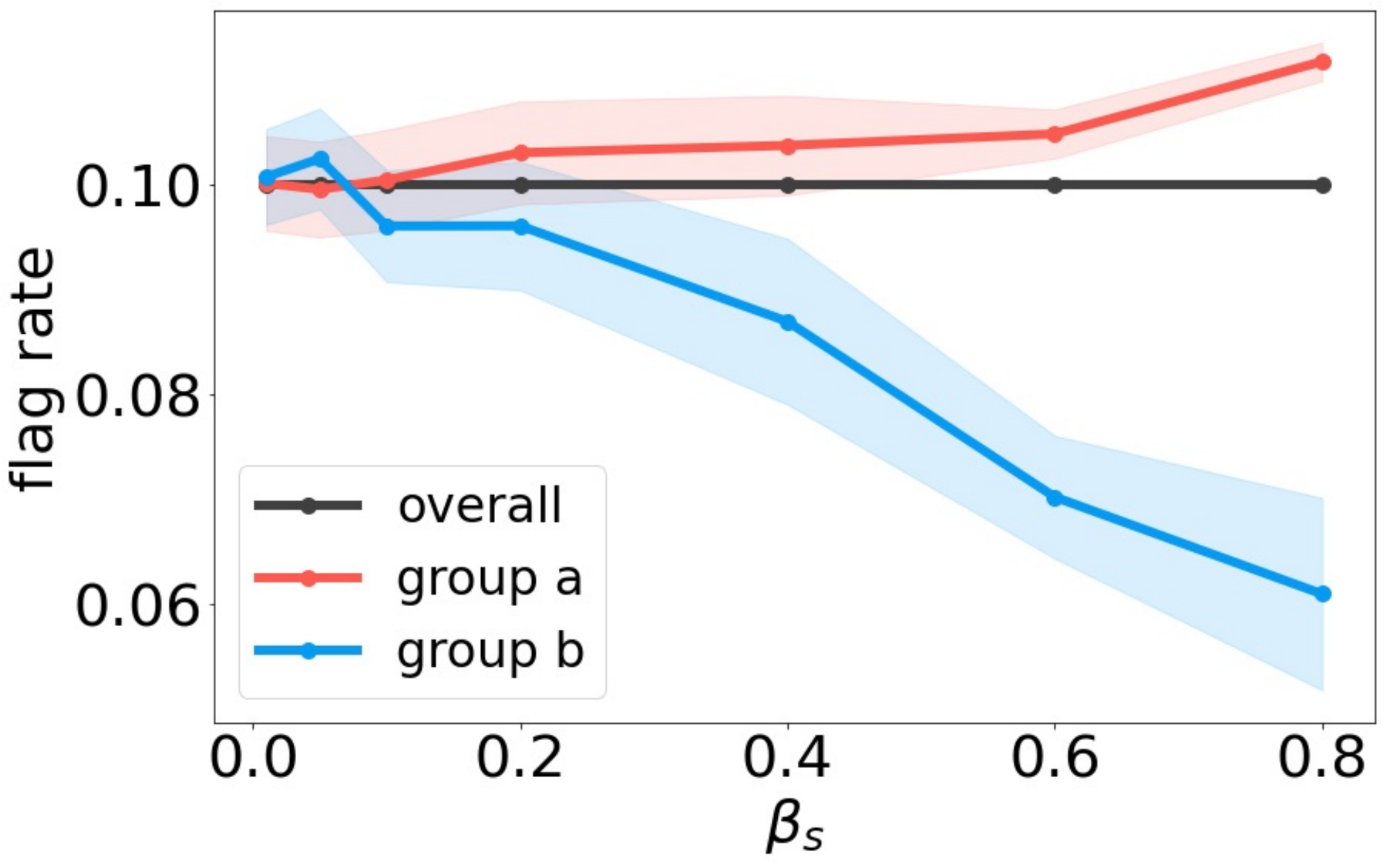}} 
    &
    \hspace{-0.1in}{\includegraphics[width=0.195\textwidth]{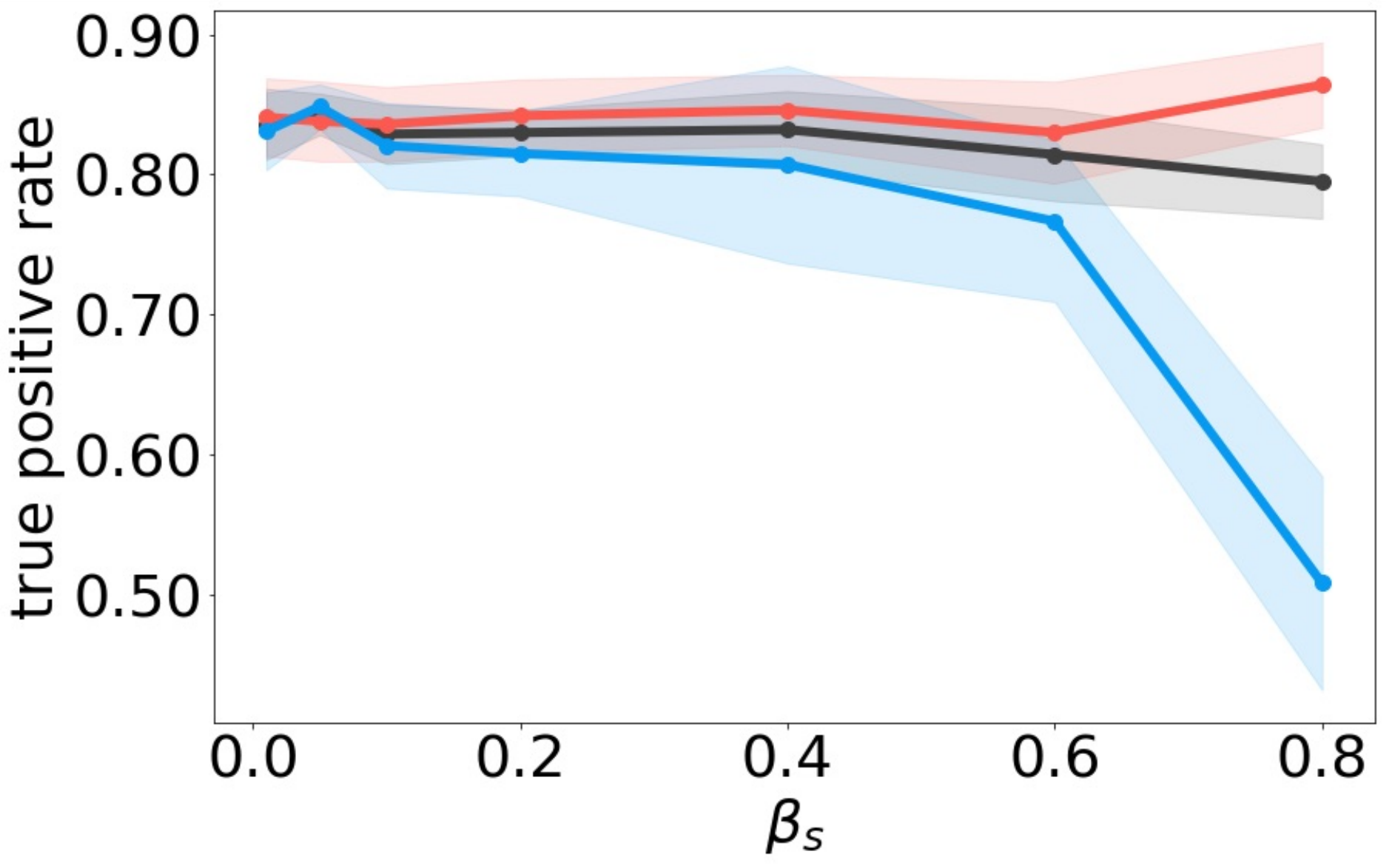}} &
     \hspace{-0.1in}{\includegraphics[width=0.195\textwidth]{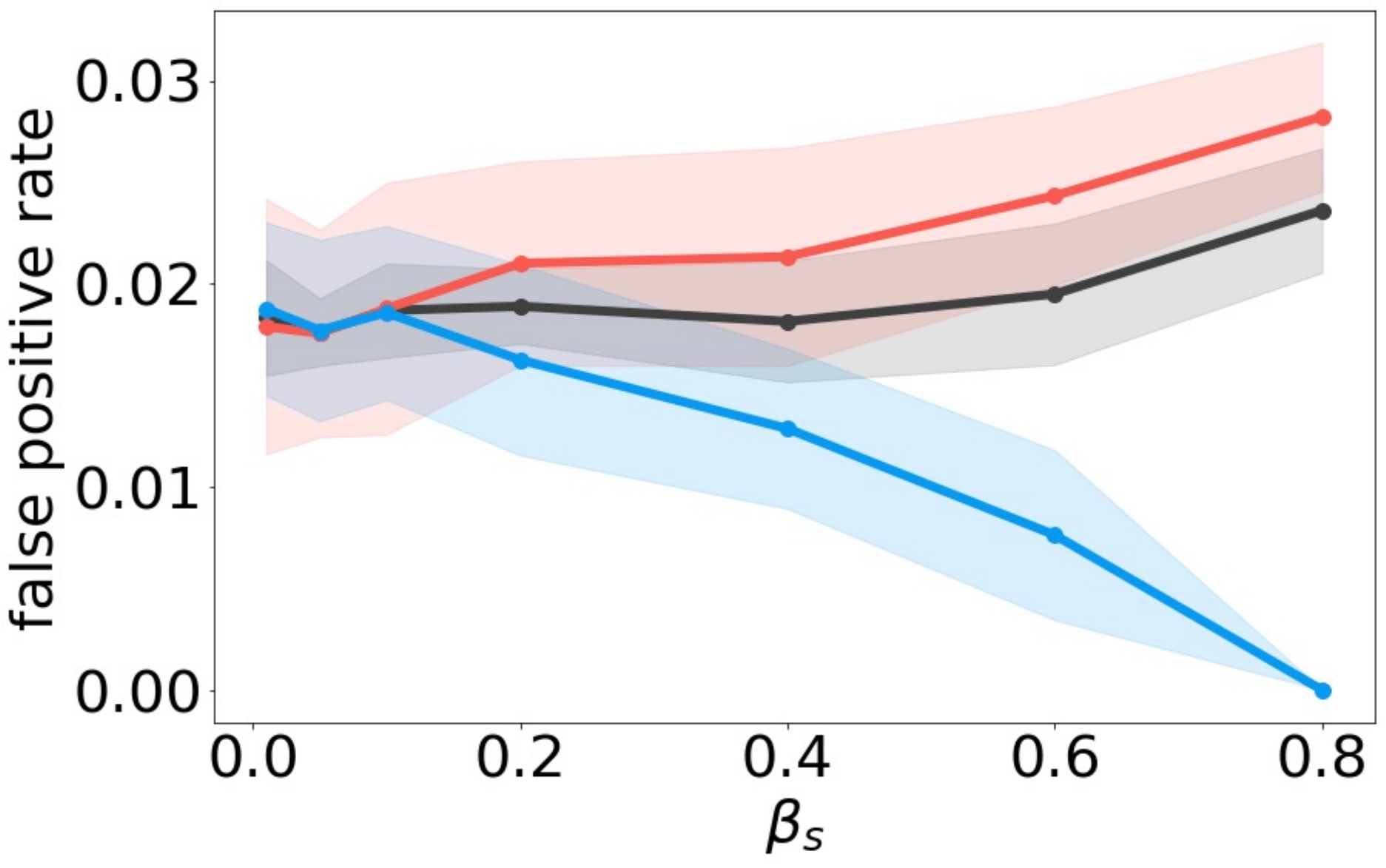}}
    &
   \hspace{-0.1in} {\includegraphics[width=0.195\textwidth]{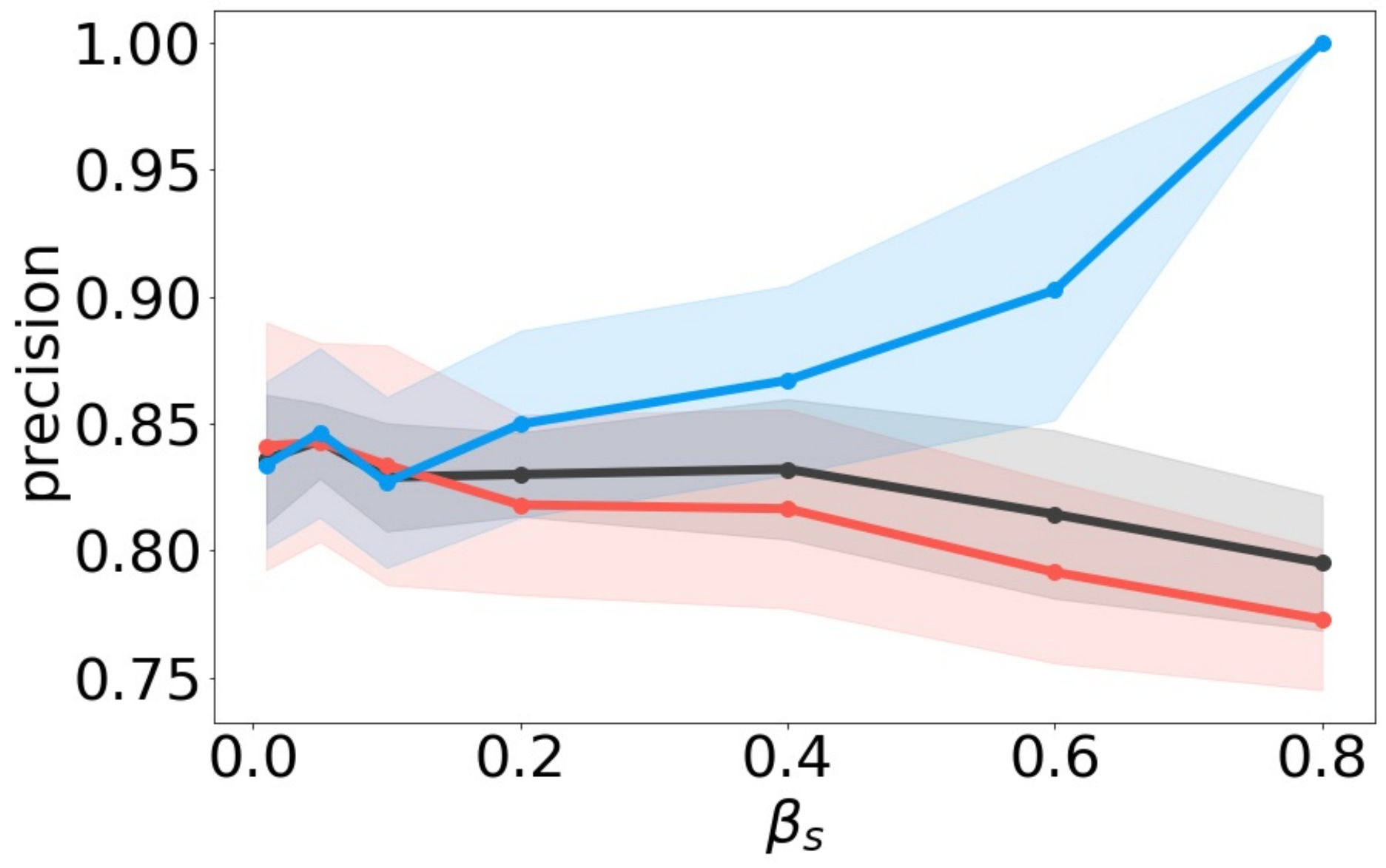}}
   &
    \hspace{-0.1in}{\includegraphics[width=0.195\textwidth]{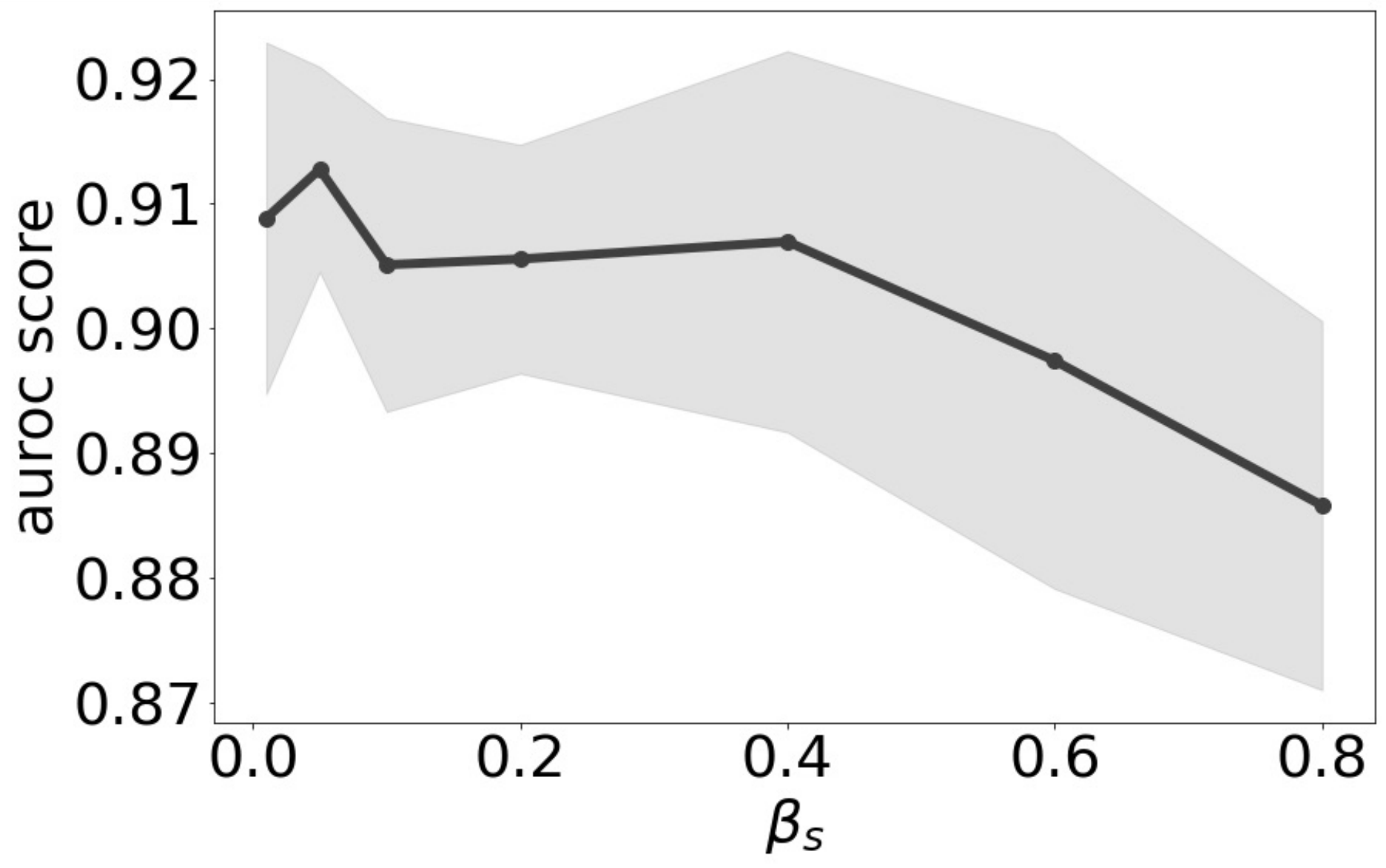}} 
   \\
     \hspace{-0.1in}{\includegraphics[width=0.195\textwidth]{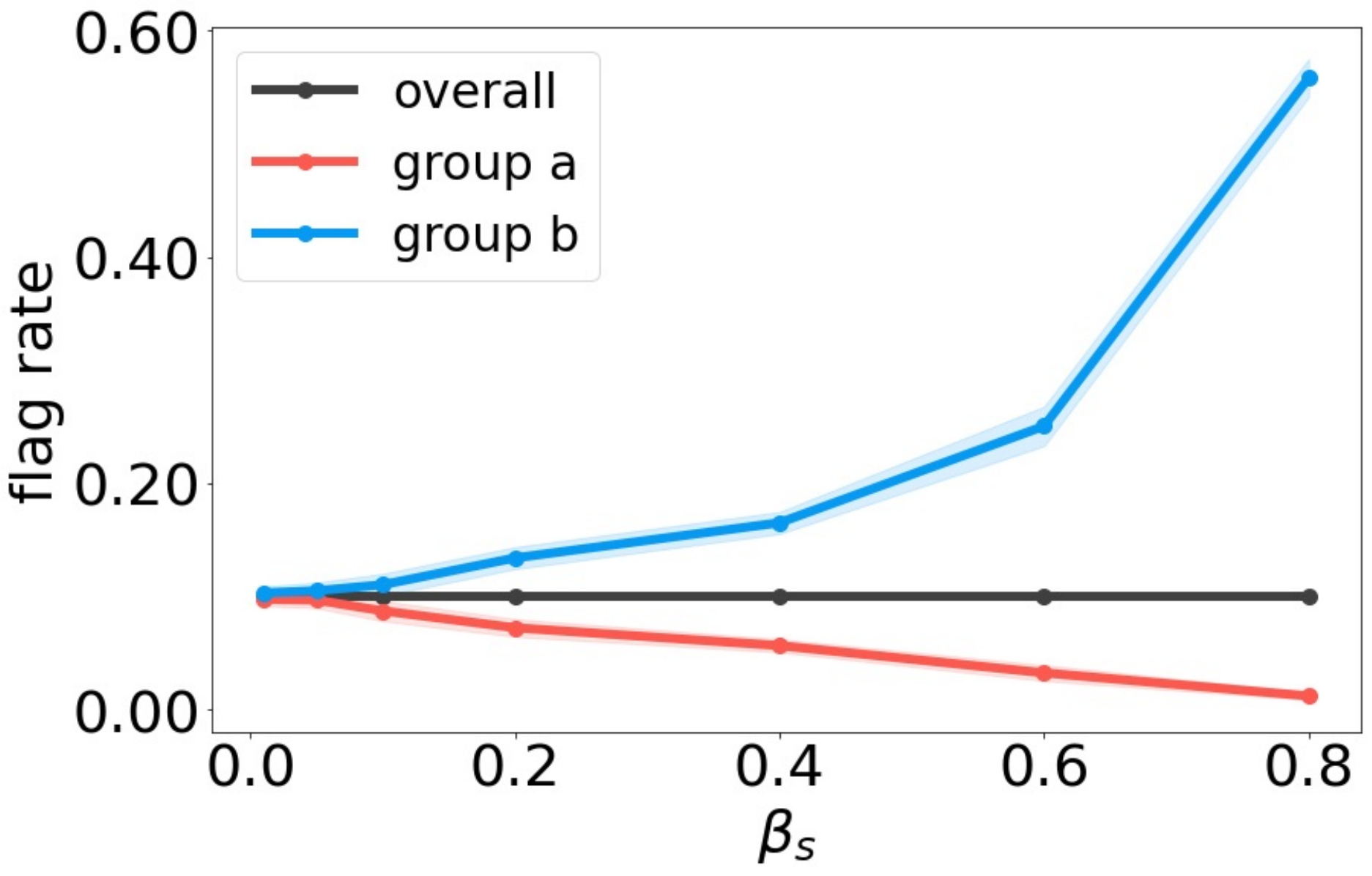}} 
    &
    \hspace{-0.1in}{\includegraphics[width=0.195\textwidth]{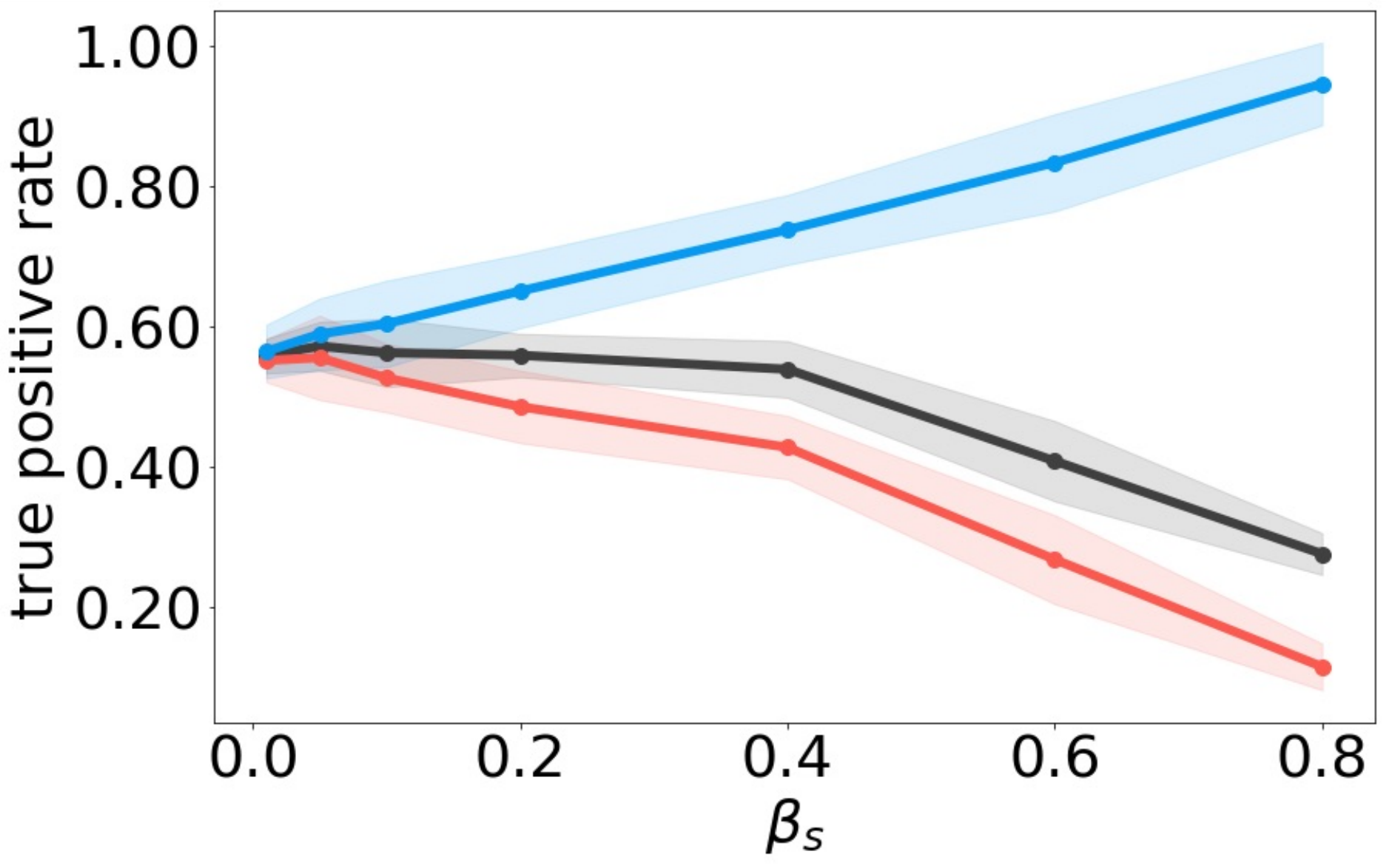}} &
     \hspace{-0.1in}{\includegraphics[width=0.195\textwidth]{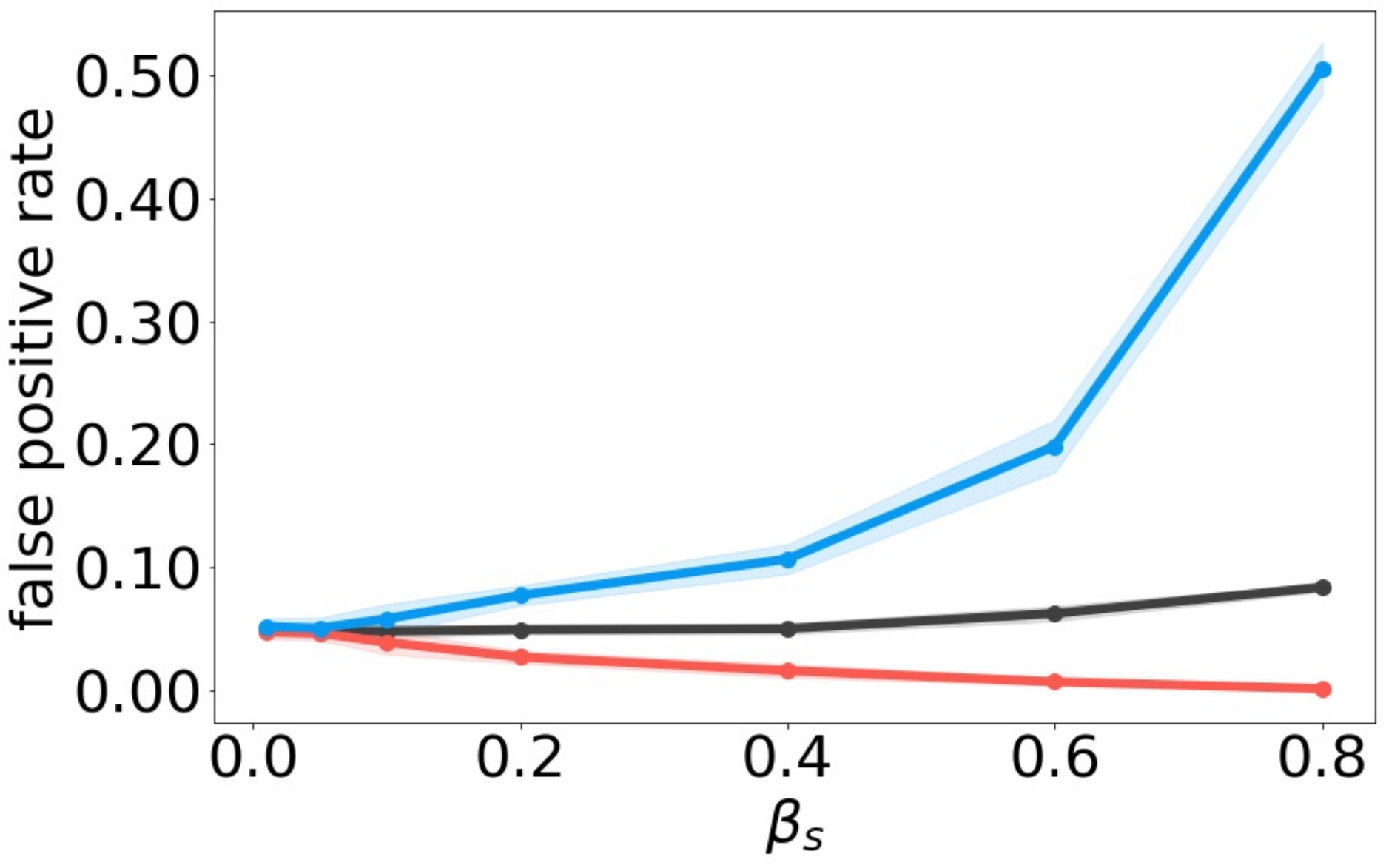}}
    &
   \hspace{-0.1in} {\includegraphics[width=0.195\textwidth]{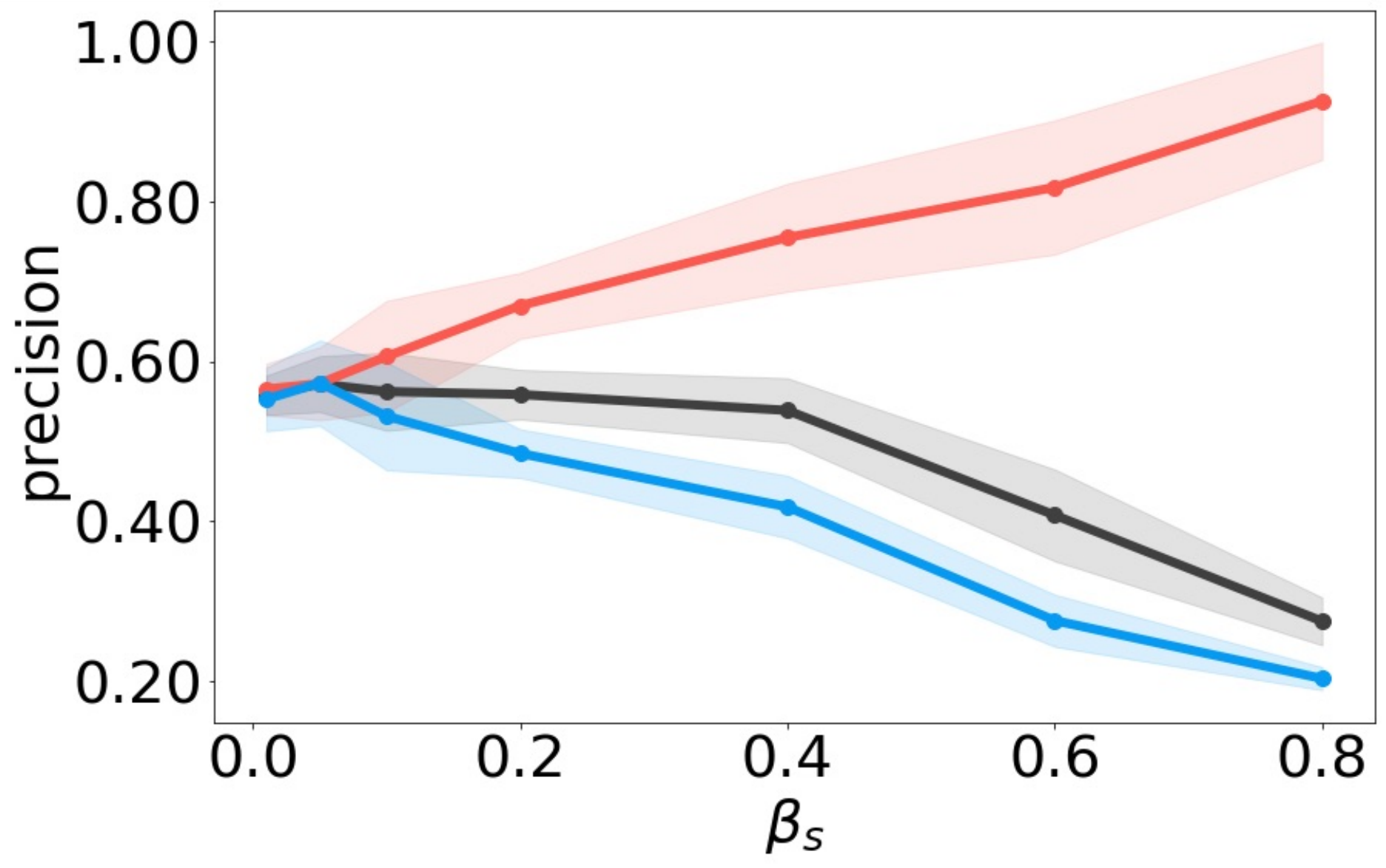}}
   &
    \hspace{-0.1in}{\includegraphics[width=0.195\textwidth]{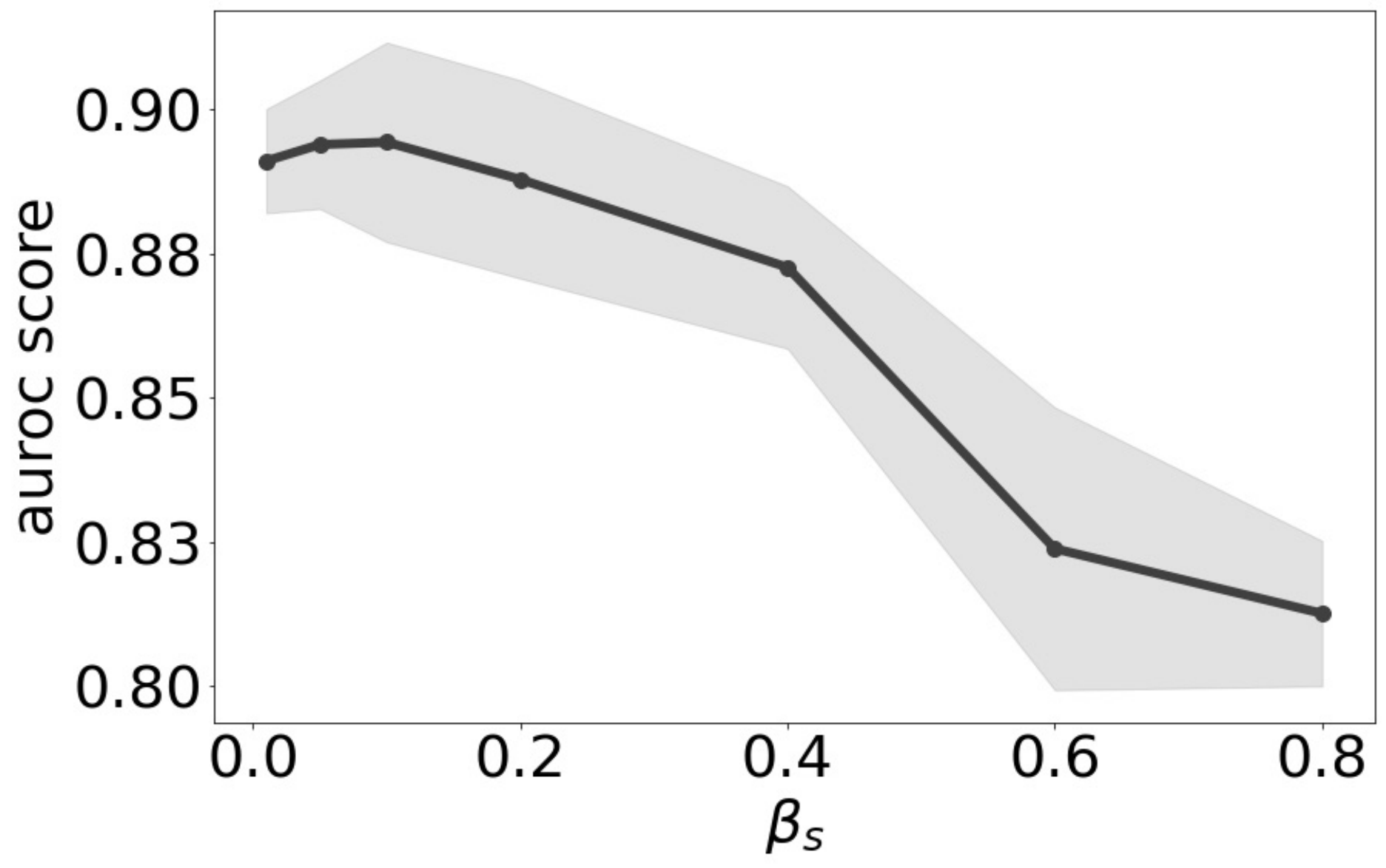}} 
   \\
    \hspace{-0.1in}{\includegraphics[width=0.195\textwidth]{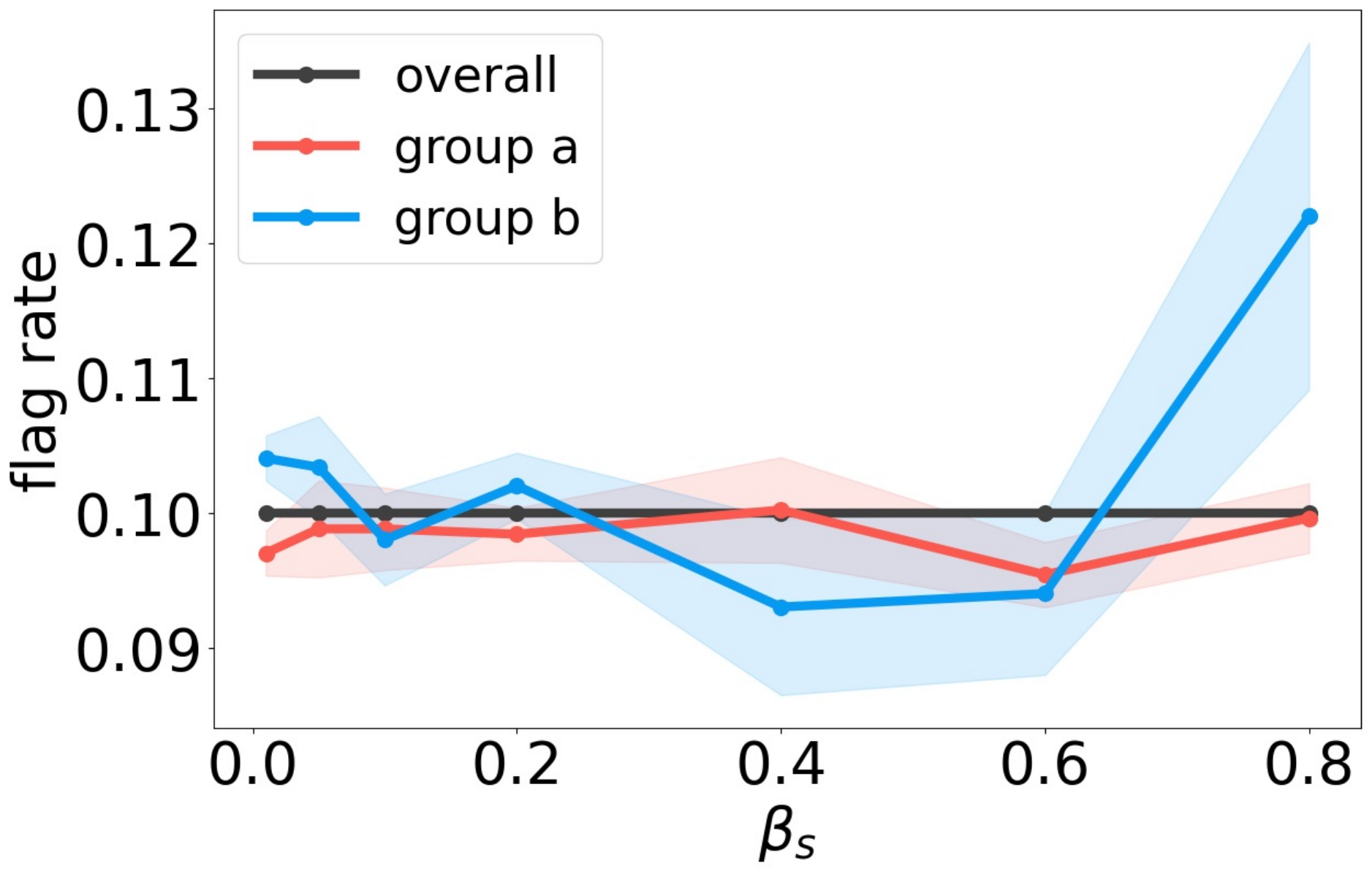}} 
    &
    \hspace{-0.1in}{\includegraphics[width=0.195\textwidth]{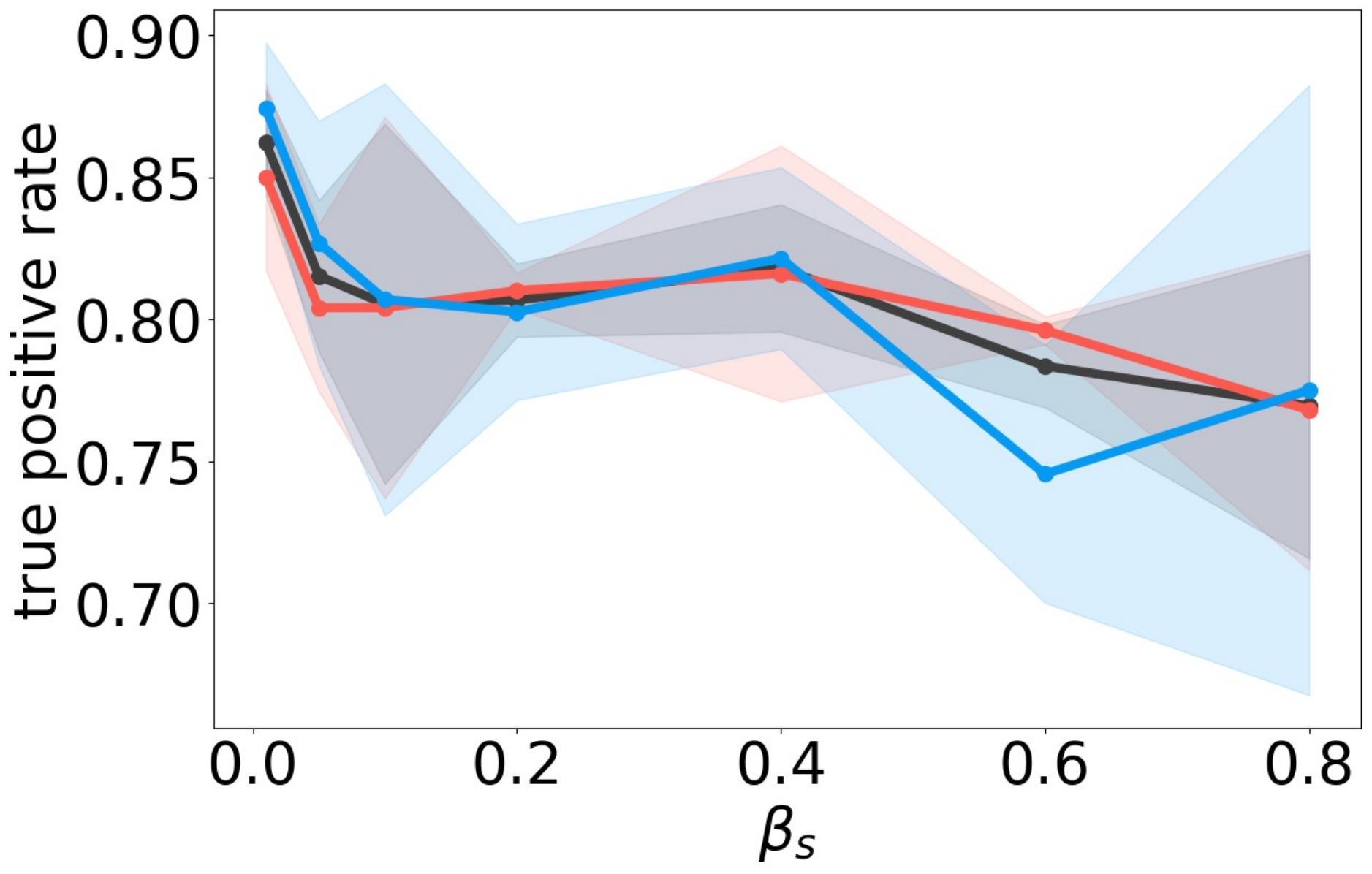}} &
     \hspace{-0.1in}{\includegraphics[width=0.195\textwidth]{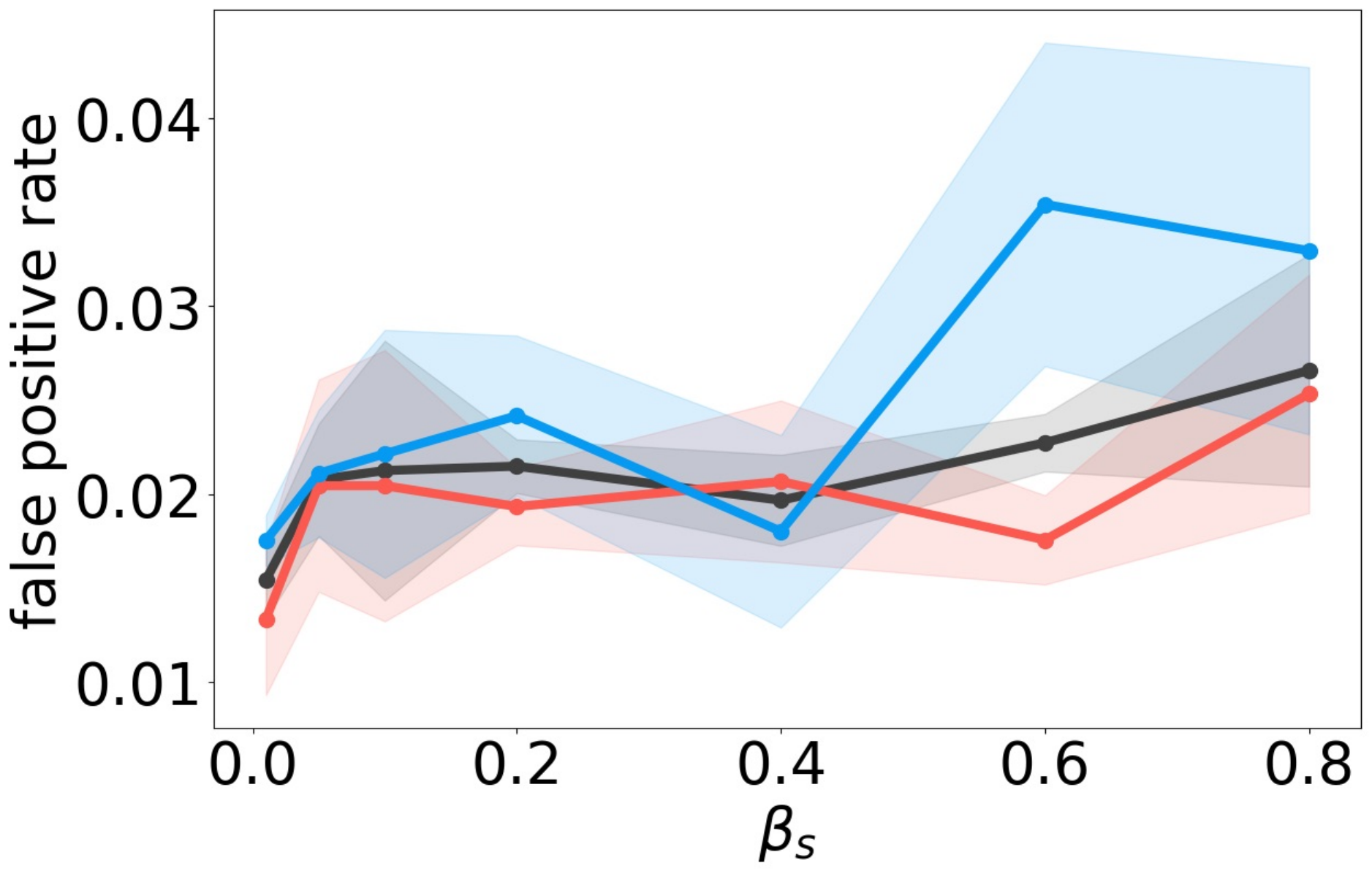}}
    &
   \hspace{-0.1in} {\includegraphics[width=0.195\textwidth]{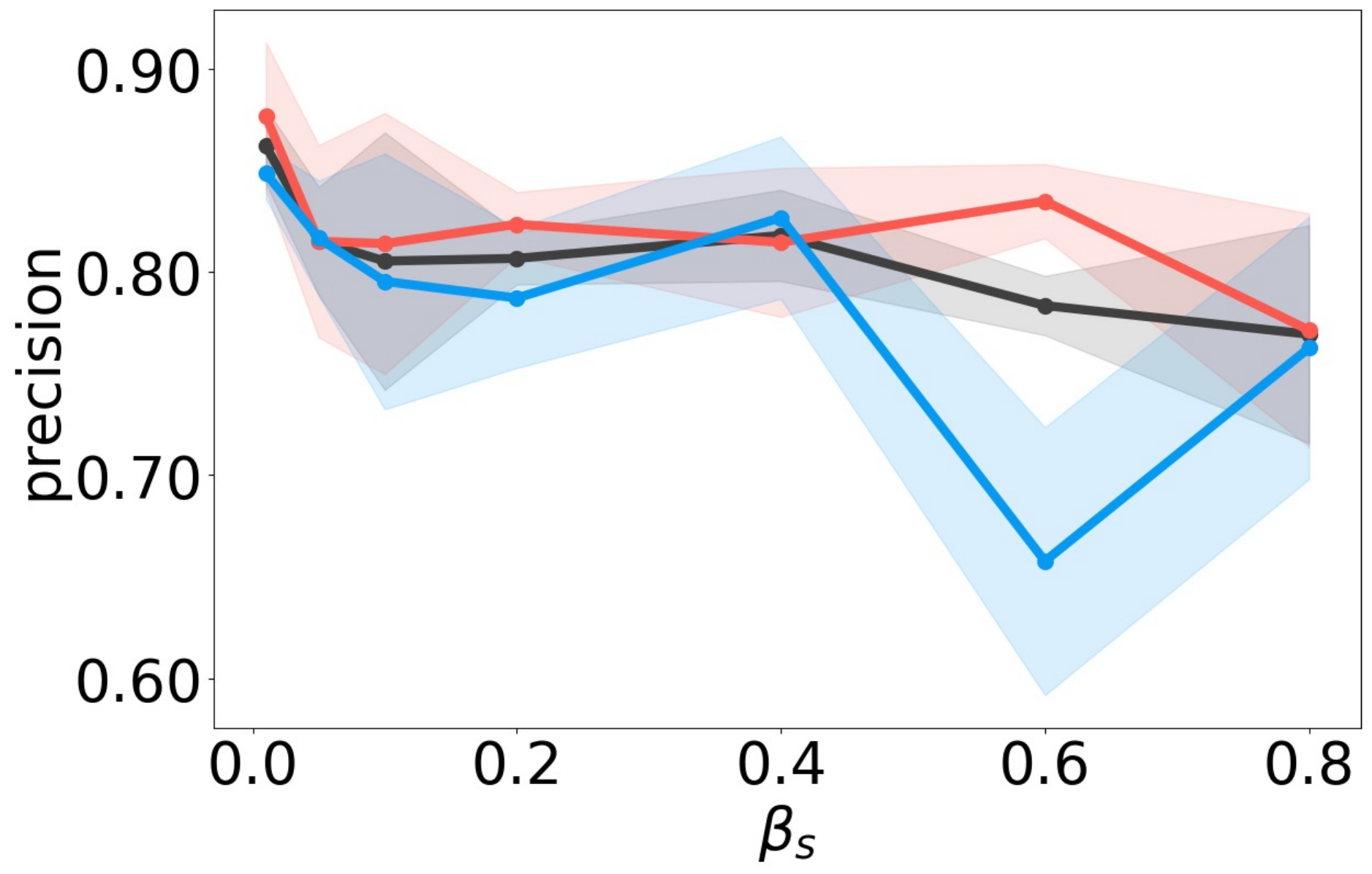}}
   &
     \hspace{-0.1in}{\includegraphics[width=0.195\textwidth]{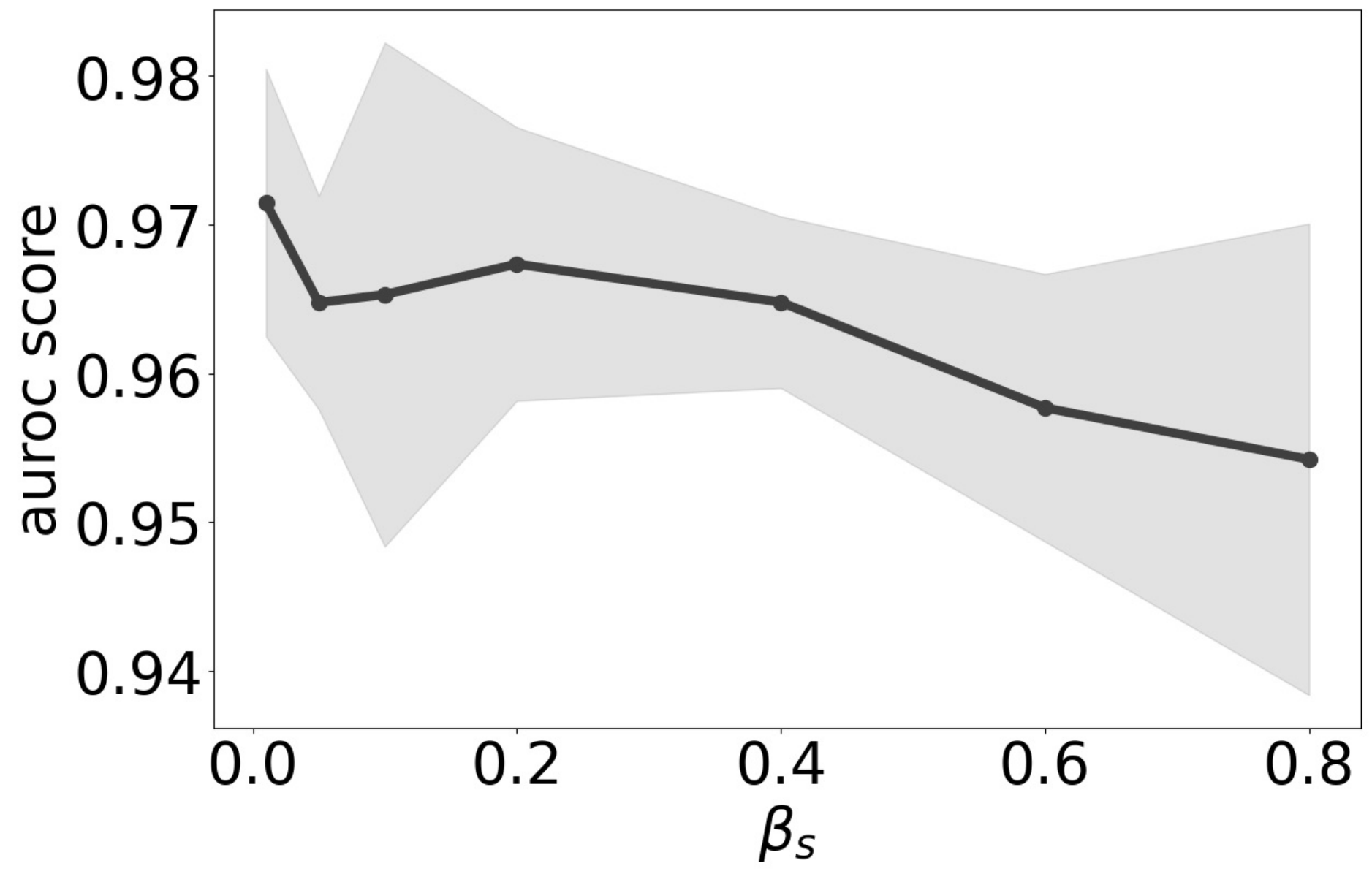}}
   \\
     \hspace{-0.1in}{\includegraphics[width=0.195\textwidth]{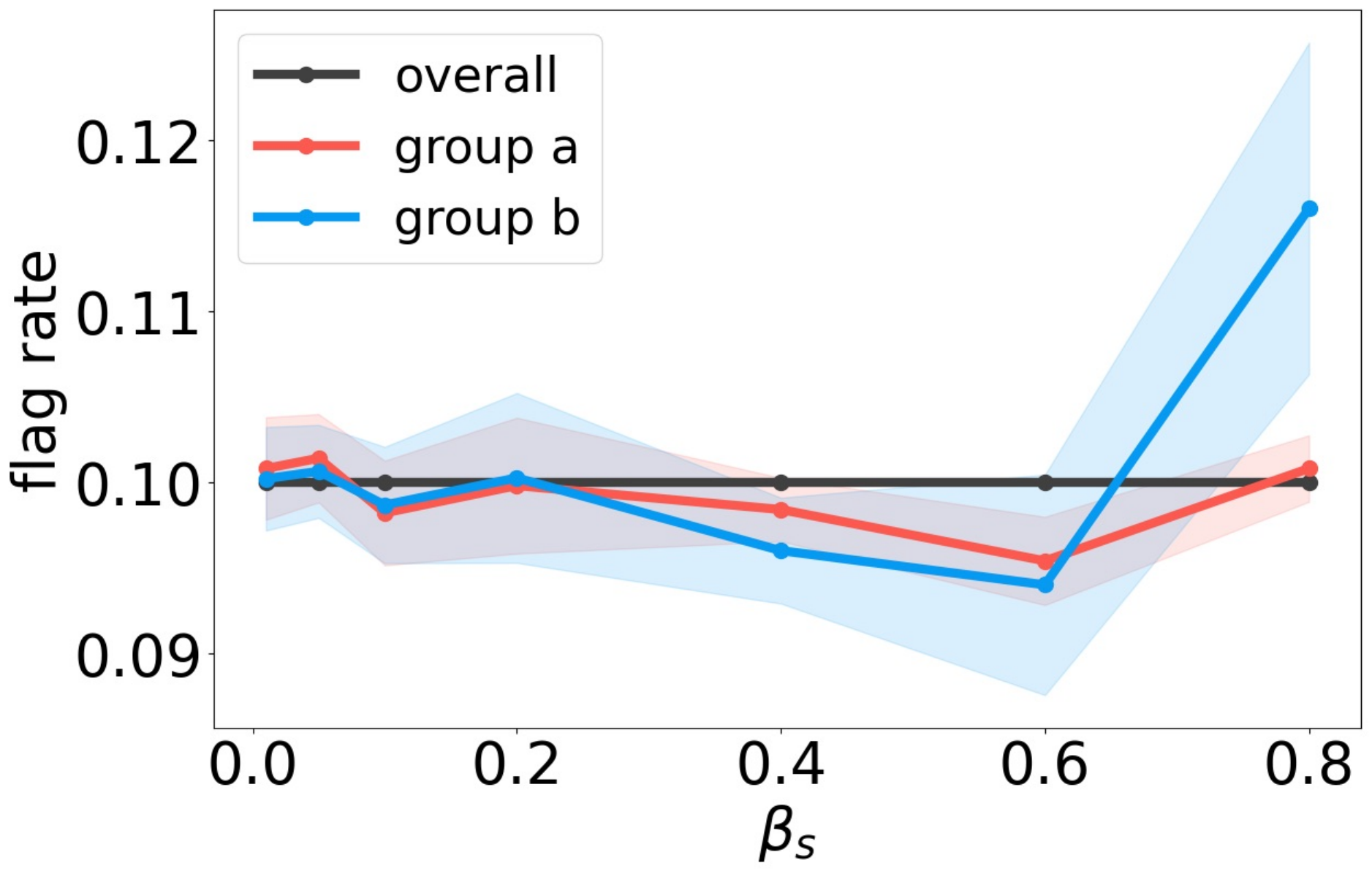}} 
    &
    \hspace{-0.1in}{\includegraphics[width=0.195\textwidth]{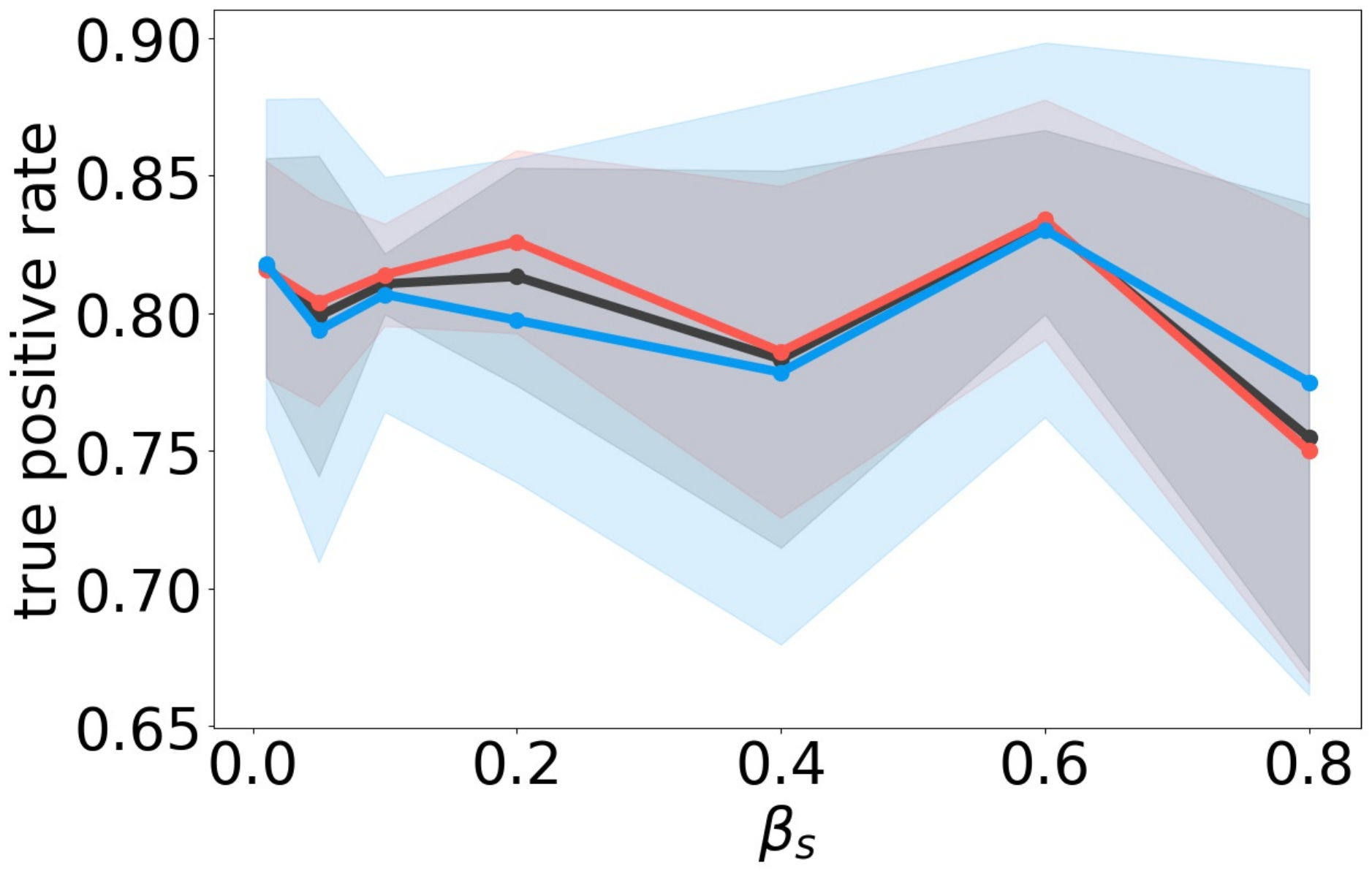}} &
     \hspace{-0.1in}{\includegraphics[width=0.195\textwidth]{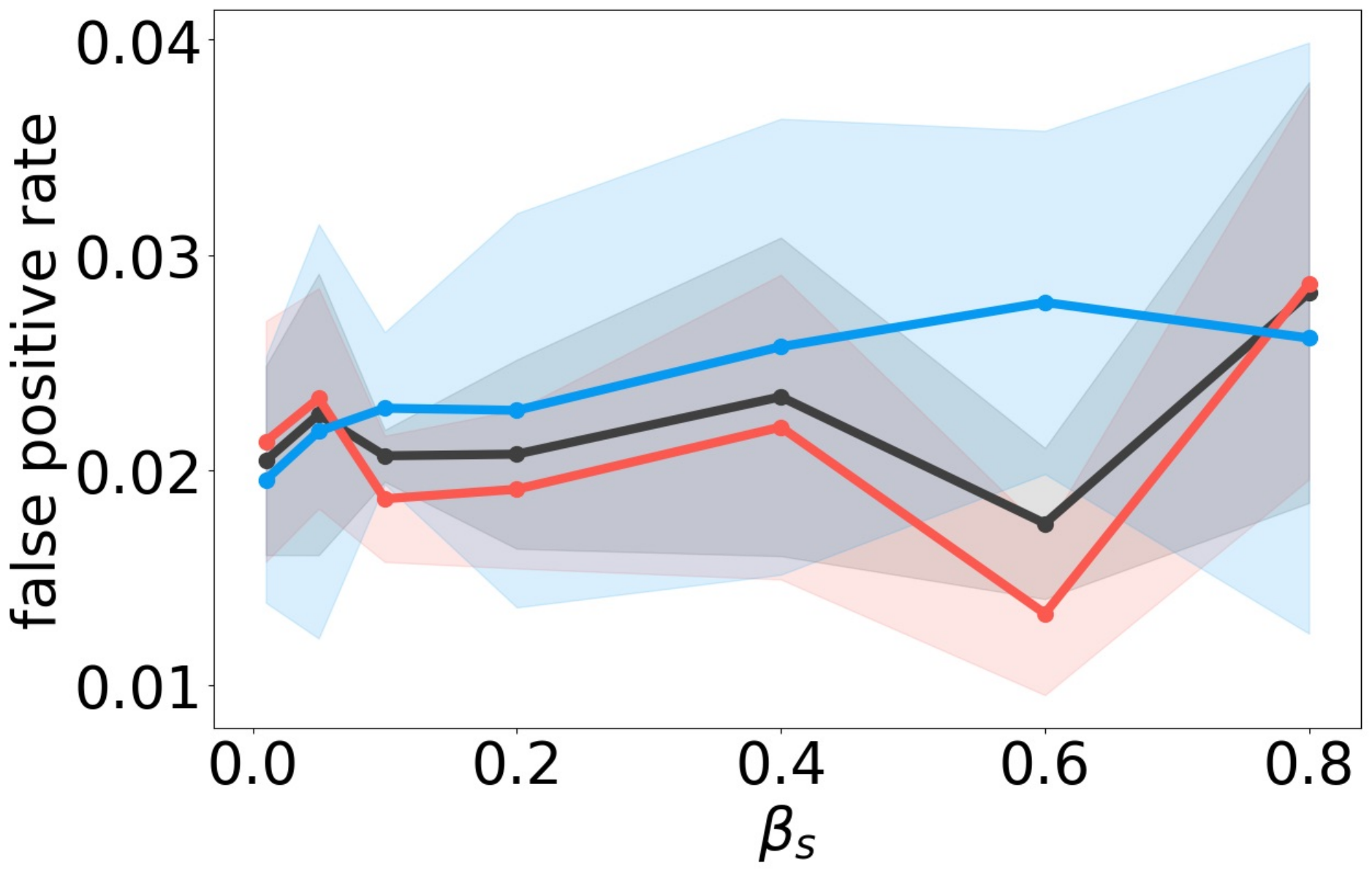}}
    &
   \hspace{-0.1in} {\includegraphics[width=0.195\textwidth]{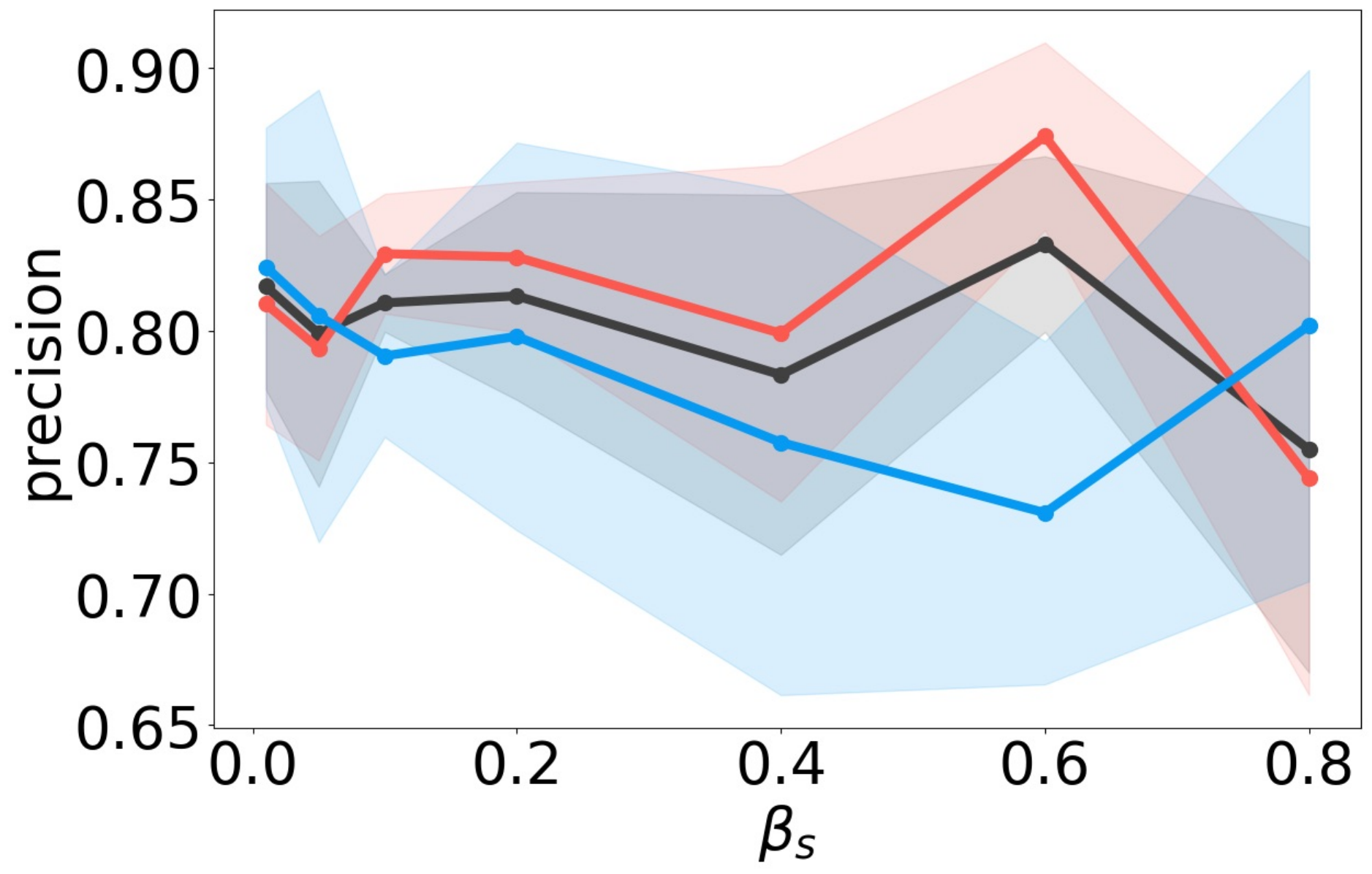}}
    &
   \hspace{-0.1in} {\includegraphics[width=0.195\textwidth]{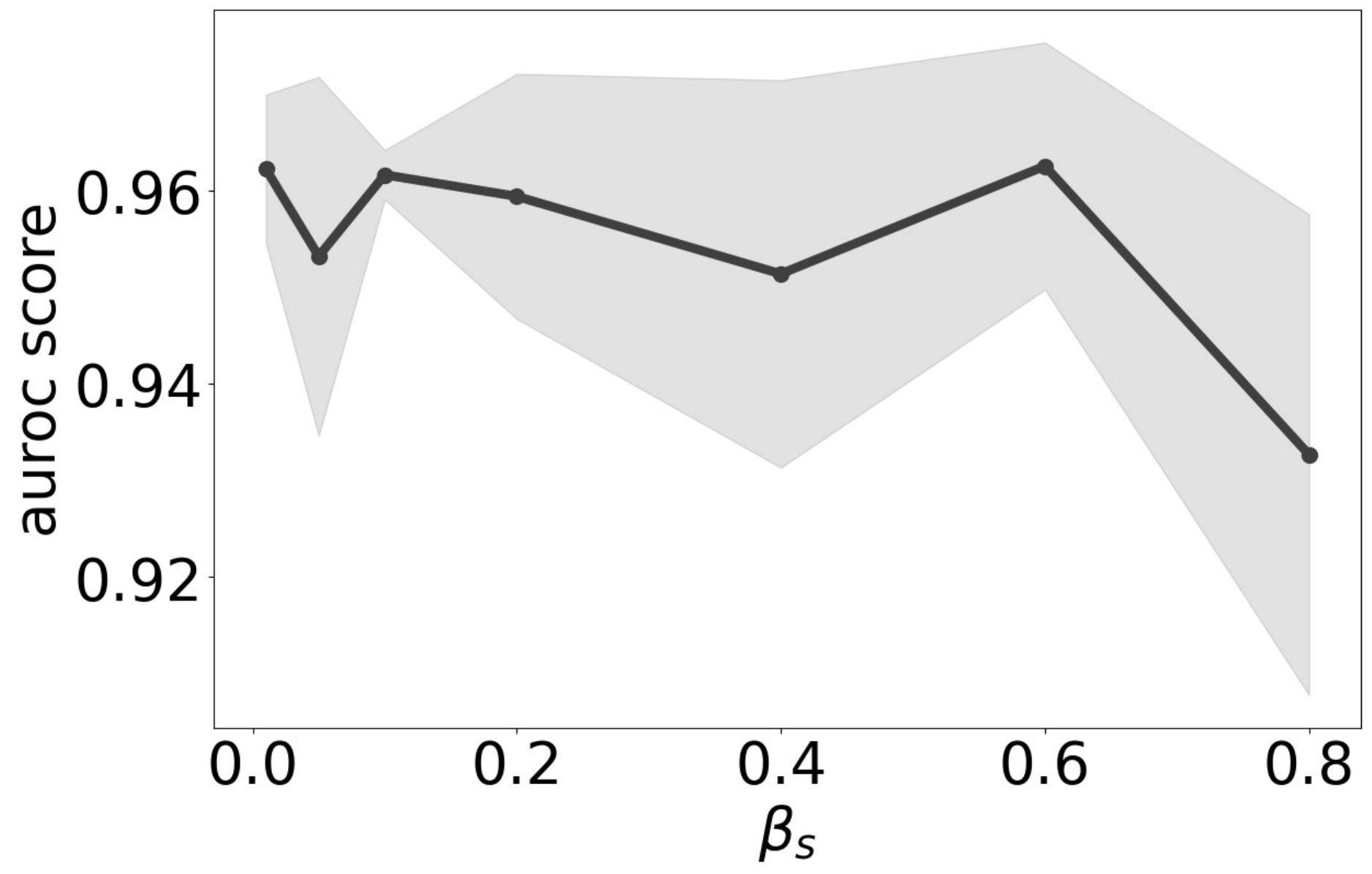}}
    \end{tabular}
    \caption{(best in color) Group-wise fairness metrics and AUROC for (top to bottom) LOF, iForest, DeepAE and FairOD under \underline{sample size bias} on \textit{scattered} outliers.}
    \label{fig:samplesize_all_scatter}
\end{figure*}

See Fig. \ref{fig:samplesize_all_scatter}.

\subsubsection{Target under-representation bias}
\label{assec:urep_scatter}

\begin{figure*}[!ht]
    \centering
    \begin{tabular}{lllll}
    \hspace{-0.1in}{\includegraphics[width=0.195\textwidth]{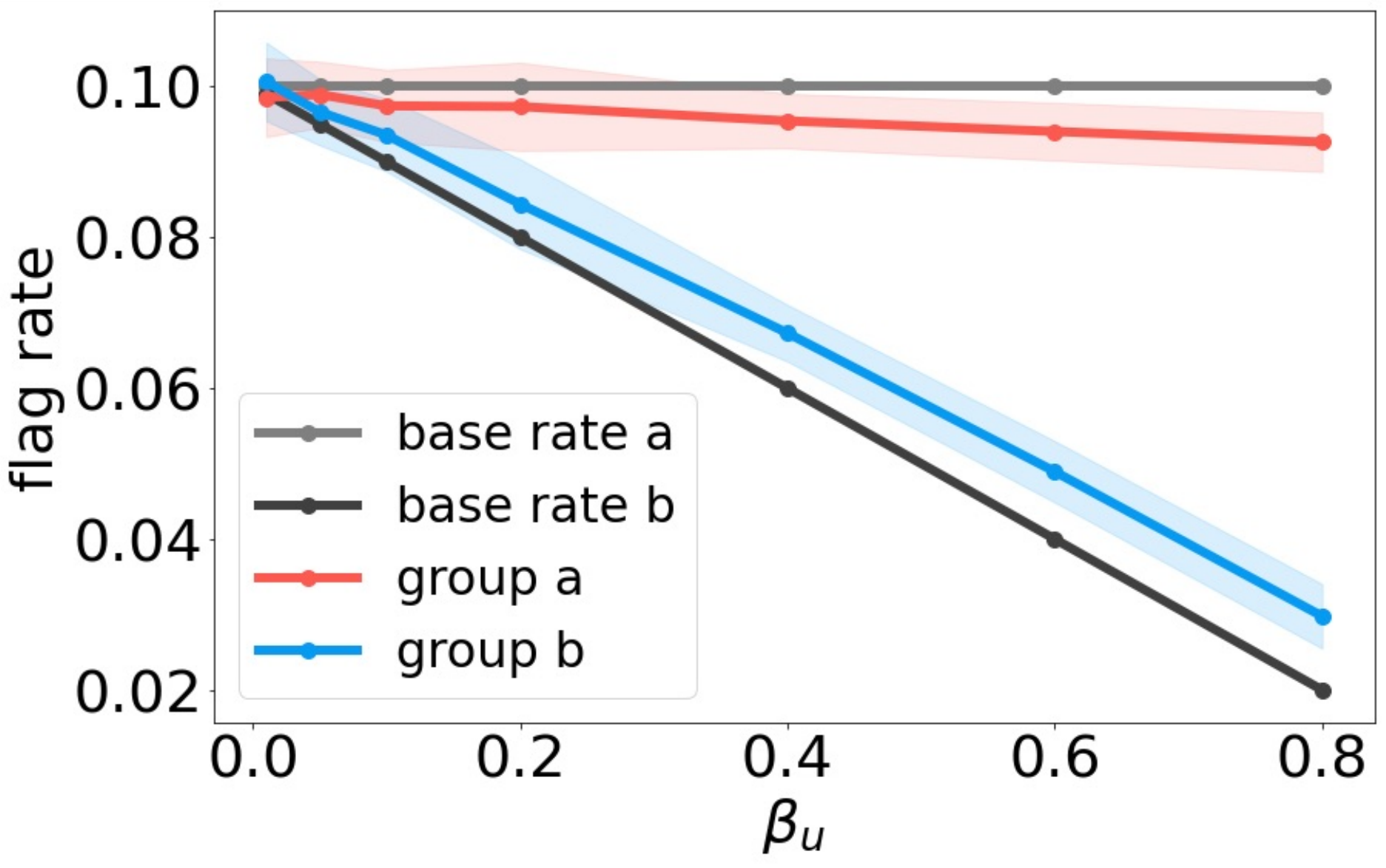}} 
    &
    \hspace{-0.1in}{\includegraphics[width=0.195\textwidth]{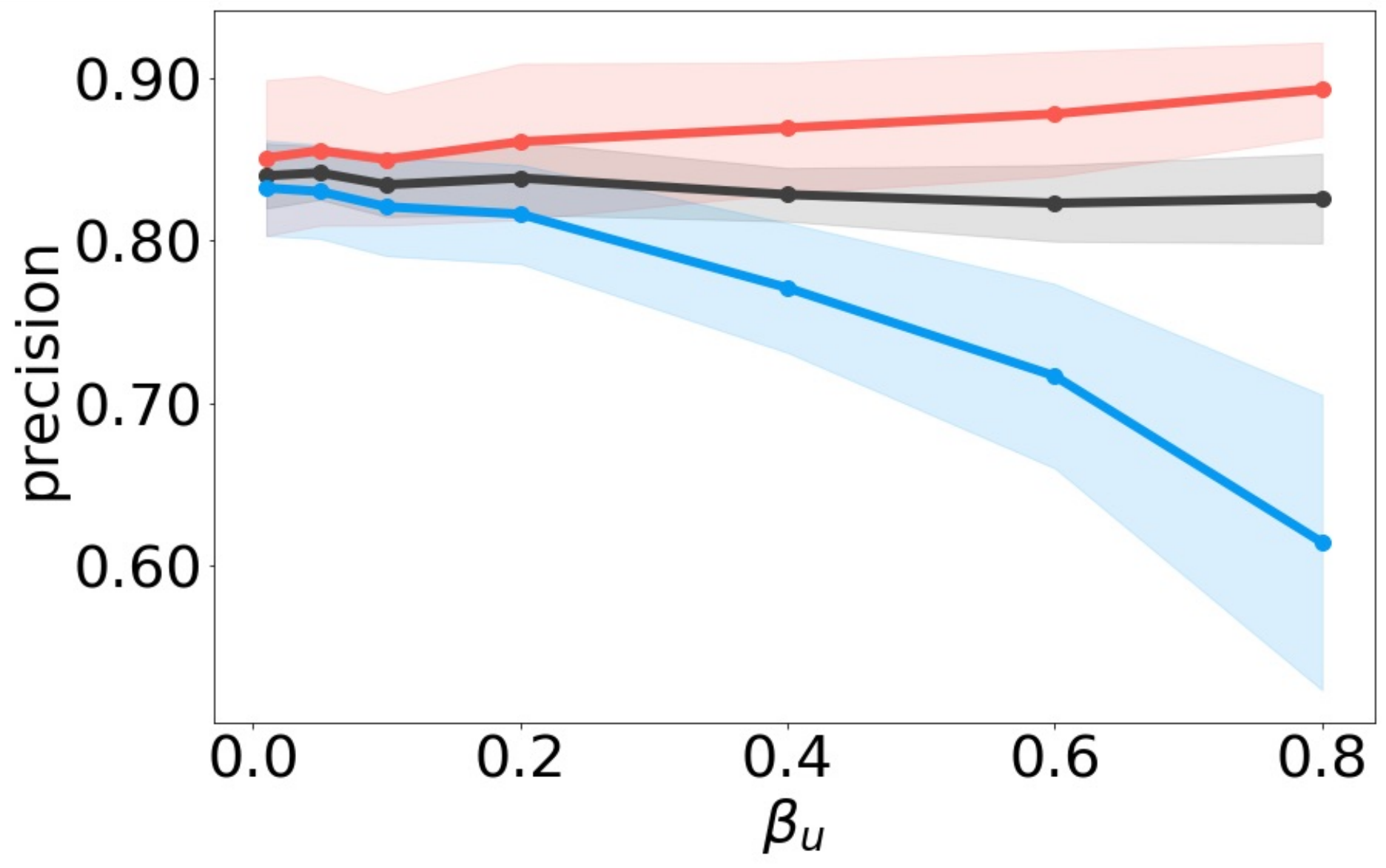}} &
     \hspace{-0.1in}{\includegraphics[width=0.195\textwidth]{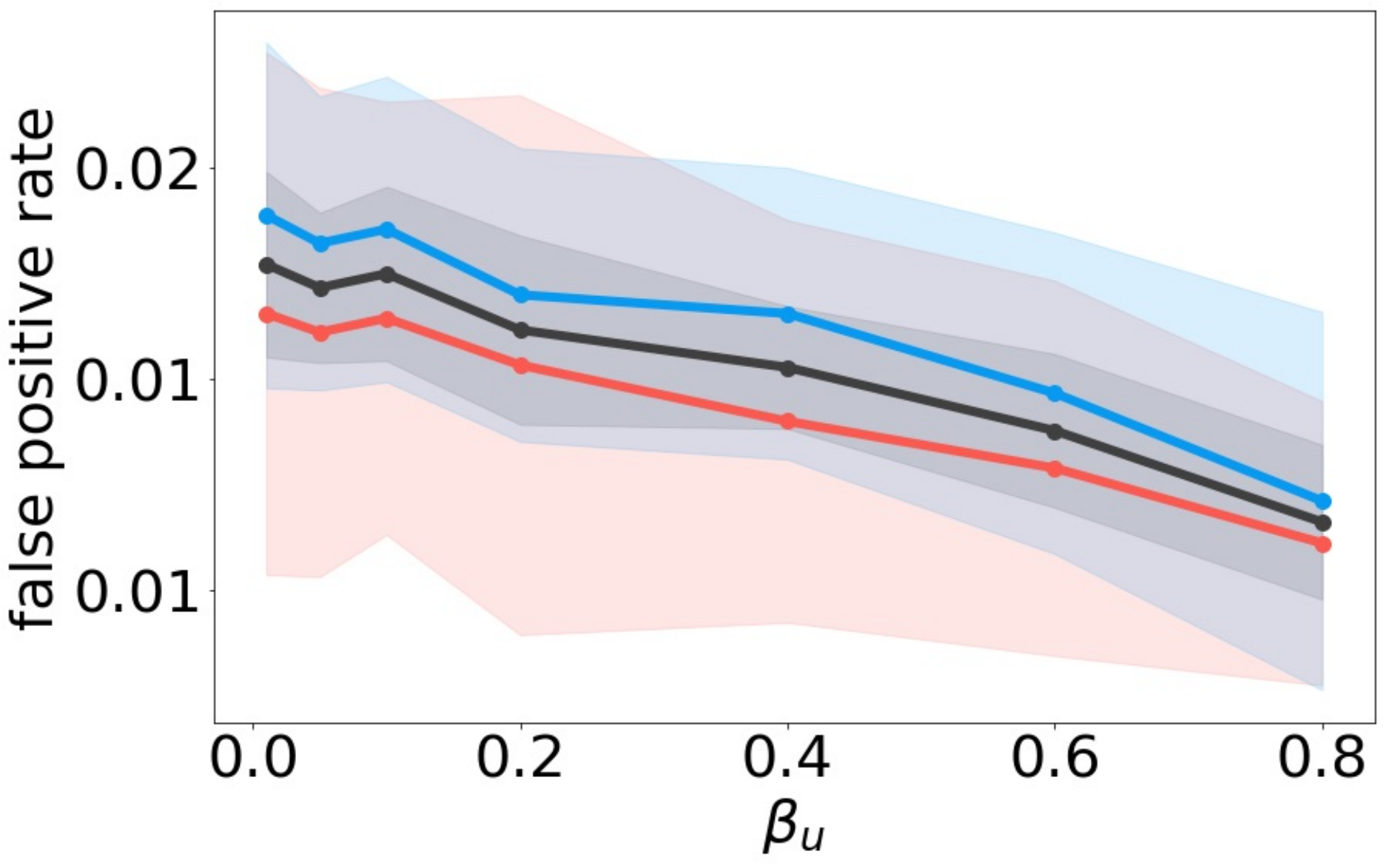}}
    &
   \hspace{-0.1in} {\includegraphics[width=0.195\textwidth]{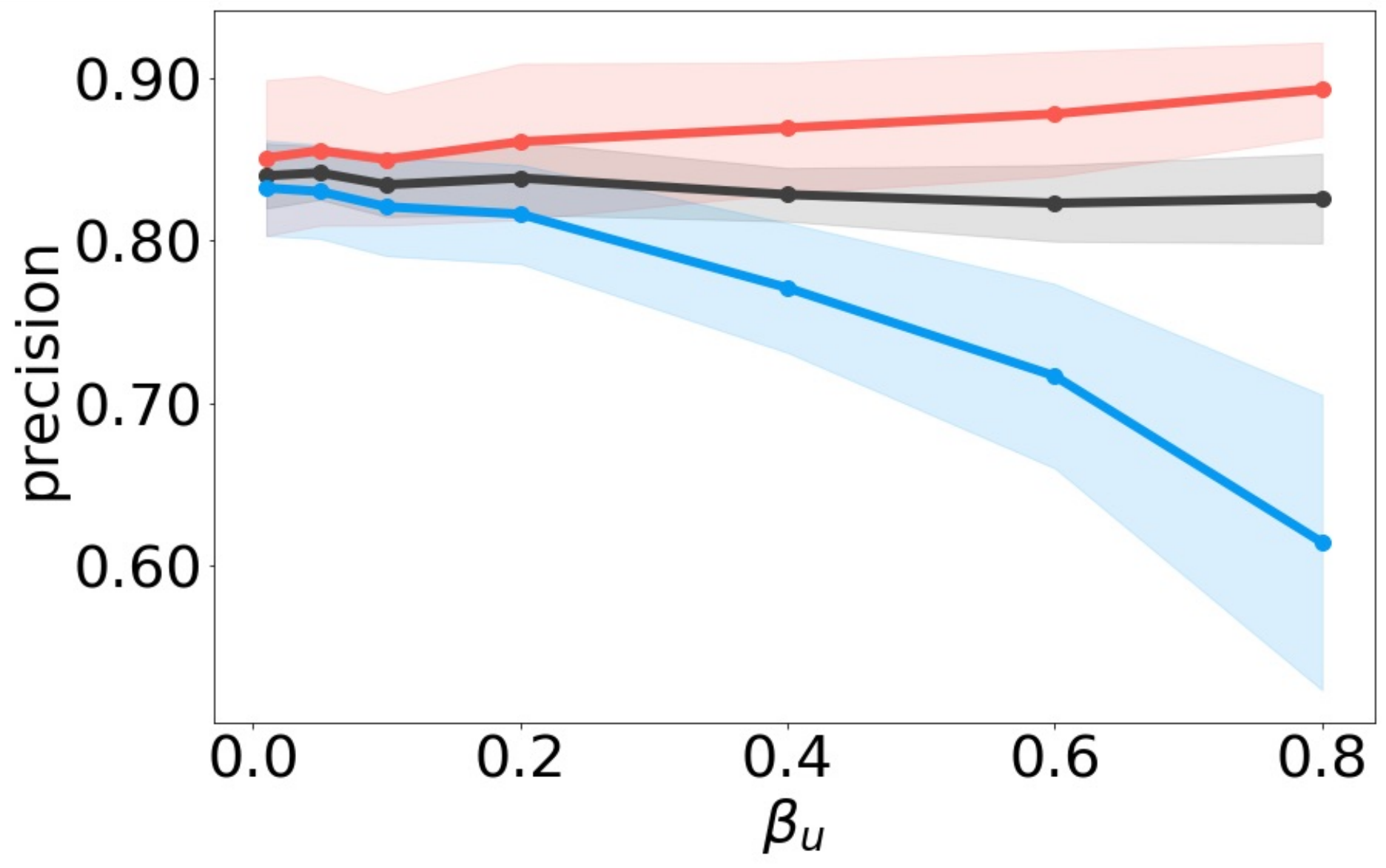}}
   &
   \hspace{-0.1in}
   {\includegraphics[width=0.195\textwidth]{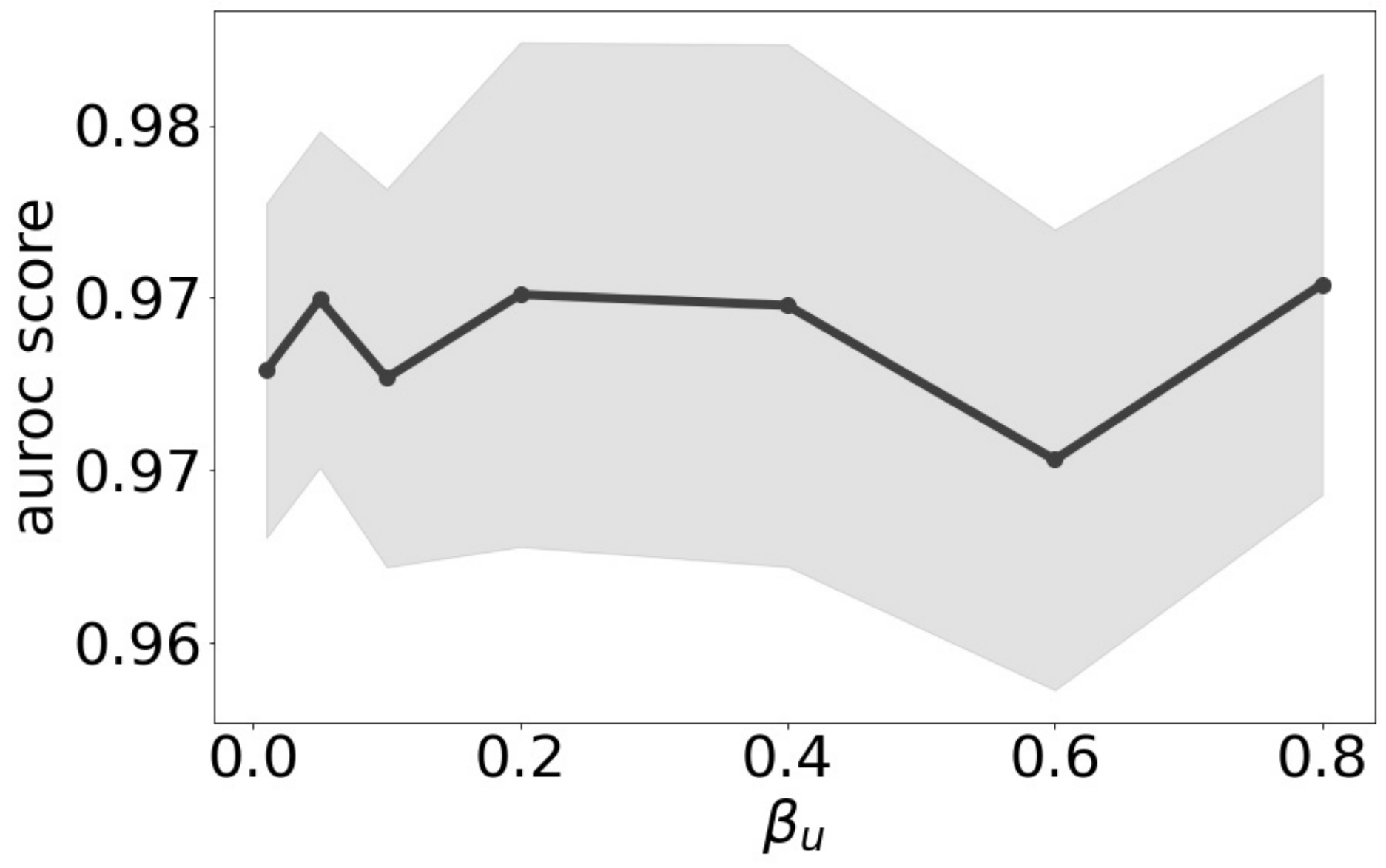}} 
\\
\hspace{-0.1in}{\includegraphics[width=0.195\textwidth]{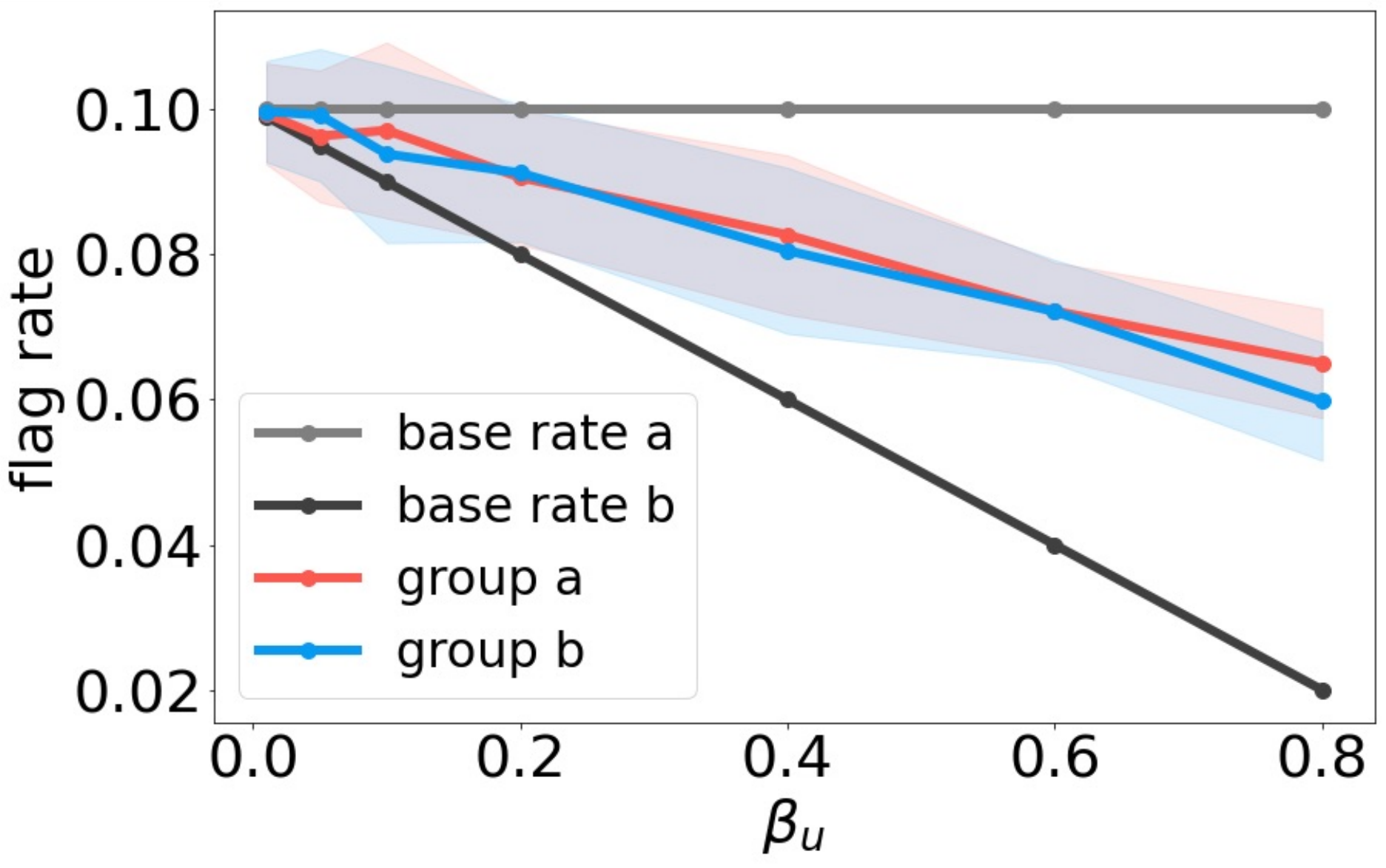}} 
    &
    \hspace{-0.1in}{\includegraphics[width=0.195\textwidth]{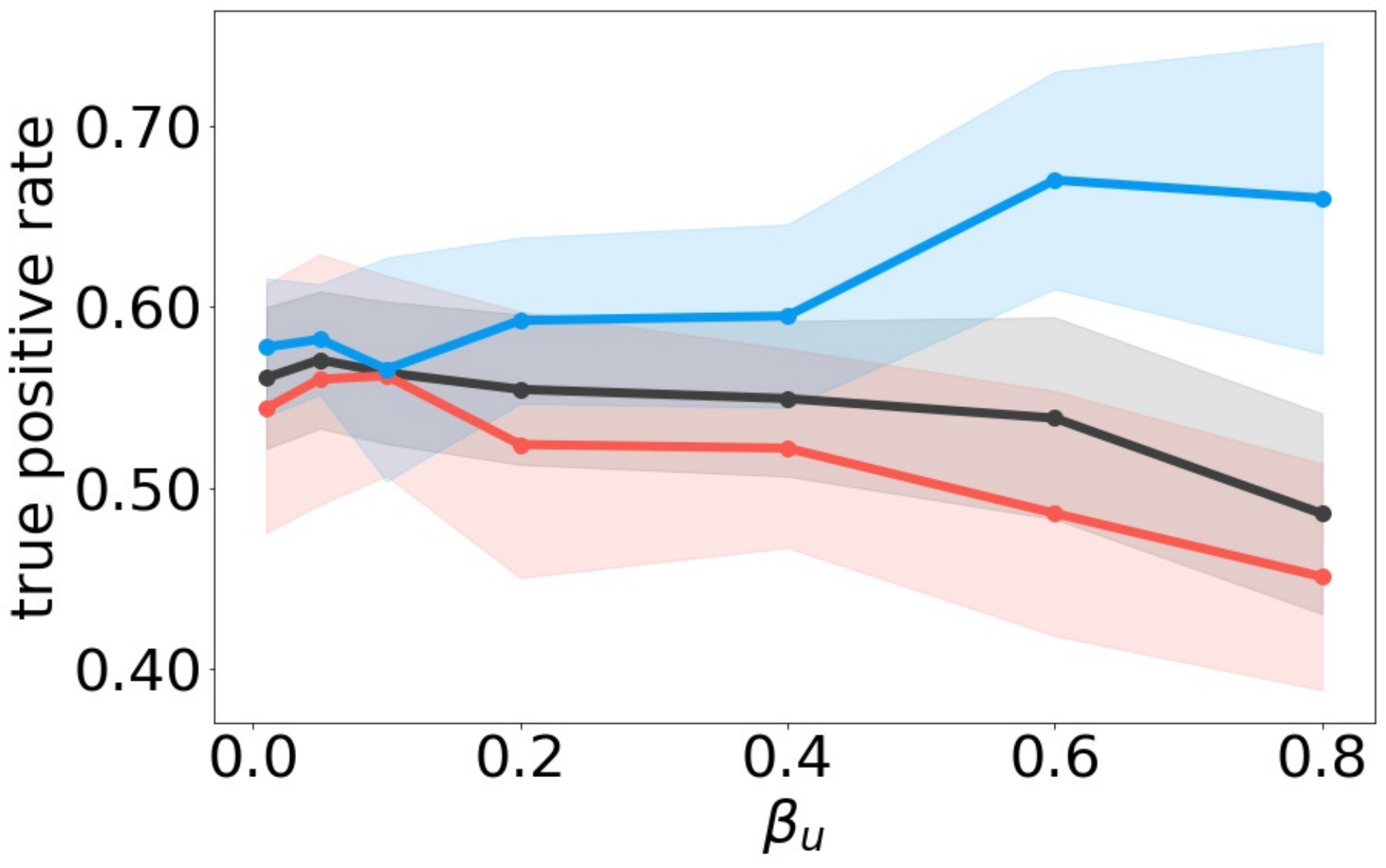}} &
     \hspace{-0.1in}{\includegraphics[width=0.195\textwidth]{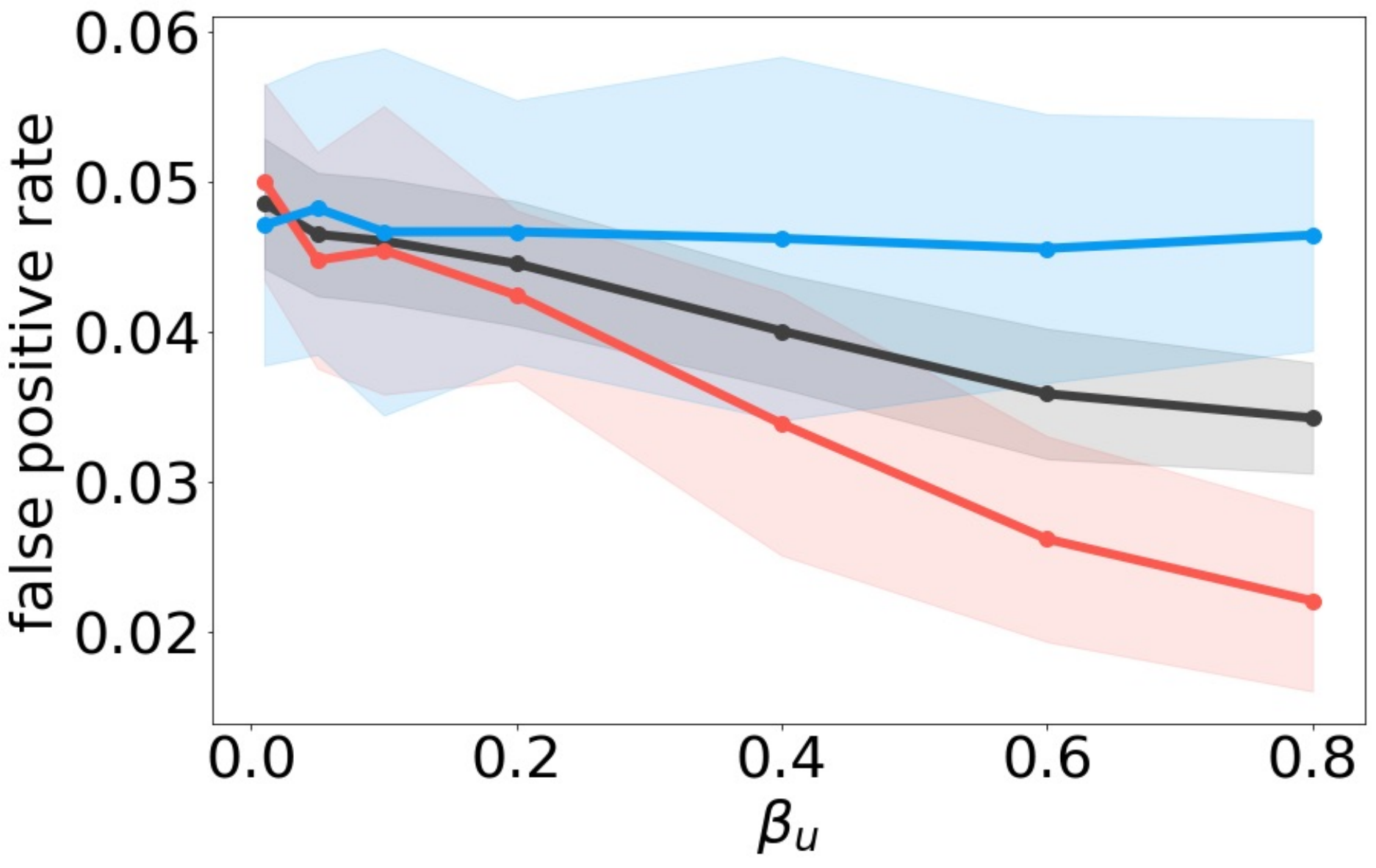}}
    &
   \hspace{-0.1in} {\includegraphics[width=0.195\textwidth]{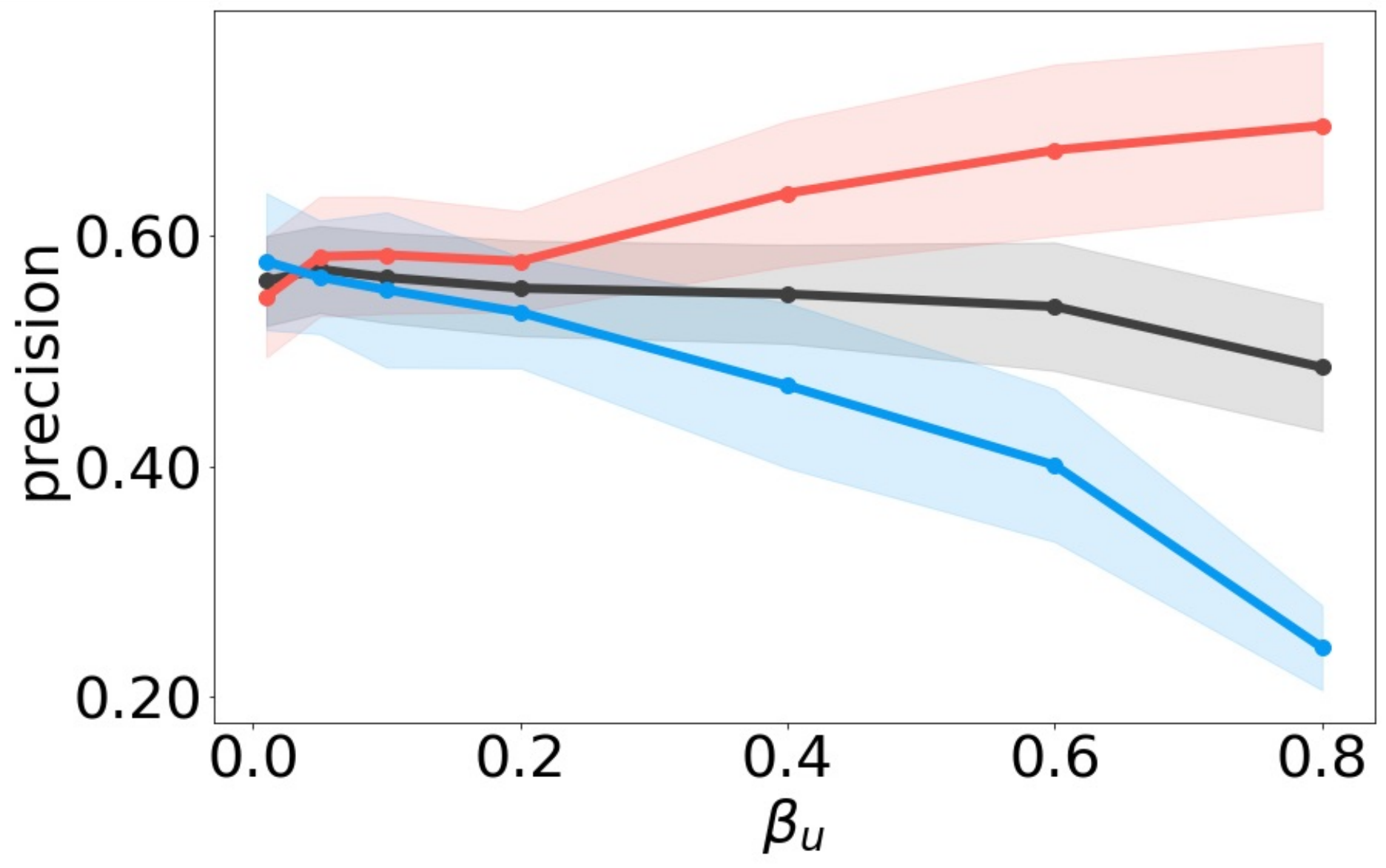}}
   &
    \hspace{-0.1in}{\includegraphics[width=0.195\textwidth]{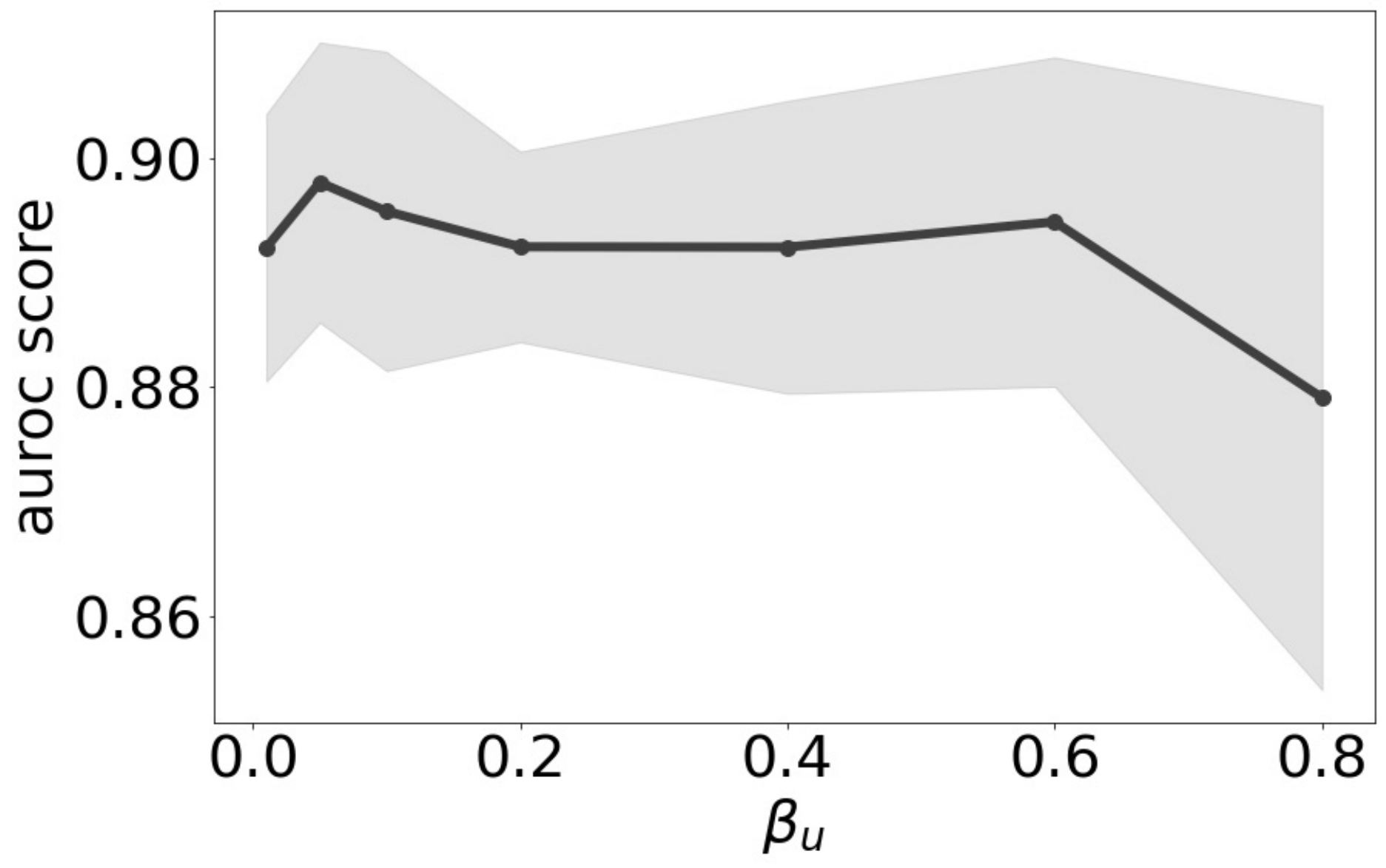}}
   \\
     \hspace{-0.1in}{\includegraphics[width=0.195\textwidth]{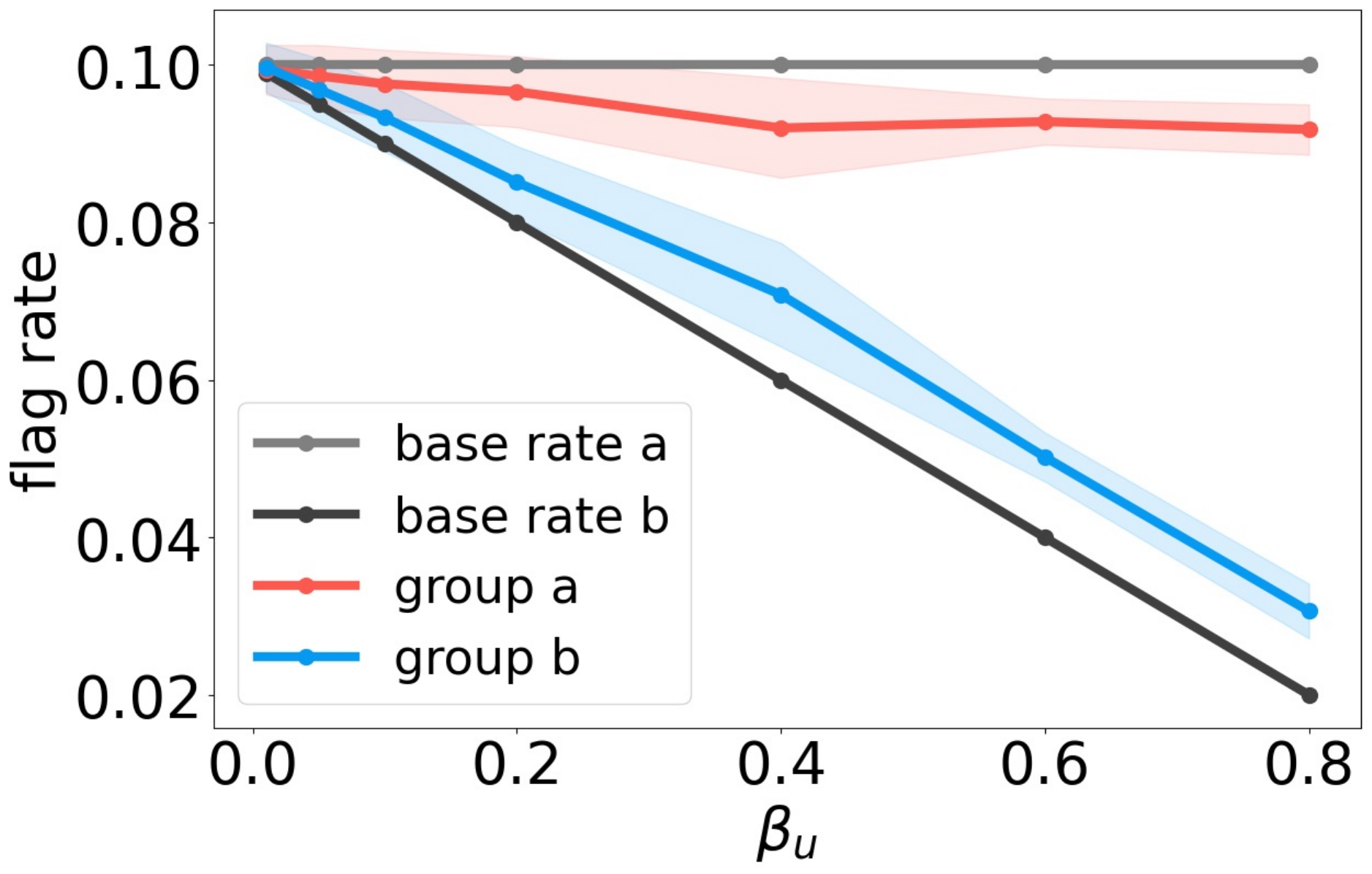}} 
    &
    \hspace{-0.1in}{\includegraphics[width=0.195\textwidth]{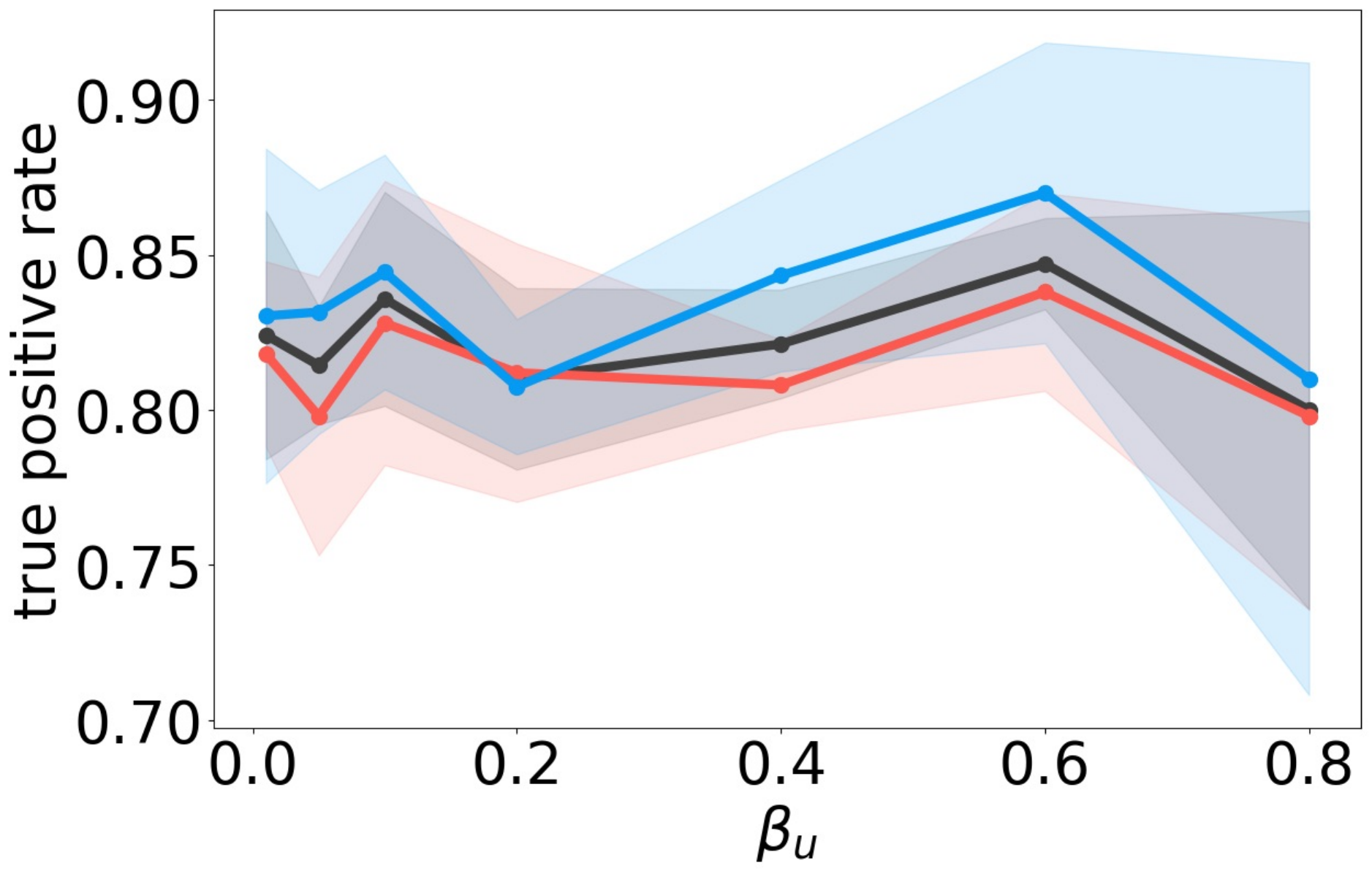}} &
     \hspace{-0.1in}{\includegraphics[width=0.195\textwidth]{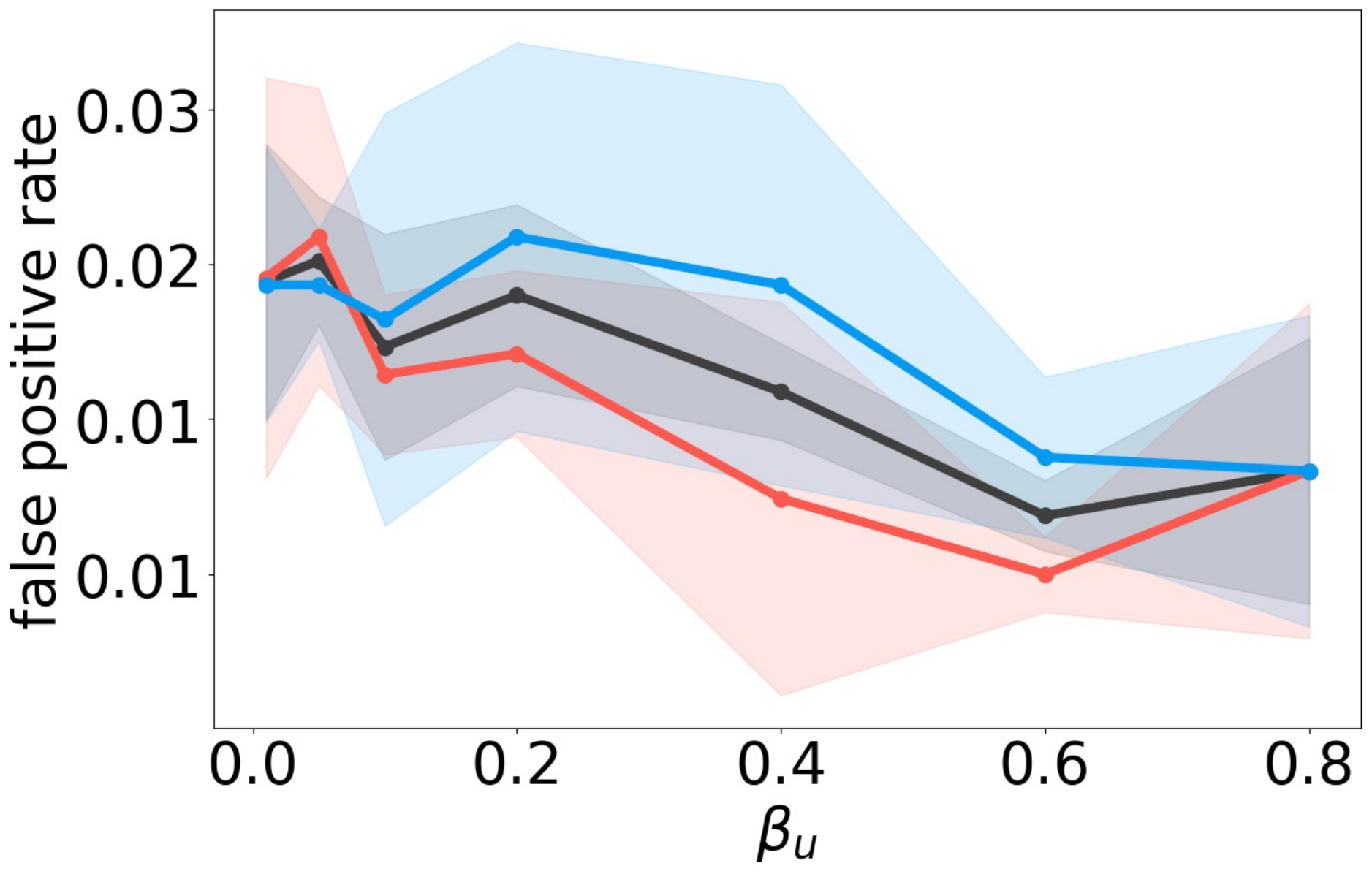}}
    &
   \hspace{-0.1in} {\includegraphics[width=0.195\textwidth]{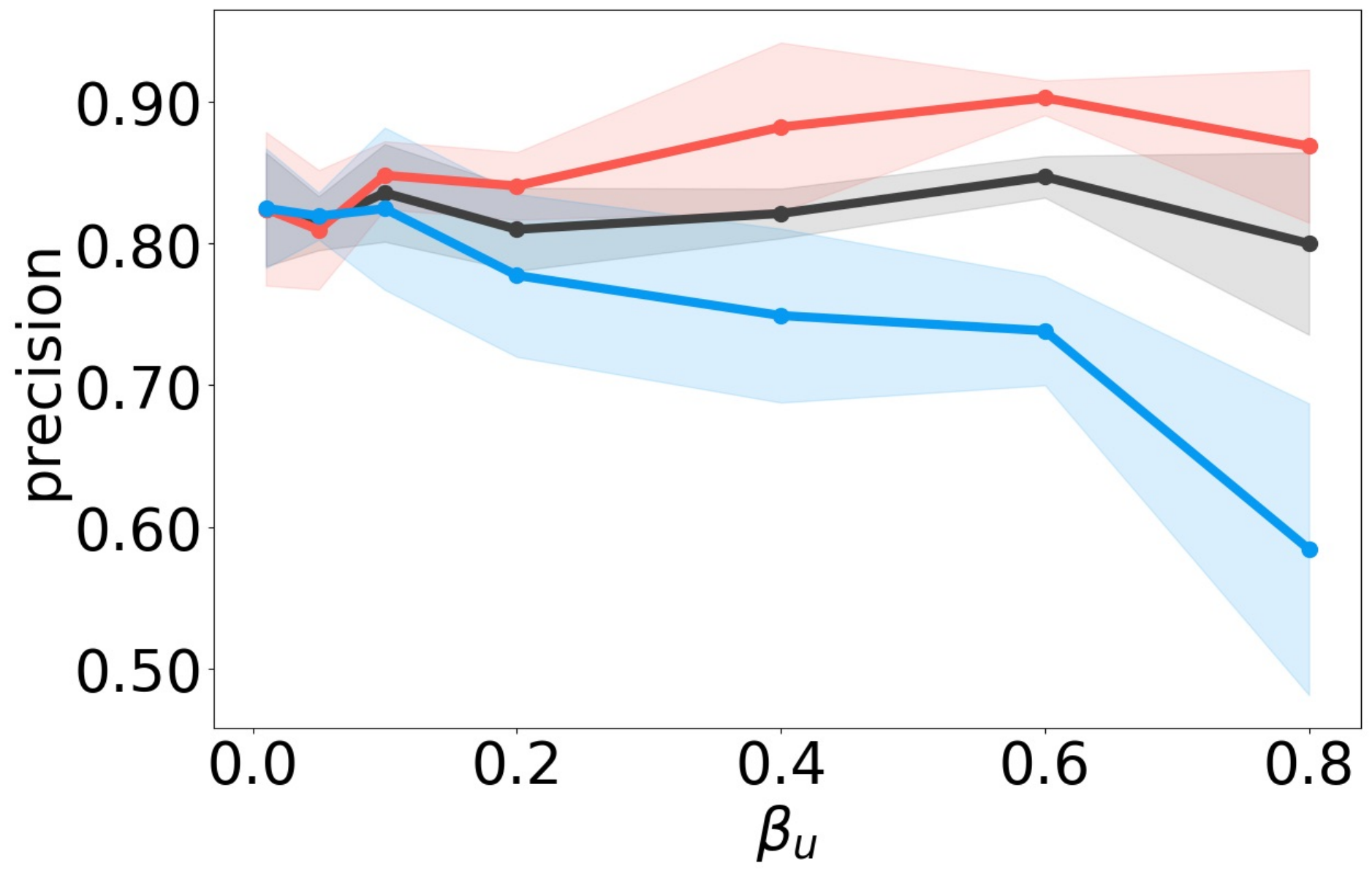}}
   &
     \hspace{-0.1in}{\includegraphics[width=0.195\textwidth]{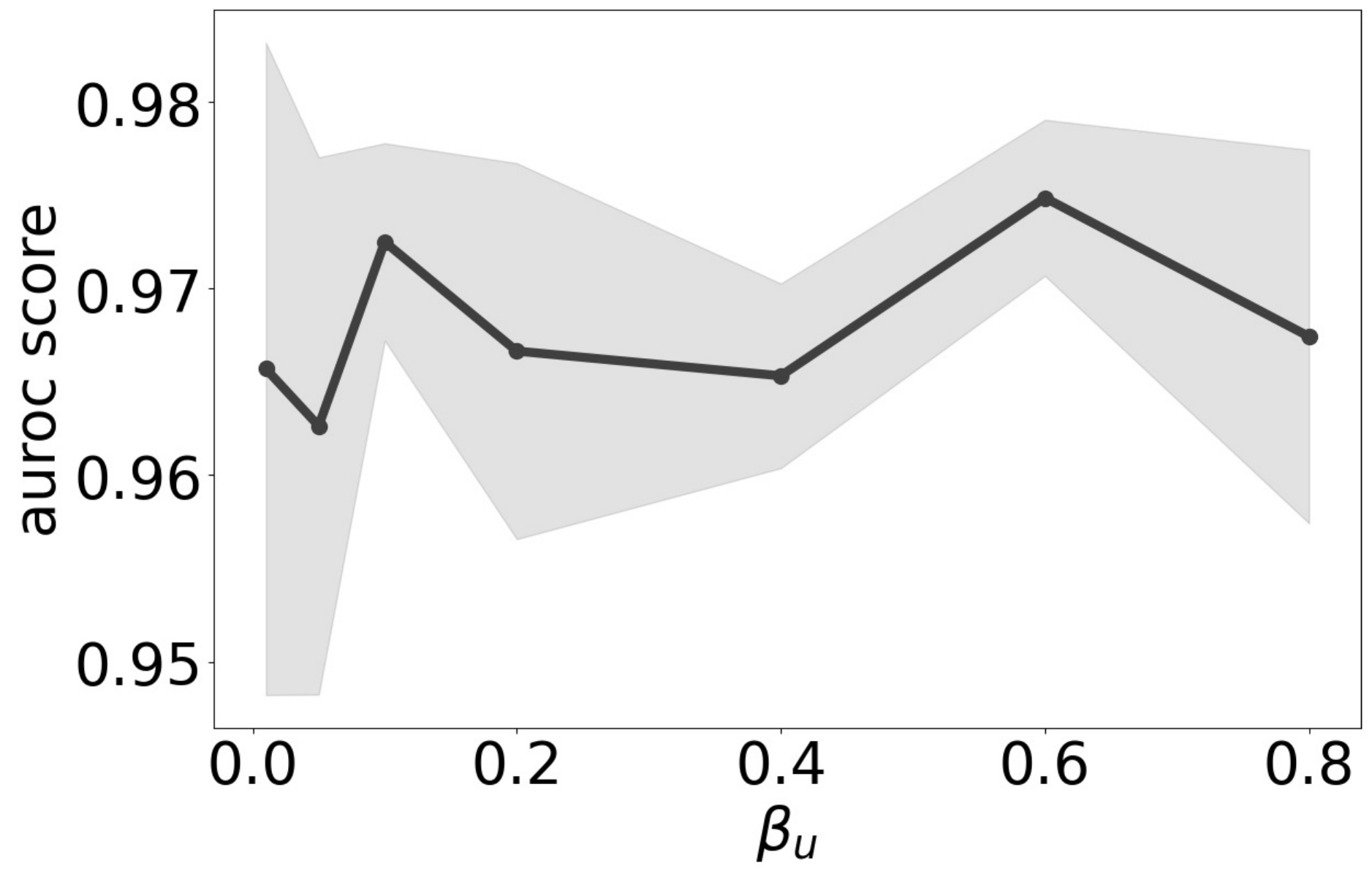}}
   \\
   \hspace{-0.1in}{\includegraphics[width=0.195\textwidth]{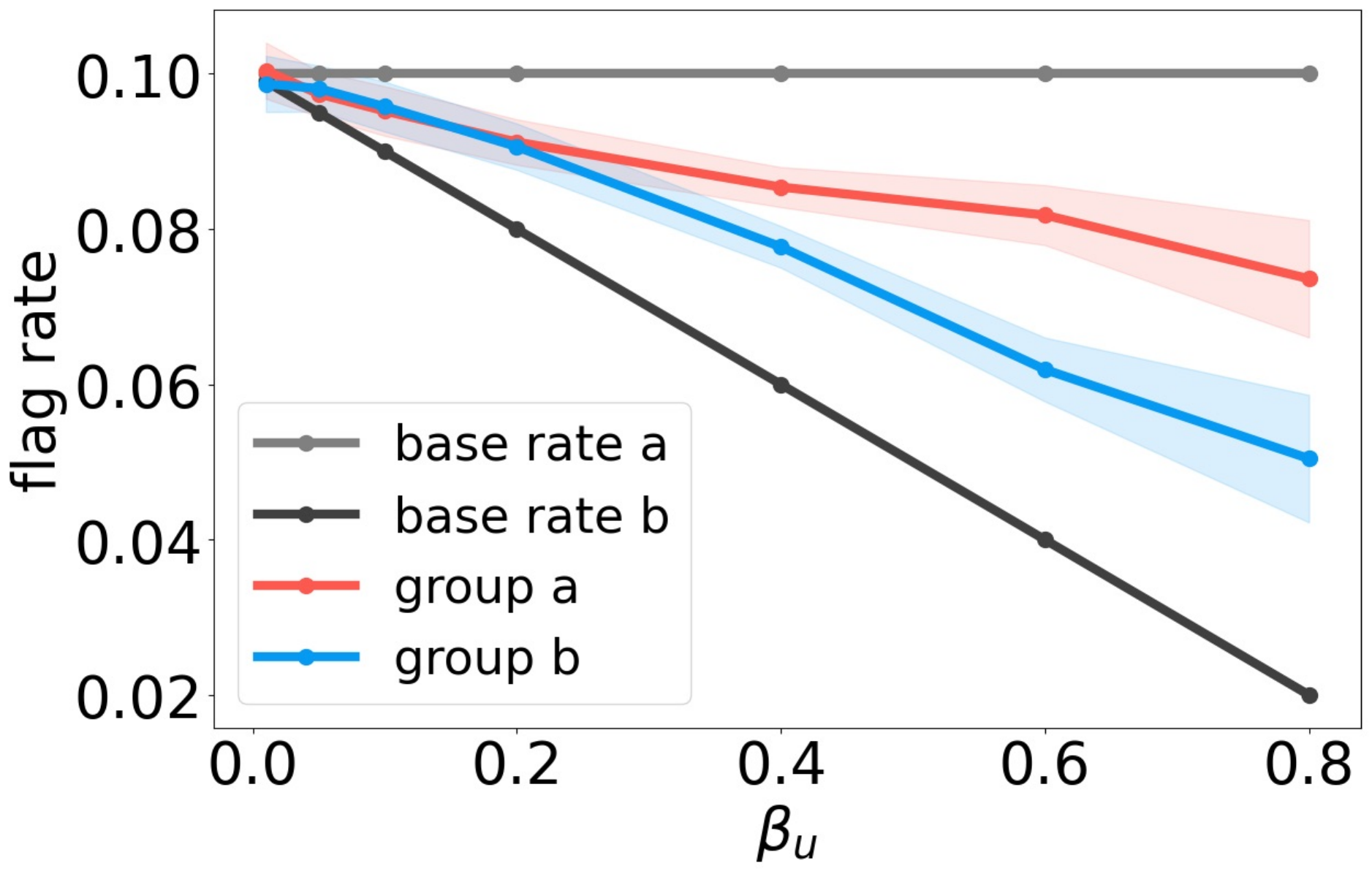}} 
    &
    \hspace{-0.1in}{\includegraphics[width=0.195\textwidth]{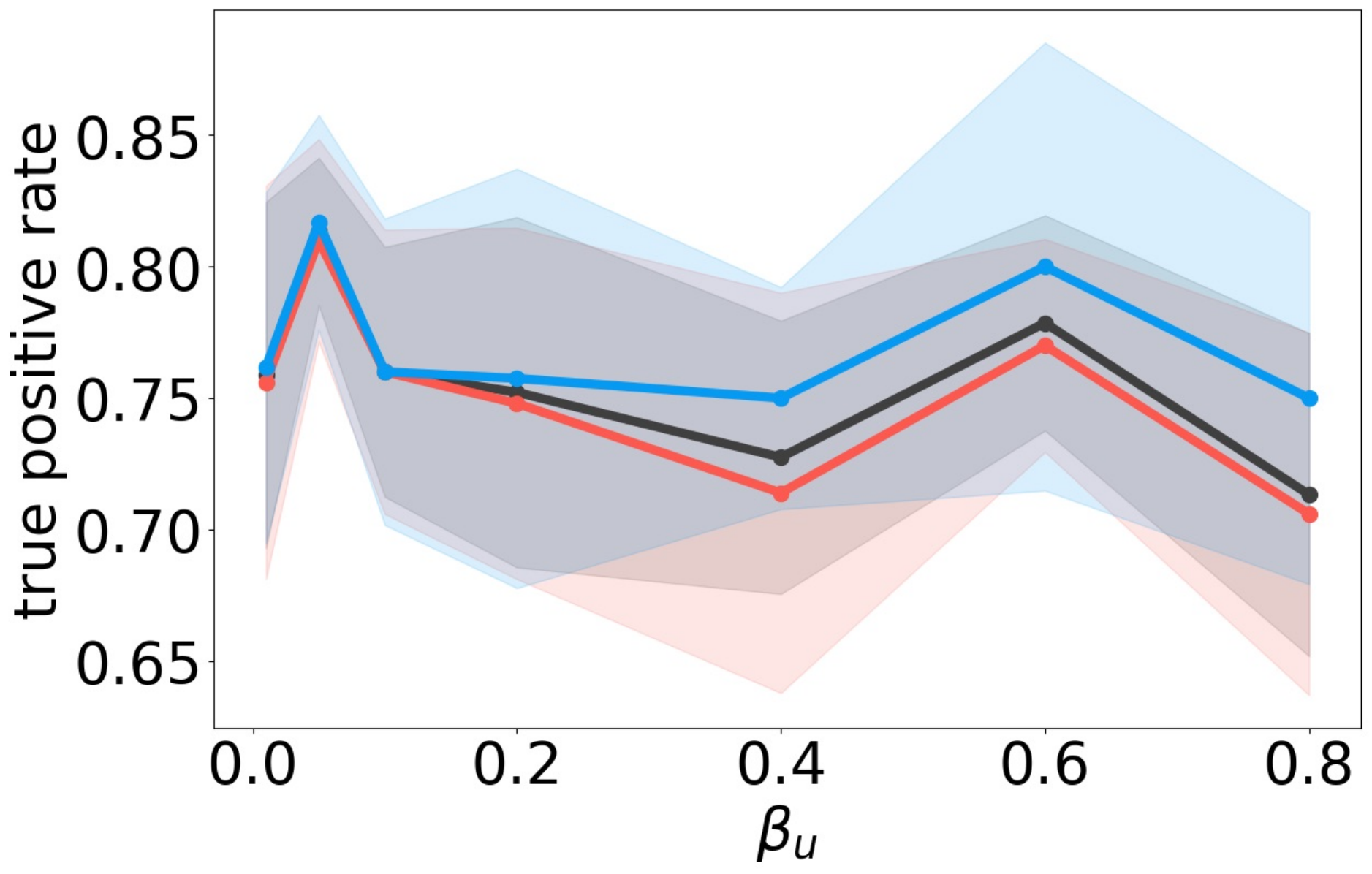}} &
     \hspace{-0.1in}{\includegraphics[width=0.195\textwidth]{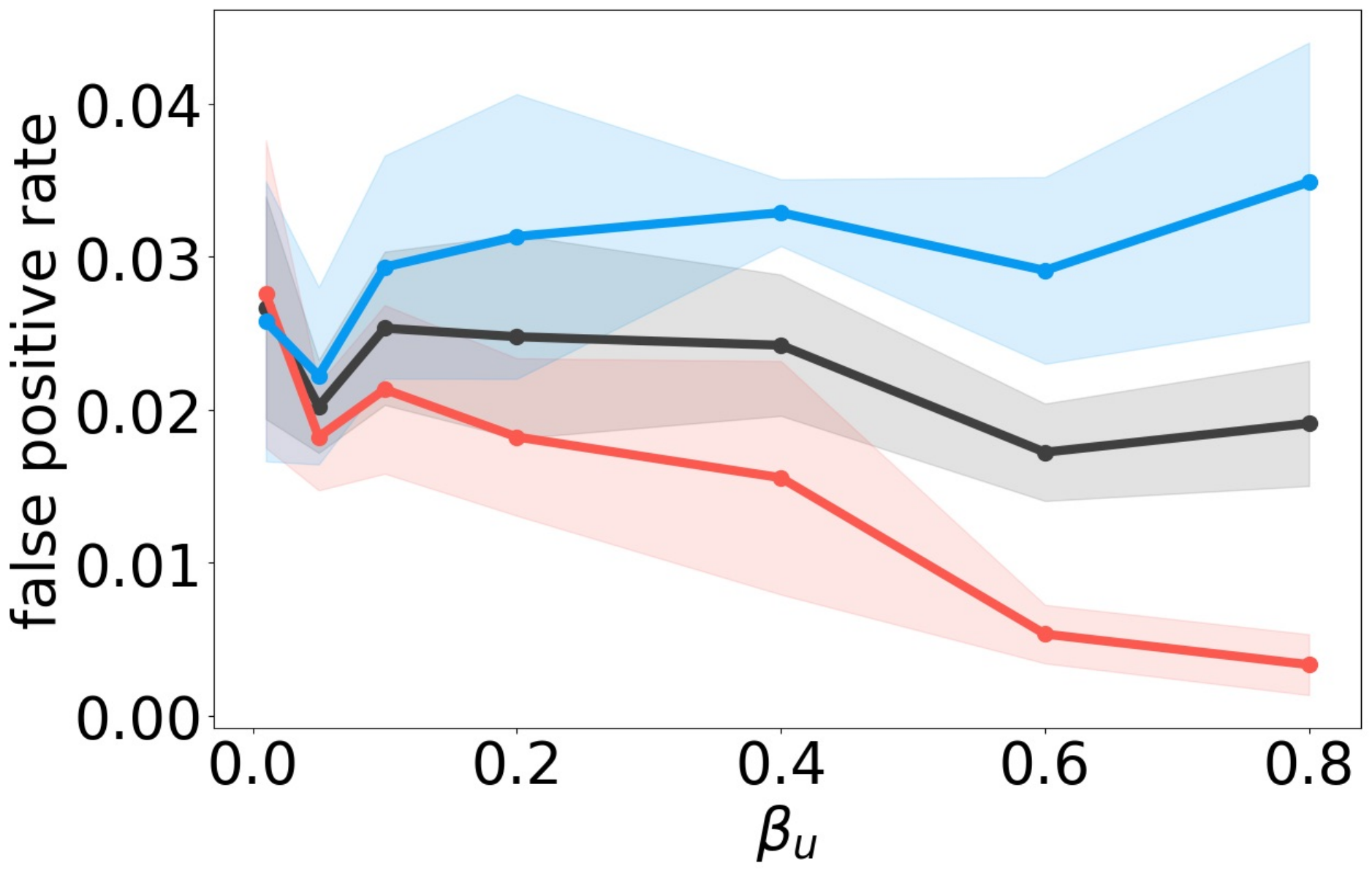}}
    &
   \hspace{-0.1in} {\includegraphics[width=0.195\textwidth]{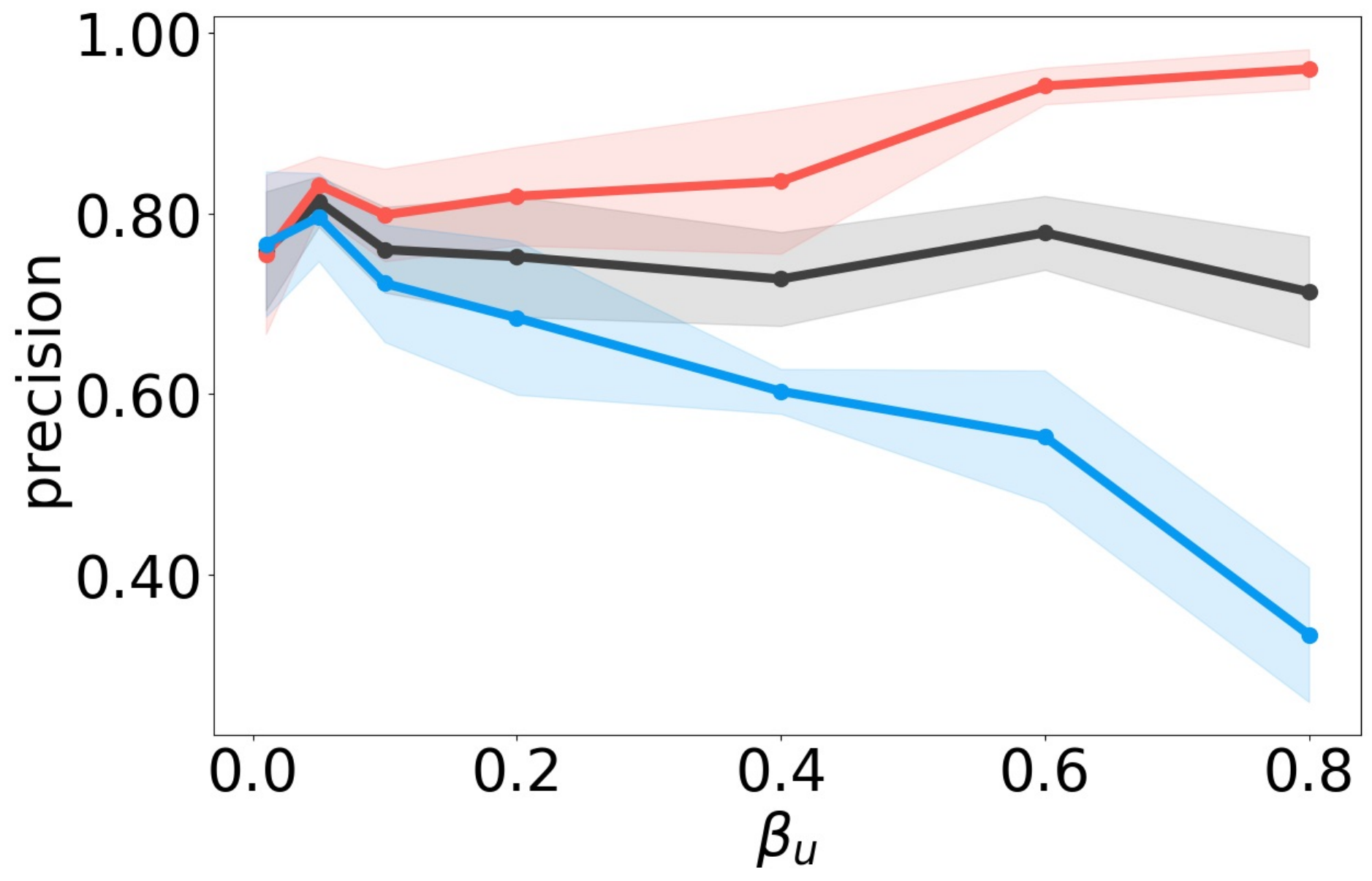}}
   &
   \hspace{-0.1in} {\includegraphics[width=0.195\textwidth]{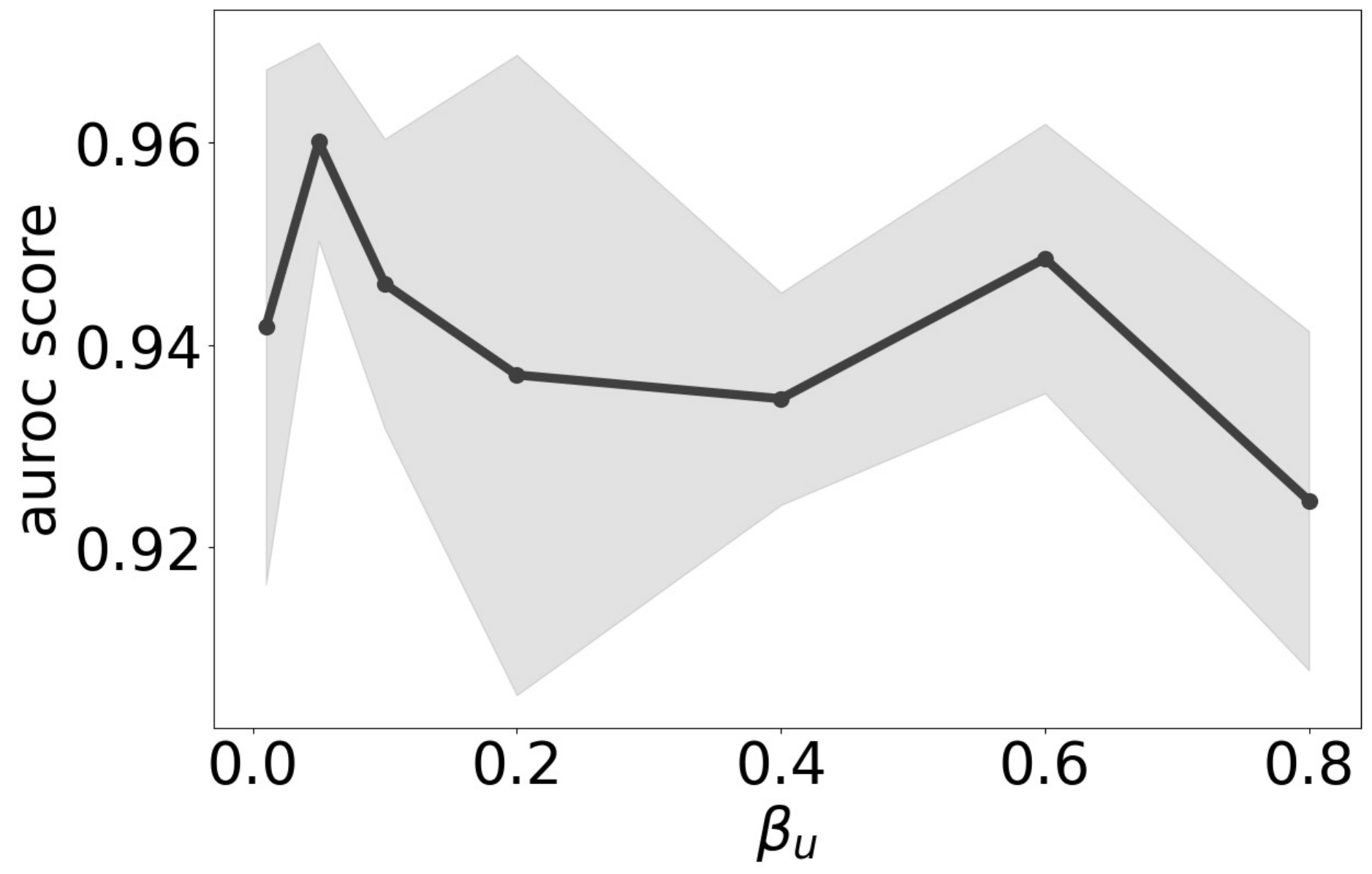}}
    \end{tabular}
    \caption{(best in color) Group-wise fairness metrics and AUROC for (top to bottom) LOF, iForest, DeepAE and FairOD under \underline{target under-representation} on \textit{scattered} outliers. }
    \label{fig:underrep_all_scatter}
\end{figure*}

See Fig. \ref{fig:underrep_all_scatter}.

\subsubsection{Feature measurement bias}
\label{assec:meas_scatter}

\begin{figure*}[!ht]
    \centering
    \begin{tabular}{lllll}
    \hspace{-0.1in}{\includegraphics[width=0.195\textwidth]{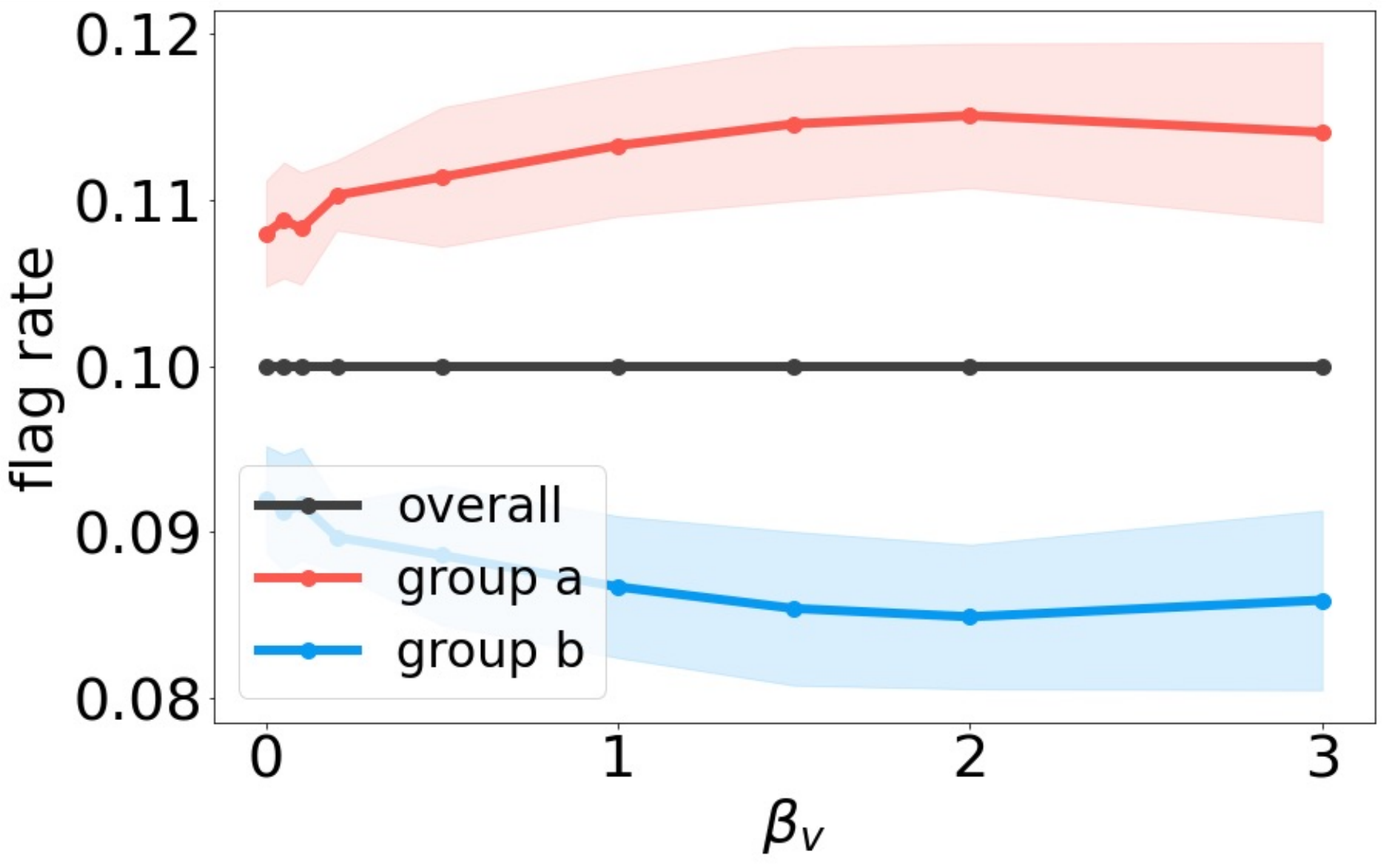}} 
    &
    \hspace{-0.1in}{\includegraphics[width=0.195\textwidth]{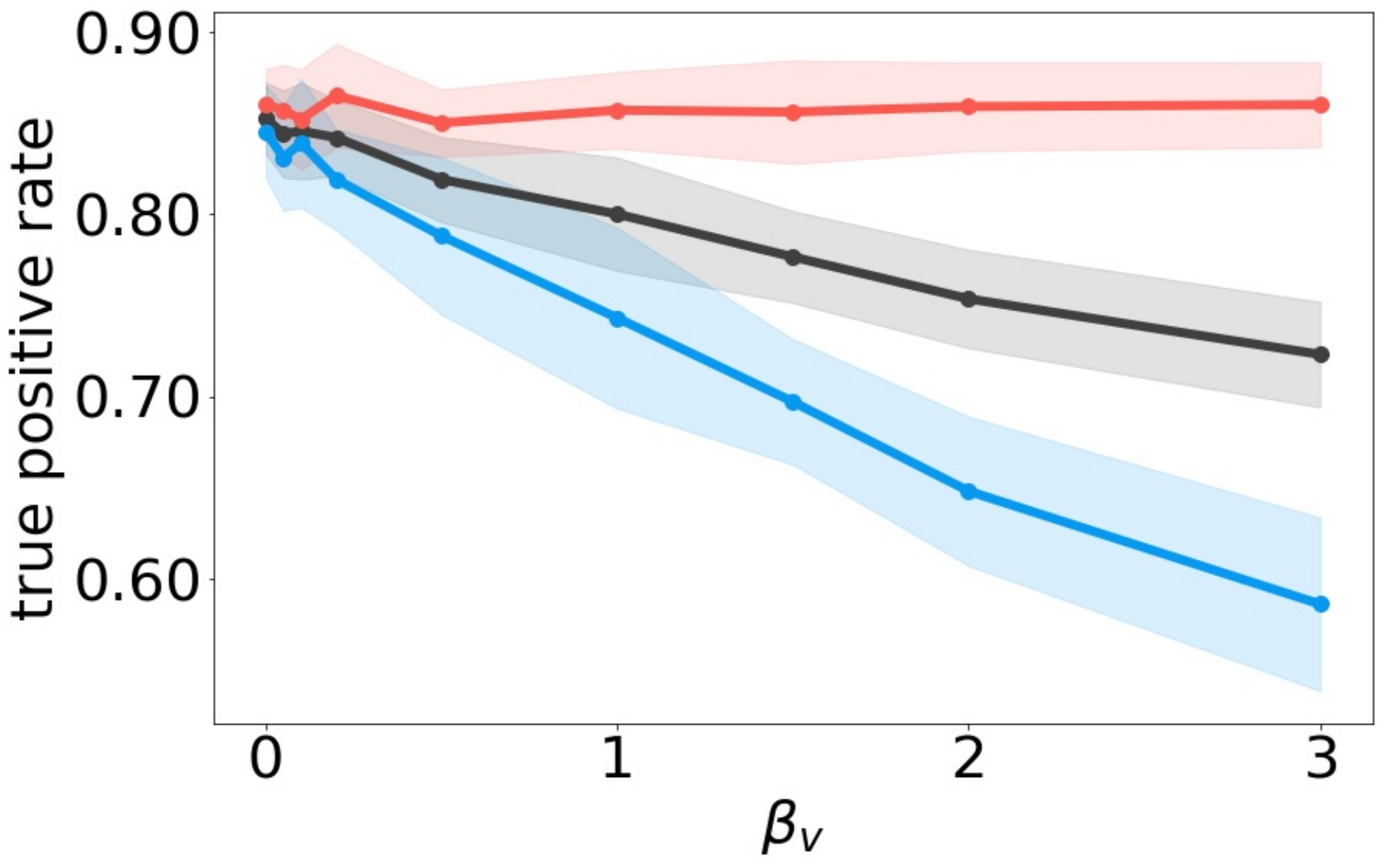}} &
     \hspace{-0.1in}{\includegraphics[width=0.195\textwidth]{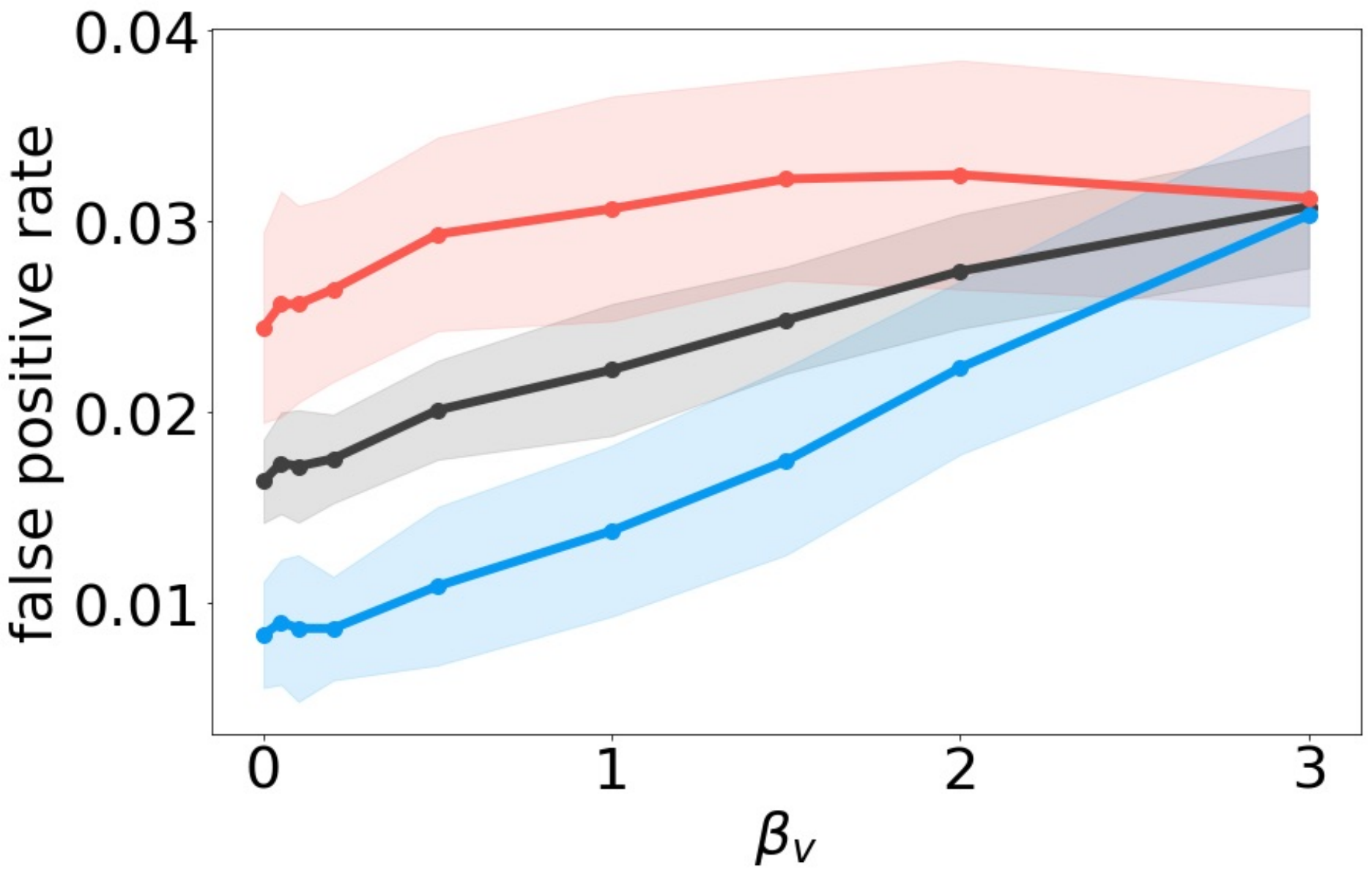}}
    &
   \hspace{-0.1in} {\includegraphics[width=0.195\textwidth]{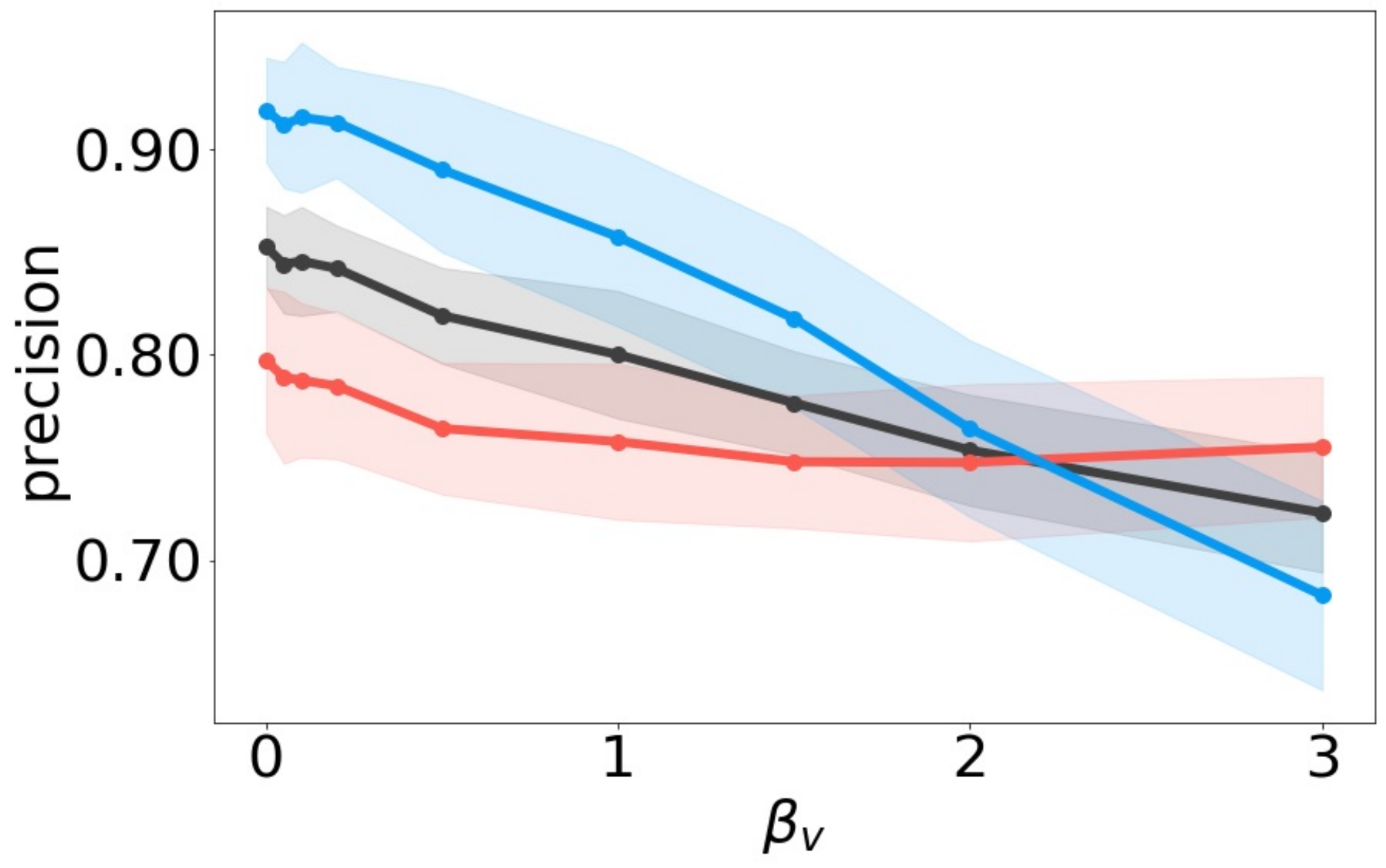}}
   &
   \hspace{-0.1in}
   {\includegraphics[width=0.195\textwidth]{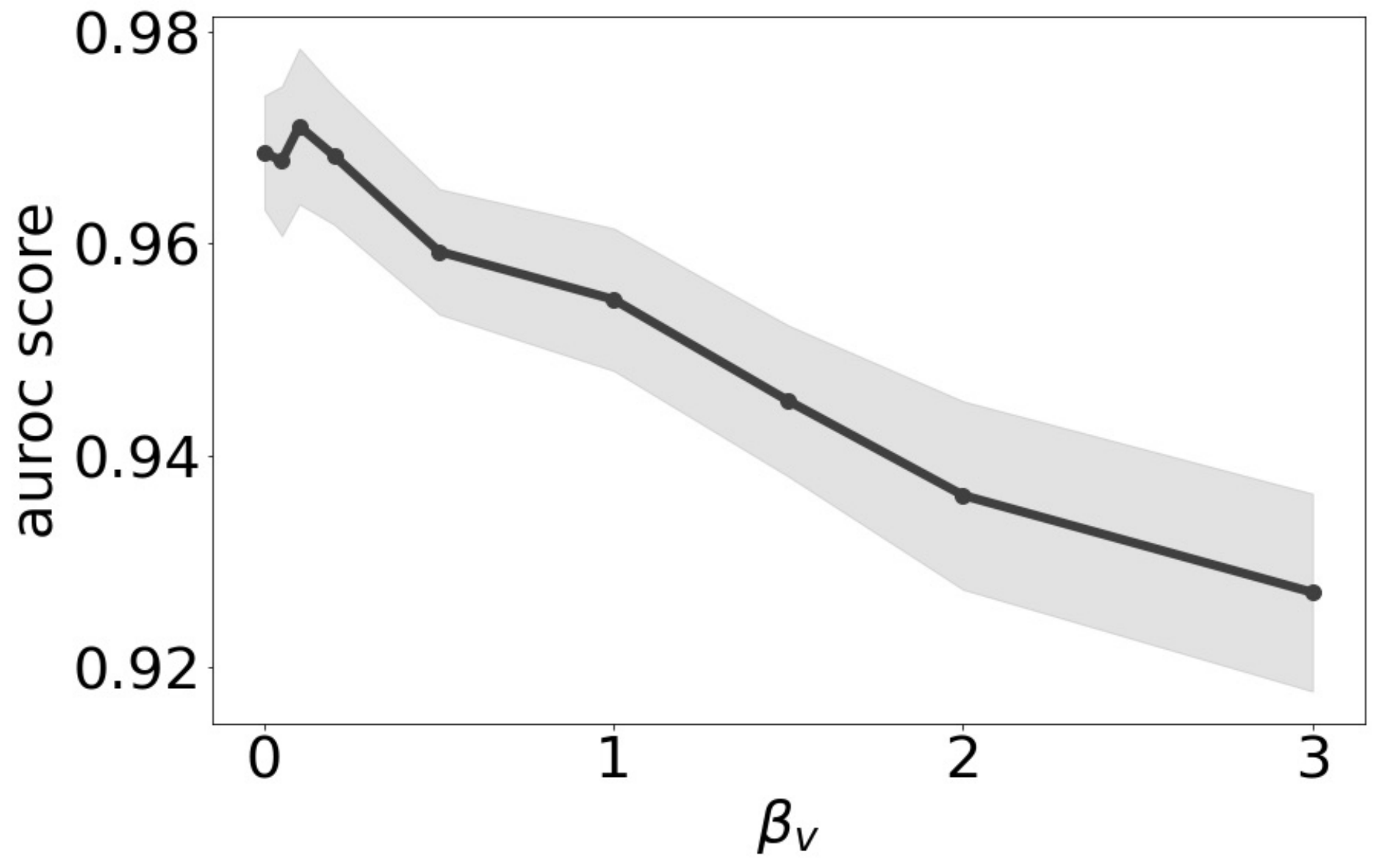}} 
\\
\hspace{-0.1in}{\includegraphics[width=0.195\textwidth]{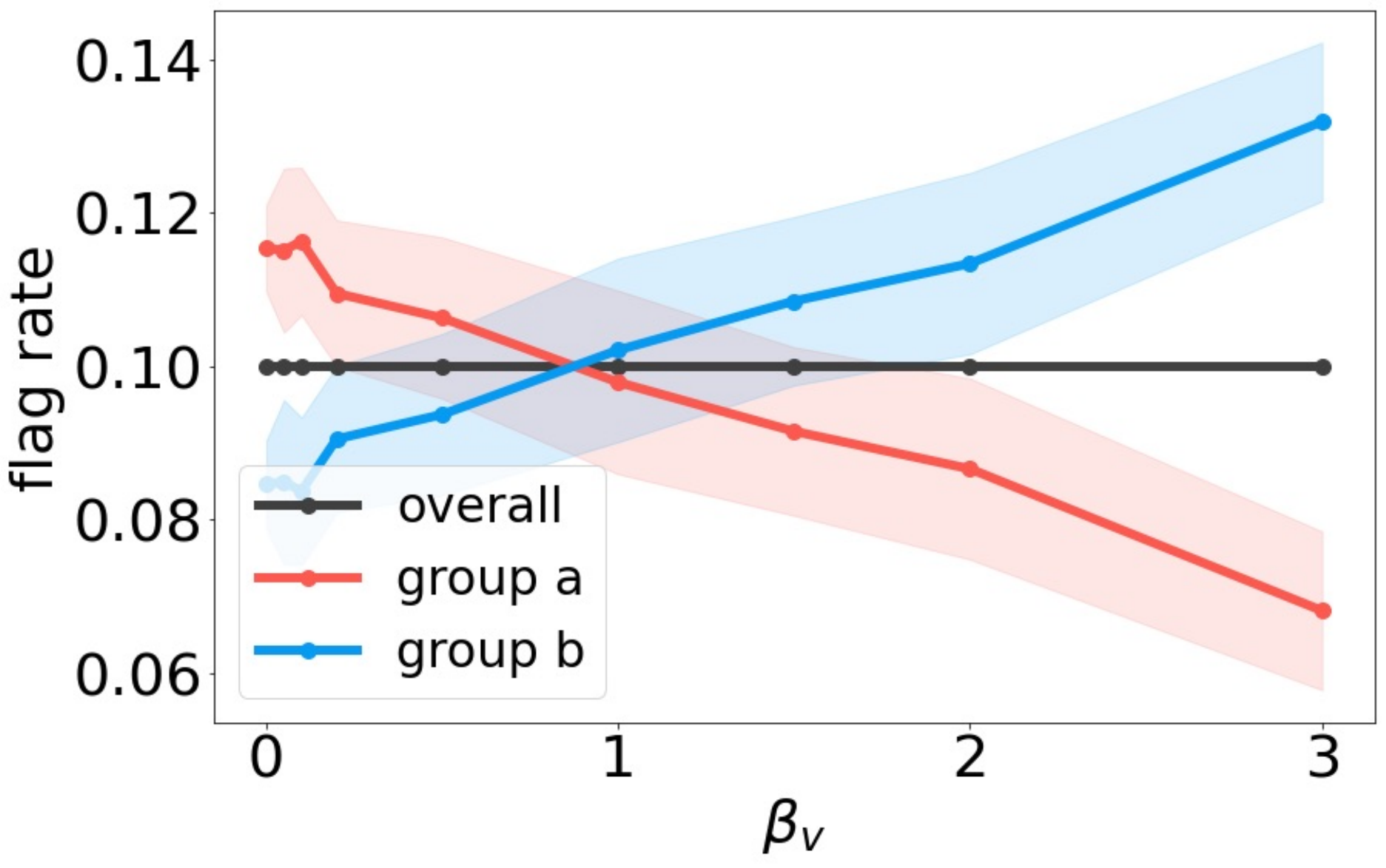}} 
    &
    \hspace{-0.1in}{\includegraphics[width=0.195\textwidth]{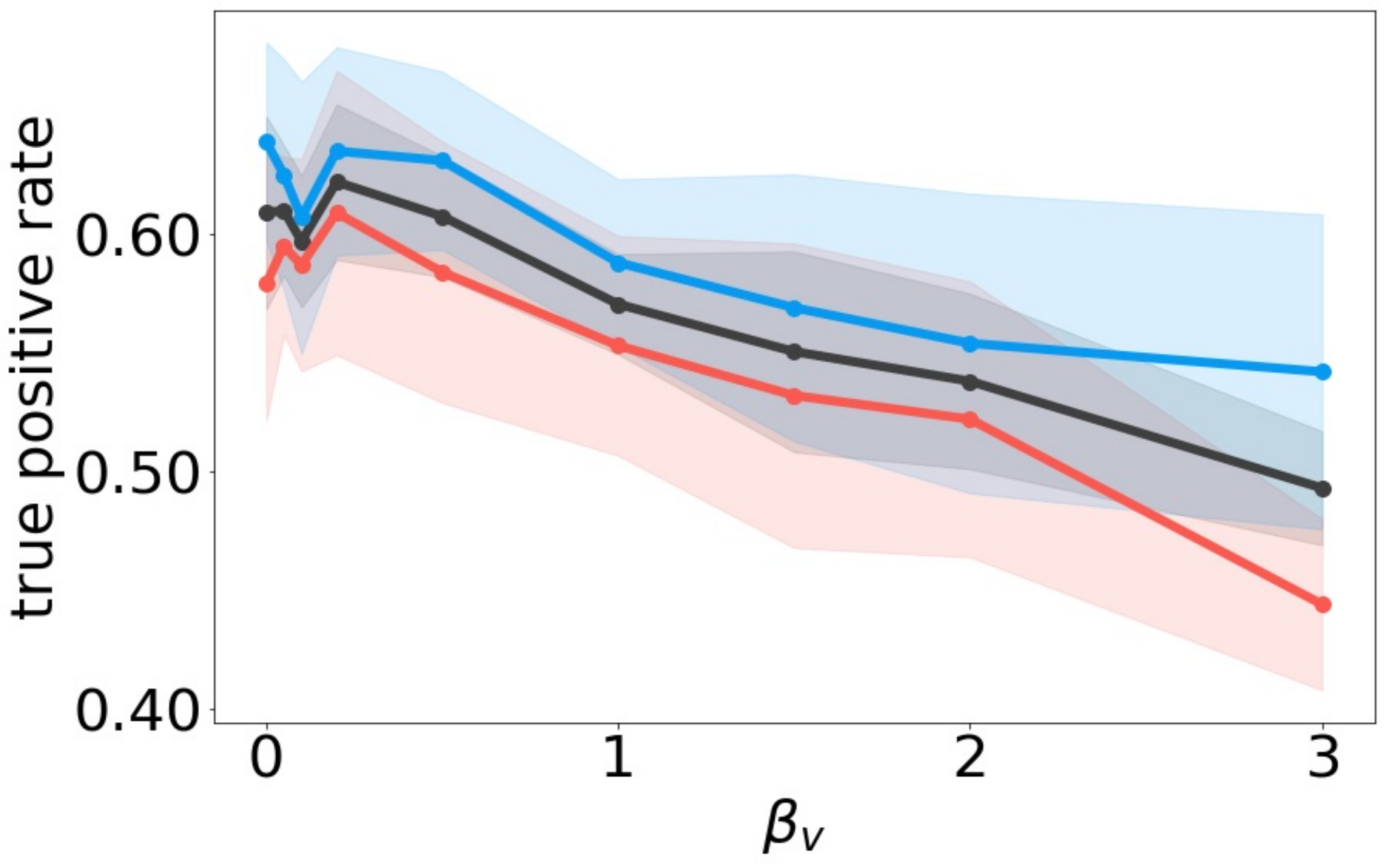}} &
     \hspace{-0.1in}{\includegraphics[width=0.195\textwidth]{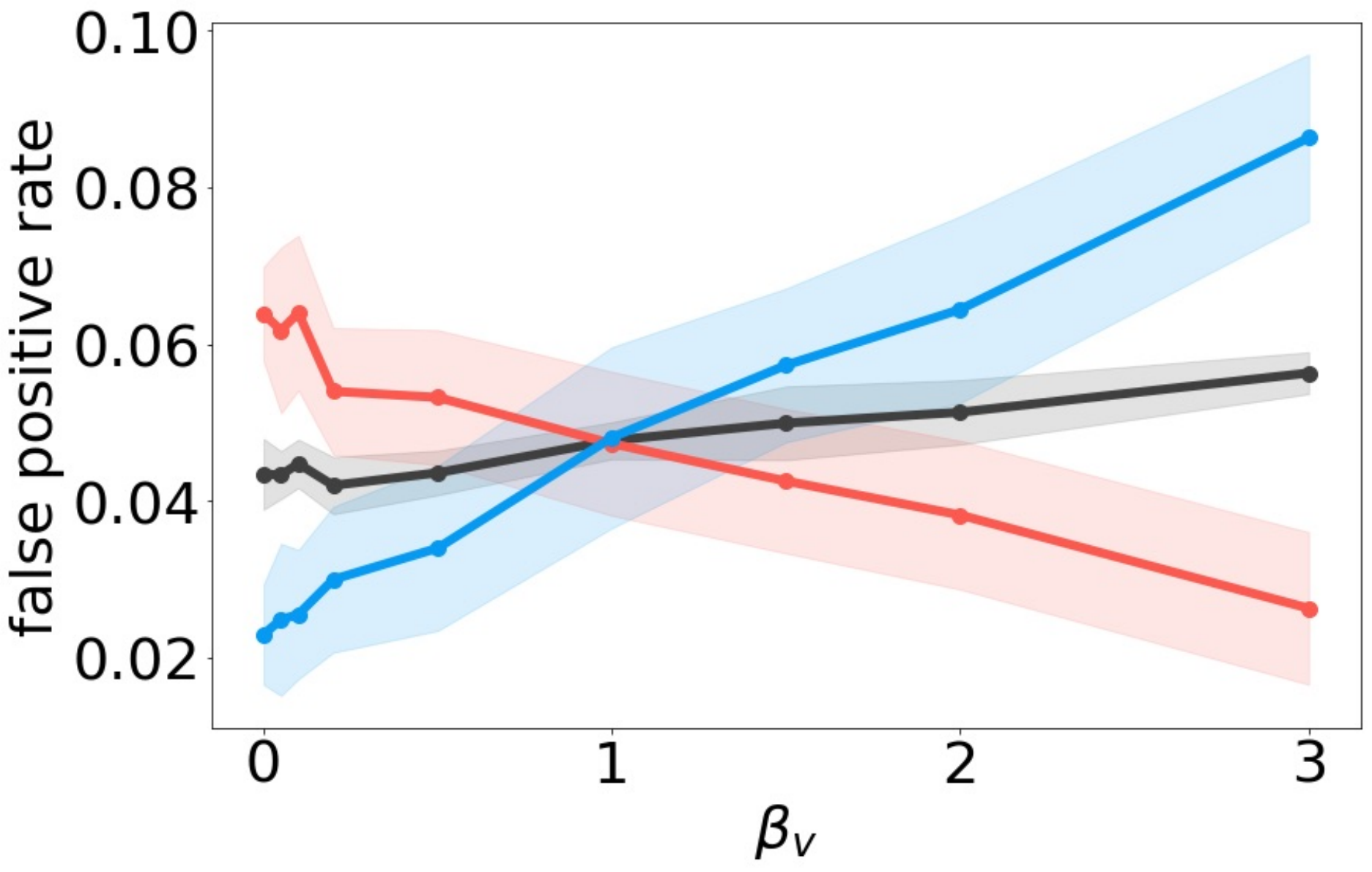}}
    &
   \hspace{-0.1in} {\includegraphics[width=0.195\textwidth]{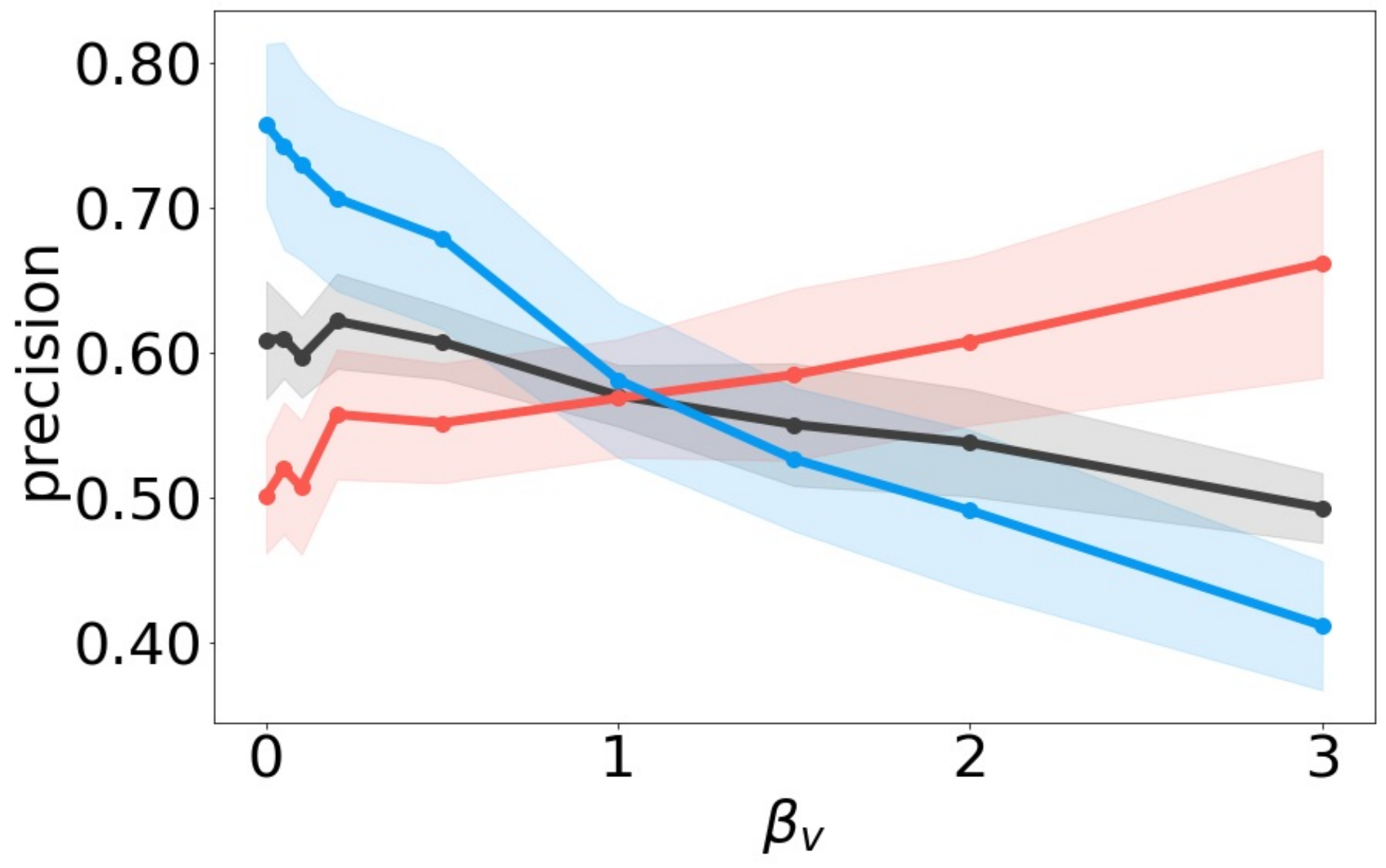}}
   &
    \hspace{-0.1in}{\includegraphics[width=0.195\textwidth]{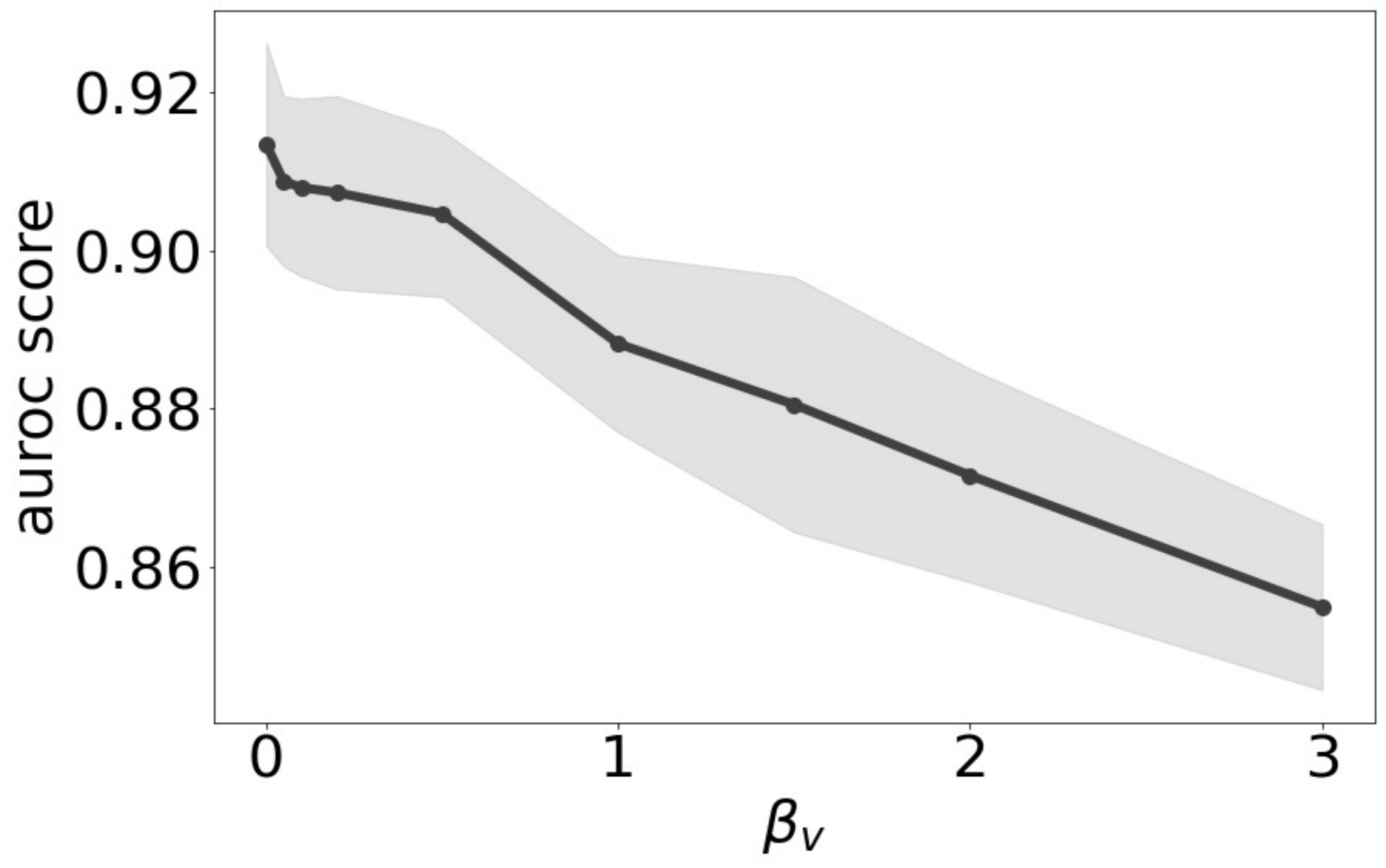}}
   \\
     \hspace{-0.1in}{\includegraphics[width=0.195\textwidth]{FIG/varianceshift/deepae/scatter/fr_legend.pdf}} 
    &
    \hspace{-0.1in}{\includegraphics[width=0.195\textwidth]{FIG/varianceshift/deepae/scatter/tpr.pdf}} &
     \hspace{-0.1in}{\includegraphics[width=0.195\textwidth]{FIG/varianceshift/deepae/scatter/fpr.pdf}}
    &
   \hspace{-0.1in} {\includegraphics[width=0.195\textwidth]{FIG/varianceshift/deepae/scatter/prec.pdf}}
   &
     \hspace{-0.1in}{\includegraphics[width=0.195\textwidth]{FIG/varianceshift/deepae/scatter/auroc.pdf}}
   \\
   \hspace{-0.1in}{\includegraphics[width=0.195\textwidth]{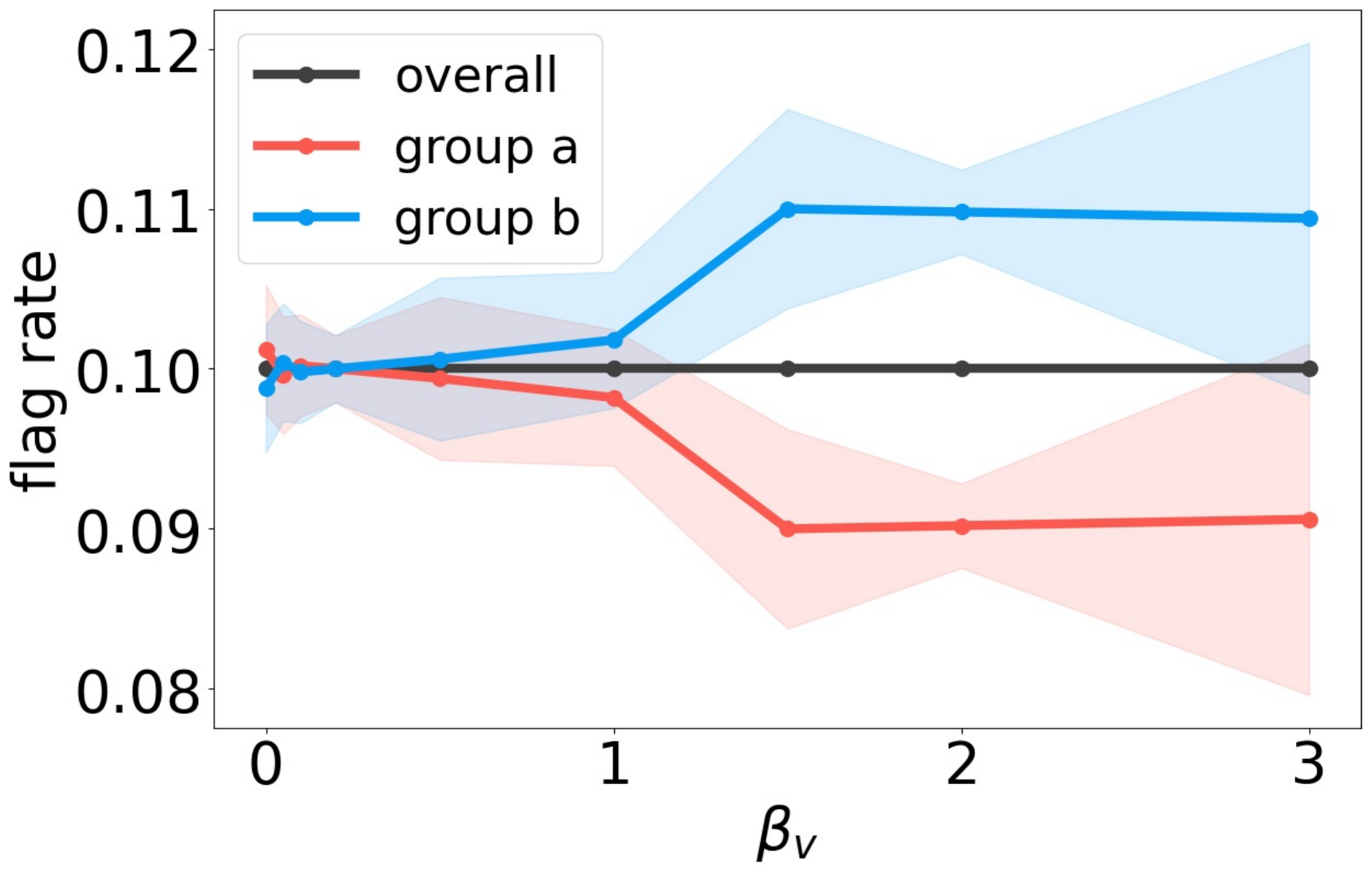} }
    &
    \hspace{-0.1in}{\includegraphics[width=0.195\textwidth]{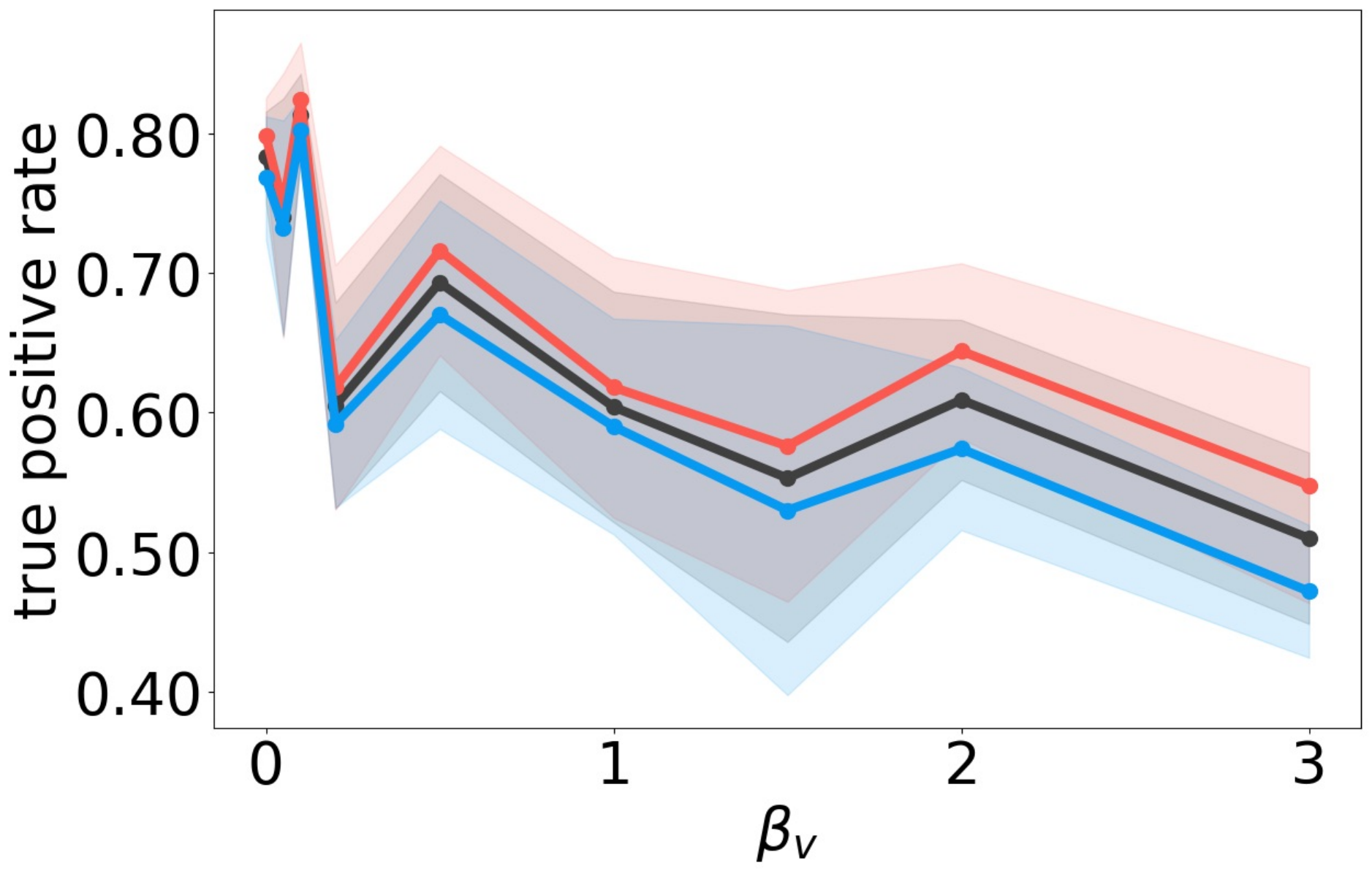}} &
     \hspace{-0.1in}{\includegraphics[width=0.195\textwidth]{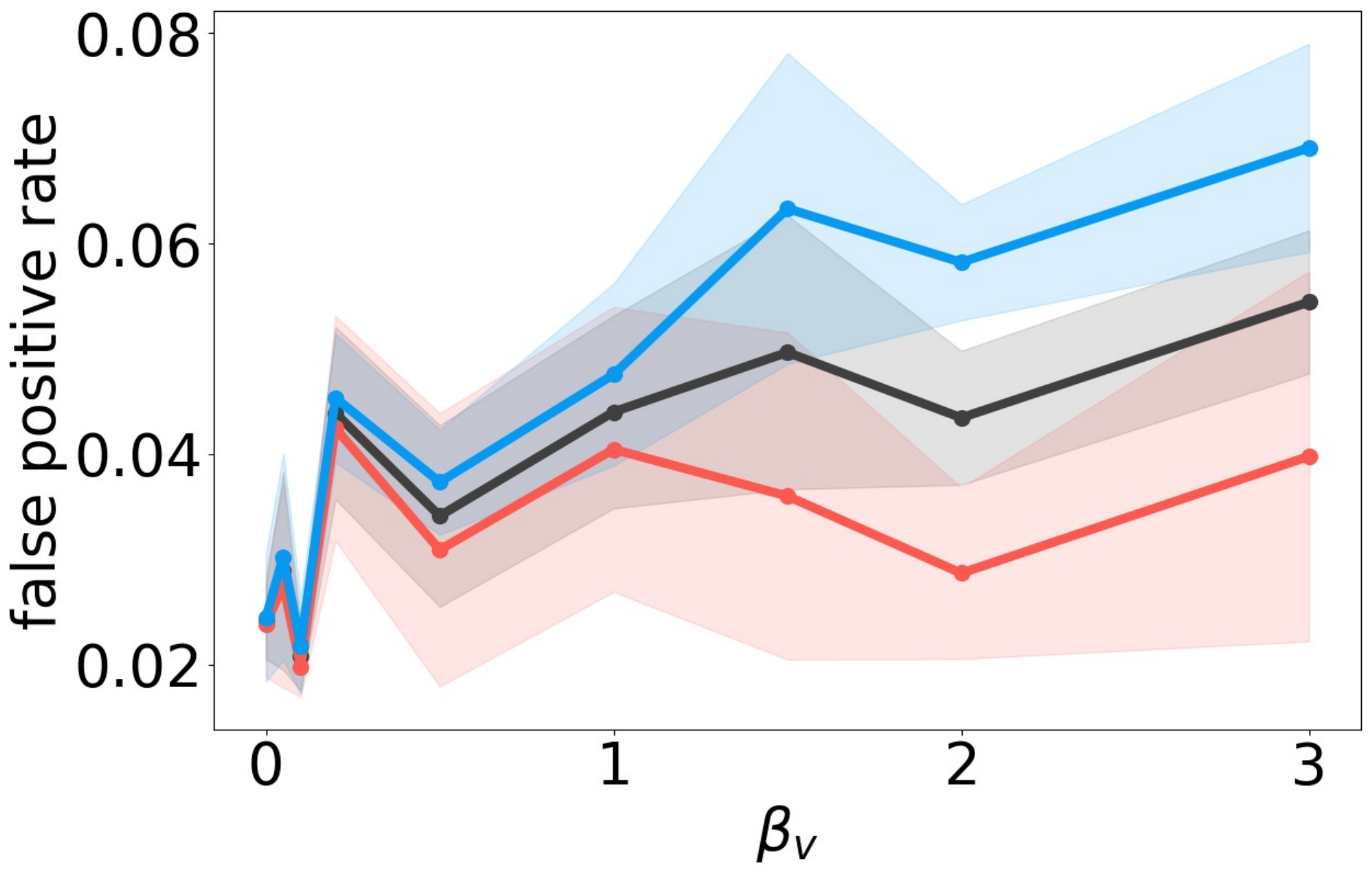}}
    &
   \hspace{-0.1in} {\includegraphics[width=0.195\textwidth]{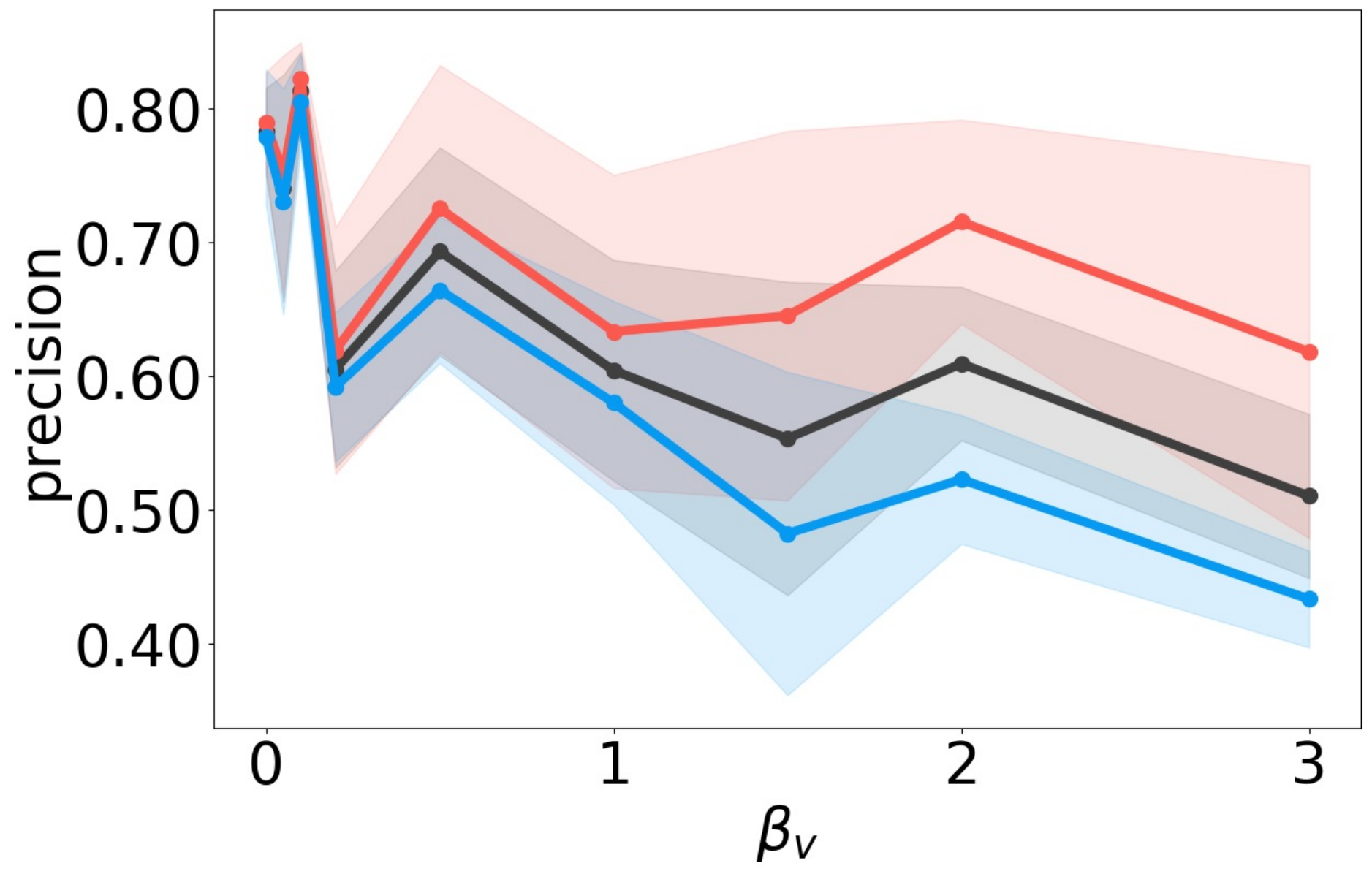}}
   &
   \hspace{-0.1in} {\includegraphics[width=0.195\textwidth]{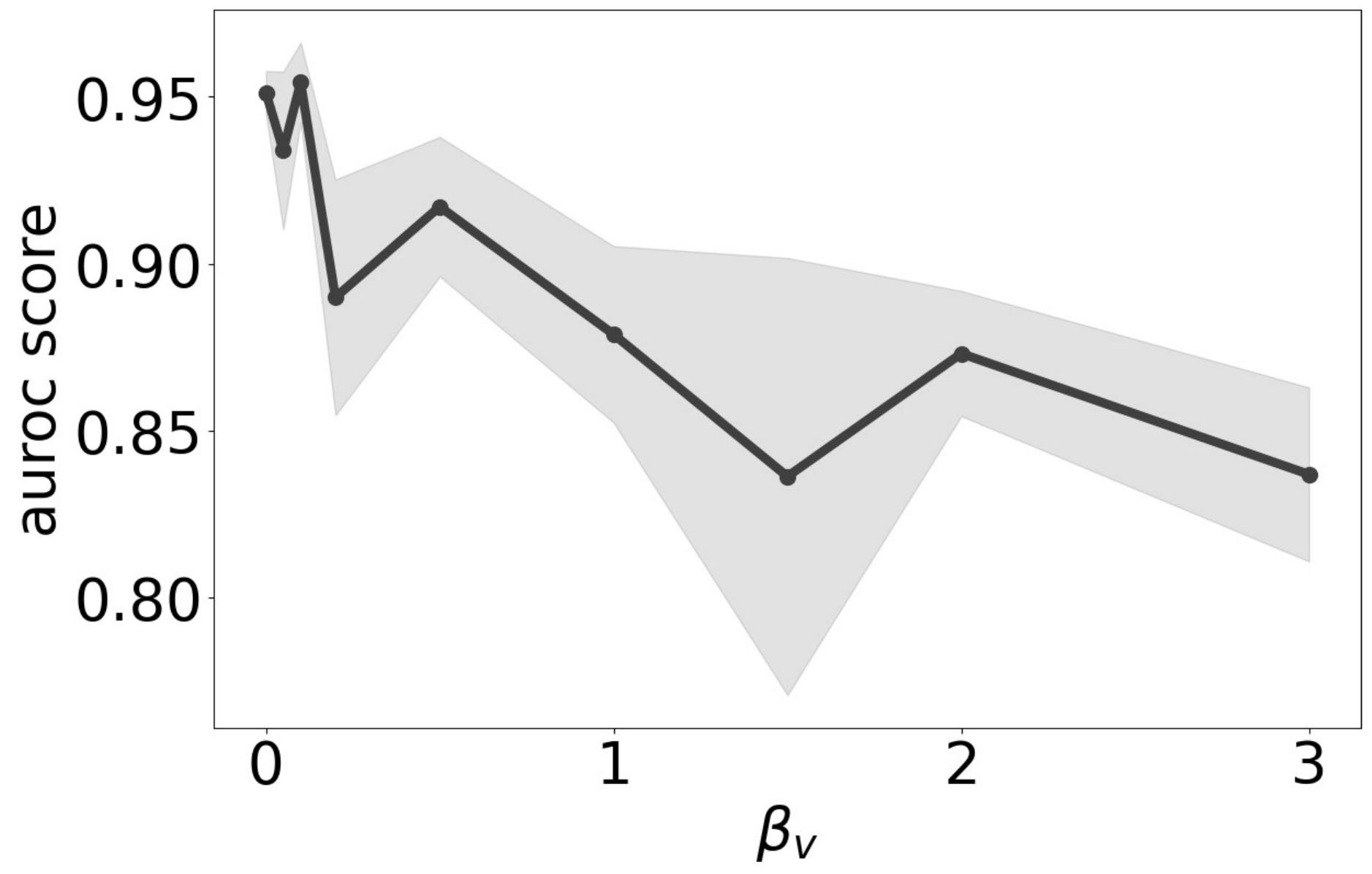}}
    \end{tabular}

    \caption{(best in color) Group-wise fairness metrics and AUROC for (top to bottom) LOF, iForest, DeepAE and FairOD under \underline{feature measurement bias} (variance shift) on \textit{scattered} outliers. }
    \label{fig:varianceshift_all_scatter}
\end{figure*}

\begin{figure*}[!ht]
    \centering
    \begin{tabular}{lllll}
    \hspace{-0.1in}{\includegraphics[width=0.195\textwidth]{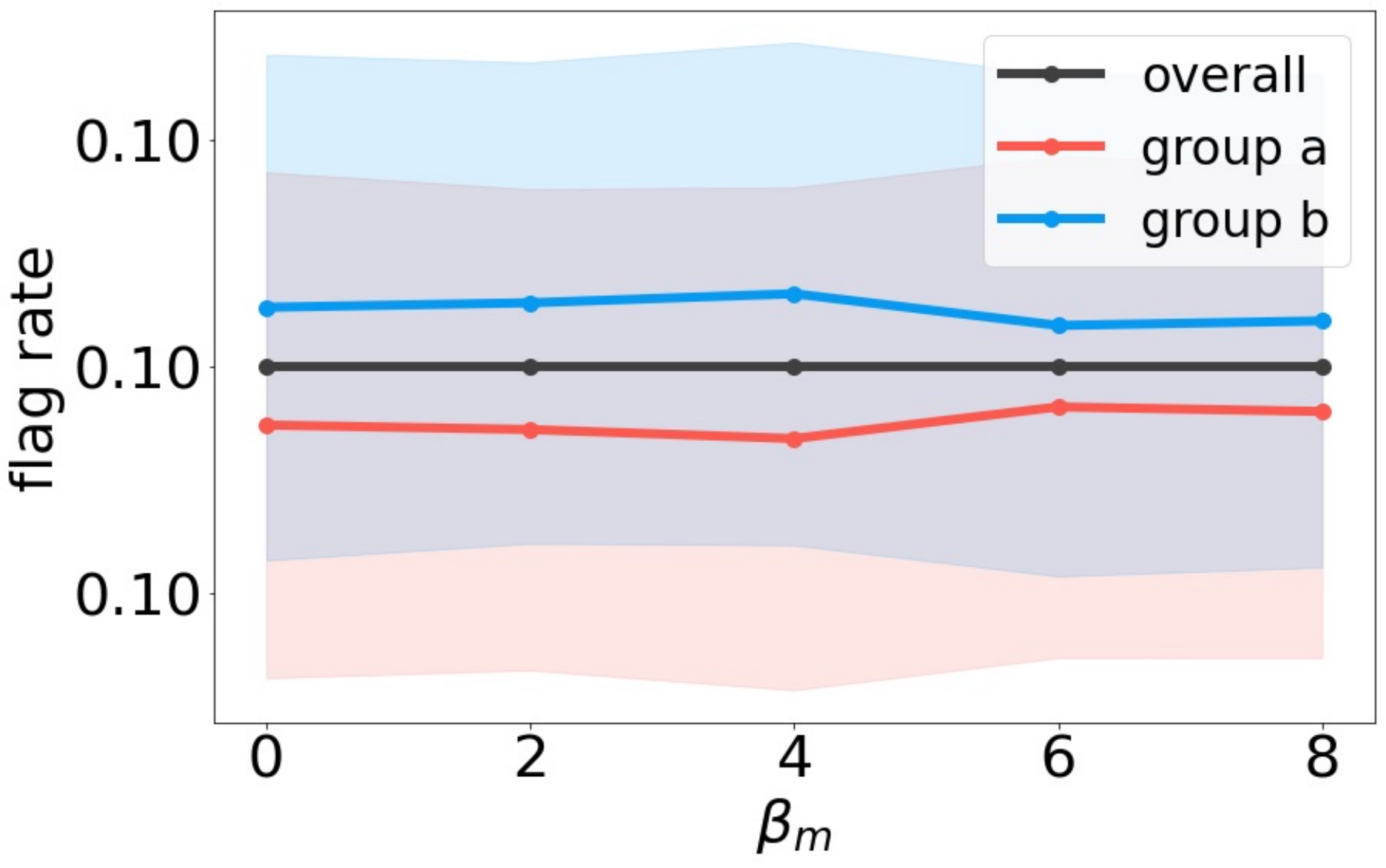}} 
    &
    \hspace{-0.1in}{\includegraphics[width=0.195\textwidth]{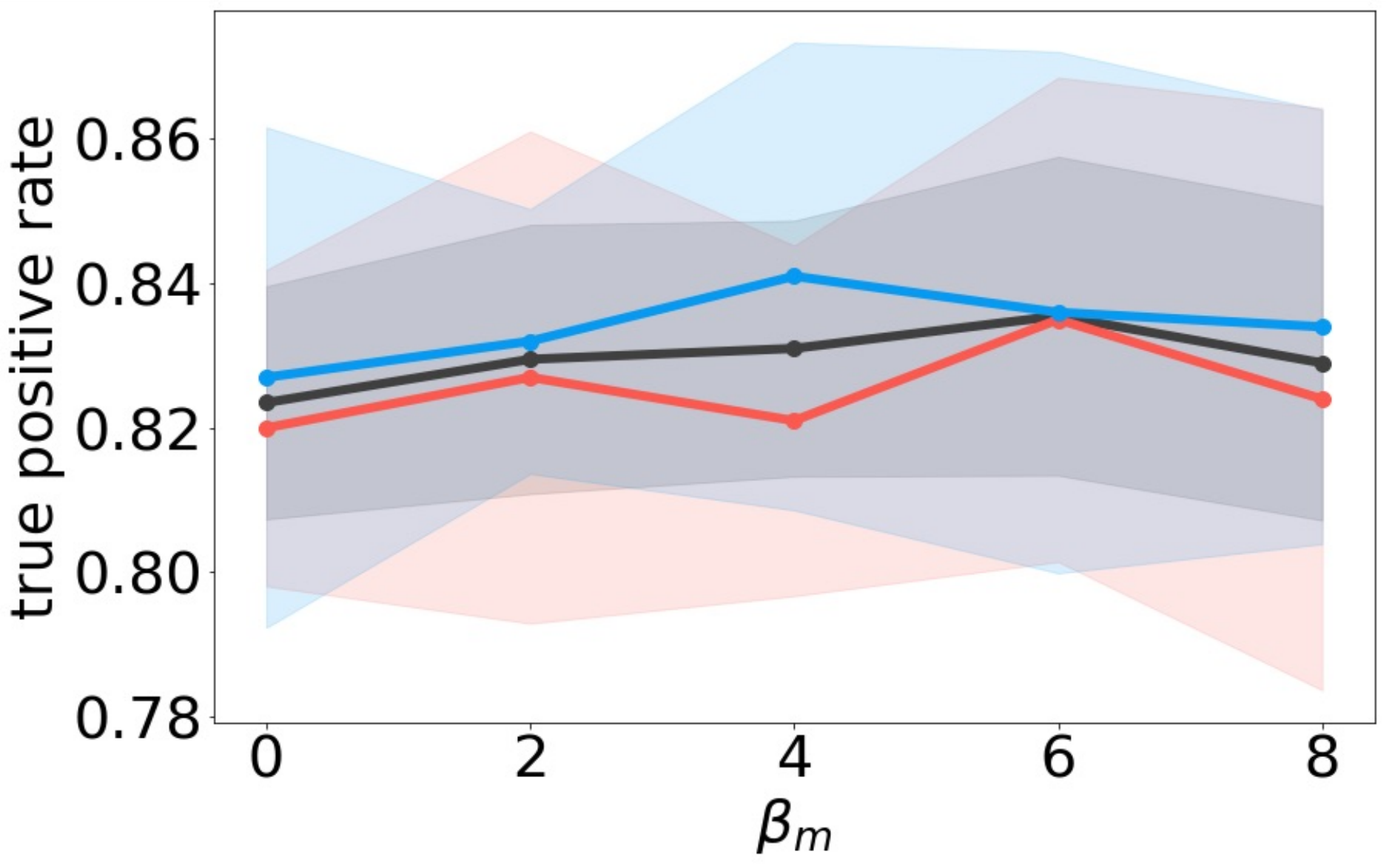}} &
     \hspace{-0.1in}{\includegraphics[width=0.195\textwidth]{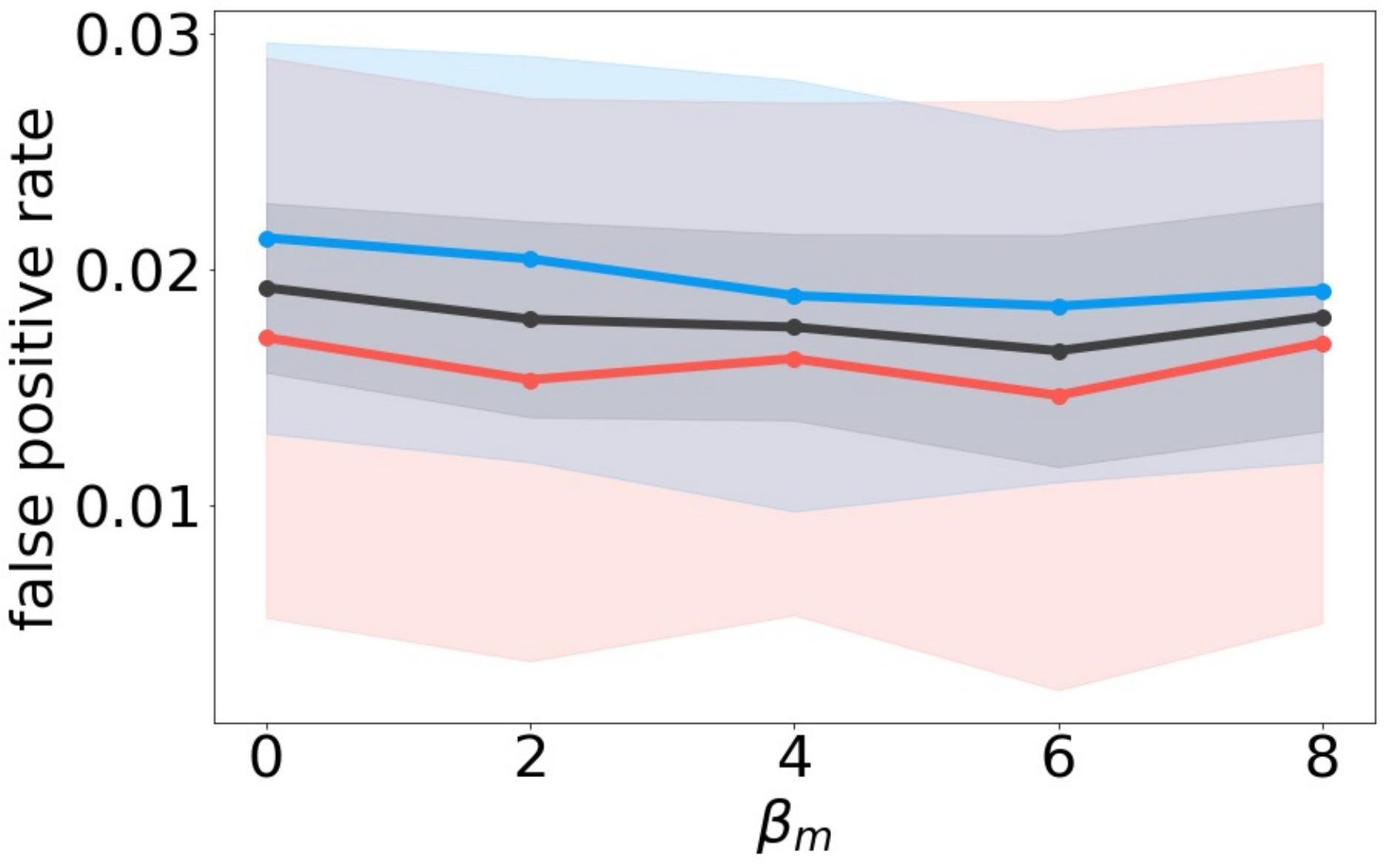}}
    &
   \hspace{-0.1in} {\includegraphics[width=0.195\textwidth]{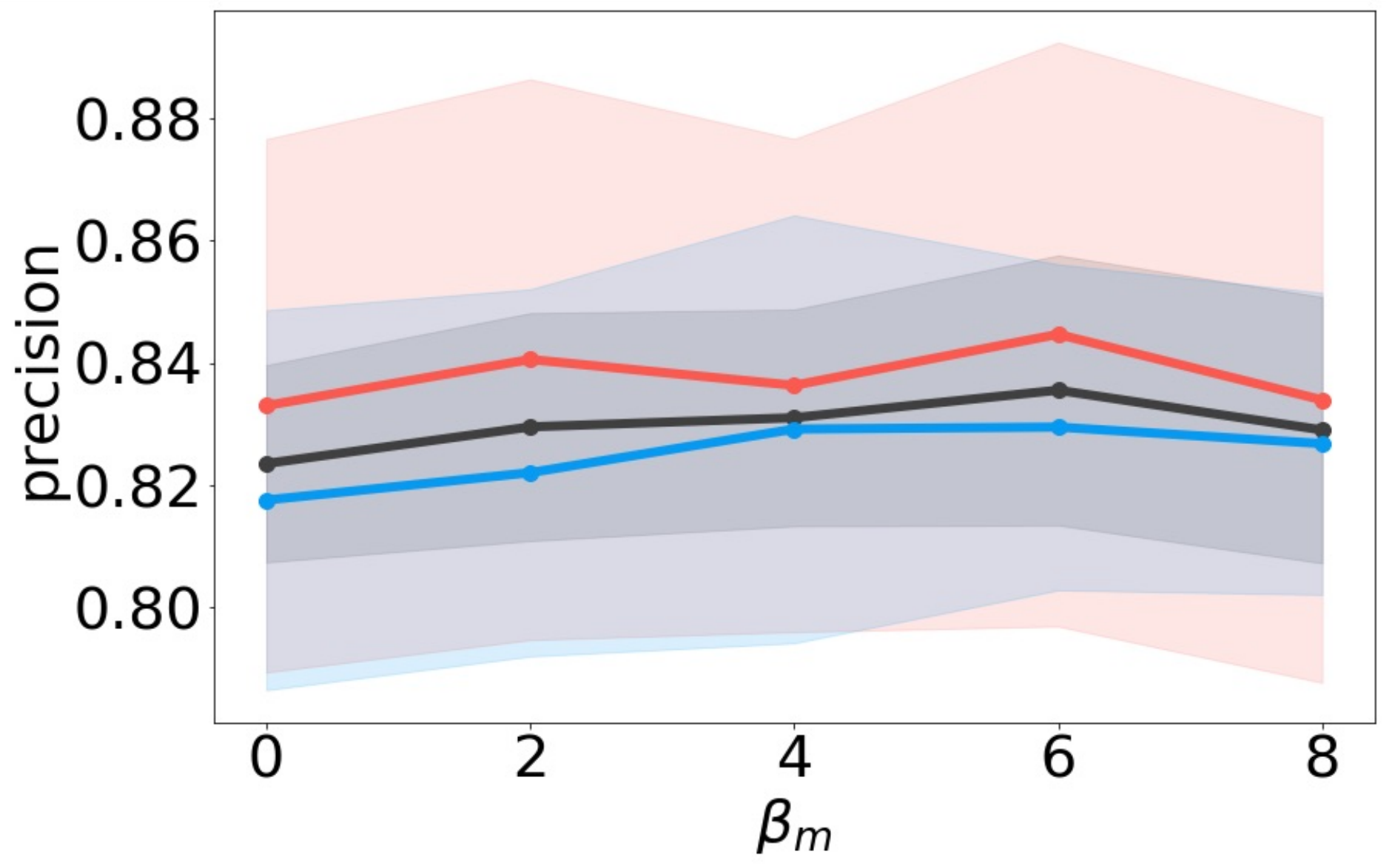}}
   &
   \hspace{-0.1in}
   {\includegraphics[width=0.195\textwidth]{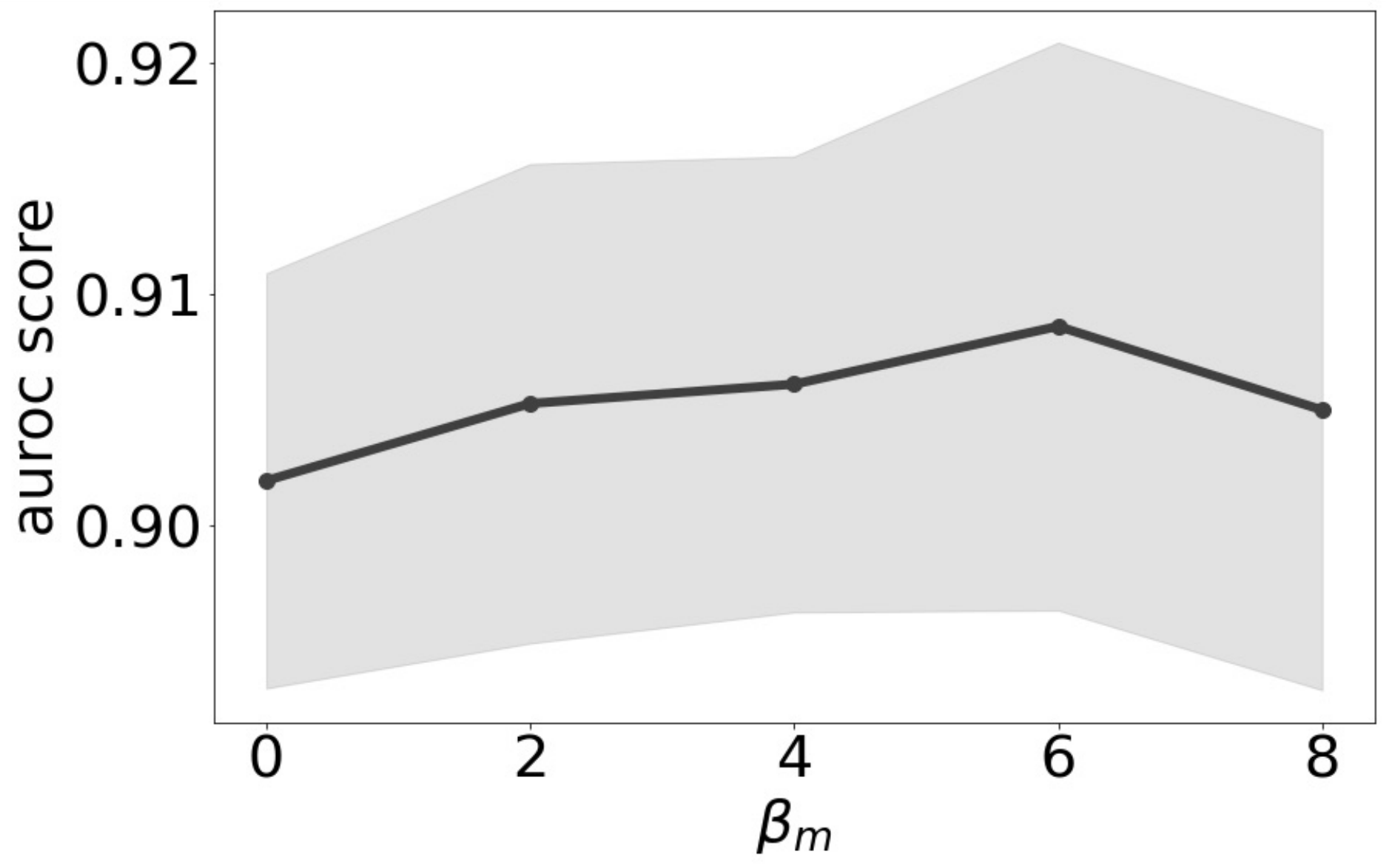}} 
\\
\hspace{-0.1in}{\includegraphics[width=0.195\textwidth]{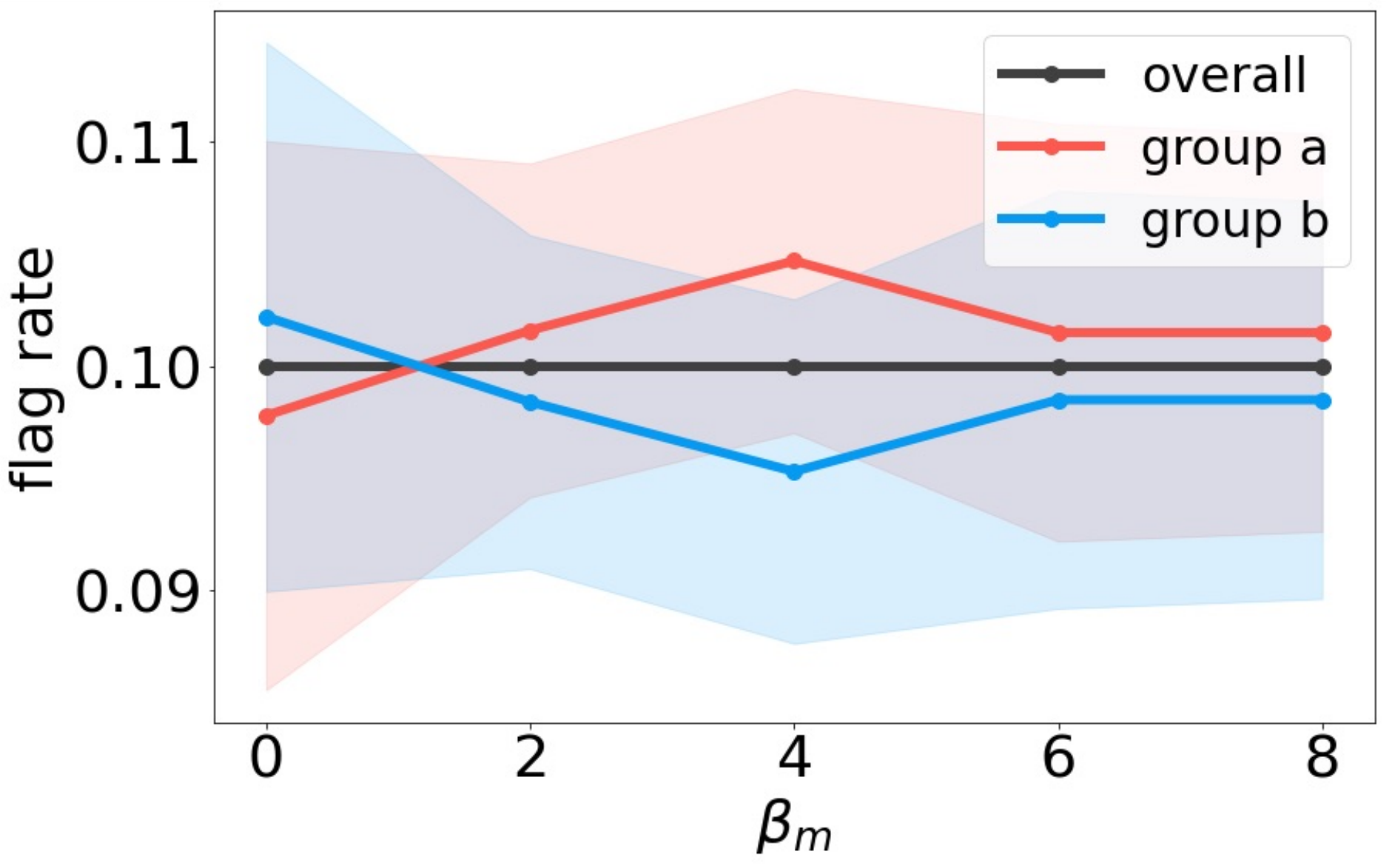}} 
    &
    \hspace{-0.1in}{\includegraphics[width=0.195\textwidth]{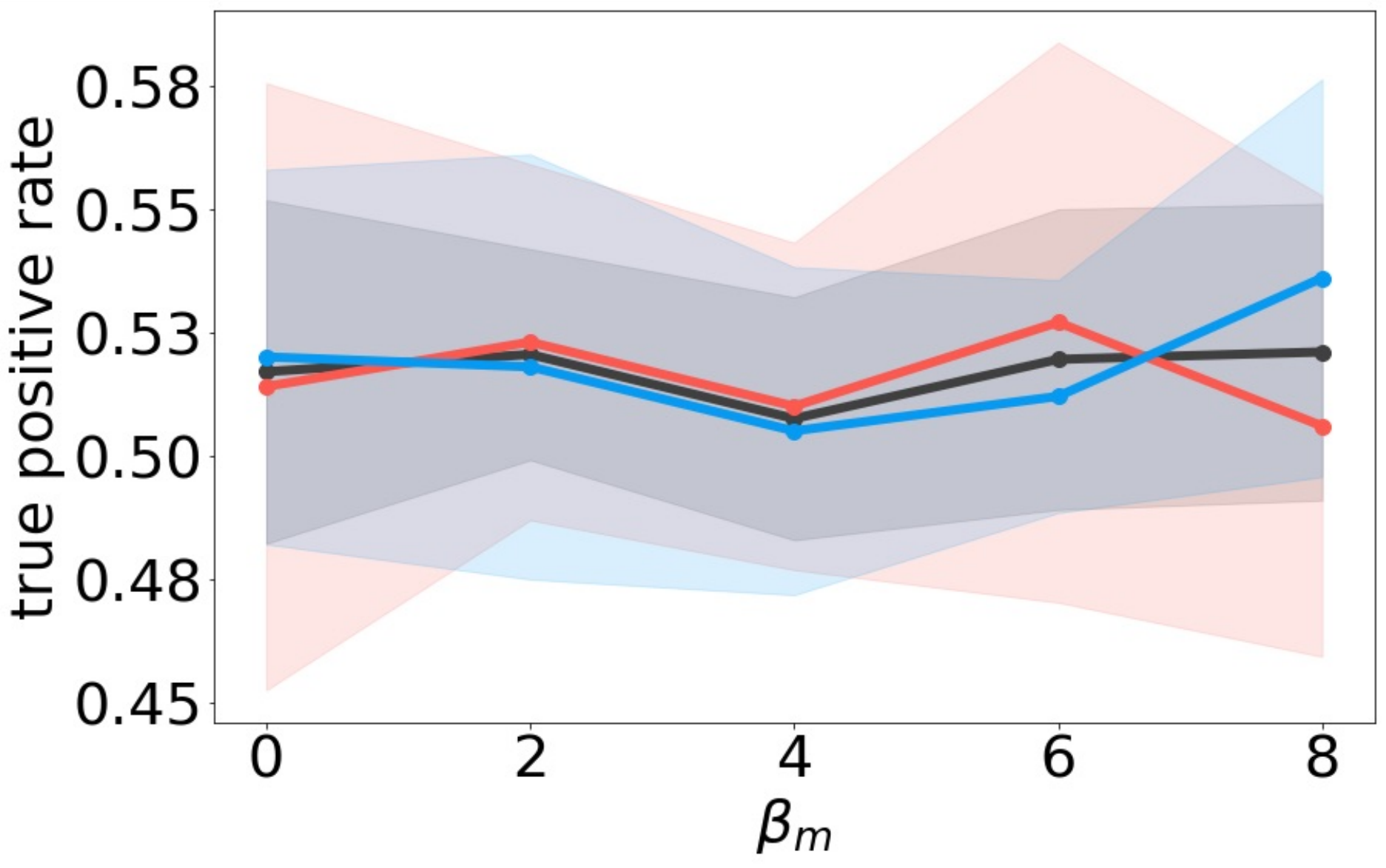}} &
     \hspace{-0.1in}{\includegraphics[width=0.195\textwidth]{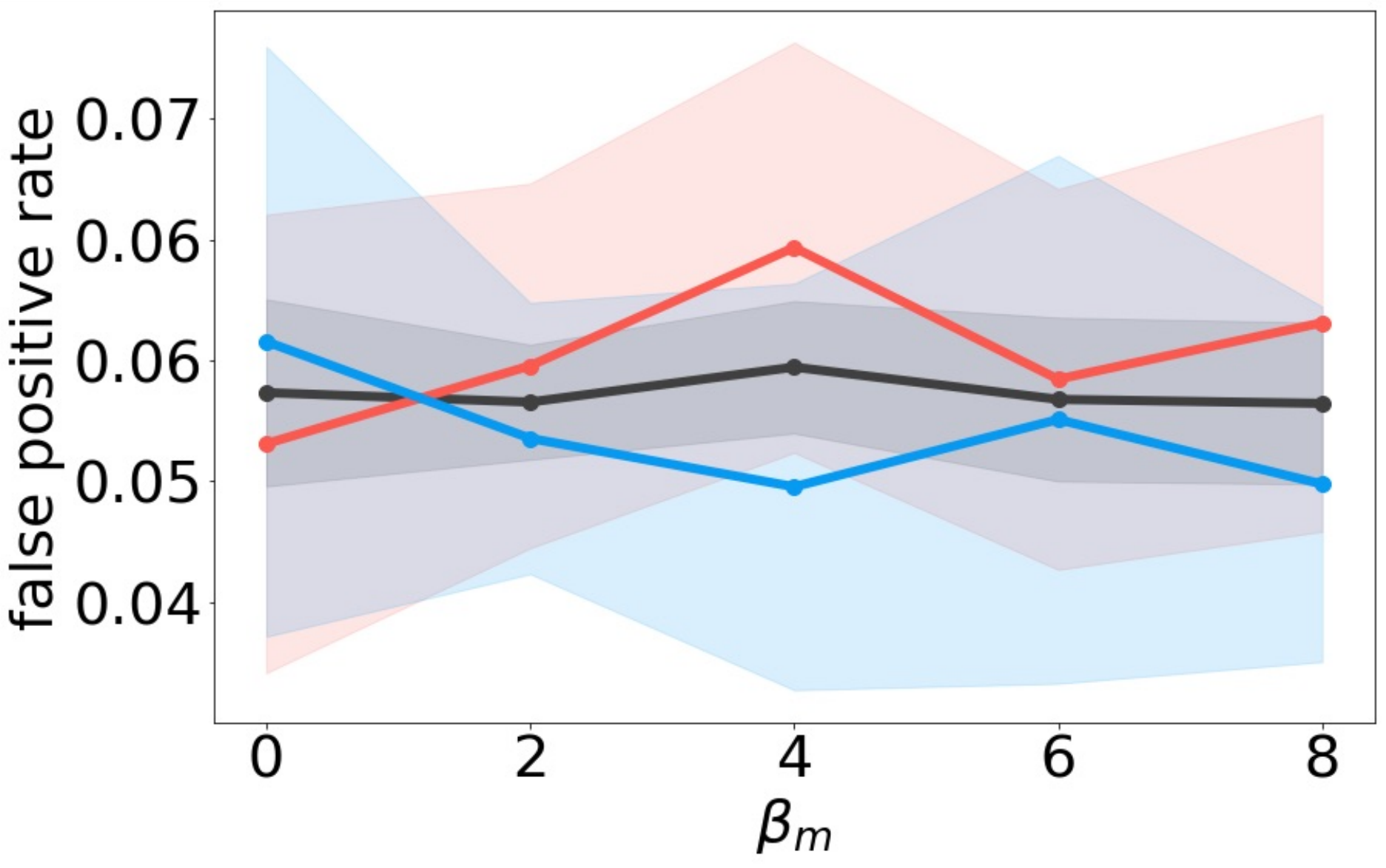}}
    &
   \hspace{-0.1in} {\includegraphics[width=0.195\textwidth]{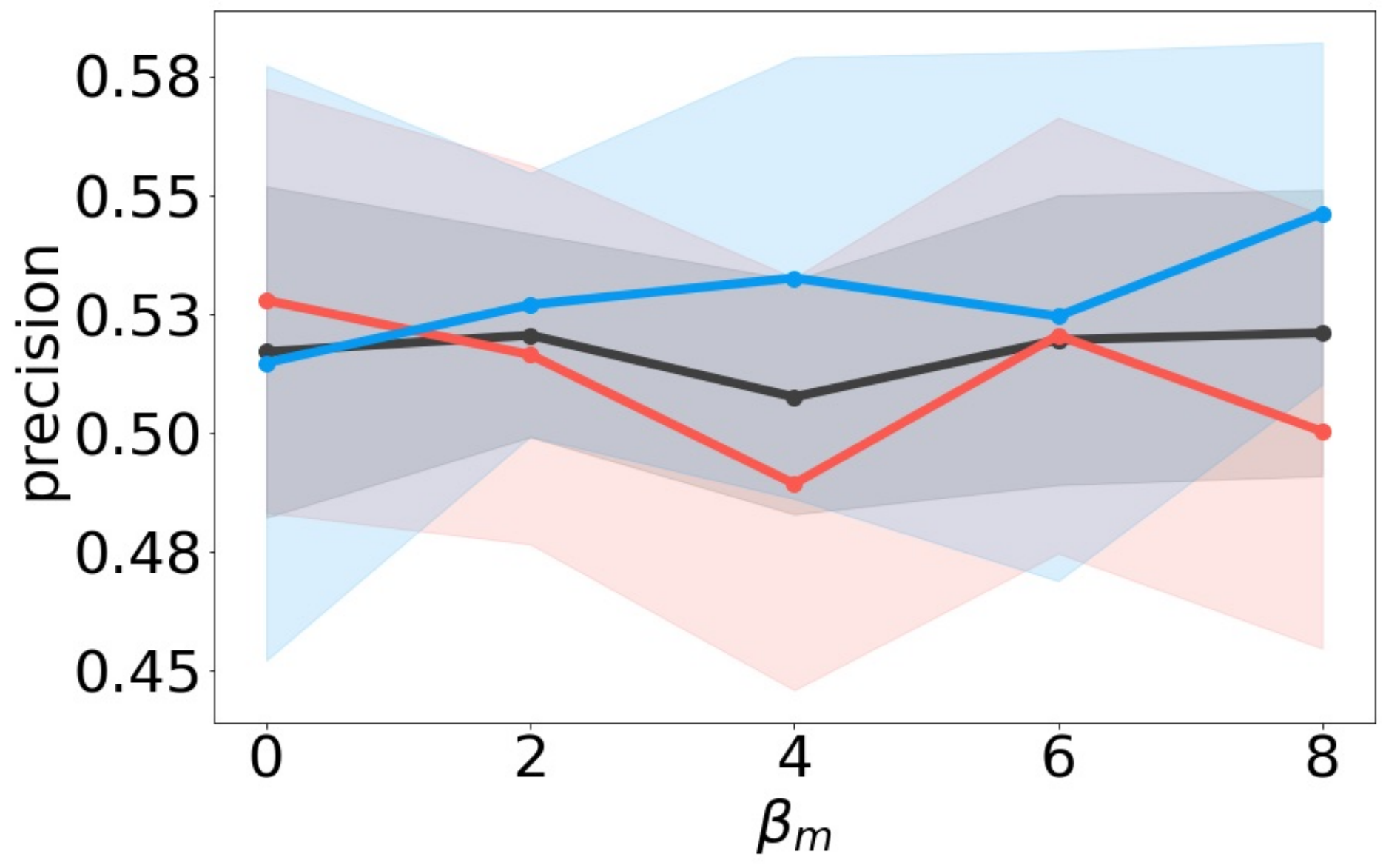}}
   &
    \hspace{-0.1in}{\includegraphics[width=0.195\textwidth]{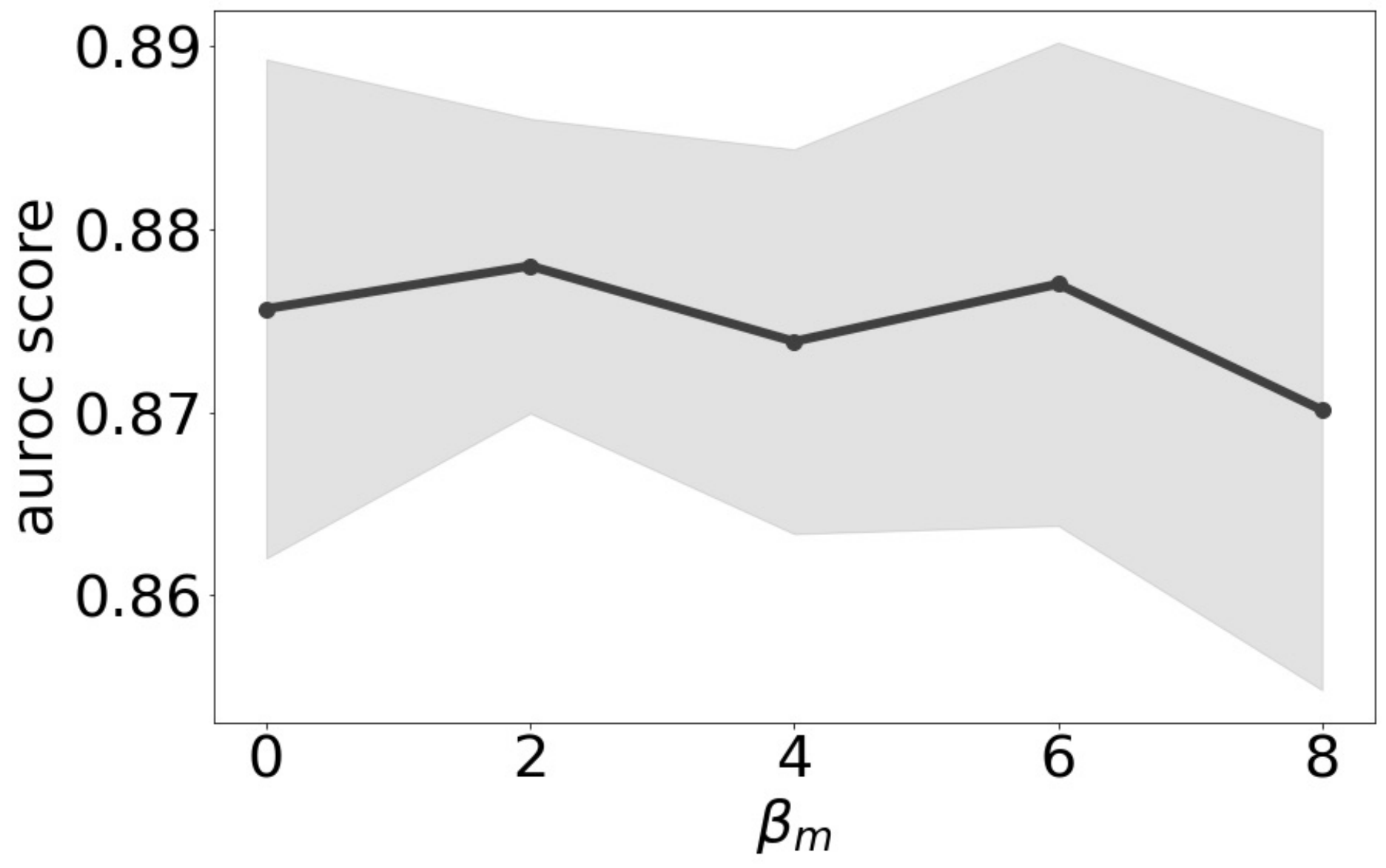}}
    \end{tabular}

    \caption{(best in color) Group-wise fairness metrics and AUROC for (top) LOF and (bottom) iForest under \underline{feature measurement bias} (mean shift) on \textit{scattered} outliers. }
    \label{fig:meanshift_lofif_scatter}
\end{figure*}

See Fig. \ref{fig:varianceshift_all_scatter} for variance shift and Fig. 
\ref{fig:meanshift_lofif_scatter} for mean shift results (LOF and iForest only for brevity).

\subsubsection{Membership obfuscation bias}
\label{assec:obf_scatter}

\begin{figure*}[!ht]
    \centering
    \begin{tabular}{lllll}
    \hspace{-0.1in}{\includegraphics[width=0.195\textwidth]{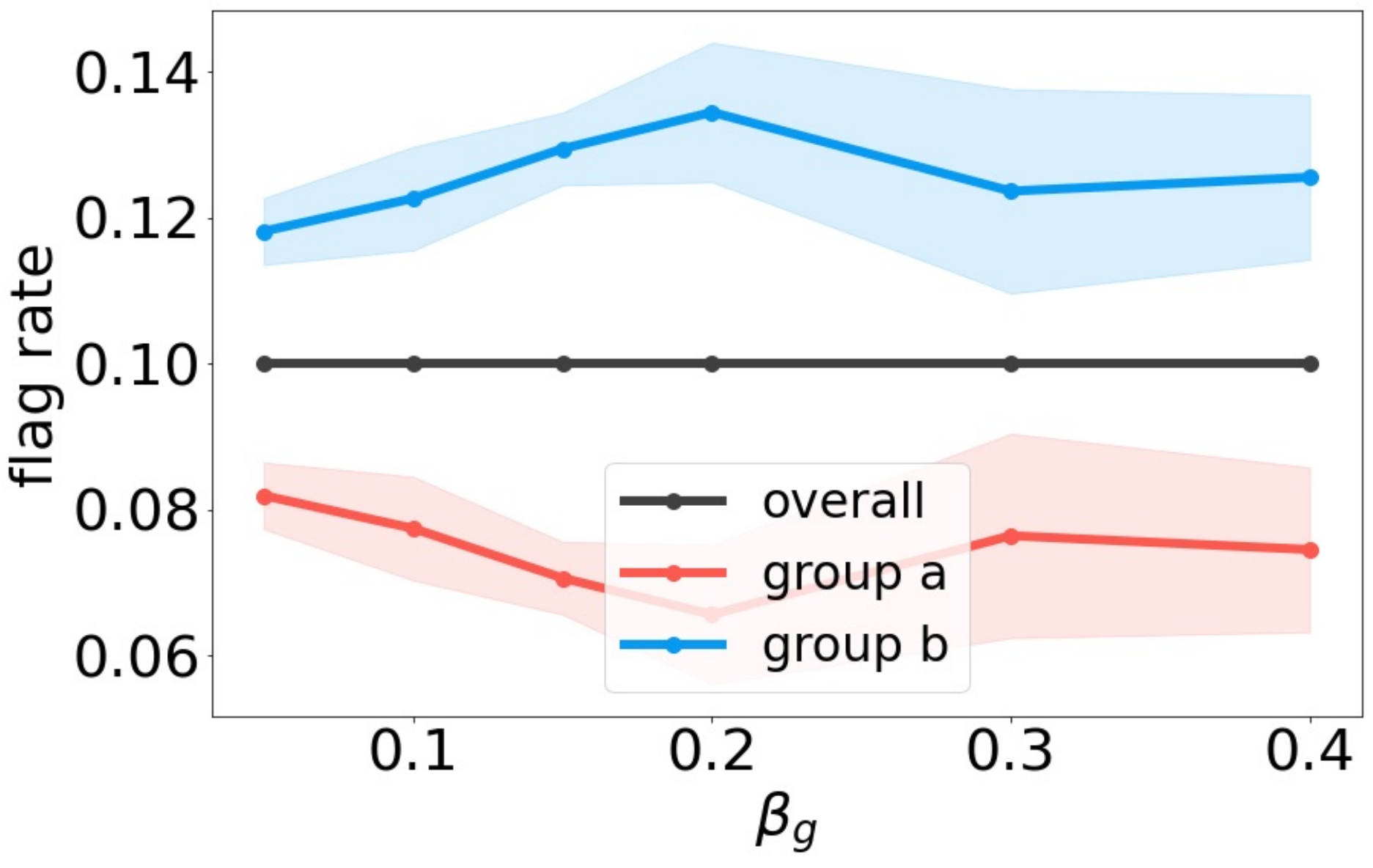}} 
    &
    \hspace{-0.1in}{\includegraphics[width=0.195\textwidth]{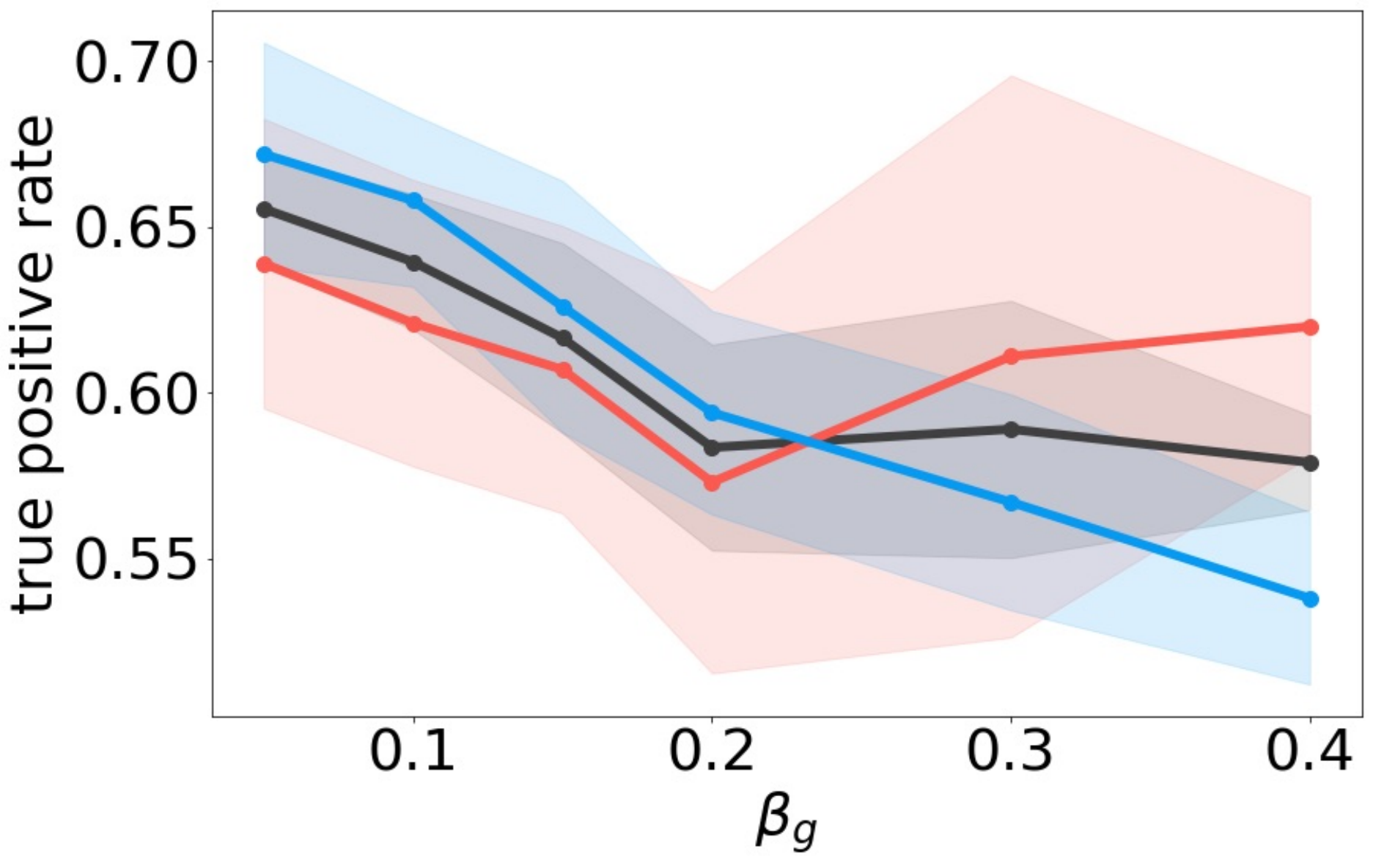}} &
     \hspace{-0.1in}{\includegraphics[width=0.195\textwidth]{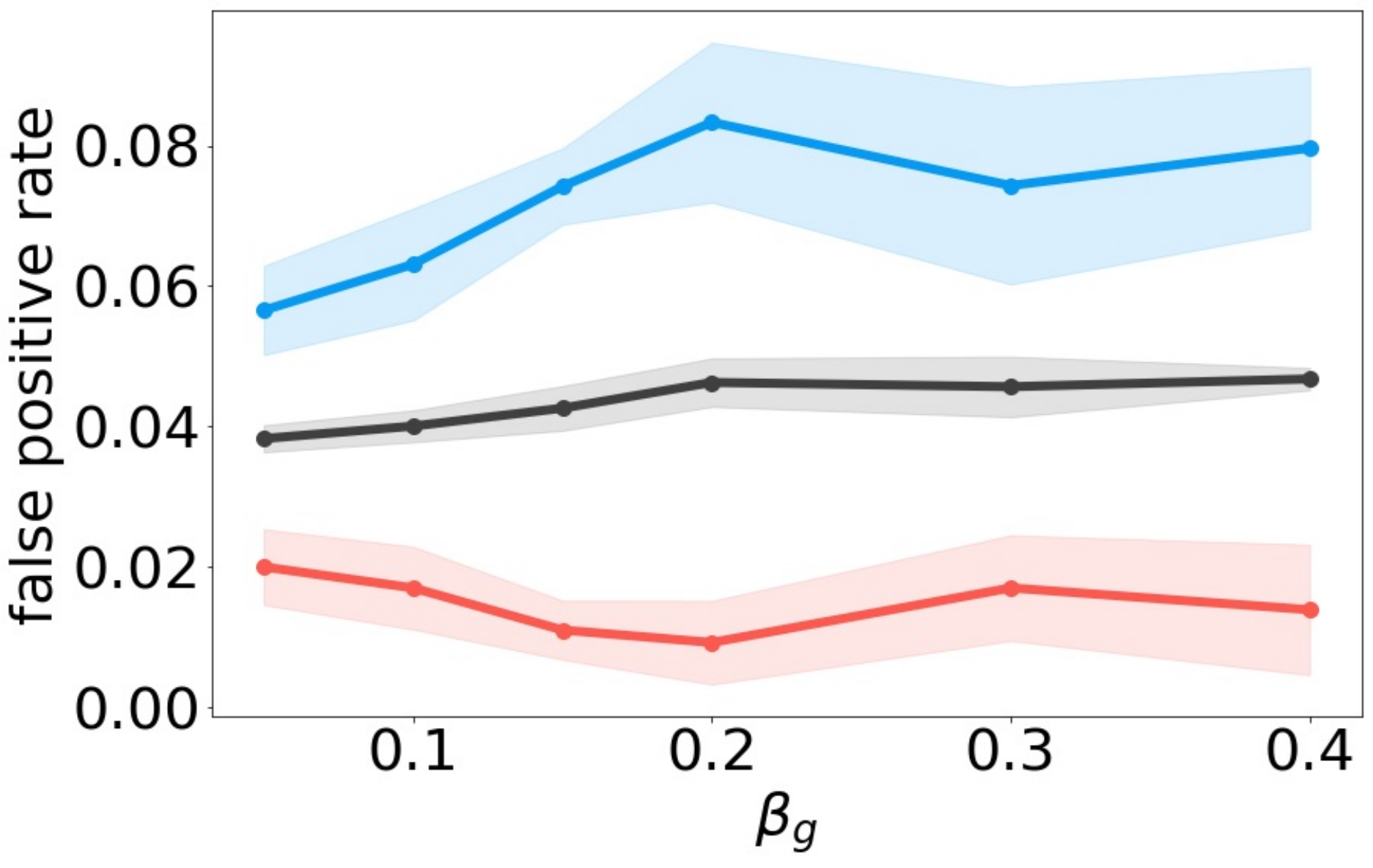}}
    &
   \hspace{-0.1in} {\includegraphics[width=0.195\textwidth]{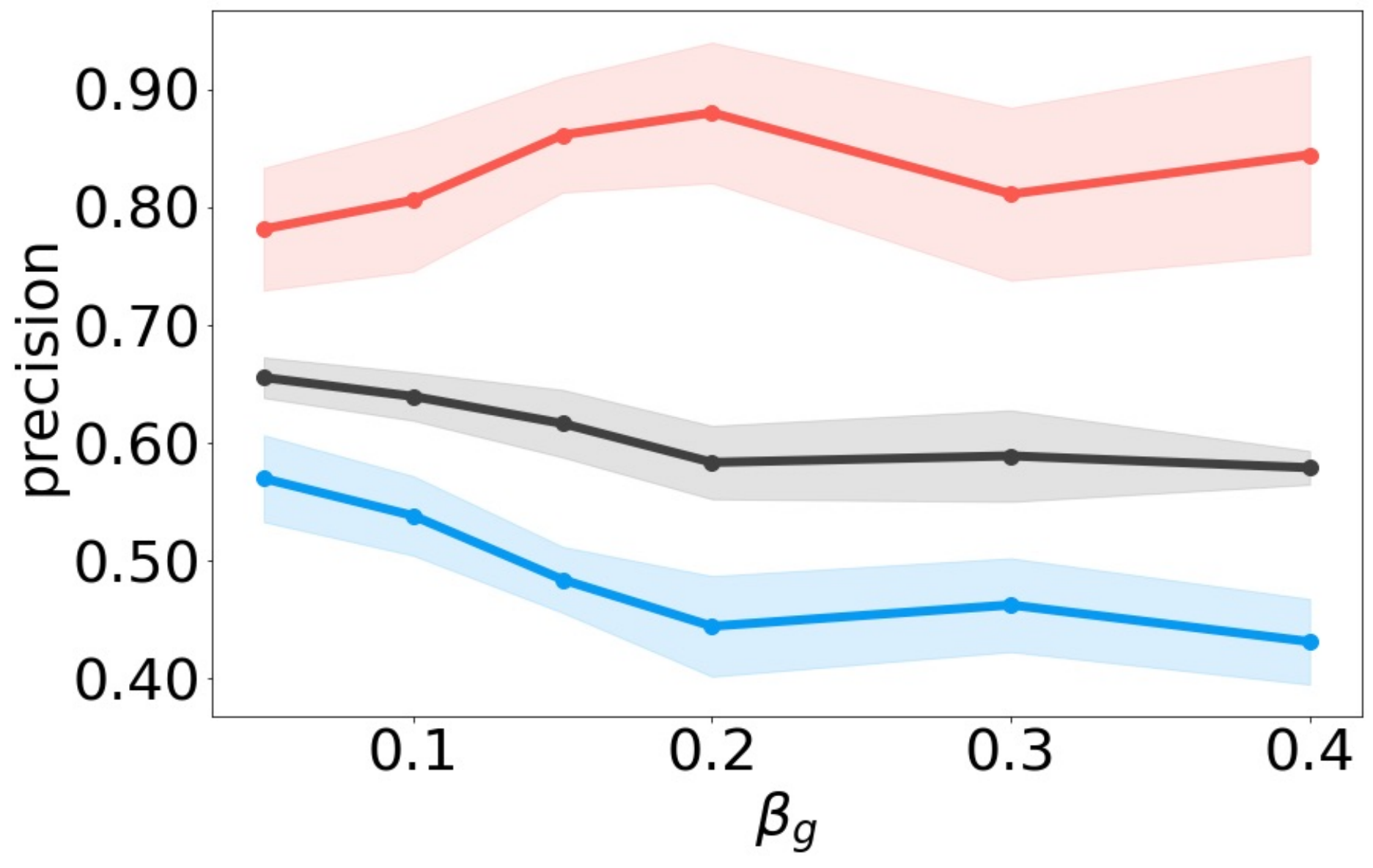}}
   &
   \hspace{-0.1in}
   {\includegraphics[width=0.195\textwidth]{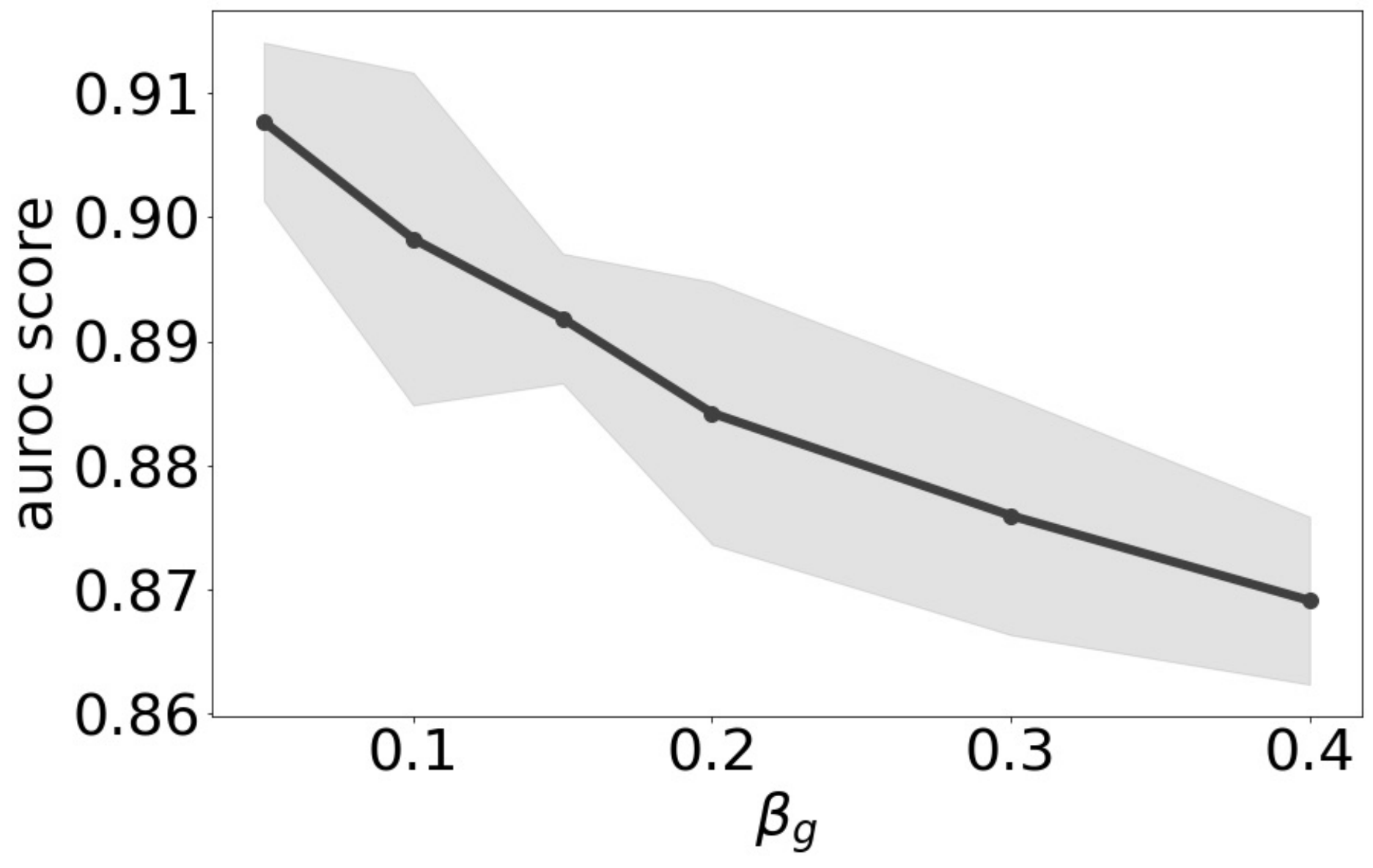}} 
\\
\hspace{-0.1in}{\includegraphics[width=0.195\textwidth]{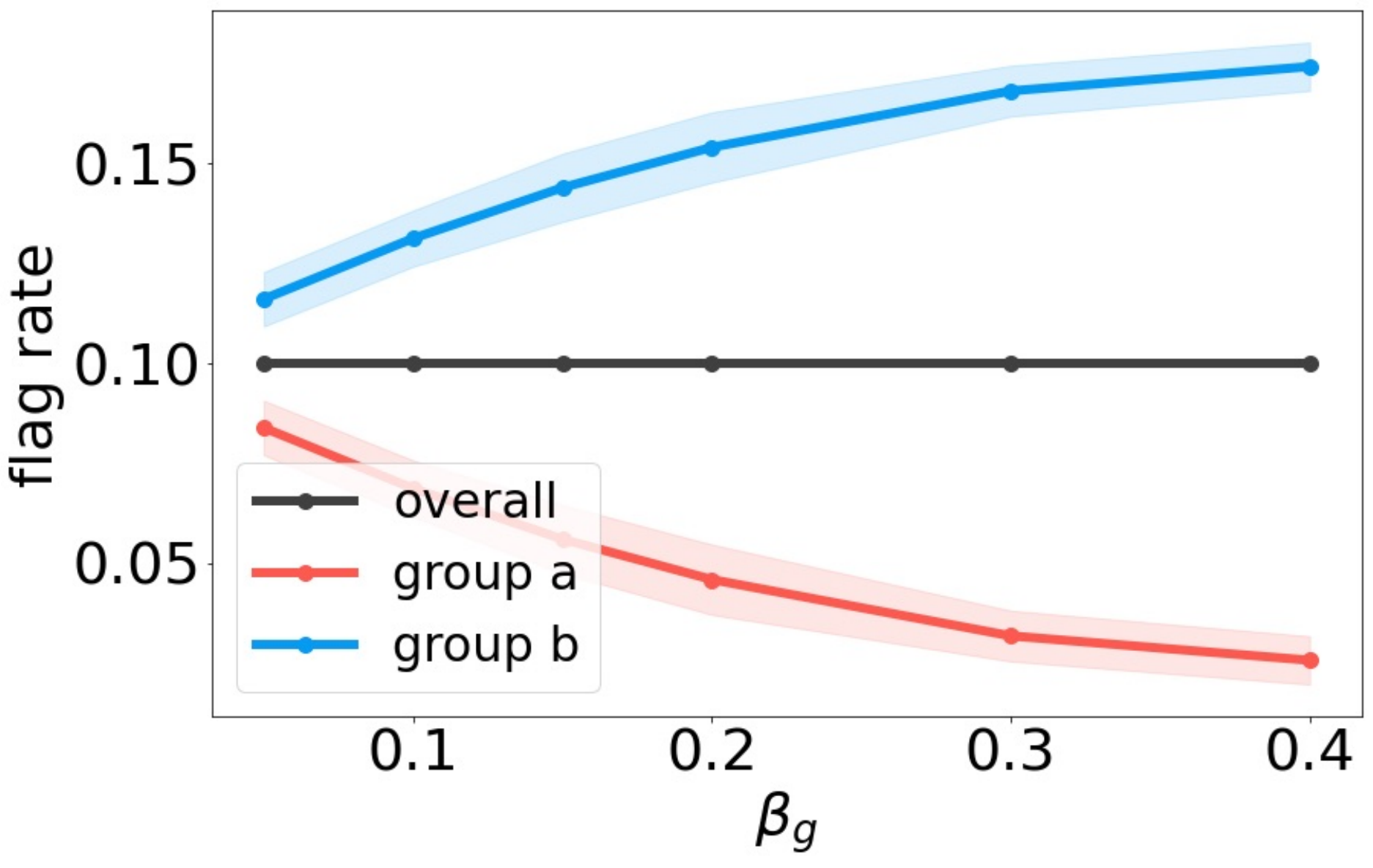}} 
    &
    \hspace{-0.1in}{\includegraphics[width=0.195\textwidth]{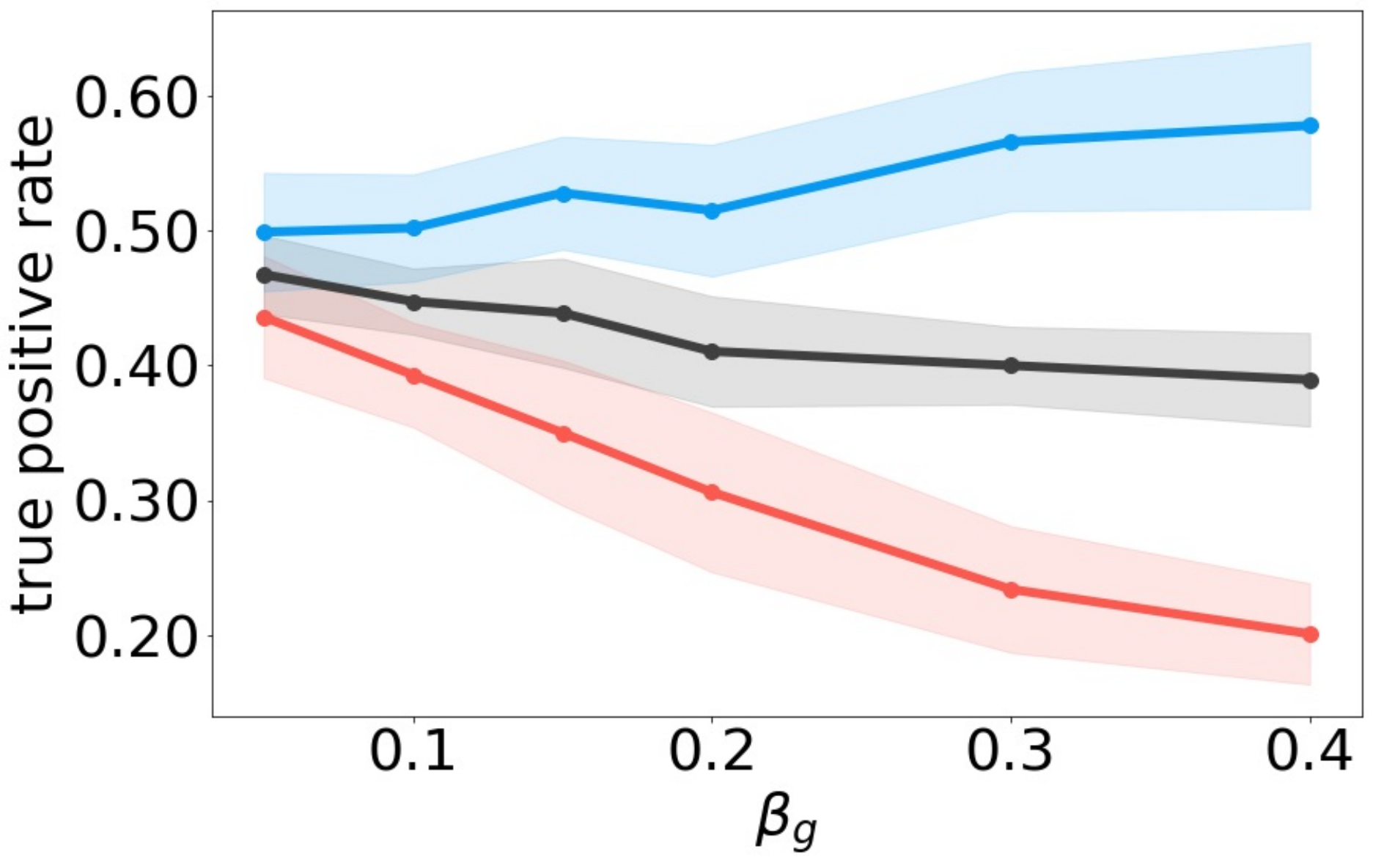}} &
     \hspace{-0.1in}{\includegraphics[width=0.195\textwidth]{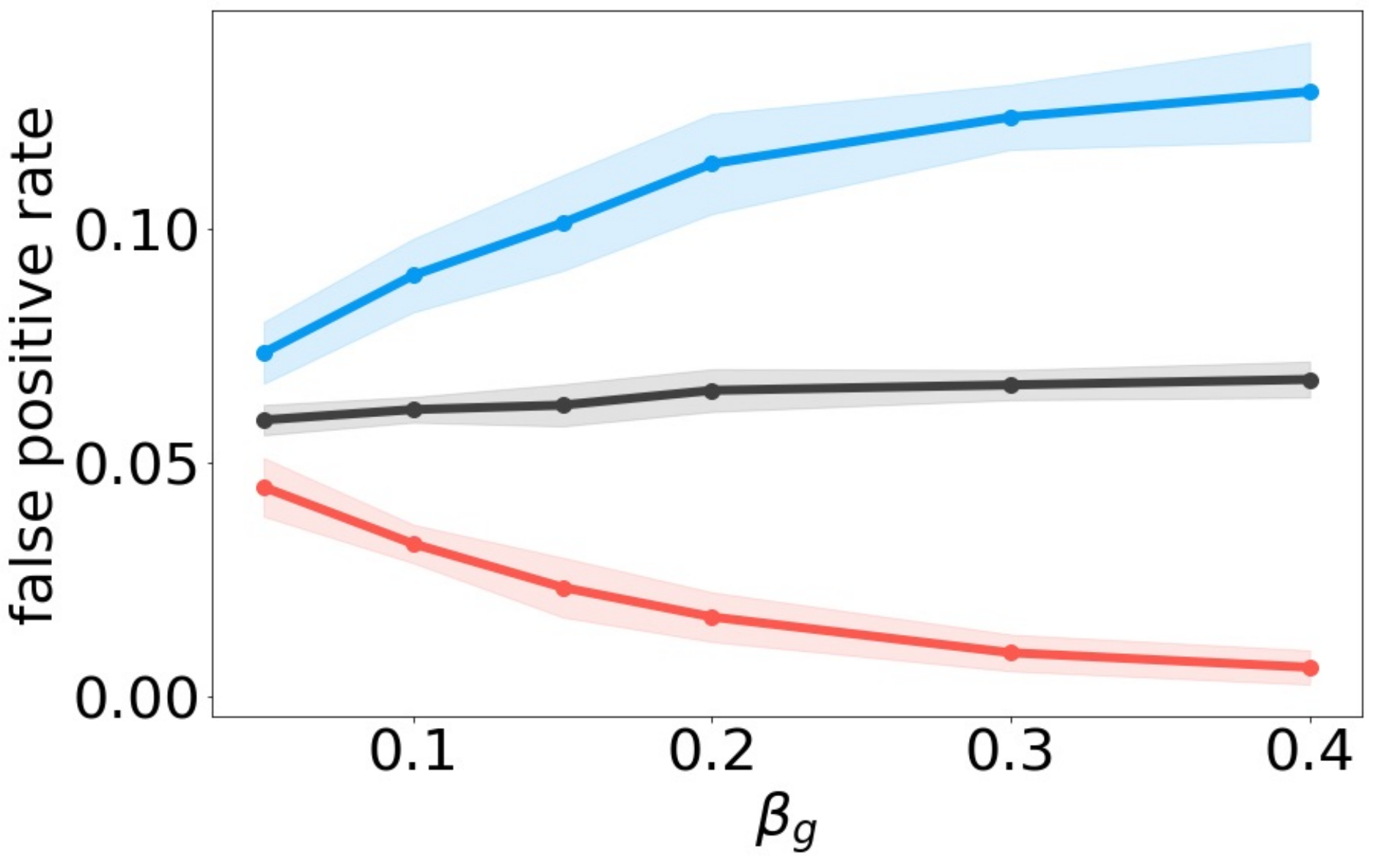}}
    &
   \hspace{-0.1in} {\includegraphics[width=0.195\textwidth]{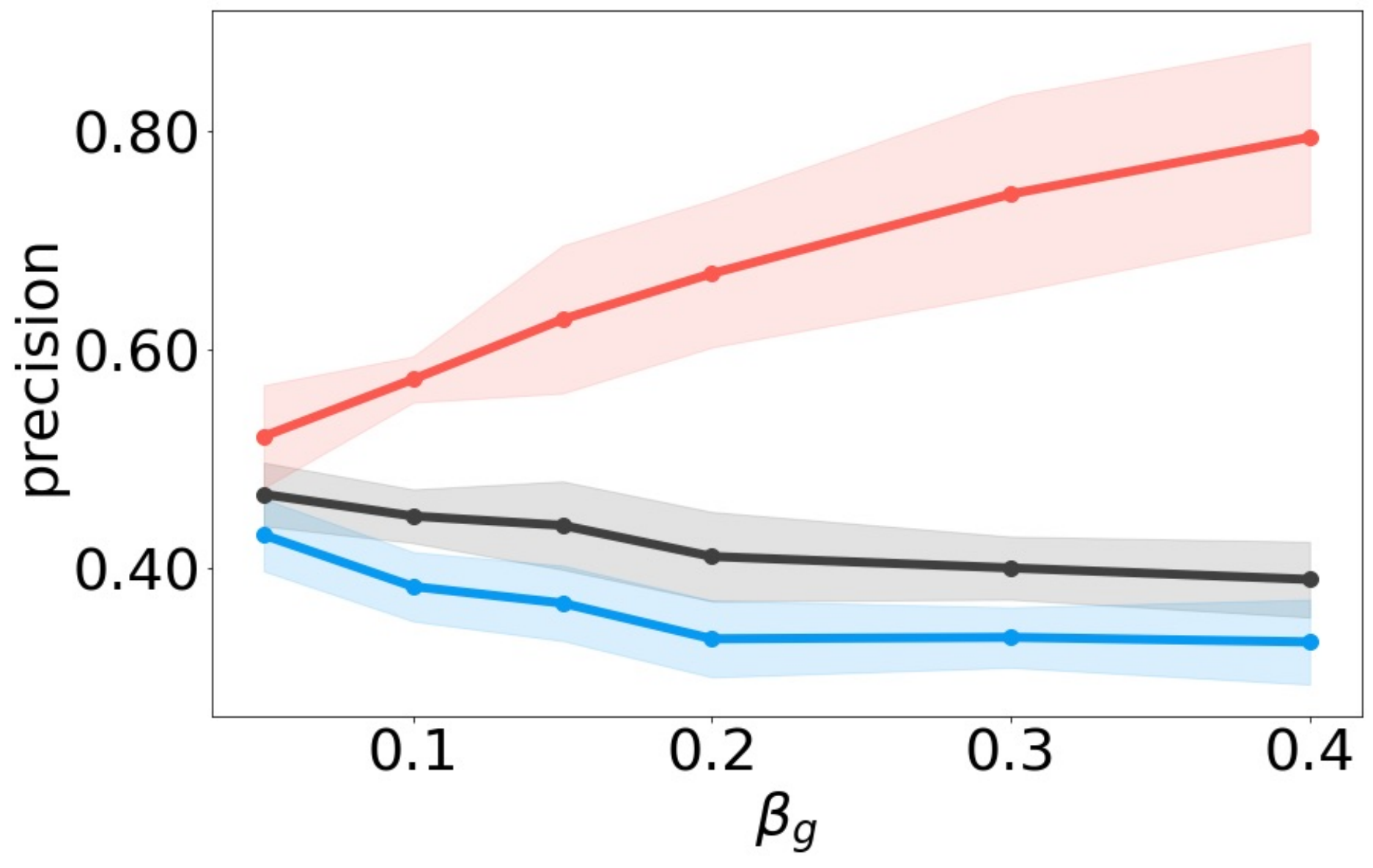}}
   &
    \hspace{-0.1in}{\includegraphics[width=0.195\textwidth]{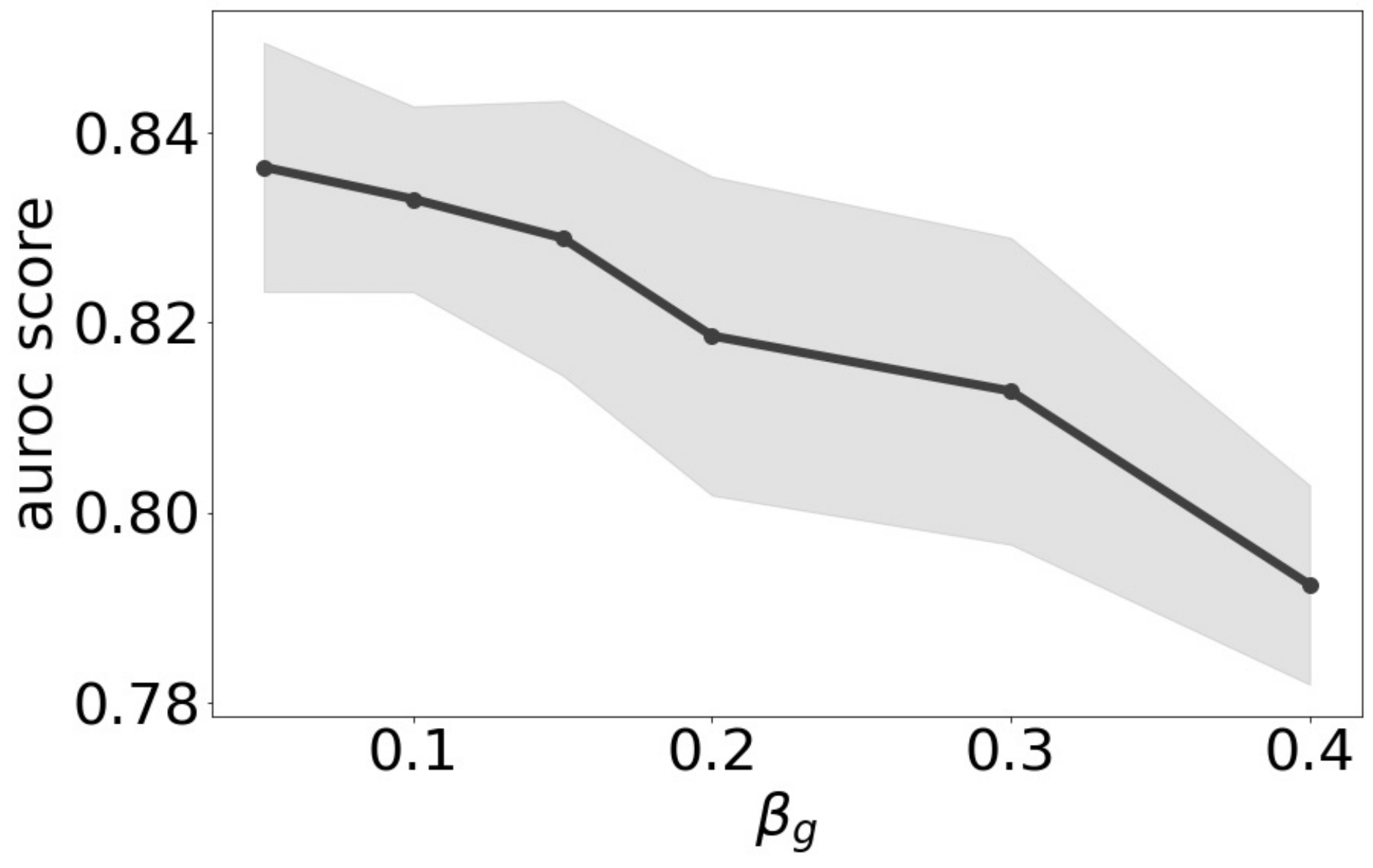}}
   \\
     \hspace{-0.1in}{\includegraphics[width=0.195\textwidth]{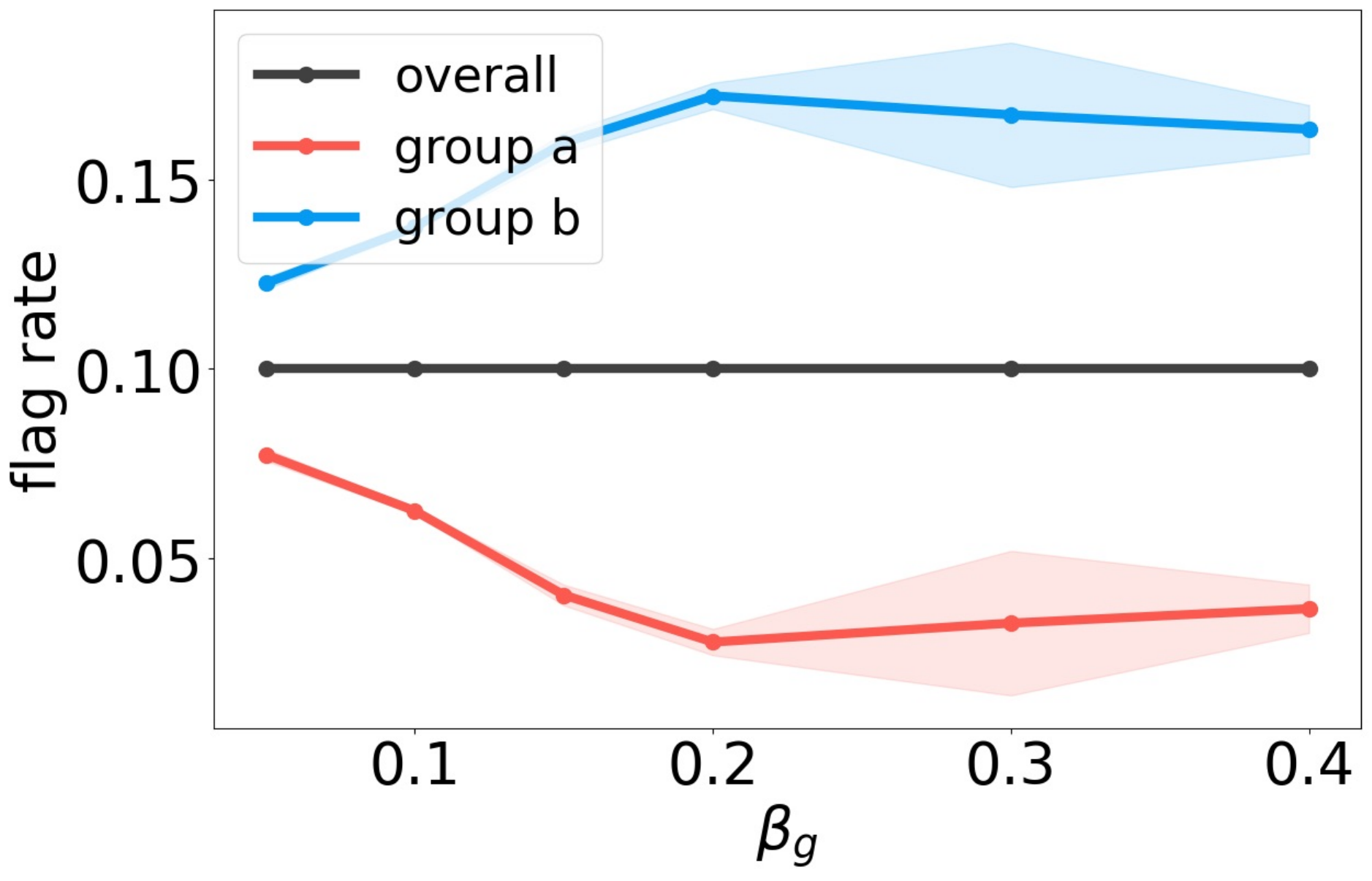}} 
    &
    \hspace{-0.1in}{\includegraphics[width=0.195\textwidth]{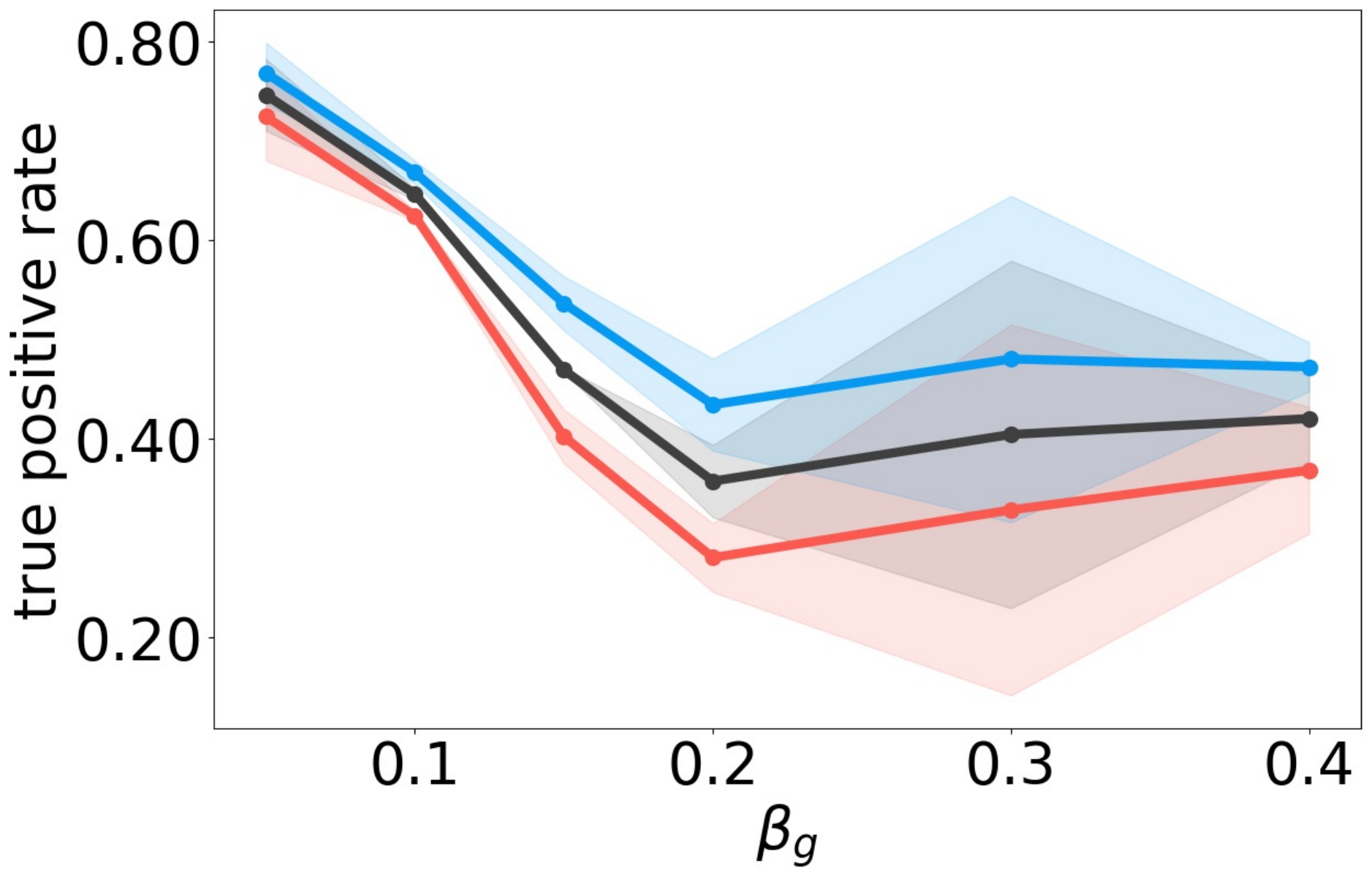}} &
     \hspace{-0.1in}{\includegraphics[width=0.195\textwidth]{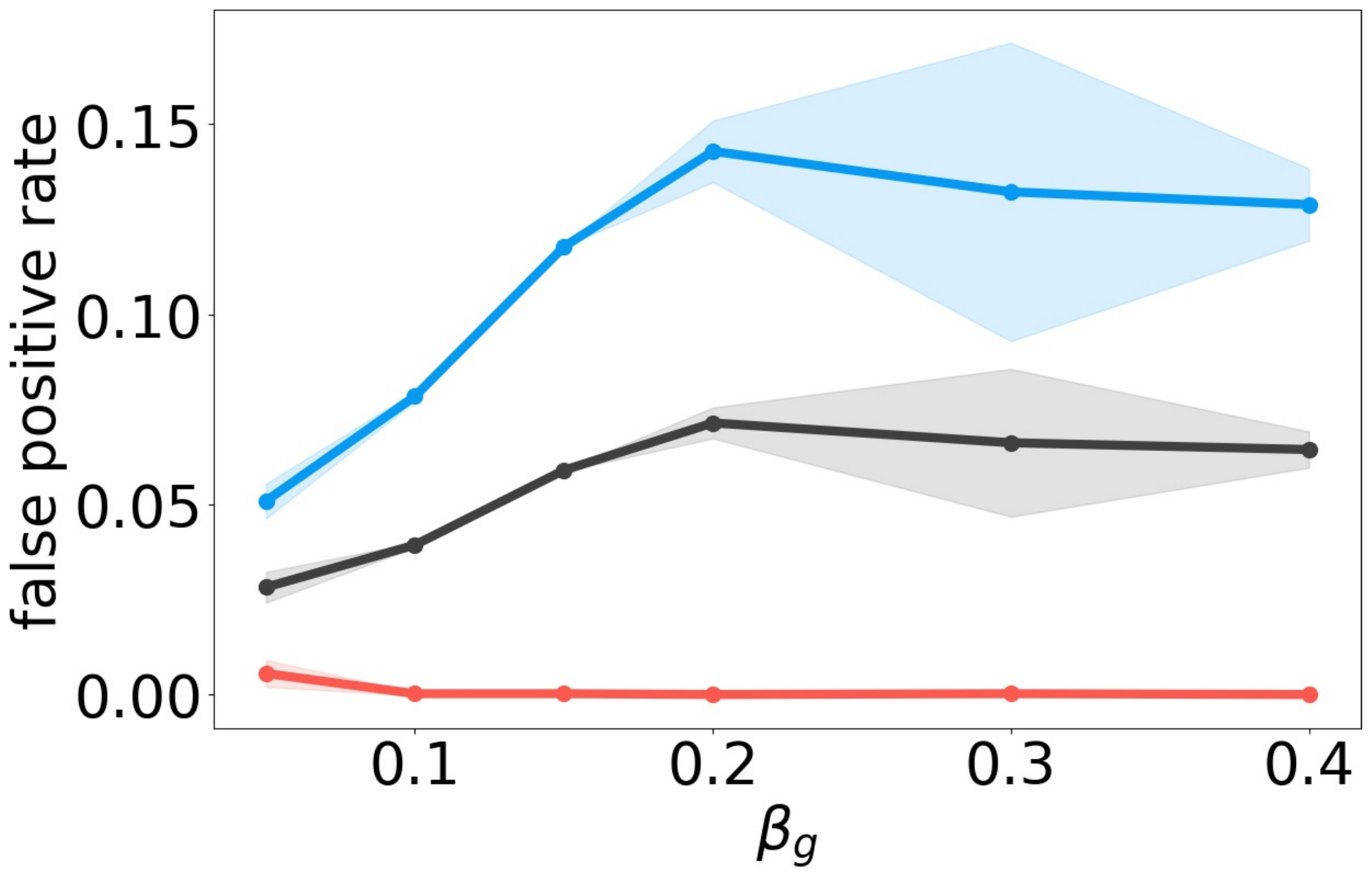}}
    &
   \hspace{-0.1in} {\includegraphics[width=0.195\textwidth]{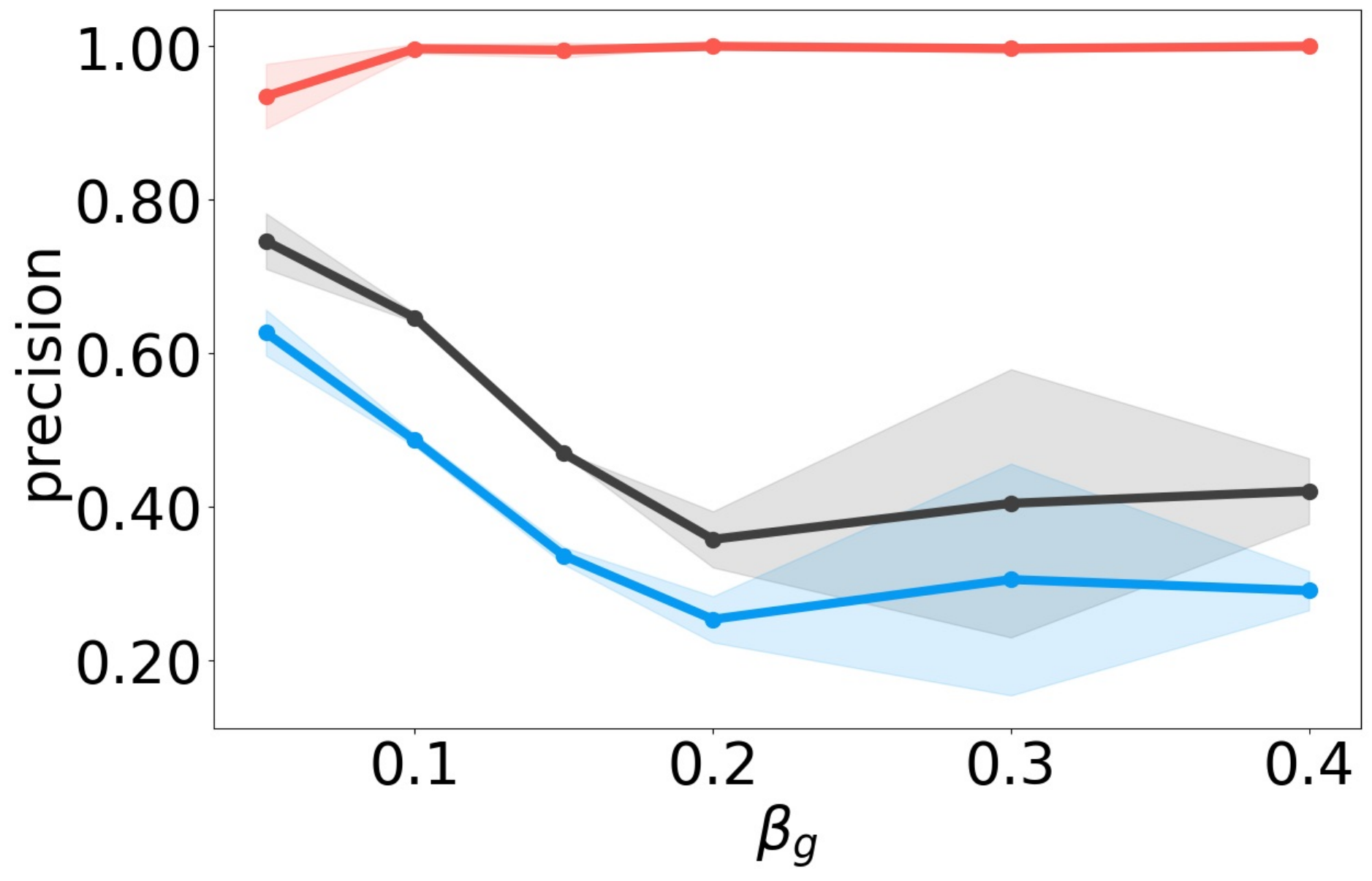}}
   &
     \hspace{-0.1in}{\includegraphics[width=0.195\textwidth]{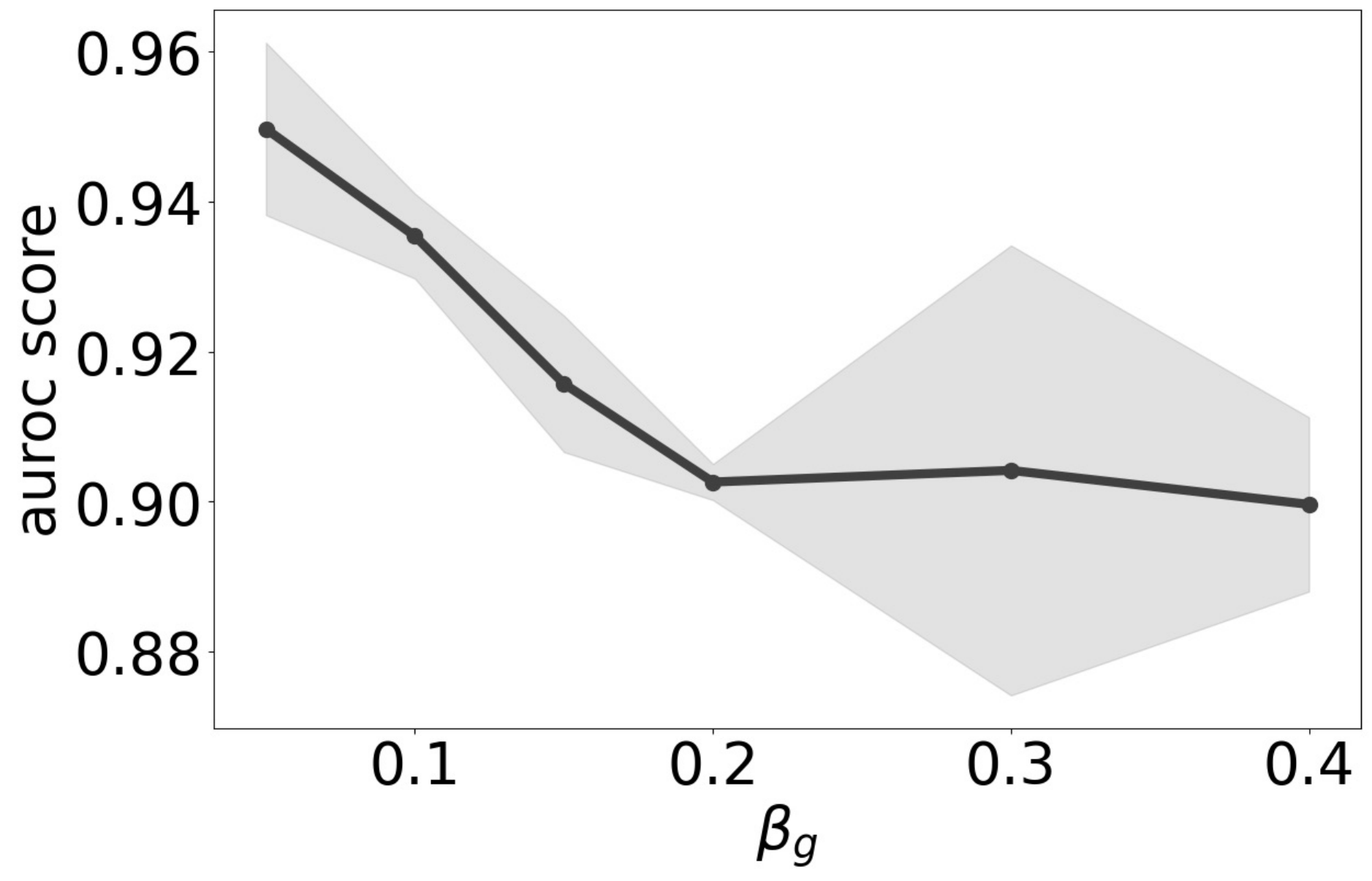}}
   \\
   \hspace{-0.1in}{\includegraphics[width=0.195\textwidth]{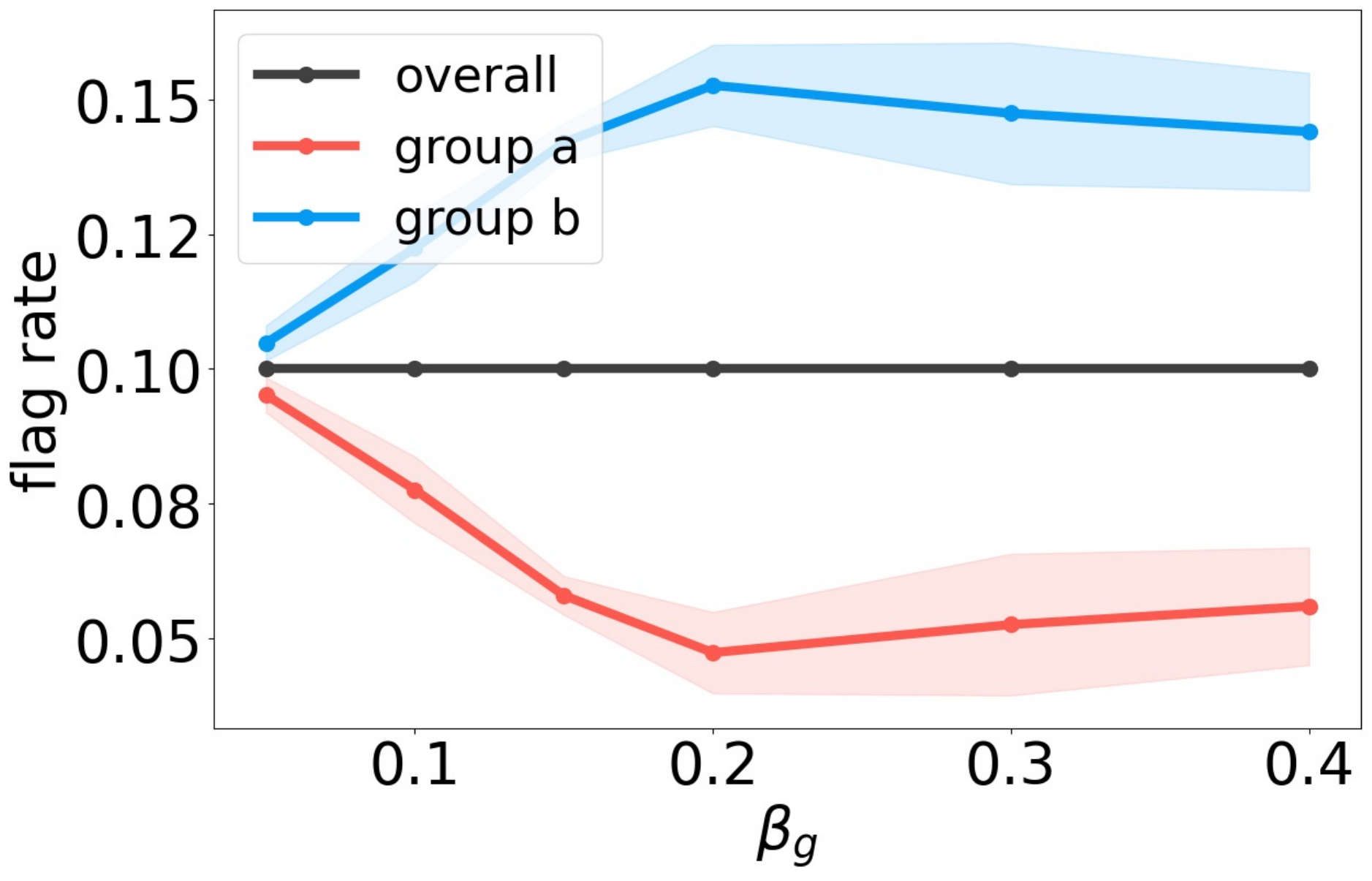} }
    &
    \hspace{-0.1in}{\includegraphics[width=0.195\textwidth]{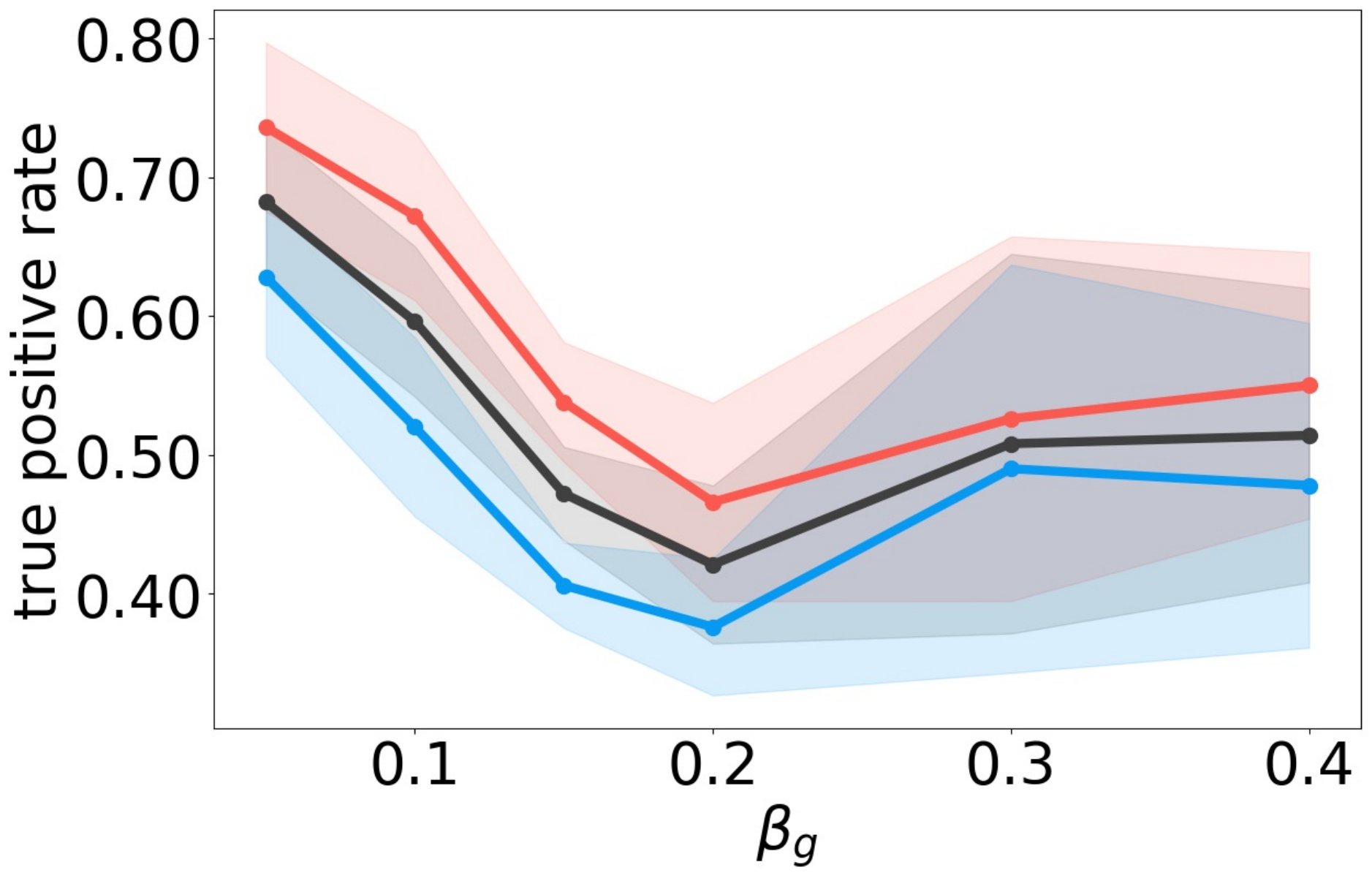}} &
     \hspace{-0.1in}{\includegraphics[width=0.195\textwidth]{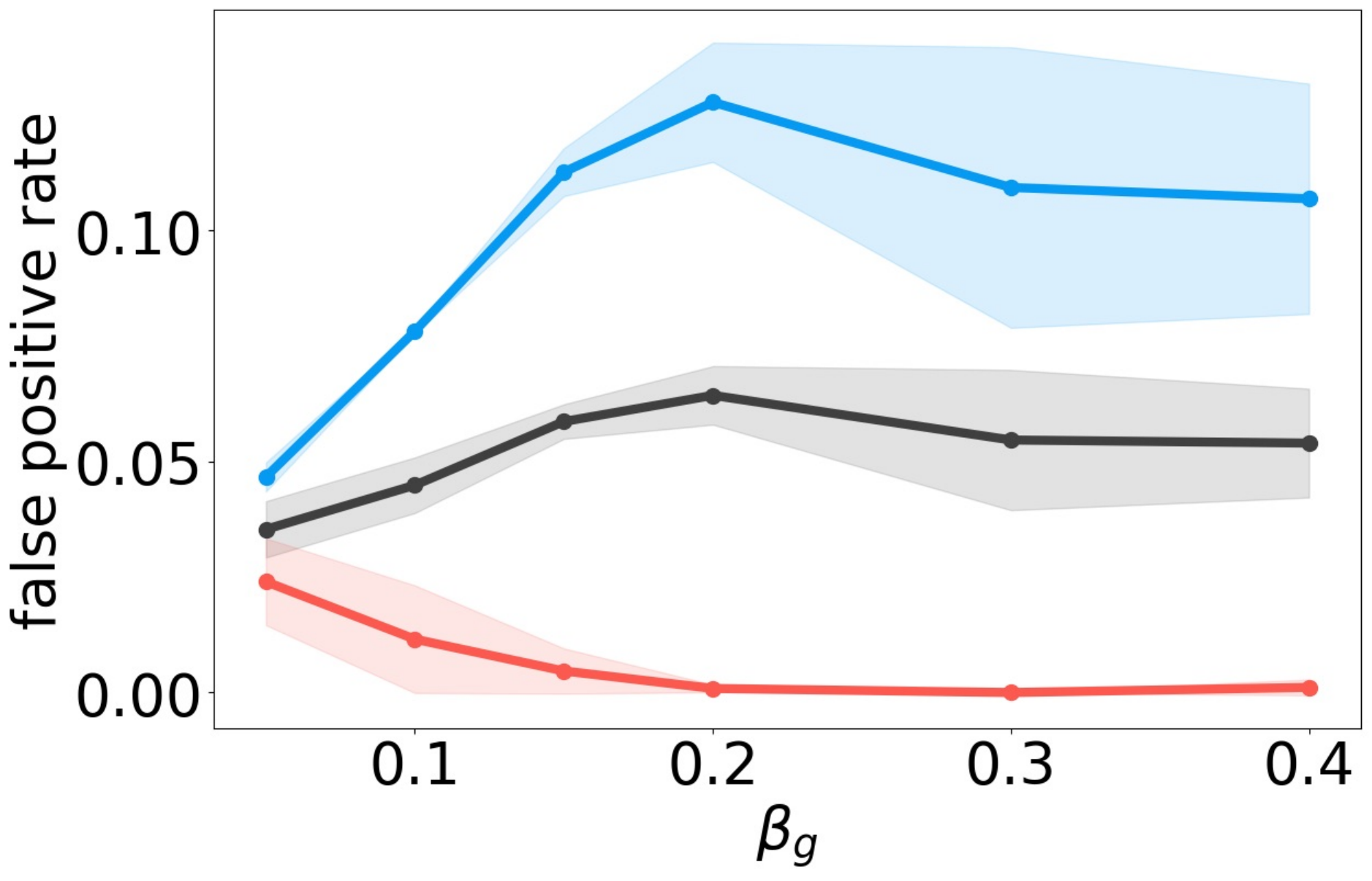}}
    &
   \hspace{-0.1in} {\includegraphics[width=0.195\textwidth]{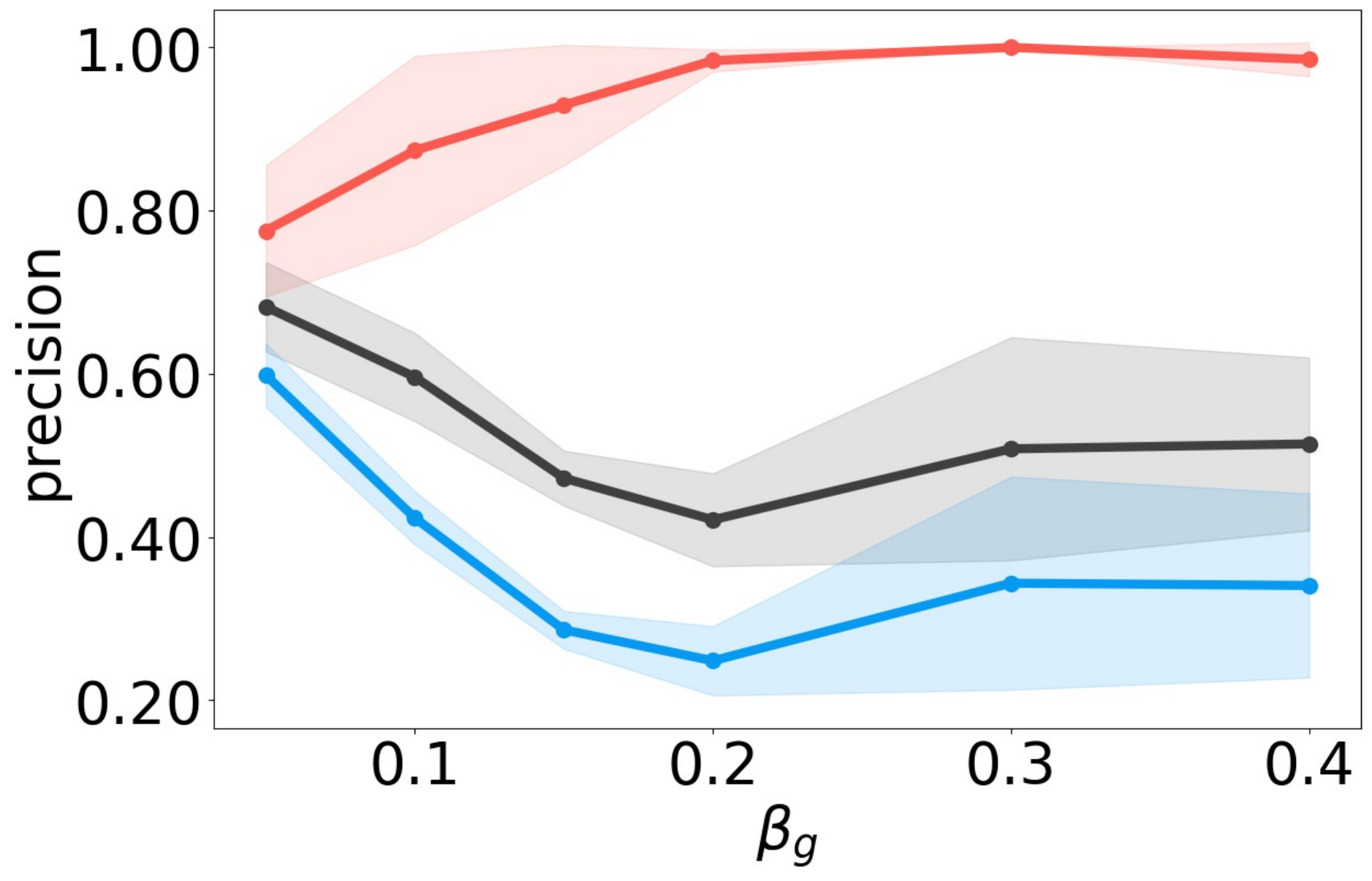}}
   &
   \hspace{-0.1in} {\includegraphics[width=0.195\textwidth]{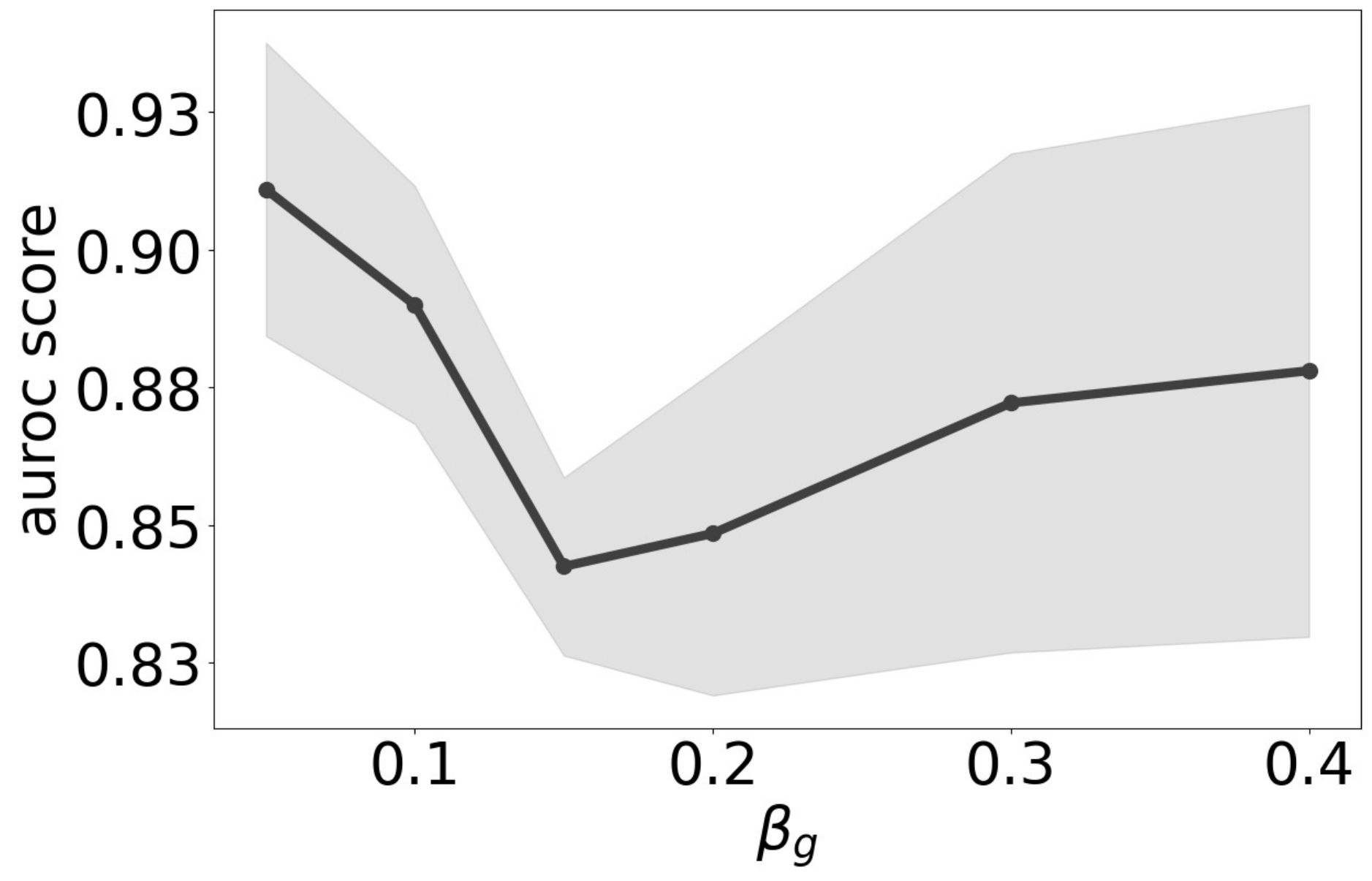}}
    \end{tabular}
    
    \caption{(best in color) Group-wise fairness metrics and AUROC for (top to bottom) LOF, iForest, DeepAE and FairOD under \underline{membership obfuscation bias} on \textit{scattered} outliers. }
    \label{fig:obfuscation_all_scatter}
\end{figure*}

See Fig. \ref{fig:obfuscation_all_scatter}.

\subsubsection{Unequal base rates}
\label{assec:ubs_scatter}

\begin{figure*}[!ht]
    \centering
    \begin{tabular}{lllll}
    \hspace{-0.1in}{\includegraphics[width=0.195\textwidth]{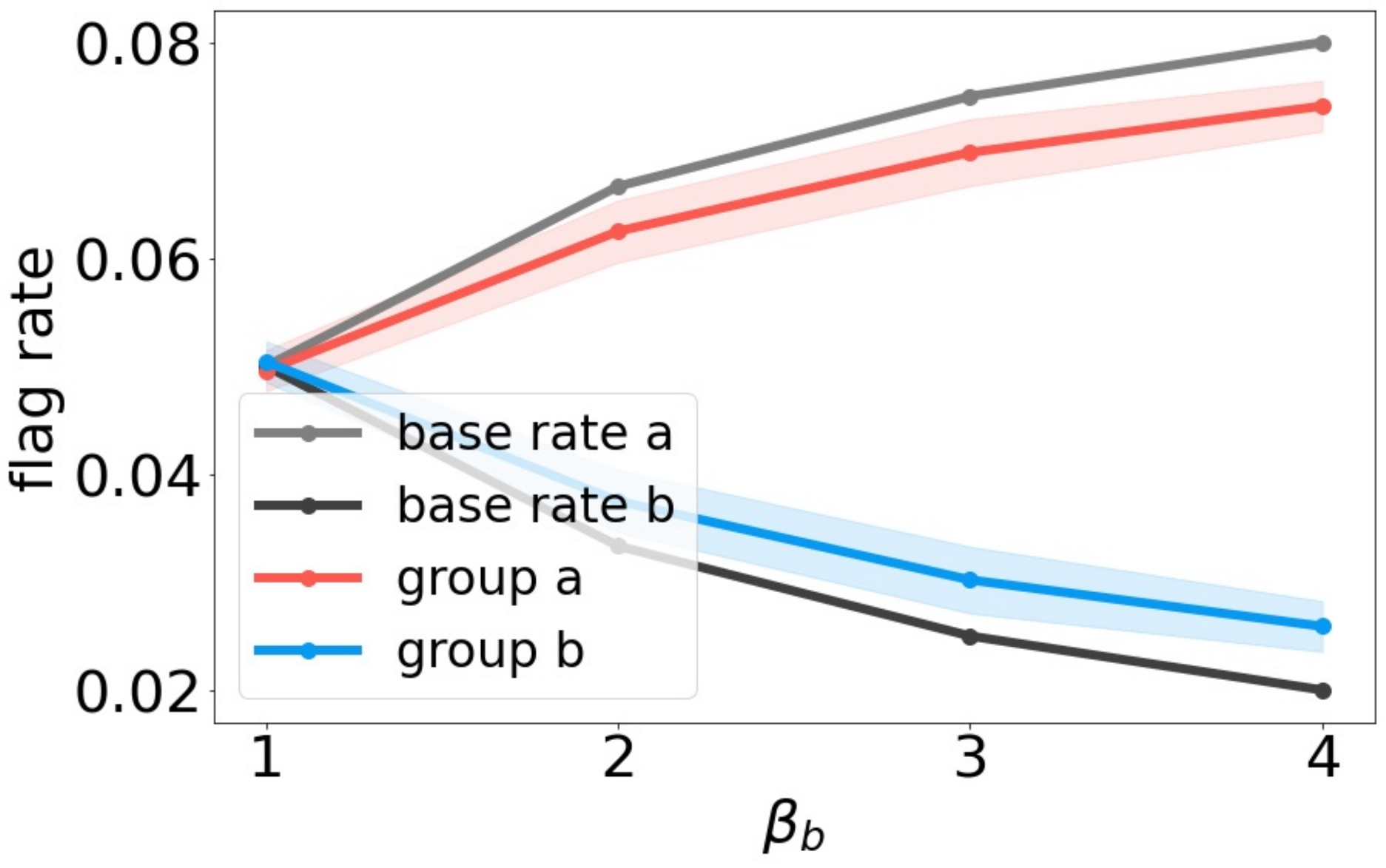}} 
    &
    \hspace{-0.1in}{\includegraphics[width=0.195\textwidth]{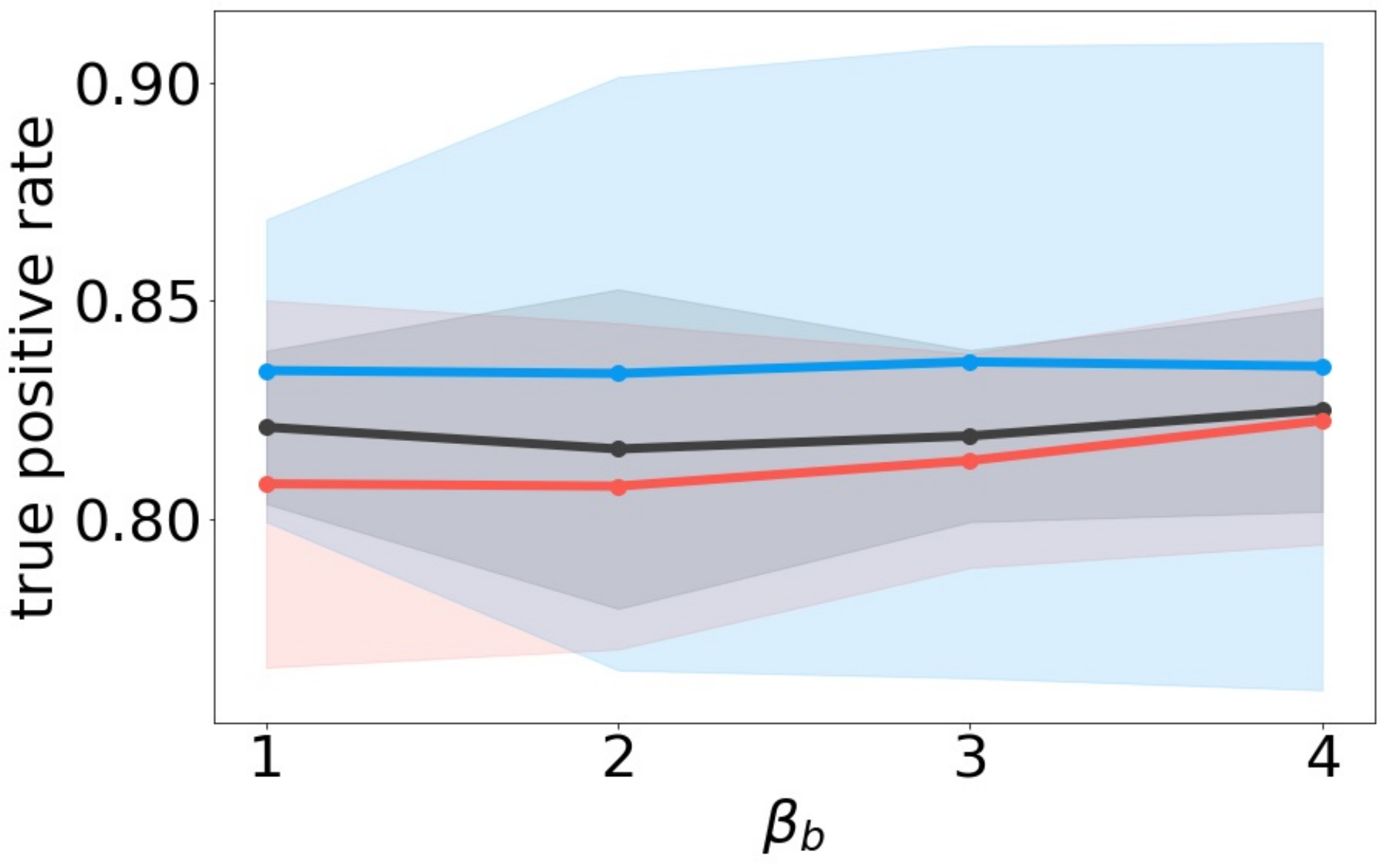}} &
     \hspace{-0.1in}{\includegraphics[width=0.195\textwidth]{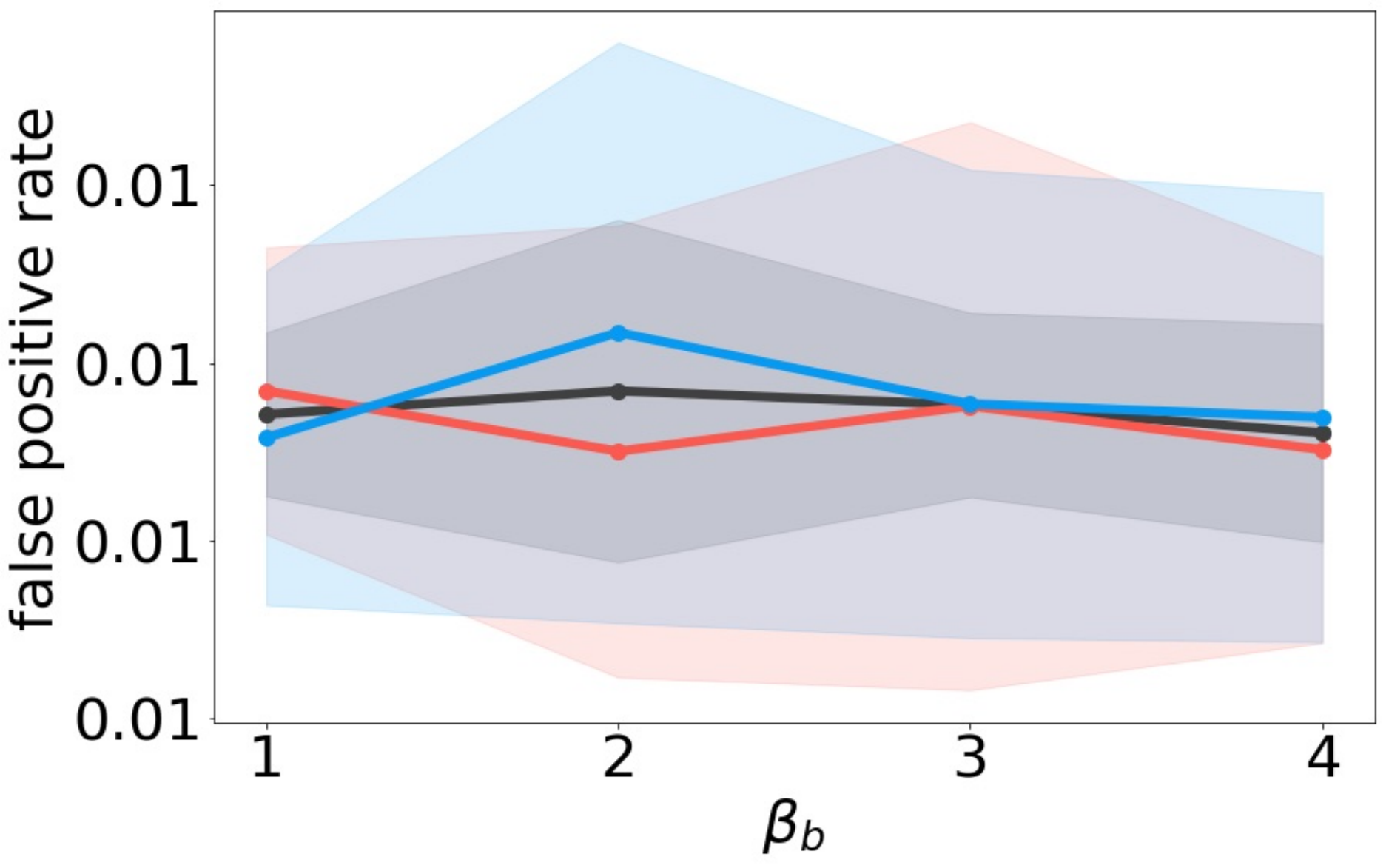}}
    &
   \hspace{-0.1in} {\includegraphics[width=0.195\textwidth]{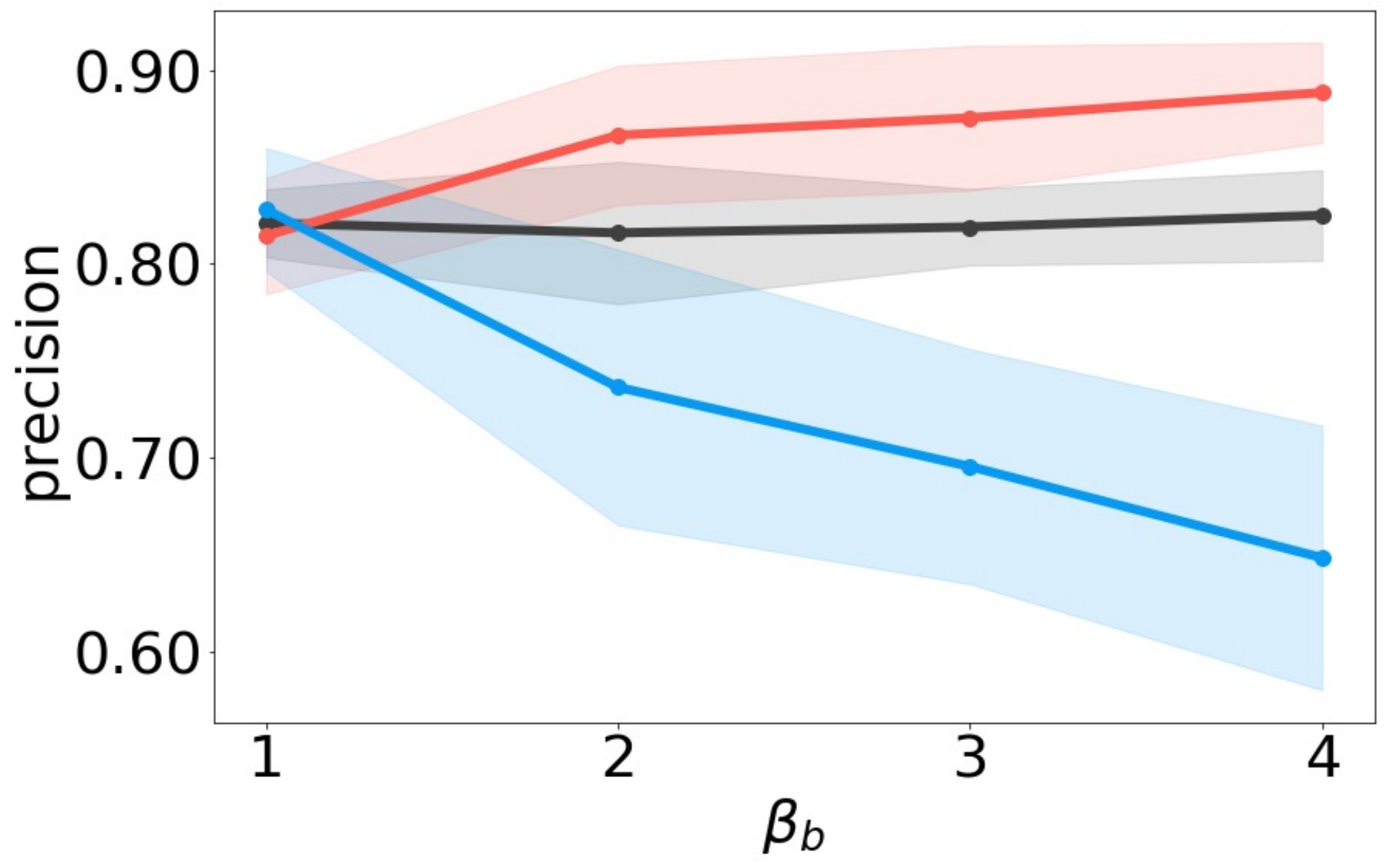}}
   &
   \hspace{-0.1in}
   {\includegraphics[width=0.195\textwidth]{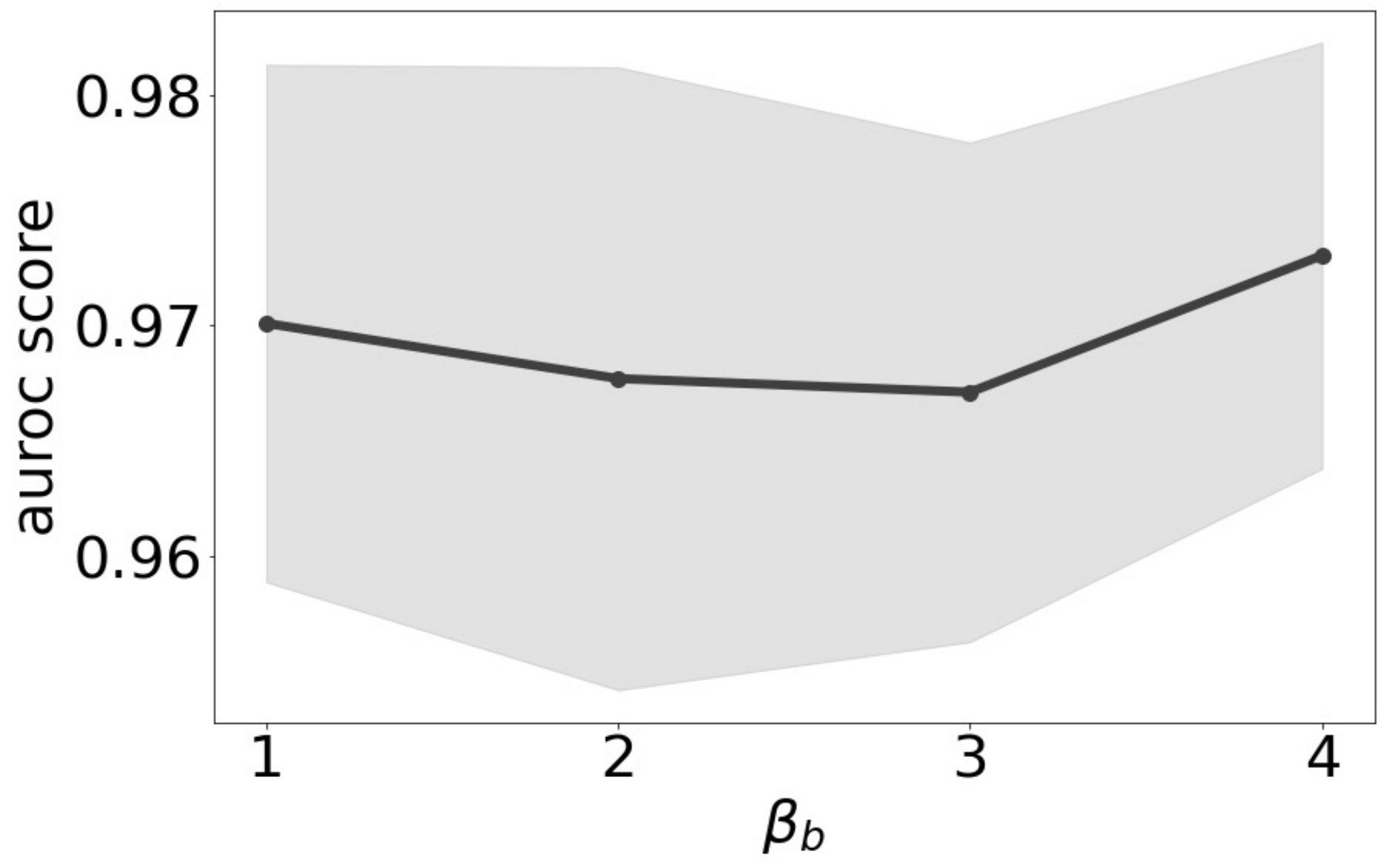}} 
\\
\hspace{-0.1in}{\includegraphics[width=0.195\textwidth]{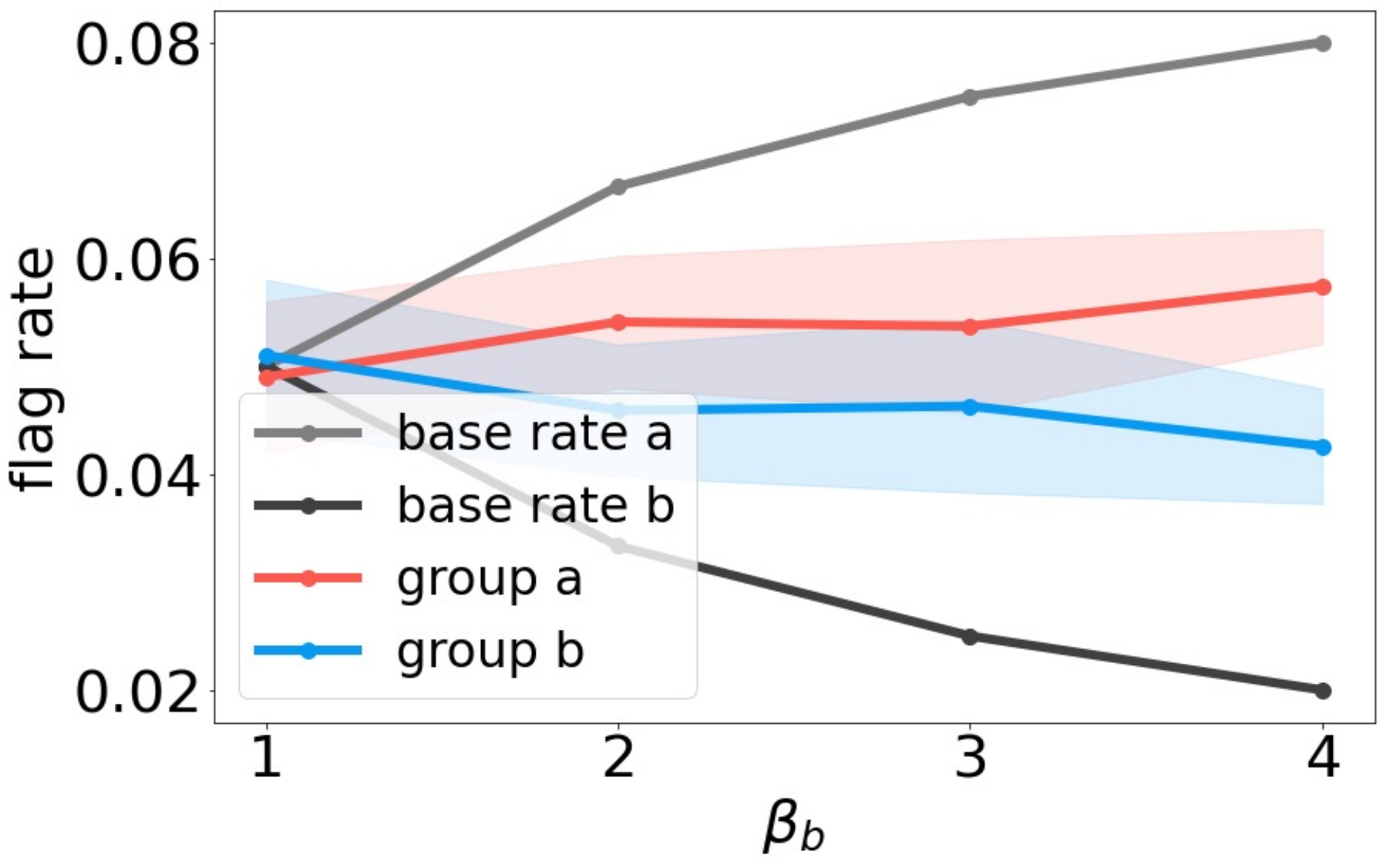}} 
    &
    \hspace{-0.1in}{\includegraphics[width=0.195\textwidth]{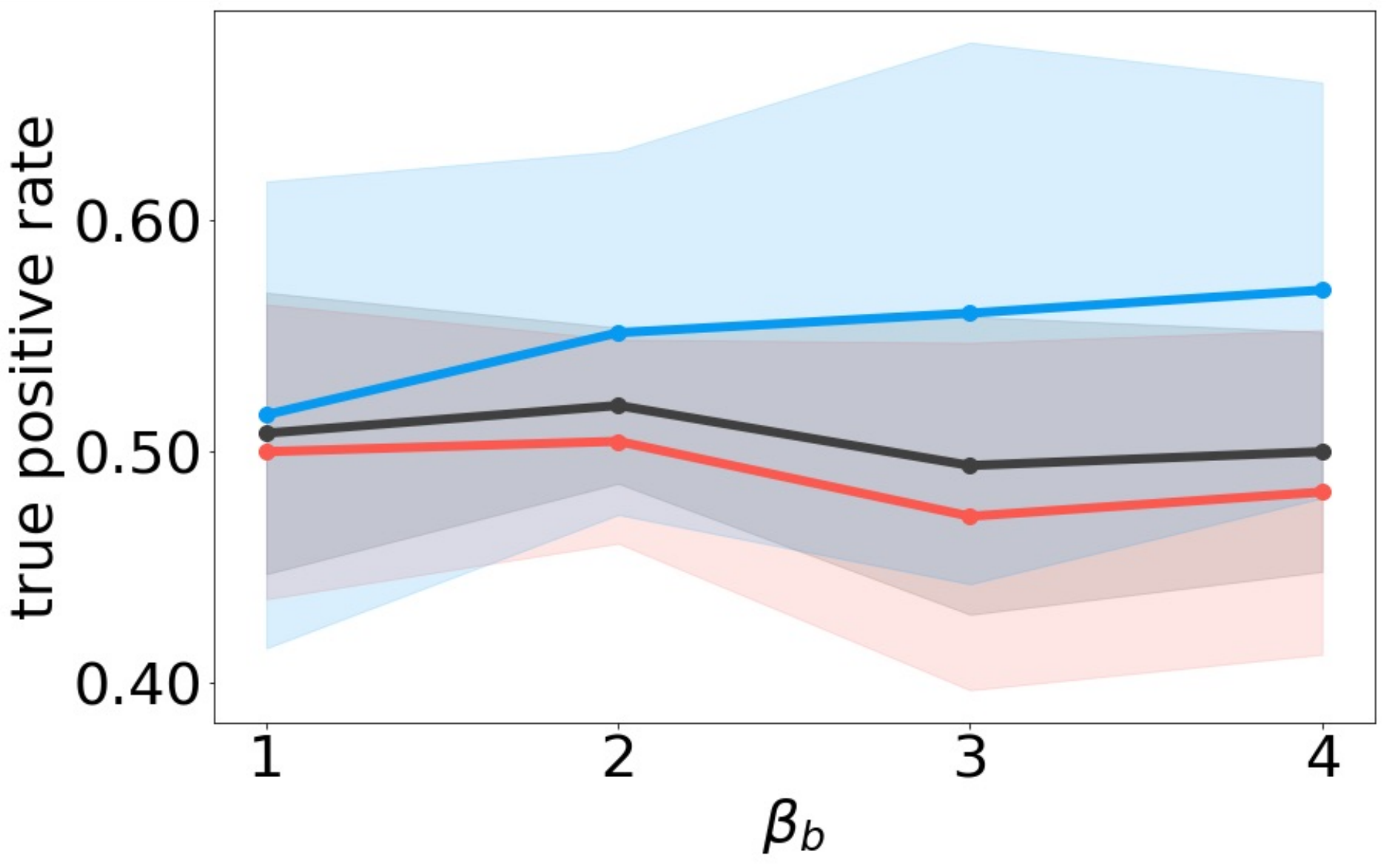}} &
     \hspace{-0.1in}{\includegraphics[width=0.195\textwidth]{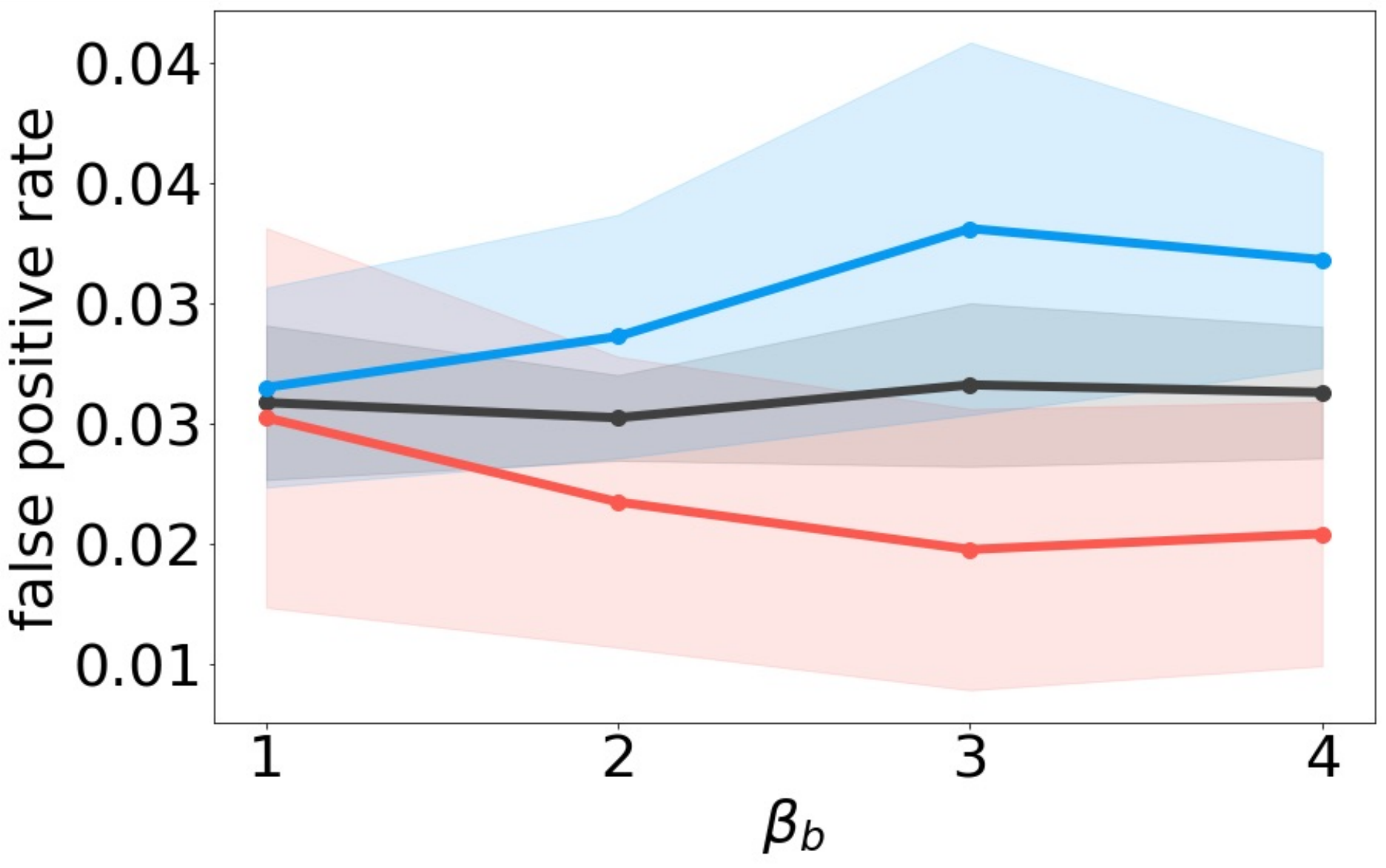}}
    &
   \hspace{-0.1in} {\includegraphics[width=0.195\textwidth]{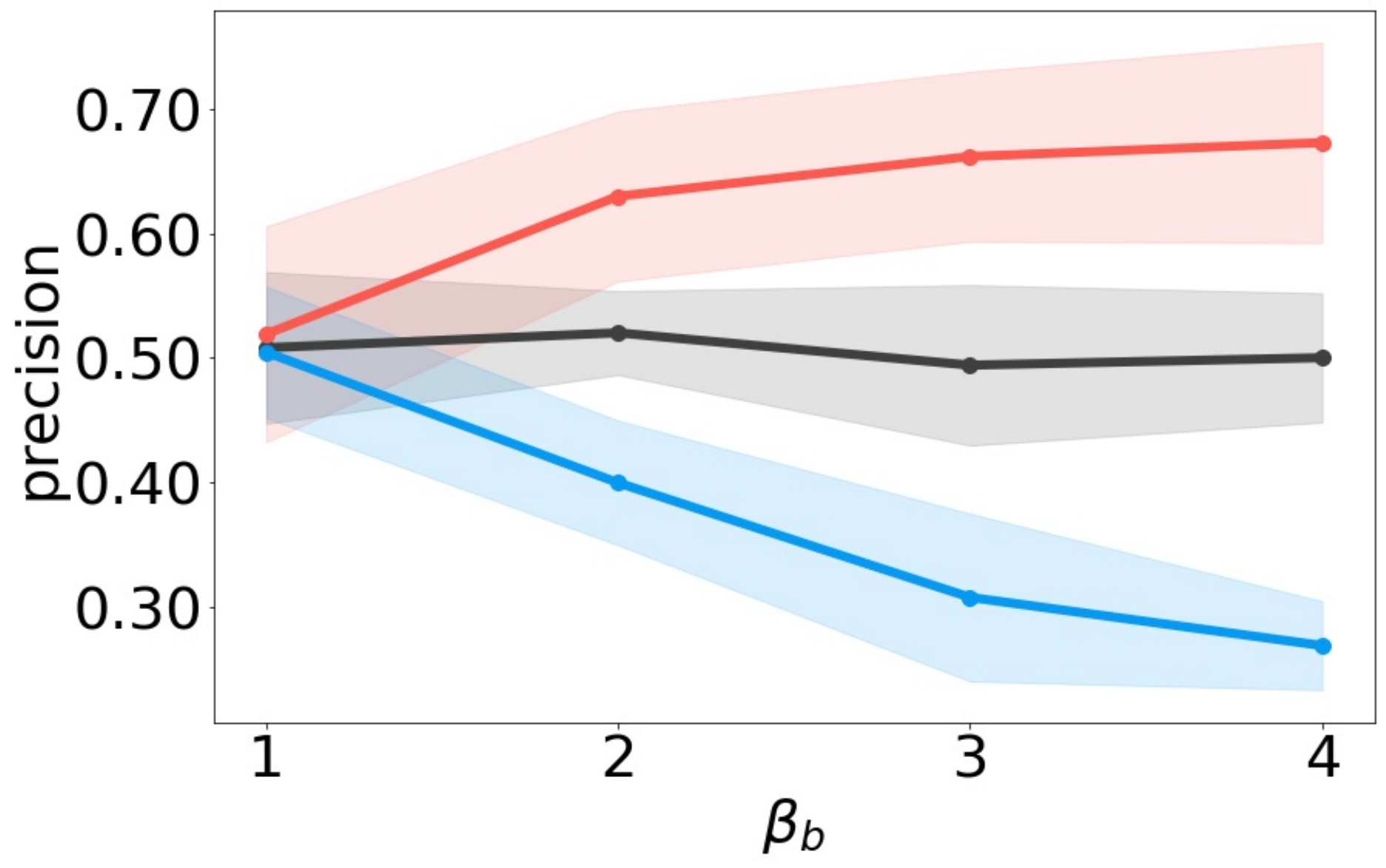}}
   &
    \hspace{-0.1in}{\includegraphics[width=0.195\textwidth]{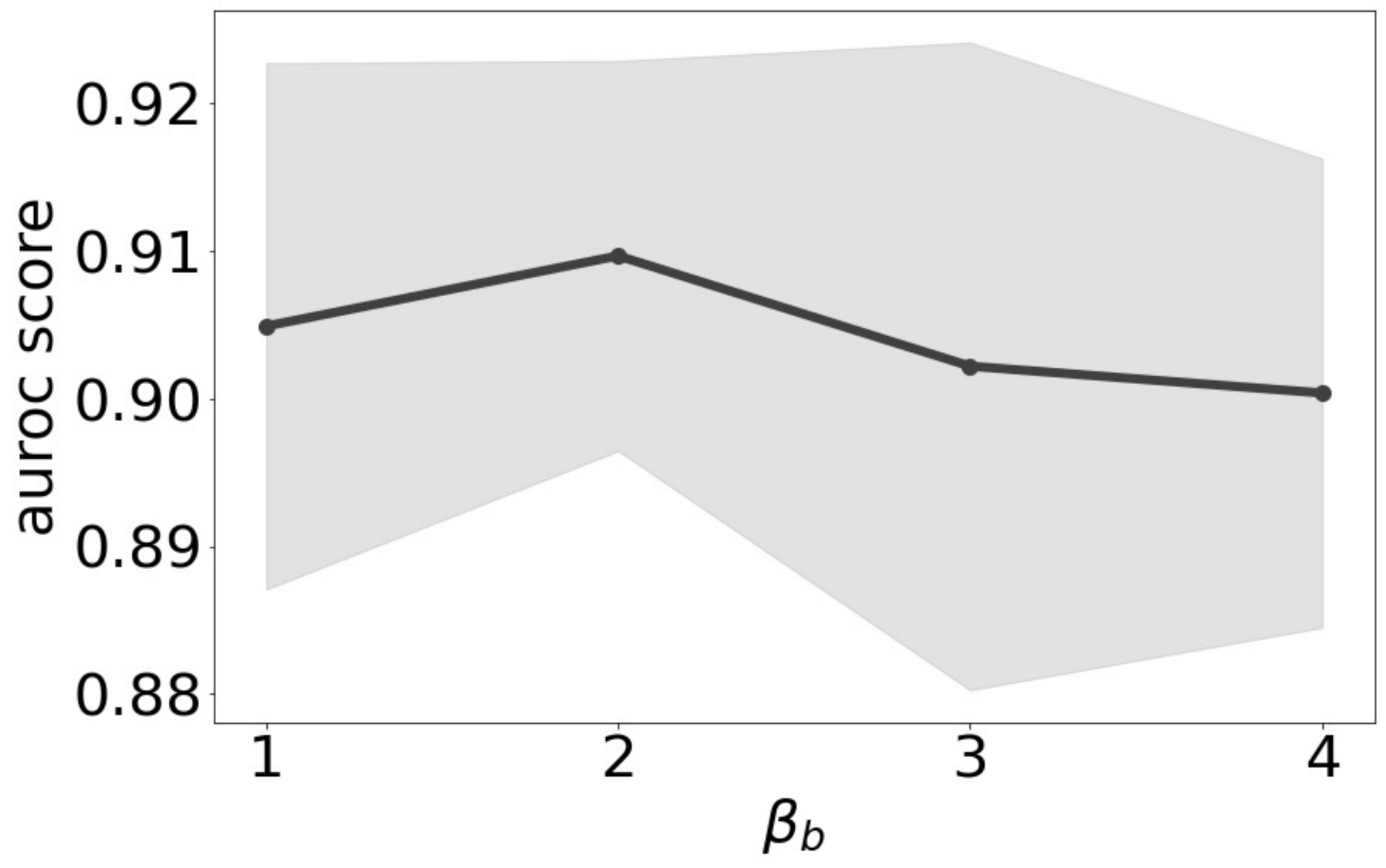}}
   \\
     \hspace{-0.1in}{\includegraphics[width=0.195\textwidth]{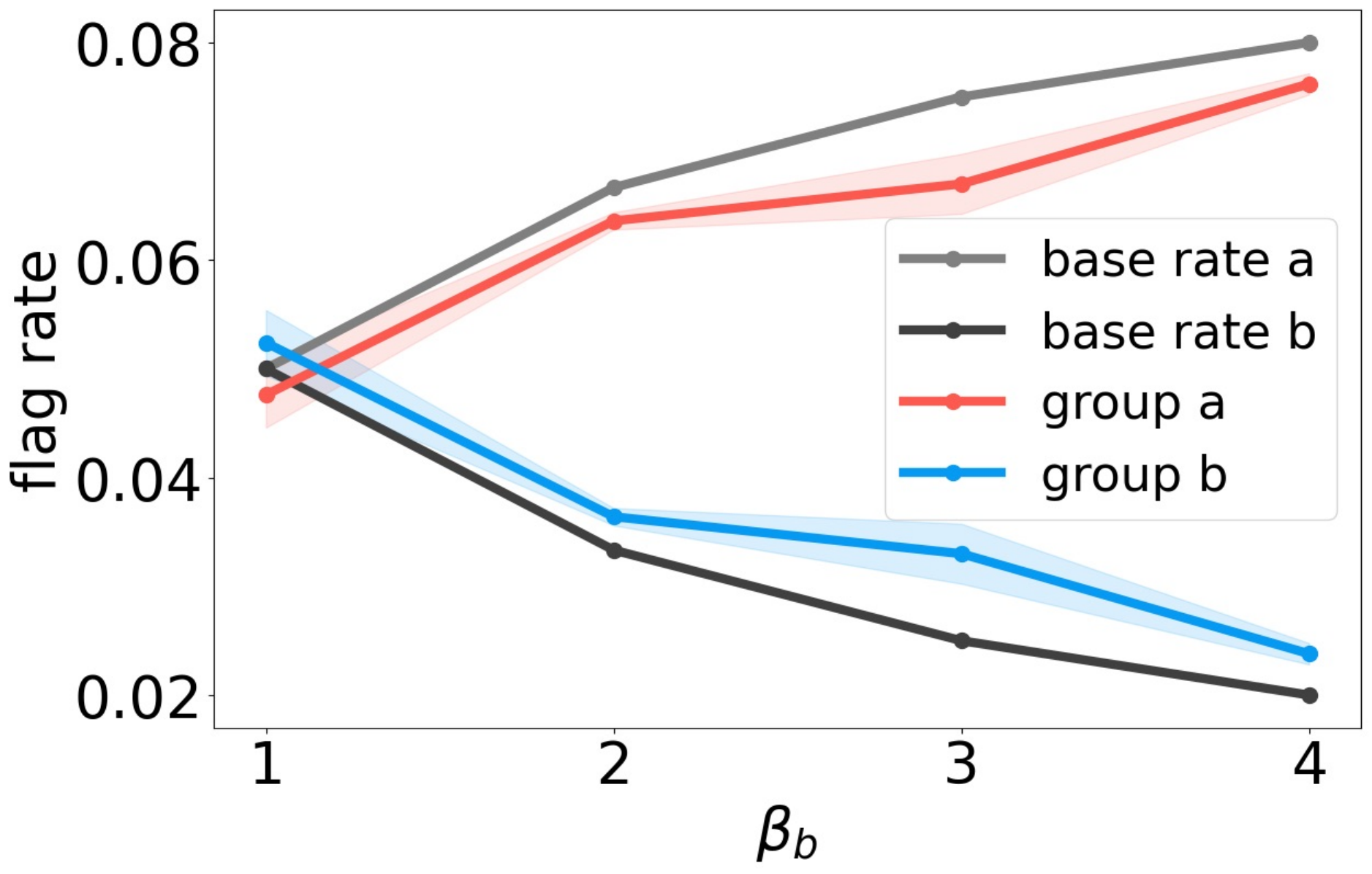}} 
    &
    \hspace{-0.1in}{\includegraphics[width=0.195\textwidth]{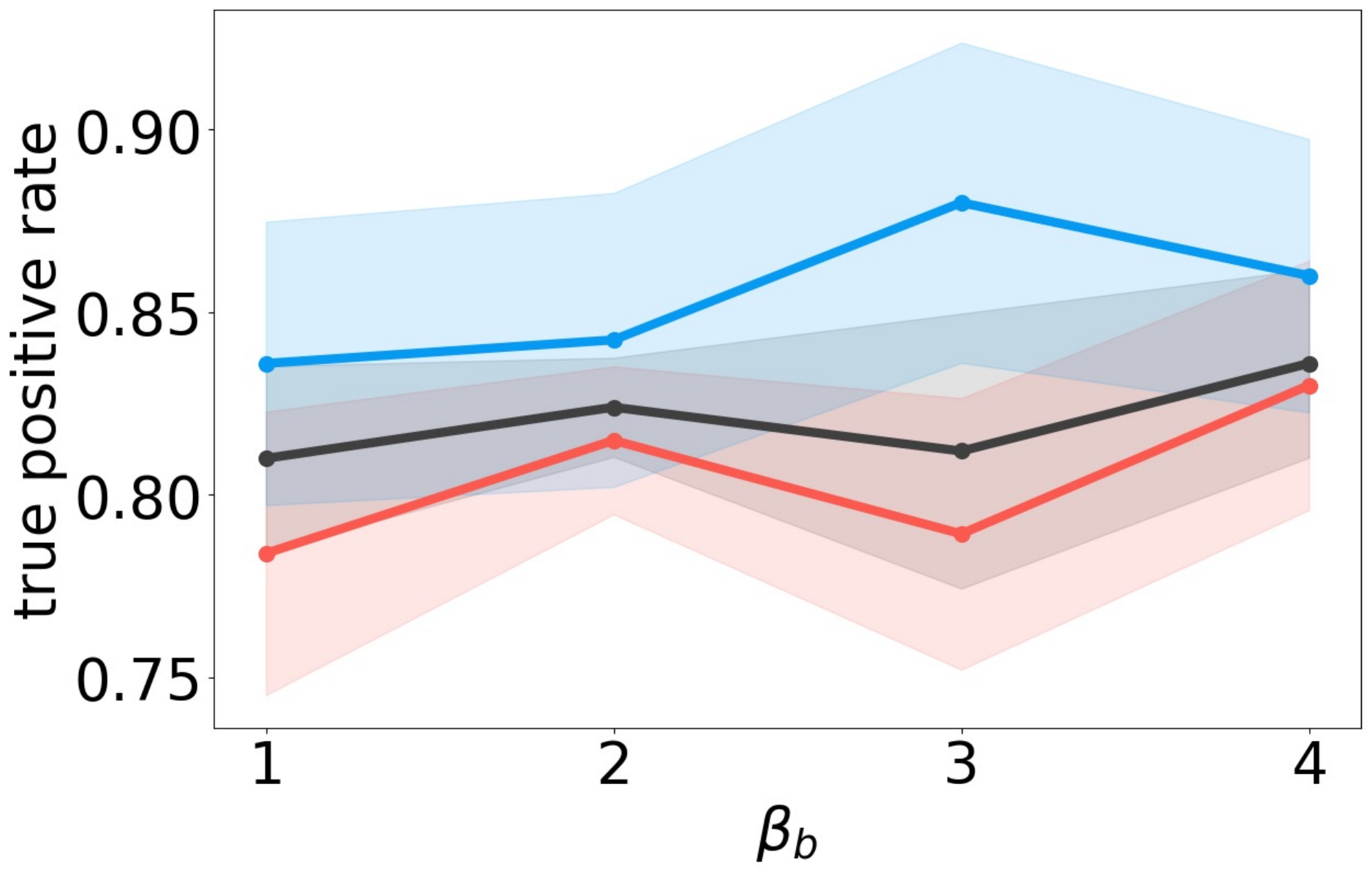}} &
     \hspace{-0.1in}{\includegraphics[width=0.195\textwidth]{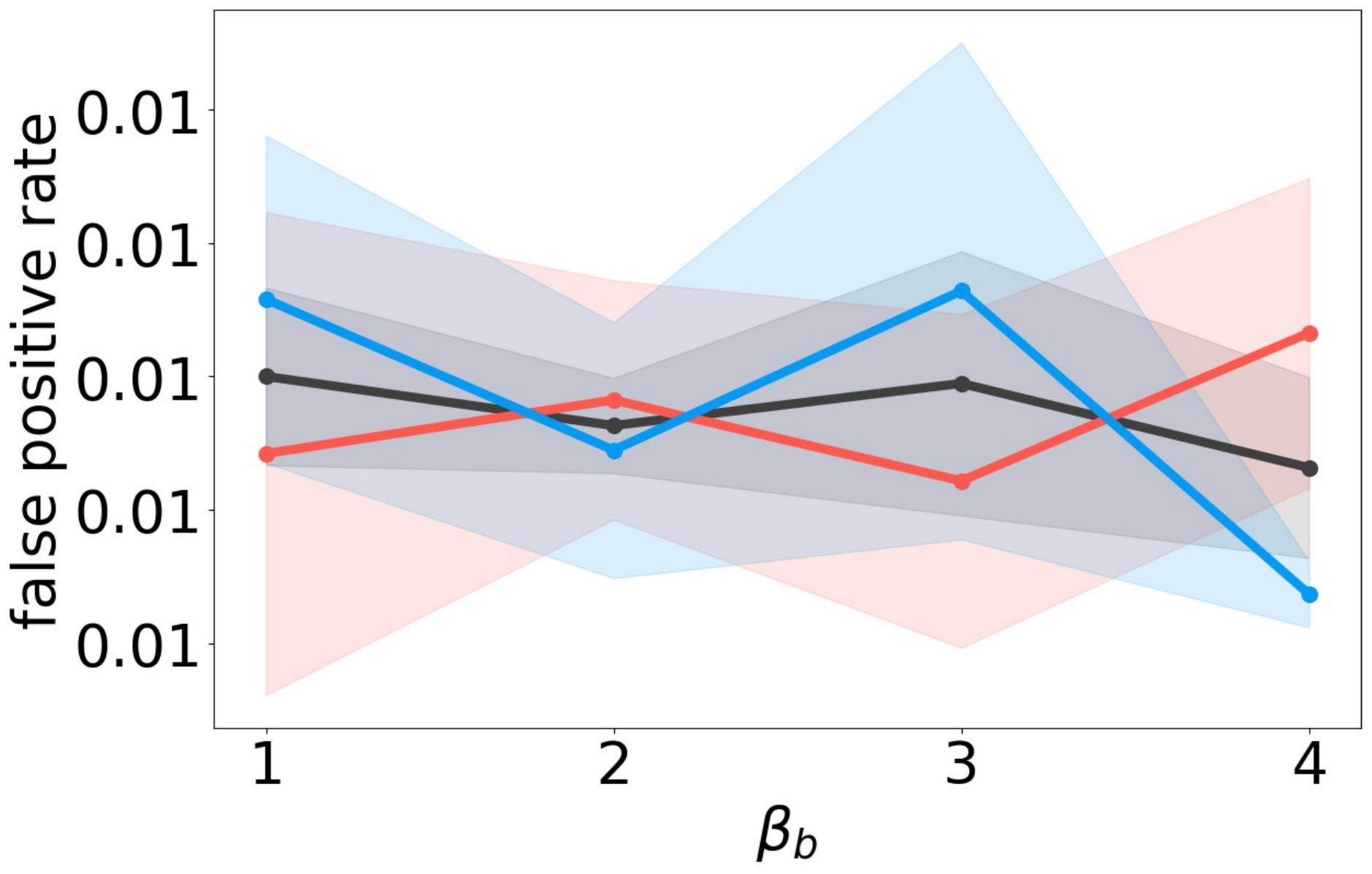}}
    &
   \hspace{-0.1in} {\includegraphics[width=0.195\textwidth]{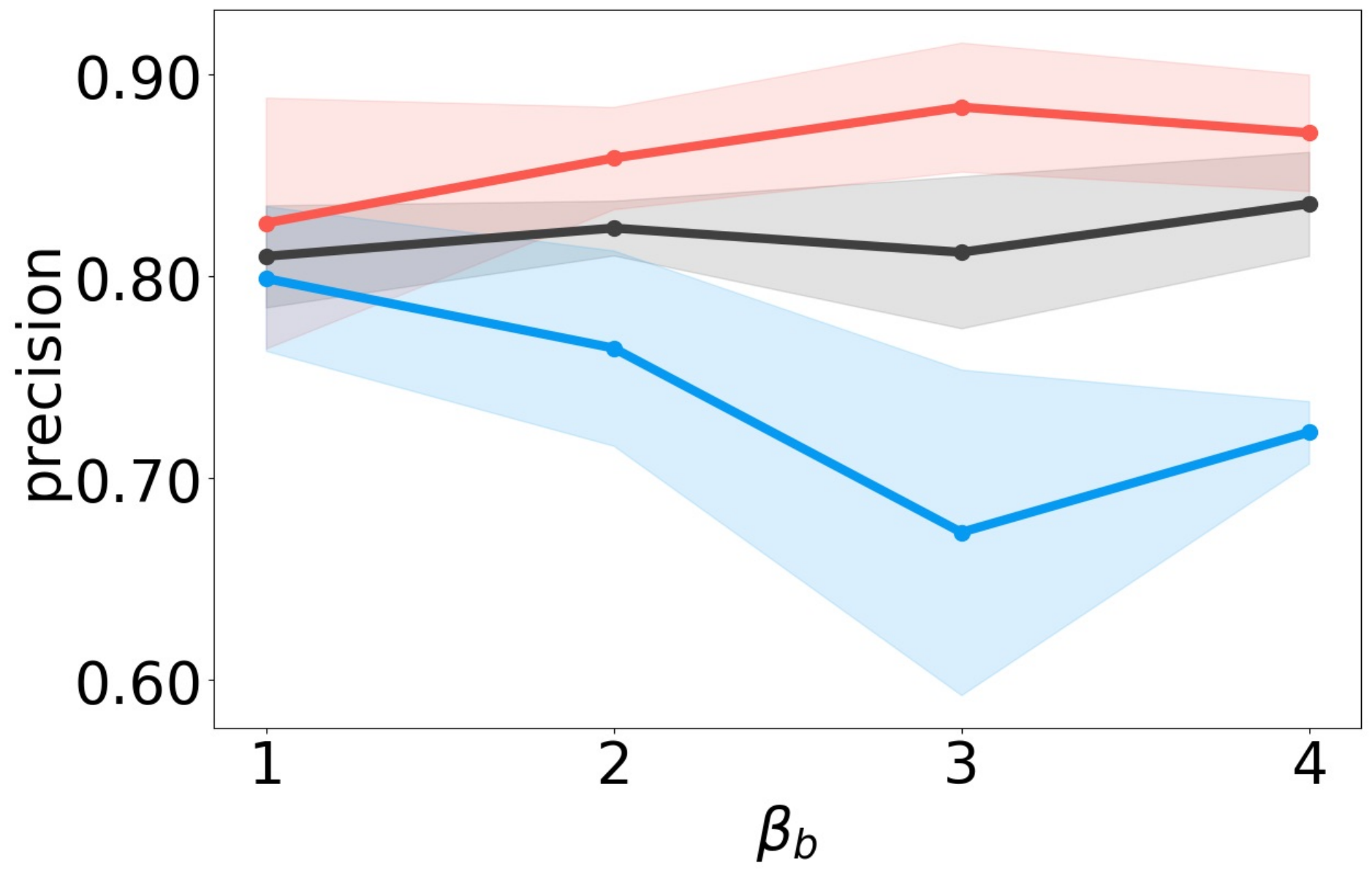}}
   &
     \hspace{-0.1in}{\includegraphics[width=0.195\textwidth]{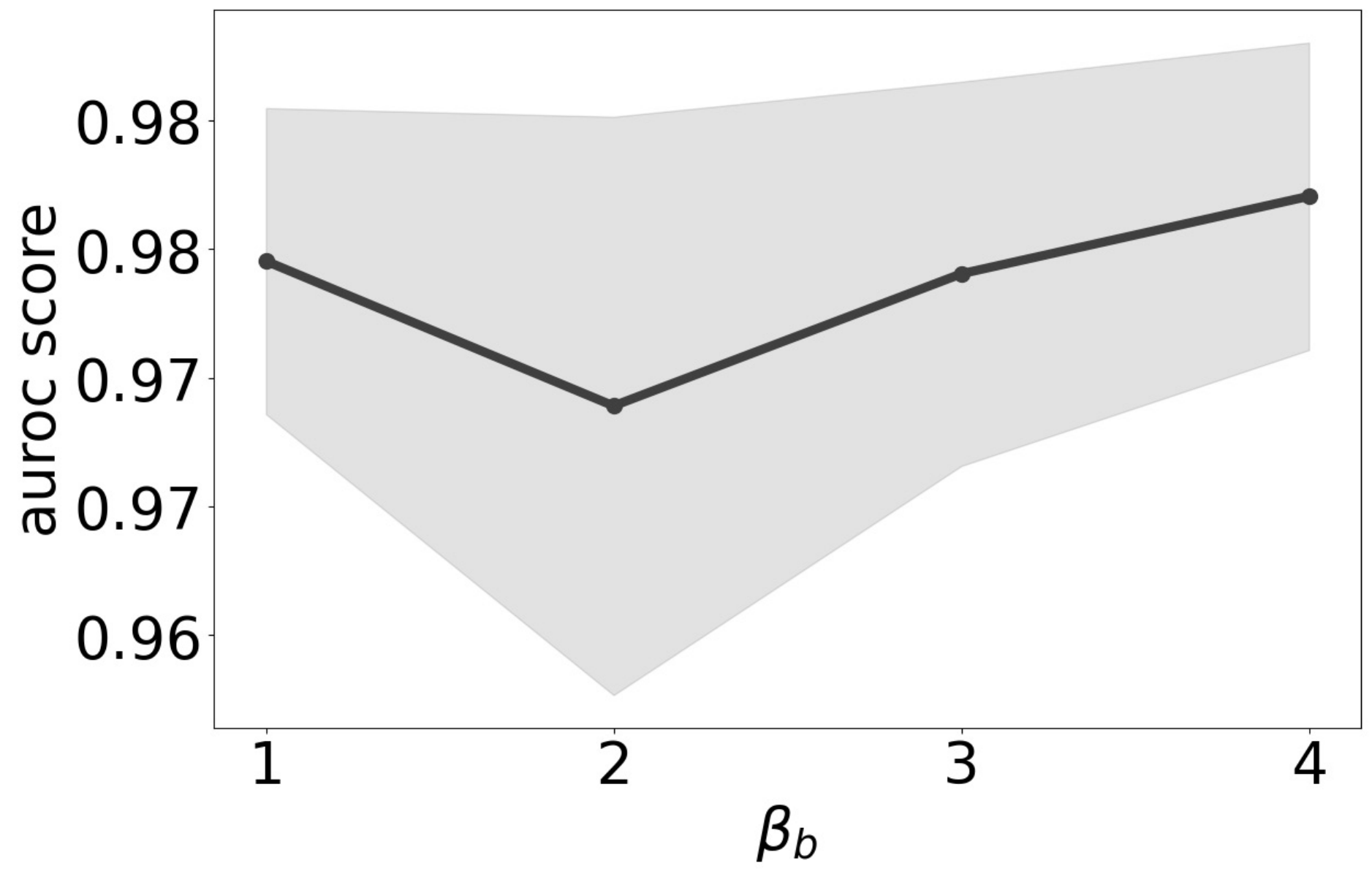}}
   \\
   \hspace{-0.1in}{\includegraphics[width=0.195\textwidth]{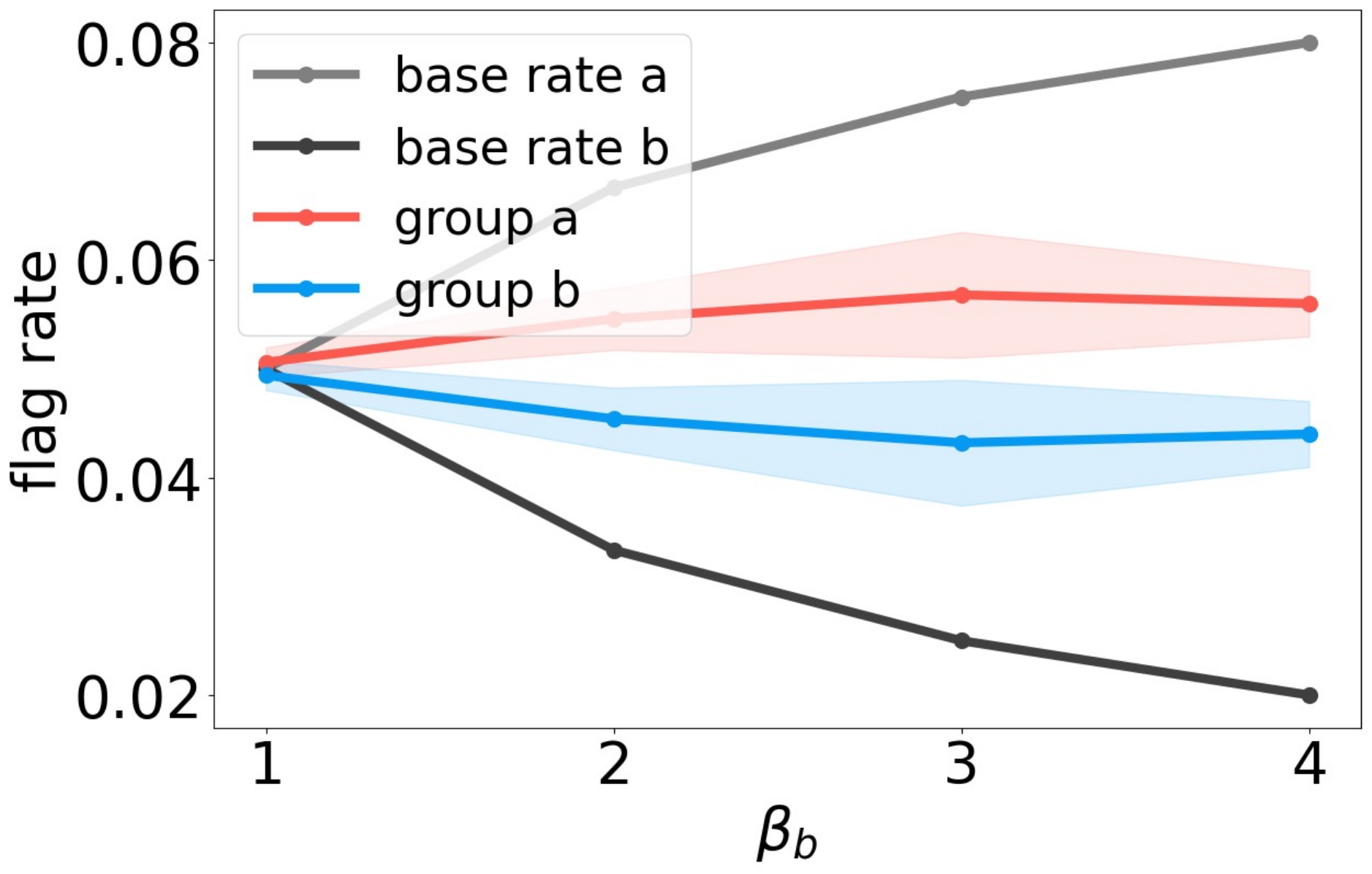} }
    &
    \hspace{-0.1in}{\includegraphics[width=0.195\textwidth]{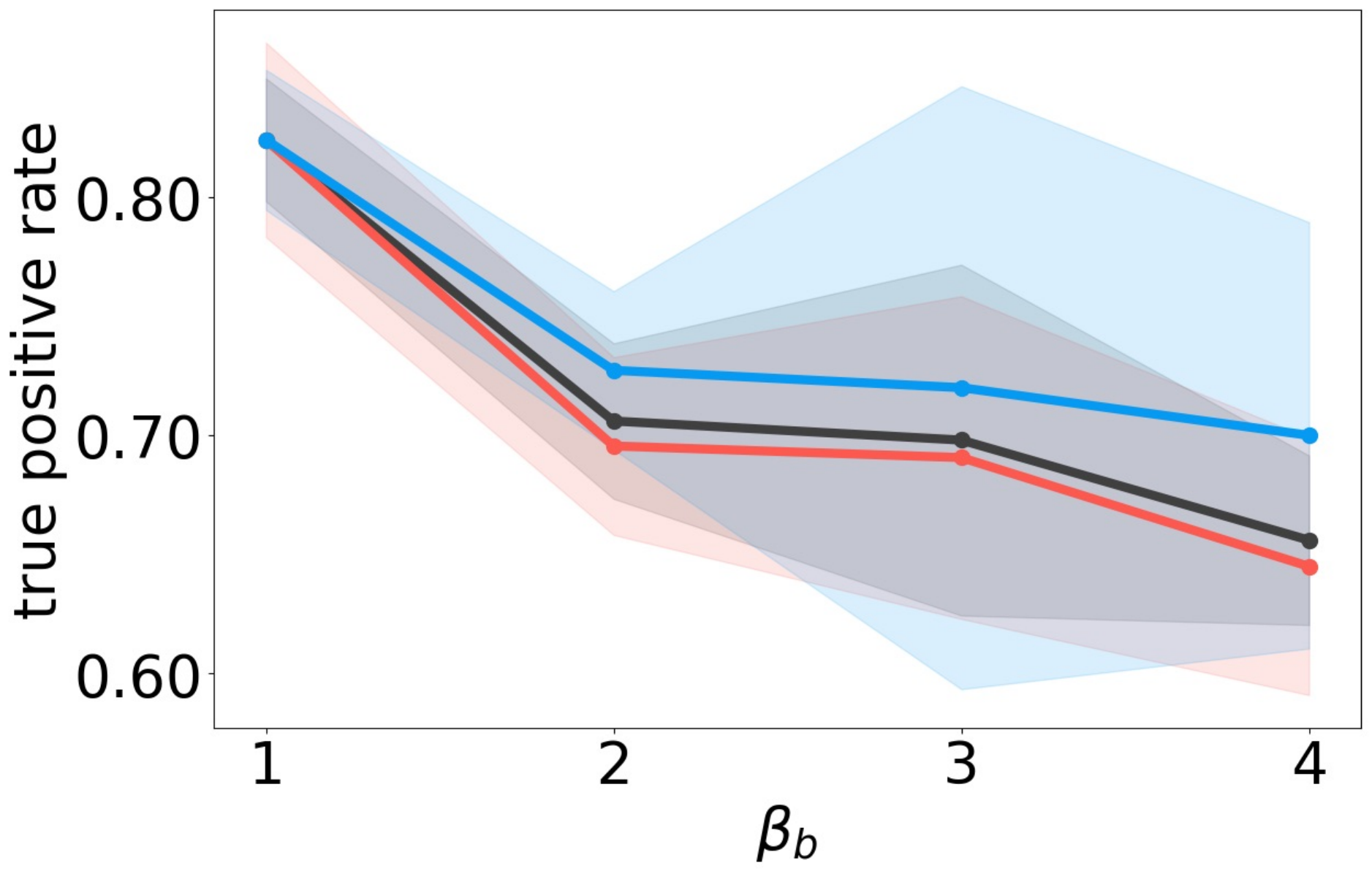}} &
     \hspace{-0.1in}{\includegraphics[width=0.195\textwidth]{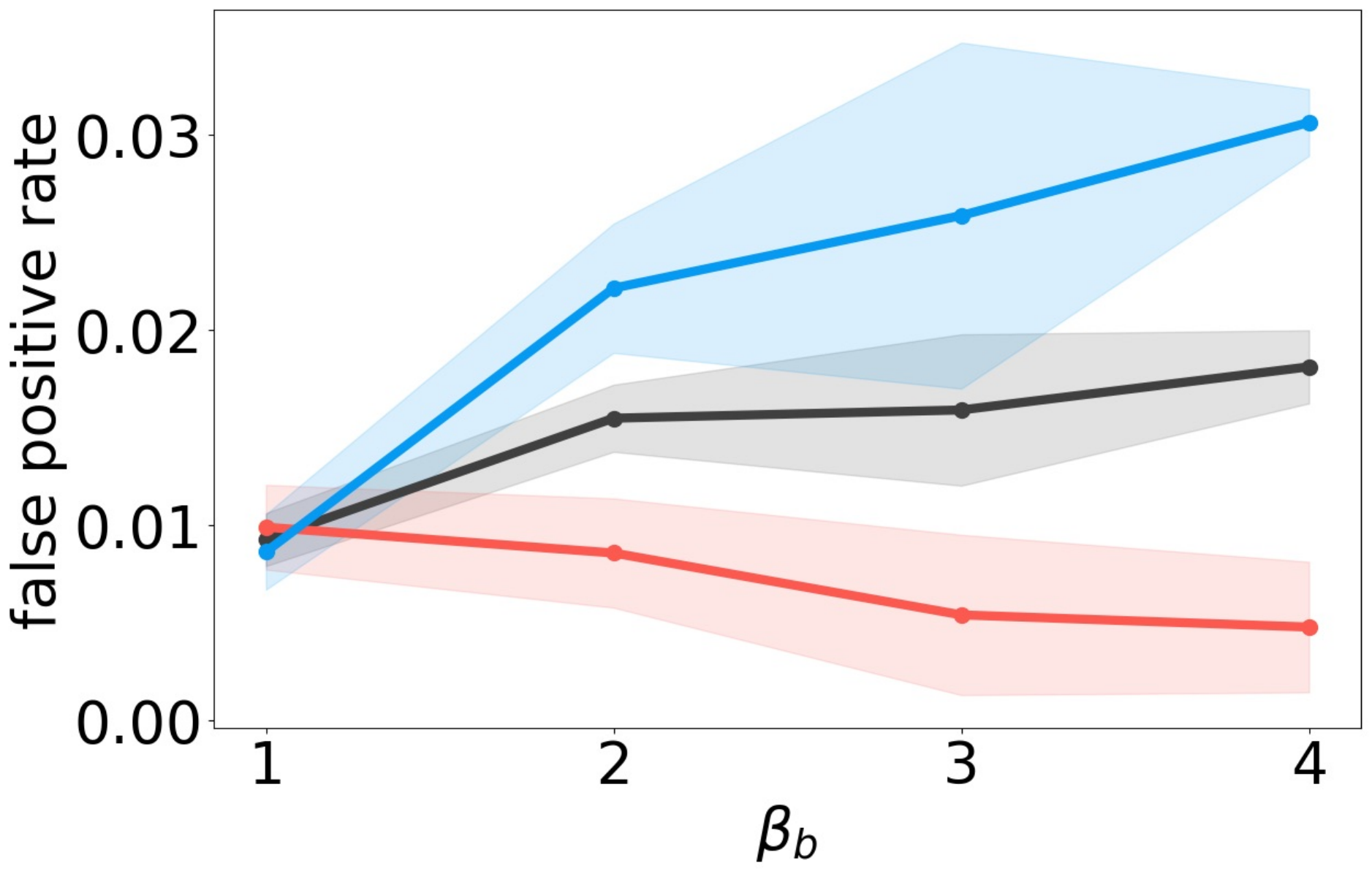}}
    &
   \hspace{-0.1in} {\includegraphics[width=0.195\textwidth]{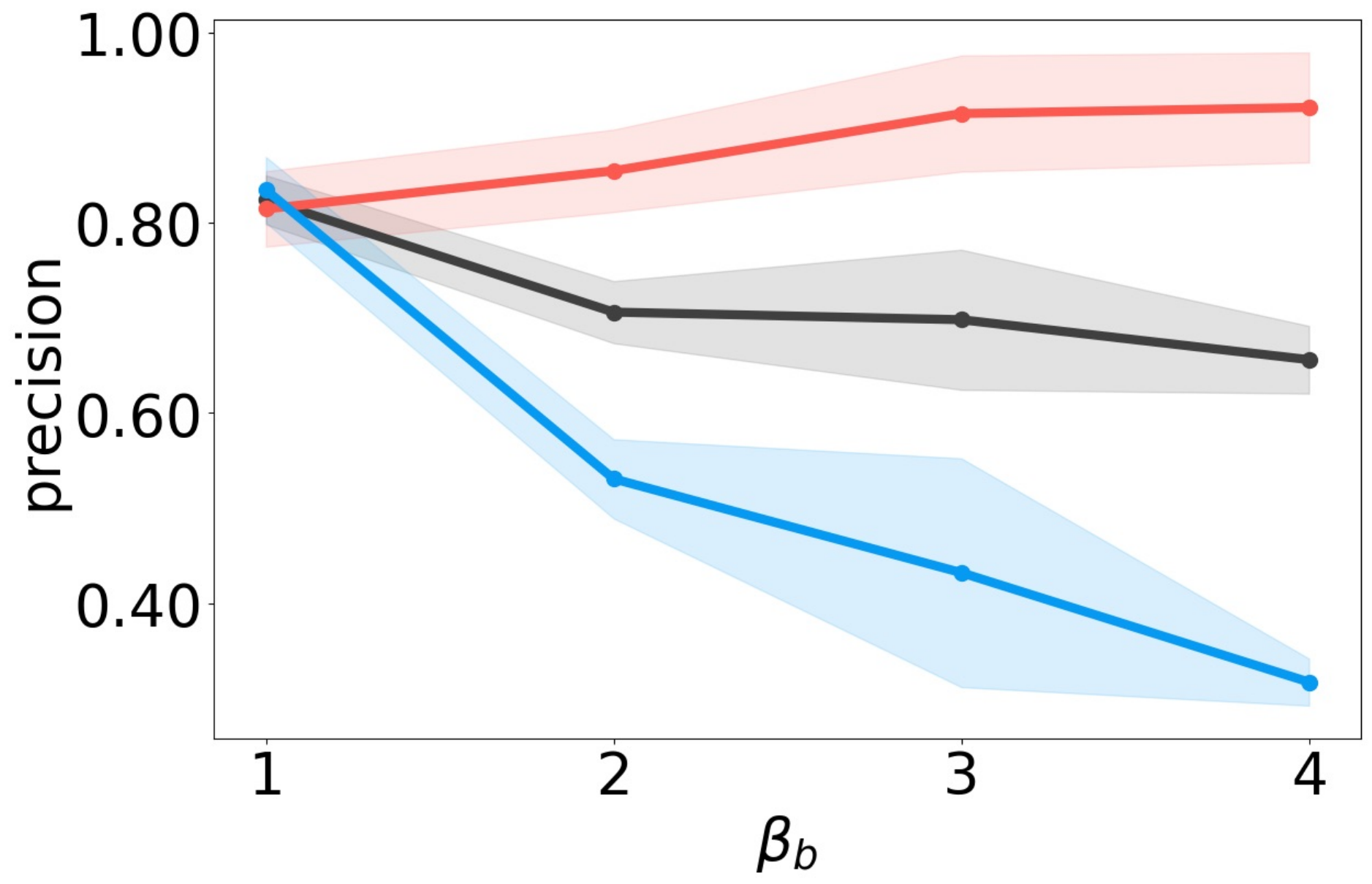}}
   &
   \hspace{-0.1in} {\includegraphics[width=0.195\textwidth]{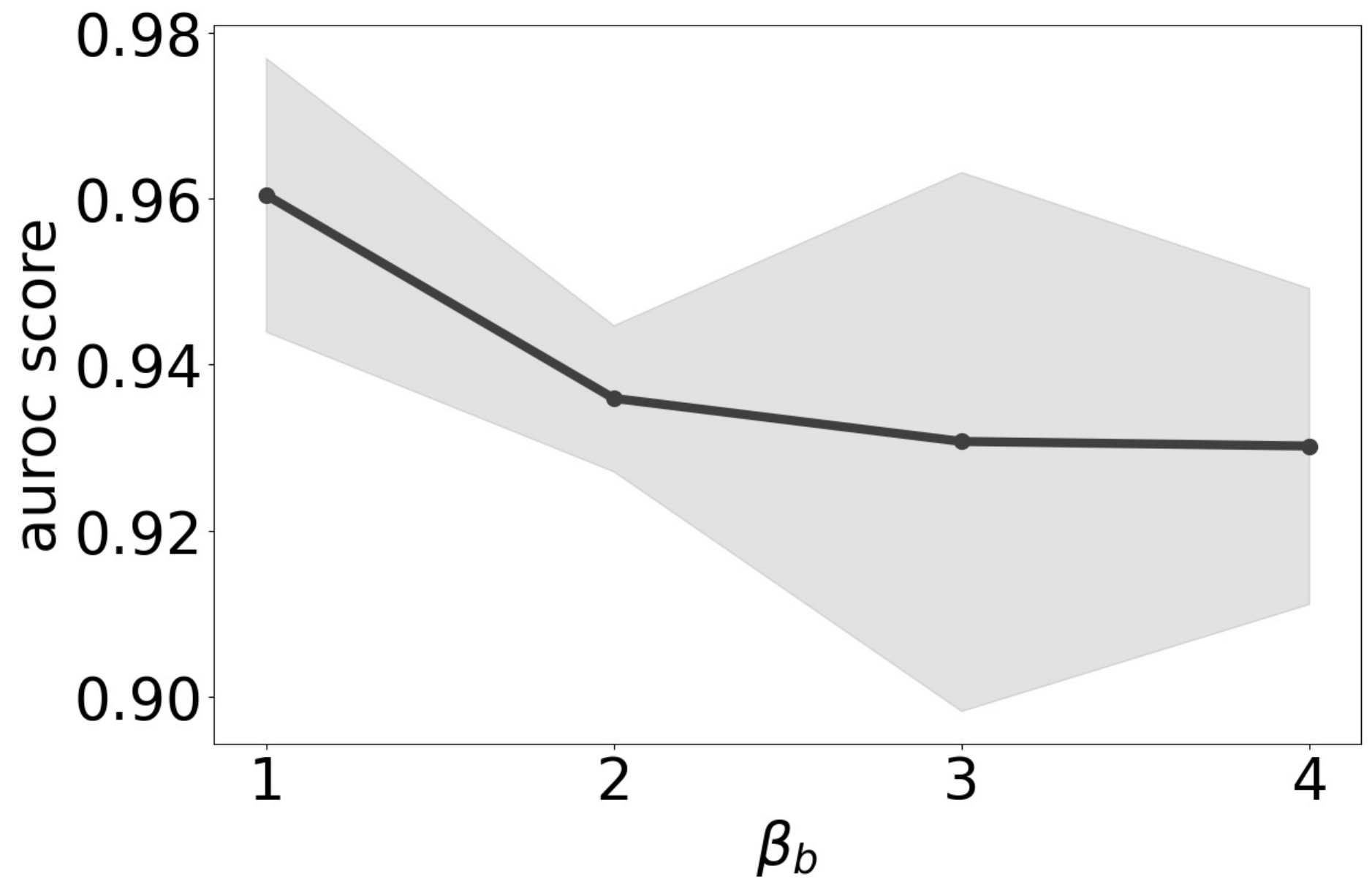}}
    \end{tabular}
    \caption{(best in color) Group-wise fairness metrics and AUROC for (top to bottom) LOF, iForest, DeepAE and FairOD under \underline{unequal base rates} on \textit{scattered} outliers. }
    \label{fig:baserate_all_scatter}
\end{figure*}

See Fig. \ref{fig:baserate_all_scatter}.

\FloatBarrier %
\clearpage %
\newpage %

\section{Theoretical Analysis}
\label{asec:theory}

In this section, we derive theoretical justification of why LOF and iForest can adapt well (or poorly) to different injected biases. We first introduce the mathematical notations in Apdx. \ref{assec:notations}, the working mechanism and the underlying quantities for these two OD models in Apdx. \ref{assec:preliminary}, and then provide theoretical results in Apdx. \ref{assec:theory_result}.

\subsection{Notations}
\label{assec:notations}
\begin{table}[h!]
\centering
\begin{tabular}{c|c}
\hline
Name & Variable \\
\hline
Size of population   & $n$   \\
Intra-group distance   & $\intra$  \\
Intra-group dist. of Group $a$   & $\intraa$  \\
Intra-group dist. of Group $b$   & $\intrab$  \\
Avg. dist. between inliers and outliers & $\Delta$ \\
Avg. dist. between inliers and outliers of Group $a$ & $\dista$ \\
Avg. dist. between inliers and outliers of Group $b$ & $\distb$ \\
Avg. dist. between subpopulations in $\mathcal{X}_g$ & $\Delta_g$ \\
LOF number of nearest neighbors & $k$ \\
Sensitive attribute of two protected groups & $G$ \\
Input features & $\mathbf{X}$ \\
Target variable & $Y$ \\
Outlier score & $S$ \\
Outlier label of OD model & $O$ \\
Base rate of a group $G$ & $br_G$\\
Flag rate of a group $G$ & $fr_G$ \\
(Proxy) Variables correlated w/ group membership & $\mathbf{X}_g$ \\
 (Incriminating) Variables correlated w/ target & $\mathbf{X}_c$\\
Non-incriminating occlusion variables & $\mathbf{X}_o$\\
Set of $k$-nearest neighbors around a point $p$ & $N_k(p)$ \\
Set of outliers & $\mathcal{O}$ \\
Set of inliers & $\mathcal{I}$ \\
Set of local neighborhoods in scattered outliers & $\mathcal{M}$ \\
An iTree's \underline{s}plit \underline{p}oint & $sp$ \\
\hline
\end{tabular}
\end{table}

\subsection{Preliminaries}
\label{assec:preliminary}

\subsubsection{\bf Probability and Outlier Score}
In the main text, we compare flag rates $P(O =1| G)$ to measure the fairness of an OD algorithm. However, our outlier detection algorithms rarely give a direct estimation of the true probability scores. For example, LOF algorithm uses local outlier factor(LOF) as the outlier score, iForest measures the outlierness by average path lengths of iTrees, and DeepAE outputs the reconstruction score of decoder.

Given any outlier detection algorithm, the probability of flagging  a point as outlier $P(O=1)$ can be written as
\begin{align} 
\nonumber
P(O=1) &= \Big[P(O=1 |G=a, Y=1)P(Y=1|a)  \\
\nonumber &+ P(O=1 |G=a, Y=0) P(Y=0 |a) \Big]P(G=a)  \\
\nonumber &+ \Big[P(O=1 |G=b, Y=1)P(Y=1| b) \\ &+ P(O=1 |G=b, Y=0) P(Y=0 |b) \Big]P(G=b)
\end{align}

During our analysis, we assume that protected groups $a$ and $b$ have equal size $n_a$, $n_b$ and equal base rate $P(Y=1|a) = P(Y=1|b)$. We also assume that the outlier detection algorithm flags a constant number of outliers (equal to the true count of outliers in the data), making $P(O=1)$ remain constant throughout our analysis. 

 Then, we can relate outlier scores to probability. If we assign higher outlier scores $S$ to the points (either inlier or outlier points) from either protected group $a$ or $b$, it implies that the probability of flagging the points $P(O=1 | G,Y)$ increases for that group, with possible decrease in probabilities of flagging the data points (either inlier or outlier points) from the other group.

 \subsubsection{\bf Mechanism of LOF}
 LOF computes the local density deviation of a given data point with respect to its neighborhood points, and using the ratio of density deviation as the outlier score $S$. A score $S$ larger than $1$ indicates that a point $p$ is more likely an outlier with higher $P(O=1|p)$. 
 
 To compute the local density deviation, one needs to compute the reachability distance. The detailed calculations are outlined by the definitions below.

\begin{definition}[k-distance] Assume the distance between two data points is $d(p,o)$. Then, given a dataset $\mathcal{D}$, a data $p \in \mathcal{D}$ and a positive integer $k$, where $o_k$ is the $k$-th nearest neighbor point of $p$, the $k$-distance of  $p$ is defined as  
\[\text{k-distance(p)} = dist(p,o_k) \;. \]
\end{definition}

\begin{definition}[Reachability Distance] Reachability distance (RD) is the distance to travel from a point to its neighbor. For any two data points $p,o \in \mathcal{D}$, the reachability distance is given as 
\[\text{RD}_k (p,o) = \max (\text{k-distance}(o),dist(p,o) )\;.\] 
\end{definition}

\begin{definition}[Local Reachability Density] For any point $p$, denote the set of $k$-nearest neighbor points by $N_k(p) \subseteq \mathcal{D}$. then, the local reachability density (LRD) of $p$ is:
\[\text{LRD}_k(p) = \frac{1}{
\sum_{o \in N_k(p)} RD_k (p,o)/ k
}\;.
\] 
\end{definition}

\begin{definition} [LOF Score] The core of the LOF algorithm is the calculation of local outlier factor (LOF), which measures density deviation between a point and its neighbors. Formally, LOF of $p$ is calculated as

\[S(p) = \text{LOF}_k(p)= \frac{\sum_{o \in N_k(p)} \text{LRD}_k(o)}{ k \cdot \text{LRD}_k(p)} \;. \]
\end{definition}

An LOF score that is notably larger than $1$ associates with relatively smaller local density and hence larger outlierness, while the factor being  close to $1$ means that the local density  is similar to that of the neighbors', suggesting an inlier.

\subsubsection{\bf Mechanism of iForest}
 iForest considers the average number of steps (i.e. randomized splits) required to isolate a point from the others. It starts with building iTrees, which makes random threshold splits sequentially on randomly chosen features. A iTree is complete when  all points are the same in a leaf node, or a leaf node contains only a single point. 
 
 The outlier score of a point $p$ is the average path length across an ensemble of iTrees to reach $p$'s leaf from the root. 

\begin{definition}[Path Length] Path length denoted as $h(p)$ of a point $p \in \mathcal{D}$ is measured by the number of edges $p$ traverses an iTree from the root node until the traversal is terminated at a leaf node.
\end{definition} 

\begin{definition}[iForest Score] For a  point $p \in \mathcal{D}$ where $|\mathcal{D}| = n$, the score is calculated as
\[S(p, n) = 2^{-\frac{\mathbb{E}[h(p)]}{c(n)}} \;,\]
\end{definition} 
where $c(n) = 2H(n - 1) - (2(n-1)/n)$ and $H(i)$ is the harmonic number.

Note that as $c(n)$ is fixed by the size of dataset,  $p$ stands out as outlier the smaller is the $\mathbb{E}[h(p)]$ (average path length across different iTrees), which indicates that $p$ is easily separated (i.e. isolated) from rest of the points via random threshold-splits. 

It is difficult to directly estimate the average path length over an ensemble of iTrees, given that each iTree is built using a subsample of data based on consecutively splitting on random features and random threshold. Therefore, we analyze \textit{how likely} it is  for an iTree to split on a feature (or multiple features) and isolate a point as outlier, where we associate a higher likelihood with a shorter expected path length (hence higher outlier score $S$). 

\subsection{Theoretical Analysis}
\label{assec:theory_result}

Propositions for LOF stand on the following two lemmas we first prove, which relate to the value of the LOF hyperparameter $k$.

\subsubsection{\bf Lemma 1} \textit{For the clustered outliers setting, let $\inliers$ and  $\outliers$ denote the set of inliers and outliers, where $|\outliers| < |\inliers|$. Within $\inliers$ and $\outliers$, the average intra-group distances are denoted by  $\intra$. Between $\inliers$ and $\outliers$, the average distance  is denoted by $\Delta$ s.t. $\Delta > \intra$. Then, for LOF to be able to bypass masking and flag  all clustered true outliers, number of neighbors $k$ needs to be chosen s.t. $k > |\outliers|$.}

\textit{\bf Proof:}
We first prove that outliers would mask 
when $k \leq |\outliers|$. 
Consider an outlier point $p \in \outliers$.
Since $\Delta > \intra$,  $p$'s $k$ nearest neighbors consist only of outliers, i.e. are points within the set $\outliers$, $N_k(p) \subseteq \outliers$.

Then,  the expected LOF (outlier score $S$) of $p$ is calculated as follows.
\begin{align}
\mathbb{E}[
\text{LOF}_k(p)] &= \mathbb{E} \Bigg [  \frac{\sum_{o \in N_k(p)} \text{LRD}_k(o)}{ |N_k(p)| \cdot \text{LRD}_k(p)} \Bigg] \\  &=  \frac{\sum_{o \in N_k(p)} \mathbb{E} \big [ \text{LRD}_k(o) \big] }{ k \cdot  \mathbb{E} \big [\text{LRD}_k(p)\big]}  \\
 &= \frac{\sum_{o \in N_k(p)} \mathbb{E} \big [\frac{1}{\sum_{j \in N_k(o)} \text{RD}_k(o,j)/ |N_k(o)|} \big]}{k \mathbb{E}\big[ \frac{1}{\sum_{m \in N_k(p)} \text{RD}_k (m,p)/ |N_k(p)| } \big]} \\ &= \frac{\sum_{o \in N_k(p)} \frac{1}{\sum_{j \in N_k(o)} \mathbb{E} [\text{RD}_k(o,j)] / k} }{k  \frac{1}{\sum_{m \in N_k(p)} 
 \mathbb{E} [\text{RD}_k (m,p)]/ k } }\label{eq:expected_LOF} \\ 
&\approx \frac{k \cdot \frac{1}{ k \cdot \intra / k} }{k  \frac{1}{k \cdot \intra / k } }  = 1
\end{align}

 Similarly, for an inlier point $z \in \inliers$, 
 $z$'s $k$ nearest neighbors consist only of inliers, when $k \leq |\outliers| < |\inliers|$.
When $N_k(z) \subset \inliers$, the average reachability distances of $z$ and $z$'s neighbors are all $c$. Therefore, expected LOF of $z$ is $1$. 

As a result, LOF algorithm fails to detect any outliers due to both outliers and inliers having the same expected outlier score $S$.

In contrast, when $|\inliers| > k > |\outliers|$, we instead have $ |\outliers| \subseteq N_k(p)$. The numerator of Eq. \eqref{eq:expected_LOF}, that is the local reachability density of neighborhood points  becomes 
\begin{align}
\nonumber  &\sum_{o \in \outliers} \frac{1}{\sum_{j \in N_k(o)} \mathbb{E} [\text{RD}_k(o,j)] / k} \\  \nonumber &+ \sum_{m \in N_k(p)\backslash \outliers}  \frac{1}{\sum_{n \in N_k(m)} \mathbb{E} [\text{RD}_k(m,n)] / k} \\ =&\sum_{o \in \outliers} \frac{1}{\Delta} + \sum_{m \in N_k(p)\backslash \outliers} \frac{1}{ 
 \intra}  \\
 =& \frac{|\outliers|}{\Delta} + \frac{k-|\outliers|}{\intra} 
\end{align}

The denominator of Eq. \eqref{eq:expected_LOF}, that is the local reachability density of $p$ becomes
\begin{align}
&k \frac{1}{\sum_{m \in N_k(p)} \mathbb{E} [\text{RD}_k (m,p)]/ k } \\
 &=   \frac{k}{\sum_{m \in \outliers} \mathbb{E} [\text{RD}_k (m,p)]/k  + \sum_{n \in N_k(p) \backslash \outliers} \mathbb{E} [\text{RD}_k (n,p)] /k} \label{eq:RD_distance_inlier_outlier} \\ 
 &= \frac{k}{|\outliers| \cdot \Delta/k + (k-|\outliers|) \cdot \Delta/k} = \frac{k}{\Delta}
 \label{eq:denominator}
\end{align}

Note that in Eq. \eqref{eq:RD_distance_inlier_outlier}, for both outlier or inlier neighbors of $p$, the reachability distance is $\Delta$, since $k$-distance is the furthest reachable point for the $k$-neighbors (recall Definition 2). Combining both, the expected LOF becomes
\begin{align}
\mathbb{E}[
\text{LOF}_k(p)] = \frac{\frac{ |\outliers|} {\Delta} + \frac{k-|\outliers|}{\intra} }{ \frac{k}{\Delta}} > 1
\end{align}

Since $k < |\inliers|$, the expected LOF remains $1$ for any inlier. As such, $p \in \outliers$ is assigned a higher $S$ than $z \in \inliers$ only when $|\inliers| > k > |\outliers|$, which concludes the proof. \qed

\subsubsection{\bf Lemma 2} 
\textit{In clustered outlier setting, consider multiple outlier sets $\outliers_i$, $i \in \{1,...,j\}$ and one inlier set $\inliers$, with the average intra-group distance for each $\outliers_i$ and $\inliers$ is denoted by $c$, and the average distance between outliers and inliers for each $\outliers_i$ is denoted $\Delta$ s.t. $\Delta > c$. Then, for LOF algorithm to detect all true outliers in the micro-clusters, the number of neighbors $k > \max(|\outliers_1|,..., |\outliers_j|)$.}

\textit{\bf Proof}: Based on Lemma 1, to detect all outliers from $\outliers_i, i \in \{1,...,j\}$, we need $k > |\outliers_i|$ for all $i$. Therefore, $k > \max(|\outliers_1|,..., |\outliers_j|)$. \qed

\subsubsection{\bf Proposition 1} 
\begin{figure}[h]
\centering
\includegraphics[width=0.76\linewidth]{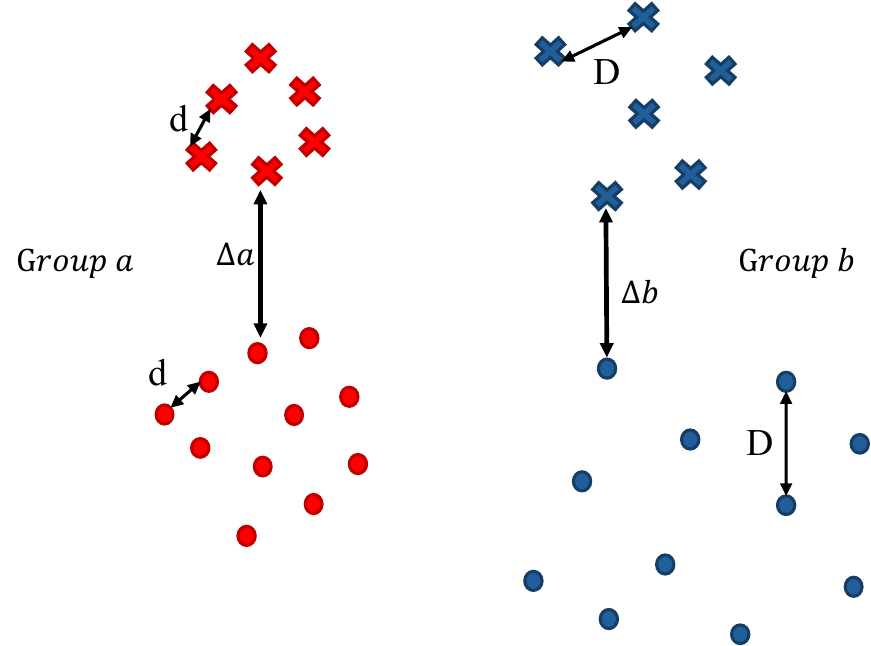}
\caption{For clustered outliers, illustration of density variation between group $a$ and $b$ (Proposition 1).}
\label{fig:prop1}
\end{figure}

\textit{In the clustered outliers setting, let groups $a$ and $b$ have equal size $n_a=n_b$ and equal base rate $P(Y=1|a)=P(Y=1|b)$.
Let $\intraa$ and $\intrab$ denote intra-group distance between inlier pairs (same for outlier pairs) for group $a$ and $b$, respectively, where $\intrab>\intraa$ as group $b$ is sparser. 
Also denote by $\dista$ and $\distb$ the average distance between outliers and inliers in each group (See Fig. \ref{fig:prop1}).
Then, assuming LOF hyperparameter $k$ is set s.t. $k > n_a \cdot P(Y=1|a)$ and $\dista = \distb$, LOF tends to assign higher  scores to group $a$-outliers, increasing flag rate $P(O=1|a)$.
Further, when $\Delta_b \approx D$, LOF score of $b$-outliers is $\approx 1$, i.e. close to inlier scores, leading to low TPR for group $b$ due to masking.
}

\textit{\bf Proof}:
Let us denote by $\inliers_a$, $\inliers_b$ for the sets of inliers  and by $\outliers_a$, $\outliers_b$  the sets of outliers in groups $a$ and $b$, respectively. Based on Lemma 1,  $k > |\outliers_a| = n_a \cdot P(Y=1|a) = |\outliers_b| =  n_b \cdot P(Y=1|b)$.
Then, for an outlier point $p_a \in \outliers_a$, the expected LOF is given as 
\begin{align}
\mathbb{E}[
\text{LOF}_k(p_a)] &= \mathbb{E} \Bigg [  \frac{\sum_{o \in N_k(p_a)} \text{LRD}_k(o)}{ |N_k(p_a)| \cdot \text{LRD}_k(p_a)} \Bigg] \\  &=  \frac{\sum_{o \in N_k(p_a)} \mathbb{E} \big [ \text{LRD}_k(o) \big] }{ k \cdot  \mathbb{E} \big [\text{LRD}_k(p_a)\big]} \\ 
 &= \frac{\sum_{o \in \outliers_a} \frac{1}{\sum_{j \in N_k(o)} \mathbb{E} [\text{RD}_k(o,j)] / k} }{k \cdot  \mathbb{E} \big [\text{LRD}_k(p_a)\big]}  \\ \nonumber &+ \frac{ \sum_{m \in N_k(p_a)\backslash \outliers_a}  \frac{1}{\sum_{n \in N_k(m)} \mathbb{E} [\text{RD}_k(m,n)] / k} }{k \cdot  \mathbb{E} \big [\text{LRD}_k(p_a)\big]} \\
 &= \frac{\frac{|\outliers_a|}{\dista} + \frac{k - |\outliers_a|}{\intraa}}{\frac{k}{\dista}} \;.
\end{align}

Using the same calculation, the expected LOF for $p_b \in \outliers_b$ is:
\begin{align}
\label{eq:pblof}
\mathbb{E}[
\text{LOF}_k(p_b)] = \frac{\frac{|\outliers_b|}{\distb} + \frac{k - |\outliers_b|}{\intrab}}{\frac{k}{\distb}} \;.
\end{align}

Since $\intrab > \intraa$ (group $b$ is sparser) and $\dista = \distb$,
\begin{align}
 \frac{\frac{|\outliers_a|}{\dista} + \frac{k - |\outliers_a|}{\intraa}}{\frac{k}{\dista}} &> \frac{\frac{|\outliers_b|}{\distb} + \frac{k - |\outliers_b|}{\intrab}}{\frac{k}{\distb}} \\
 \implies \mathbb{E}[
\text{LOF}_k(p_{a})] &> \mathbb{E}[\text{LOF}_k(p_{b})] \;.
\end{align}

 Therefore, for the outlier scores $S_a$ and $S_b$ associated with $p_a$ and $p_b$, we have $S_a$ > $S_b$.  Flag rate $P(O= 1| G=a) = P(O=1|G=b,Y=1)P(Y=1|G=b) +P(O=1|G=b,Y=0)P(Y=0|G=b)$ increases with higher outlier scores assigned to group $a$-outliers as $P(O=1|G=a,Y=1)$ increases.

 When $\intrab \approx \distb$, group $b$ outliers $\outliers_b$ become indistinguishable from the  inliers $\inliers_b$, where the expected LOF for $p_b$ in Eq. \eqref{eq:pblof} becomes 
 \begin{align}
\mathbb{E}[
\text{LOF}_k(p_{b})] &= \frac{\frac{|\outliers_b|}{\distb} + \frac{k - |\outliers_b|}{\distb}}{\frac{k}{\distb}} = 1 \;.
\end{align}

Then, group $b$-outliers mask as they blend with the inlier points with similar LOF score ($\approx 1$), leading to low TPR for group $b$. \qedsymbol{}

\subsubsection{\bf Proposition 2} 
\begin{figure}[h]
\centering
\includegraphics[width=0.9\linewidth]{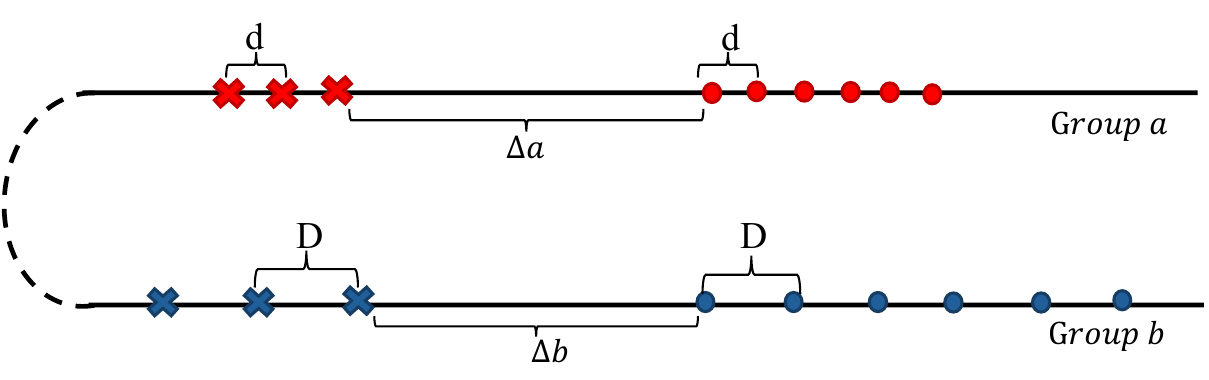}
\caption{For clustered outliers, illustration of density variation between group $a$ and $b$ in 1-d for iForest (Proposition 2).}
\label{fig:prop2}
\end{figure}

\textit{In the clustered outliers setting,
 let $sp$ denote an iTree's split, and $P(x_a < sp < x_b)$ denote the probability for a split to occur between points $x_a$ and $x_b$. Let groups $a$ and $b$ have equal base rate $P(Y=1|a)=P(Y=1|b)$
 and equal average distance between outliers and inliers $\dista = \distb$.
  As group $b$ is sparser, $\intraa< \intrab$ denotes the intra-group distances between inlier and outlier pairs for groups $a$ and $b$, respectively (See Fig. \ref{fig:prop2}). Let $\{\outliers_a$, $\inliers_a\}$, $\{\outliers_b$, $\inliers_b\}$ represent the outliers, inliers sets for groups $a$ and $b$. 
 Then, for $o_b, p_b \in \outliers_b$, and $o_a, p_a \in \outliers_a$, we have $P(p_b < sp < o_b) > P(p_a < sp < o_a)$, indicating an iTree is more likely to split among group-$b$ outliers. The difference between $P(p_b < sp < o_b)$ and $P(p_a < sp < o_a)$ becomes larger when iTree is built in higher dimensions. In addition, for $q_b \in \inliers_b$, $q_a \in \inliers_a$, $P(o_b < sp < q_b) = P(o_a < sp < q_a)$, i.e. an iTree is equally likely to split between the inliers and outliers clusters. }

\textit{\bf Proof:} For simplicity, we first consider the one-dimensional setting. Since iTree is built based on uniformly random splits between points and $\intrab>\intraa$, for $o_b, p_b \in \outliers_b$, and $o_a, p_a \in \outliers_a$, we have
\begin{align}
P(p_b < sp < o_b) &= \frac{\intrab}{\max_{x,y \in \inliers \cup \outliers} d(x,y) } \\ P(p_a < sp < o_a) &= \frac{\intraa}{\max_{x,y \in \inliers \cup \outliers} d(x,y) } \\
\implies P(p_b < sp < o_b) &> P(p_a < sp < o_a) \;. \label{eq:ficuts}
\end{align}

In words, iTree is more likely to split among $b$-outliers than $a$-outliers.  iForest assigns higher outlier scores to points that are isolated via fewer splits, yielding shorter root-to-leaf paths. Then, based on Eq. \eqref{eq:ficuts}, an iTree is more likely to isolate $b$-outliers and thus associate them with shorter paths, which explains why $P(O=1|G=b) > P(O=1 |G=a)$ for iForest.

On the other hand, for $p_b \in \outliers_b$, $q_b \in \inliers_b$, and $p_a \in \outliers_a$, $q_a \in \inliers_a$, we have
 \begin{align}
 P(p_b < sp < q_b) &= \frac{\distb}{\max_{x,y \in \inliers \cup \outliers} d(x,y) } \\ &= \frac{\dista}{\max_{x,y \in \inliers \cup \outliers} d(x,y) }= P(p_a < sp < q_a)
 \label{eq:iforst_equal_outlier_inlier}
 \end{align}

That is, iTree is equally likely to split between the inliers and outliers for both groups $a$ and $b$. This single split is crucial for both groups for ruling out the outliers collectively as a cluster. Due to this special split, iForest can flag certain fraction of $a$-outliers, leading to smaller decrease of TPR for group $a$.

 Extending the analysis to the multiple $h$-dimensional setting,  we assume the intra-group distance remains $\intrab>\intraa$ on each dimension, and distance $\dista = \distb$ holds between outliers and inliers for both groups $a$ and $b$. Then, for $\mathbf{o}_b, \mathbf{p}_b \in \outliers_b$, and $\mathbf{o}_a, \mathbf{p}_a \in \outliers_a$, we have
\begin{align}
&P(\mathbf{p}_b < sp < \mathbf{o}_b) - P(\mathbf{p}_a < sp < \mathbf{o}_a) \label{eq:iforst_n_dim}\\
=& \frac{\intrab^h}{\prod_{j=1}^h(\max_{\mathbf{x}, \mathbf{y} \in \inliers \cup \outliers} d(\mathbf{x}_j,\mathbf{y}_j)) }-  \frac{\intraa^h}{\prod_{j=1}^h(\max_{\mathbf{x}, \mathbf{y} \in \inliers \cup \outliers} d(\mathbf{x}_j,\mathbf{y}_j)) }
\end{align}

Since $\intrab > \intraa$, $D^h - d^h$ grows with increasing $h$. While the denominator is roughly the same, the discrepancy between two probabilities increases with larger $h$.  \qedsymbol{}

\subsubsection{\bf Proposition 3} 
\begin{figure}[h]
\centering
\includegraphics[width=0.8\linewidth]{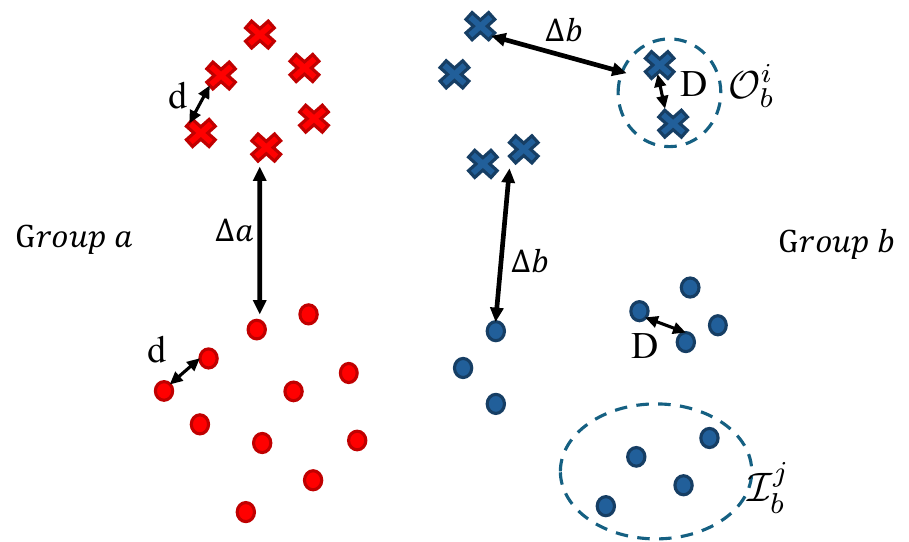}
\caption{For clustered outliers, illustration of membership obfuscation bias in group $b$ (Proposition 3).}
\label{fig:prop3}
\end{figure}

\textit{In the clustered outliers setting, let groups $a$ and $b$ have equal size $n_a=n_b$ and equal base rate $P(Y=1|a)=P(Y=1|b)$. Assume group $b$-inliers form $f$ smaller populations, $\inliers_b^{i}$, for $i \in [\![1,f]\!]$, and group $b$-outliers form $g$ smaller populations, $\outliers_b^{j}$ for $j \in [\![1,g]\!]$. 
Let $\intraa$ and $\intrab$ denote intra-group distance between inlier pairs (same for outlier pairs) for group $a$ and each sub-population in $b$, respectively, where $\intrab=\intraa$. Denote $\dista$ as the average distance between outliers and inliers in group $a$, and   $\distb$ as the average distance between  each subpopulation in group $b$ for both inliers and outliers (See Fig. \ref{fig:prop3}). Then, assuming LOF hyperparameter $k$ is set s.t. $k > n_a \cdot P(Y=1|a)$, and $\dista = \distb$ while $\dista > \intraa$ and $\distb > \intrab$, LOF tends to assign higher scores to group $b$-outliers, increasing flag rate $P(O=1|b)$.
Further, when $|\inliers_b^j| < |\outliers_b^i| < k$, LOF score of $b$-inliers is larger than of $b$-outliers, leading to both high FPR and low TPR for group $b$.}

\textit{\bf Proof}: For simplicity, we assume each subpopulation in group $b$ maintains the average intra-group distance $\intrab$, and the distances between each subpopulation remains at $\distb$. However, altering certain subpopulation's intra-group distance or distance to other subpopulations can cause one subpopulation to have lower/higher LOF outlier scores, thus complicating the analysis on detecting group $a$ and $b$ outliers. 

Let us denote by $\inliers_a$ and $\outliers_a$ the sets of inliers and outliers in group $a$.
Based on Lemma 2, LOF hyperparameter $k > n_a \cdot P(Y=1|a) > \max(|\outliers_b^{1}|, ..., |\outliers_b^{f}|)$. For an outlier point $p_b^i$ in subpopulation $\outliers_b^i$ for any $i \in [1,...,f]$, the average local reachability density is $\frac{k}{\distb}$ by the same calculation as Eq. \eqref{eq:denominator}. On the other hand, the local reachability density of $p_b^i$'s $k$-neighborhood is calculated as
\begin{align}
\frac{|\outliers_b^i|}{\distb} + \frac{k - |\outliers_b^i|}{\intrab}
\end{align}

In comparison to the expected LOF score of an outlier point $p_a \in \outliers_a$, with $\dista=\distb$, $\intraa=\intrab$, $\dista > \intraa$, and $\distb > \intrab$, we derive the following inequality.
\begin{align}
\frac{\frac{|\outliers_b^i|}{\distb} + \frac{k - |\outliers_b^i|}{\intrab}}{\frac{k}{\distb}} &>\frac{\frac{|\outliers_a|}{\dista} + \frac{k - |\outliers_a|}{\intraa}}{\frac{k}{\dista}} \\
\implies \mathbb{E}[
\text{LOF}_k(p_b^i)] &> \mathbb{E}[\text{LOF}_k(p_a)] \;.
\end{align}

Therefore, $p_b^i$ from group $b$ receives higher outlier score than $p_a$ ($S_b$ > $S_a$). Flag rate $P(O=1 |G=b)$ increases due to increased $P(O=1 |G=b, Y=1)$. 

In addition, when $|\inliers_b^j| <k$, an inlier $p_b^{j} \in \inliers_b^j$ has the following expected LOF score.
\begin{align}
\mathbb{E}[
\text{LOF}_k(p_b^{j})]
= \frac{\frac{|\inliers_b^j|}{\distb} + \frac{k - |\inliers_b^j|}{\intrab}}{\frac{k}{\distb}} 
\end{align}
If $|\inliers_b^j| < |\outliers_b^i|$, then  $\mathbb{E}[
\text{LOF}_k(p_b^{j})] > \mathbb{E}[
\text{LOF}_k(p_b^{i})]$, where $p_b^i \in \outliers_b^i$ is a true outlier; i.e. LOF score of $b$-inliers is higher than for $b$-outliers, in this case. As $P(O=1 | G=b,Y=0) > (O=1 |G=b, Y=1)$, FPR increases and TPR drops for group $b$.  \qedsymbol{}

\subsubsection{\bf Proposition 4}
\begin{figure}[h]
\centering
\includegraphics[width=0.9\linewidth]{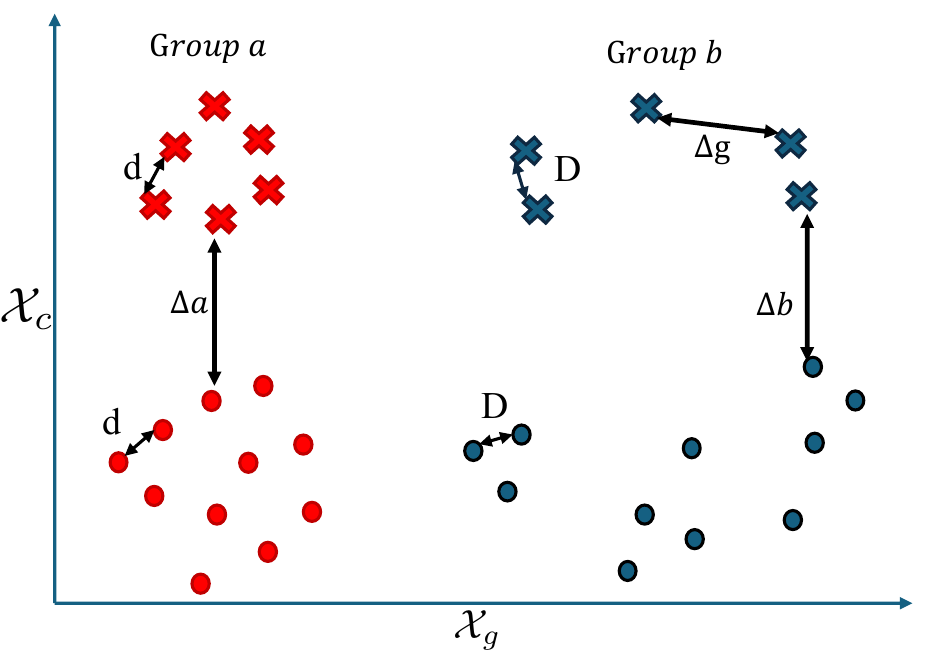}
\caption{For clustered outliers, illustration of group membership obfuscation bias in group $b$ is only in $\mathcal{X}_g$ (x-axis) but not in $\mathcal{X}_c$ (y-axis), for iForest (Proposition 4).}
\label{fig:prop4}
\end{figure}

\textit{In the clustered outliers setting,
let $sp$ denote an iTree's split point, and  $P(x_a < sp < x_b)$ the probability for a split to occur between points $x_a$ and $x_b$. Let groups $a$ and $b$ have equal base rates, $P(Y=1|a)=P(Y=1|b)$.  Let $\{\outliers_a$, $\inliers_a\}$, $\{\outliers_b$, $\inliers_b\}$ represent  the outlier and inlier sets for groups $a$ and $b$. Assume group $b$-inliers form $f$ smaller populations, $\inliers_b^{i}$, for $i \in [\![1,f]\!]$, and $b$-outliers form $g$ smaller populations, $\outliers_b^{j}$ for $j \in [\![1,g]\!]$. Let $\intraa$ and $\intrab$ denote intra-group distance between inlier pairs (same for outlier pairs) for group $a$ and $b$ among both $\mathcal{X}_g$ and $\mathcal{X}_c$ dimensions where $\intrab=\intraa$. 
Also denote by $\dista$ the average distance between outliers and inliers for group $a$ among $\mathcal{X}_c$, by $\distb$ the average distance between outliers and inliers for group $b$ among $\mathcal{X}_c$, and by $\Delta_g$ the average distance among outlier subpopulations and inlier subpopulations of group $b$ among $\mathcal{X}_g$, s.t. $\Delta_g > \intraa$ (See Fig. \ref{fig:prop4}). Then among $\mathcal{X}_g$ features, we have $P(p_b < sp < o_b) > P(p_a < sp < o_a)$, where $o_b\in \outliers_b^i$, $p_b \in \outliers_b^l$, $l \neq i$, and $o_a, p_a \in \outliers_a$. That is, an iTree split on $\mathcal{X}_g$   between group $b$-subpopulations is more likely than between group $a$-outliers. In addition, among $\mathcal{X}_c$,  we have $P(p_b < sp < q_b) = P(p_a < sp < q_a)$, where $p_b \in \outliers_b$, $q_b \in \inliers_b$, $p_a \in \outliers_a$, $q_a \in \inliers_a$. That is, an iTree split on $\mathcal{X}_c$ between inlier and outlier clusters is equally like for both groups.  }

\label{assec:prop4}

\textit{\bf Proof}: When group $b$ exhibits subpopulations due to group-membership obfuscation or naturally, it fragments into inlier and outlier micro-clusters based on varying feature values within $\mathcal{X}_g$. Then among $\mathcal{X}_g$ features, for $o_b\in \outliers_b^i$, $p_b \in \outliers_b^l$, $l \neq i$, and $o_a, p_a \in \outliers_a$, we have
\begin{align}
P(p_b < sp < o_b) &= \frac{\Delta_g}{\max_{x,y \in \inliers \cup \outliers} d(x,y) } \\
&> \frac{\intraa}{\max_{x,y \in \inliers \cup \outliers} d(x,y) } = P(p_a < sp < o_a)
\end{align}

In words, iTree is more likely to split among group $b$-outliers than $a$-outliers, due to larger distances between group $b$'s subpopulations. Therefore, $P(O=1|G=b) > P(O=1 |G=a)$ for iForest. Moreoever, among $\mathcal{X}_g$ dimensions, for $r_b\in \inliers_b^i$, $s_b \in \inliers_b^l$, $l \neq i$, 

\begin{align}
P(r_b < sp < s_b) &= \frac{\Delta_g}{\max_{x,y \in \inliers \cup \outliers} d(x,y) } \\
&= P(p_b < sp < o_b)
\end{align}

That is, iTree is equally likely to split between group $b$'s outlier micro-clusters and inlier micro-clusters among $\mathcal{X}_g$. Therefore, iForest can have $P(O=1 |G=b,Y=1) = P(O=1 |G=b, Y=0)$ and incur an increase in group $b$'s FPR, if  only $\mathcal{X}_g$ dimensions alone were considered. When combined with $\mathcal{X}_c$ dimensions,  for $p_b \in \outliers_b$, $q_b \in \inliers_b$, $p_a \in \outliers_a$, $q_a \in \inliers_a$, we can show that $P(p_b < sp < q_b) = P(p_a < sp < q_a)$ using the same calculation as Eq. \eqref{eq:iforst_equal_outlier_inlier}. Then, because  iTree is equally likely to split between inlier and outlier clusters for both groups $a$ and $b$ among $\mathcal{X}_c$, iForest successfully assigns high outlier scores to points in both $\outliers_b$ and $\outliers_a$. However, as group $b$'s flag rate is increased due to increased FPR, flag rate of $a$ is reduced leading to lower TPR for group $a$. \qedsymbol{}

\subsubsection{\bf Proposition 5}
\label{assec:prop5}
\begin{figure}[!t]
\centering
\includegraphics[width=0.9\linewidth]{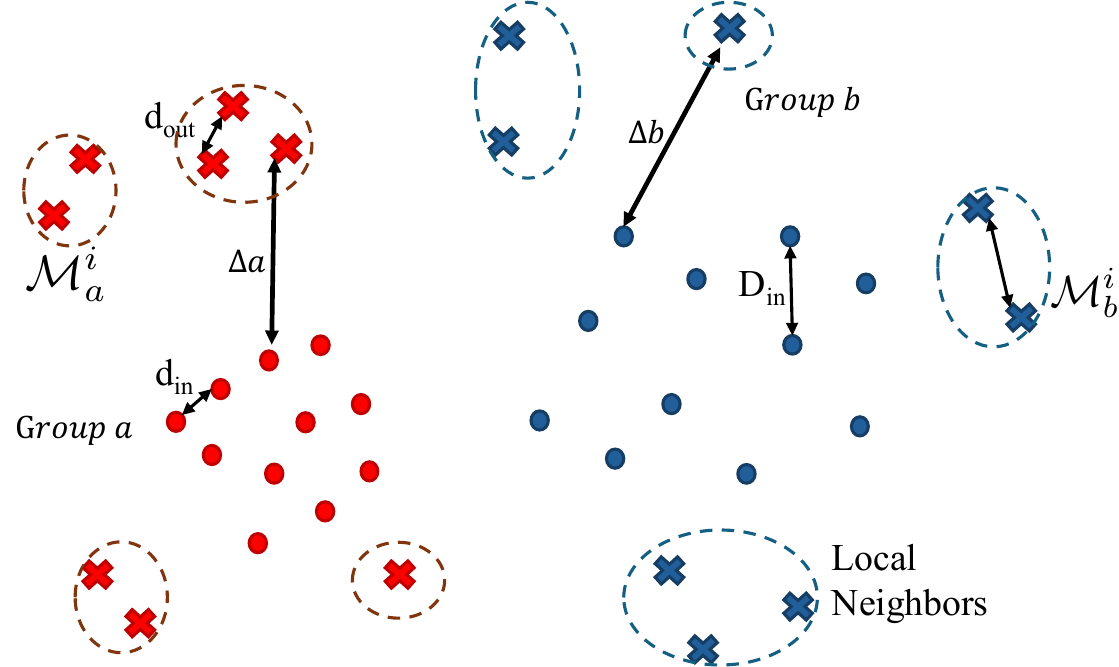}
\caption{For scattered outliers, illustration of density variation between group $a$ and $b$ (Proposition 5).}
\label{fig:prop5}
\end{figure}

\textit{For scattered outliers, let groups $a$ and $b$ have equal size $n_a = n_b$ and equal base rate $P(Y=1|a) = P(Y=1|b)$. Instead of forming a single cluster, scattered outliers form local neighbors (See Fig. \ref{fig:prop5}). We assume that group $a$'s outliers form $g$ local neighborhoods, each denoted by $\mathcal{M}_a^i$, $i \in \{1,...,g\}$, and  group $b$'s outliers form $f$ local neighborhoods, each denoted by $\mathcal{M}_b^j$, $j \in \{1,...,f\}$. In other words, we have $\cup_{1,...,f} \mathcal{M}_b^j = \mathcal{O}_b$ and $\cup_{1,...,g} \mathcal{M}_a^i = \mathcal{O}_a$. Let $\intraa_{in}$ and $\intrab_{in}$ denote the average intra-group distance between inliers for group $a$ and $b$, respectively, where $\intraa_{in} < \intrab_{in}$ (group $b$ is sparser). Similarly, let $\intraa_{out}$ and $\intrab_{out}$ denote the average intra-group distance between outliers of the local neighborhoods for group $a$ and $b$, respectively, where  $\intraa_{out} < \intrab_{out}$. 
We also have $\intraa_{out} > \intraa_{in}$ and $\intrab_{out} > \intrab_{in}$ due to the scattered nature of the outliers. Let $\dista$ and $\distb$ denote the average distance between outliers and inliers in each group, and $m$ the maximum size of the local neighborhoods. Then, assuming LOF hyperparameter $k$ is set s.t. $k > m$ and $\distb = \dista \geq \intrab_{out}$, LOF tends to assign higher scores to group $a$-outliers, increasing flag rate $P(O=1 |a)$. }

\textit{\bf Proof:}
Let $\inliers_a$, $\inliers_b$ denote the sets of inliers in group $a$ and group $b$.  We have that $\dista \geq \intraa_{out}$ and $\distb \geq \intrab_{out}$. Otherwise, when a local neighborhood has intra-neighborhood distance larger than distances between outliers and inliers (i.e. $\intraa_{out} > \dista$, $\intrab_{out} > \distb$ ), the local neighborhood can instead form several smaller neighborhoods. Based on Lemma 1 and 2, LOF could flag micro-clusters of local neighborhood outliers when $k > m$, the size of the largest local neighborhood for scattered outliers in both groups.

 For an outlier point in any of group $a$'s local neighborhoods, $p_a^i \in \mathcal{M}_a^i$ for $i \in \{1,...,g\}$, the expected LOF is
\begin{align}
\mathbb{E}[
\text{LOF}_k(p_{a}^i)] &= \mathbb{E} \Bigg [  \frac{\sum_{o \in N_k(p_{a}^i)} \text{LRD}_k(o)}{ |N_k(p_a^i)| \cdot \text{LRD}_k(p_{a}^i)} \Bigg] \\  &=  \frac{\sum_{o \in N_k(p_{a}^i)} \mathbb{E} \big [ \text{LRD}_k(o) \big] }{ k \cdot  \mathbb{E} \big [\text{LRD}_k(p_{a}^i)\big]} \\ 
&= \frac{\frac{|\mathcal{M}_a^i|}{\dista} + \frac{k - |\mathcal{M}_a^i|}{\intraa_{in}}}{ \frac{k}{\dista}} \label{eq:scatterd_variance_a}
\end{align}

In Eq. \eqref{eq:scatterd_variance_a}, at least $k-|\mathcal{M}_a^i|$ neighbors of $p_a^i$ contain only inliers and therefore the average reachability distance is $\intraa_{in}$. Assume that there exists a $p_b^j \in \mathcal{M}_b^j$, where $|\mathcal{M}_b^j| = |\mathcal{M}_a^i|$, Then, we have $\mathbb{E}[\text{LOF}_k(p_{a}^i)] > \mathbb{E}[\text{LOF}_k(p_{b}^i)] $ due to $\intraa_{in} < \intrab_{in}$. That is, if local neighborhoods are of similar sizes, LOF assigns higher score to group-$a$ outliers, increasing flag rate $P(O=1|G=a)$.

 When $\intrab_{out} \approx \distb$, outliers are equally distanced from one another and from the inliers, and thus local neighborhoods disappear. However for $p_b \in \mathcal{O}_b$, as long as  $\alpha>0$ number of group $b$-inliers exist among the $k$-neighbors of $p_b$, the expected LOF of $p_b$ is
 \begin{align}
\mathbb{E}[
\text{LOF}_k(p_{b})] &= \frac{\frac{k-\alpha}{\dista} + \frac{\alpha}{\intrab_{in}}}{ \frac{k}{\dista}}
\end{align}
\noindent
which prevents LOF  to assign an outlier score of $1$ for $p_b$. The adjacency to inliers prevents LOF from incurring very low TPR on group $b$, unlike the clustered outlier setting in Proposition 1. \qedsymbol{}

\subsection{Limitations of Analysis}
\label{asec:limitations_additional}

While our theoretical analyses provide explanations for various OD outcome discrepancies between protected groups $a$ and $b$, we acknowledge they exhibit several limitations.

First, while some mechanistic/non-learning algorithms (LOF and iForest in our study) can be explained by analyzing their working mechanisms, it is hard to directly analyze the behavior of  other algorithms  in reaction to different data biases. Specifically, the learning-based algorithms DeepAE and FairOD that rely on a multi-layer autoencoder structure are hard to analyze mathematically due to randomness in initialization and training. %

Second, while we explicitly and clearly stated, our theoretical analyses are based on certain assumptions.  Specifically, we assumed that each group or subpopulation of outliers and inliers follow the same average pairwise intra-group distances, which is smaller than the distances between inliers and outliers. The uniformity of data distributions generally does not hold true, especially for real-world datasets. We also assumed the groups to be well separated in the proxy feature space so that inter-group distances do not interfere with the behavior of LOF and iForest on each group. While it may be a weaker assumption that groups are not too alike, the extent of inter-group similarity could affect our propositions in practice.

\end{document}